\documentclass[final]{cvpr}

\usepackage{booktabs}
\usepackage{multirow}
\usepackage{multicol}
\usepackage{times}
\usepackage{epsfig}
\usepackage{graphicx}
\usepackage{amsmath}
\usepackage{amssymb}
\usepackage[inline]{enumitem}
\usepackage{bm}
\usepackage{rotating}
\usepackage{gensymb}
\usepackage{nicefrac}
\usepackage[dvipsnames]{xcolor}
\usepackage{calc}
\usepackage{microtype}  
\usepackage{cuted}  
\usepackage{cite}
\usepackage[bottom]{footmisc}
\usepackage{xr}  

\usepackage[pagebackref=true,breaklinks=true,colorlinks,bookmarks=false]{hyperref}
\hypersetup{
    citecolor=ForestGreen,
    urlcolor=MidnightBlue,
}

\def\cvprPaperID{2989} 
\def\confYear{CVPR 2021}

\newif\ifproceedings  
\proceedingsfalse
\newif\ifsupponly
\supponlyfalse

\ifproceedings
    \newcommand{\supp}{Supplemental} 
\else
    \newcommand{\supp}{Appendix} 
\fi

\newcommand{\PAR}[1]{\vskip4pt \noindent{\bf #1~}}
\renewcommand{\*}[1]{\bm{\mathrm{#1}}}
\renewcommand{\b}[1]{\textbf{#1}}
\newcommand{\red}[1]{\textcolor{red}{#1}}
\newcommand{\green}[1]{\textcolor{green}{#1}}
\newcommand{\gray}[1]{\textcolor{gray}{#1}}
\newcommand{\blue}[1]{\textcolor{blue}{#1}}
\newcommand{\0}{\phantom{0}}

\makeatletter
\newcommand{\printfnsymbol}[1]{%
  \textsuperscript{\@fnsymbol{#1}}%
}
\makeatother

\begin{document}

\title{Back to the Feature: Learning Robust Camera Localization from Pixels to Pose}

\author{Paul-Edouard Sarlin$^{1}$\thanks{denotes equal contribution} \hspace{.03in}
Ajaykumar Unagar$^{2}$\printfnsymbol{1}\hspace{.03in}
M{\aa}ns Larsson$^{3,4}$\hspace{.03in}
Hugo Germain$^{5}$ \hspace{.03in}
Carl Toft$^{3}$\\
Viktor Larsson$^{1}$\hspace{.01in}
Marc Pollefeys$^{1,6}$\hspace{.01in}
Vincent Lepetit$^{5}$\hspace{.01in}
Lars Hammarstrand$^{3}$\hspace{.01in}
Fredrik Kahl$^{3}$\hspace{.01in}
Torsten Sattler$^{3,7}$
\vspace{0.05in}\\
$^{1}$ Department of Computer Science, ETH Zurich\hspace{0.07in}
$^{2}$ ETH Zurich\hspace{0.07in}
$^{3}$ Chalmers University of Technology\\
$^{4}$ Eigenvision\hspace{0.07in}
$^{5}$ Ecole des Ponts\hspace{0.07in}
$^{6}$ Microsoft\hspace{0.07in}
$^{7}$ Czech Technical University in Prague
}

\ifsupponly\else 
\maketitle
\ifproceedings\iftoggle{cvprfinal}{\pagestyle{empty}}\fi  
\ifproceedings\iftoggle{cvprfinal}{\thispagestyle{empty}}\fi  

\begin{abstract}
Camera pose estimation in known scenes is a 3D geometry task recently tackled by multiple learning algorithms.
Many regress precise geometric quantities, like poses or 3D points, from an input image.
This either fails to generalize to new viewpoints or ties the model parameters to a specific scene.
In this paper, we go Back to the Feature: we argue that deep networks should focus on learning robust and invariant visual features, while the geometric estimation should be left to principled algorithms.
We introduce PixLoc, a scene-agnostic neural network that estimates an accurate 6-DoF pose from an image and a 3D model.
Our approach is based on the direct alignment of multiscale deep features, casting camera localization as metric learning. 
PixLoc learns strong data priors by end-to-end training from pixels to pose and exhibits exceptional generalization to new scenes by separating model parameters and scene geometry.
The system can localize in large environments given coarse pose priors but also improve the accuracy of sparse feature matching by jointly refining keypoints and poses with little overhead.
The~code will be publicly available at \small{\href{https://github.com/cvg/pixloc}{\texttt{github.com/cvg/pixloc}}}.
\end{abstract}
\vspace{-3mm}

\section{Introduction}

\begin{figure}[t]
    \centering
    \includegraphics[width=\linewidth]{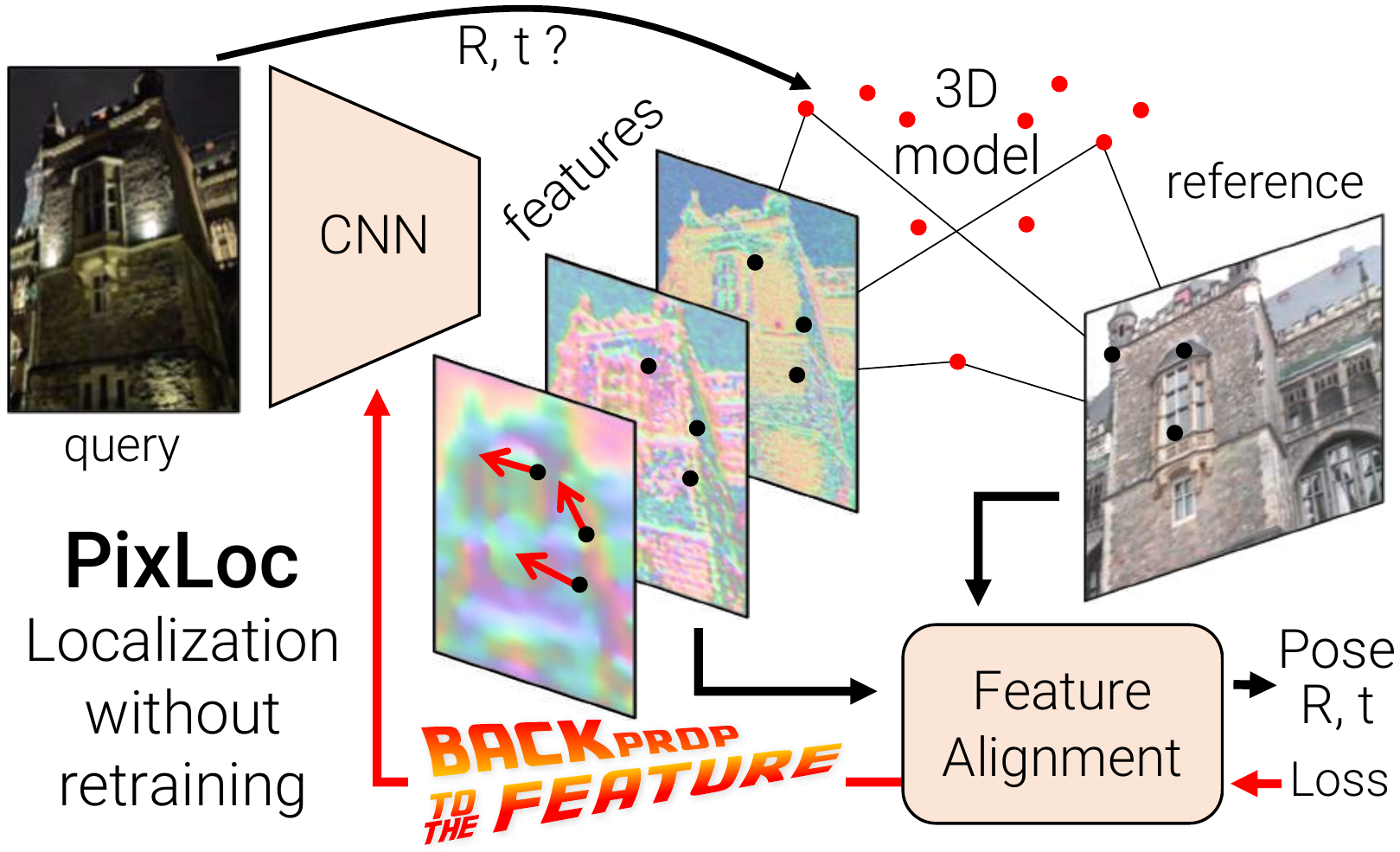}%
    \caption{\textbf{Learning scene-agnostic localization.} Deep neural networks should not have to rediscover well-understood geometric principles. We only need to learn good features: PixLoc is trained end-to-end to estimate the pose of an image by aligning deep features with a reference 3D model via a differentiable optimization.
    }
    \label{fig:teaser}%
\end{figure}

Visual localization is the problem of estimating the camera position and orientation for a given image in a known scene.
Solving this problem is a key step towards truly autonomous robots such as self-driving cars and is a pre-requisite for Augmented and Virtual Reality systems.

State-of-the-art approaches to visual localization commonly rely on correspondences between 2D pixel positions and 3D points in the scene~\cite{sarlin2019coarse,sattler2016efficient,brachmann2020dsacstar,cavallari2019let,taira2018inloc,Toft2020TPAMI,sarlin2020superglue,svarm2017city,Lynen2020IJRR,germain2020s2dnet}.
Such a formulation estimates the camera pose using a Perspective-n-Point (PnP) solver~\cite{Kukelova13ICCV,Albl2016CVPR,Kneip2011CVPR,Bujnak08CVPR,Haralick94IJCV} inside a RANSAC  loop~\cite{fischler1981random,Lebeda2012BMVC,barath2019magsac,Chum08PAMI}. 
These 2D-3D correspondences are traditionally computed by matching local image features.
Recent localization systems can handle large scenes with complex geometry and appearance changes over time.
They~leverage deep neural networks that learn to extract such features~\cite{dusmanu2019d2,revaud2019r2d2,superpoint,bhowmik2020reinforced,schonberger2018semantic,pautrat2020online,yang2020ur2kid}, to match them~\cite{sarlin2020superglue,pautrat2020online}, and to filter outlier correspondences~\cite{sarlin2020superglue,MLarsson2019ICCV,Toft2018ECCV,brachmann2019neural,moo2018learning}.

Training a feature matching pipeline in an end-to-end manner is challenging and unstable as its complexity hinders gradients propagation~\cite{bhowmik2020reinforced}. 
An alternative is to train a convolutional neural network (CNN) to regress geometric quantities such as camera poses~\cite{kendall2015posenet,kendall2017geometric,walch2017image,balntas2018relocnet,laskar2017camera,ding2019camnet,Zhou2020ICRA} or the 3D scene coordinate corresponding to each pixel~\cite{shotton2013scene,cavallari2017fly,cavallari2019let,brachmann2017dsac,brachmann2018learning,brachmann2019expert,brachmann2020dsacstar,li2020hierarchical,yang2019sanet}.
While these approaches can be trained end-to-end, they come with their own drawbacks. Absolute pose and coordinate regression are scene-specific and require to be trained for or adapted to new scenes~\cite{cavallari2017fly, cavallari2019let}.
Generalization to new viewing conditions, \eg, localizing night-time images when training only on daytime photos, and handling larger, more complex scenes~\cite{taira2018inloc,schonberger2018semantic} are open challenges for such approaches. 
Additionally, absolute or relative pose regression has limited accuracy and often fails to generalize to new viewpoints~\cite{sattler2019understanding,Zhou2020ICRA}.
While regressing poses relative to a set of reference images~\cite{balntas2018relocnet,laskar2017camera,ding2019camnet,Zhou2020ICRA} is in theory scene-agnostic,
generalization to strongly differing scenes without a significant drop in pose accuracy~\cite{sattler2019understanding,Zhou2020ICRA} has, to the best of our knowledge, not been shown so far.

What hinders the generalization of existing end-to-end regression methods is that they predict camera poses or 3D geometry solely from image information.
In practice, such quantities are often readily available.
Pose priors can be obtained via image retrieval or sensors such as GPS. 
At the same time, the 3D scene geometry is often provided as a by-product of the 3D reconstruction systems that generate the training poses, \eg with Structure-from-Motion or SLAM.

\begin{figure}[t]
    \def\iwidth{0.49}
    \begin{minipage}{\iwidth\linewidth}
        \centering
        \footnotesize{Reference image}%
    \end{minipage}%
    \hspace{0.05mm}
    \begin{minipage}{\iwidth\linewidth}
        \centering
        \footnotesize{Aligned query}
    \end{minipage}
    
    \begin{minipage}{\iwidth\linewidth}
        \includegraphics[width=\linewidth]{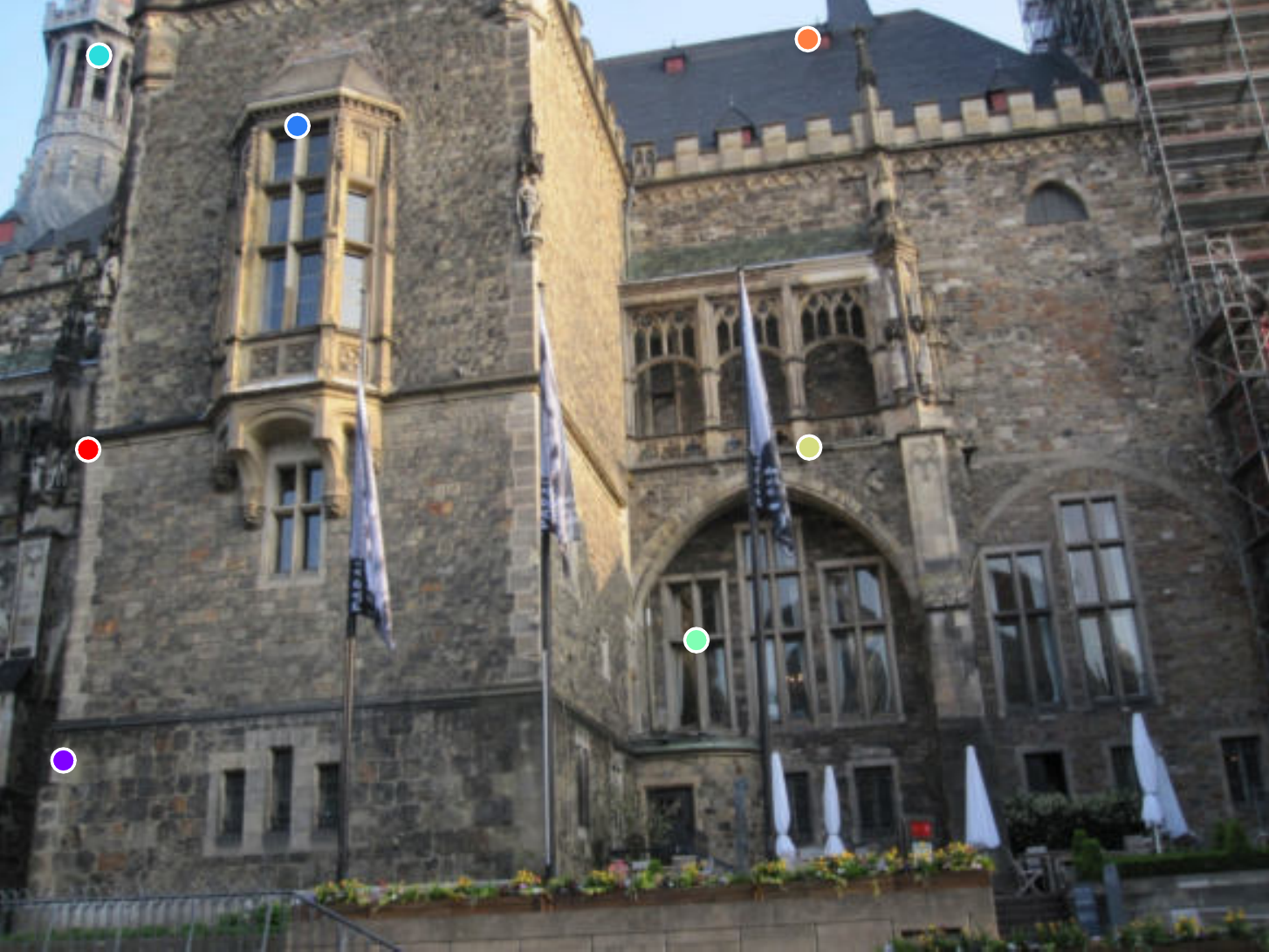}%
    \end{minipage}%
    \hspace{0.05mm}
    \begin{minipage}{\iwidth\linewidth}
        \includegraphics[width=\linewidth]{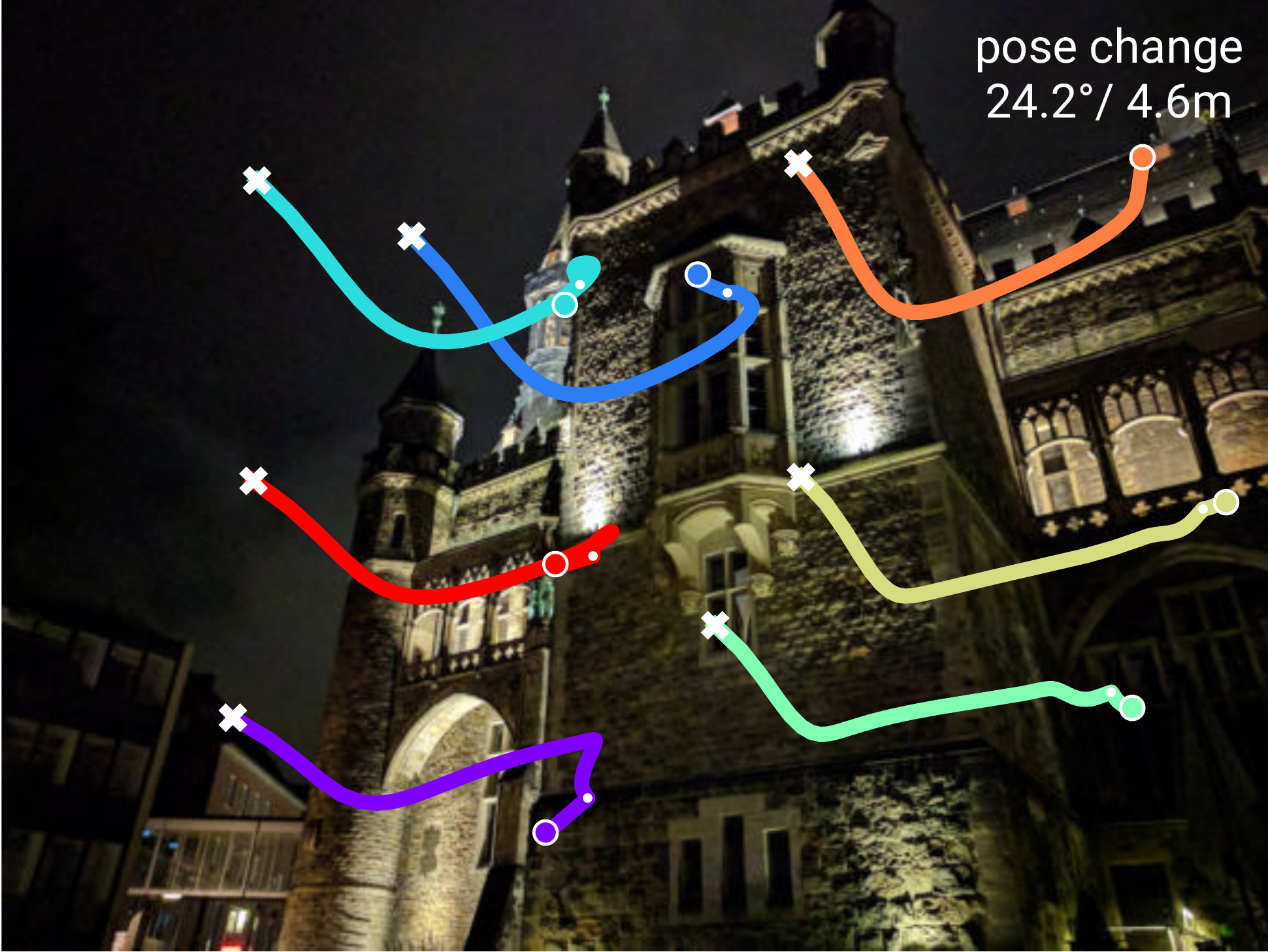}%
    \end{minipage}
    
    \begin{minipage}{0.425\linewidth}
        \includegraphics[width=\linewidth, trim=0 0 0 80, clip]{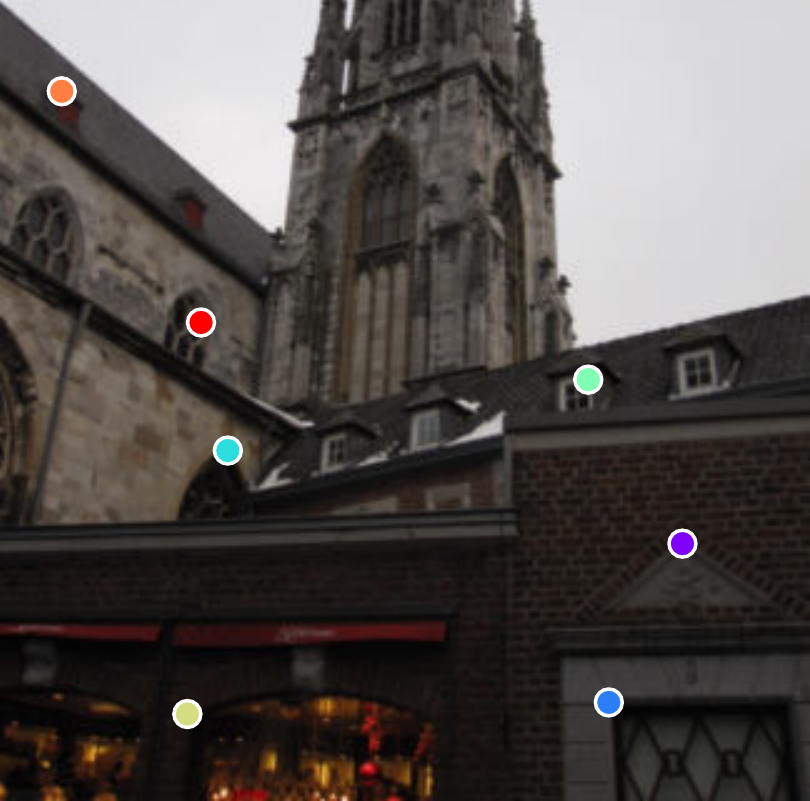}%
    \end{minipage}%
    \hspace{0.05mm}
    \begin{minipage}{0.555\linewidth}
        \includegraphics[width=\linewidth]{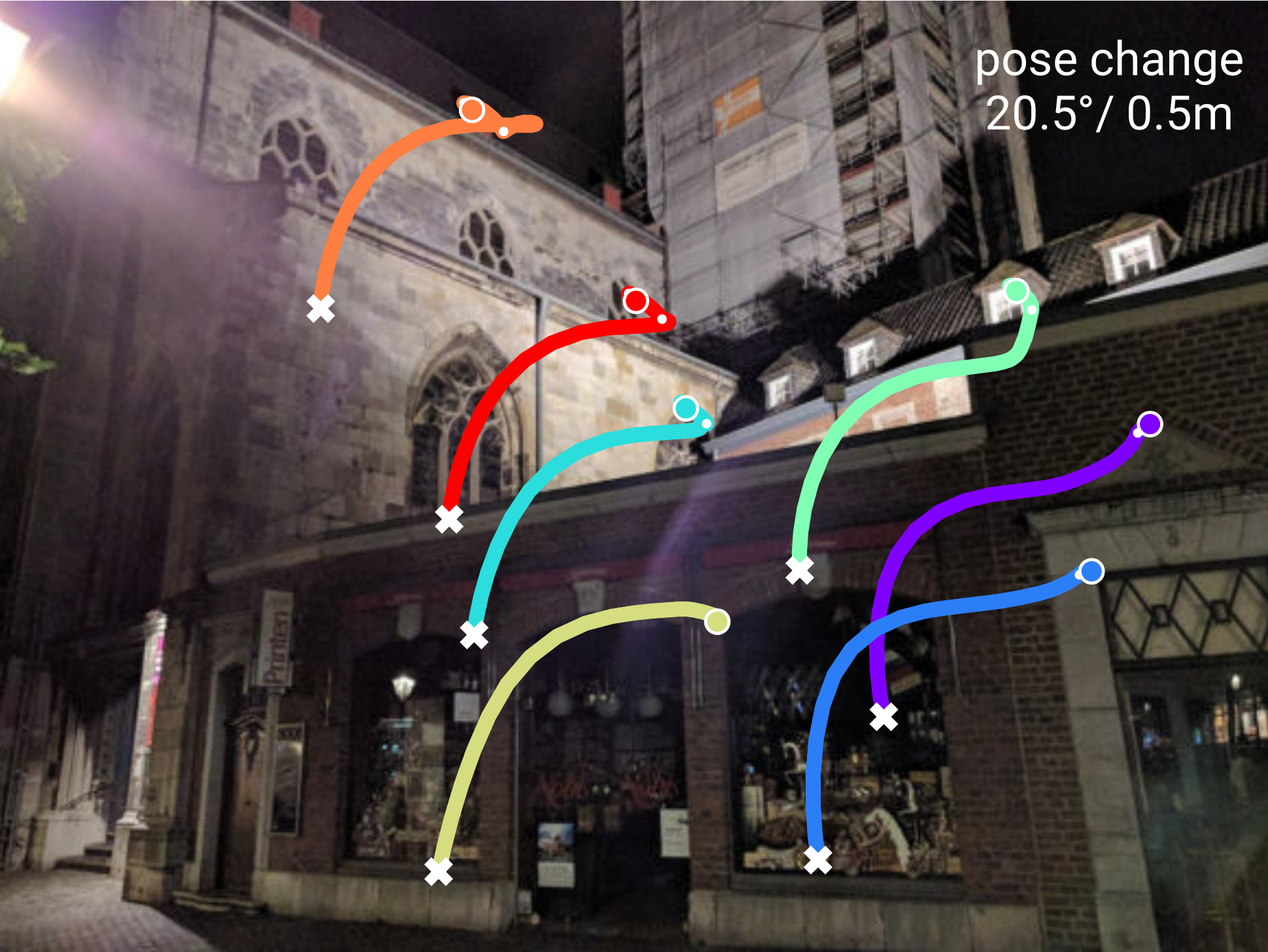}%
    \end{minipage}
    \vspace{1mm}
    \caption{\textbf{Alignment for localization.}
    Although only based on local gradients, direct alignment works well thanks to deep features, despite the coarse initial pose estimate and strong appearance changes. Here points travel from crosses to colored dots.
    }
    \label{fig:tracks}%
\end{figure}

Inspired by direct image alignment~\cite{engel14eccv,engel2017direct,park2017icra,Czarnowski_2017_ICCV,von2020gn,von2020lm} and learned image representations for outlier rejection~\cite{MLarsson2019ICCV}, 
we argue that end-to-end visual localization algorithms should focus on representation learning. 
Rather than devoting model capacity and data to learn basic geometric relations or encode 3D maps, they should rely on well-understood geometric principles and instead learn robustness to appearance and structural changes. 

In this paper, we introduce a trainable algorithm, PixLoc, that localizes an image by aligning it to an explicit 3D model of the scene based on dense features extracted by a CNN (Figure~\ref{fig:teaser}).
By relying on classical geometric optimization, the network does not need to learn pose regression itself, but only to extract suitable features, making the algorithm accurate and scene-agnostic.
We train PixLoc end-to-end, from pixels to pose, by unrolling the direct alignment and supervising only the pose.
Given an initial pose obtained by image retrieval, our formulation results in a simple localization pipeline competitive with complex state-of-the-art approaches, even when the latter are trained specifically per scene. 
PixLoc can also refine poses estimated by any existing approach as a lightweight post-processing step.
Through detailed experiments, we show that our method generalizes well to new scenes, \eg, from outdoor to indoor scenes, and challenging viewing conditions. 
To the best of our knowledge, PixLoc is the first end-to-end visual localization approach to exhibit such exceptional generalization. 

\begin{figure*}[!t]
    \centering
    \includegraphics[width=0.98\linewidth, clip]{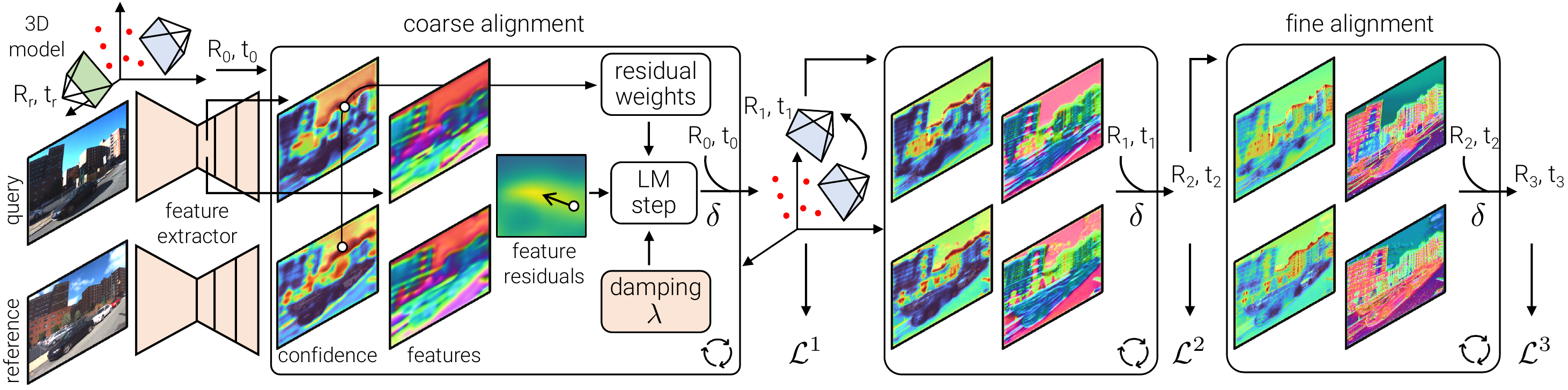}%
    \caption{\textbf{Pose estimation with PixLoc.}
    Given a sparse 3D model and a coarse initial pose $\left(\*R_0,\*t_0\right)$, PixLoc extracts multilevel features with pixelwise confidences for query and reference images.
    The Levenberg-Marquardt optimization then aligns corresponding features according to the 3D points, guided by the confidence, from the coarse to the fine level. We only supervise the pose predicted at each level.
    }
\label{fig:detail}%
\end{figure*}

\section{Related work}
\PAR{Accurate visual localization} commonly relies on estimating correspondences between 2D pixel positions and 3D scene coordinates.
Such approaches detect, describe~\cite{lowe2004distinctive, bay2006surf}, and match~\cite{irschara2009structure, li2012worldwide, sattler2016efficient, svarm2017city, Zeisl_2015_ICCV, liu2017efficient} local features, maintain an explicit sparse 3D representation of the environment, and sometimes leverage image retrieval~\cite{torii201524, vlad} to scale to large scenes~\cite{middelberg, taira2018inloc, torii2019large, sarlin2019coarse, sattler2012image, irschara2009structure}.
Recently, many of these components have been learned with great success~\cite{yi2016lift, superpoint, dusmanu2019d2, revaud2019r2d2, arandjelovic2016netvlad, delf, radenovic2018fine, moo2018learning, sarlin2020superglue}, but often independently and not end-to-end due to the complexity of such systems. Here we introduce a simpler alternative to feature matching, finally enabling stable end-to-end training.
Our solution can learn more powerful priors than individual blocks, yet remains highly flexible and interpretable. 

\PAR{End-to-end learning for localization} has recently received much attention.
Common approaches encode the scene into a deep network by regressing from an input image to an absolute pose~\cite{kendall2015posenet, kendall2017geometric, walch2017image, radwan2018vlocnet++, naseer2017deep} or 3D scene coordinates~\cite{shotton2013scene, brachmann2017dsac, brachmann2020dsacstar, cavallari2017fly, cavallari2019let}.
Pose regression lacks geometric constraints and thus does not generalize well to novel viewpoints or appearances~\cite{sattler2019understanding, schonberger2018semantic}, while coordinate regression is more robust.
Both do not scale well due to the limited network capacity~\cite{taira2018inloc, brachmann2019expert} and require for each new scene either costly retraining or adaptation~\cite{cavallari2017fly,cavallari2019let}.
ESAC~\cite{brachmann2019expert} improves the scalability by training an ensemble of regressors, each specialized in a scene subset, but is still significantly less accurate than feature-based methods in larger environments.

Differently, some approaches regress a camera pose relative to one or more training images~\cite{balntas2018relocnet, laskar2017camera, ding2019camnet, Zhou2020ICRA}, often after an explicit retrieval step.
They do no memorize the scene geometry and are thus scene-agnostic, but, similar to absolute regressors, are less accurate than feature-based methods~\cite{Zhou2020ICRA,sattler2019understanding}.
Closer to ours, SANet~\cite{yang2019sanet} takes the scene representation out of the network by regressing 3D coordinates from an input 3D point cloud.
Critically, all top-performing learnable approaches are at least trained per-dataset, if not per-scene, and are limited to small environments~\cite{kendall2015posenet, shotton2013scene}.
In this work we demonstrate the first end-to-end learnable network that generalizes across scenes, including from outdoor to indoor, and that delivers performance competitive with complex pipelines on large real-world datasets, thanks to a differentiable pose solver.

\PAR{Learning camera pose optimization} can be tackled by unrolling the optimizer for a fixed number of steps~\cite{clark2018ls, wang2018deep, tang2018ba, lv2019taking, ma2020deep, xu2020deep}, computing implicit derivatives~\cite{jorgensen2019, russell2019fixing, campbell2020solving, chen2020bpnp, brachmann2020dsacstar}, or crafting losses to mimic optimization steps~\cite{von2020gn, von2020lm}.
Multiple works have proposed to learn components of these optimizers~\cite{clark2018ls, lv2019taking, tang2018ba}, with added complexity and unclear generalization.
Some of these formulations optimize reprojection errors over sparse points, while others use direct objectives for (semi\nobreakdash-)dense image alignment.
The latter are attractive for their simplicity and accuracy, but usually do not scale well.
Like their classical counterparts~\cite{engel2017direct, kerl2013dense}, they also suffer from a small basin of convergence, limiting them to frame tracking.
In contrast, PixLoc is explicitly trained for wide-baseline cross-condition camera pose estimation from sparse measurements (Figure~\ref{fig:tracks}).
By focusing on learning good features, it shows good generalization yet learns sensible data priors that shape the optimization objective.

\section{PixLoc: from pixels to pose}
\PAR{Overview:} PixLoc localizes by aligning query and reference images according to the known 3D structure of the scene.
The alignment consists of a few steps that minimize an error over deep features predicted from the input images by a CNN (Figure~\ref{fig:detail}).
The CNN and the optimization parameters are trained end-to-end from ground truth poses.

\PAR{Motivation:}
In absolute pose and scene coordinate regression from a single image, a deep neural network learns to
\begin{enumerate*}[label=\roman*)]
\item\label{item:regress:1} recognize the approximate location in a scene, 
\item\label{item:regress:2} recognize robust visual features tailored to this scene, and 
\item\label{item:regress:3} regress accurate geometric quantities like pose or coordinates.
\end{enumerate*}
Since CNNs can learn features that generalize well across appearances and geometries, i) and ii) do not need to be tied to a specific scene, and i) is already solved by image retrieval. 
On the other hand, iii) is tackled by classical geometry using feature matching~\cite{fischler1981random, Chum08PAMI, chum2003locally} or image alignment~\cite{LK, LK20years, engel2017direct, engel14eccv} and a 3D representation.
We should thus focus on learning robust and generic features, making the pose estimation scene-agnostic and tightly constrained by geometry.
The challenge lies in how to define good features to localize. We solve this by making the geometric estimation differentiable and supervise only the final pose estimate.
Differently from pose or coordinate regression, we assume that a 3D scene representation is available. This requirement is easily met in practice since the reference poses are usually obtained by sparse or dense 3D reconstruction.

\PAR{Problem formulation:} Our goal is to estimate the 6-DoF pose $(\*R, \*t) \in \mathbf{SE}(3)$ of a query image $\*I_q$, where $\*R$ is a rotation matrix and $\*t$ is a translation vector in the camera frame. We are given a 3D representation of the environment, such as a sparse or dense 3D point cloud $\{\*P_i\}$ and posed reference images $\{\*I_k\}$, collectively called the reference data.

\subsection{Localization as image alignment}

\PAR{Image Representation:} The sparse alignment is performed over learned feature representations of the images. We leverage CNNs and their ability to extract a hierarchy of features at multiple levels. For each query image $\*I_q$ and reference image $\*I_k$, a CNN extracts a $D_l$-dimensional feature map $\*F^l \in \mathbb{R}^{W_l\times H_l \times D_l}$ at each level $l \in \{L, ..., 1\}$. Those have decreasing resolution and progressively encode richer semantic information and a larger spatial context of the image. The features are $L_2$-normalized along the channels to improve their robustness and generalization across datasets.

This learned representation, inspired by past works on handcrafted and learned features for camera tracking~\cite{park2017icra, Czarnowski_2017_ICCV, wang2018deep, tang2018ba, lv2019taking, von2020gn}, is robust to large illumination or viewpoint changes and provides meaningful gradients for successful alignments despite poor initial pose estimates. In contrast, classical direct alignment~\cite{LK, LK20years, engel2017direct, engel14eccv} operates on the original image intensity, which is not robust to long-term changes encountered in common localization scenarios, and resorts to Gaussian image pyramids, which still largely limits the convergence to frame-to-frame tracking.

\PAR{Direct alignment:} The goal of the geometric optimization is to find the pose $(\*R, \*t)$ which minimizes the difference in appearance between the query image and each reference image. For a given feature level $l$ and each 3D point~$i$ observed in each reference image~$k$, we define a residual:
\begin{equation}
    \*r_k^i = \*F_q^l\left[\*p_q^i\right] - \*F_k^l\left[\*p_k^i\right] \in \mathbb{R}^D \enspace,
\end{equation}
where $\*p_q^i = \Pi\left(\*R\*P_i+\*t\right)$ is the projection of $i$ in the query given its current pose estimate and $\left[\cdot\right]$ is a lookup with sub-pixel interpolation. The total error over $N$ observations is
\begin{equation}
    \label{eq:cost}
    E_l(\*R, \*t) = \sum_{i,k} w_k^i\ \rho\left(\left\Vert\*r_k^i\right\Vert_2^2\right) \enspace,
\end{equation}
where $\rho$ is a robust cost function~\cite{hampel1986robust} with derivative $\rho'$ and $w_k^i$ is a per-residual weight. This nonlinear least-squares cost is iteratively minimized from an initial estimate $(\*R_0, \*t_0)$ using the Levenberg-Marquardt~(LM)  algorithm~\cite{levenberg1944method, marquardt1963algorithm}.

To maximize the convergence basin, we optimize each feature level successively, starting with the coarsest level $l{=}1$, 
and initialize each with the result of the previous level. Low-resolution feature maps are thus responsible for the robustness of the pose prediction while finer features 
enhance its accuracy.
Each pose update $\*\delta \in \mathbb{R}^6$ is parametrized on the $\*{SE}(3)$ manifold using its Lie algebra. We stack all residuals into $\*r\in\mathbb{R}^{ND}$ and all weights into $\*W = \mathrm{diag}_{i,k}\left(w_k^i\,\rho'\right)$ and write the Jacobian and Hessian matrices as
\begin{equation}
    \*J_{i,k} = \frac{\partial \*r_k^i}{\partial \*\delta}
    = \frac{\partial\*F_q}{\partial\*p_q^i}\frac{\partial\*p_q^i}{\partial\*\delta}
    \quad \text{and} \quad \*H = \*J^\top\*W\*J \enspace.
\end{equation}
The update is computed by damping the Hessian and solving the linear system:
\begin{equation}
    \label{eq:lm_update}
    \*\delta = -\left(\*H + \lambda\,\mathrm{diag}\left(\*H\right)\right)^{-1}\*J^\top\*W\*r \enspace,
\end{equation}
where $\lambda$, the damping factor, interpolates between the Gauss-Newton~($\lambda{=}0$) and gradient descent~($\lambda\!\!\to\!\!\infty$) formulations and is usually adjusted at each iteration using diverse heuristics~\cite{levenberg1944method, marquardt1963algorithm, madsen2004methods}. Finally, the new pose is computed by left-multiplication on the manifold as
\begin{equation}
    \begin{bmatrix}\*R^+& \*t^+\end{bmatrix}
    = \exp{\left(\*\delta^\wedge\right)}^\top
    \begin{bmatrix}\*R & \*t \\ \*0 & 1\end{bmatrix} \enspace,
\end{equation}
where $\cdot^\wedge$ is the skew operator. The optimization stops when the update $\*\delta$ is small enough.

\PAR{Infusing visual priors:} The steps described above are identical to the classical photometric alignment~\cite{LK, LK20years, engel2017direct}. The CNN is however capable of learning complex visual priors -- we therefore would like to give it the ability to steer the optimization towards the correct pose. To this end, the CNN predicts an uncertainty map $\*U_k^l \in \mathbb{R}_{>0}^{W_l\times H_l}$ along with each feature map. The pointwise uncertainties of the query and reference images are combined into a per-residual weight as
\begin{equation}
    w_k^i = u_q^i\,u_k^i = \frac{1}{1 + \*U_q^l\left[\*p^i_q\right]}\,\frac{1}{1 + \*U_k^l\left[\*p^i_k\right]} \in [0, 1] \enspace.
\end{equation}
The weight is $1$ if the 3D point projects into a location with low uncertainty in both the query and the reference images. It tends to $0$ as either of the location is uncertain.
Here $w_k^i$ is not explicitly supervised, but rather learned as to maximize the pose accuracy.
A similar formulation was applied to direct RGB-D frame tracking in a concurrent work~\cite{xu2020deep}.

This weighting can capture multiple scenarios.
First, the network can learn to be uncertain when it cannot predict invariant features, \eg, because of domain shift, similarly to an aleatoric uncertainty~\cite{kendall2017what}.
The uncertainty can also be high for locations that can be well described by the CNN, but which consistently push the optimization away from the correct pose by introducing local minima in the cost landscape.
This encompasses dynamic objects or repeated patterns and symmetries, as shown in Figures~\ref{fig:uncertainties} and~\ref{fig:examples}. The uncertainty is different for each level, as different cues might be useful at different stages of the optimization.

\begin{figure}[t]
    \centering
    \begin{minipage}{0.9\linewidth}
    \centering
    \def\iwidth{0.49}
    \begin{minipage}{\iwidth\linewidth}
        \centering
        \footnotesize{Input Image}%
    \end{minipage}%
    \hspace{0.05mm}
    \begin{minipage}{\iwidth\linewidth}
        \centering
        \footnotesize{Learned Weighting $u_q$}
    \end{minipage}
    
    \begin{minipage}{\iwidth\linewidth}
        \includegraphics[width=\linewidth]{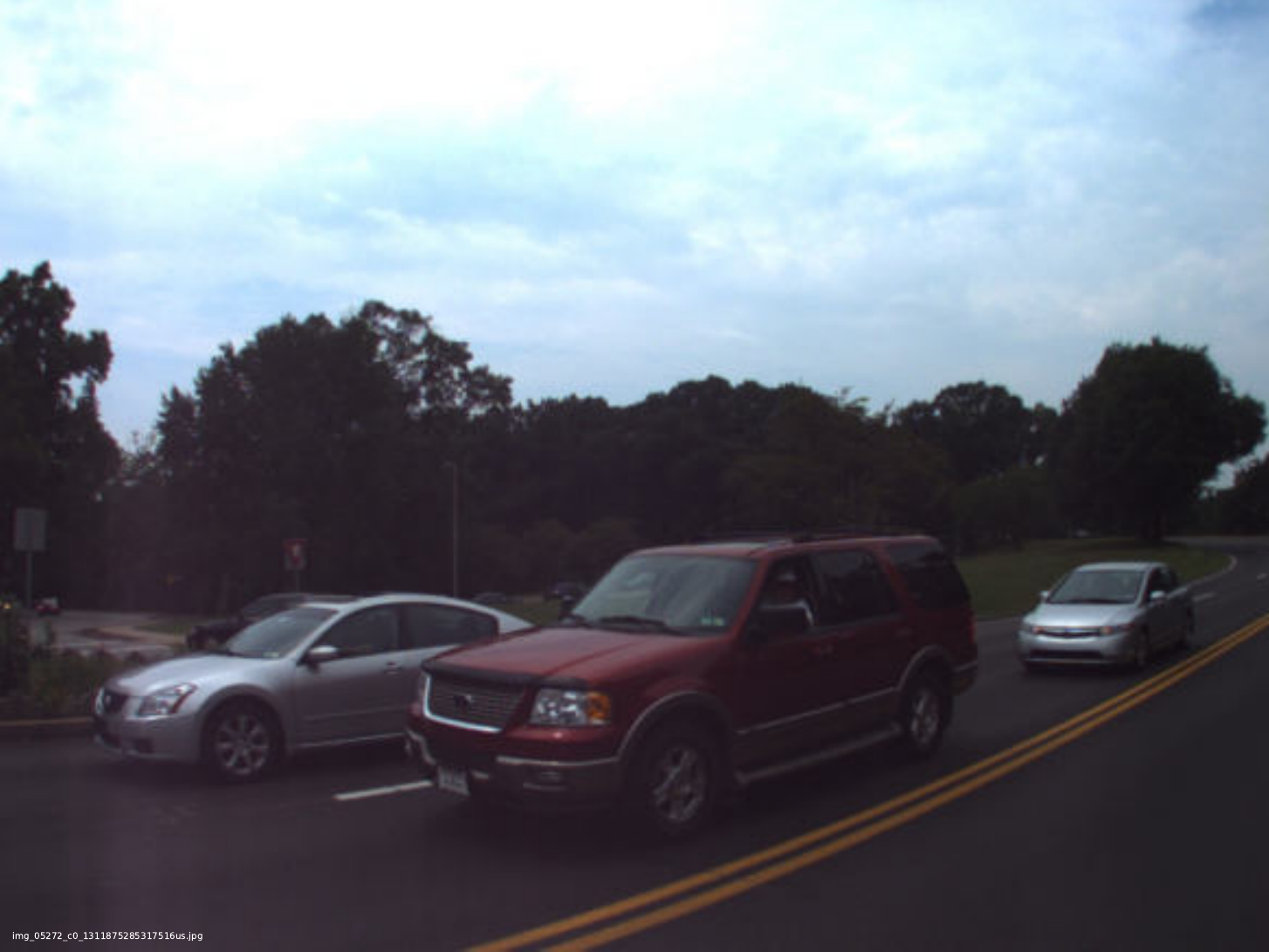}%
    \end{minipage}%
    \hspace{0.05mm}
    \begin{minipage}{\iwidth\linewidth}
        \includegraphics[width=\linewidth]{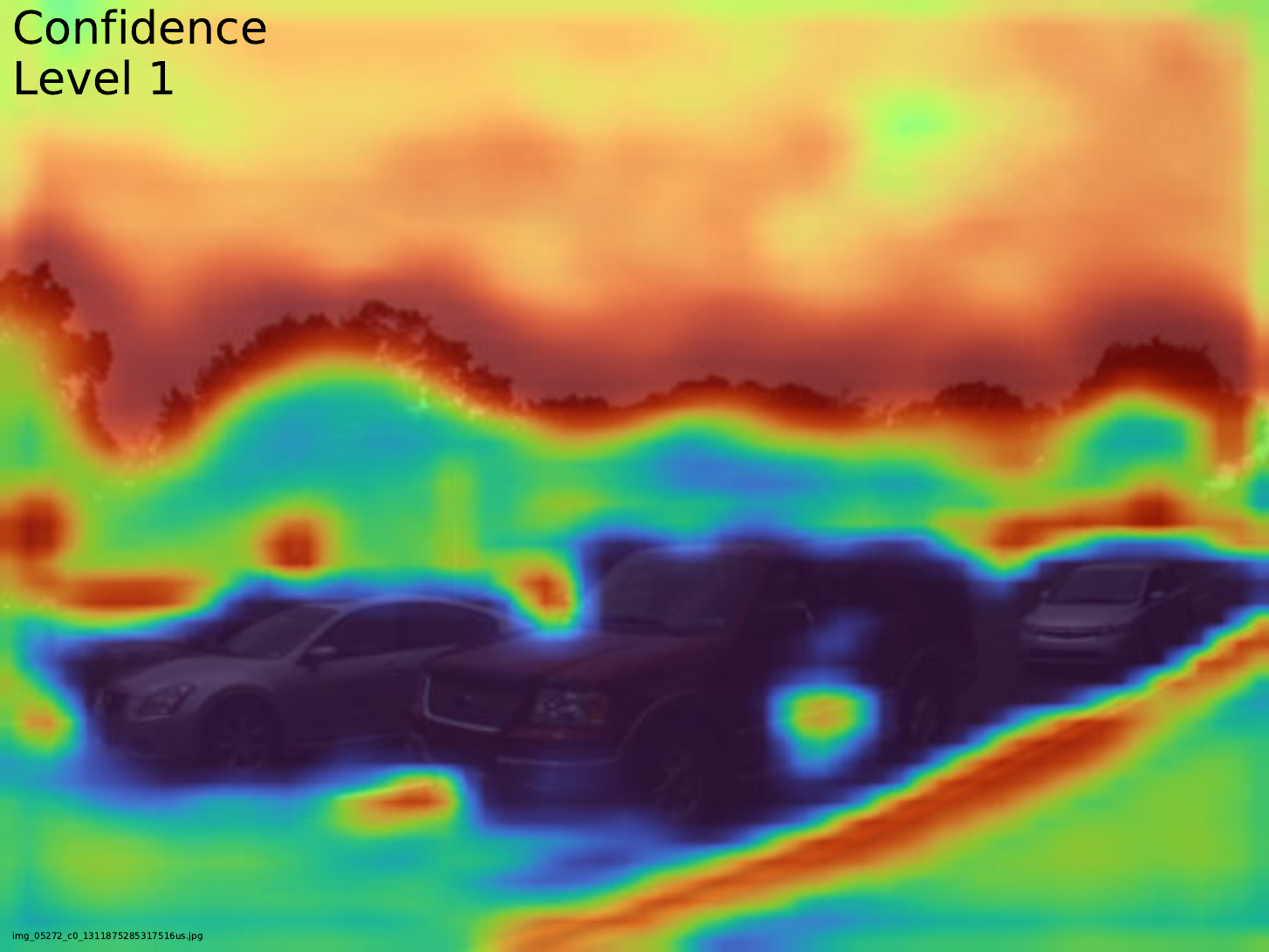}%
    \end{minipage}
    
    \vspace{0.5mm}
    \begin{minipage}{\iwidth\linewidth}
        \includegraphics[width=\linewidth]{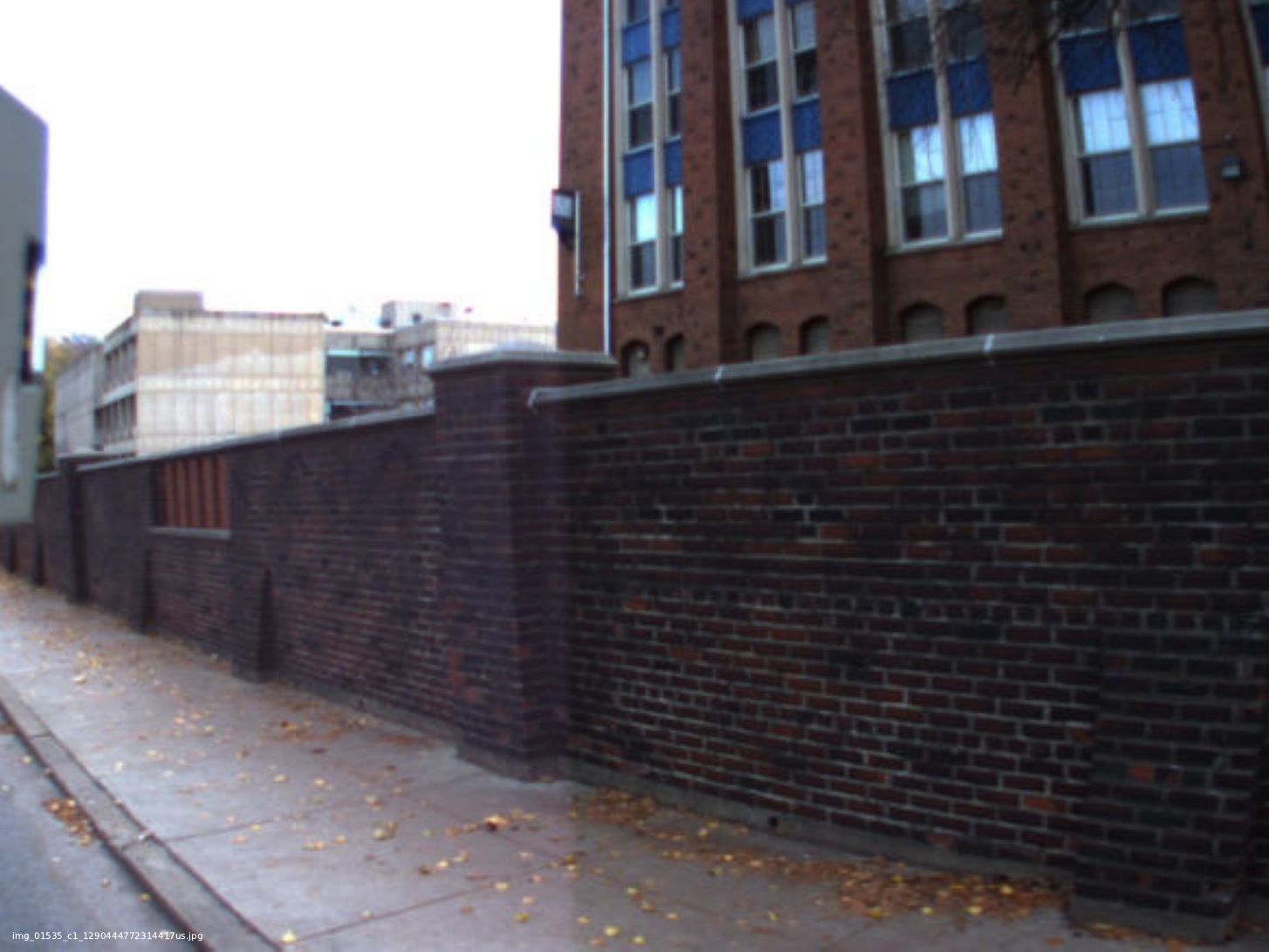}%
    \end{minipage}%
    \hspace{0.05mm}
    \begin{minipage}{\iwidth\linewidth}
        \includegraphics[width=\linewidth]{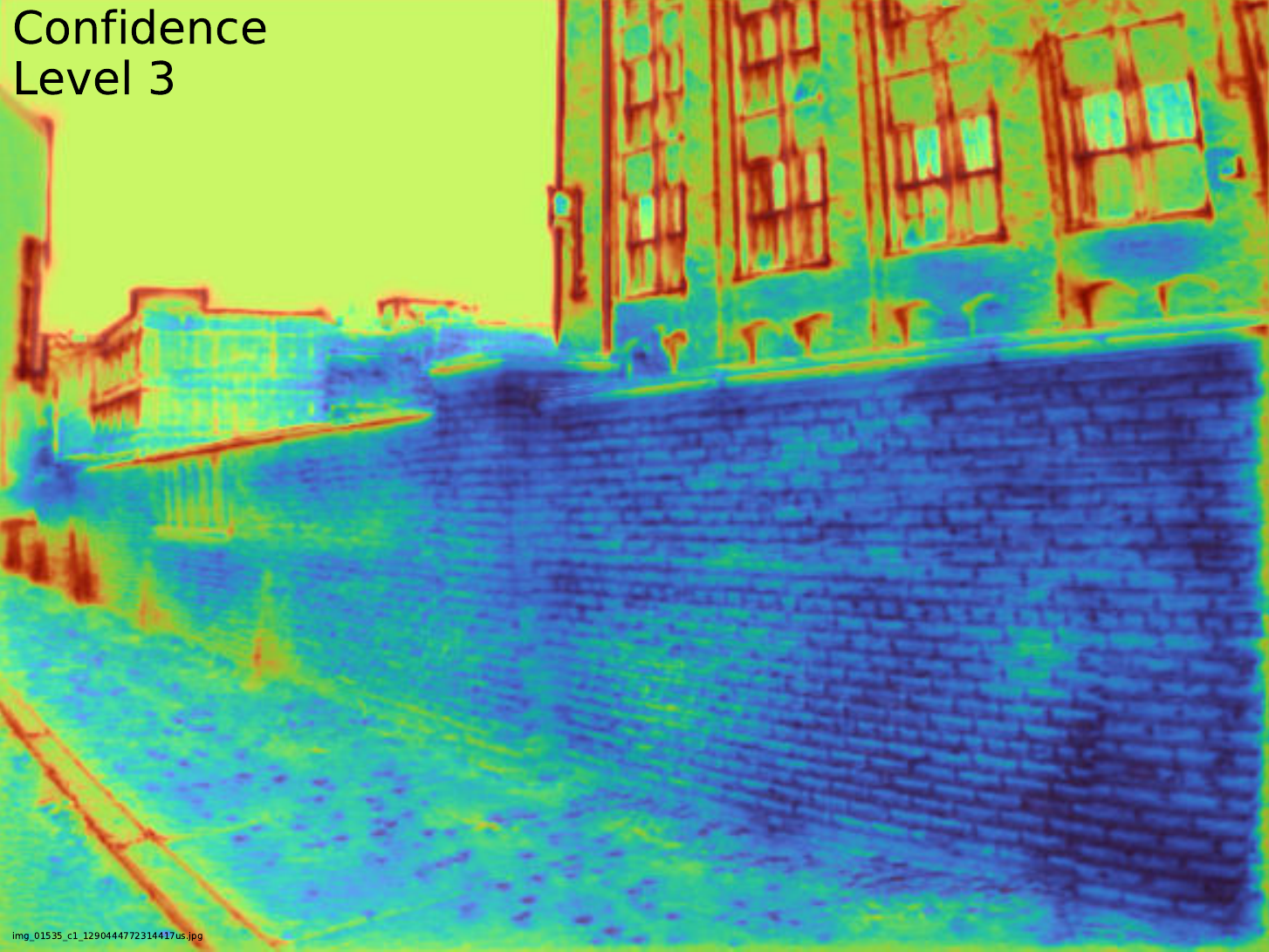}%
    \end{minipage}
    \end{minipage}%
    \hspace{0.1mm}
    \begin{minipage}{0.028\linewidth}
        \centering
        \vspace{2.5mm}
        \includegraphics[width=\linewidth]{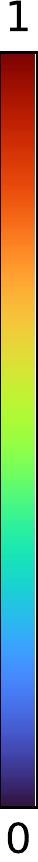}%
    \end{minipage}
    \begin{minipage}{0.02\linewidth}
        \centering
        \vspace{2.7mm}
        \rotatebox[origin=c]{90}{\footnotesize{ignored \hspace{2.5cm} useful}}
    \end{minipage}
    \vspace{1mm}
    \caption{\textbf{Good features to localize.} PixLoc learns to ignore dynamic objects like cars (top) or fallen leaves (bottom) and repeated patterns like the brick wall. It focuses on road markings, silhouettes of trees, or prominent structures on buildings. See also Figure~\ref{fig:examples}.
    }
    \label{fig:uncertainties}%
\end{figure}

\PAR{Fitting the optimizer to the data:} Levenberg-Marquardt is a generic optimization algorithm that involves several heuristics, such as the choice of robust cost function $\rho$ or of the damping factor $\lambda$. Past works on learned optimization employ deep networks to predict $\rho'$~\cite{lv2019taking}, $\lambda$~\cite{tang2018ba, lv2019taking}, or even the pose update $\*\delta$~\cite{clark2018ls, ma2020deep}, from the residuals and visual features. We argue that this can greatly impair the ability to generalize to new data distributions, as it ties the optimizer to the visual-semantic content of the training data. Instead, it is desirable to fit the optimizer to the distribution of poses or residuals but not to their semantic content. As such, we propose to make $\lambda$ a fixed model parameter and learn it by gradient descent along with the CNN.

Importantly, we learn a different factor for each of the 6 pose parameters and for each feature level, replacing the scalar $\lambda$ by $\*\lambda_l\in\mathbb{R}^6$, parametrized by $\*\theta_l$ as
\begin{equation}
    \log_{10}\*\lambda_l = \lambda_{\min} + \mathrm{sigmoid}\left(\*\theta_l\right)\left(\lambda_{\max} - \lambda_{\min}\right) \enspace.
\end{equation}
This adjusts the curvature of the individual pose parameters during training, and directly learns motion priors from the data.
For example, when the camera is mounted on a car or a robot that is mostly upright, we expect the damping for the in-plane rotation to be large.
In contrast, common heuristics treat all pose parameters equally and do not permit a per-parameter damping.
We show in \supp~\ref{sec:supp:diffdataset} that the learned damping parameters vary with the training data.

\subsection{Learning from poses}
As the CNN never sees 3D points, PixLoc can generalize to any 3D structure available. This includes sparse SfM point clouds, dense depth maps from stereo or RGBD sensors, meshes, Lidar scans, but also lines and other primitives.

\PAR{Training:} The optimization algorithm presented here is end-to-end differentiable and only involves operations commonly supported by deep learning frameworks. Gradients thus flow from the pose all the way to the pixels, through the feature and uncertainty maps and the CNN. Thanks to the uncertainties and robust cost, PixLoc is robust to incorrect 3D geometry and works well with noisy reference data like sparse SfM models. During training, an imperfect 3D representation is sufficient -- our approach does not require accurate or dense 3D models.

\PAR{Loss function:} Our approach is trained by comparing the pose estimated at each level $(\*R_l, \*t_l)$ to its ground truth $(\bar{\*R}, \bar{\*t})$. We minimize the reprojection error of the 3D points:
\begin{equation}
    \mathcal{L} = \frac{1}{L} \sum_l \sum_i
    \left\Vert \Pi\left(\*R_l\*P_i+\*t_l\right) - \Pi\left(\bar{\*R}\*P_i+\bar{\*t}\right)\right\Vert_\gamma,
\end{equation}
where $\gamma$ is the Huber cost. This loss weights the supervision of the rotation and translation adaptively for each training example~\cite{kendall2017geometric} and is invariant to the scale of the scene, making it possible to train with data generated by SfM.
To prevent hard examples from smoothing the fine features, we apply the loss at a given level only if the previous one succeeded in bringing the pose sufficiently close to the ground truth. 
Otherwise, the subsequent loss terms are ignored.

\subsection{Comparisons to existing approaches}
\PAR{PixLoc vs.\ sparse matching:} Pose estimation via local feature matching comprises multiple operations that are non-differentiable, such as keypoint and correspondence selection or RANSAC. Bhowmik \etal~\cite{bhowmik2020reinforced} proposed a formulation based on reinforcement learning, which suffers from high variance and thus requires a strong pretraining. In contrast, our approach is extremely simple and converges well from generic weights trained for image classification.

\PAR{PixLoc vs.\ GN-Net:} Von Stumberg \etal~\cite{von2020gn, von2020lm} recently trained deep features for cross-season localization via direct alignment. Their works focus on small-baseline scenarios and require accurate pixelwise ground truth correspondences and substantial hyperparameter tuning. In contrast, we leverage the power of differentiable programming to match the test and training conditions and learn additional strong priors from noisy data. We compare with their loss in Section~\ref{sec:insights}.


\begin{figure}[t]
    \centering
    \def\iwidth{0.49}
    \begin{minipage}{\iwidth\linewidth}
        \includegraphics[width=\linewidth]{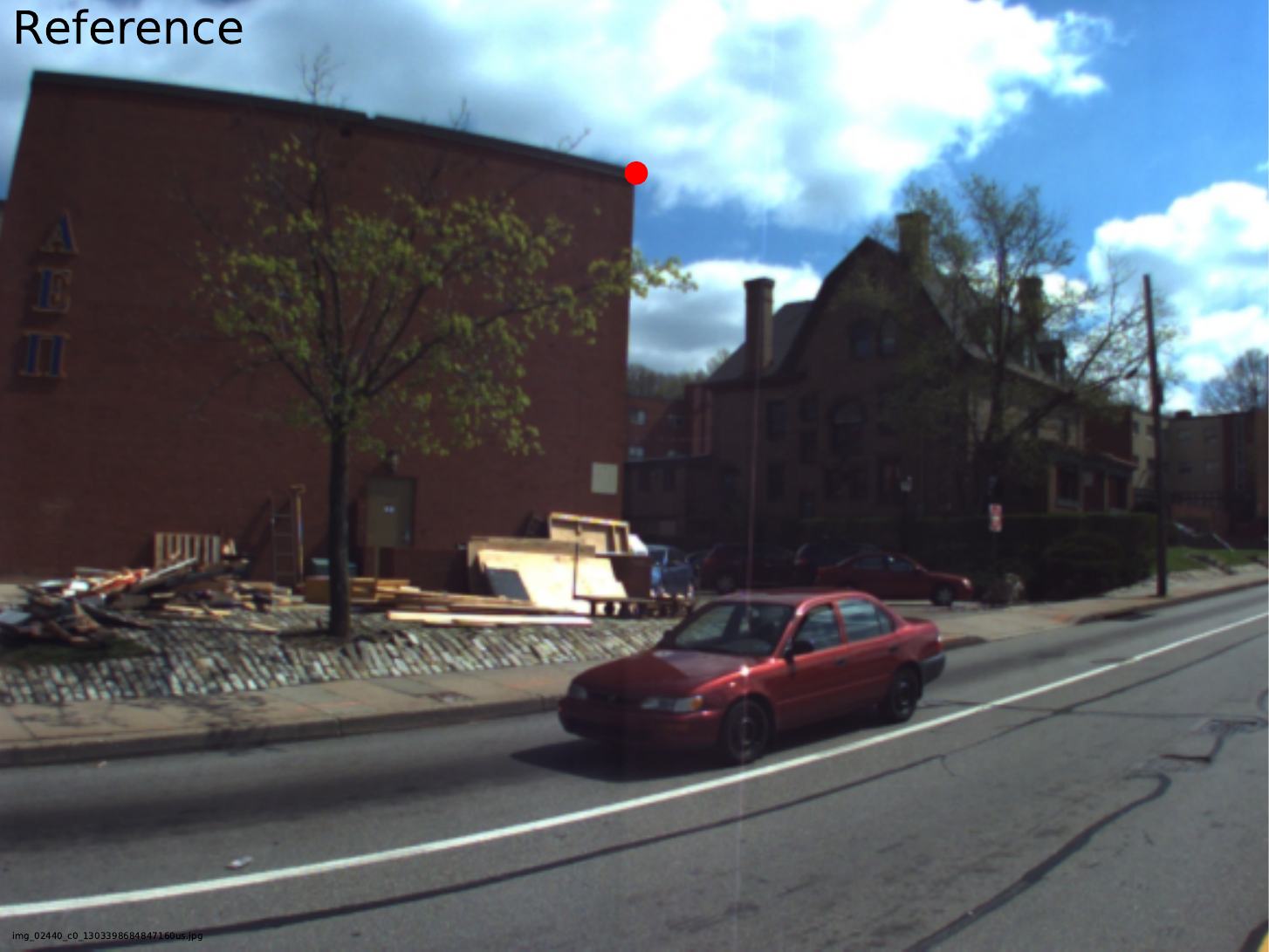}%
    \end{minipage}%
    \hspace{0.05mm}
    \begin{minipage}{\iwidth\linewidth}
        \includegraphics[width=\linewidth]{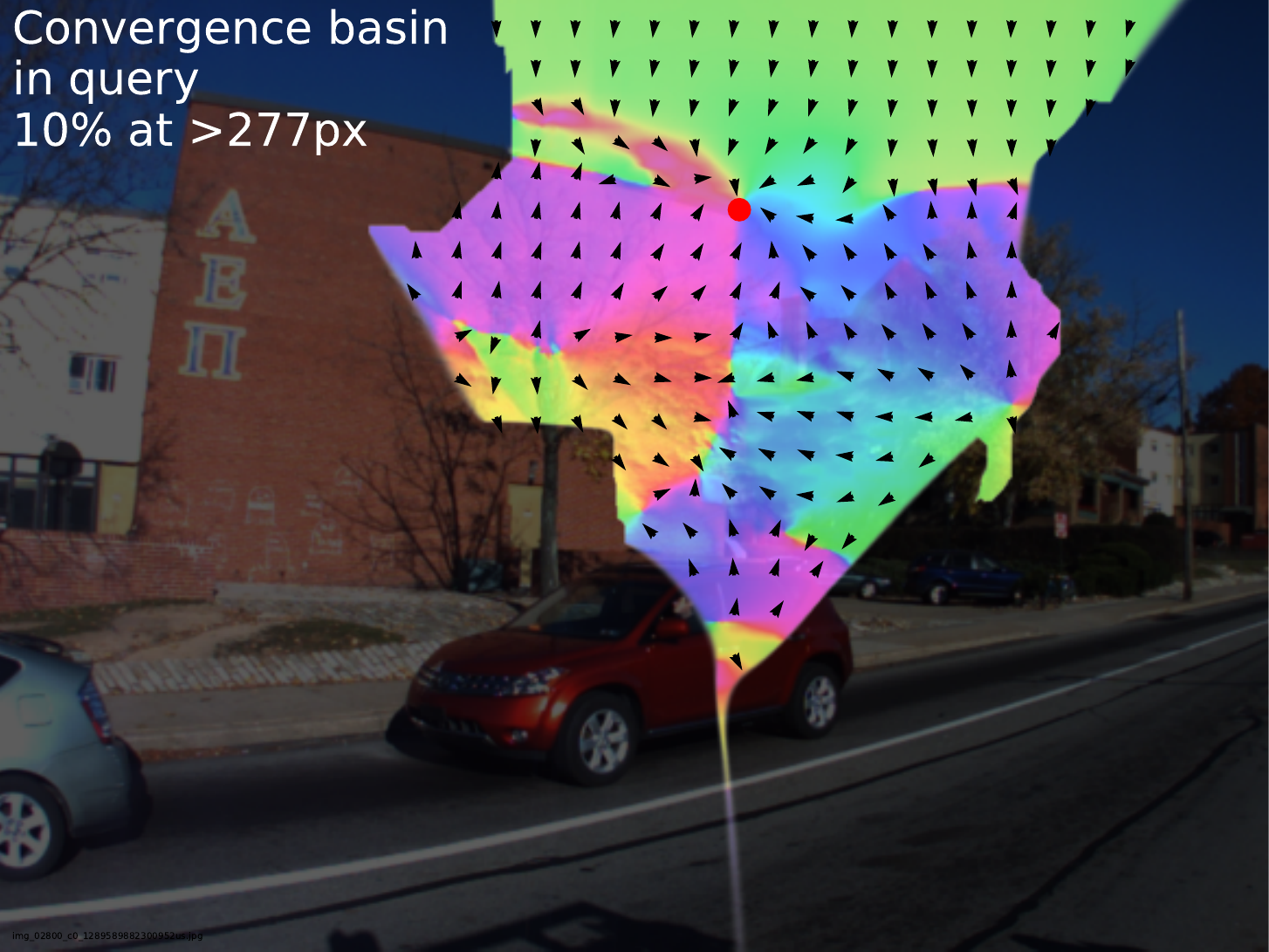}%
    \end{minipage}
    \vspace{1mm}
    \caption{\textbf{Wide convergence.} For a \red{red} point in the reference image (left), we highlight in the query (right) the multilevel basin of attraction colored by the 2D gradient angle $\nicefrac{\partial\*F_q}{\partial\*p_q^i}^\top \*r_k^i$. Deep features ensure a wide convergence despite appearance changes.}
    \label{fig:convergence}%
\end{figure}

\begin{table*}[ht]
\centering
\footnotesize{
\setlength\tabcolsep{2.7pt}\
\begin{tabular}{llccccccccccccc}
    \toprule
    &\multirow{2}{1cm}[-.4em]{Method} & \multicolumn{5}{c}{Cambridge Landmarks - outdoor}
    & \multicolumn{8}{c}{7Scenes - indoor}\\
    \cmidrule(lr){3-7}
    \cmidrule(lr){8-15}
    && Court & King's & Hospital & Shop & St. Mary's & Chess & Fire & Heads & Office & Pumpkin & Kitchen & Stairs & Recall$\uparrow$\\
    \midrule
    \multirow{2}{*}{\begin{sideways}IR\end{sideways}}
    & DenseVLAD~\cite{torii201524} & - & 280/5.7  & 401/7.1 & 111/7.6 & 231/8.0 & 21/12.5 & 33/13.8 & 15/14.9 & 28/11.2 & 31/11.3 & 30/12.3 & 25/15.8 & - \\
    & Oracle & 207/7.0 & 137/7.2 & 323/8.3 & 133/7.8 & 204/8.1 & 16/12.3 & 26/13.6 & 12/14.7 & 20/11.5 & 19/14.0 & 18/15.0 & 17/18.1 & 0.17\\
    \midrule
    \multirow{3}{*}{\begin{sideways}FM\end{sideways}}
    & AS~\cite{sattler2016efficient}$^\dag$ & 24/0.13 & 13/0.22 & 20/0.36 & \04/0.21 & \08/0.25 & \03/0.87 & \02/1.01 & \01/0.82 & \04/1.15 & \07/1.69 & \05/1.72 & \04/1.01 & 68.7\\
    & InLoc~\cite{taira2018inloc} & - & - & - & - & - & \03/1.05 & \03/1.07 & \02/1.16 & \03/1.05 & \05/1.55 & \04/1.31 & \09/2.47 & 66.3\\
    & hloc~\cite{sarlin2019coarse} & 16/0.11 & 12/0.20 & 15/0.30 & \04/0.20 & \07/0.21 & \02/0.85 & \02/0.94 & \01/0.75 & \03/0.92 & \05/1.30 & \04/1.40 & \05/1.47 & 73.1 \\
    \midrule
    \multirow{6}{*}{\begin{sideways} end-to-end \end{sideways}}
    & \red{DSAC*}~\cite{brachmann2020dsacstar} & 49/0.3\0 & 15/0.3\0 & 21/0.4\0 & \0\b{5}/0.3\0 & 13/0.4\0 & \0\b{2}/1.10 & \0\b{2}/1.24 & \0\b{1}/1.82 & \0\b{3}/1.15 & \0\b{4}/1.34 & \04/1.68 & \0\b{3}/1.16 & \b{85.2}\\
    & \red{HACNet}~\cite{li2020hierarchical} & \b{28}/0.2\0 & 18/0.3\0 & 19/\b{0.3}\0 & \06/0.3\0 & \0\b{9}/\b{0.3}\0 & \0\b{2}/\b{0.7}\0 & \0\b{2}/0.9\0 & \0\b{1}/0.9\0 & \0\b{3}/\b{0.8}\0 & \0\b{4}/\b{1.0}\0 & \04/\b{1.2}\0 & \0\b{3}/\b{0.8}\0 & 84.8 \\
    & \red{CAMNet}~\cite{ding2019camnet} & - & - & - & - & - & \04/1.73 & \03/1.74 & \05/1.98 & \04/1.62 & \0\b{4}/1.64 & \04/1.63 & \04/1.51 & -\\
    & SANet~\cite{yang2019sanet} & 328/1.95 & 32/0.54 & 32/0.53 & 10/0.47 & 16/0.57 & \03/0.88 & \03/1.08 & \02/1.48 & \0\b{3}/1.00 & \05/1.32 & \04/1.40 & 16/4.59 & 68.2 \\
    & \b{PixLoc} & 30/\b{0.14} & \b{14}/\b{0.24} & \b{16}/\b{0.32} & \0\b{5}/\b{0.23} & 10/\b{0.34} & \0\b{2}/0.80 & \0\b{2}/\b{0.73} & \0\b{1}/\b{0.82} & \0\b{3}/\b{0.82} & \0\b{4}/1.21 & \0\b{3}/\b{1.20} & \05/1.30 & 75.7\\
    & \gray{+ Oracle prior} & \gray{21/0.12} & \gray{13/0.24} & \gray{16/0.31} & \gray{\05/0.22} & \gray{\09/0.28} &
    \gray{\02/0.80} & \gray{\02/0.70} & \gray{\01/0.78} & \gray{\03/0.80} & \gray{\04/1.13} & \gray{\03/1.14} & \gray{\04/1.08} & \gray{81.7} \\
    \bottomrule
\end{tabular}
}
\vspace{.02in}
\caption{\textbf{Visual localization on the Cambridge Landmarks and 7Scenes datasets.} We report the median translation~(cm) and rotation~(\degree) errors and the average recall at~(5cm, 5\degree).
Despite its simplicity, PixLoc is competitive with complex feature matching (FM) pipelines and performs similarly to, and often better than,  geometric regression models, including those specifically trained per scene (\red{red}).
Our model, trained solely on outdoor data, generalizes well to unseen outdoor and indoor scenes, and can benefit from improved image retrieval (IR).
The best results in the end-to-end category are in bold (oracle excluded). $^\dag$The results for AS were kindly provided by the authors.
}
\label{tab:small-loc}
\end{table*}

\section{Localization pipeline}
PixLoc can be a competitive standalone localization module when coupled with image retrieval, but can also refine poses obtained by previous approaches.
It only requires a 3D model and a coarse initial pose, which we now discuss.

\PAR{Initialization:} 
How accurate the initial pose should be depends on the basin of convergence of the alignment.
Features from a deep CNN with a large receptive field ensure a large basin (Figure~\ref{fig:convergence}).
To further increase it, we apply PixLoc to image pyramids, starting at the lowest resolution, yielding coarsest feature maps of size $W{=}16$.
To keep the pipeline simple, we select the initial pose as the pose of the first reference image returned by image retrieval.
This results in a good convergence in most scenarios.
When retrieval is not sufficiently robust and returns an incorrect location, as in the most challenging conditions, one could improve the performance by reranking using covisiblity clustering~\cite{sattler2016efficient, sarlin2019coarse} or pose verification with sparse~\cite{Zeisl_2015_ICCV, Sattler16CVPR} or dense matching~\cite{taira2018inloc}.

\PAR{3D structure:}
For simplicity and unless mentioned, for both training and evaluation, we use sparse SfM models triangulated from posed reference images using hloc~\cite{hloc, sarlin2019coarse} and COLMAP~\cite{schoenberger2016sfm,schoenberger2016mvs}.
Given a subset of reference images, \eg~top-5 retrieved, we gather all the 3D points that they observe, extract multilevel features at their 2D observations, and average them based on their confidence.

\section{Experiments}
We first compare against existing learning-based localization approaches and show that PixLoc often performs better than those trained for each scene and generalizes well across environments.
We then compare PixLoc with state-of-the-art feature matching pipelines on a large-scale benchmark 
and show that it delivers competitive accuracy, but can also enhance them when used as a post-processing.
Finally, we provide insights into PixLoc through an ablation study.

\PAR{Architecture:} We employ a UNet feature extractor based on a VGG19 encoder pretrained on ImageNet, and extract $L{=}$3 feature maps with strides 1, 4, and 16, and dimensions $D_l{=}$32, 128, and 128, respectively.
PixLoc is implemented in PyTorch~\cite{pytorch},
extracts features for an image in around 100ms,
and optimizes the pose in 200ms to 1s depending on the number of points. More details are in the \supp. 
\PAR{Training:} We train two versions of PixLoc to demonstrate its ability to learn environment-specific priors.
The benefits of such priors are analyzed in \supp~\ref{sec:supp:diffdataset}.
One version is trained on the MegaDepth dataset~\cite{li2018megadepth}, composed of crowd-sourced images depicting popular landmarks around the world, and the other on the training set of the Extended CMU Seasons dataset~\cite{sattler2018benchmarking, badino2011visual,Toft2020TPAMI}, a collection of sequences captured by car-mounted cameras in urban and rural environments. The latter dataset exhibits large seasonal changes with often only natural structures like trees being visible in the images, which are challenging for feature matching.
We sample covisible image pairs and simulate the localization of one image with respect to the other, given its observed 3D points. The optimization runs for 15 iterations at each level and is initialized with the pose of the reference image.

\begin{table*}
\centering
\footnotesize{
\setlength\tabcolsep{2.7pt}\
\begin{tabular}{llccccccc}
    \toprule
    &\multirow{2}{1cm}[-.4em]{Method} & \multicolumn{2}{c}{Aachen Day-Night}
    & \multicolumn{2}{c}{RobotCar Seasons}
    & \multicolumn{3}{c}{Extended CMU Seasons}\\
    \cmidrule(lr){3-4}
    \cmidrule(lr){5-6}
    \cmidrule(lr){7-9}
    && Day & Night & Day & Night & Urban & Suburban & Park\\
    \midrule
    \multirow{3}{*}{\begin{sideways}IR\end{sideways}}
    &   DenseVLAD~\cite{torii201524} &
    \00.0 / \00.1 / 22.8& \00.0 / \01.0 / 19.4& \07.6 / 31.2 / 91.2& \01.0 / \04.4 / 22.7 & 14.7 / 36.3 / 83.9 & \05.3 / 18.7 / 73.9 & \05.2 / 19.1 / 62.0\\
    & NetVLAD~\cite{arandjelovic2016netvlad} &
    \00.0 / \00.2 / 18.9 & \00.0 / \00.0 / 14.3 & \06.4 / 26.3 / 90.9 & \00.3 / \02.3 / 15.9 & 12.2 / 31.5 / 89.8 & \03.7 / 13.9 / 74.7 & \02.6 / 10.4 / 55.9\\
    & \gray{Oracle} & \gray{\00.0 / \00.2 / 22.1} & \gray{\00.0 / \01.0 / 22.4} & \gray{\09.6 / 38.1 / 96.3} & \gray{\04.3 / 16.4 / 84.9} & \gray{21.2 / 52.2 / 98.2} & \gray{\08.6 / 29.5 / 94.3} & \gray{\08.2 / 31.5 / 90.2}\\
    \midrule
    \multirow{3}{*}{\begin{sideways}E2E\end{sideways}}
    & ESAC~\cite{brachmann2019expert} & 
	42.6 / 59.6 / 75.5	& \06.1 / 10.2 / 18.4 &
    - & - & - & - & - \\
    & \b{Pixloc} &
    64.3 / 69.3 / 77.4 & 51.0 / 55.1 / 67.3 & 
	52.7 / 77.5 / 93.9 &	12.0 / 20.7 / 45.4 &
	88.3 / 90.4 / 93.7 & 79.6 / 81.1 / 85.2 & 61.0 / 62.5 / 69.4 \\
    & \gray{+ Oracle prior} &
    \gray{68.0 / 74.6 / 80.8} & \gray{57.1 / 69.4 / 76.5} &  
    \gray{55.8 / 80.8 / 96.4} & \gray{23.6 / 40.3 / 77.8} & \gray{92.8 / 95.1 / 98.5} & \gray{91.9 / 93.4 / 95.8} & \gray{84.0 / 85.8 / 90.9} 
    \\
    \midrule
    \multirow{5}{*}{\begin{sideways}FM\end{sideways}}
    & AS~\cite{sattler2016efficient} &
	85.3 / 92.2 / 97.9 & 39.8 / 49.0 / 64.3 &
	50.9 / 80.2 / 96.6 & 6.9 / 15.6 / 31.7 &
    81.0 / 87.3 / 92.4 & 62.6 / 70.9 / 81.0 & 45.5 / 51.6 / 62.0 \\
    & D2-Net ~\cite{dusmanu2019d2} &
    84.3 / 91.9 / 96.2 & 75.5 / 87.8 / 95.9 &
    54.5 / 80.0 / 95.3 & 20.4 / 40.1 / 55.0 & 
    94.0 / 97.7 / 99.1 & 93.0 / \b{95.7} / \b{98.3} & \b{89.2} / \b{93.2} / \b{95.0}\\
    & S2DNet ~\cite{germain2020s2dnet} &
    84.5 / 90.3 / 95.3 & 74.5 / 82.7 / 94.9 &  
    53.9 / 80.6 / 95.8 & 14.5 / 40.2 / 69.7 &
    - & - & -\\
    & hloc ~\cite{sarlin2019coarse} &
    \b{89.6} / \b{95.4} / \b{98.8} & \b{86.7} / \b{93.9} / \b{100.} &
    \b{56.9} / 81.7 / \b{98.1} & 33.3 / 65.9 / 88.8 &
    95.5 / 98.6 / \b{99.3} & 90.9 / 94.2 / 97.1 & 85.7 / 89.0 / 91.6 \\
    & \b{+ PixLoc refine} &
	84.7 / 94.2 / \b{98.8} & 81.6 / \b{93.9} / \b{100.} &
    \b{56.9} / \b{82.0} / \b{98.1} & \b{34.9} / \b{67.7} / \b{89.5} &
    \b{96.9} / \b{98.9} / \b{99.3} & \b{93.3} / 95.4 / 97.1 & 87.0 / 89.5 / 91.6 \\
    \bottomrule
\end{tabular}
}
\caption{\textbf{Large-scale localization on the Aachen, RobotCar, and CMU datasets.} PixLoc, when initialized from image retrieval (IR), can substantially improve IR accuracy. It consistently outperforms the only scalable end-to-end (E2E) method ESAC, and performs reasonably compared to complex feature matching (FM) pipelines. PixLoc can also improve their accuracy by refining their local features (+ refine). 
}
\label{tab:large-loc}
\end{table*}

\subsection{Comparison to learned approaches}
\label{sec:end2end}
We first evaluate on the Cambridge Landmarks~\cite{kendall2015posenet} and 7Scenes~\cite{shotton2013scene} datasets, which are commonly used to compare learning-based approaches.

\PAR{Evaluation:} The two datasets contain 5 outdoor and 7 indoor scenes, respectively, each composed of posed reference images and query images captured along different trajectories and conditions.
We report for each scene the median translation (cm) and rotation (\degree) errors~\cite{kendall2015posenet}, as well as the average localization recall at (5cm, 5\degree) for 7Scenes~\cite{shotton2013scene}.

\PAR{Baselines:} We compare with multiple state-of-the-art learning-based approaches. Those trained per scene include 3D coordinate regression networks DSAC*~RGB~\cite{brachmann2020dsacstar} and HACNet~\cite{li2020hierarchical}, and CAMNet~\cite{ding2019camnet}, which regresses a relative pose following image retrieval.
SANet~\cite{yang2019sanet} is scene-agnostic.
All methods, including PixLoc, use 3D points from SfM and dense depth maps for Cambridge and 7Scenes, respectively.

We report image retrieval with DenseVLAD~\cite{torii201524} but not PoseNet and its variants as they perform similarly~\cite{sattler2019understanding}.
We also compare with feature matching pipelines. Active Search~(AS)~\cite{sattler2016efficient} performs global matching with SIFT~\cite{lowe2004distinctive}.
InLoc~\cite{taira2018inloc} and hloc~\cite{sarlin2019coarse} first perform image retrieval before matching features to the retrieved images.
The former matches dense deep descriptors and relies on a dense reference 3D model,
while hloc matches SuperPoint~\cite{superpoint} features with SuperGlue~\cite{sarlin2020superglue} and builds a sparse 3D SfM reference point cloud.
PixLoc, trained on MegaDepth, is initialized with image retrieval obtained with either DenseVLAD~\cite{torii201524} or an oracle, which returns the reference image containing the largest number of inlier matches found by hloc.
This oracle shows the benefits of better image retrieval using a more complex pipeline without ground truth information.

\PAR{Results:} The evaluation results are reported in Table~\ref{tab:small-loc}.
On outdoor data, PixLoc consistently outperforms the only end-to-end scene-agnostic method, SANet, and performs similarly to, or better than scene-specific approaches. It is competitive for indoor scenes, despite being trained on outdoor Internet data only. This confirms that deep features are all we need for accurate localization and that they generalize well despite end-to-end training. PixLoc performs comparably to the best feature matching localizer hloc -- a~complex pipeline that integrates learned feature detection, description, and matching.
Localizing with the oracle prior only marginally improves the performance, confirming that image retrieval can be sufficiently accurate for the pose optimization to converge to the correct minimum.

\subsection{Large-scale localization}
\label{sec:exp:vloc}
We now evaluate on a large-scale, long-term localization benchmark~\cite{sattler2018benchmarking} that exhibits considerably more diversity in geometry and appearance than Cambridge and 7Scenes. 

\PAR{Evaluation:} The benchmark is composed of three datasets.
The Aachen Day-Night~\cite{sattler2012image,sattler2018benchmarking} dataset is captured by handheld devices. 
The RobotCar~\cite{Maddern2017IJRR,sattler2018benchmarking} and the Extended CMU~\cite{badino2011visual,Toft2020TPAMI} seasons datasets are captured by car-mounted cameras across different seasons, weather, and times, in urban and rural areas. All datasets have posed reference images, SfM models, and query images.
We report the localization recall at thresholds (25cm,~2\degree), (50cm,~5\degree), and (5m,~10\degree).

\PAR{Baselines:} Multiple past works~\cite{sattler2019understanding, taira2018inloc, schonberger2018semantic, brachmann2019expert} report that end-to-end learning-based methods cannot be stably trained on such large-scale datasets.
The only exception is ESAC~\cite{brachmann2019expert}, which reports results for Aachen only.
We additionally compare against image retrieval with DenseVLAD~\cite{torii201524} and NetVLAD~\cite{arandjelovic2016netvlad} and feature matching pipelines based on Active Search~\cite{sattler2016efficient}, D2-Net~\cite{dusmanu2019d2}, S2DNet~\cite{germain2020s2dnet}, and hloc~\cite{sarlin2019coarse}.
PixLoc is trained on MegaDepth (CMU) when evaluated on Aachen (RobotCar and CMU).
It is initialized by the weighted average~\cite{pion3DV2020} of the top-3 poses retrieved by NetVLAD for Aachen and top-1 for RobotCar and CMU.
The oracle prior is identical to Section~\ref{sec:end2end}.

\PAR{Results:} We report the results in Table~\ref{tab:large-loc}.
When the initial pose prior is provided by image retrieval, PixLoc is a simple localization system that is more accurate than ESAC, especially in the challenging condition of night. 
This improvement is not brought by the significantly less accurate image retrieval.
PixLoc is however less robust than the feature matching pipelines, which is mostly due to the naive pose prior, as our algorithm cannot converge if the retrieval returns the incorrect location. 
Using the oracle prior partially bridges the gap, and makes PixLoc competitive on driving datasets like CMU and RobotCar.
It however lags behind on Aachen, where the reference images are significantly sparser and the initial priors are therefore much coarser. 
Naturally, this is challenging for direct alignment, irrespective of the daytime or nighttime condition. 
PixLoc is nevertheless the only end-to-end trained method that can scale to this large extent without requiring retraining.

\begin{figure*}[t]
    \centering
    \input{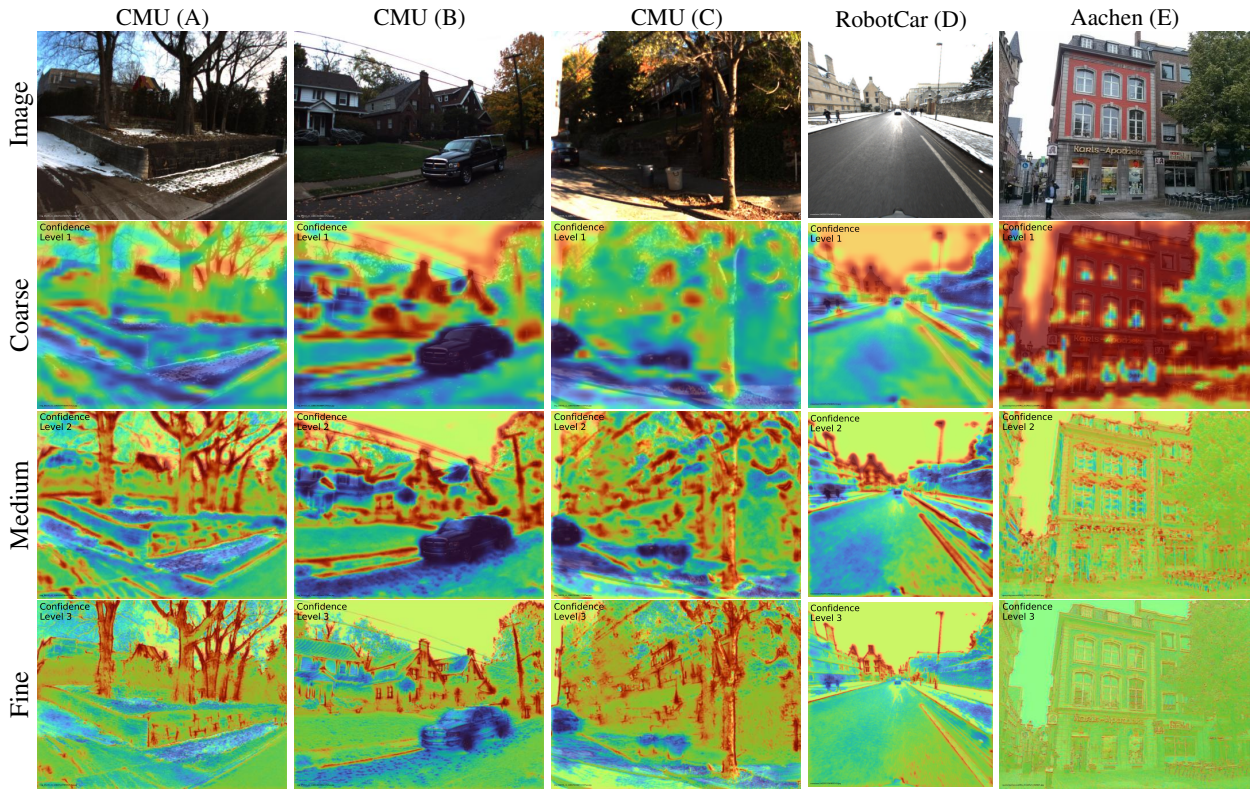}%
    \caption{\textbf{Which features matter?}
    In driving scenarios (A-D), besides dynamic objects such as cars, PixLoc learns to ignore (in blue) more subtle short-term entities like snow (A), fallen leaves (B), trash bins (C), or shadows at all feature levels.
    Instead, it focuses (in red) on poles, tree trunks, road markings, power lines, or building silhouettes.
    Repetitive structures like windows or road cracks are often ignored at first but later on used for fine alignment.
    Differently, when trained on urban scenes (E), it ignores trees as buildings are more stable structures.
    }
    \label{fig:examples}%
\end{figure*}

\begin{table}[tb]
\centering
\begin{minipage}{.57\linewidth}
    \centering
    \footnotesize{\setlength\tabcolsep{2pt}\
\begin{tabular}{lcc}
    \toprule
    \multirow{2}{1cm}[-.4em]{Method} & \multicolumn{2}{c}{AUC}\\
    \cmidrule(lr){2-3}
    & 25cm & 1m \\
    \midrule
    {\large $\color{black}\bullet$} Photometric optimization & \01.2 & \03.3\\
    {\large $\color{blue}\bullet$} + deep features (GN loss) & 13.2 & 21.4 \\
    {\large$\color{green!40!black}\bullet$} + unroll (fixed damping $\lambda$) & 49.2 & 67.0 \\
    {\large$\color{orange}\bullet$} + confidence $w_{i,k}$ & 53.3 & 72.5 \\
    {\large$\color{red}\bullet$} + learned $\lambda$ \b{(PixLoc-full)} & \b{59.8} & \b{79.0} \\
    {\large$\color{cyan}\bullet$} $D{=}128\!\!\to\!\!16$ (PixLoc-light) & 50.4 & 70.2 \\
    \bottomrule
\end{tabular}}
\end{minipage}
\hspace{2mm}%
\begin{minipage}{0.38\linewidth}
    \centering
    \includegraphics[width=\linewidth]{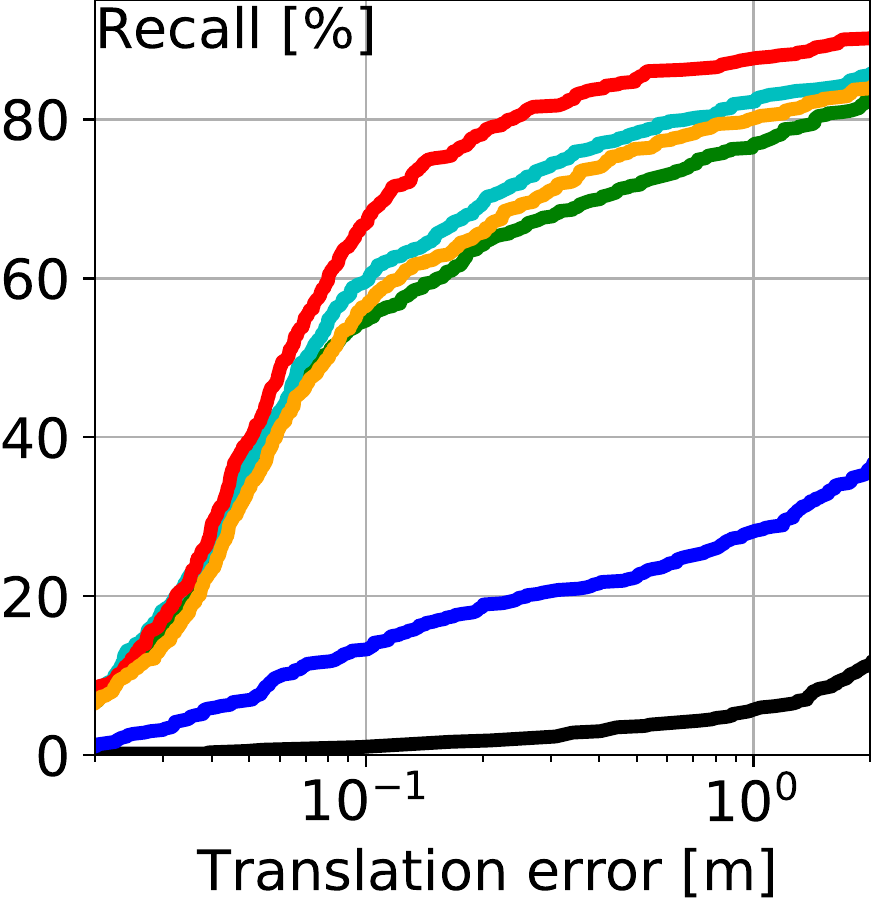}
\end{minipage}
\vspace{1mm}
\caption{\textbf{Ablation study.} Unrolling the optimizer and learning features, damping factor, and confidences all contribute to the performance of PixLoc over classical photometric alignment. Learning compact features as in past works~\cite{lv2019taking, von2020gn} results in a drop of performance compared to high-dimensional representations.
}
\label{tab:ablation}
\end{table}

\subsection{Pose post-processing with PixLoc}
\label{sec:exp:post-processing}
We showed that too large baselines between query and reference images can cause PixLoc to converge to an incorrect local minima. 
Naturally, PixLoc can also serve as a post-processing step for any other localization pipeline. 

\PAR{Refinement in challenging conditions:}
We apply PixLoc to refine the poses estimated by hloc in the previous localization experiment. 
We consider all 3D points that have at least one inlier match.
The results are shown in the last row of Table~\ref{tab:large-loc}. 
PixLoc brings consistent improvement on CMU, especially in the fine threshold, with up to +2.4\% recall.
It also increases the pose accuracy at all thresholds on RobotCar Night, which exhibits significant motion blur, a difficult condition for sparse keypoint detection.
However, no improvement can be observed on RobotCar Day, while the refinement is detrimental on Aachen at 0.25m.
This might be due to inaccurate ground truth poses or camera intrinsics, for which we provide evidence in \supp~\ref{sec:supp:poses}.

\subsection{Additional insights}
\label{sec:insights}

\PAR{Ablation study:} We justify our design decisions by comparing different variants of PixLoc.
We have attempted to train our CNN with the Gauss-Newton loss~\cite{von2020gn}, but it fails to converge on our challenging training data despite extensive hyperparameter tuning.
We select difficult query-reference pairs in the CMU validation set and report the recall curve and its area (AUC) in Table~\ref{tab:ablation}.
As can be seen, all components significantly contribute to PixLoc's performance.

\PAR{Interpretability:} Visualizing the weight maps $u_q$ learned by PixLoc helps us discover what cues are useful or detrimental for localizing in which environments.
We show visualizations in Figure~\ref{fig:examples} and in \supp~\ref{sec:supp:qualitative}.

\PAR{Limitations:}
PixLoc relies on gradients of CNN features, which can only encode a limited context.
It is thus a local method and can fall into incorrect minima for excessively large initial reprojection errors arising from large viewpoint changes.
We analyze the convergence in \supp~\ref{sec:supp:initposes}.
PixLoc can also fail for large outliers ratios due prominent occluders and is more sensitive to camera miscalibration.

{\footnotesize%
\PAR{Acknowledgements:}
The authors thank Mihai Dusmanu, Rémi Pautrat, and Xingxing Zuo for their thoughtful comments.
This work received funding through the EU Horizon 2020 project RICAIP (grant agreeement No 857306), the European Regional Development Fund under project IMPACT No. CZ.02.1.01/0.0/0.0/15 003/0000468, the Chalmers AI Research Centre (CHAIR) (VisLocLearn), and the Swedish Foundation for Strategic Research (Semantic Mapping and Visual Navigation for Smart Robots).
Paul-Edouard Sarlin was supported by gift funding from Huawei, and Viktor Larsson by an ETH Zurich Postdoctoral Fellowship.
\par
}

\section{Conclusion}
In this paper, we have introduced a simple solution to end-to-end learning of camera pose estimation.
In contrast to previous approaches that regress geometric quantities, we do not try to teach a deep network basic geometric principles or 3D map encoding.
Instead, we go Back to the Feature: we show that learning robust and generic features is sufficient for accurate localization by leveraging classical image alignment with existing 3D maps.
To the best of our knowledge, the resulting system, PixLoc, is the first end-to-end trainable approach capable of being deployed into new scenes widely differing from its training data without retraining or fine-tuning. 
PixLoc achieves a pose accuracy competitive with significantly more complex state-of-the-art pipelines. 
End-to-end training combined with uncertainty modeling enables PixLoc to learn complex yet interpretable priors.

PixLoc learns which features and objects matter for robust, long-term localization. 
Yet, it requires a good initialization to successfully localize.
We thus see PixLoc as a first step towards deep networks that learn and reason about long-term, extreme changes of appearance and 3D structure.
We believe that taking steps towards human-level spatiotemporal understanding will ultimately lead to robust, reliable, and accurate localization systems.

\newpage
\fi 

\ifproceedings
    \ifsupponly
        \appendix

\ifproceedings
\pagestyle{plain}
\begin{strip}
\begin{center}
    {
        {\Large \bf Back to the Feature: Learning Robust Camera Localization from Pixels to Pose}%
        \vspace{0.5cm}\\
        \large
        Paul-Edouard Sarlin$^{1}$\printfnsymbol{1} \hspace{.03in}
        Ajaykumar Unagar$^{2}$\printfnsymbol{1}\hspace{.03in}
        M{\aa}ns Larsson$^{3,4}$\hspace{.03in}
        Hugo Germain$^{5}$ \hspace{.03in}
        Carl Toft$^{3}$\\
        Viktor Larsson$^{1}$\hspace{.01in}
        Marc Pollefeys$^{1,6}$\hspace{.01in}
        Vincent Lepetit$^{5}$\hspace{.01in}
        Lars Hammarstrand$^{3}$\hspace{.01in}
        Fredrik Kahl$^{3}$\hspace{.01in}
        Torsten Sattler$^{3,7}$%
        \vspace{0.3cm}\\
        $^{1}$ Department of Computer Science, ETH Zurich\hspace{0.07in}
        $^{2}$ ETH Zurich\hspace{0.07in}
        $^{3}$ Chalmers University of Technology\\
        $^{4}$ Eigenvision\hspace{0.07in}
        $^{5}$ Ecole des Ponts\hspace{0.07in}
        $^{6}$ Microsoft\hspace{0.07in}
        $^{7}$ Czech Technical University in Prague
    }
\end{center}
\end{strip}
\cvprrulercount 0\relax
\setcounter{page}{1}
\fi
\section*{\supp}
\ifproceedings
In the following pages, we provide additional details on the experiments conducted in the main paper.
Section~\ref{sec:supp:initposes} analyzes the convergence of the alignment depending on the accuracy of the initial pose.
In Section~\ref{sec:supp:diffdataset}, we evaluate the benefits of learning environment-specific priors.
In Section~\ref{sec:supp:gtdepth}, we assess the impact of the accuracy of the 3D model.
Section~\ref{sec:supp:poses} analyzes the ground truth poses of the RobotCar dataset, supporting the results of the evaluation performed in Section~\ref{sec:exp:post-processing}.
In Section~\ref{sec:supp:basin}, we explain the computation of the convergence basin shown in Figure~\ref{fig:convergence} and provide additional examples.
In Section~\ref{sec:supp:qualitative}, we provide qualitative examples for the localization experiment of Section~\ref{sec:exp:vloc}.
Lastly, in Section~\ref{sec:supp:implementation}, we provide details on the implementation of PixLoc, its training, and the ablation study presented in Section~\ref{sec:insights}.
\fi

\section{Convergence and initial pose} 
\label{sec:supp:initposes}
\PAR{Convergence:}
The pose optimization in PixLoc tends to converge to spurious local minima if the initial pose is too coarse, such as on the Aachen dataset, in which reference images are sparse. Since the receptive field of the CNN is limited, the convergence mostly depends on the initial 2D reprojection error, which accounts for the rotation and translation errors and for the distance to the 3D structure.
The exact density of reference images required for high success thus depends on the distance to the scene.

We report in Figure~\ref{fig:convergence_threshold} the success rate for different initial reprojection errors and their distribution for the oracle retrieval, with hloc as pseudo ground truth. Convergence within 1 meter is observed for 80\% of the cases only when the initial error is smaller than 200 pixels and is significantly reduced for larger errors. 

\PAR{Initial pose:}
The 7Scenes and Cambridge datasets have reference poses with a high density.
In driving scenarios like in the RobotCar and CMU datasets, there are no rotation changes between reference and query poses.
In all these scenarios, initializing PixLoc with the pose of the first retrieved image is therefore sufficient.

To improve the performance on the Aachen dataset, the results in Table~\ref{tab:large-loc} rely on additional filtering steps.
We first cluster the top-3 retrieved reference images based on their covisibility~\cite{sattler2016efficient, sarlin2019coarse} and only retain the images that belong to the largest cluster.
We then perform a weighted average of the reference poses~\cite{markley2007averaging}, where the weights are computed from the similarity of the global descriptors~\cite{pion3DV2020}.
We compare in Table~\ref{tab:pose-approx} the results obtained with this pose averaging and with the top-1 retrieval.
To further improve the convergence, one could also rerank based on featuremetric error or initialize with poses randomly sampled around top-retrieved poses.

\section{Benefits of training on different datasets}
\label{sec:supp:diffdataset}

The training datasets CMU and MegaDepth reflect different scenarios, autonomous driving and tourism landmark photography, respectively.
Training on each one separately allows to learn task-specific priors and demonstrates the ability of PixLoc to adapt to the environment.

Each dataset depicts scenes with different semantic elements (street-level landscapes and urban landmarks, respectively) and different changes of conditions (weather and season for CMU, cameras, occluders, and viewpoints for MegaDepth). 
Figure~\ref{fig:examples} mentions that the models learn to ignore different unreliable elements depending on the training dataset.
For example, tree silhouettes are reliable on CMU due to the small viewpoint changes, but are ignored by the model trained on MegaDepth.

Cameras also exhibit different motions, as they are either car-mounted or hand-held.
Such priors are learned by the model through the damping factors, which we visualize in Figure~\ref{fig:damping}.
On CMU, the motion across query and reference images is mostly a translation along the $x$ and $z$ axis of the camera, and never along the $y$ axis (fixed height above the ground plane) or a rotation around the $z$ axis (fixed roll).
Differently, the motion on MegaDepth is more uniformly distributed among the 6 DoF, resulting in similar factors.
The relative scale between the two sets of factors is irrelevant.

These learned priors have a noticeable impact on the performance, as shown in Table~\ref{tab:cross-dataset}.
Training on CMU performs better than training on MegaDepth when evaluating on a driving dataset like RobotCar.
When evaluating on a totally different environment like Aachen, it however still performs better than a scene-specific approach like ESAC (shown in Table~\ref{tab:large-loc}).
PixLoc thus generalizes well across scenarios but can also learn and exploit their specificities.

\begin{figure}[tb]
    \centering
    \includegraphics[width=\linewidth]{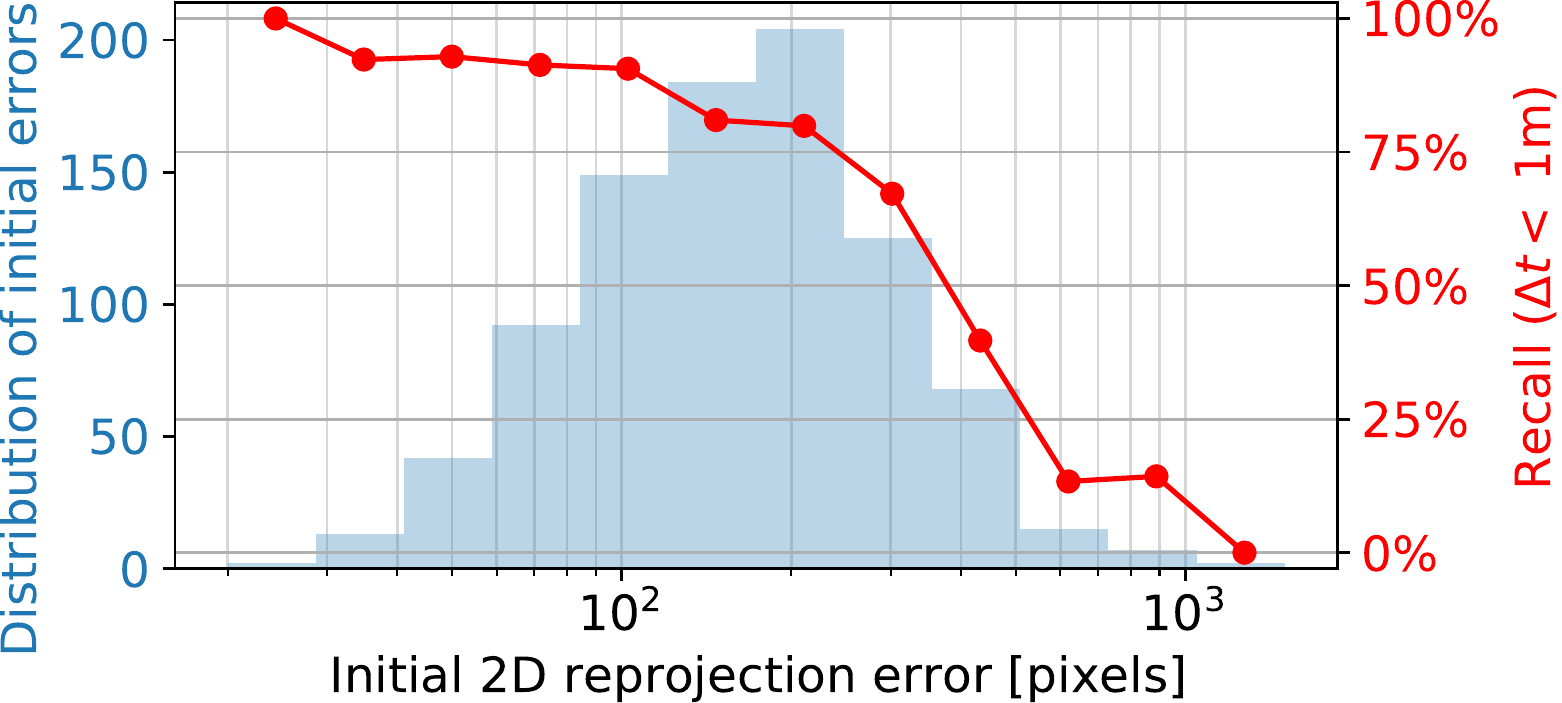}%
    \caption{\textbf{Impact of the initial pose on the Aachen dataset.} The success of the pose optimization decreases with larger initial reprojection errors, which vary significantly across the 922 queries.
    }
    \label{fig:convergence_threshold}%
\end{figure}
\begin{table}[t]
    \centering
    \footnotesize{\setlength\tabcolsep{2.7pt}
\begin{tabular}{lccc}
    \toprule
    \multirow{2}{*}[-.4em]{Initial pose} & \multicolumn{2}{c}{Aachen Day-Night} & \multicolumn{1}{c}{CMU Seasons}\\
    \cmidrule(lr){2-3}
    \cmidrule(lr){4-4}
    & Day & Night & Park\\
    \midrule
    top-1 & 61.7 / 67.6 / 74.8 & 46.9 / 53.1 / 64.3 & 61.0 / 62.5 / 69.4\\
    top-3 averaging & 64.3 / 69.3 / 77.4 & 51.0 / 55.1 / 67.3 & 64.9 / 66.8 / 71.7\\ 
    oracle prior & 68.0 / 74.6 / 80.8 & 57.1 / 69.4 / 76.5 & 84.0 / 85.8 / 90.9\\
    \bottomrule
\end{tabular}}
    \vspace{.06in}
    \caption{\textbf{Selection of the initial pose.}
    Averaging the poses of the top retrieved images improves the convergence of PixLoc compared to simply selecting the pose of the first image.
    }
    \label{tab:pose-approx}%
\end{table}
\begin{figure}[t]
    \centering
    \includegraphics[width=0.9\linewidth]{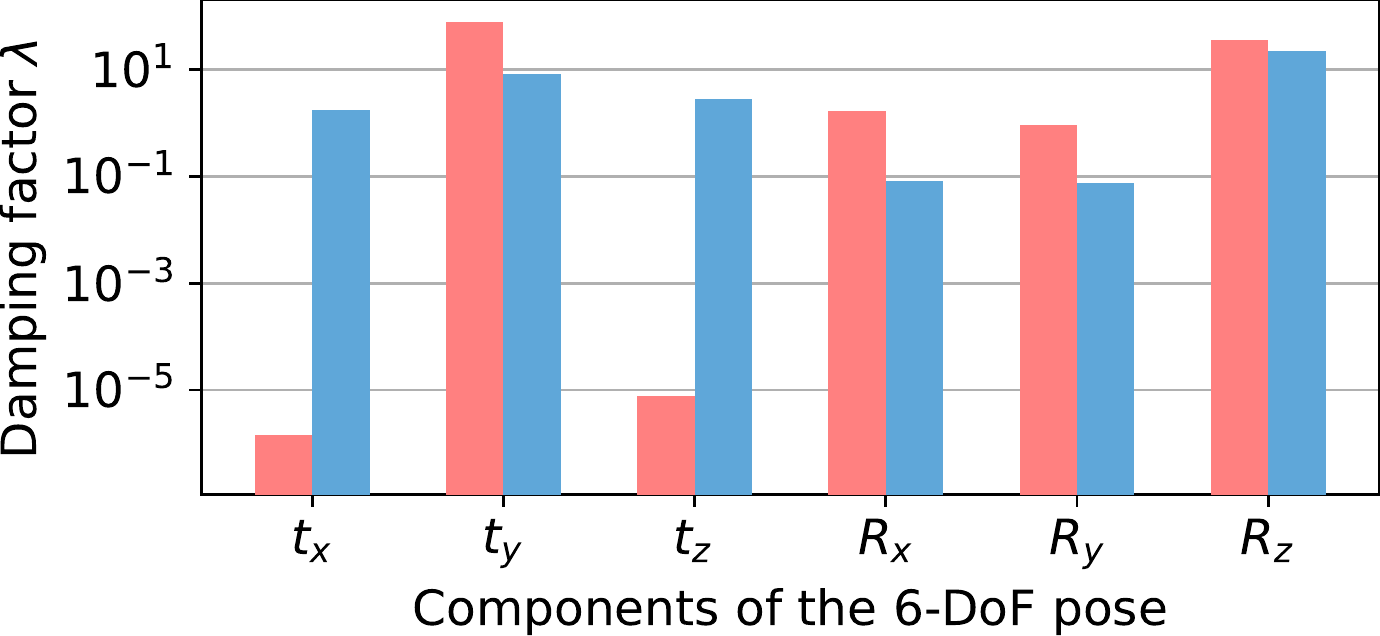}%
    \caption{\textbf{Learned motion prior.} Training on data recorded with 3-DoF car-mounted cameras (CMU, in \red{red}) or with 6-DoF hand-held devices (MegaDepth, in \blue{blue}) results in different motion priors learned by the damping factor $\*\lambda$. Larger relative values indicate smaller expected motion in the corresponding direction.
    }
    \label{fig:damping}%
\end{figure}

\section{Accuracy of the 3D model}
\label{sec:supp:gtdepth}

When localizing on the Cambridge Landmarks dataset, PixLoc relies on SfM models triangulated by hloc~\cite{sarlin2019coarse, sarlin2020superglue}.
For indoor scenes of the 7Scenes dataset, we found that the 3D SfM points are less accurate than the dense depth provided with the dataset.
The results in the main paper (Table~\ref{tab:small-loc}) are thus based on this dense depth.

More specifically, we rely on the depth maps rendered by Brachmann~\etal~\cite{brachmann2020dsacstar}, which are aligned to the color images and are less noisy than the original depth maps.
We simply replace each 3D SfM point by back-projecting one of its 2D observations using the interpolated depth and the image pose.
This 3D model has the same sparsity as the SfM point cloud but is more accurate.
This process is fair as it relies on the same data as all other learning-based approaches, which use the full dense 3D model for training.

We show in Table~\ref{tab:gt-depth} the impact on the performance of PixLoc.
Using this corrected 3D model results in more accurate localization than the triangulated SfM model.

\begin{table}[t]
    \centering
    \resizebox{\columnwidth}{!}{%
    \footnotesize{\setlength\tabcolsep{2.7pt}
\begin{tabular}{lcccc}
    \toprule
    \multirow{2}{1cm}[-.4em]{Training dataset} & \multicolumn{2}{c}{Aachen (urban scenes like MD)} & \multicolumn{2}{c}{CMU (natural scenes)}\\
    \cmidrule(lr){2-3}
    \cmidrule(lr){4-5}
    & Day & Night & Urban & Park\\
    \midrule
    MD & \b{68.0} / \b{74.6} / \b{80.8} & \b{57.1} / \b{69.4} / \b{76.5} & 78.3 / 81.8 / 94.6 & 72.5 / 75.5 / 90.3  \\
    CMU & 54.4 / 62.6 / 74.3 & 46.9 / 54.1 / 68.4 & \b{91.9} / \b{93.4} / \b{95.8} & \b{84.0} / \b{85.8} / \b{90.9} \\
    \bottomrule
\end{tabular}}%
    }
    \vspace{.03in}
    \caption{\textbf{Cross-dataset evaluation with oracle prior.}
    Training and testing in different environments does not perform as well as training for the target distribution.
    Task-specific priors learned by PixLoc, like semantics and motion, are thus largely beneficial.
    }
    \label{tab:cross-dataset}%
\end{table}

\begin{table}[t]
    \centering
    \resizebox{\columnwidth}{!}{%
    \footnotesize{\setlength\tabcolsep{2.7pt}
\begin{tabular}{lcccccccc}
    \toprule
    \multirow{2}{0.8cm}[-.4em]{3D from} & \multicolumn{7}{c}{median error in translation/rotation~(cm/\degree) $\downarrow$} & \multirow{2}{*}[-.2em]{R$\uparrow$}\\
    \cmidrule(lr){2-8}
    & Chess & Fire & Heads & Office & Pumpkin & Kitchen & Stairs & \\
    \midrule
    SfM & 3/0.90 & \b{2}/0.87 & \b{1}/\b{0.79} & \b{3}/0.96 & 5/1.42 & 4/1.44 & 6/1.38 & 69.5 \\
    RGB-D & \b{2}/\b{0.80} & \b{2}/\b{0.73} & \b{1}/0.82 & \b{3}/\b{0.82} & \b{4}/\b{1.21} & \b{3}/\b{1.20} & \b{5}/\b{1.30} & \b{75.7} \\
    \bottomrule
\end{tabular}}%
    }
    \vspace{.03in}
    \caption{\textbf{Depth sensor fusion vs.\ SfM point cloud.}
    For the 7Scenes indoor environment, localizing with 3D points obtained from depth maps fused across multiple view (RGB-D SLAM) is more accurate than with point clouds triangulated via SfM.
    }
    \label{tab:gt-depth}%
\end{table}

\section{Inaccuracy of the ground truth poses}
\label{sec:supp:poses}

The RobotCar v2 dataset has publicly available ground truth poses for a subset of the queries. We project 3D SfM points into the query images using ground truth poses and those estimated by hloc. We observe in many instances a large reprojection error, where hloc poses look qualitatively more accurate. Some examples are shown in Figure~\ref{fig:robotcar_gt}. This might explain why no method localizes more than 58\% of the daytime images at the finest threshold according to the public leaderboard~\footnote{\url{https://www.visuallocalization.net/benchmark/}}.
This might also explain why refining poses with PixLoc does not show improvements for the day-time queries of RobotCar, as observed in Section~\ref{sec:exp:post-processing}.

Similar artifacts were found in training sequences of the Extended CMU Seasons dataset, making the training supervision noisier.
We however do not know if this also applies to the poses of the test sequences because such poses are not publicly available.

\section{Qualitative examples}
\label{sec:supp:qualitative}
We show examples of successful localization on the Extended CMU Seasons dataset in Figure~\ref{fig:qualitative_cmu_success}. We show failure cases in Figure~\ref{fig:qualitative_cmu_failure}. Similarly, we show successful and failed examples on the Aachen Day-Night dataset in Figures~\ref{fig:qualitative_aachen_success} and~\ref{fig:qualitative_aachen_failure}, respectively. 
Videos and animations of the uncertainties and the optimization are available along with the code and trained weights at \href{https://github.com/cvg/pixloc}{\texttt{github.com/cvg/pixloc}}.

\section{Attraction basin}
\label{sec:supp:basin}
\PAR{Computation:} We compute the basin of attraction of a given point by backtracking feature gradients throughout the levels and scales.
For each pixel, we consider the 2 neighbors, in an 8-connected neighborhood, that are in the direction opposite to the feature gradient $\nicefrac{\partial\*F_q}{\partial\*p_q^i}^\top \*r_k^i$.
A pixel is in the basin of attraction if any of those two are themselves in the basin.
The voting is performed in a soft manner using the gradient angle, resulting in a basin score for each pixel.
We first label the point of interest as in the basin and then iteratively run  the algorithm at each level, from the finest to the coarsest level, moving to the next one when the scores stop changing.
Note that the total convergence basin of the pose, which corresponds to the aggregation of all the points, might be smaller or larger.

\PAR{Visualization:}
We show one example in Figure~\ref{fig:convergence} in the main paper, where we color pixels that belong to the basin by changing their hue according to the angle of the total gradient.
We show additional examples in Figure~\ref{fig:convergence_supp}, but showing the gradient field as arrows only.

\section{Experimental details}
\label{sec:supp:implementation}
We now provide more details about the implementation of PixLoc and the experiments.

\PAR{Implementation:}
The CNN and the optimizer are implemented in PyTorch~\cite{pytorch}.
The linear system of the Levenberg-Marquardt step (Equation~\ref{eq:lm_update}) is solved using the Cholesky decomposition.
The lookup of features and uncertainty is computed via bilinear interpolation.
We use the Cauchy robust cost function with scale 0.1.
When computing the residuals or the Jacobian matrix, we ignore points that project outside the image or within 2 pixels of the image borders.
We set $\lambda_{\mathrm{min}}{=}{-}6$ and $\lambda_{\mathrm{max}}{=}5$.

\PAR{Training:}
We train PixLoc with image pairs composed of a query image and a single reference image.
For each pair, we sample 512 3D points visible in the reference image according to the SfM covisibility information.
We apply gradient checkpointing to each block of the encoder and of the decoder to minimize the GPU memory consumption.
The network is trained for 50k iteration with a constant learning rate of $10^{-5}$ and the Adam optimizer~\cite{kingma2014adam}.
To stabilize the training, the average loss per pair is clamped to 50 pixels and the per-parameter gradients are clipped to $[-1, 1]$.

When training on the CMU dataset, we use slices 8-12 and 22-25 for training and slices 6, 13, 21 for validation.
We train with batches of 3 image pairs.
The images are resized such that their smallest dimension is 720 pixels and we sample square crops of 720 pixels
The query pose is initialized with the pose of the reference image.

When training on the MegaDepth dataset, we use the same split of scenes as Dusmanu~\etal~\cite{dusmanu2019d2} and sample image pairs with an overlap score in $[0.3, 1]$.
In addition, we rotate images that are not upright using the gravity direction of each scene.
All images are resized such that their smallest dimension is 512 pixels, and we sample square crops of 512 pixels.
PixLoc is then trained with batches of 6 image pairs.
The initial pose is sampled in the range $t\in[0.75, 1]$ of the linear interpolation between the reference pose ($t{=}0$) and the ground truth query pose ($t{=}1$). 
Sampling initial poses that are too difficult can result in coarse features that are too smooth and uninformative at the lower-resolution scale.

\PAR{Inference:}
In order to keep the runtime reasonable, we use 5 or 3 reference images when initializing from hloc or retrieved reference poses, respectively.
The optimization runs at each level for at most 100 iterations, but stops when either the gradient or the step are small enough.
When refining hloc poses, we only optimize over the medium and fine levels as the initial estimate is always sufficiently good.
All images are resized such that their longest dimension is equal to 1024 pixels.
For the multiscale inference, the resized images are successively aligned at scale 1/4 and 1.

\PAR{Ablation study:} We sample 2000 query images from slices 6, 7, 13, and 21 of the CMU dataset.
To generate challenging pairs, we select the closest reference image that is at least 4 meters away.
For the baseline based on a fixed damping factor $\lambda$, we use $\lambda{=}10^{-2}$.
As GN-Net~\cite{von2020gn} has no publicly-available implementation, we reimplemented it and trained it with our settings on the CMU dataset.
The GN-Net loss has several hyperparameters: the Gauss-Newton sampling vicinity, the weight of the contrastive loss, and the margin of its negative term.
We performed an extensive hyperparameter search and report the best results obtained.
Our training data is significantly more difficult than the one used in the original paper~\cite{von2020gn}, with significantly larger baselines and appearance changes.
This explains the large performance gap observed in Table~\ref{tab:ablation} compared to the results originally reported.

\begin{figure}[t]
    \centering
    \def\iwidth{0.49}
    \begin{minipage}{\iwidth\linewidth}
        \centering
        \footnotesize{Query image}%
    \end{minipage}%
    \hspace{0.01mm}
    \begin{minipage}{\iwidth\linewidth}
        \centering
        \footnotesize{Nearest reference image}
    \end{minipage}
    
    \begin{minipage}{\iwidth\linewidth}
        \includegraphics[width=\linewidth]{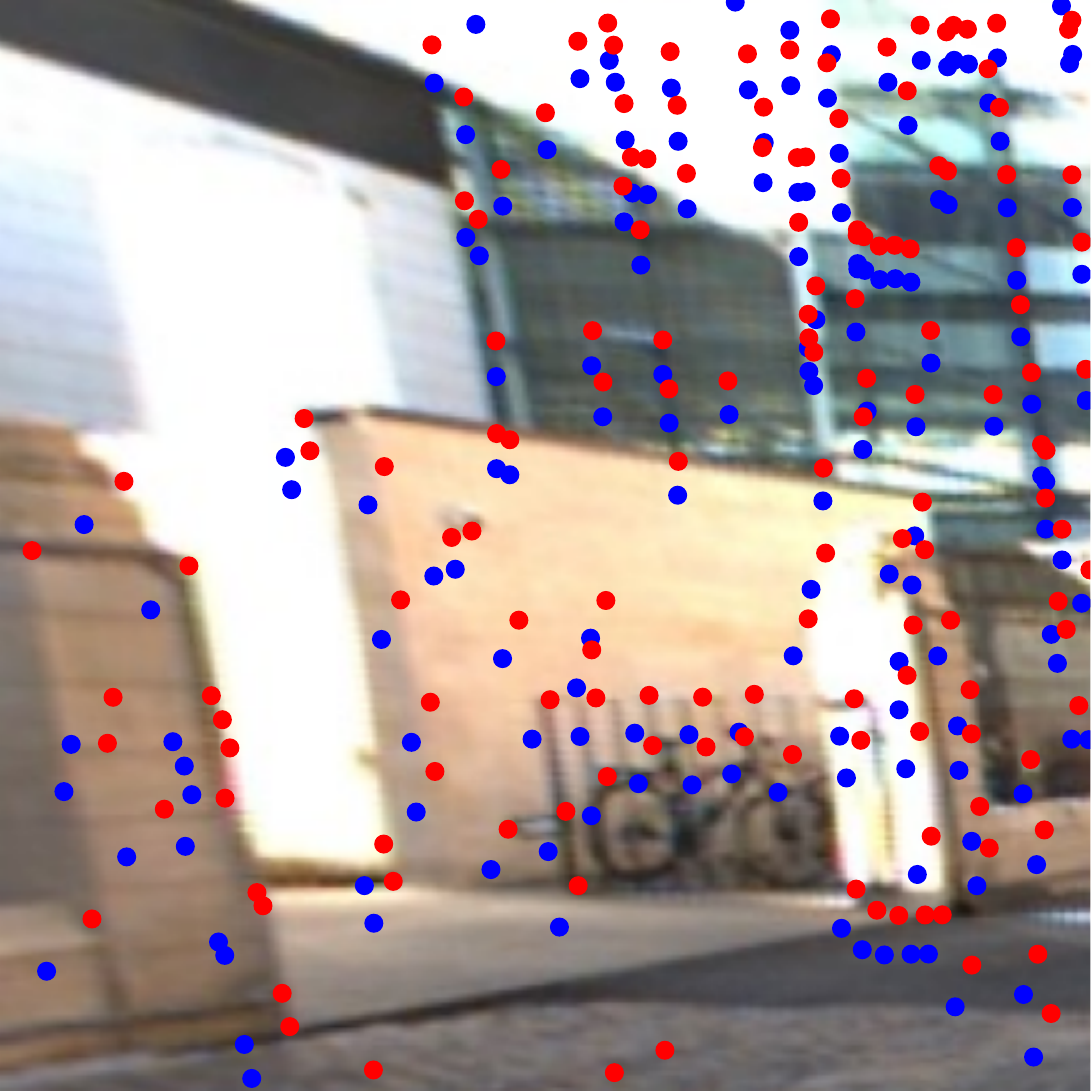}%
    \end{minipage}%
    \hspace{0.01mm}
    \begin{minipage}{\iwidth\linewidth}
        \includegraphics[width=\linewidth]{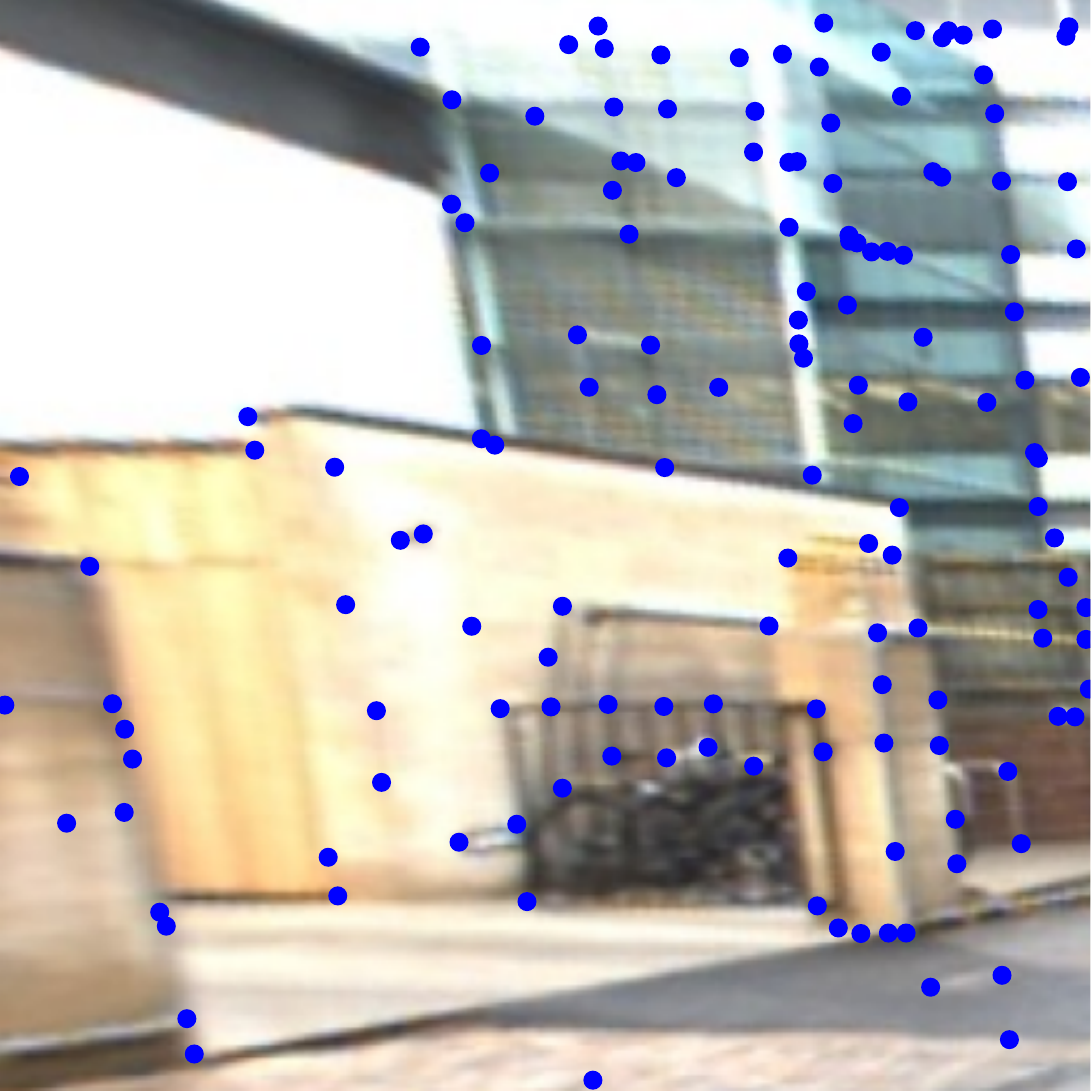}%
    \end{minipage}
    
    \vspace{1mm}
    \begin{minipage}{\iwidth\linewidth}
        \includegraphics[width=\linewidth]{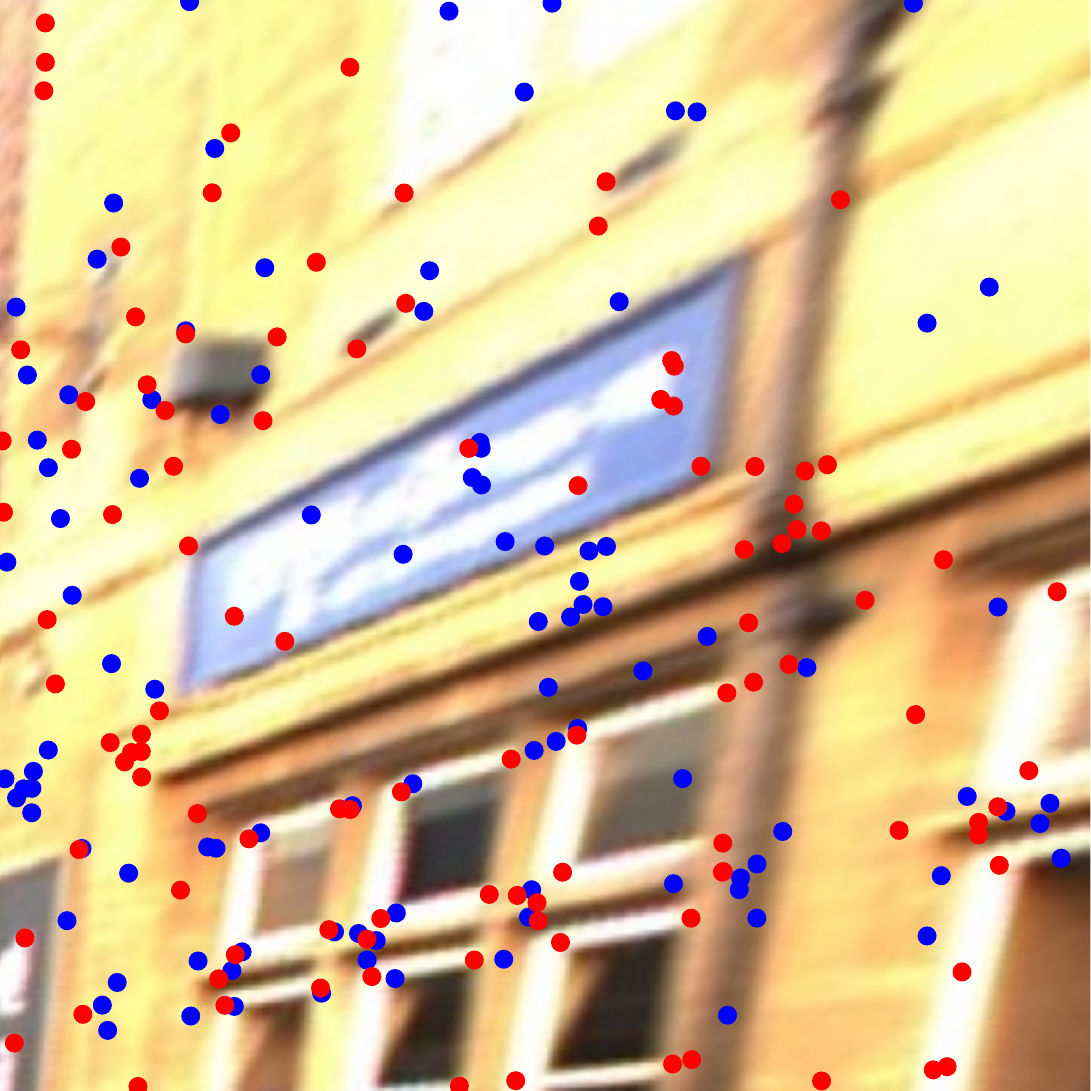}%
    \end{minipage}%
    \hspace{0.01mm}
    \begin{minipage}{\iwidth\linewidth}
        \includegraphics[width=\linewidth]{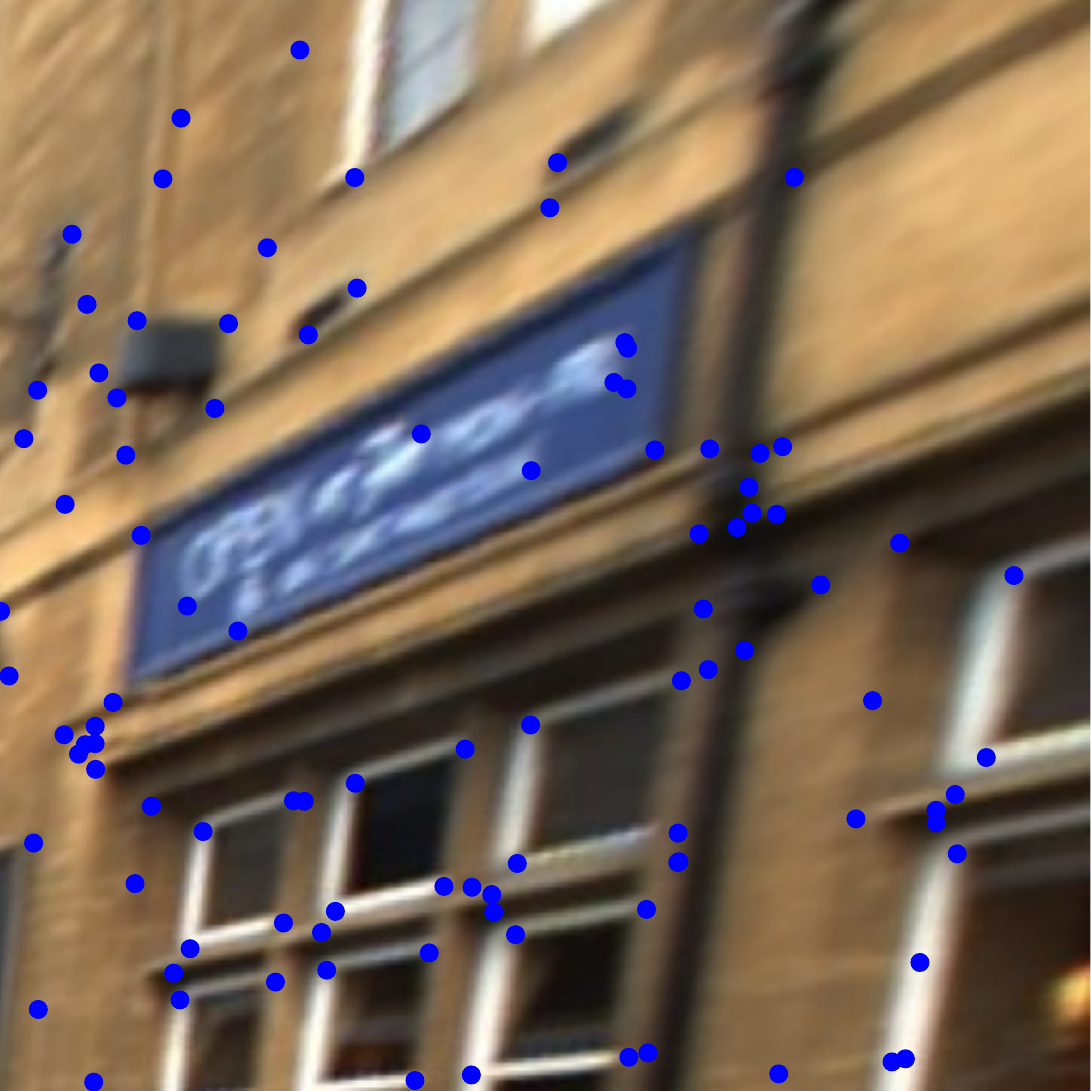}%
    \end{minipage}
    
    \vspace{1mm}
    \begin{minipage}{\iwidth\linewidth}
        \includegraphics[width=\linewidth]{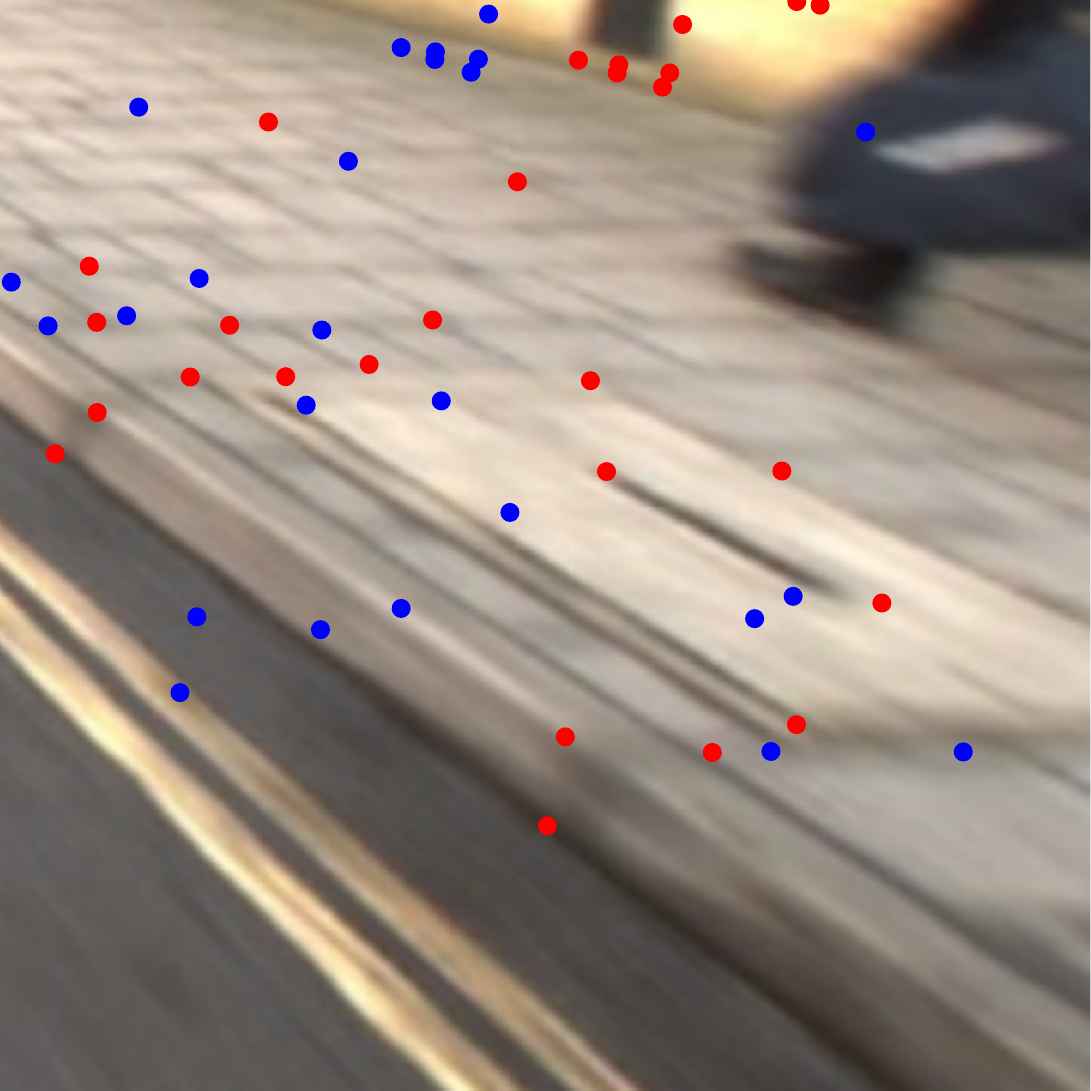}%
    \end{minipage}%
    \hspace{0.01mm}
    \begin{minipage}{\iwidth\linewidth}
        \includegraphics[width=\linewidth]{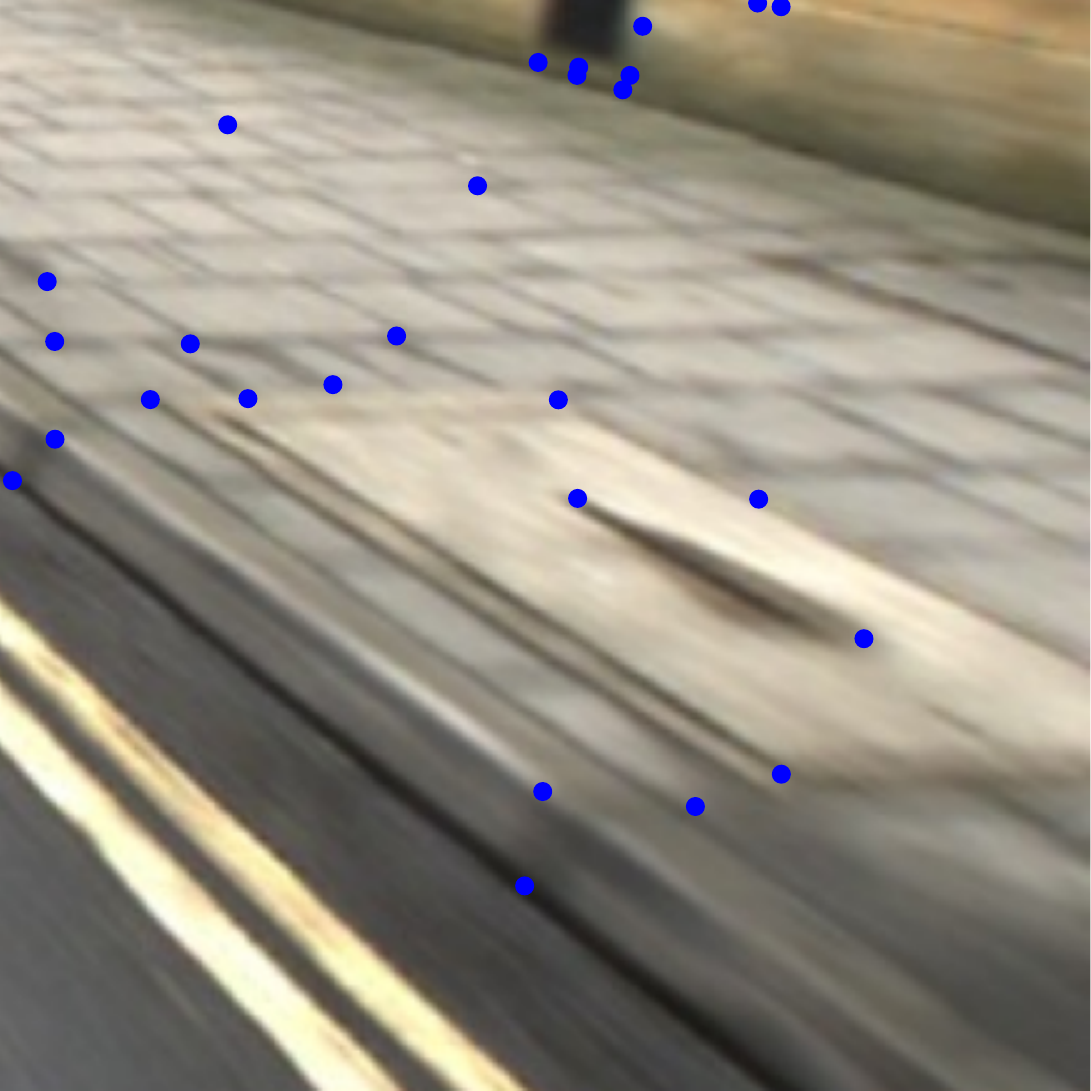}%
    \end{minipage}
    
    \vspace{1mm}
    \begin{minipage}{\iwidth\linewidth}
        \includegraphics[width=\linewidth]{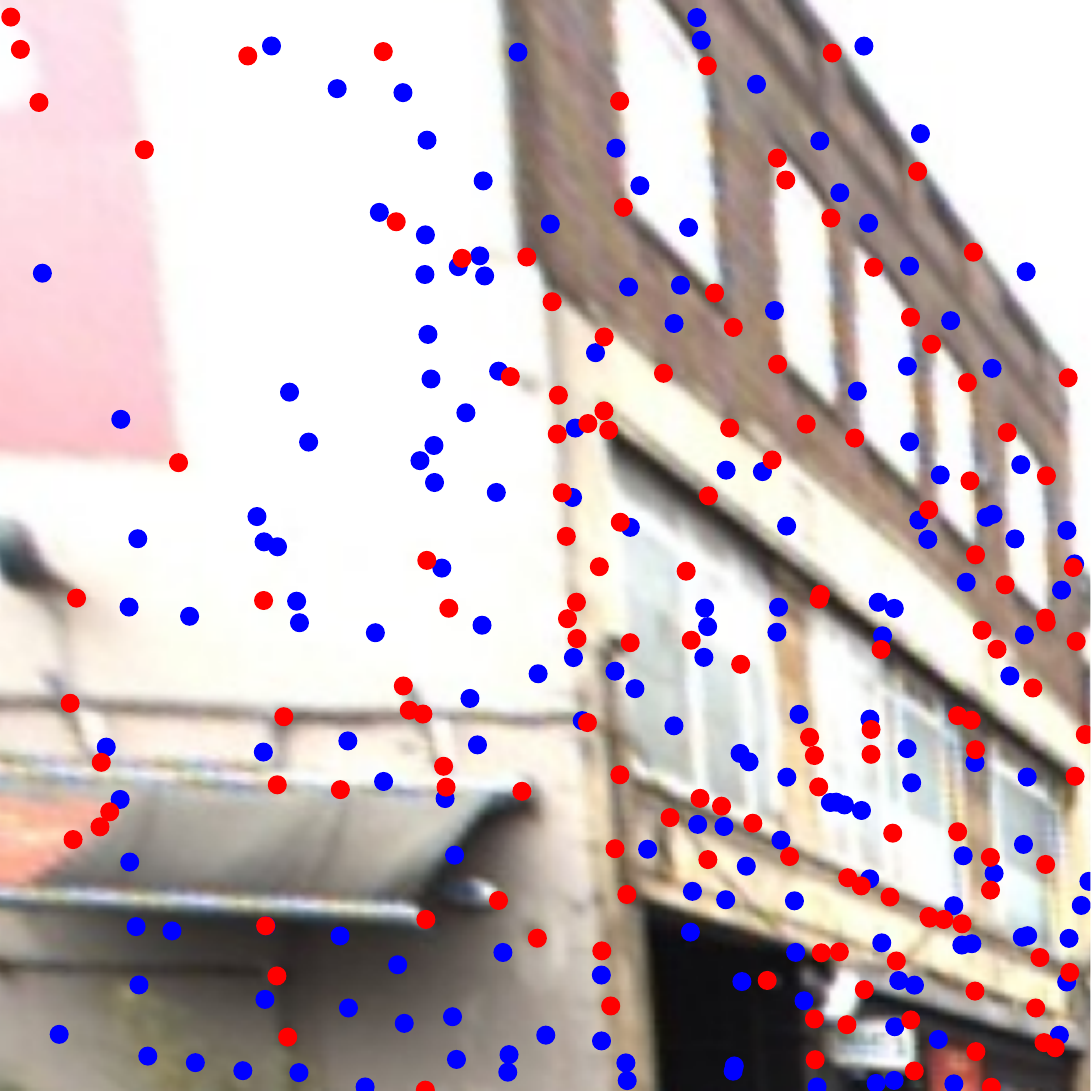}%
    \end{minipage}%
    \hspace{0.01mm}
    \begin{minipage}{\iwidth\linewidth}
        \includegraphics[width=\linewidth]{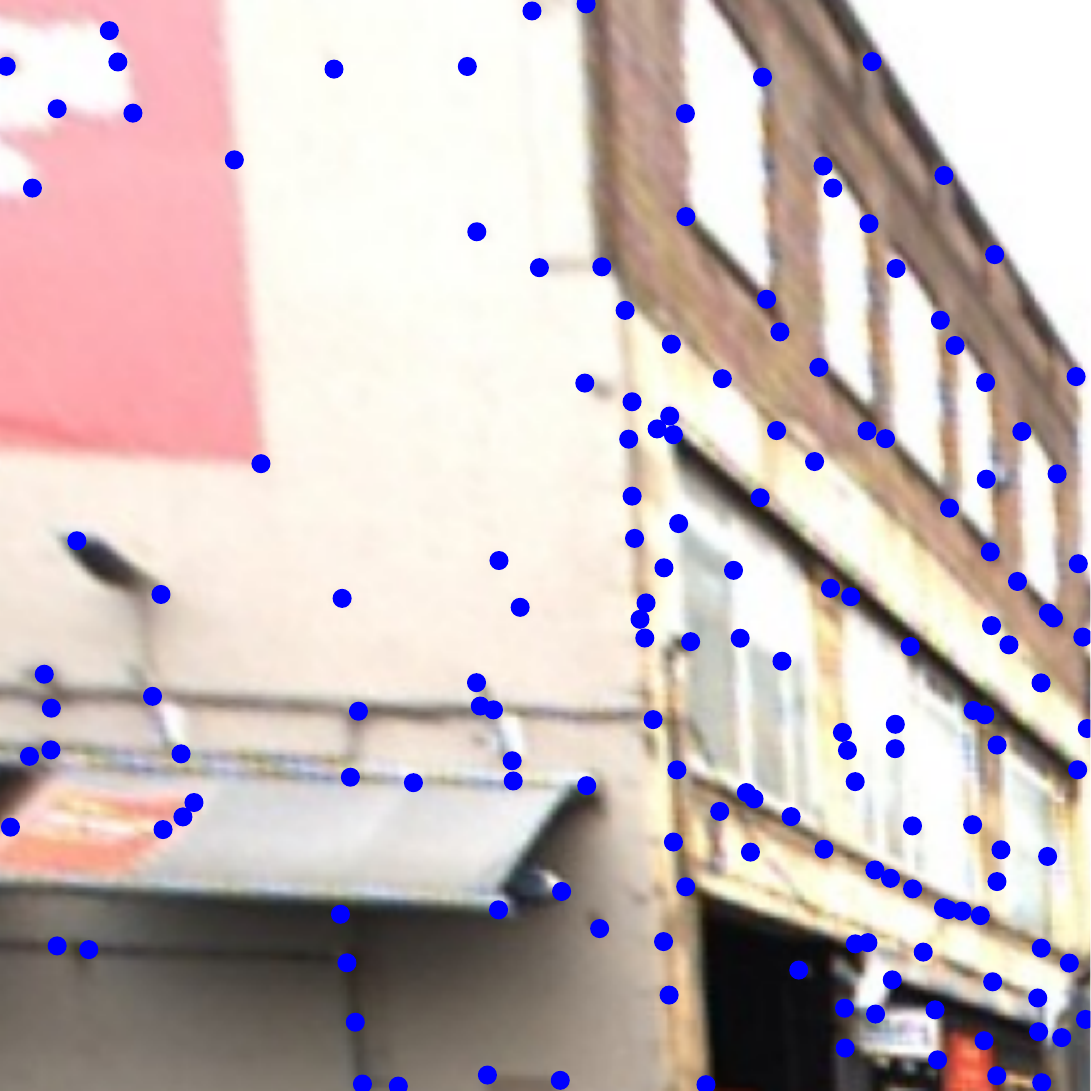}%
    \end{minipage}
    
    \vspace{1mm}
    \caption{\textbf{Inaccurate RobotCar ground truth poses.} We plot the projection of 3D SfM points in the query images according to the ground truth (in \blue{blue}) and hloc (in \red{red}) poses. We project the same points in the reference images using the reference poses (in \blue{blue}). Query points using hloc are better aligned to the reference points, indicating that the ground truth query poses are inaccurate.
    }
    \label{fig:robotcar_gt}%
\end{figure}

\begin{figure*}[t]
    \centering
\def\iwidth{0.14}
\def\pwidth{0.99}
\def\lwidth{0.020}
\def\rcwidth{0.14}

\begin{minipage}{\lwidth\textwidth}
\hfill
\end{minipage}%
\begin{minipage}{\iwidth\textwidth}
    \centering
    \small{Images}
\end{minipage}%
\begin{minipage}{\iwidth\textwidth*\real{3.0}}
    \centering
    \small{Features}
    \vspace{0.5mm}
    \hrule width 0.99\linewidth
    \vspace{0.2mm}
\end{minipage}%
\begin{minipage}{\iwidth\textwidth*\real{3.0}}
    \centering
    \small{Confidence}
    \vspace{0.5mm}
    \hrule width 0.99\linewidth
    \vspace{0.2mm}
\end{minipage}%

\begin{minipage}{\lwidth\textwidth}
\rotatebox[origin=c]{90}{Query}
\end{minipage}%
\begin{minipage}{\iwidth\textwidth}
    \centering
    \includegraphics[width=\pwidth\linewidth]{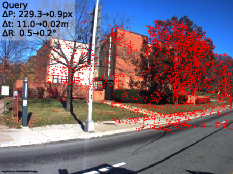}
\end{minipage}%
\begin{minipage}{\iwidth\textwidth}
    \centering
    \includegraphics[width=\pwidth\linewidth]{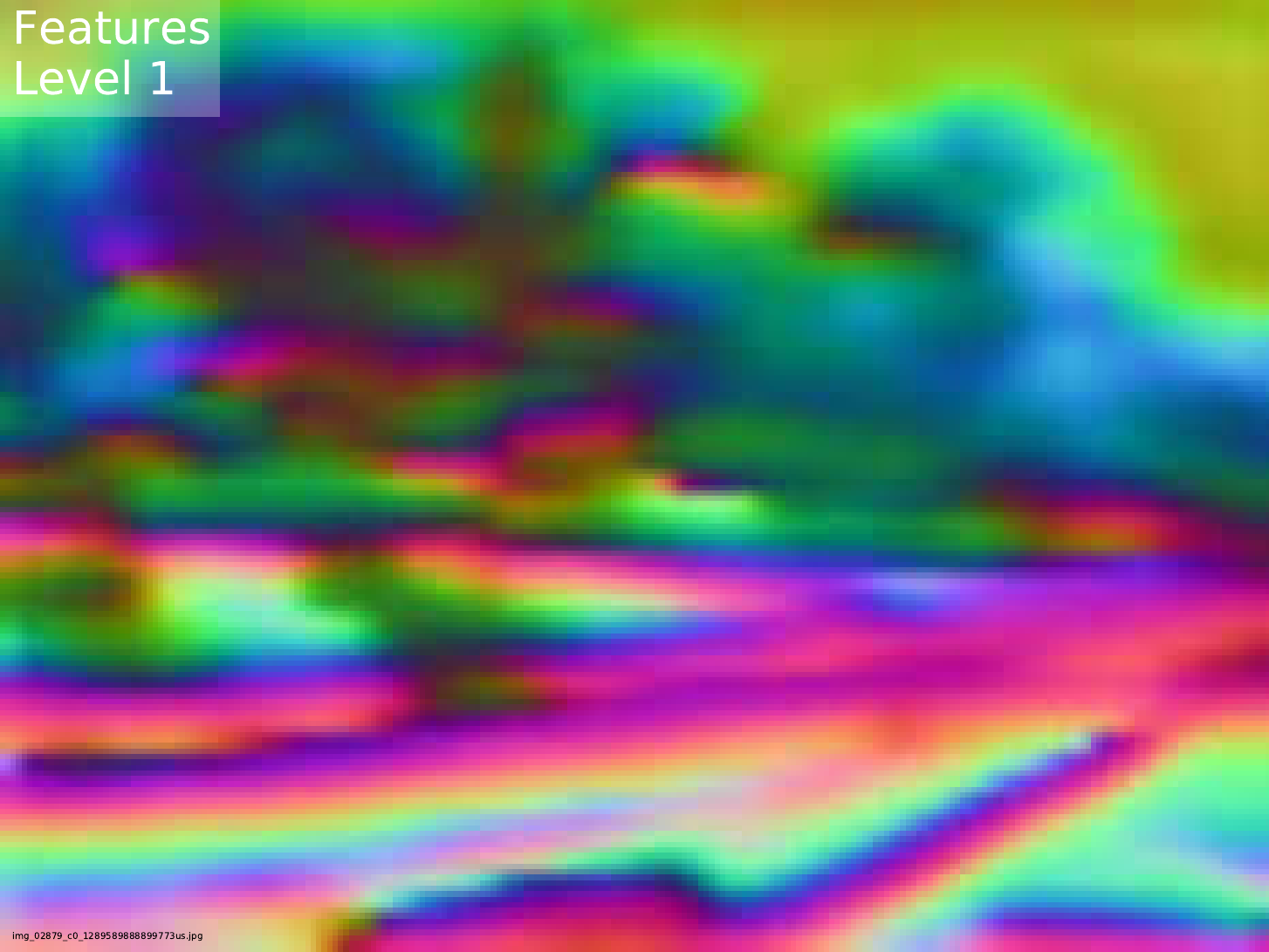}
\end{minipage}%
\begin{minipage}{\iwidth\textwidth}
    \centering
    \includegraphics[width=\pwidth\linewidth]{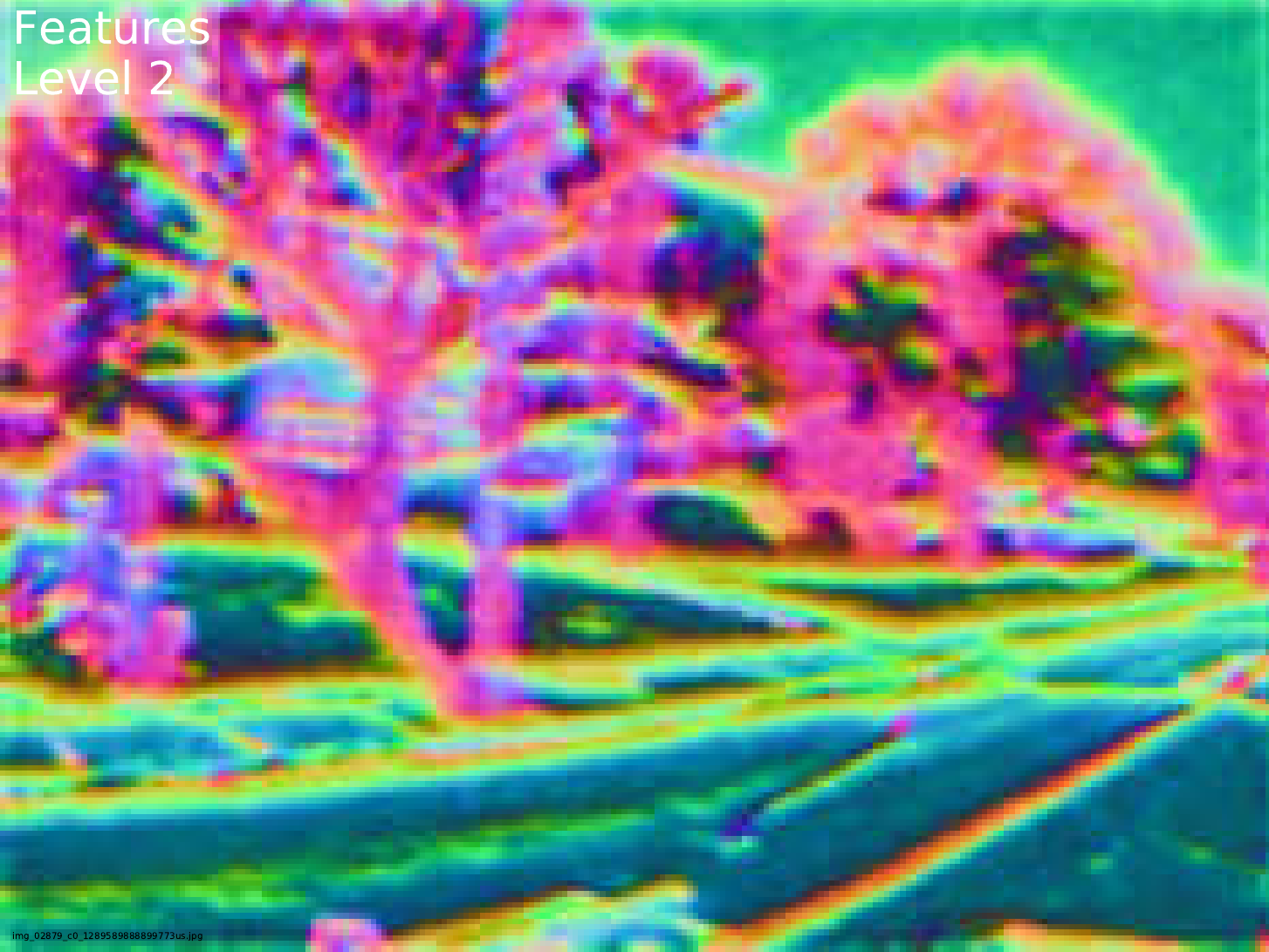}
\end{minipage}%
\begin{minipage}{\iwidth\textwidth}
    \centering
    \includegraphics[width=\pwidth\linewidth]{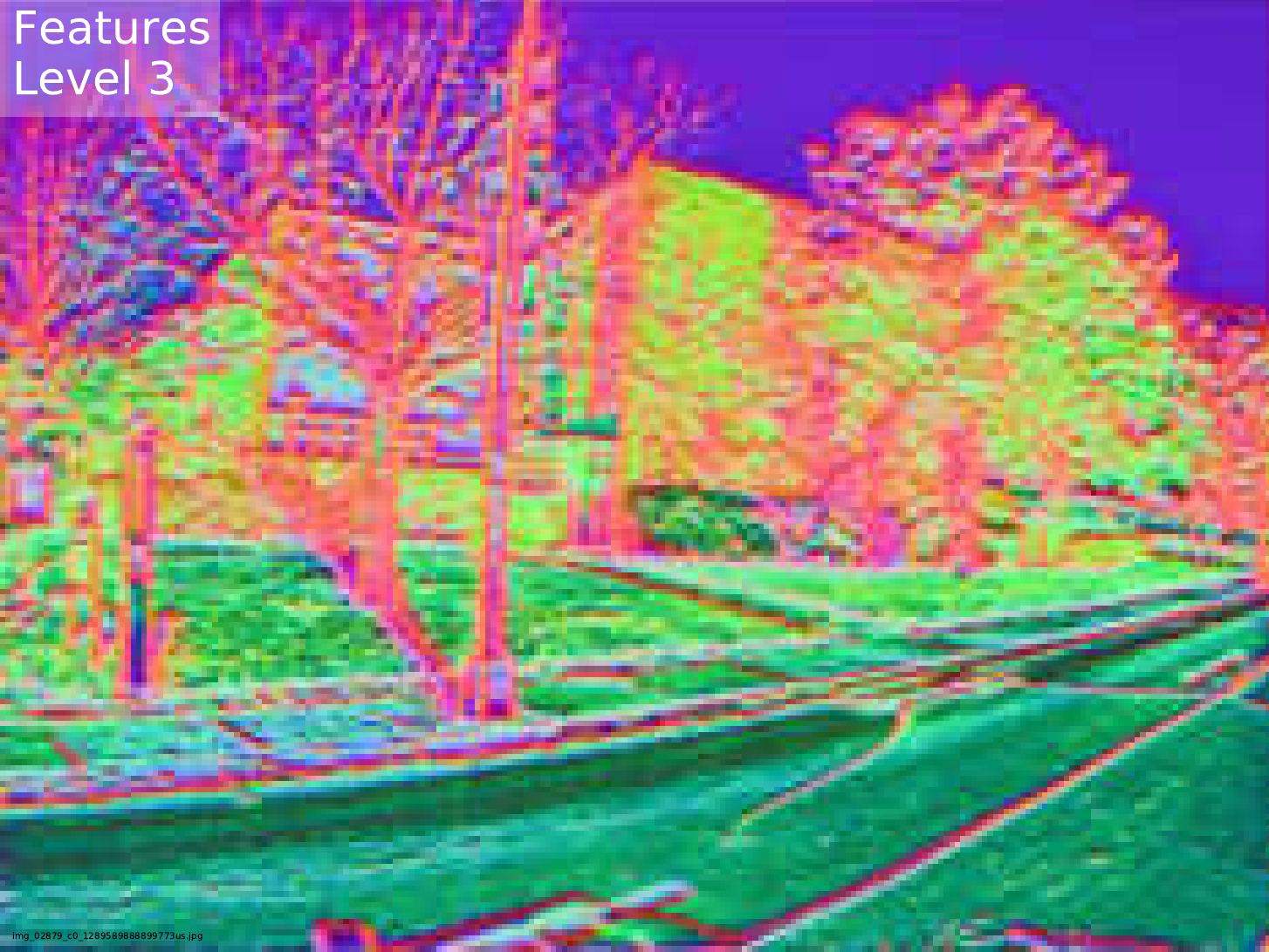}
\end{minipage}%
\begin{minipage}{\iwidth\textwidth}
    \centering
    \includegraphics[width=\pwidth\linewidth]{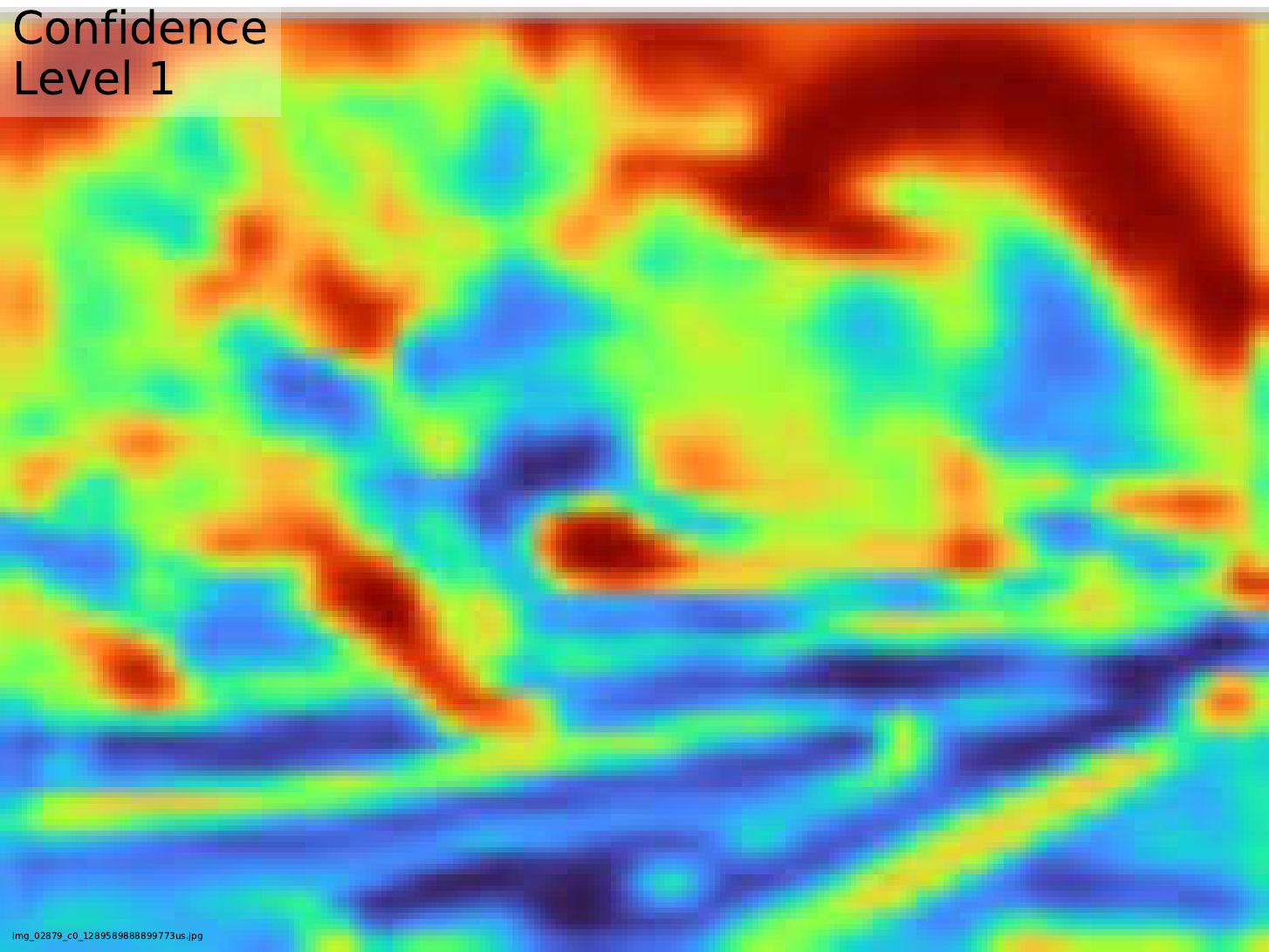}
\end{minipage}%
\begin{minipage}{\iwidth\textwidth}
    \centering
    \includegraphics[width=\pwidth\linewidth]{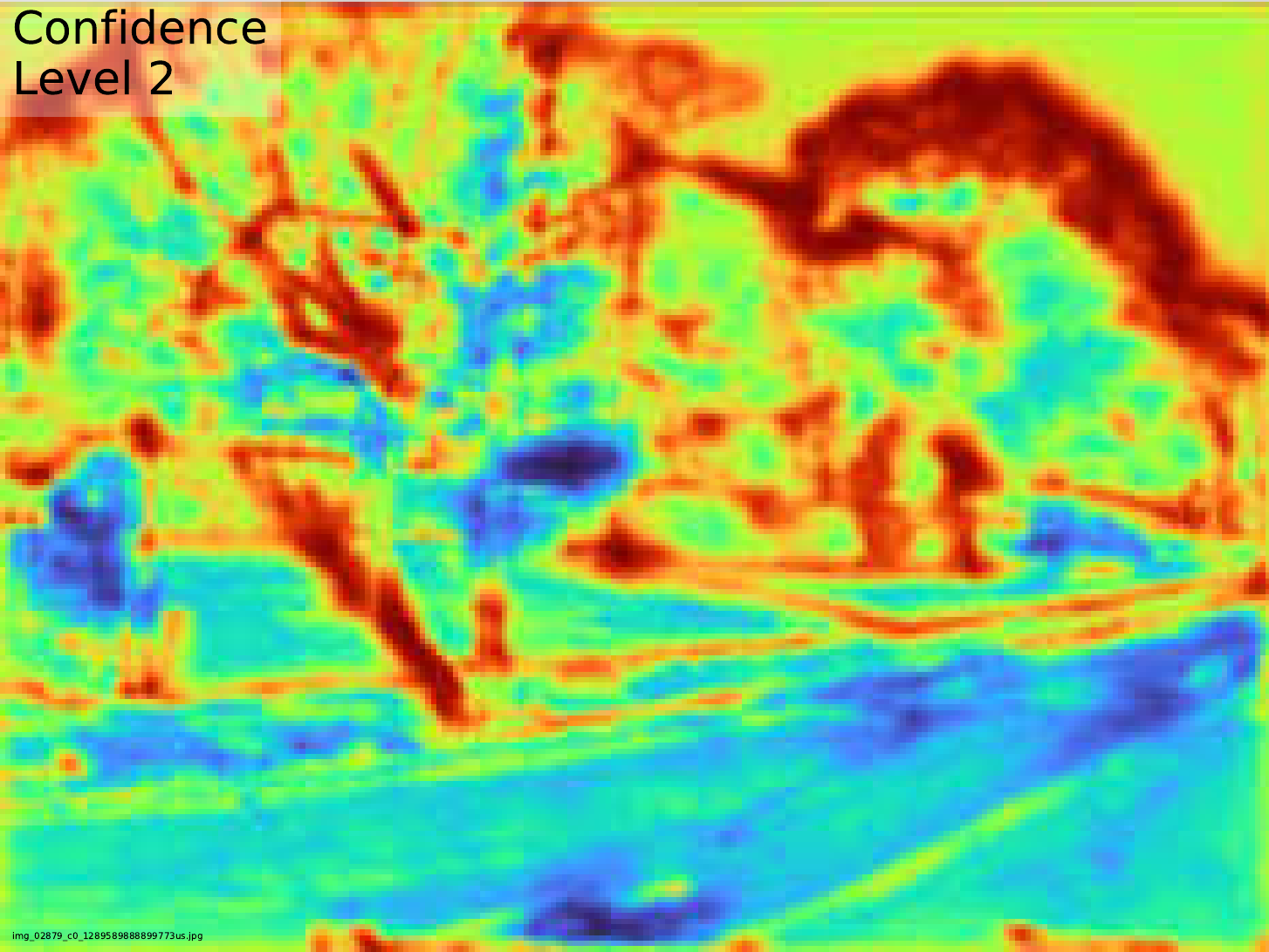}
\end{minipage}%
\begin{minipage}{\iwidth\textwidth}
    \centering
    \includegraphics[width=\pwidth\linewidth]{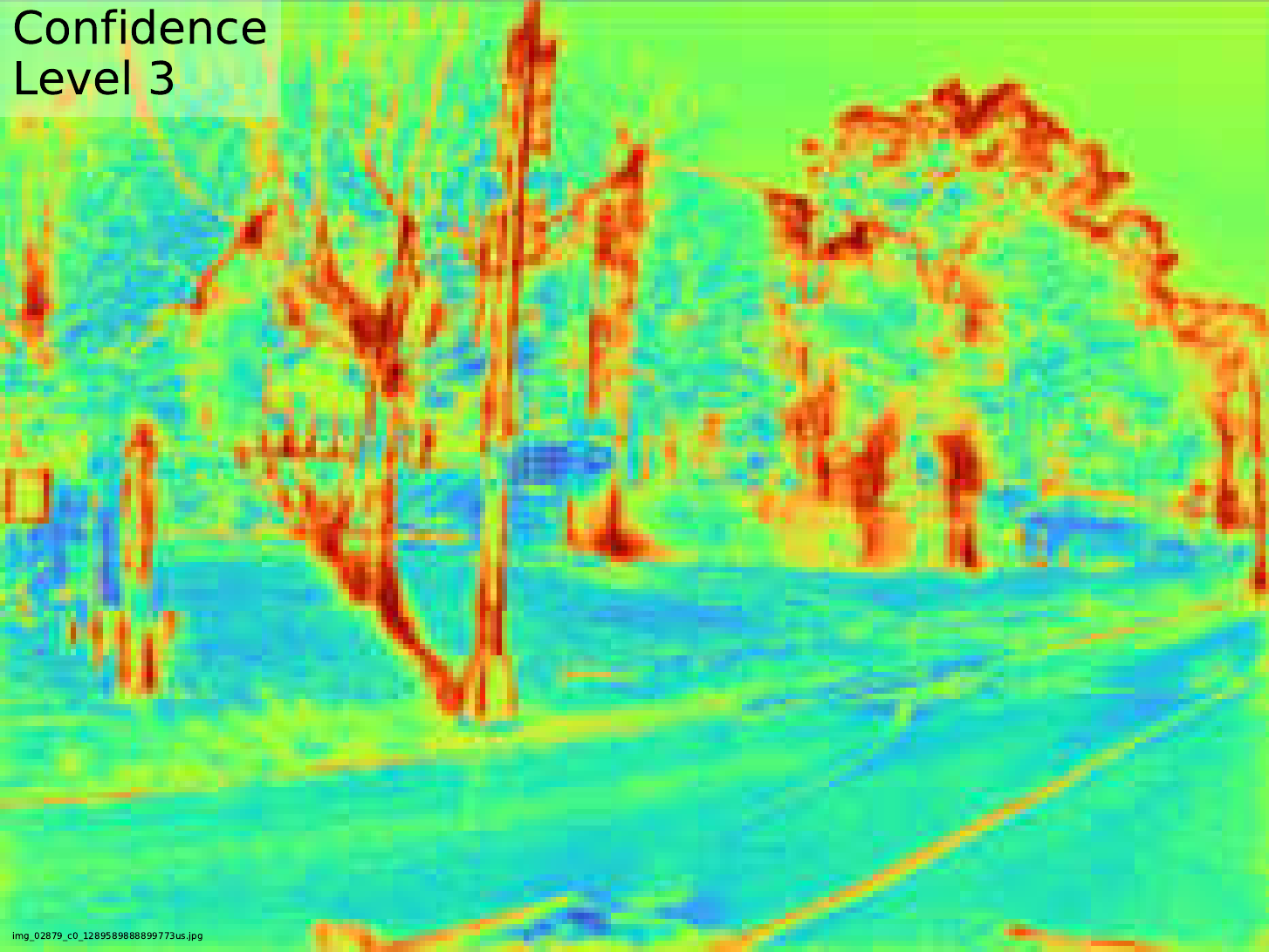}
\end{minipage}
\begin{minipage}{\lwidth\textwidth}
\rotatebox[origin=c]{90}{Reference}
\end{minipage}%
\begin{minipage}{\iwidth\textwidth}
    \centering
    \includegraphics[width=\pwidth\linewidth]{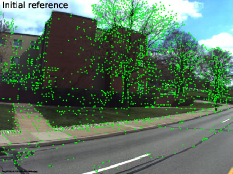}
\end{minipage}%
\begin{minipage}{\iwidth\textwidth}
    \centering
    \includegraphics[width=\pwidth\linewidth]{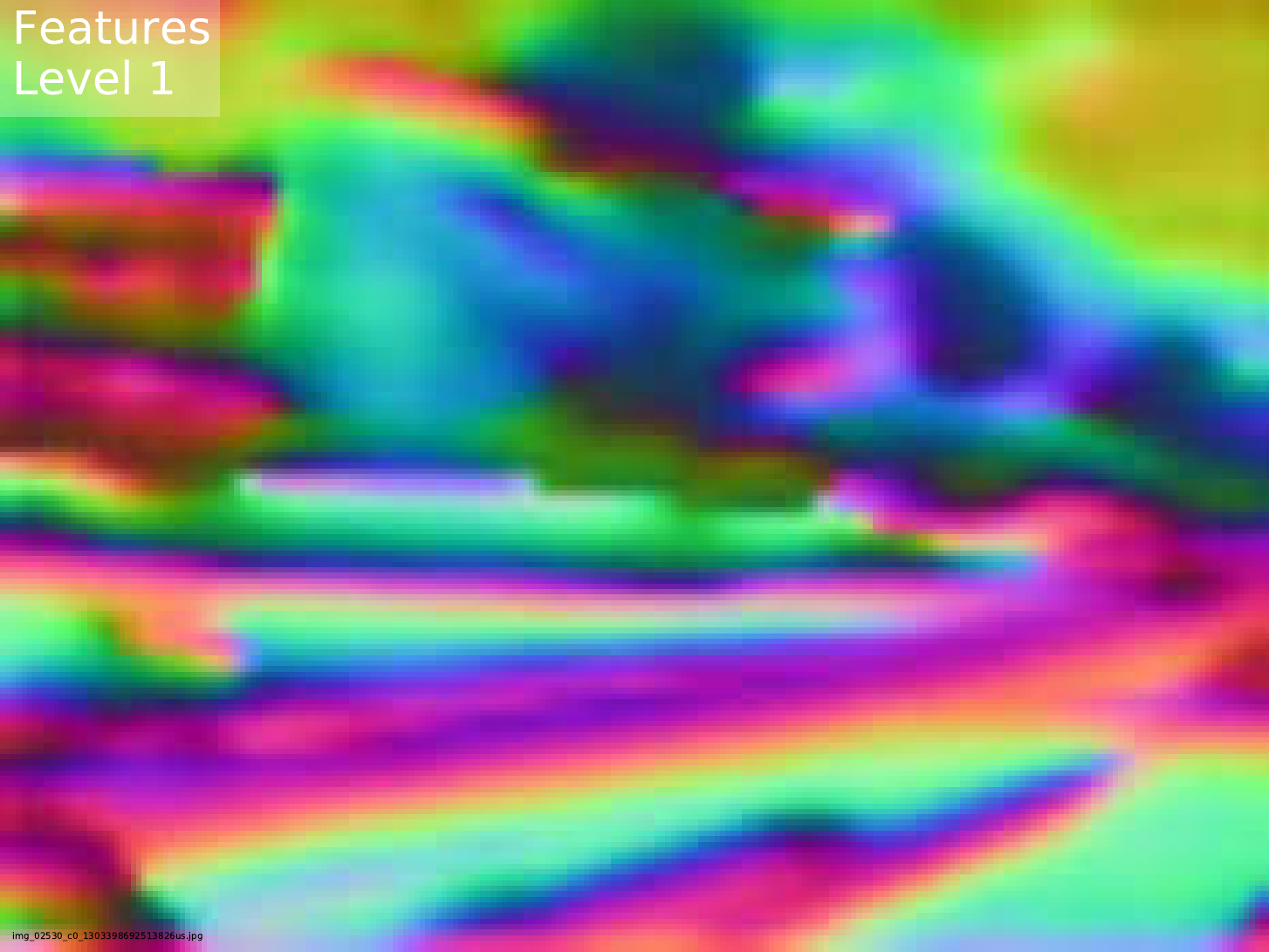}
\end{minipage}%
\begin{minipage}{\iwidth\textwidth}
    \centering
    \includegraphics[width=\pwidth\linewidth]{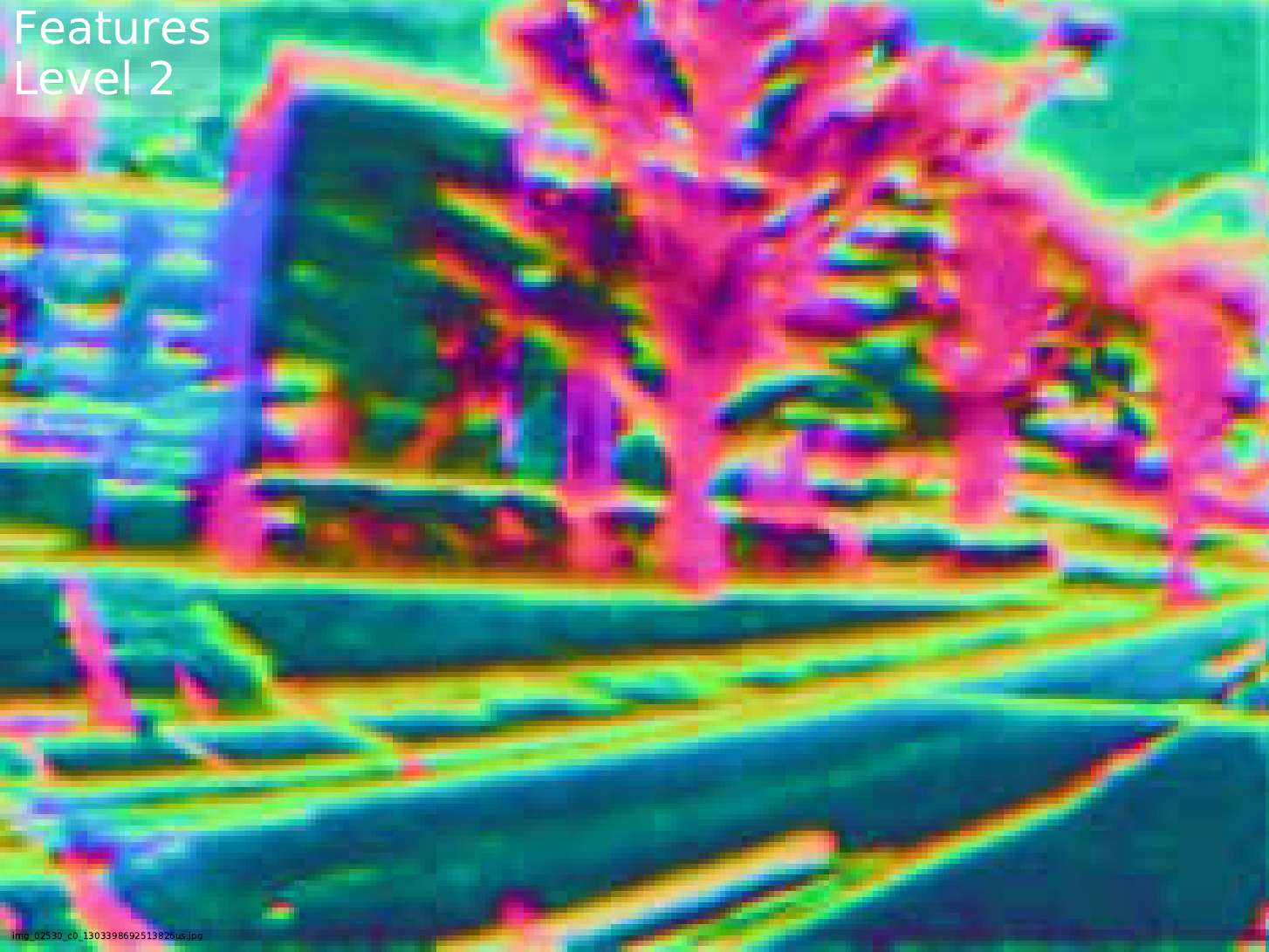}
\end{minipage}%
\begin{minipage}{\iwidth\textwidth}
    \centering
    \includegraphics[width=\pwidth\linewidth]{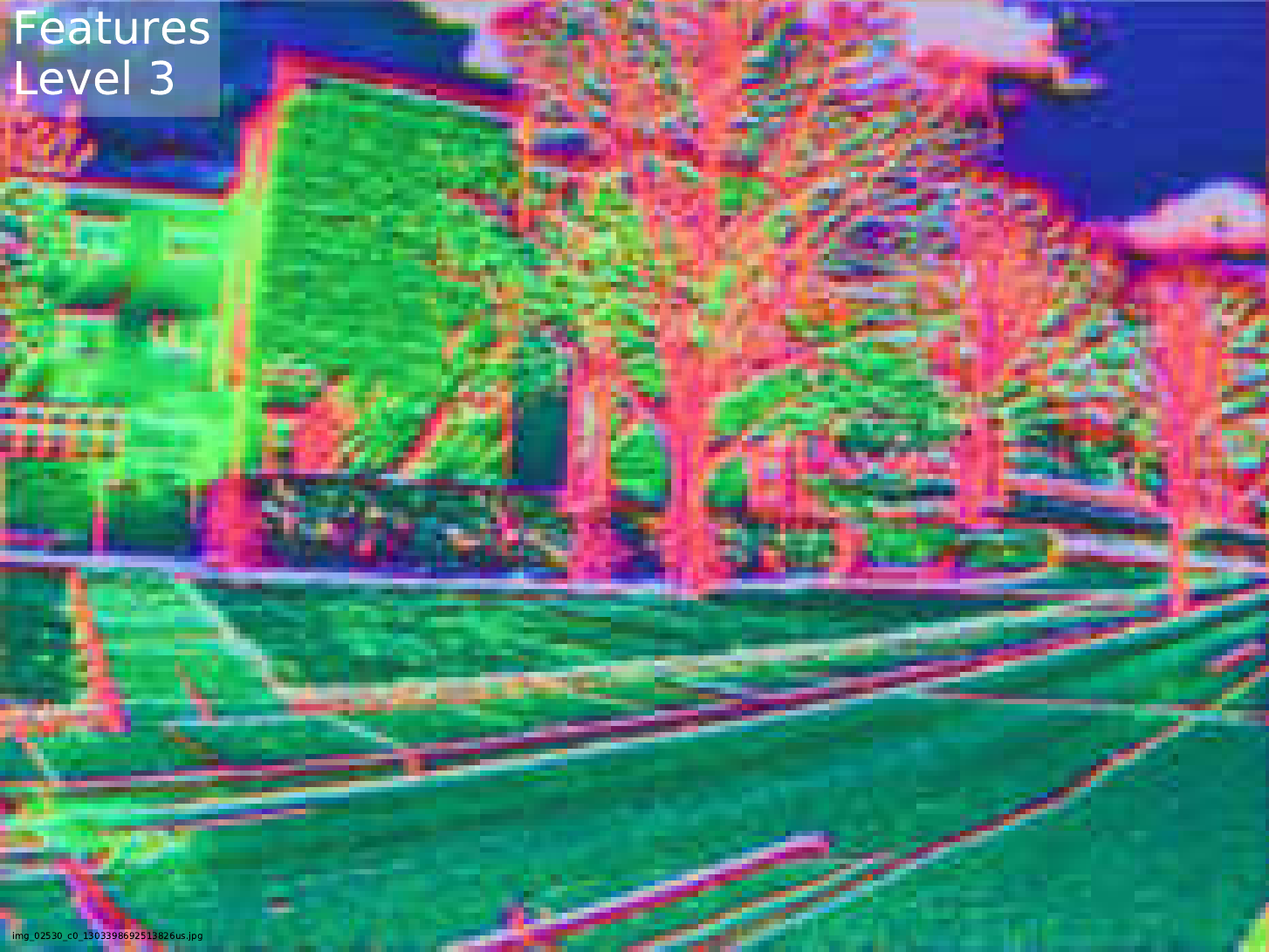}
\end{minipage}%
\begin{minipage}{\iwidth\textwidth}
    \centering
    \includegraphics[width=\pwidth\linewidth]{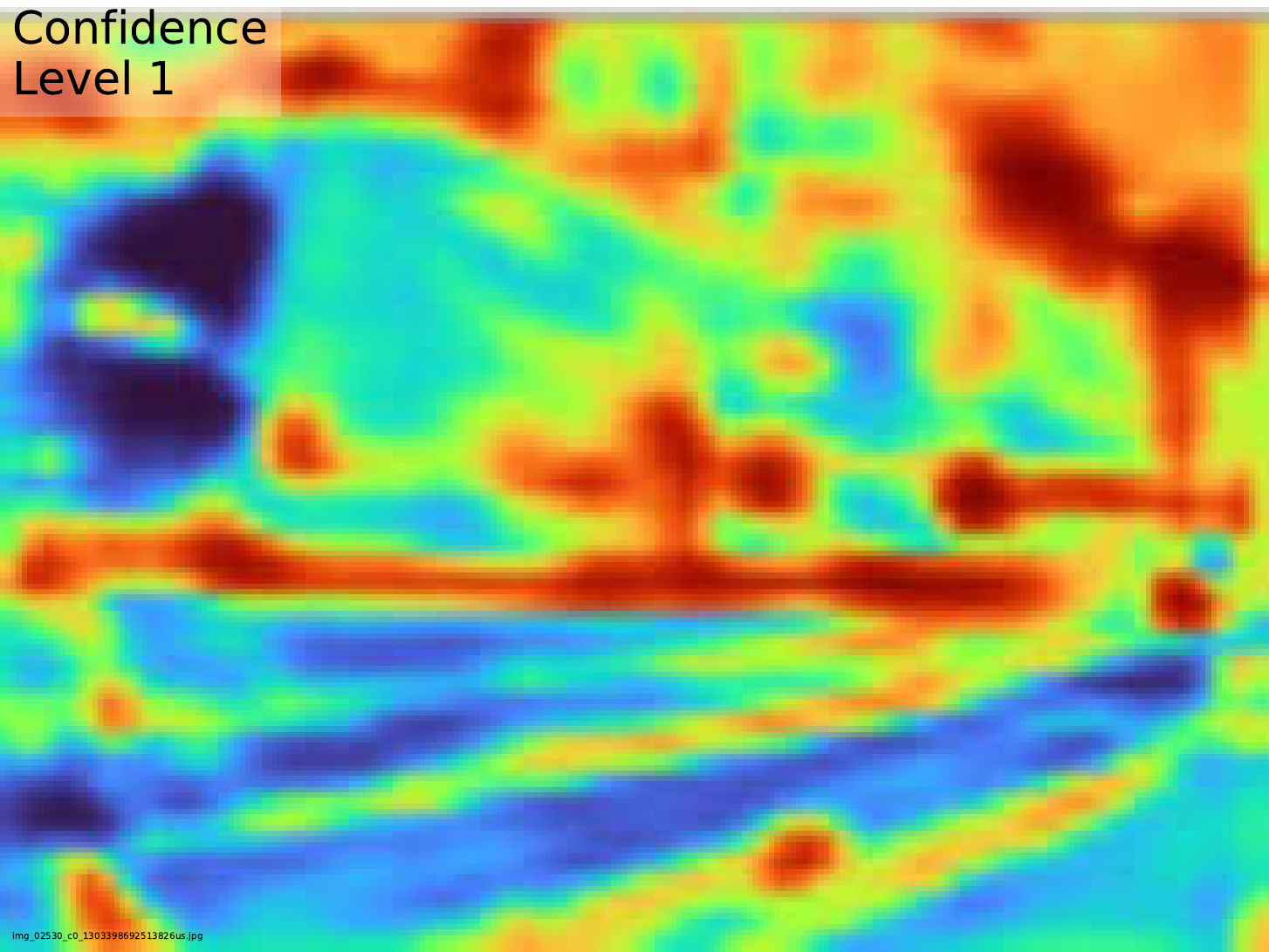}
\end{minipage}%
\begin{minipage}{\iwidth\textwidth}
    \centering
    \includegraphics[width=\pwidth\linewidth]{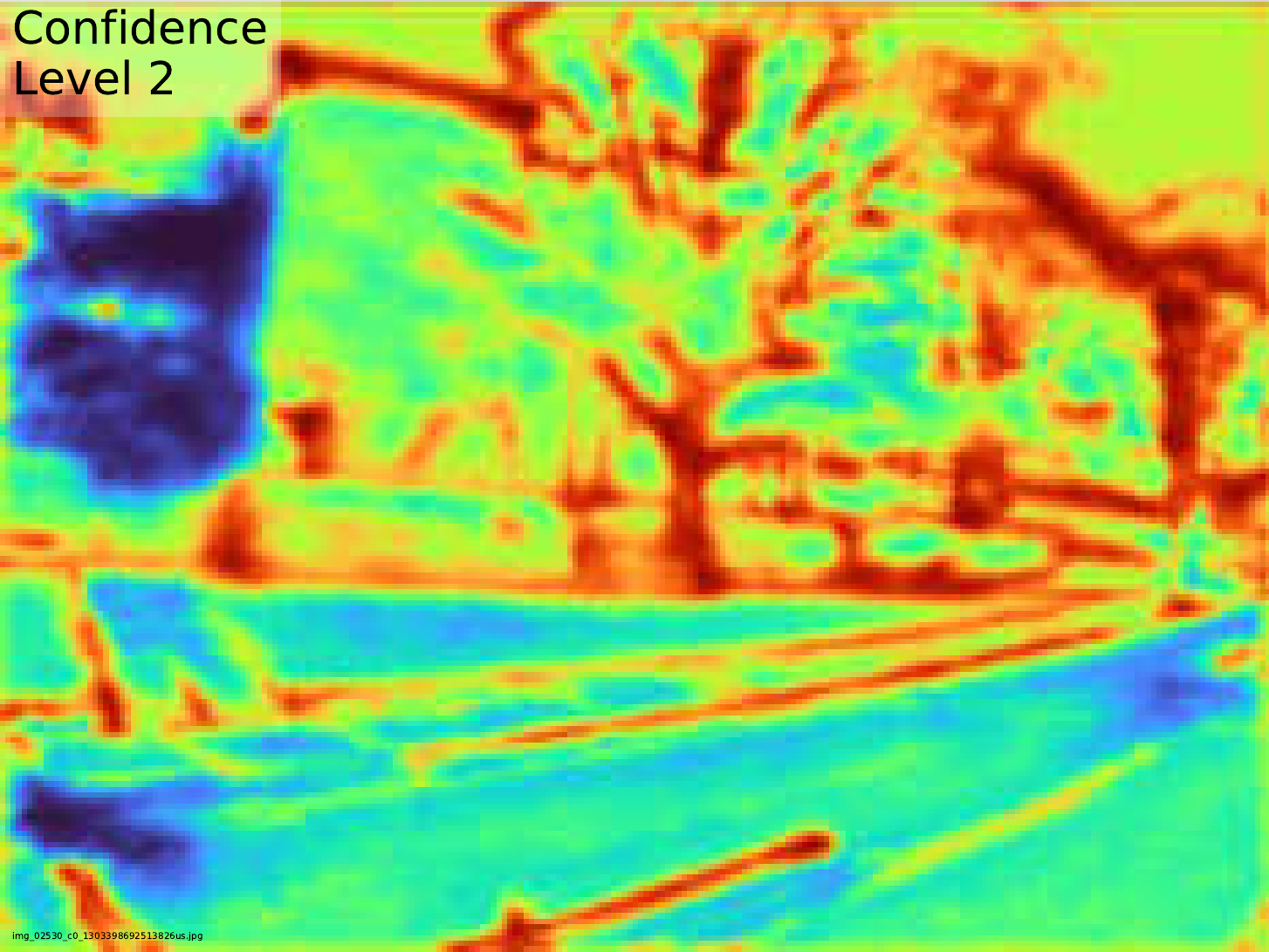}
\end{minipage}%
\begin{minipage}{\iwidth\textwidth}
    \centering
    \includegraphics[width=\pwidth\linewidth]{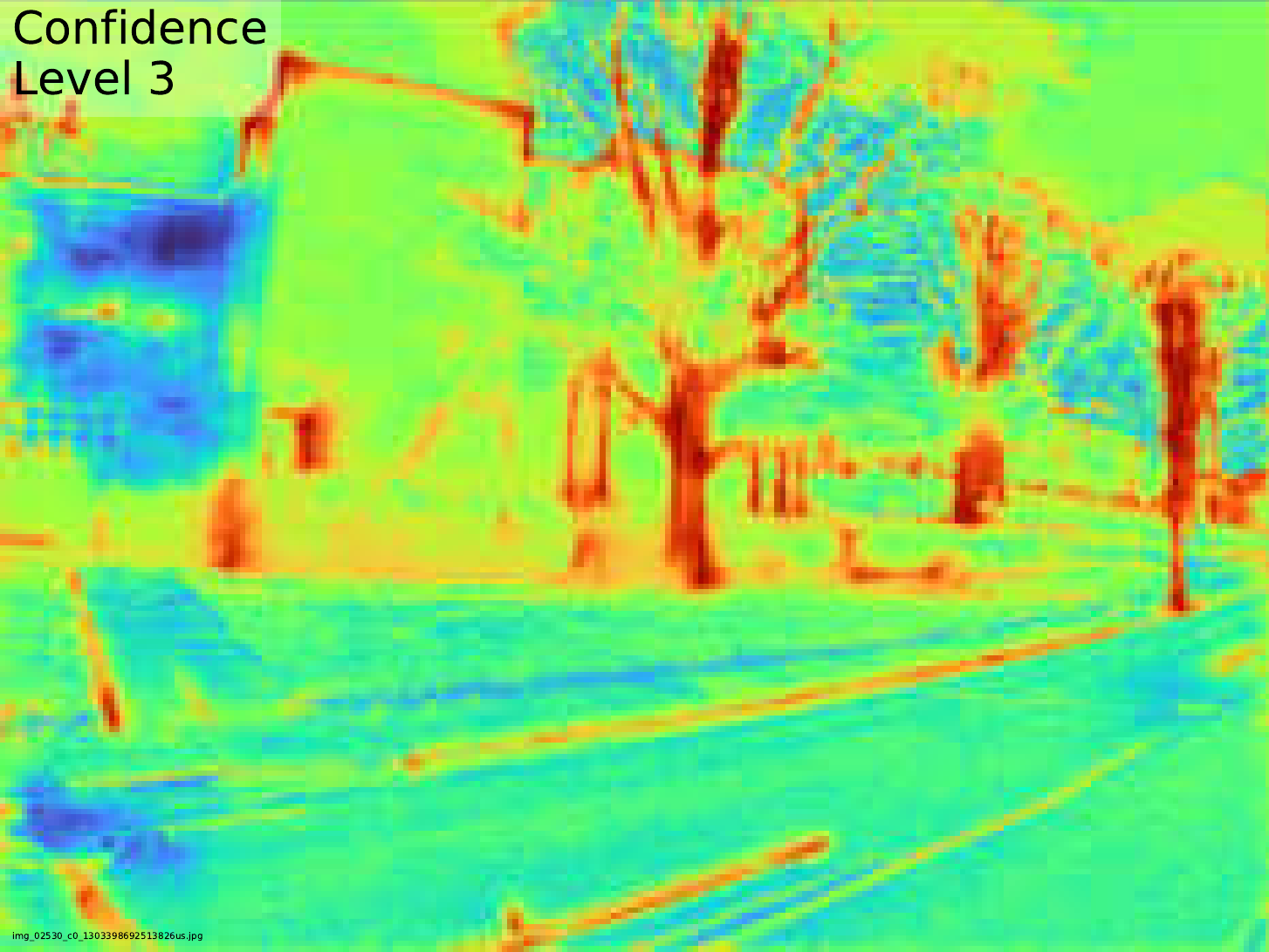}
\end{minipage}
\vspace{2mm}

\begin{minipage}{\lwidth\textwidth}
\rotatebox[origin=c]{90}{Query}
\end{minipage}%
\begin{minipage}{\iwidth\textwidth}
    \centering
    \includegraphics[width=\pwidth\linewidth]{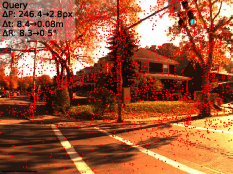}
\end{minipage}%
\begin{minipage}{\iwidth\textwidth}
    \centering
    \includegraphics[width=\pwidth\linewidth]{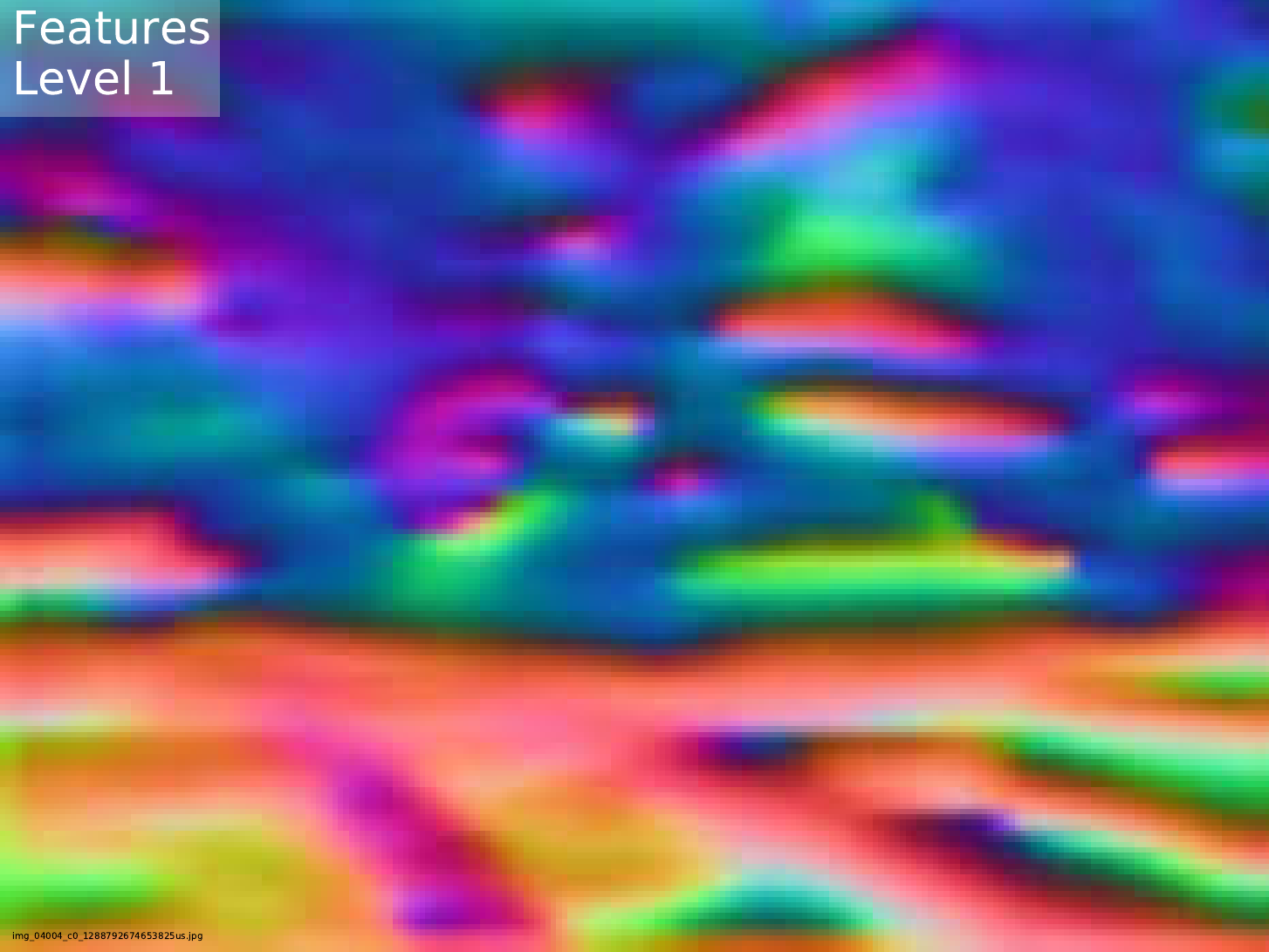}
\end{minipage}%
\begin{minipage}{\iwidth\textwidth}
    \centering
    \includegraphics[width=\pwidth\linewidth]{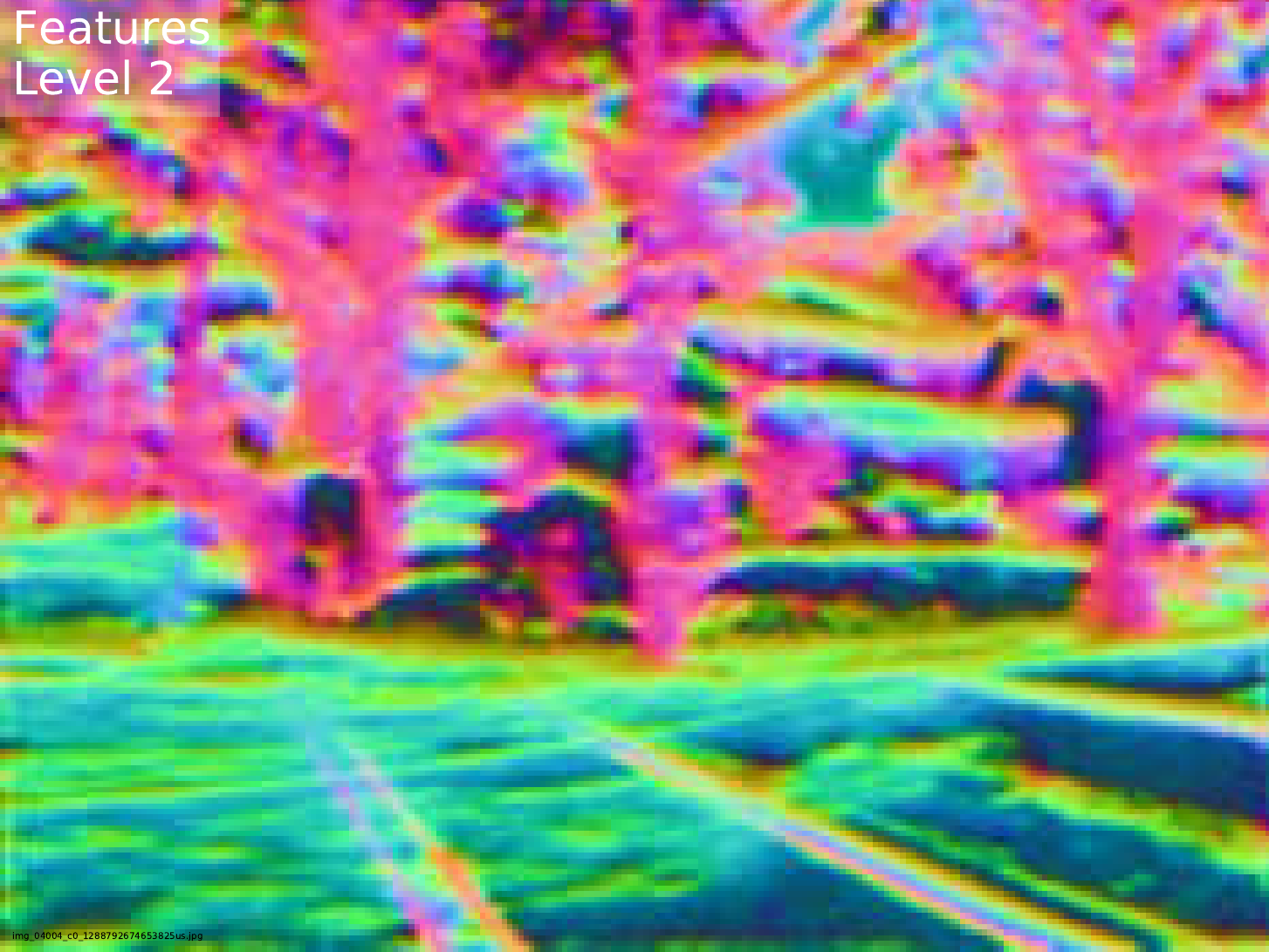}
\end{minipage}%
\begin{minipage}{\iwidth\textwidth}
    \centering
    \includegraphics[width=\pwidth\linewidth]{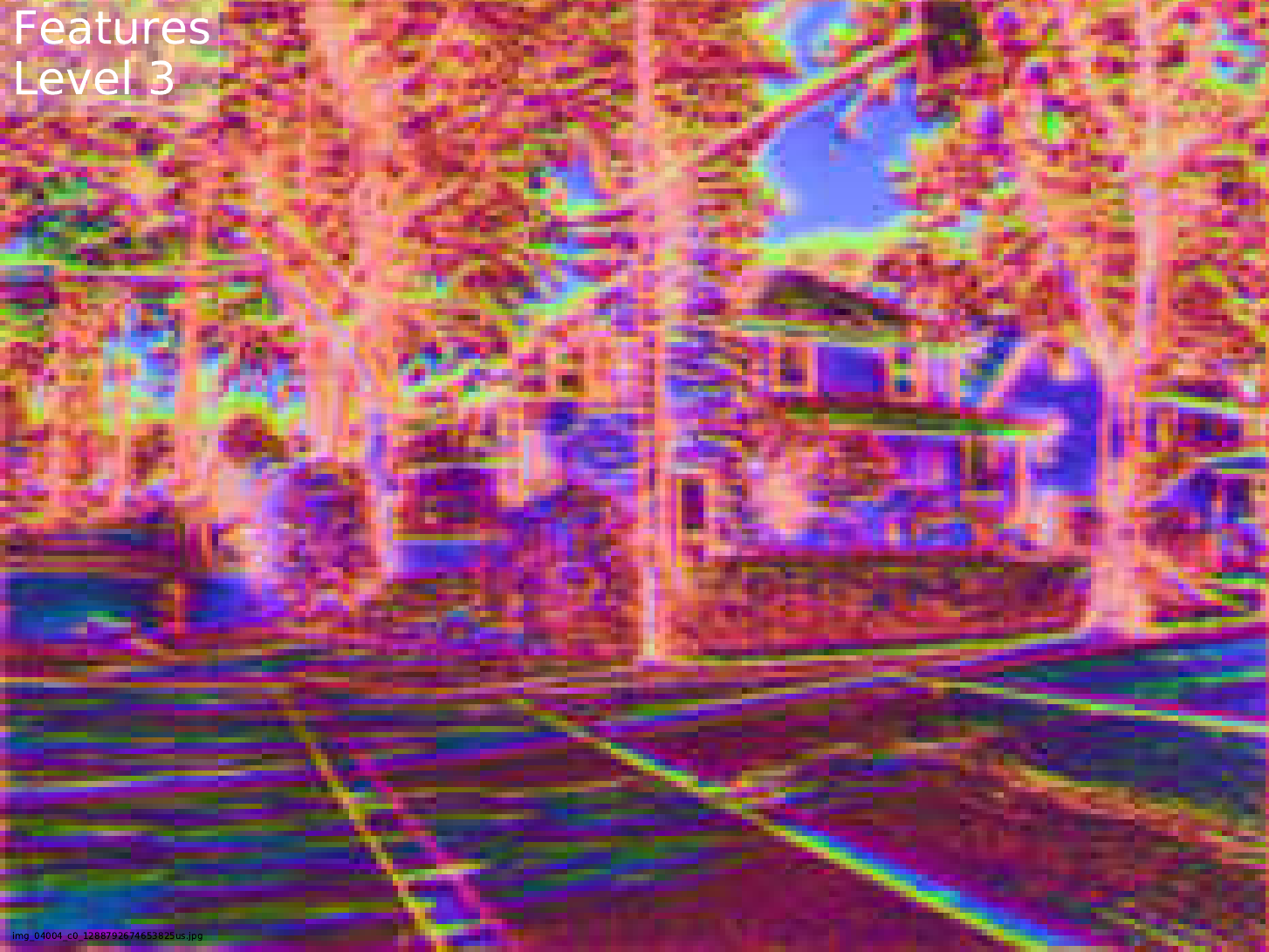}
\end{minipage}%
\begin{minipage}{\iwidth\textwidth}
    \centering
    \includegraphics[width=\pwidth\linewidth]{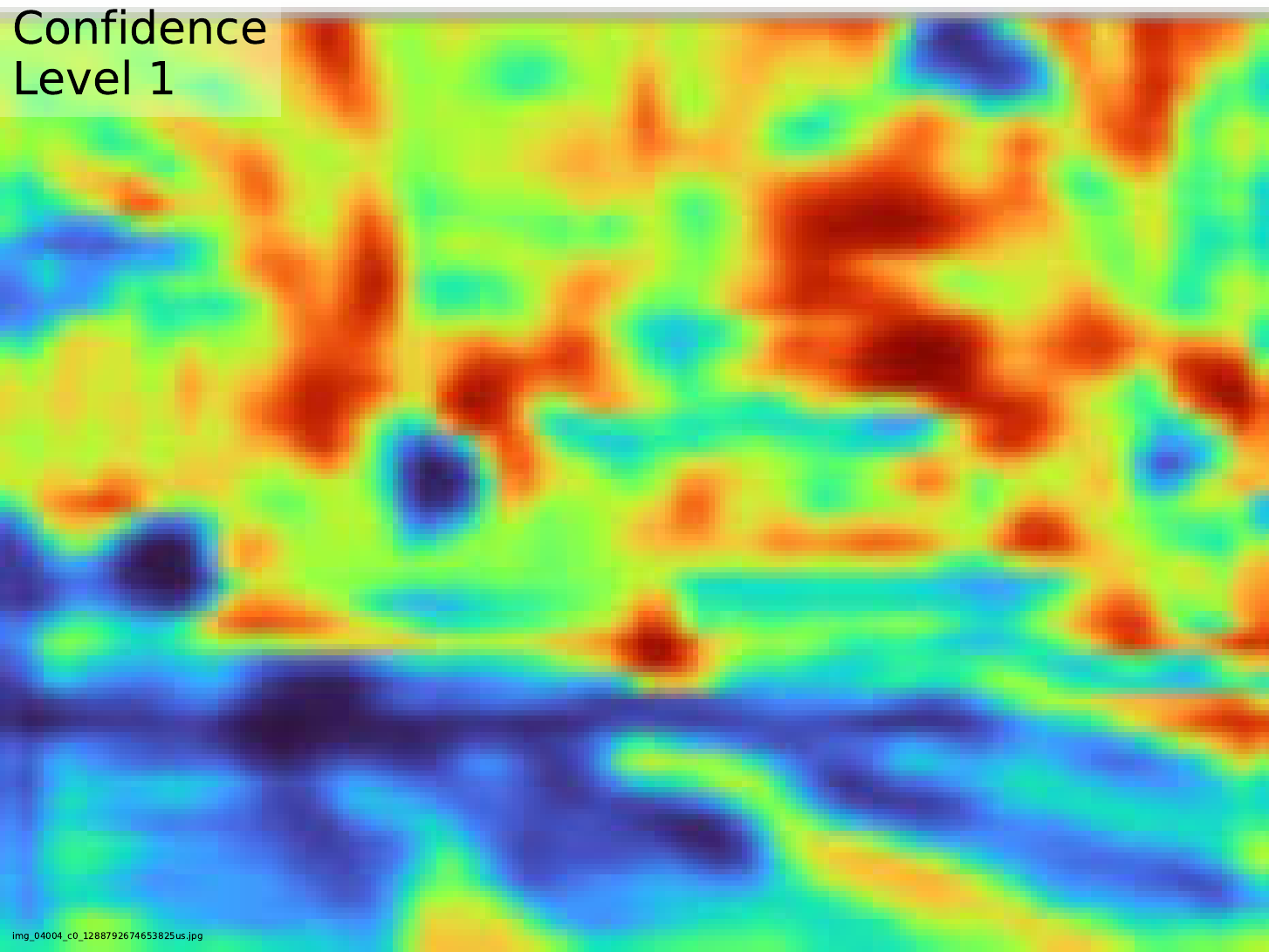}
\end{minipage}%
\begin{minipage}{\iwidth\textwidth}
    \centering
    \includegraphics[width=\pwidth\linewidth]{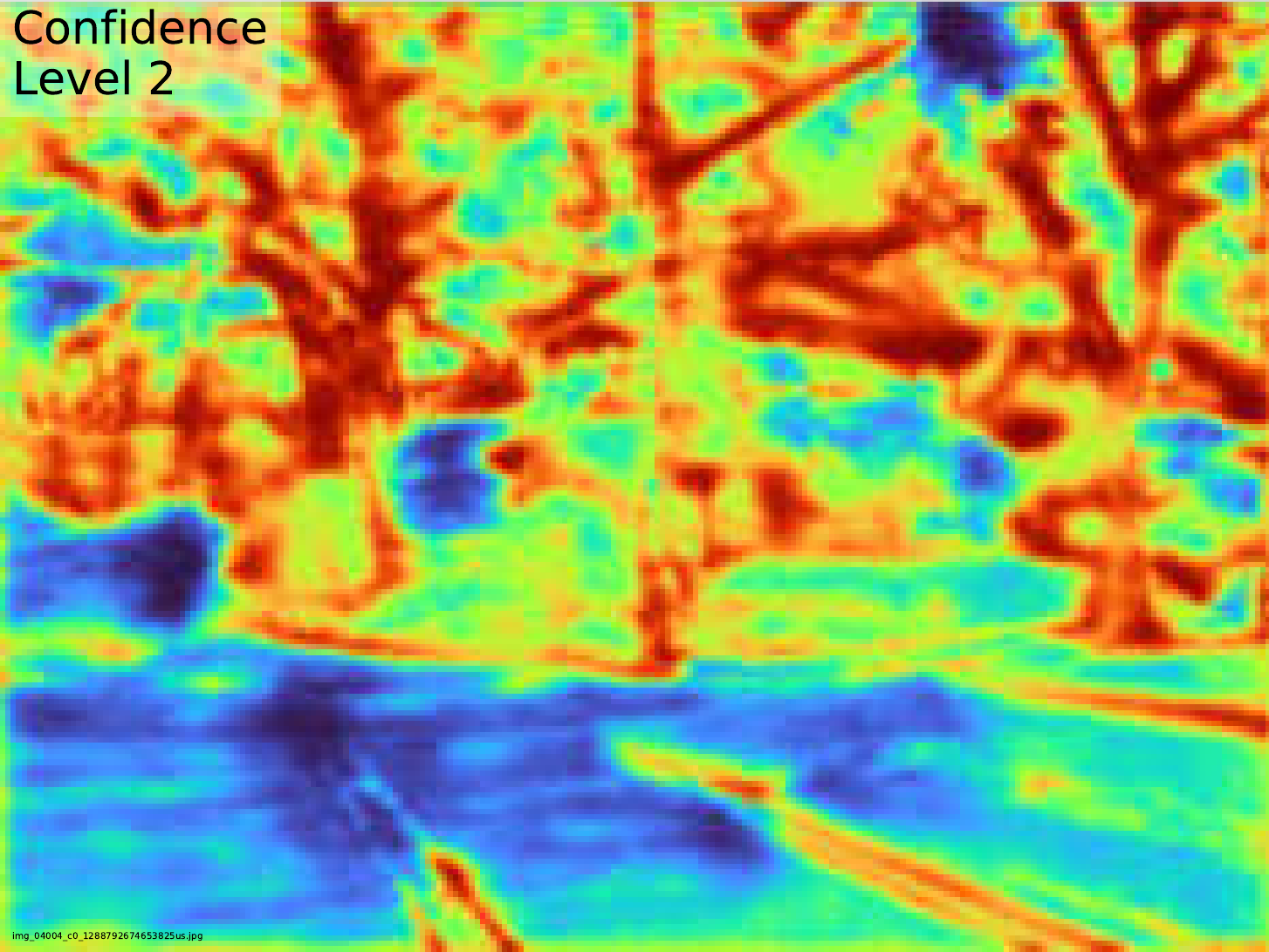}
\end{minipage}%
\begin{minipage}{\iwidth\textwidth}
    \centering
    \includegraphics[width=\pwidth\linewidth]{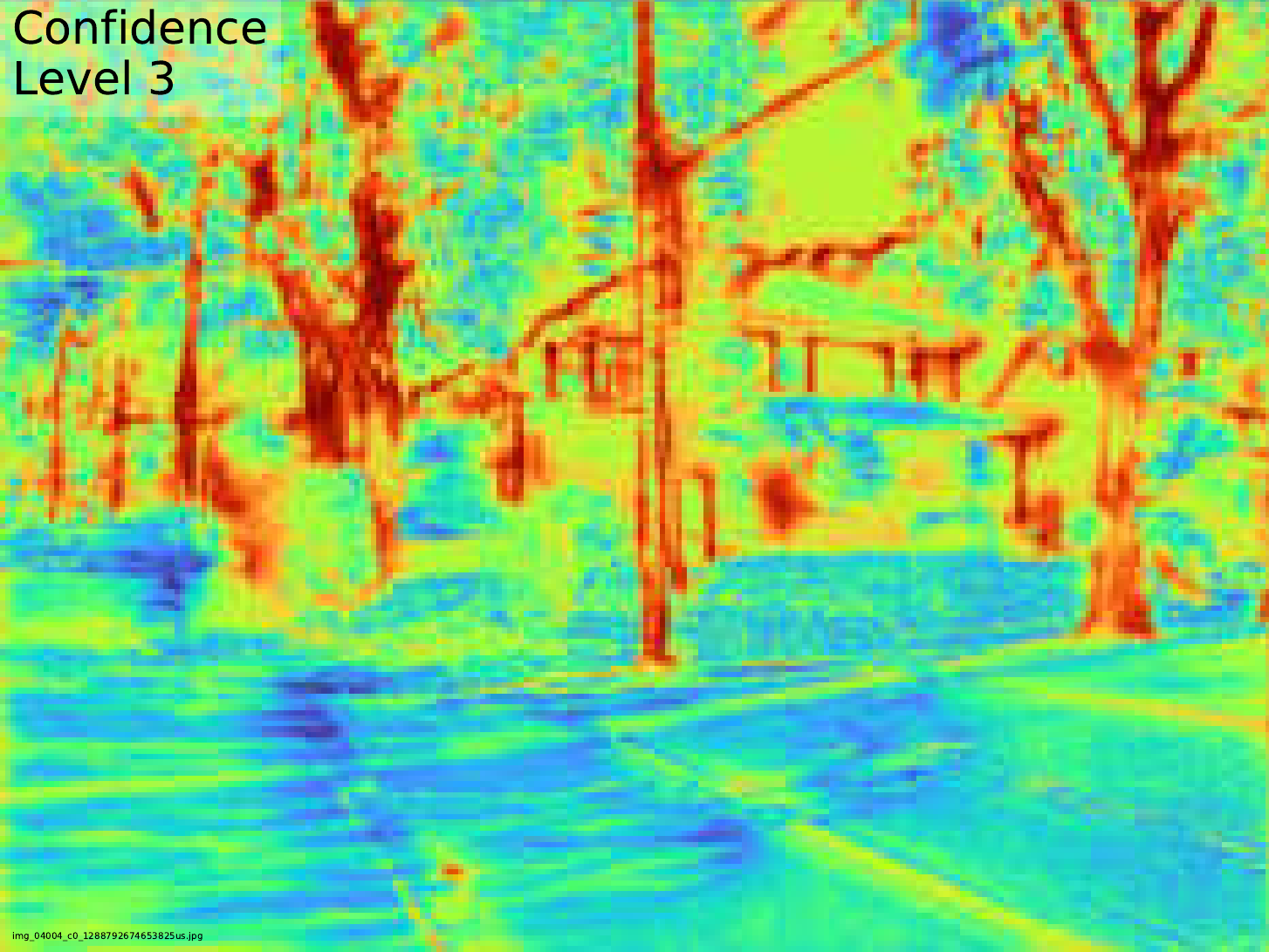}
\end{minipage}
\begin{minipage}{\lwidth\textwidth}
\rotatebox[origin=c]{90}{Reference}
\end{minipage}%
\begin{minipage}{\iwidth\textwidth}
    \centering
    \includegraphics[width=\pwidth\linewidth]{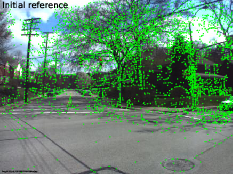}
\end{minipage}%
\begin{minipage}{\iwidth\textwidth}
    \centering
    \includegraphics[width=\pwidth\linewidth]{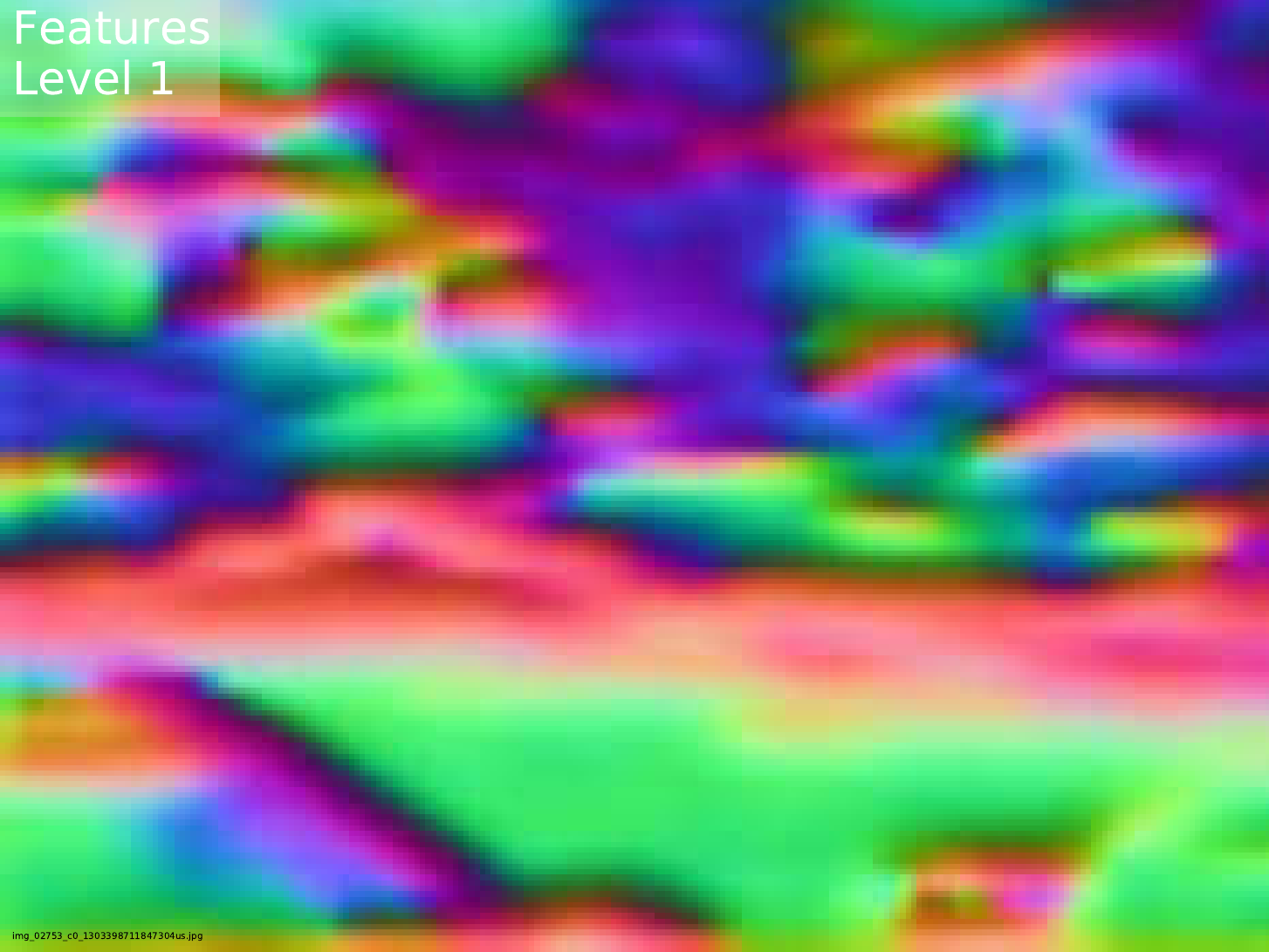}
\end{minipage}%
\begin{minipage}{\iwidth\textwidth}
    \centering
    \includegraphics[width=\pwidth\linewidth]{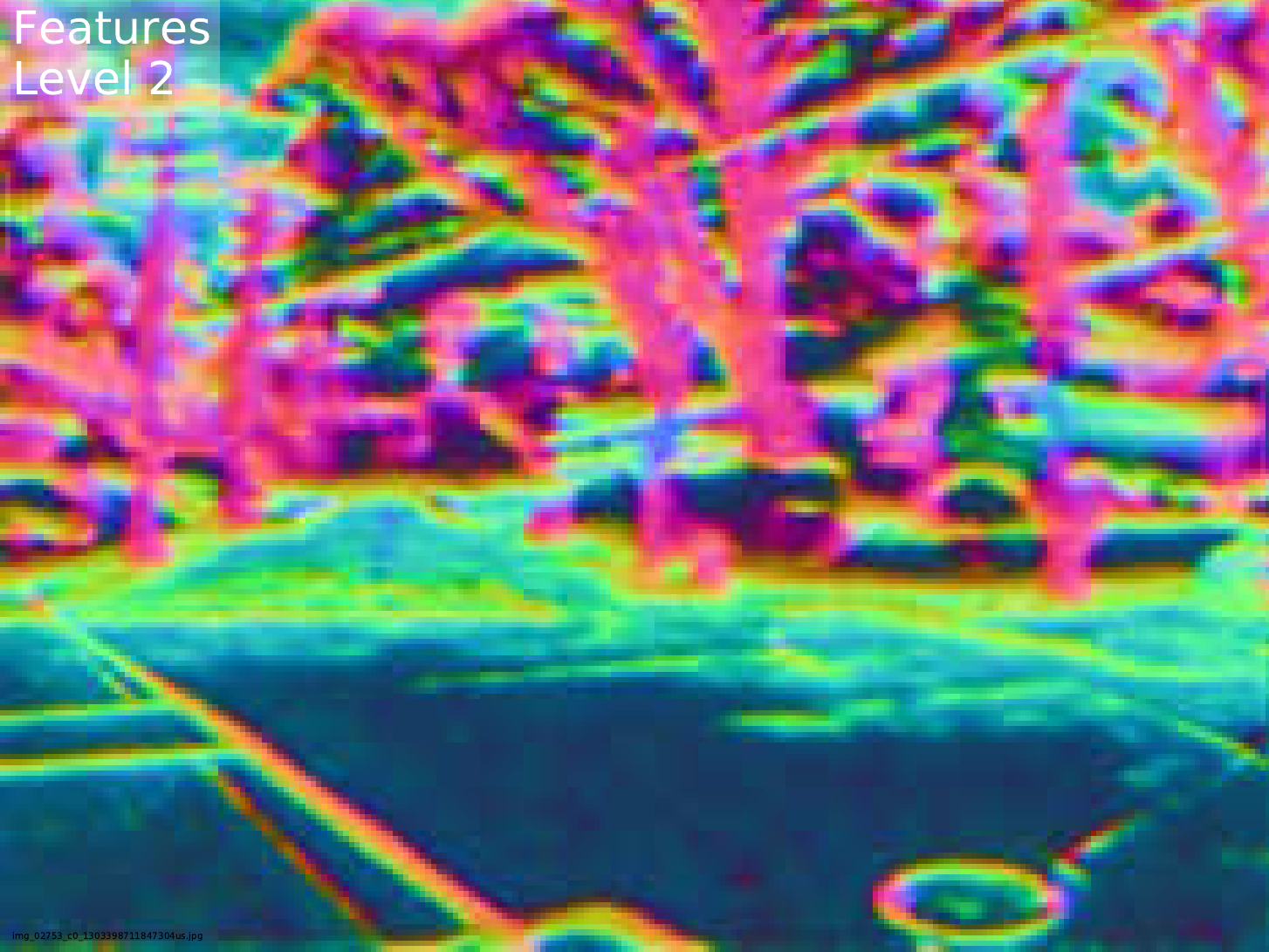}
\end{minipage}%
\begin{minipage}{\iwidth\textwidth}
    \centering
    \includegraphics[width=\pwidth\linewidth]{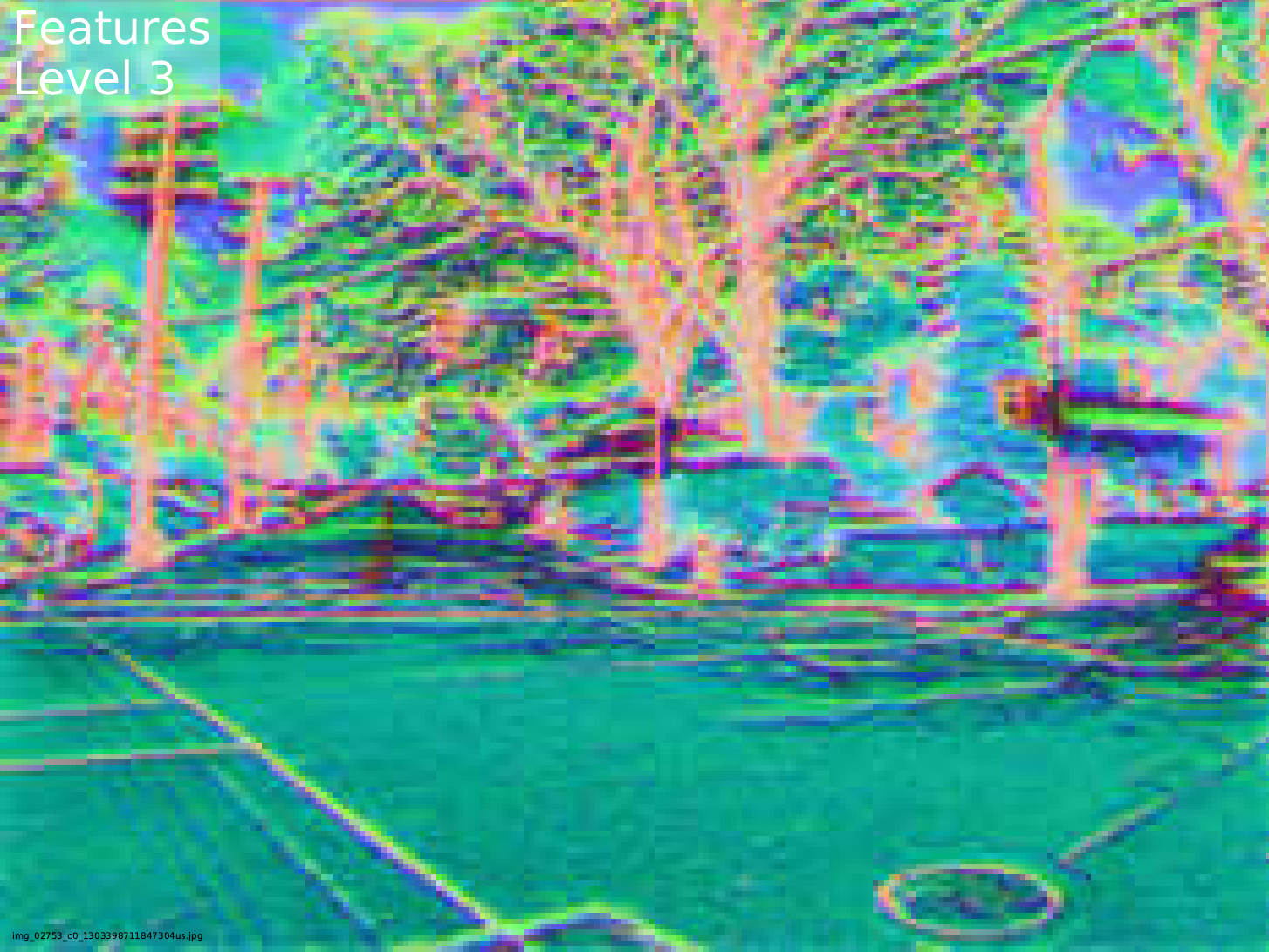}
\end{minipage}%
\begin{minipage}{\iwidth\textwidth}
    \centering
    \includegraphics[width=\pwidth\linewidth]{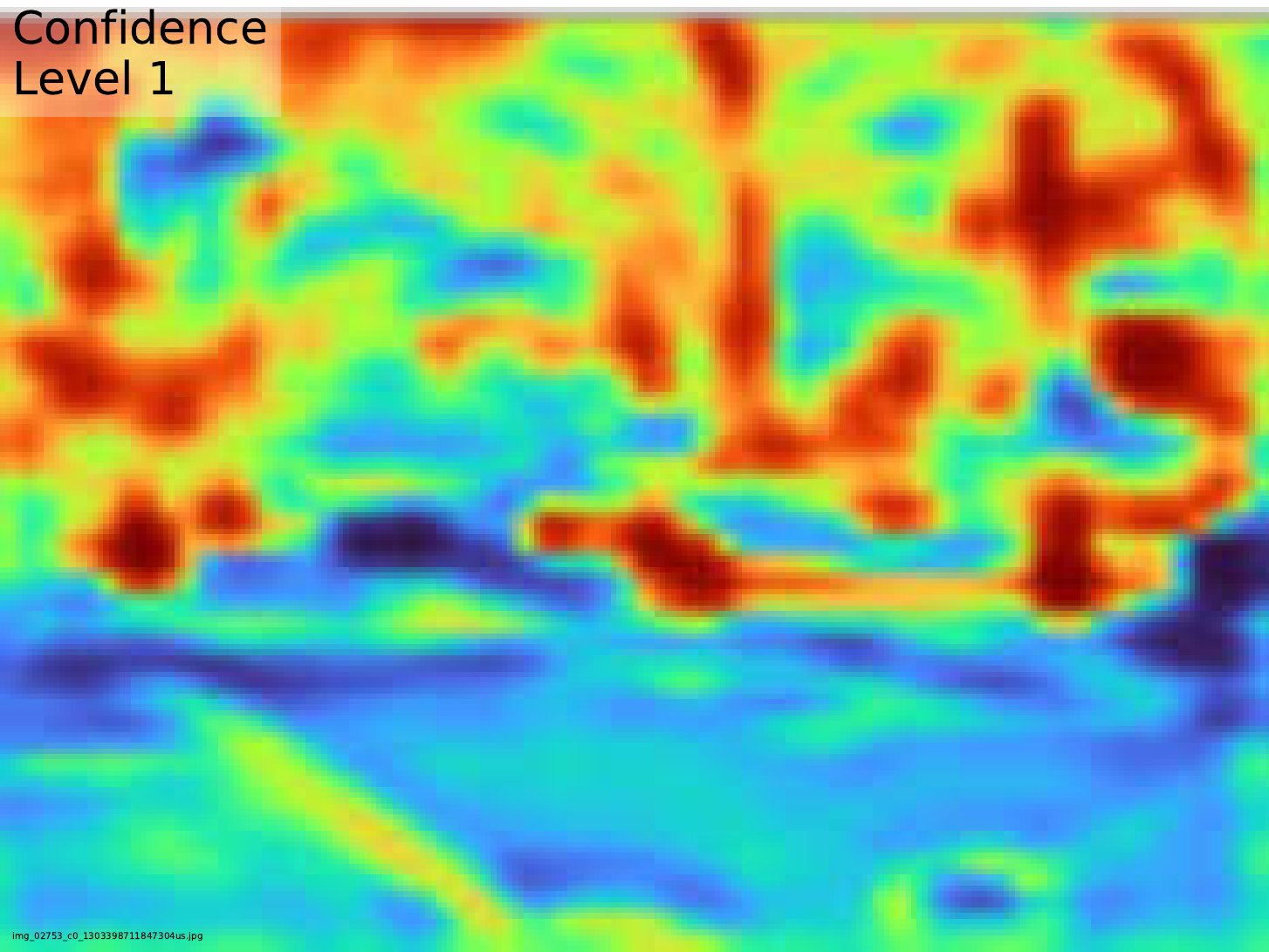}
\end{minipage}%
\begin{minipage}{\iwidth\textwidth}
    \centering
    \includegraphics[width=\pwidth\linewidth]{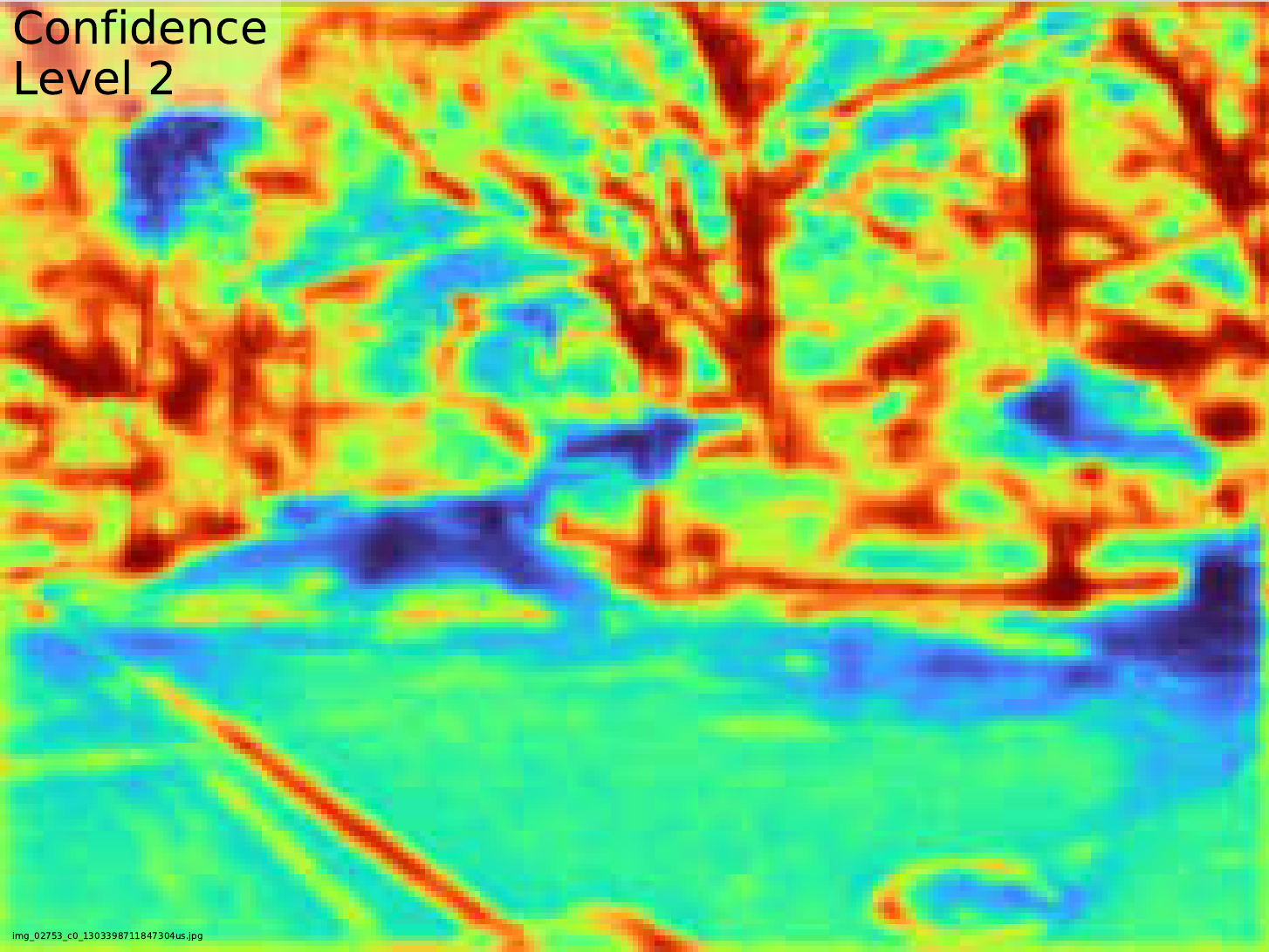}
\end{minipage}%
\begin{minipage}{\iwidth\textwidth}
    \centering
    \includegraphics[width=\pwidth\linewidth]{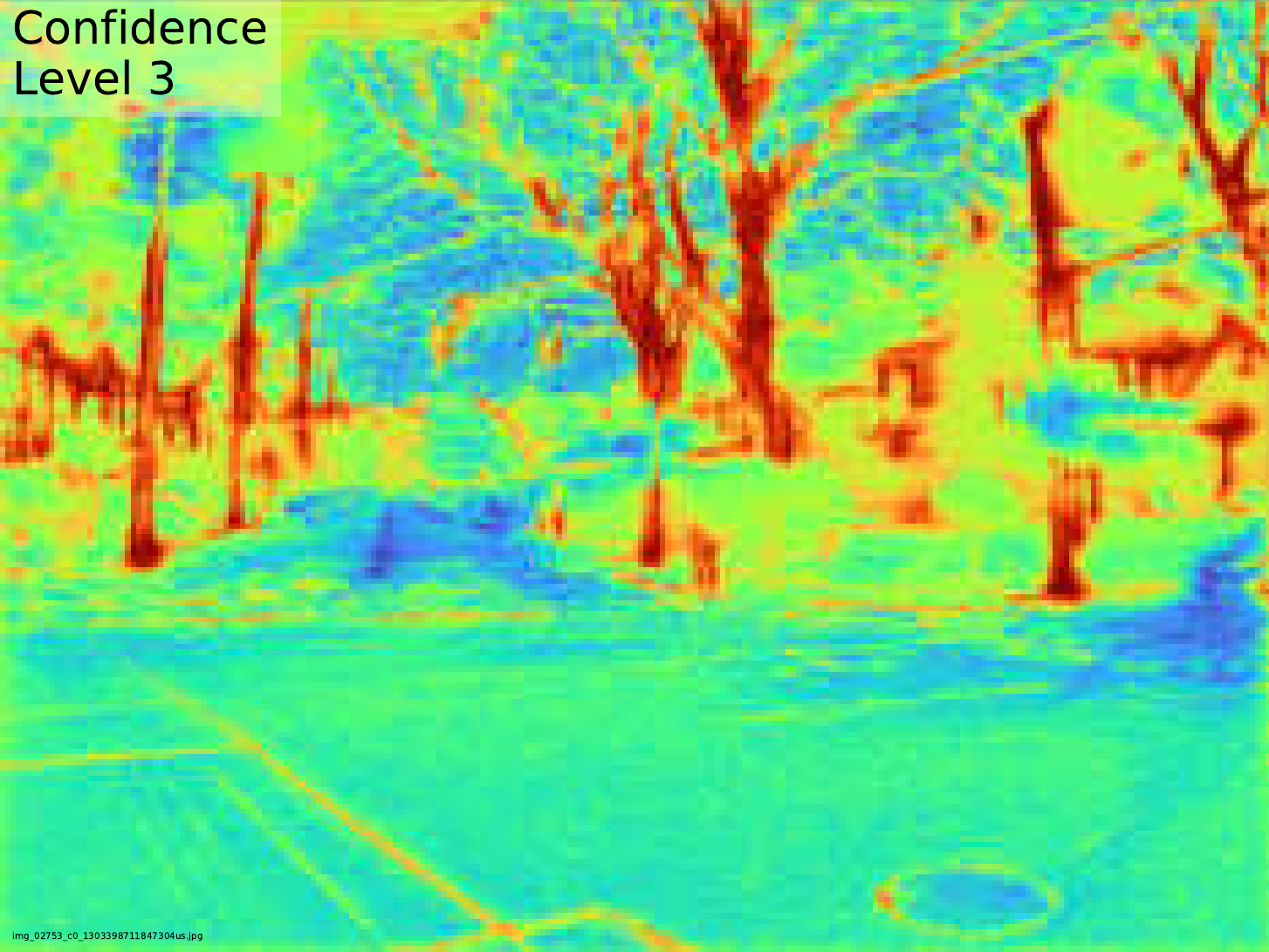}
\end{minipage}
\vspace{2mm}

\begin{minipage}{\lwidth\textwidth}
\rotatebox[origin=c]{90}{Query}
\end{minipage}%
\begin{minipage}{\iwidth\textwidth}
    \centering
    \includegraphics[width=\pwidth\linewidth]{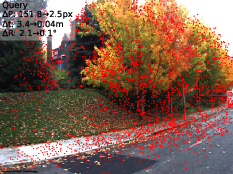}
\end{minipage}%
\begin{minipage}{\iwidth\textwidth}
    \centering
    \includegraphics[width=\pwidth\linewidth]{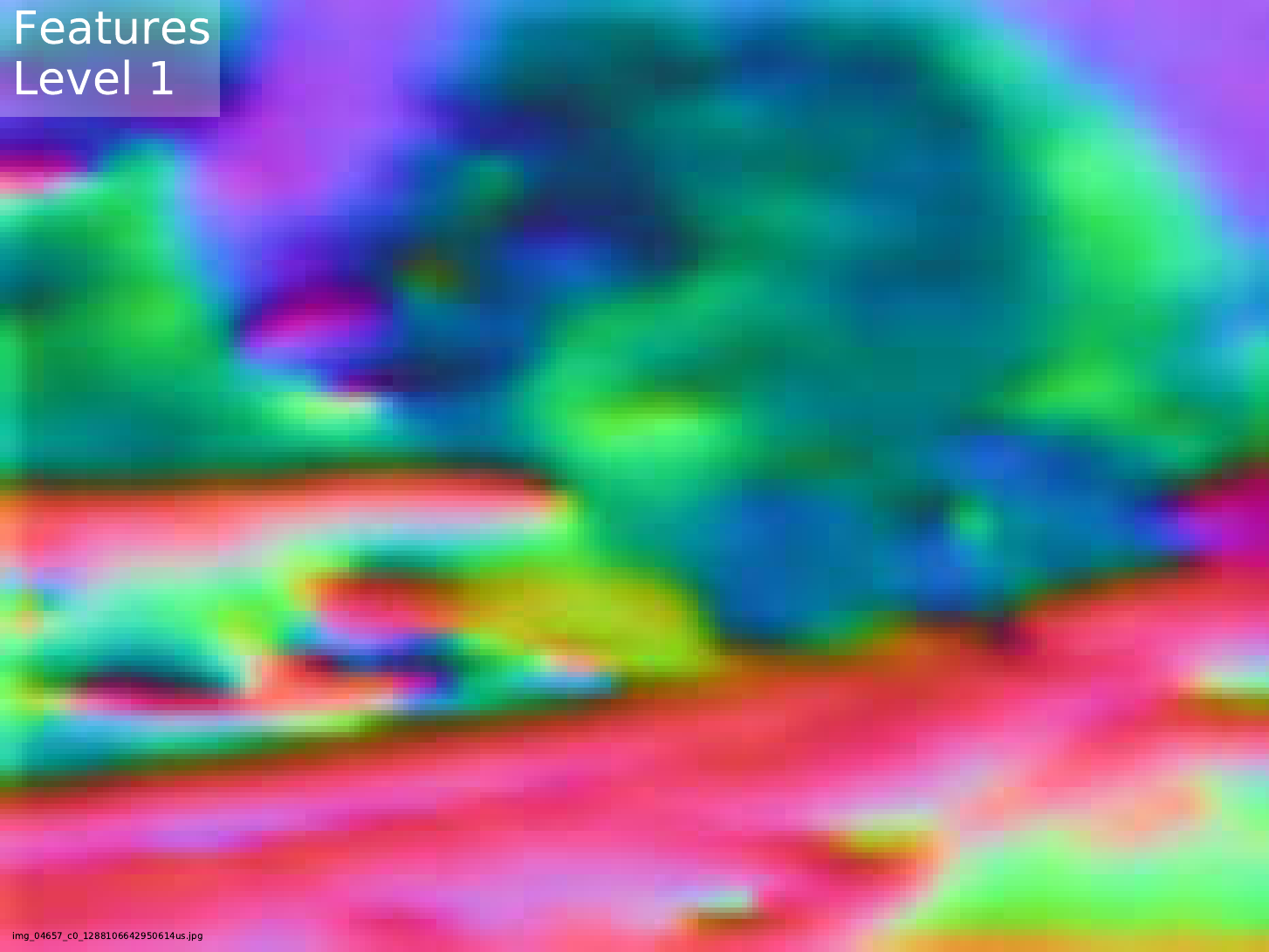}
\end{minipage}%
\begin{minipage}{\iwidth\textwidth}
    \centering
    \includegraphics[width=\pwidth\linewidth]{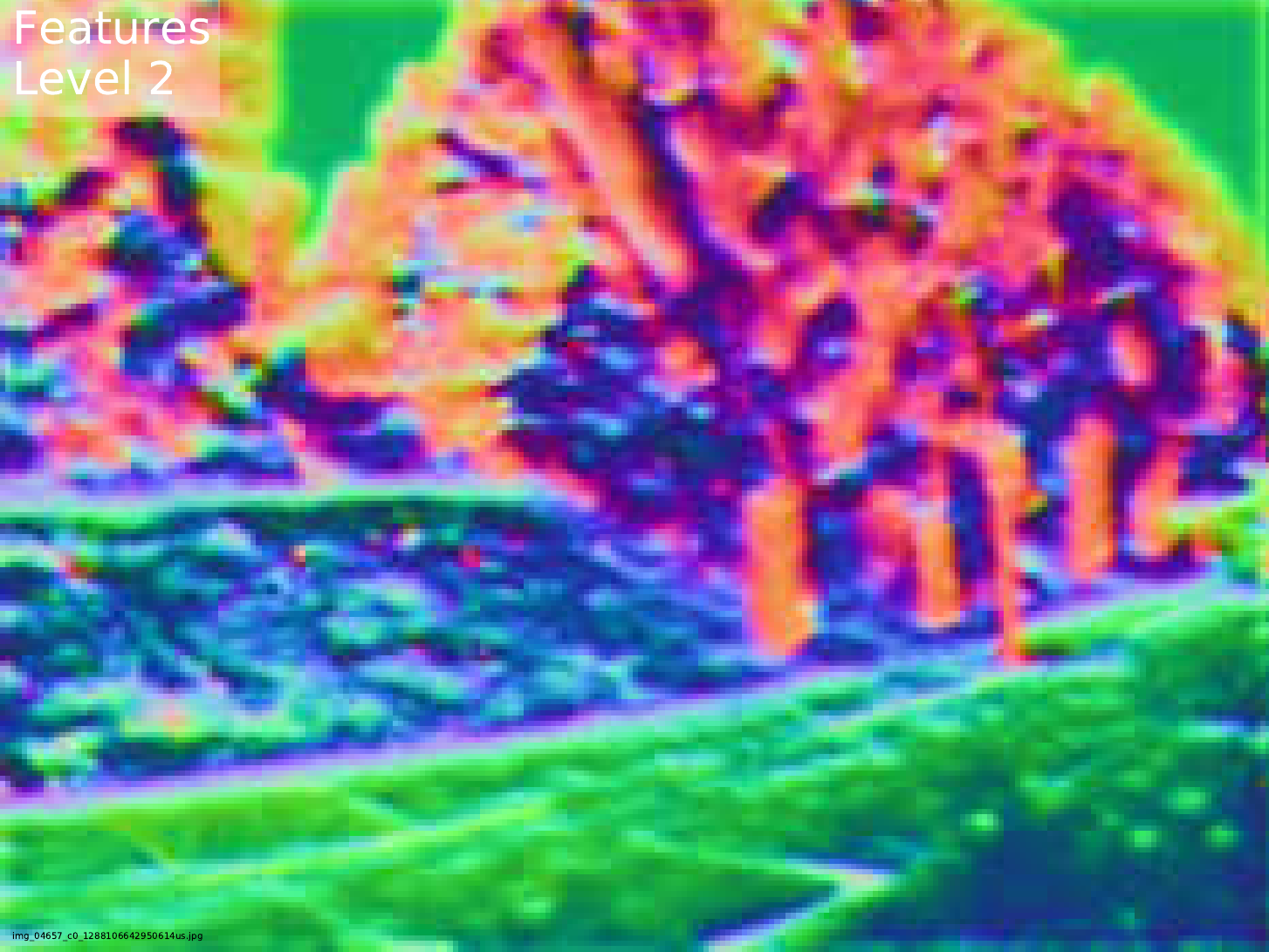}
\end{minipage}%
\begin{minipage}{\iwidth\textwidth}
    \centering
    \includegraphics[width=\pwidth\linewidth]{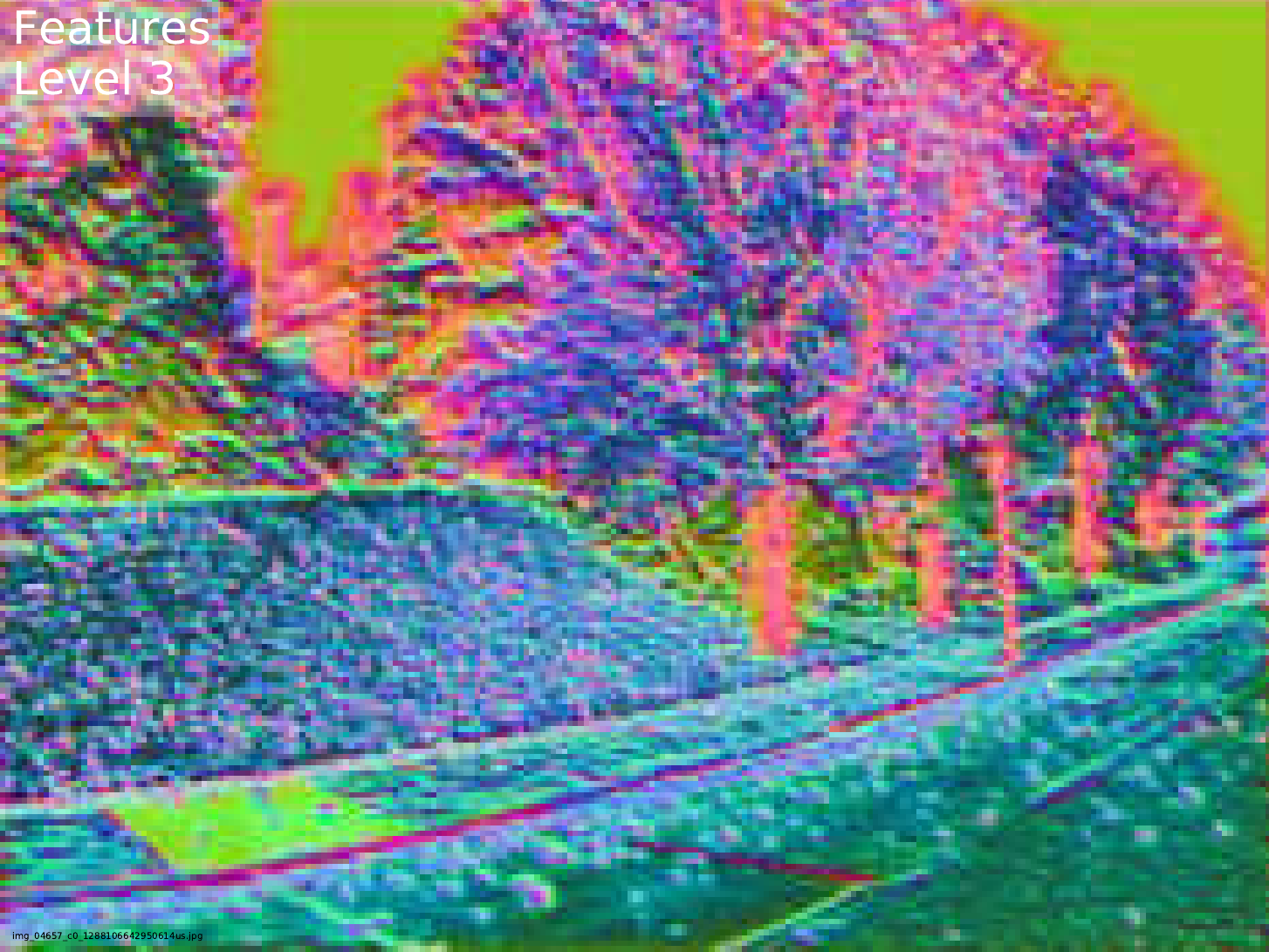}
\end{minipage}%
\begin{minipage}{\iwidth\textwidth}
    \centering
    \includegraphics[width=\pwidth\linewidth]{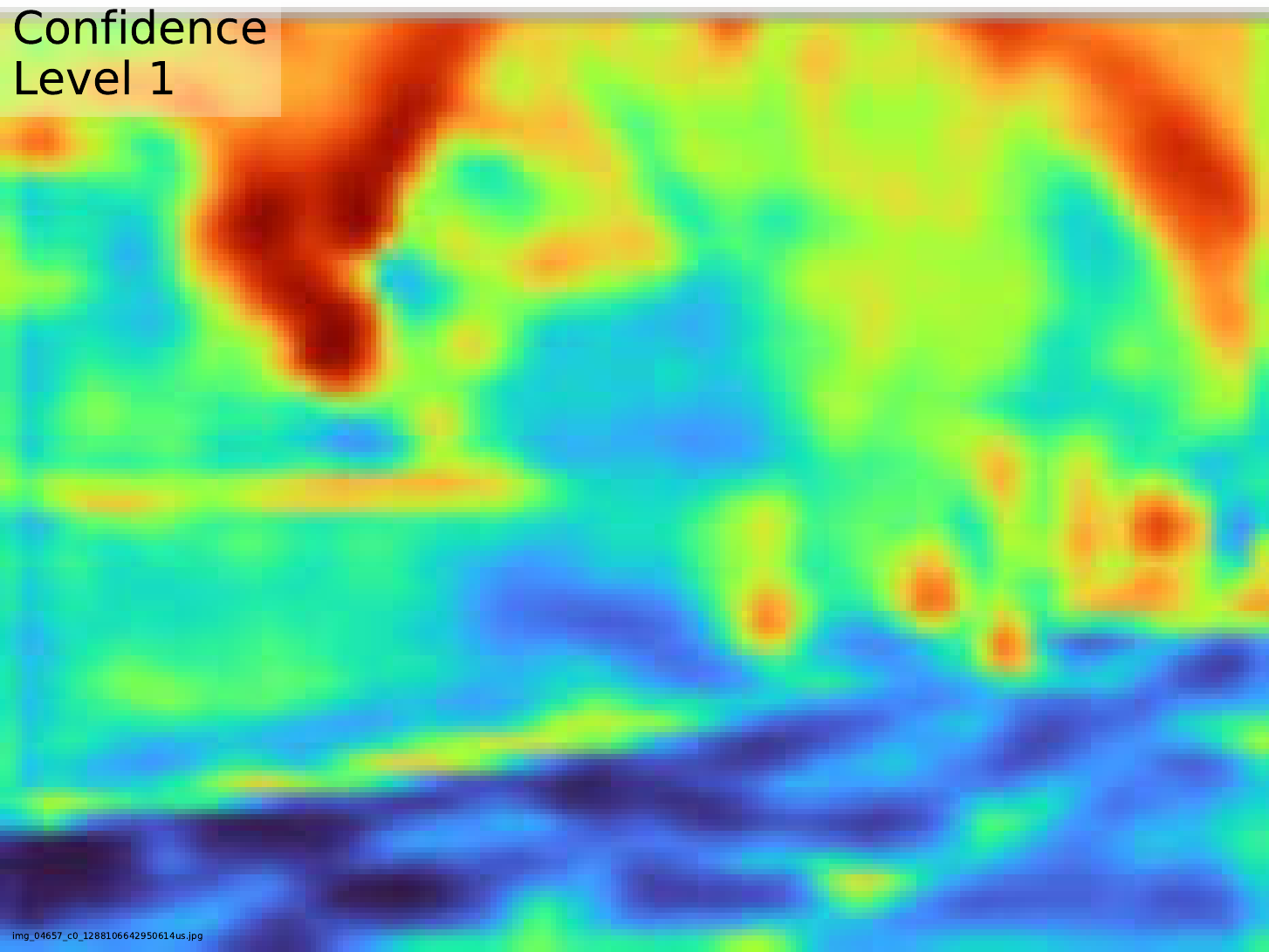}
\end{minipage}%
\begin{minipage}{\iwidth\textwidth}
    \centering
    \includegraphics[width=\pwidth\linewidth]{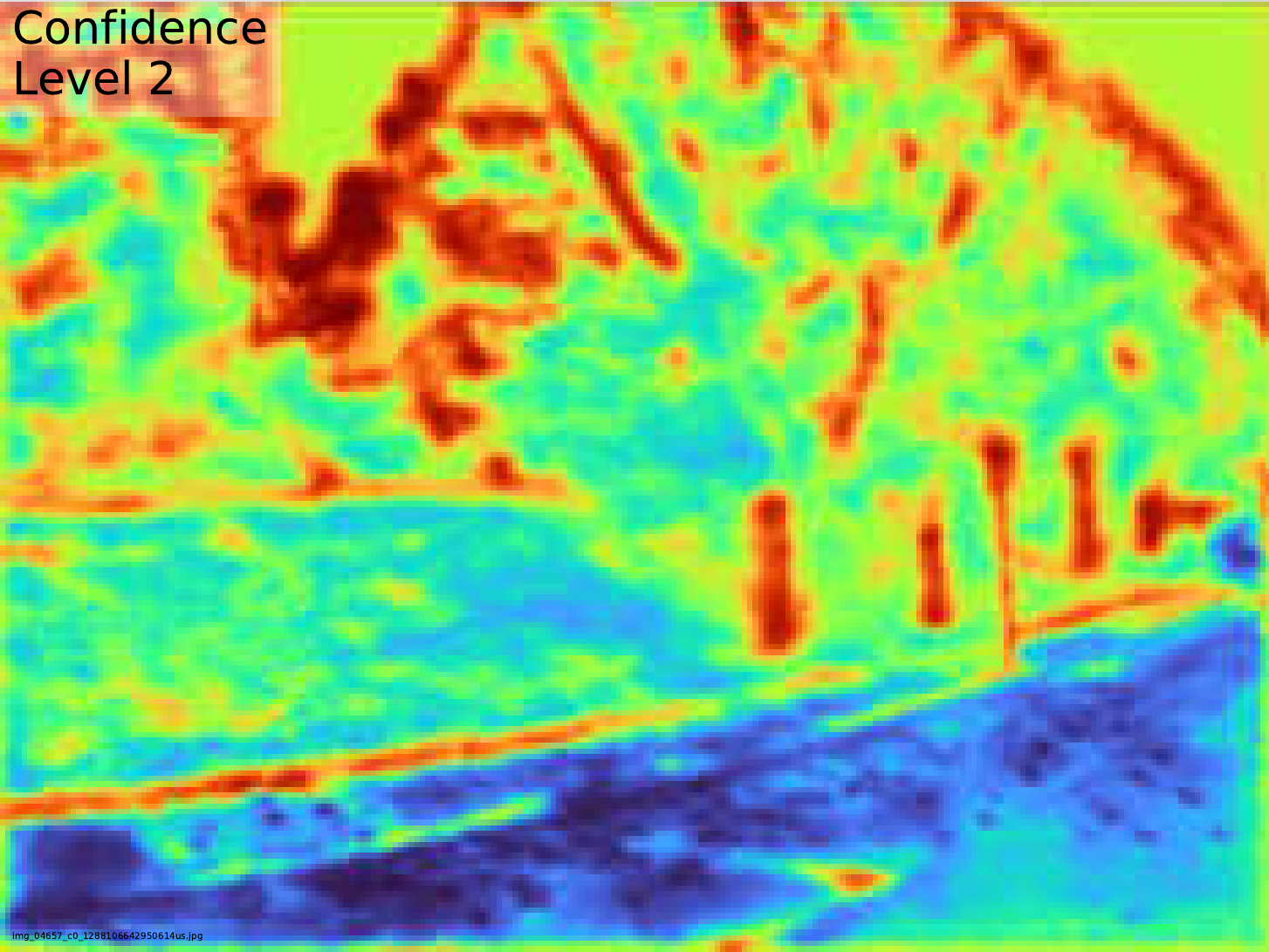}
\end{minipage}%
\begin{minipage}{\iwidth\textwidth}
    \centering
    \includegraphics[width=\pwidth\linewidth]{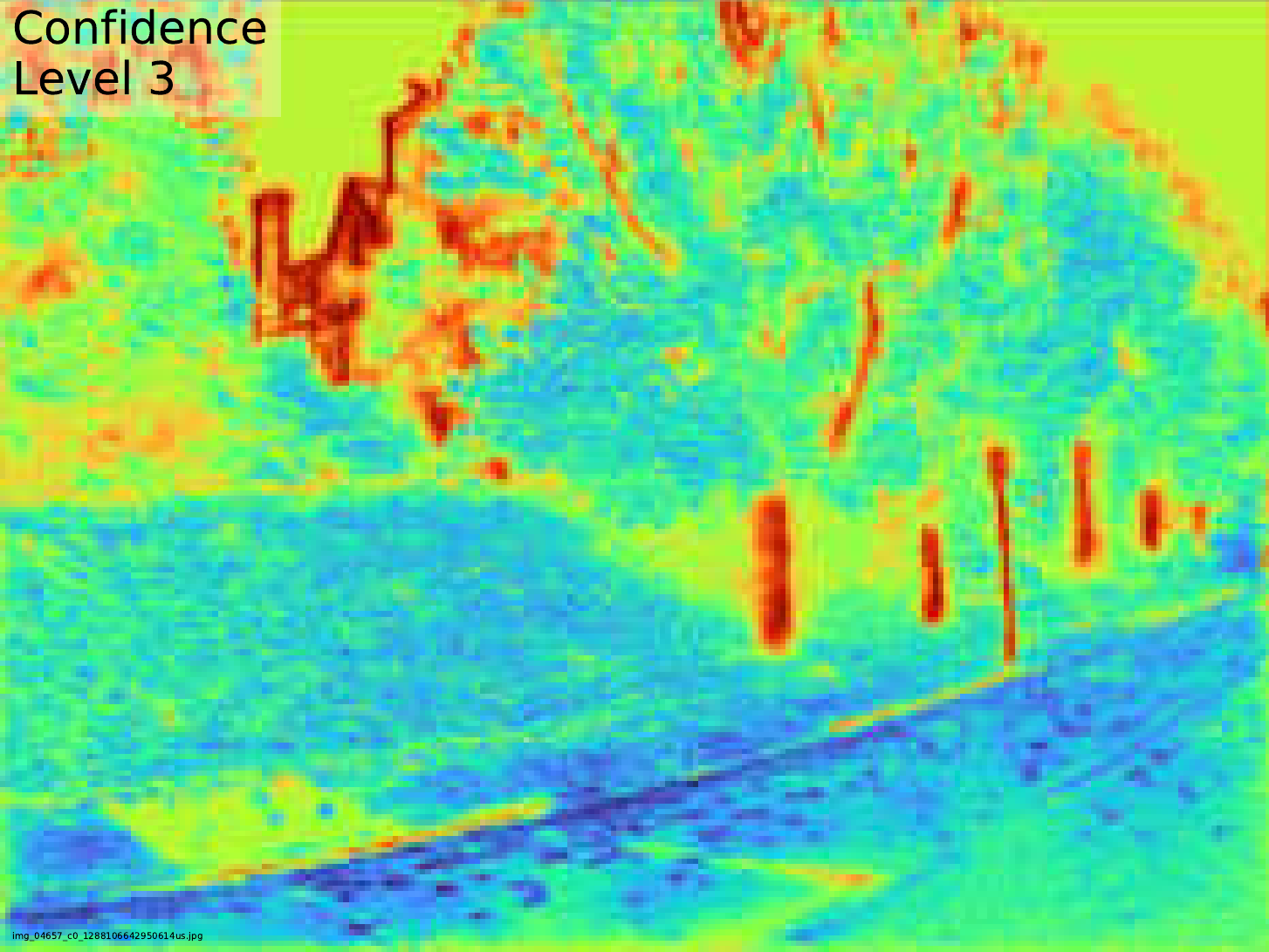}
\end{minipage}
\begin{minipage}{\lwidth\textwidth}
\rotatebox[origin=c]{90}{Reference}
\end{minipage}%
\begin{minipage}{\iwidth\textwidth}
    \centering
    \includegraphics[width=\pwidth\linewidth]{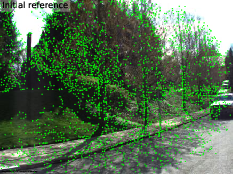}
\end{minipage}%
\begin{minipage}{\iwidth\textwidth}
    \centering
    \includegraphics[width=\pwidth\linewidth]{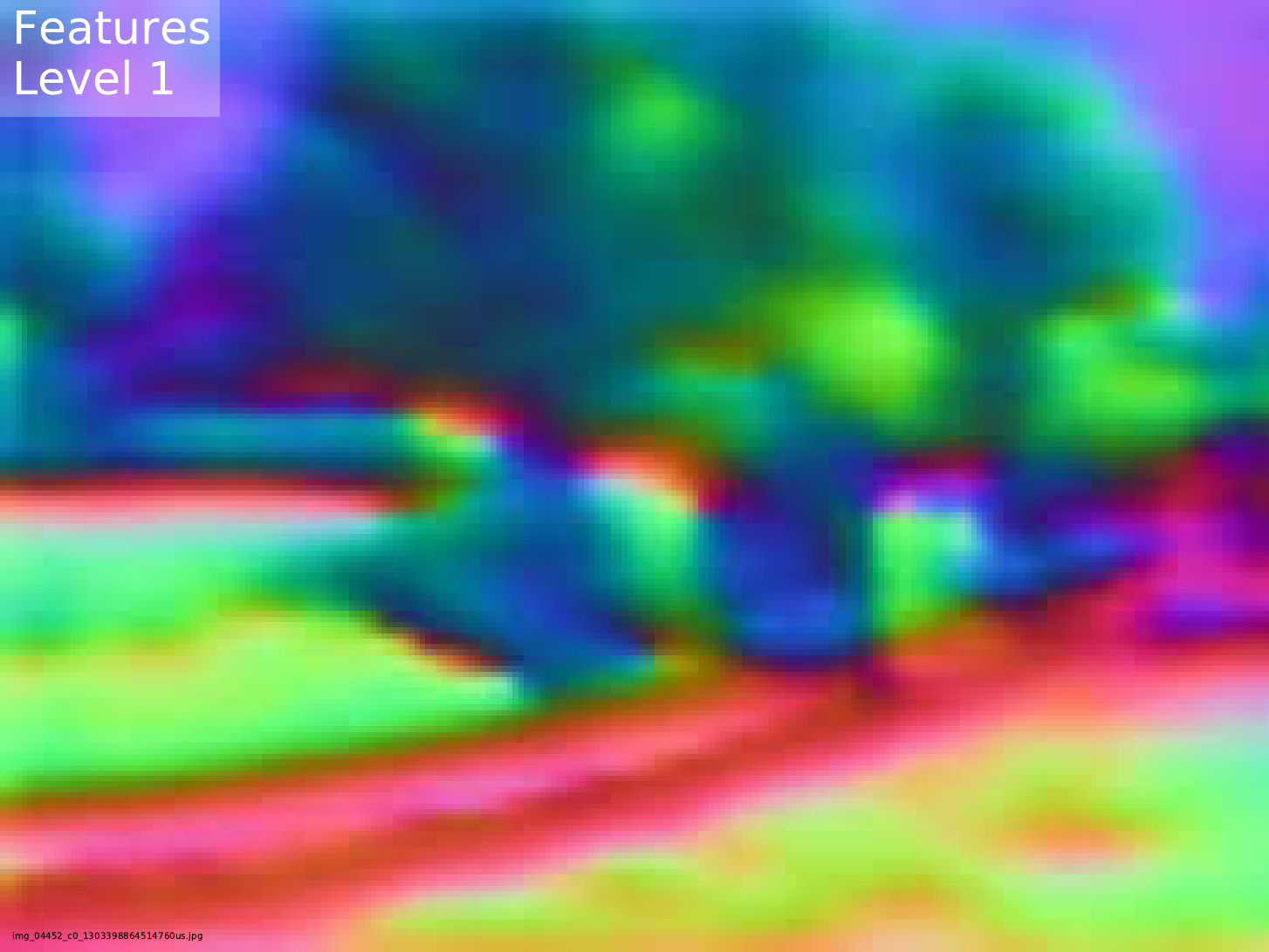}
\end{minipage}%
\begin{minipage}{\iwidth\textwidth}
    \centering
    \includegraphics[width=\pwidth\linewidth]{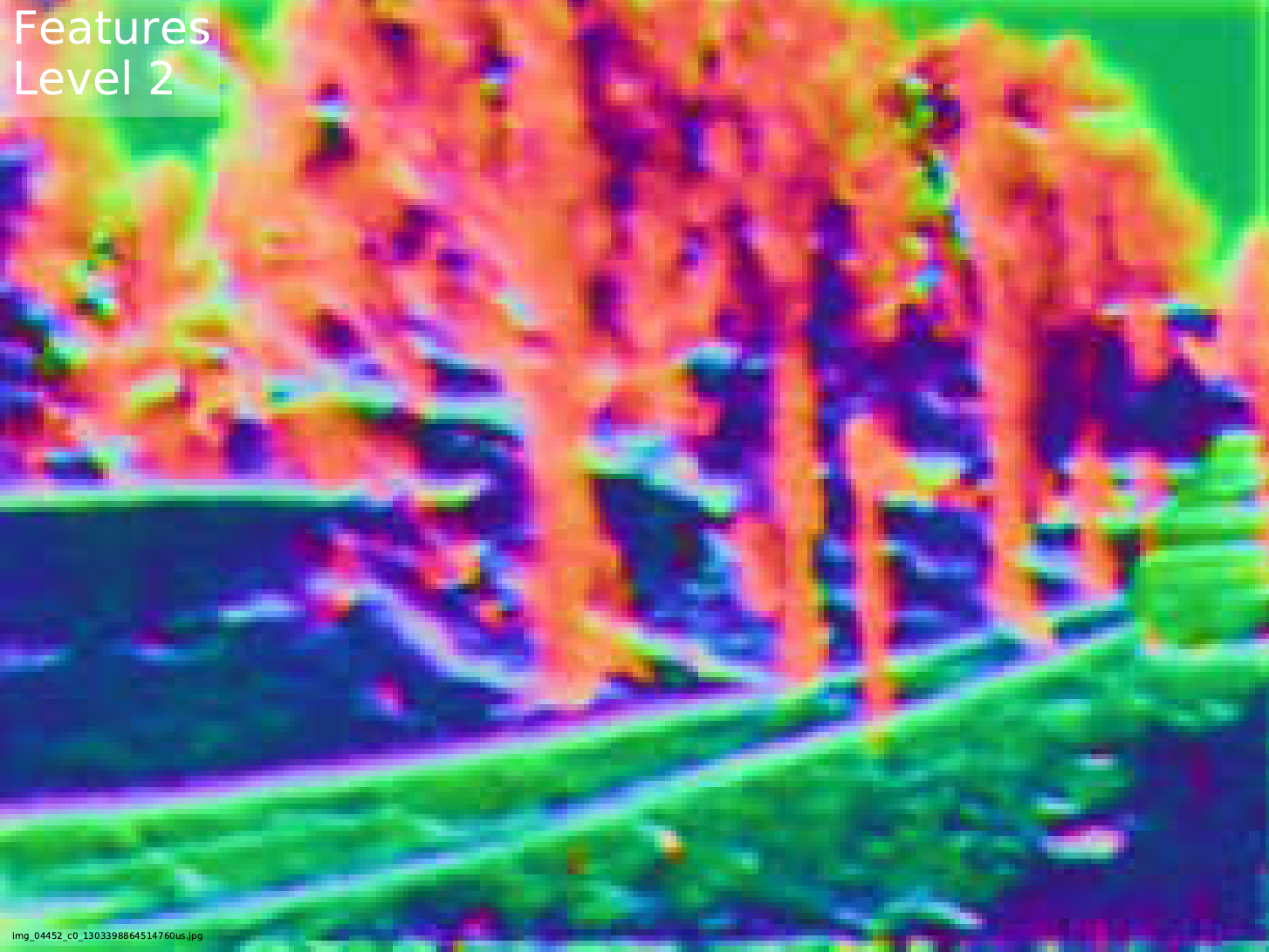}
\end{minipage}%
\begin{minipage}{\iwidth\textwidth}
    \centering
    \includegraphics[width=\pwidth\linewidth]{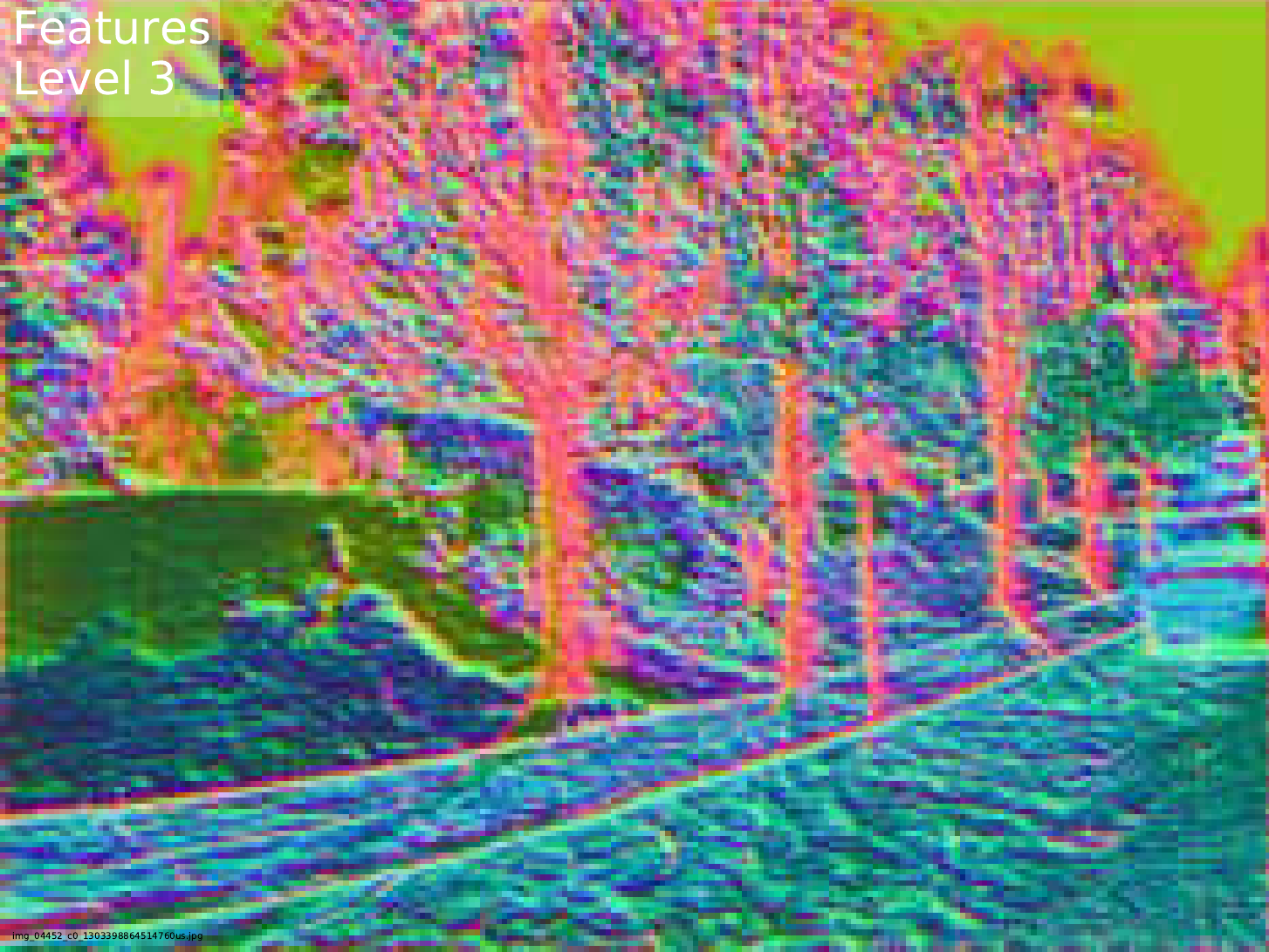}
\end{minipage}%
\begin{minipage}{\iwidth\textwidth}
    \centering
    \includegraphics[width=\pwidth\linewidth]{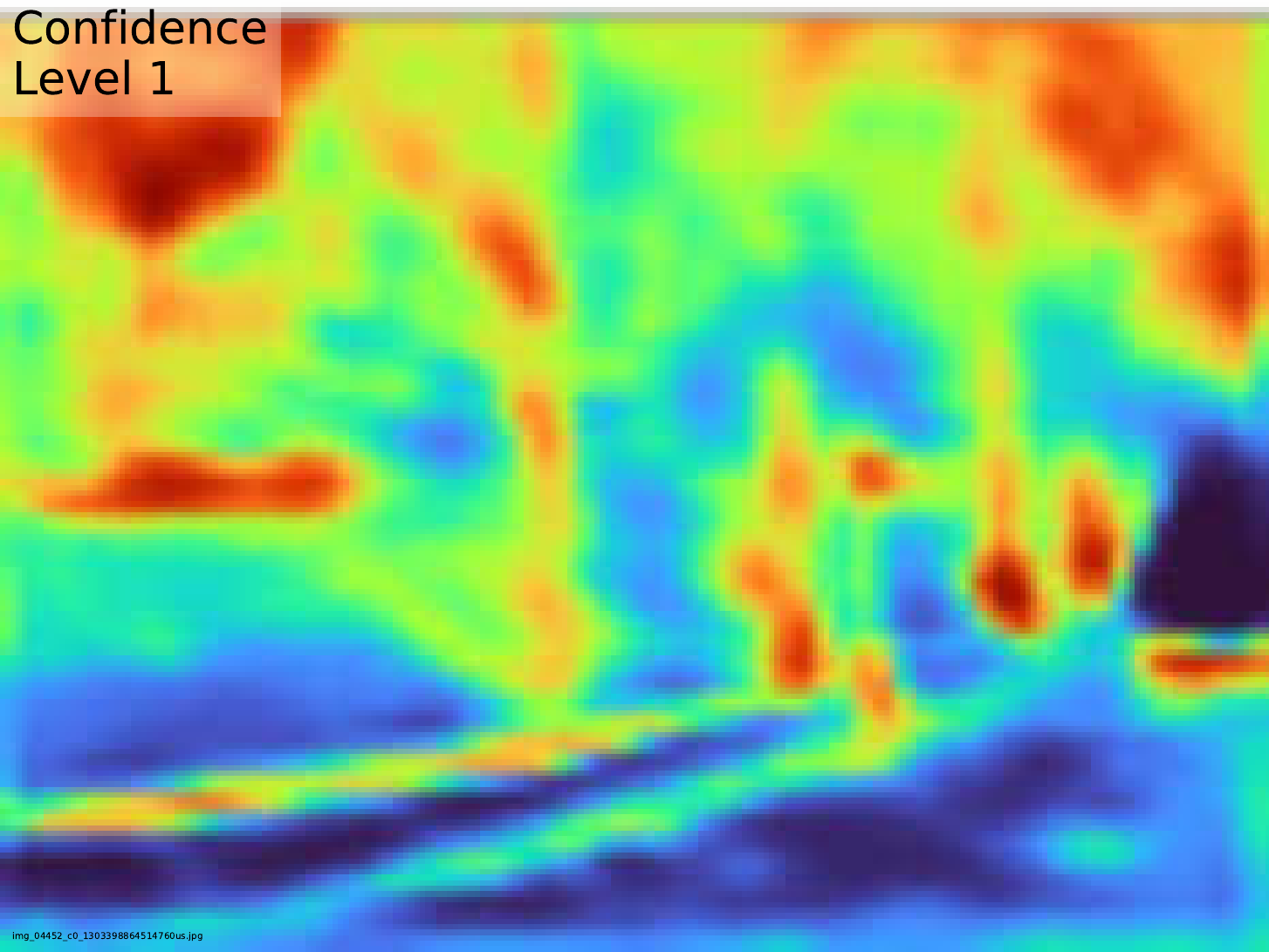}
\end{minipage}%
\begin{minipage}{\iwidth\textwidth}
    \centering
    \includegraphics[width=\pwidth\linewidth]{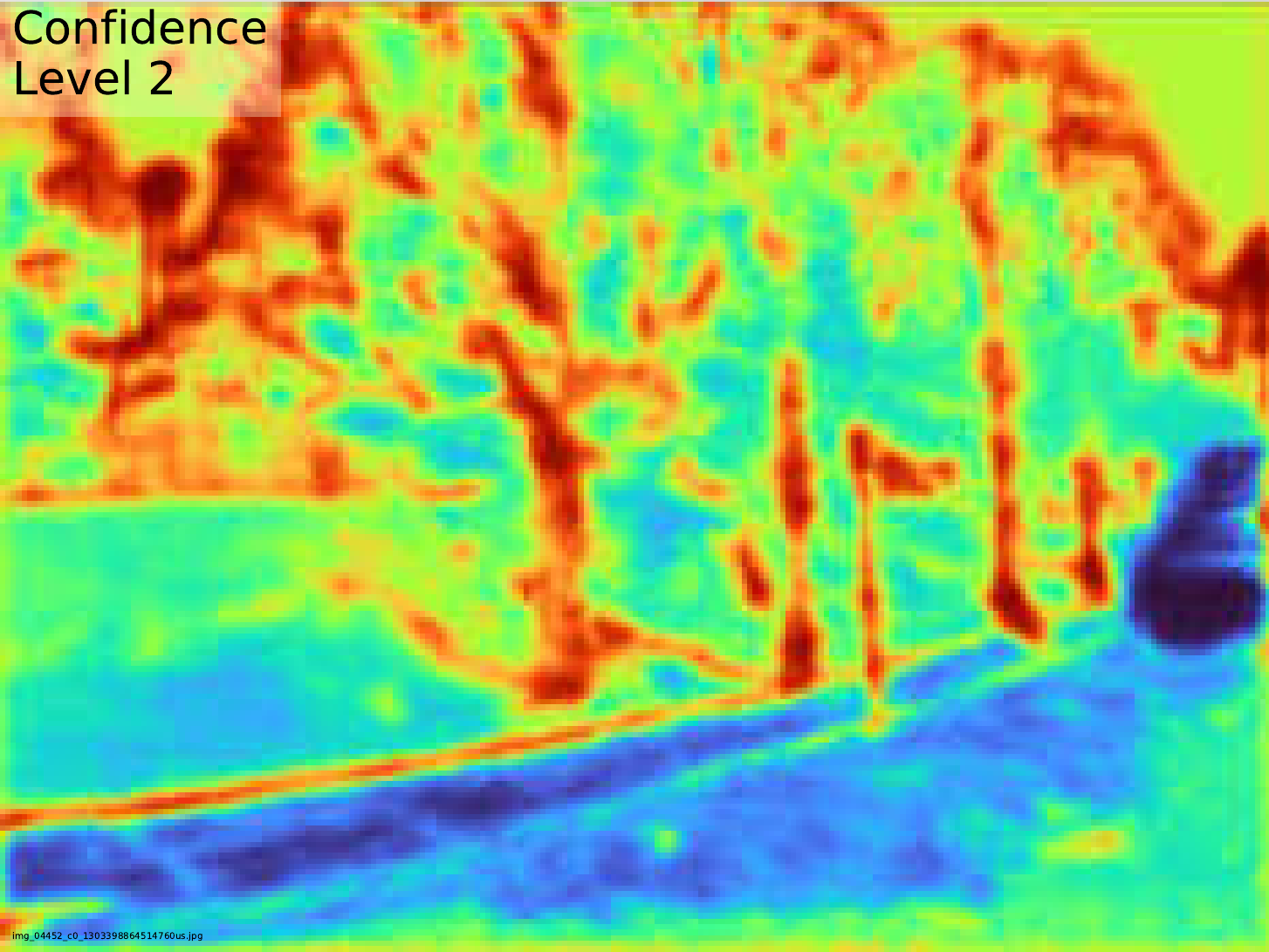}
\end{minipage}%
\begin{minipage}{\iwidth\textwidth}
    \centering
    \includegraphics[width=\pwidth\linewidth]{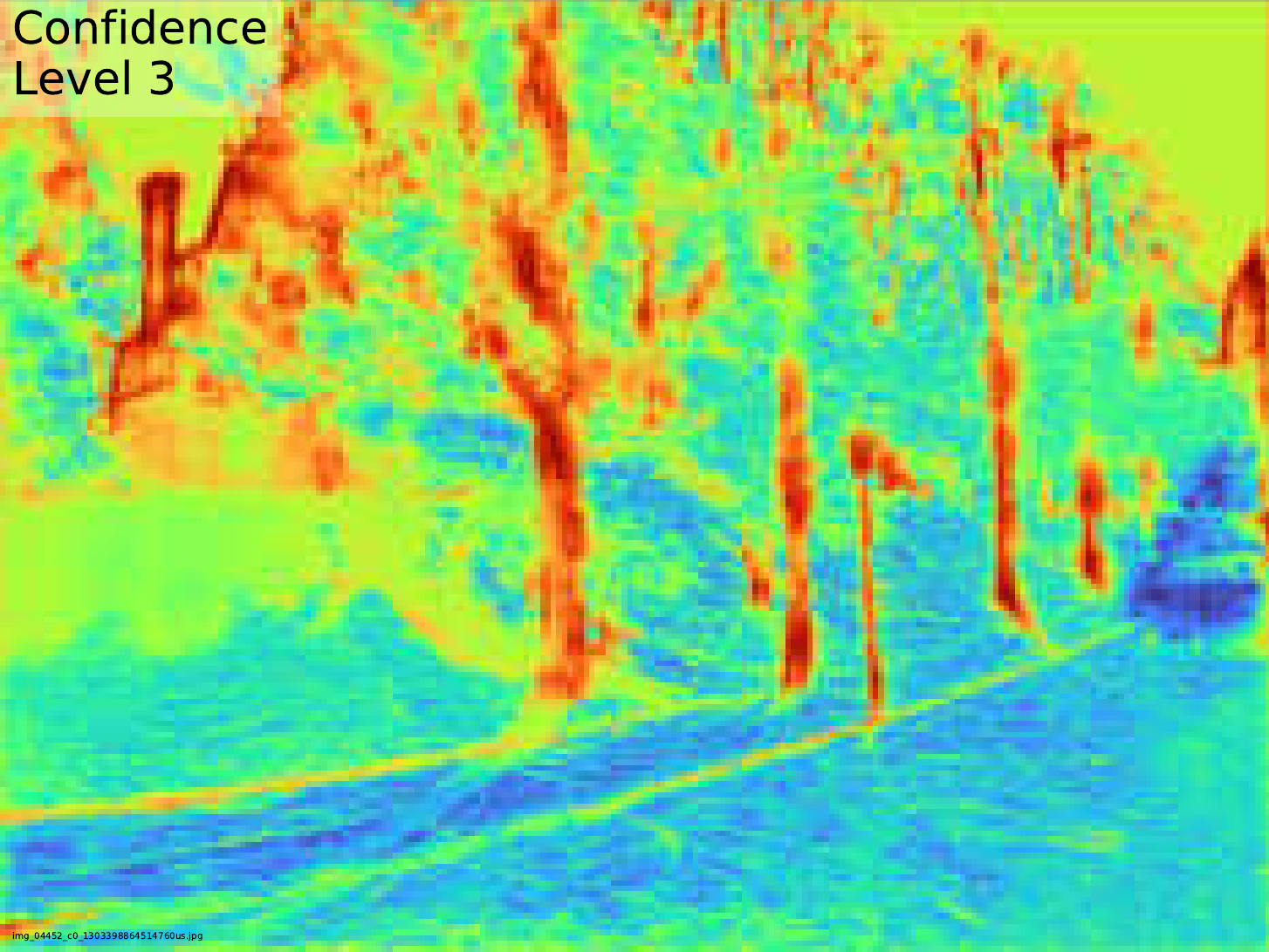}
\end{minipage}
\vspace{2mm}

\begin{minipage}{\lwidth\textwidth}
\rotatebox[origin=c]{90}{Query}
\end{minipage}%
\begin{minipage}{\iwidth\textwidth}
    \centering
    \includegraphics[width=\pwidth\linewidth]{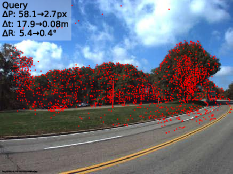}
\end{minipage}%
\begin{minipage}{\iwidth\textwidth}
    \centering
    \includegraphics[width=\pwidth\linewidth]{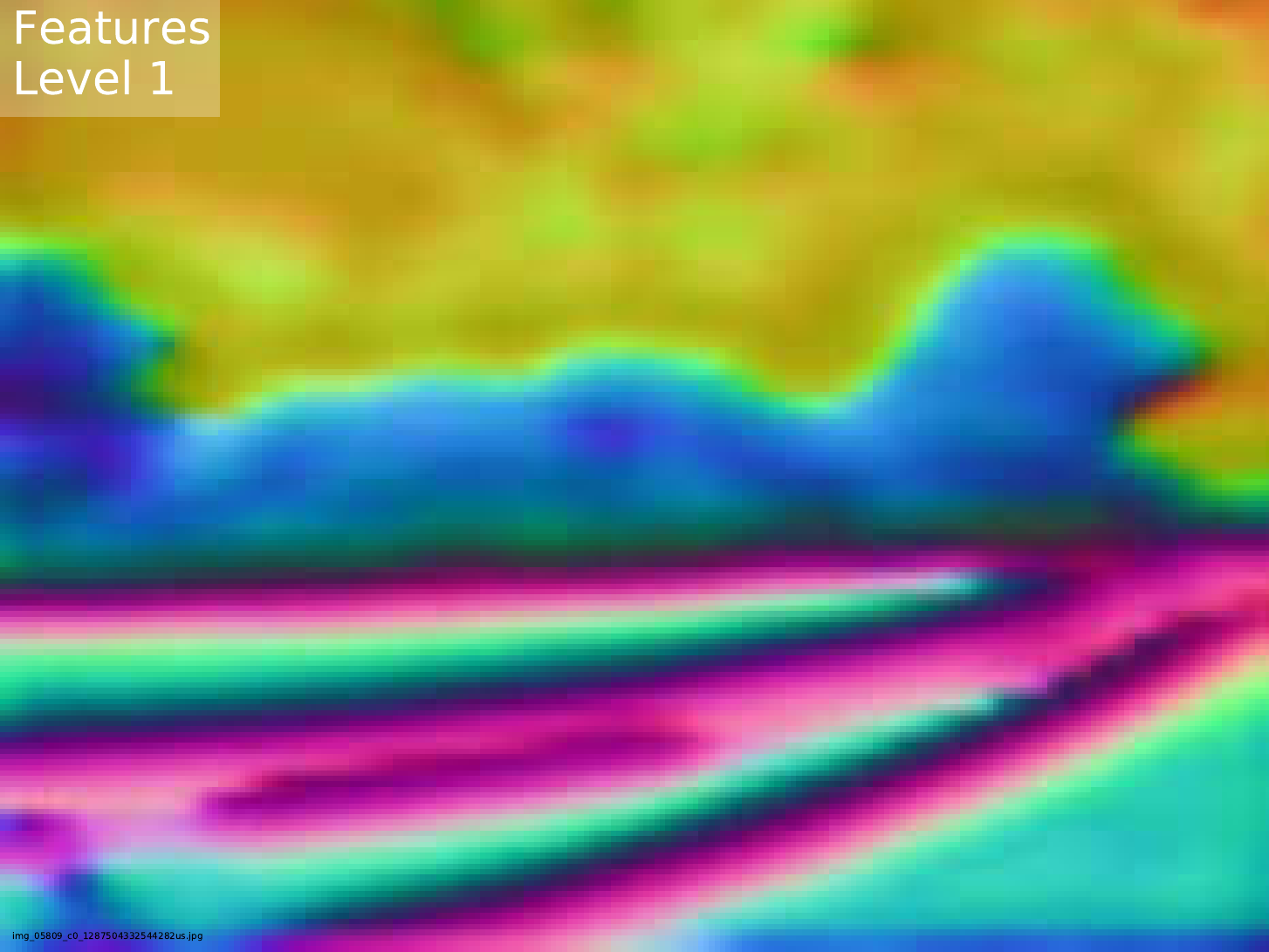}
\end{minipage}%
\begin{minipage}{\iwidth\textwidth}
    \centering
    \includegraphics[width=\pwidth\linewidth]{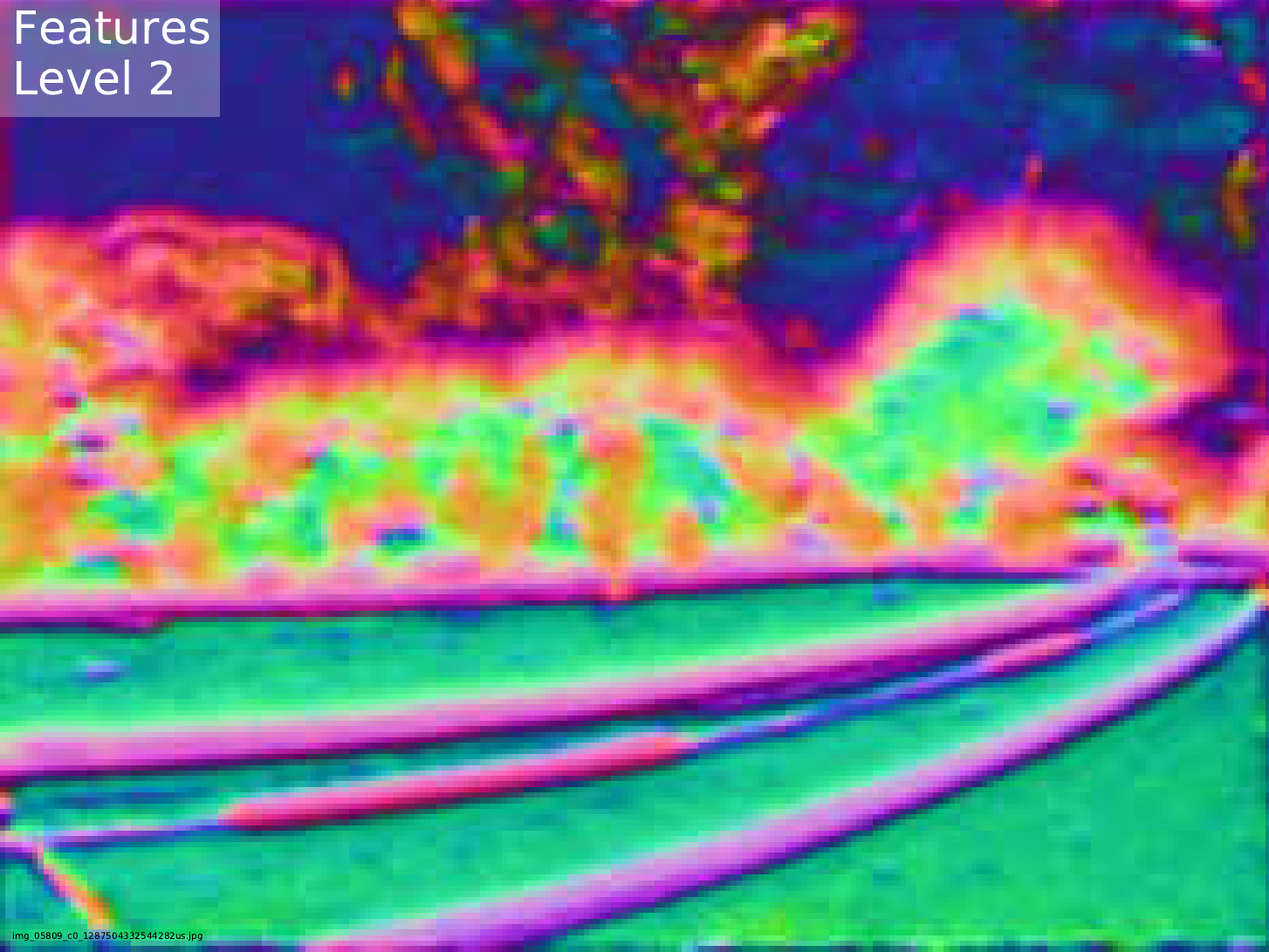}
\end{minipage}%
\begin{minipage}{\iwidth\textwidth}
    \centering
    \includegraphics[width=\pwidth\linewidth]{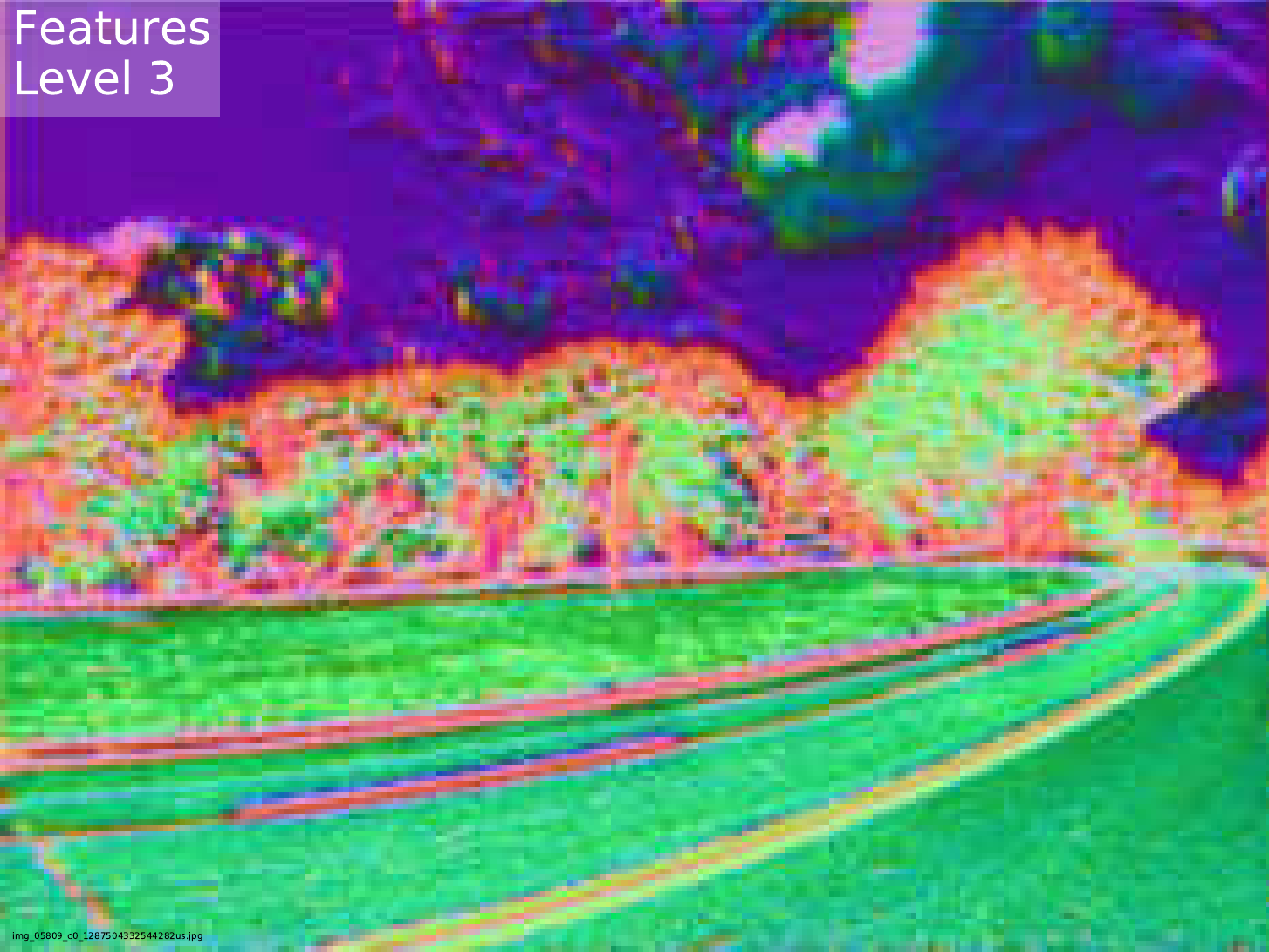}
\end{minipage}%
\begin{minipage}{\iwidth\textwidth}
    \centering
    \includegraphics[width=\pwidth\linewidth]{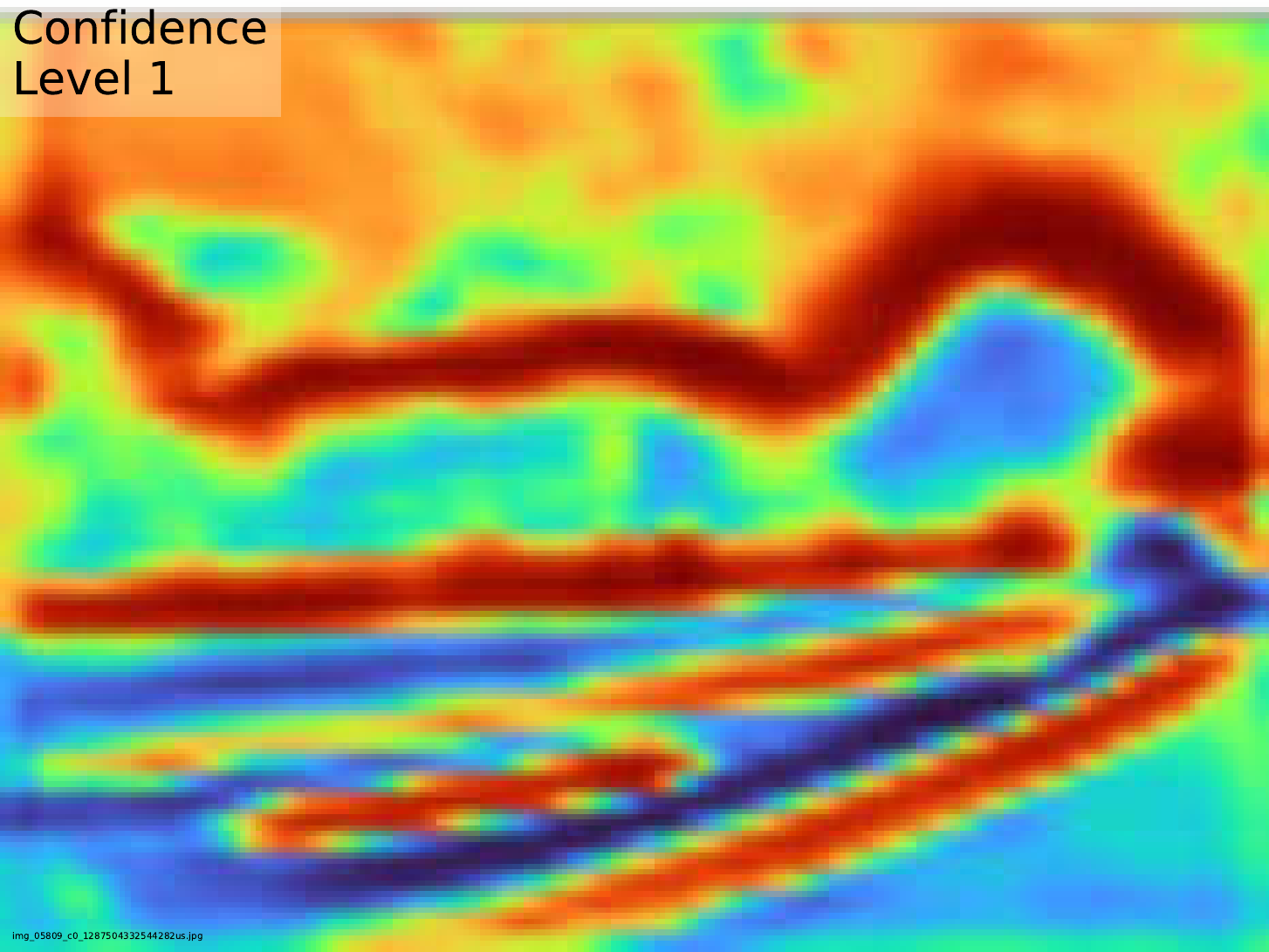}
\end{minipage}%
\begin{minipage}{\iwidth\textwidth}
    \centering
    \includegraphics[width=\pwidth\linewidth]{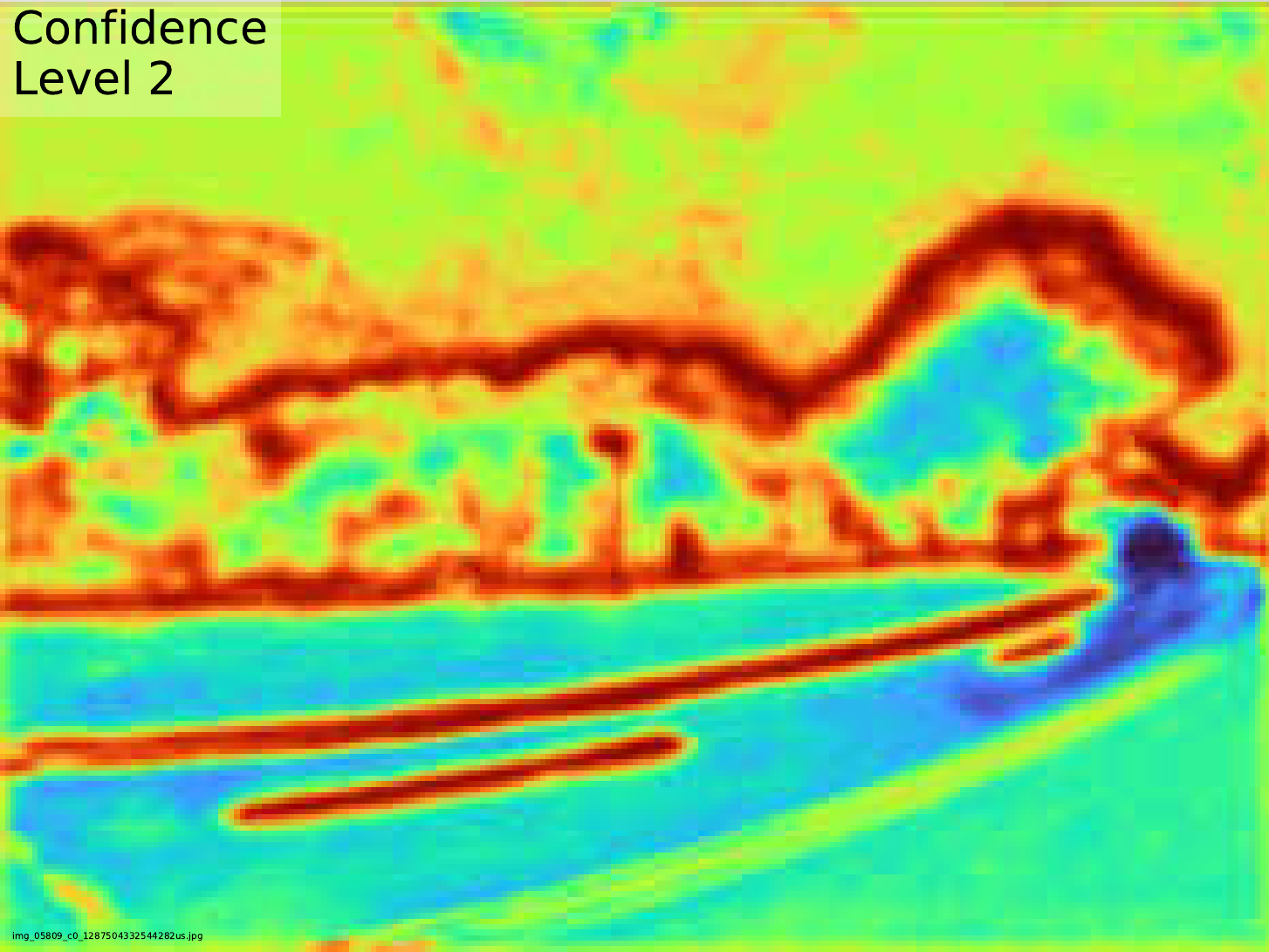}
\end{minipage}%
\begin{minipage}{\iwidth\textwidth}
    \centering
    \includegraphics[width=\pwidth\linewidth]{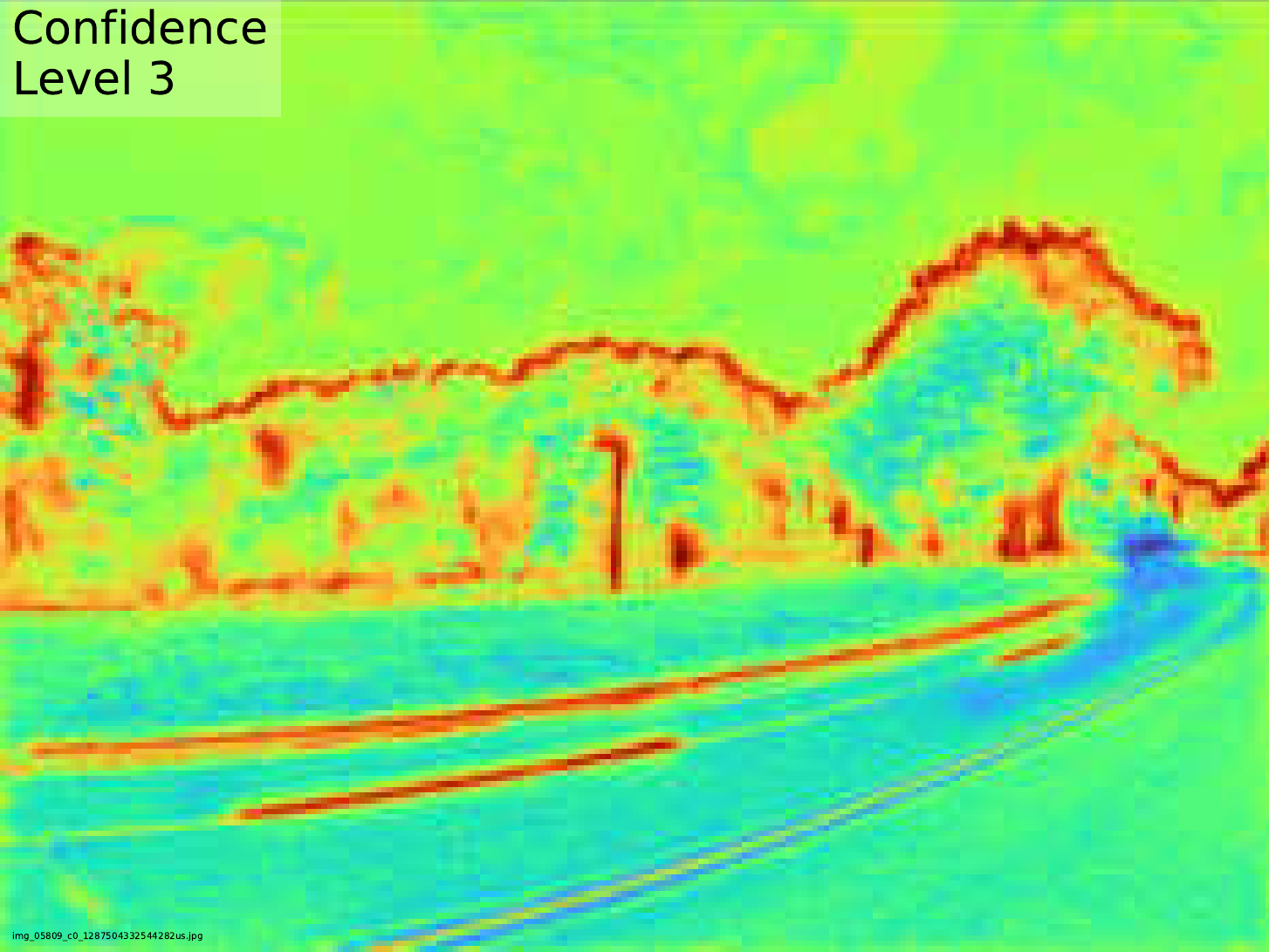}
\end{minipage}
\begin{minipage}{\lwidth\textwidth}
\rotatebox[origin=c]{90}{Reference}
\end{minipage}%
\begin{minipage}{\iwidth\textwidth}
    \centering
    \includegraphics[width=\pwidth\linewidth]{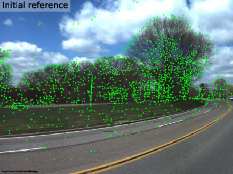}
\end{minipage}%
\begin{minipage}{\iwidth\textwidth}
    \centering
    \includegraphics[width=\pwidth\linewidth]{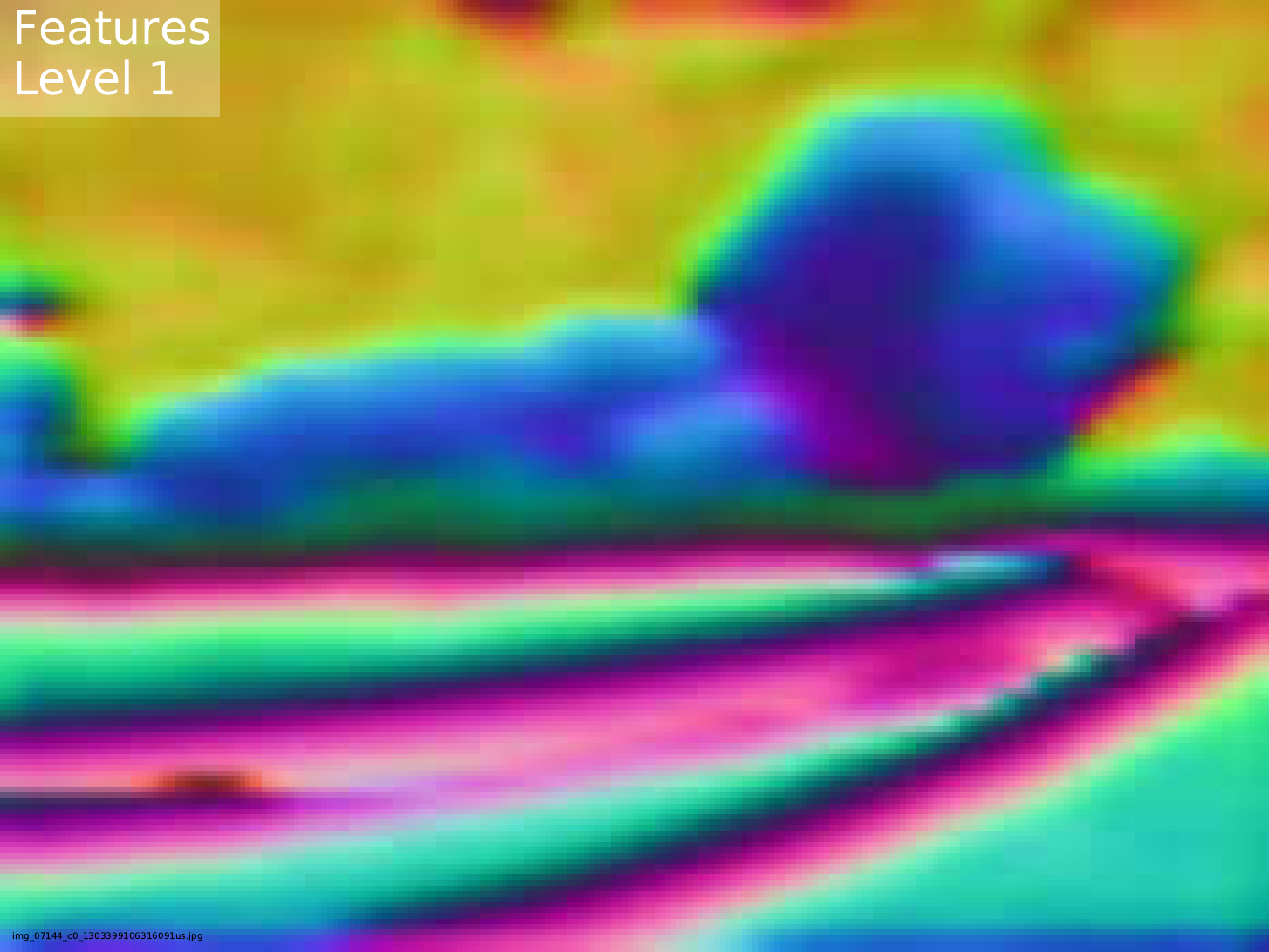}
\end{minipage}%
\begin{minipage}{\iwidth\textwidth}
    \centering
    \includegraphics[width=\pwidth\linewidth]{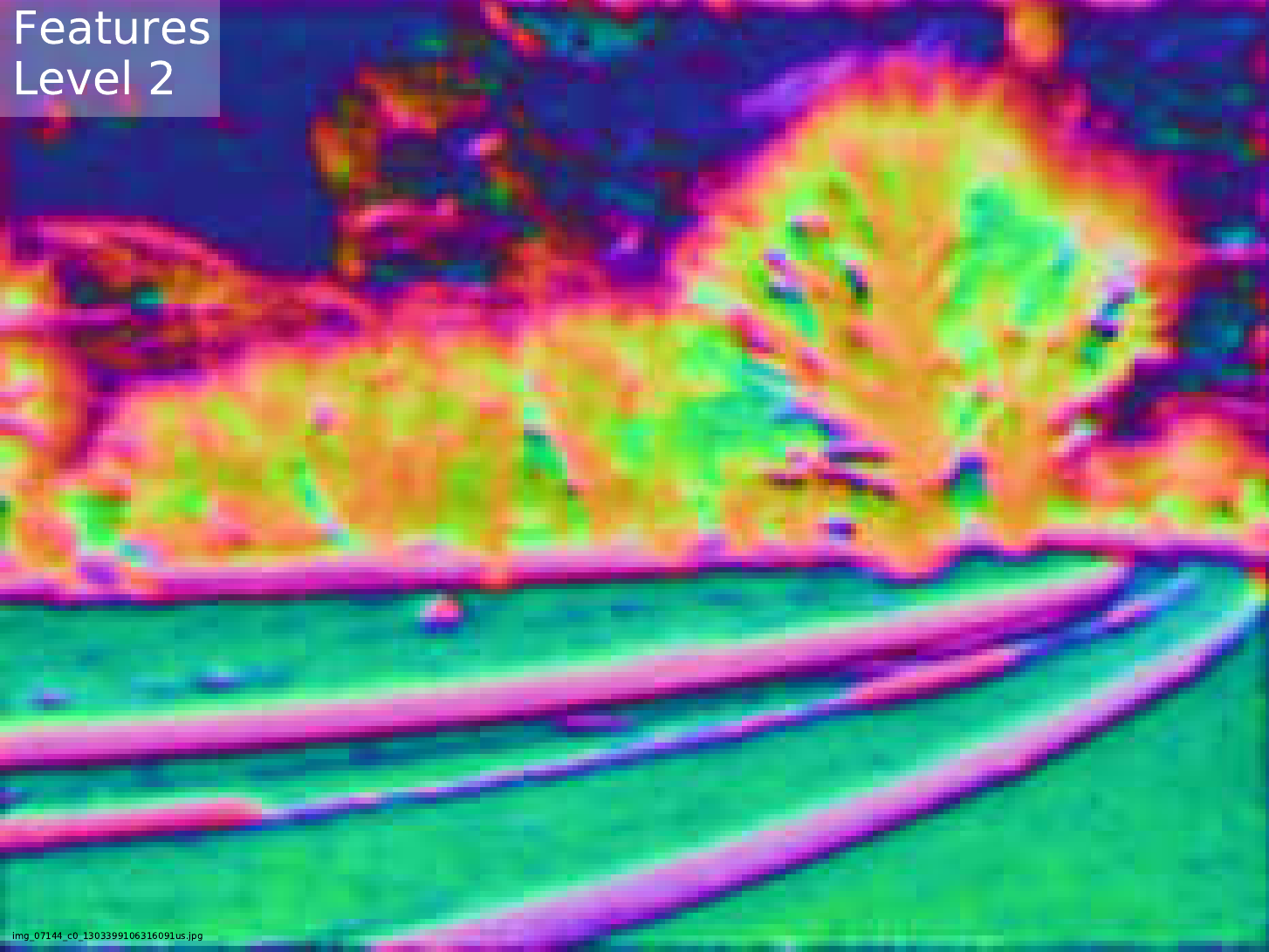}
\end{minipage}%
\begin{minipage}{\iwidth\textwidth}
    \centering
    \includegraphics[width=\pwidth\linewidth]{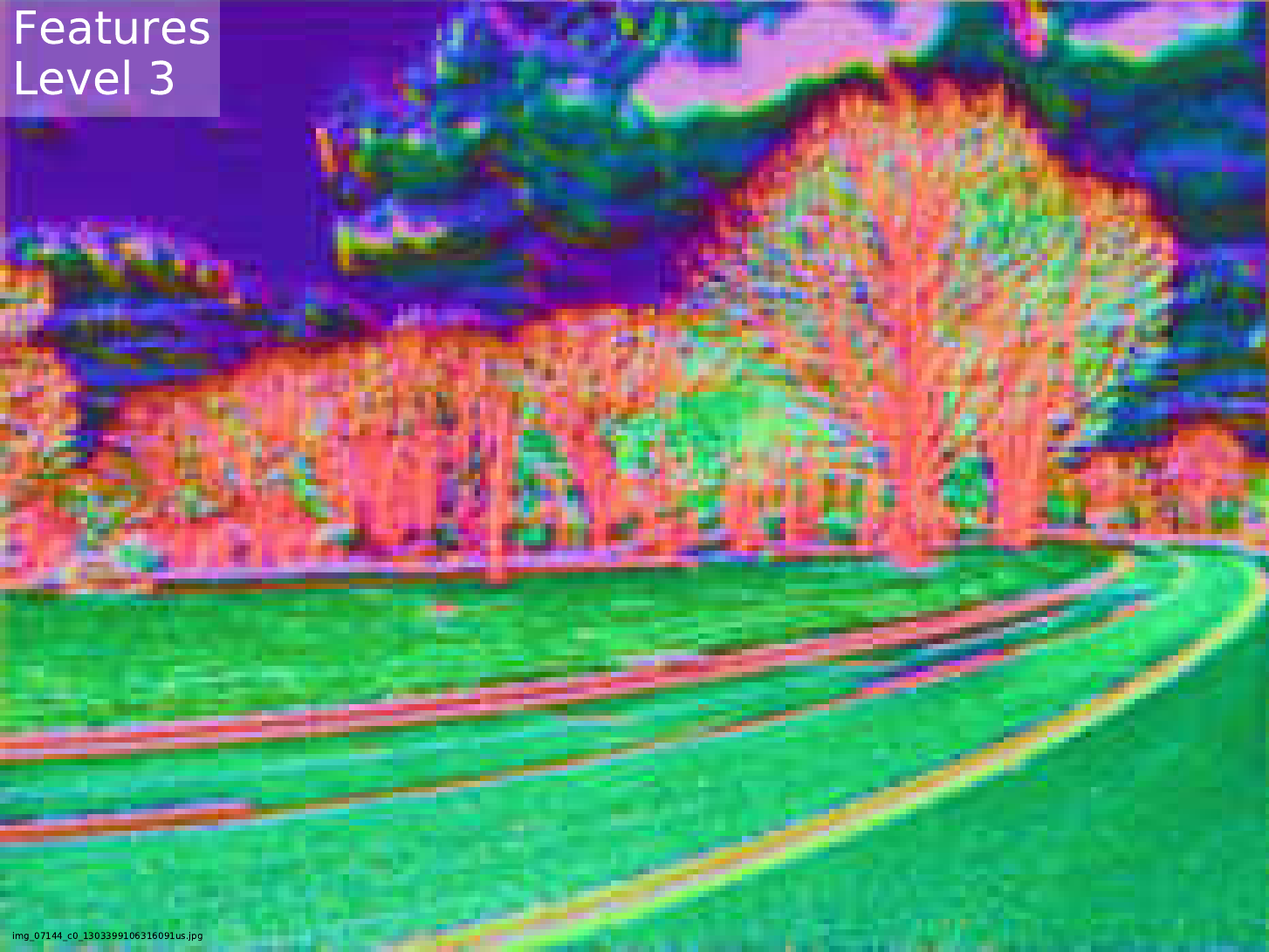}
\end{minipage}%
\begin{minipage}{\iwidth\textwidth}
    \centering
    \includegraphics[width=\pwidth\linewidth]{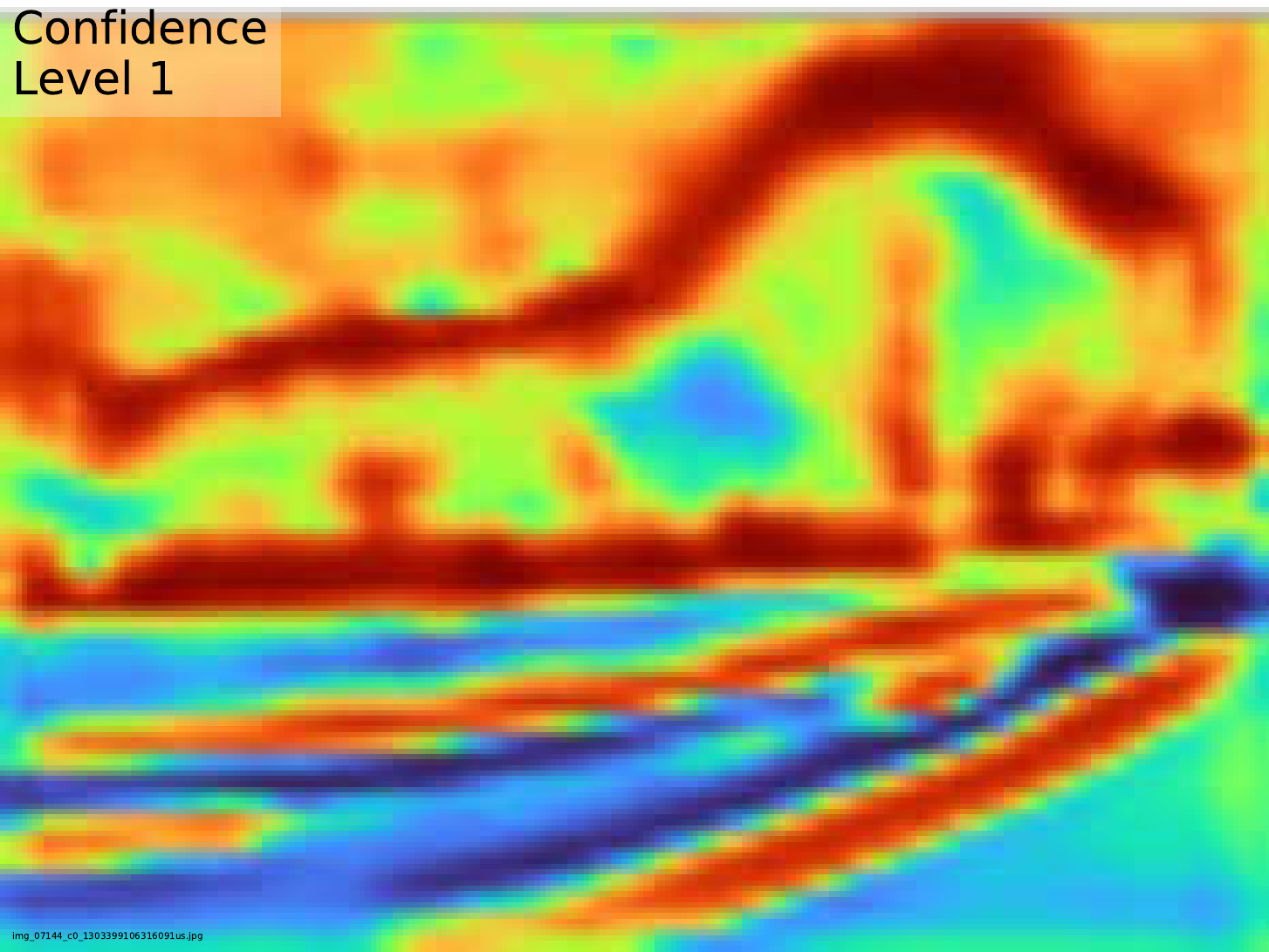}
\end{minipage}%
\begin{minipage}{\iwidth\textwidth}
    \centering
    \includegraphics[width=\pwidth\linewidth]{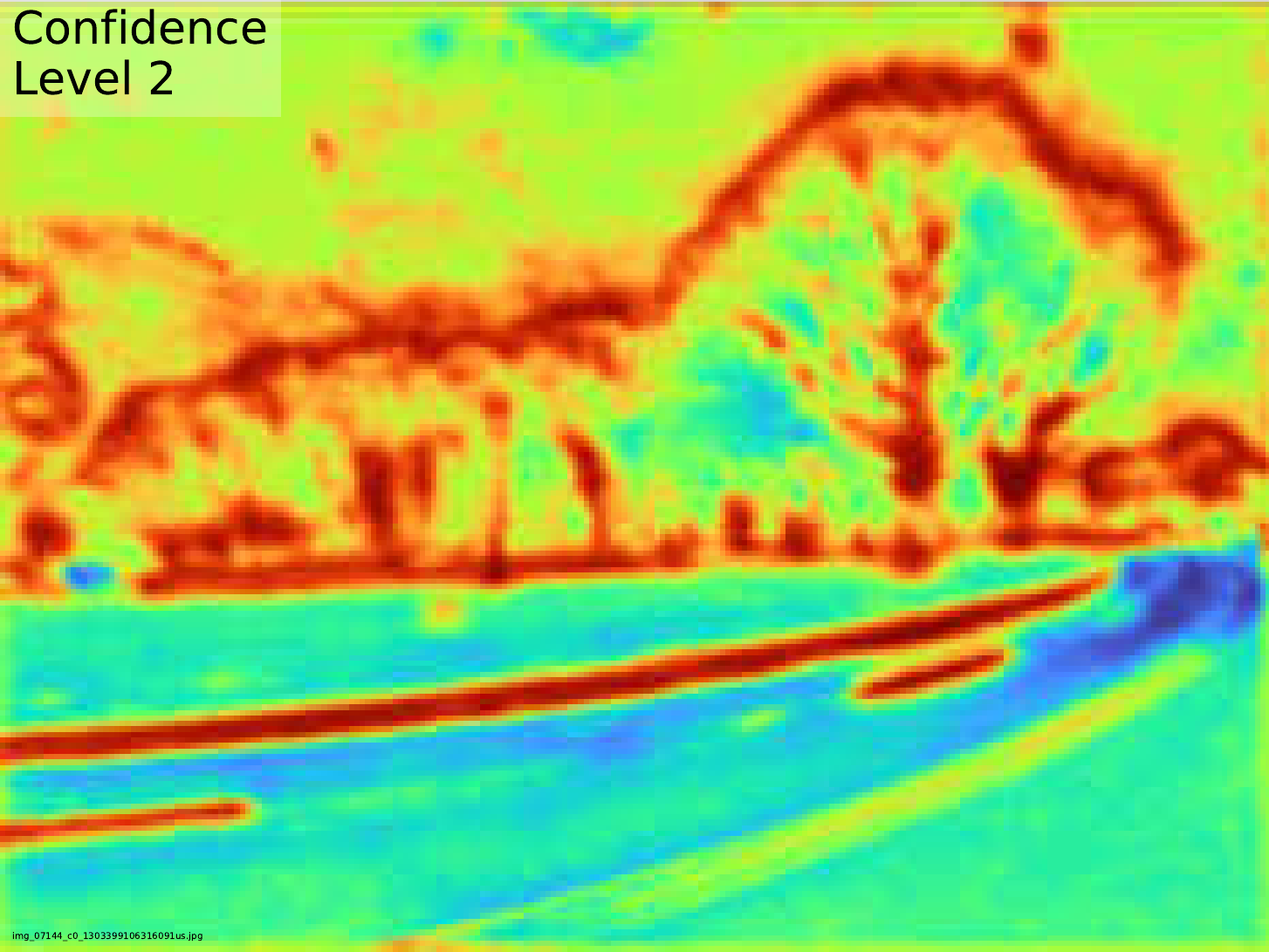}
\end{minipage}%
\begin{minipage}{\iwidth\textwidth}
    \centering
    \includegraphics[width=\pwidth\linewidth]{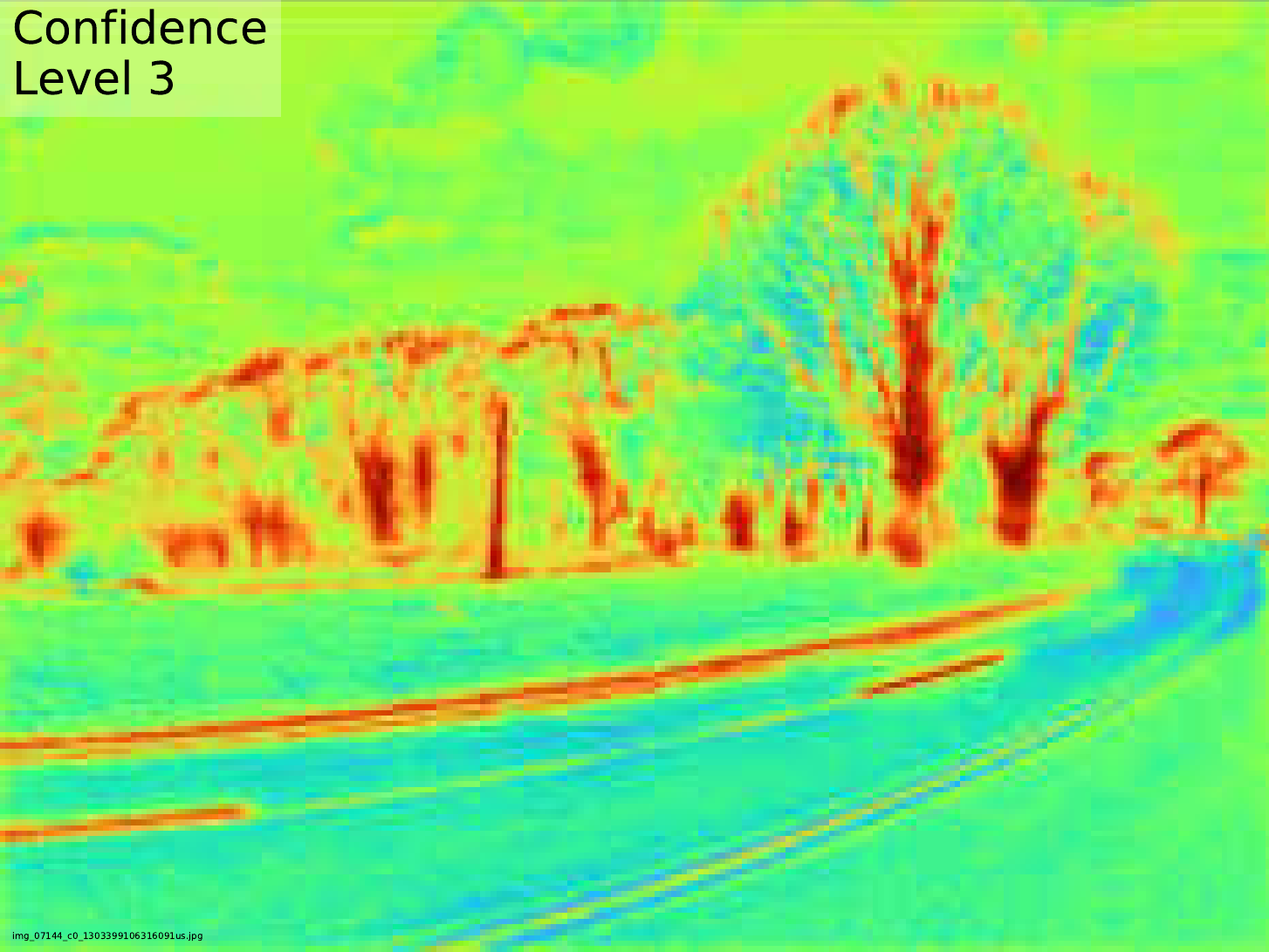}
\end{minipage}
\vspace{2mm}

\begin{minipage}{\lwidth\textwidth}
\rotatebox[origin=c]{90}{Query}
\end{minipage}%
\begin{minipage}{\iwidth\textwidth}
    \centering
    \includegraphics[width=\pwidth\linewidth]{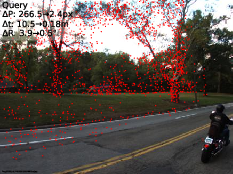}
\end{minipage}%
\begin{minipage}{\iwidth\textwidth}
    \centering
    \includegraphics[width=\pwidth\linewidth]{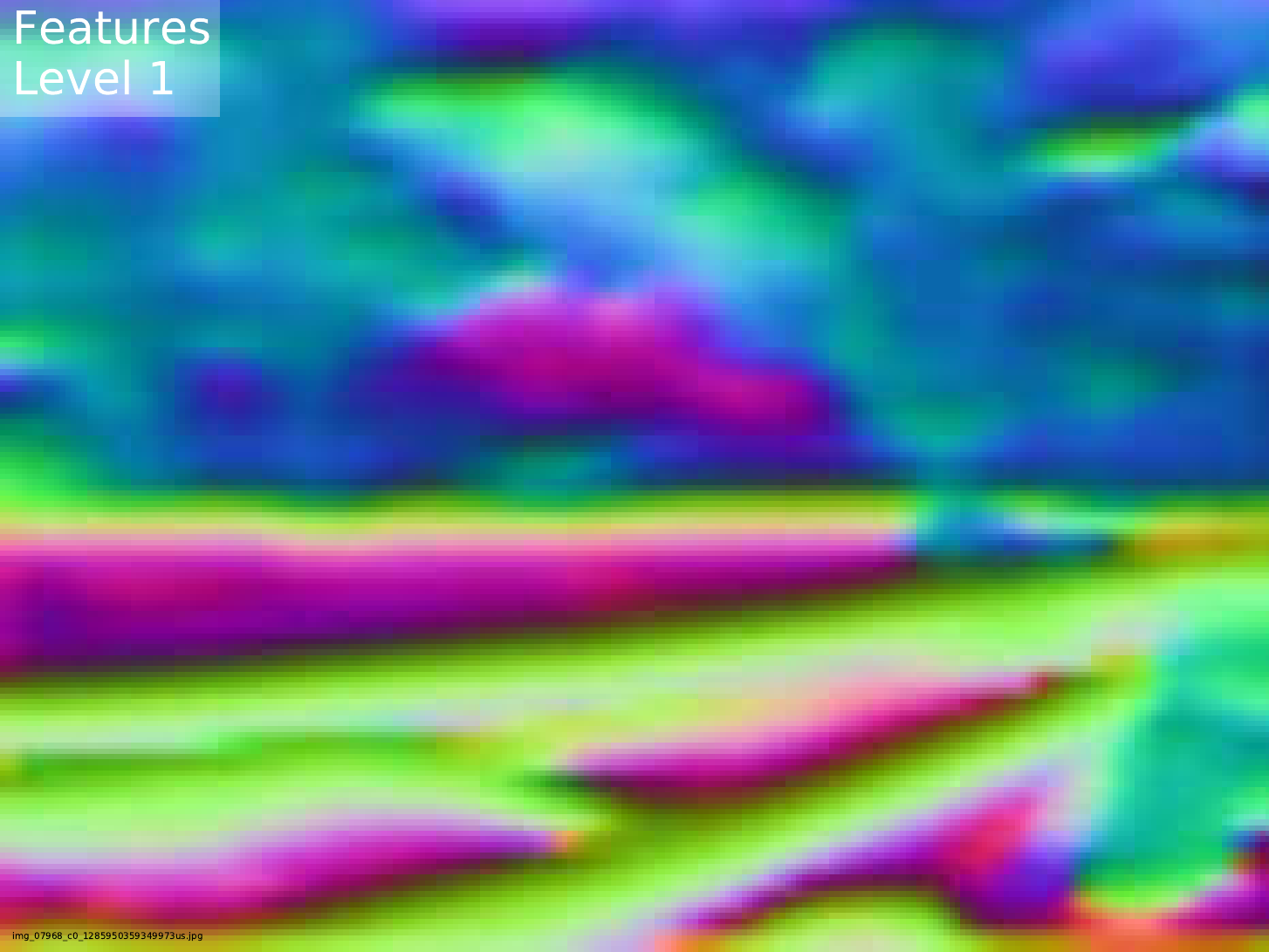}
\end{minipage}%
\begin{minipage}{\iwidth\textwidth}
    \centering
    \includegraphics[width=\pwidth\linewidth]{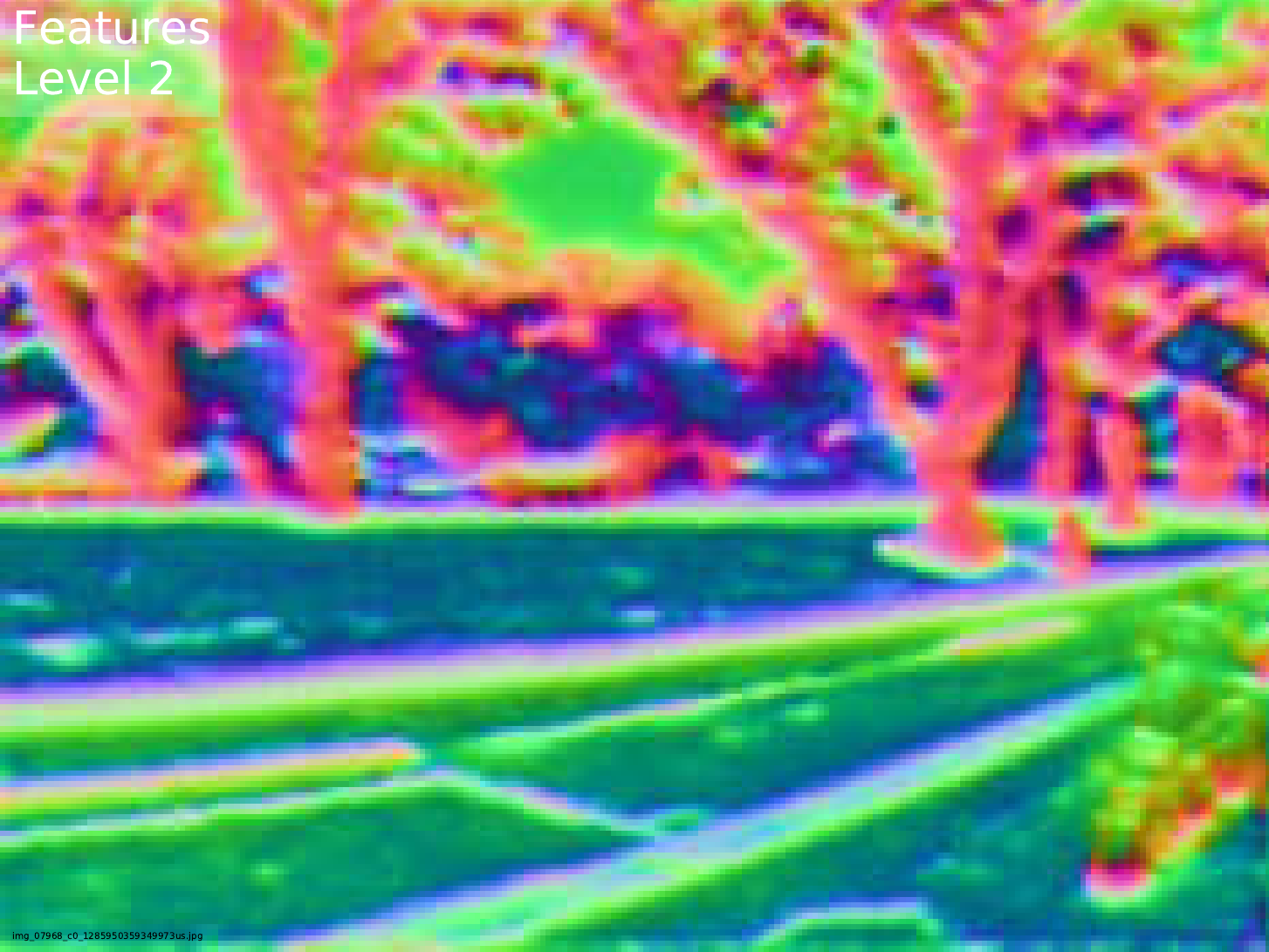}
\end{minipage}%
\begin{minipage}{\iwidth\textwidth}
    \centering
    \includegraphics[width=\pwidth\linewidth]{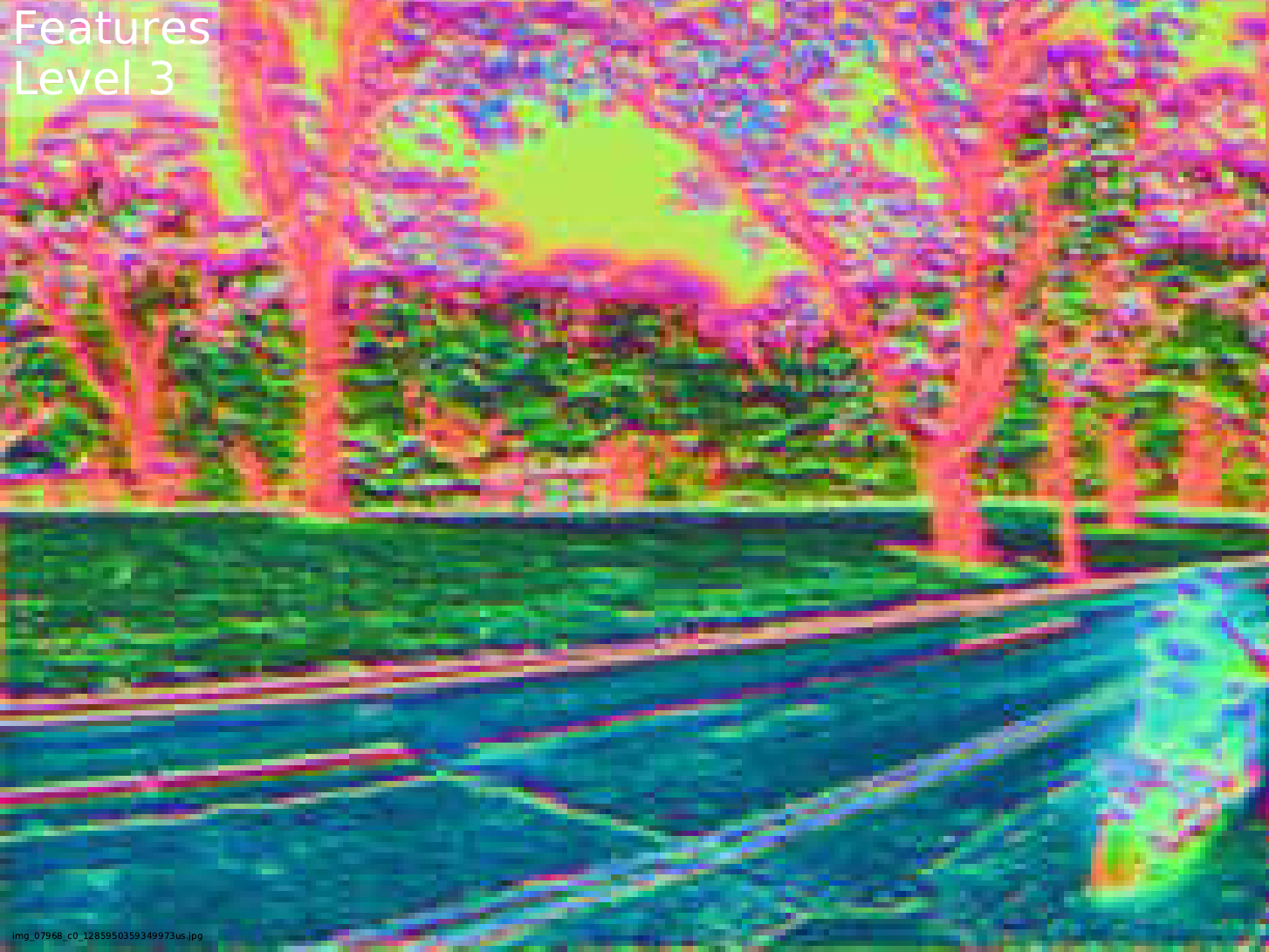}
\end{minipage}%
\begin{minipage}{\iwidth\textwidth}
    \centering
    \includegraphics[width=\pwidth\linewidth]{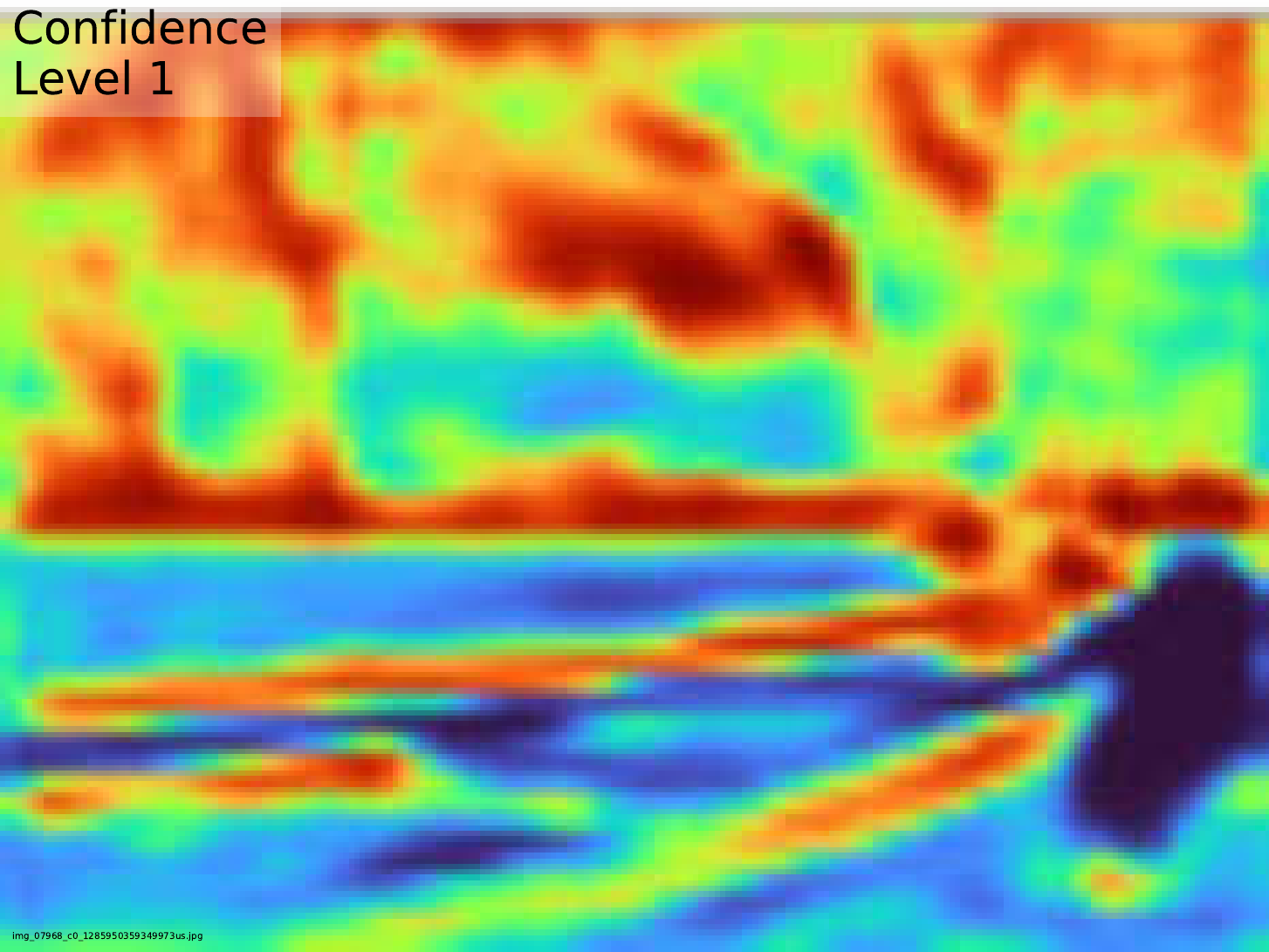}
\end{minipage}%
\begin{minipage}{\iwidth\textwidth}
    \centering
    \includegraphics[width=\pwidth\linewidth]{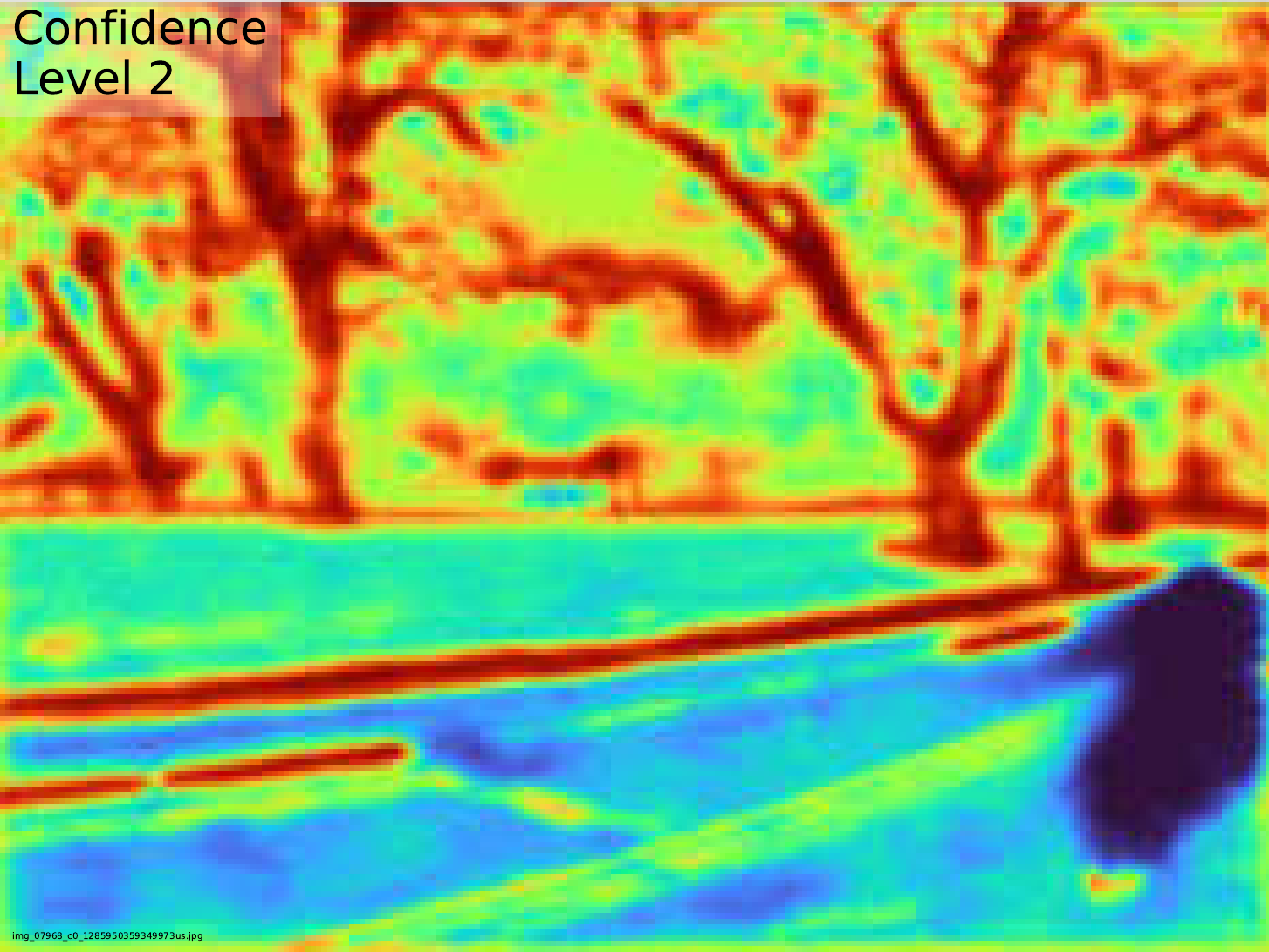}
\end{minipage}%
\begin{minipage}{\iwidth\textwidth}
    \centering
    \includegraphics[width=\pwidth\linewidth]{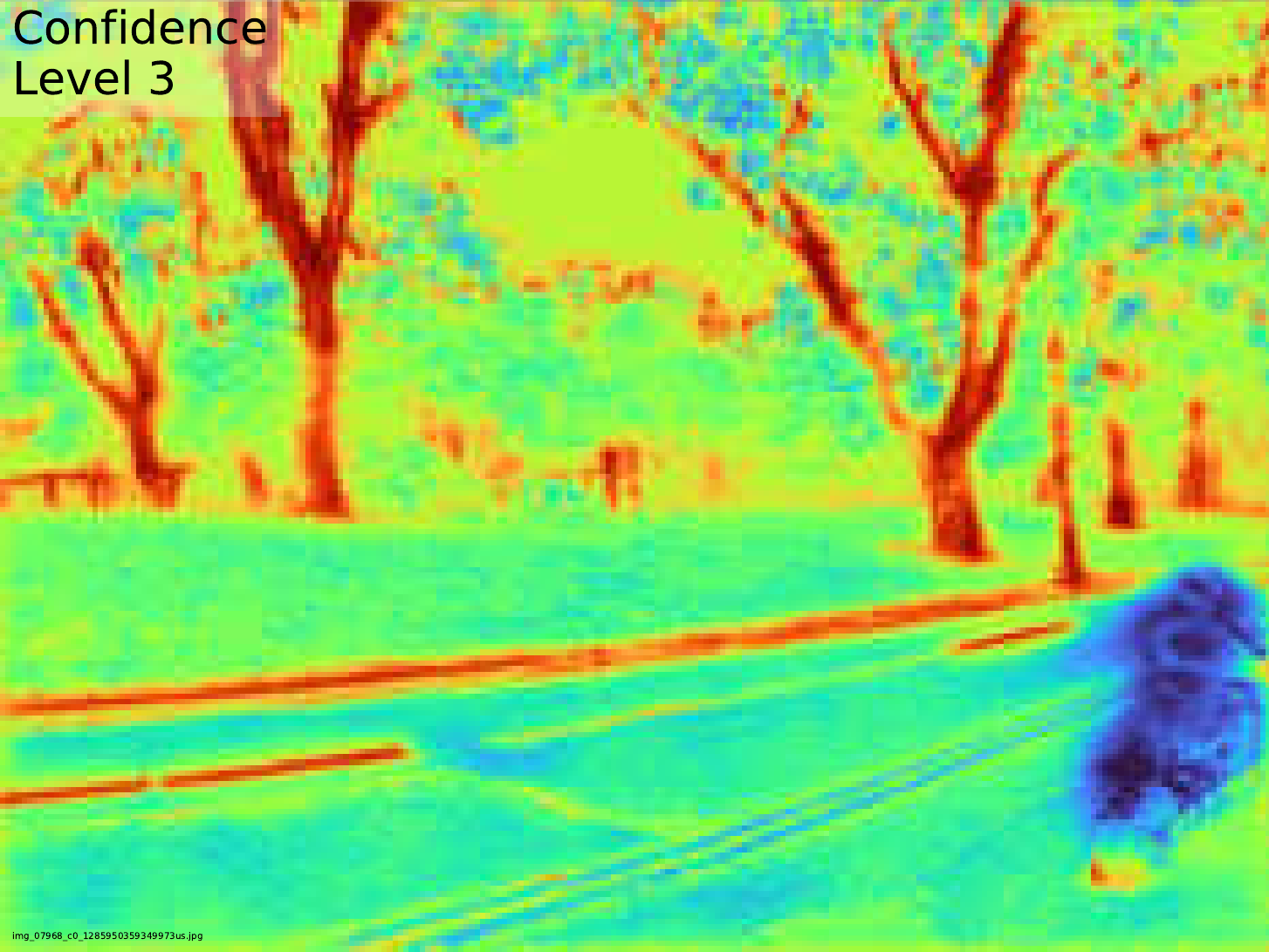}
\end{minipage}
\begin{minipage}{\lwidth\textwidth}
\rotatebox[origin=c]{90}{Reference}
\end{minipage}%
\begin{minipage}{\iwidth\textwidth}
    \centering
    \includegraphics[width=\pwidth\linewidth]{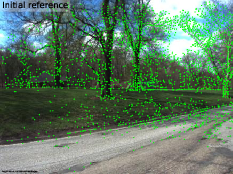}
\end{minipage}%
\begin{minipage}{\iwidth\textwidth}
    \centering
    \includegraphics[width=\pwidth\linewidth]{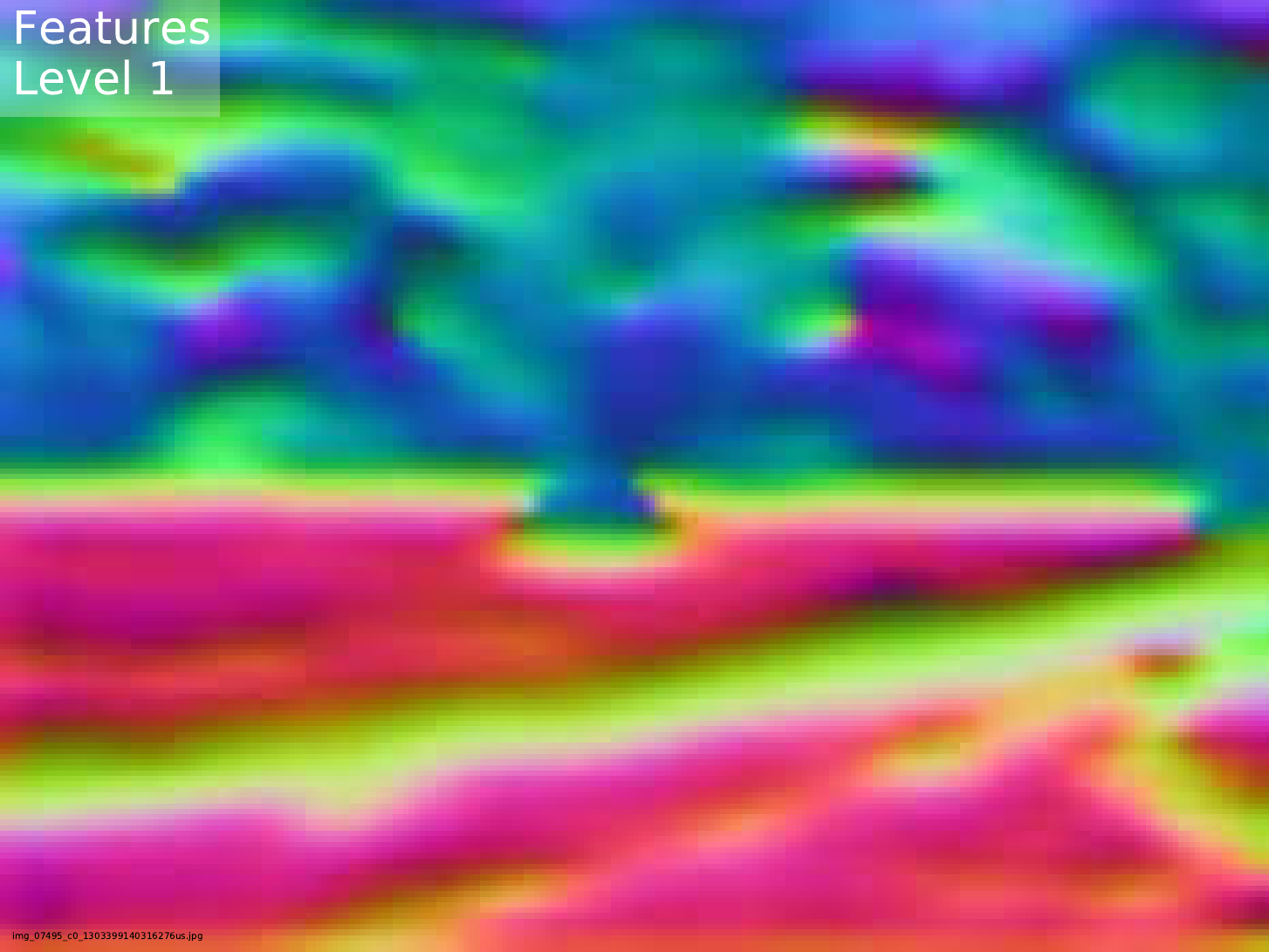}
\end{minipage}%
\begin{minipage}{\iwidth\textwidth}
    \centering
    \includegraphics[width=\pwidth\linewidth]{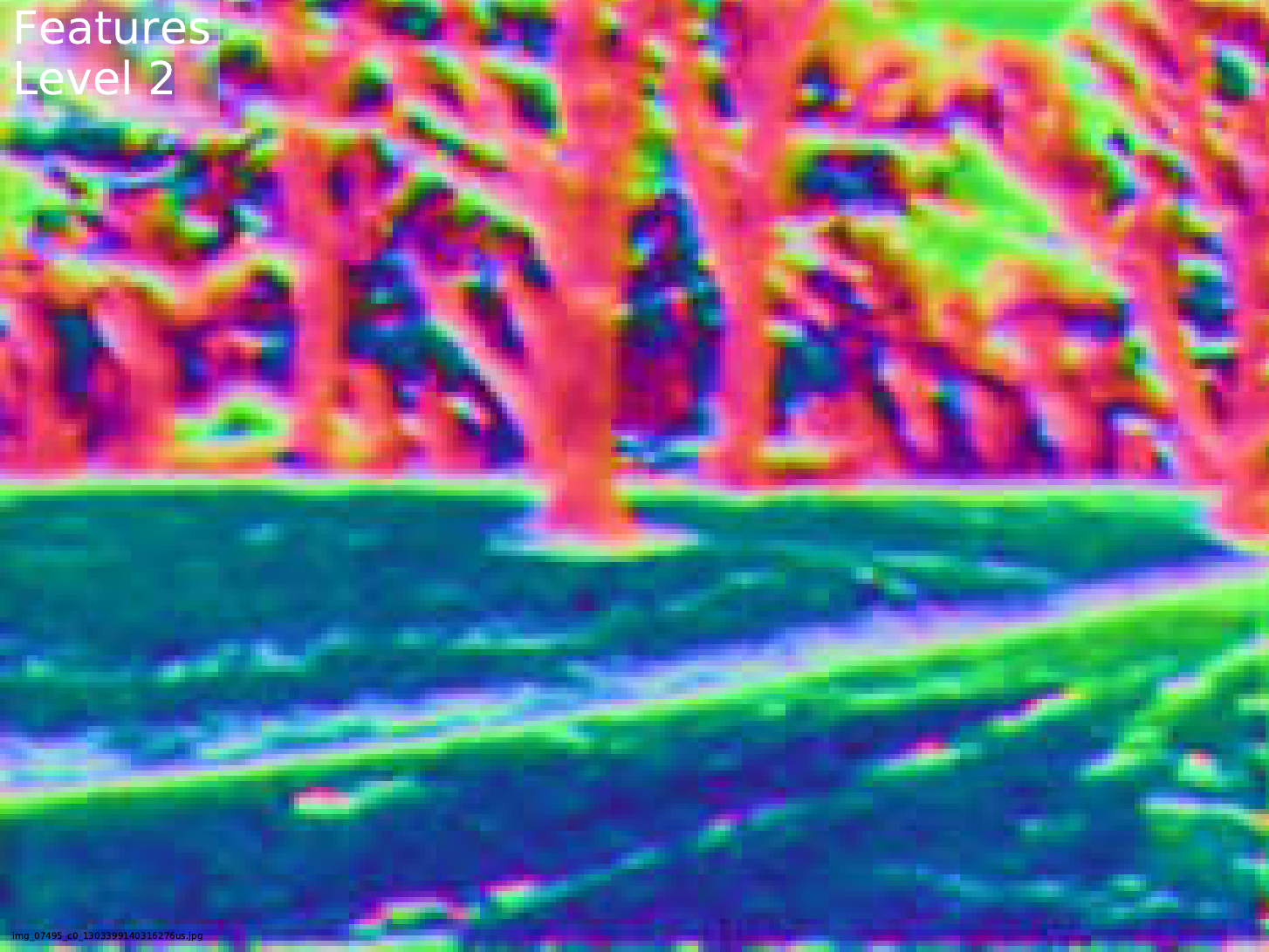}
\end{minipage}%
\begin{minipage}{\iwidth\textwidth}
    \centering
    \includegraphics[width=\pwidth\linewidth]{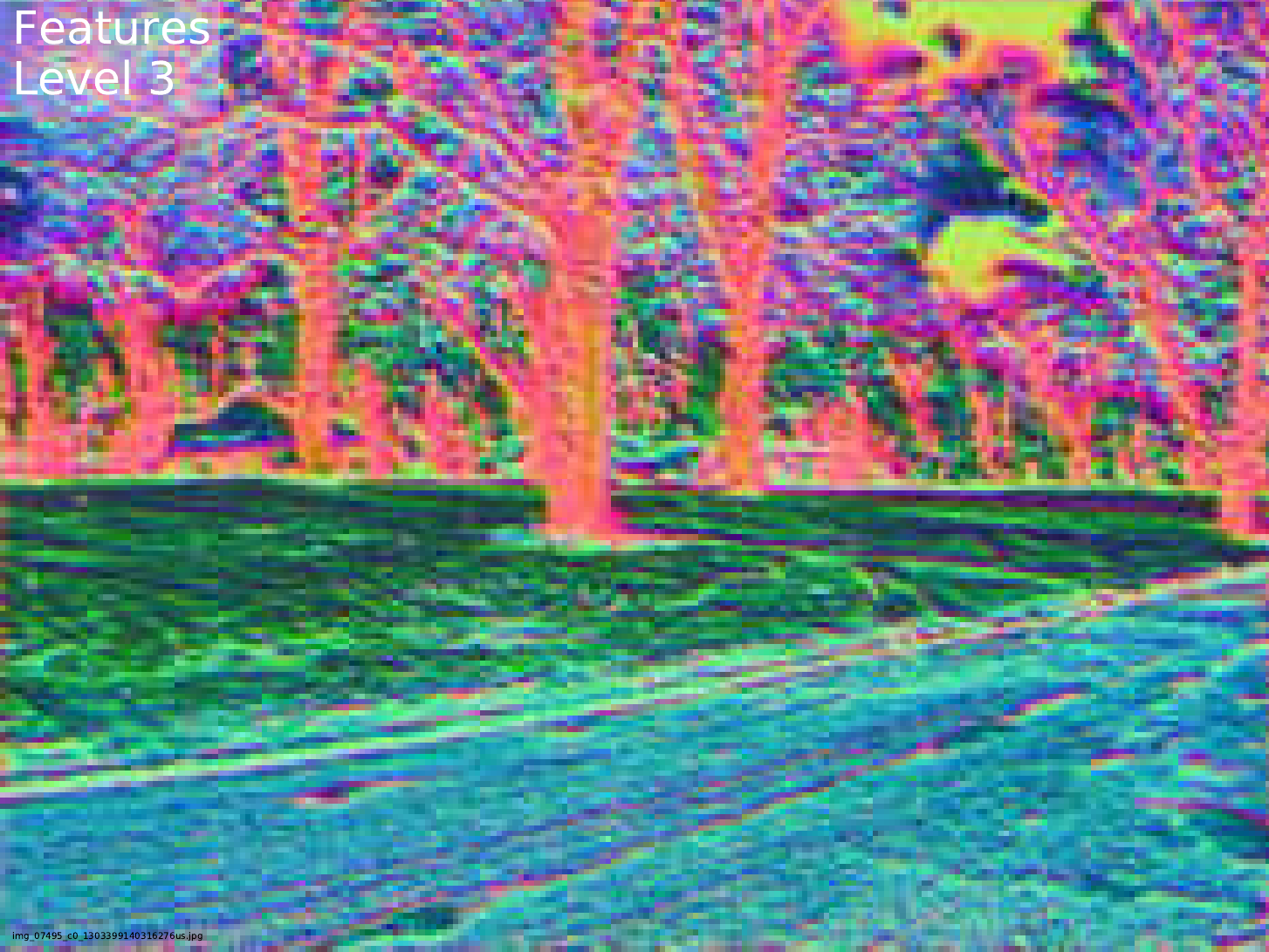}
\end{minipage}%
\begin{minipage}{\iwidth\textwidth}
    \centering
    \includegraphics[width=\pwidth\linewidth]{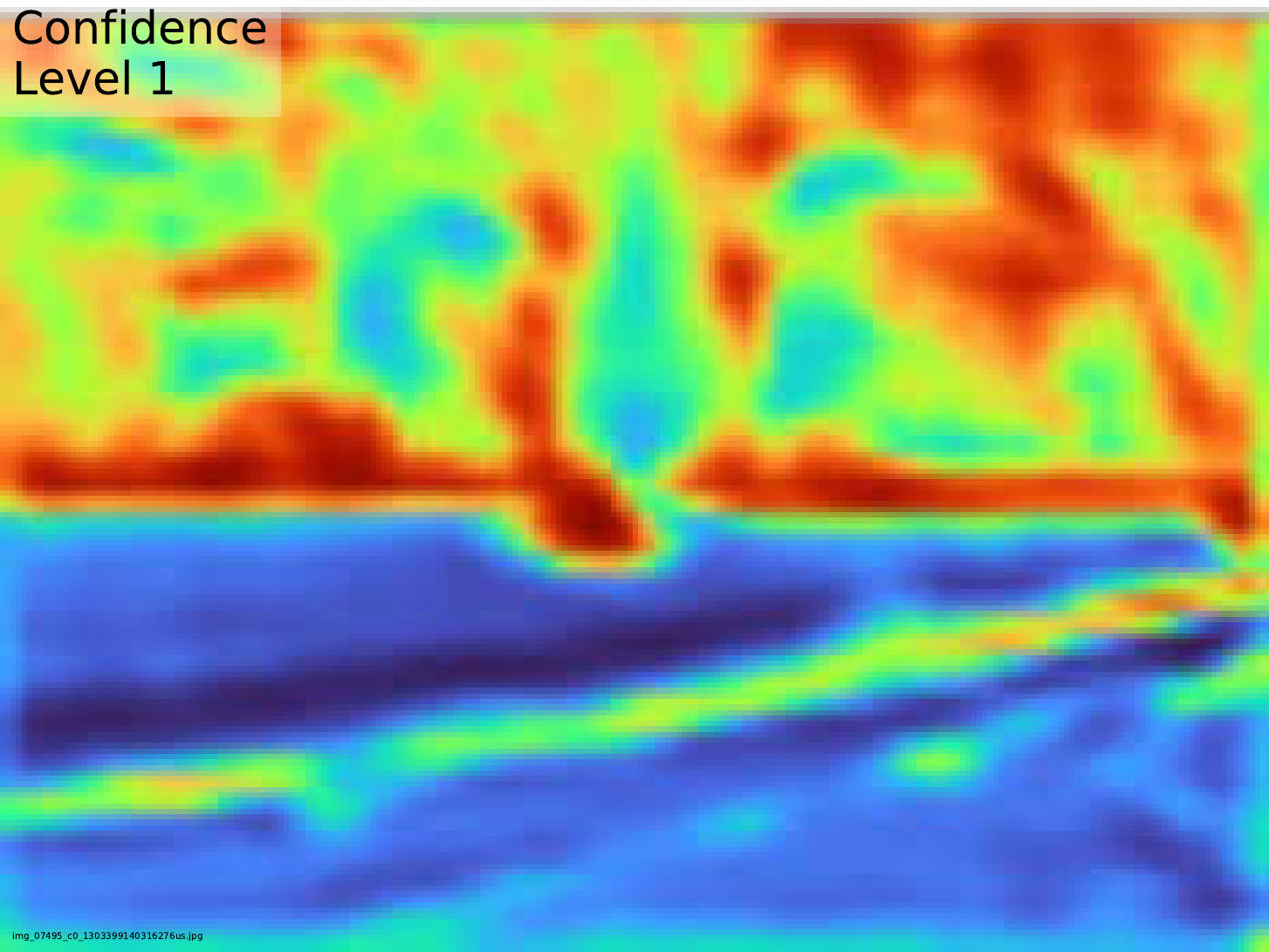}
\end{minipage}%
\begin{minipage}{\iwidth\textwidth}
    \centering
    \includegraphics[width=\pwidth\linewidth]{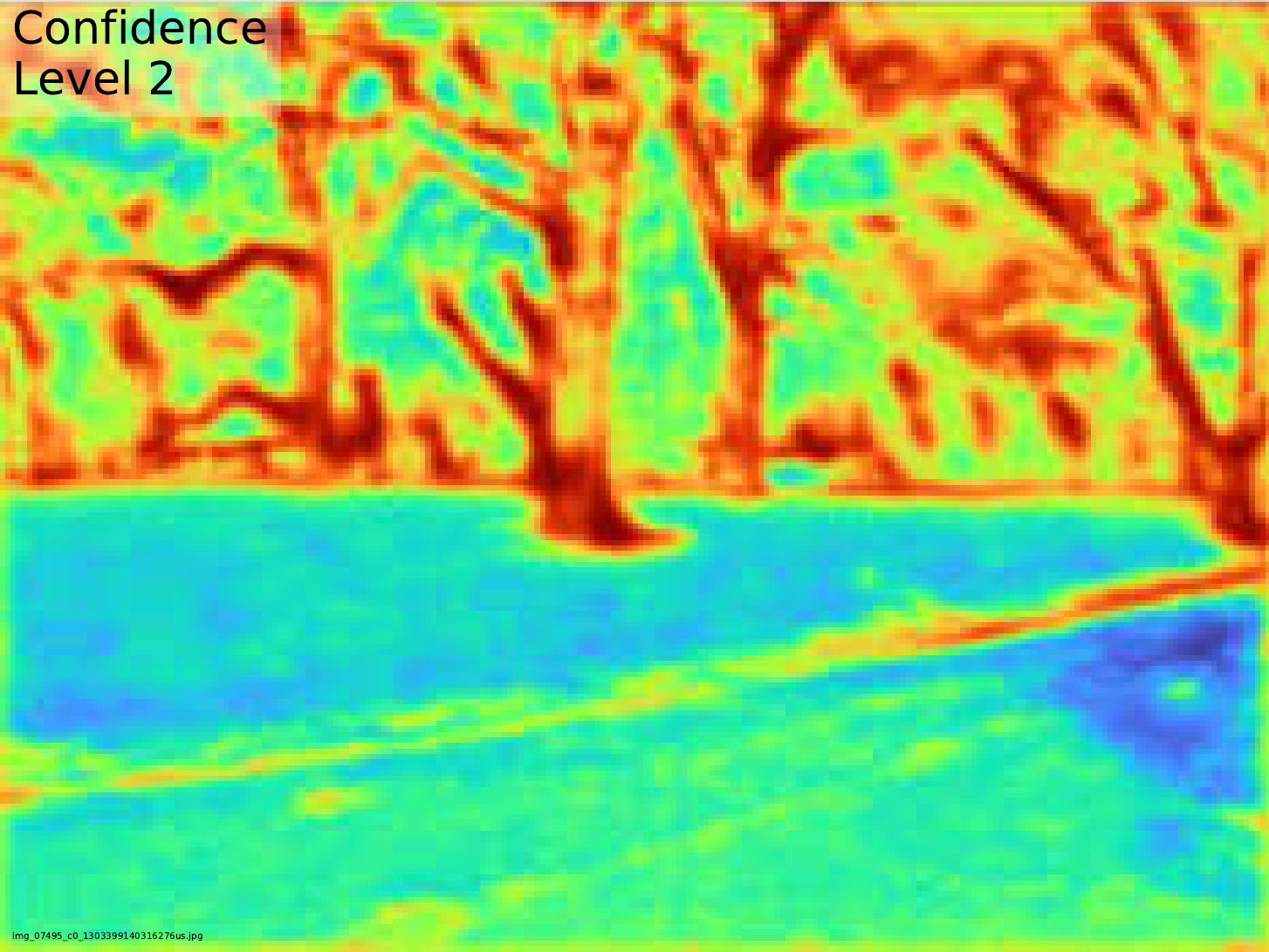}
\end{minipage}%
\begin{minipage}{\iwidth\textwidth}
    \centering
    \includegraphics[width=\pwidth\linewidth]{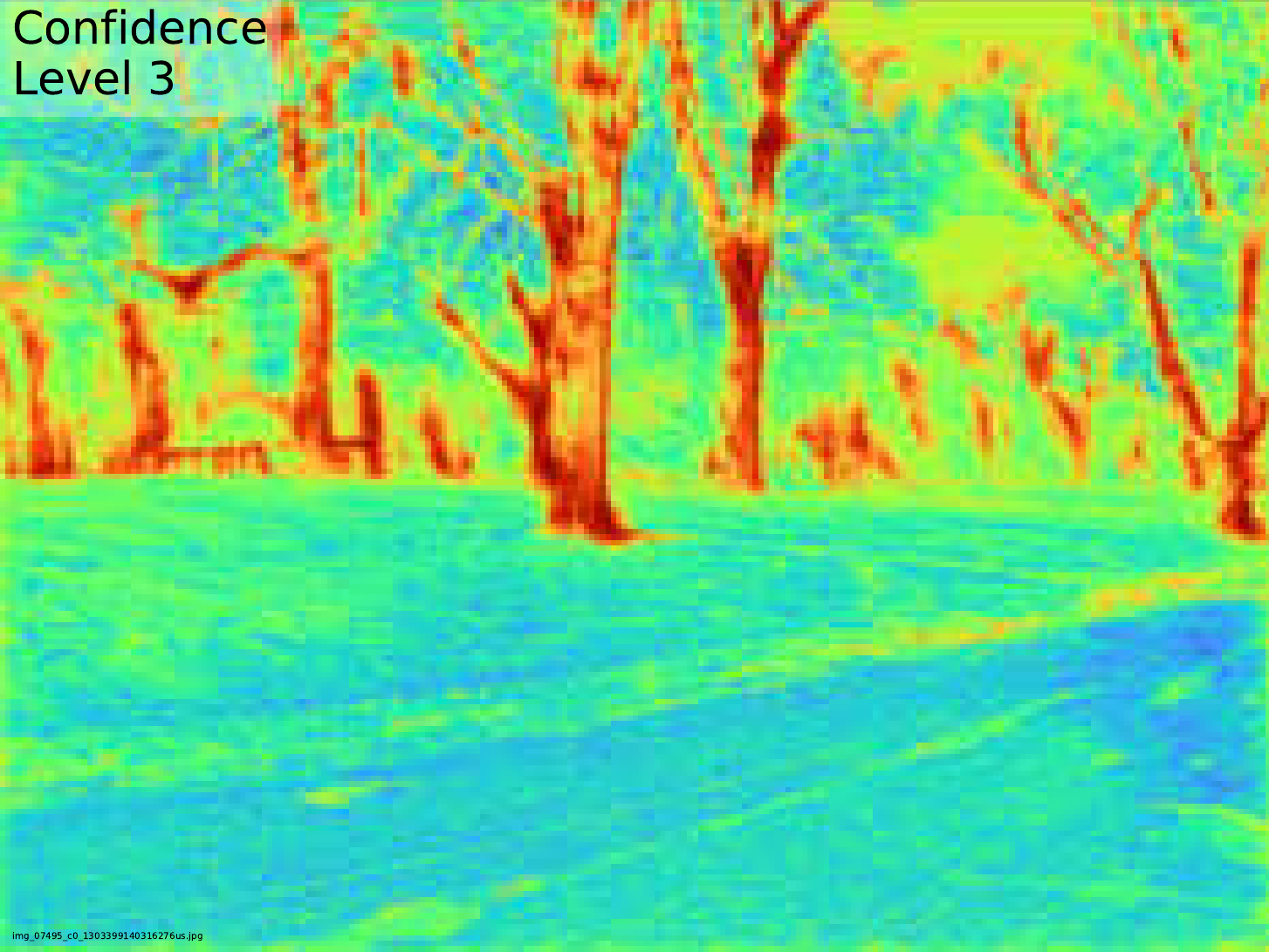}
\end{minipage}
\vspace{2mm}

    \caption{\textbf{Successful localization on the CMU dataset.}
    We show 5 challenging queries with large initial errors and large cross-season appearance changes that are successfully localized by PixLoc.
    We project 3D SfM points into the initial reference image (in \green{green}) and into the query image using the estimated pose (in \red{red}).
    We show the features at the 3 different levels, mapping them to RGB using PCA.
    We also show the confidence maps, where blue pixels are ignored while red ones are more important for the optimization.
    Features useful for localization are invariant across seasons and thus appear in similar colors.
    }
    \label{fig:qualitative_cmu_success}%
\end{figure*}

\begin{figure*}[t]
    \centering
\def\iwidth{0.14}
\def\pwidth{0.99}
\def\lwidth{0.020}
\def\rcwidth{0.14}

\begin{minipage}{\lwidth\textwidth}
\hfill
\end{minipage}%
\begin{minipage}{\iwidth\textwidth}
    \centering
    \small{Images}
\end{minipage}%
\begin{minipage}{\iwidth\textwidth*\real{3.0}}
    \centering
    \small{Features}
    \vspace{0.5mm}
    \hrule width 0.99\linewidth
    \vspace{0.2mm}
\end{minipage}%
\begin{minipage}{\iwidth\textwidth*\real{3.0}}
    \centering
    \small{Confidence}
    \vspace{0.5mm}
    \hrule width 0.99\linewidth
    \vspace{0.2mm}
\end{minipage}%

\begin{minipage}{\lwidth\textwidth}
\rotatebox[origin=c]{90}{Query}
\end{minipage}%
\begin{minipage}{\iwidth\textwidth}
    \centering
    \includegraphics[width=\pwidth\linewidth]{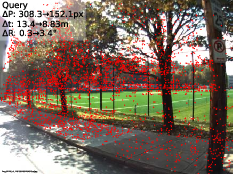}
\end{minipage}%
\begin{minipage}{\iwidth\textwidth}
    \centering
    \includegraphics[width=\pwidth\linewidth]{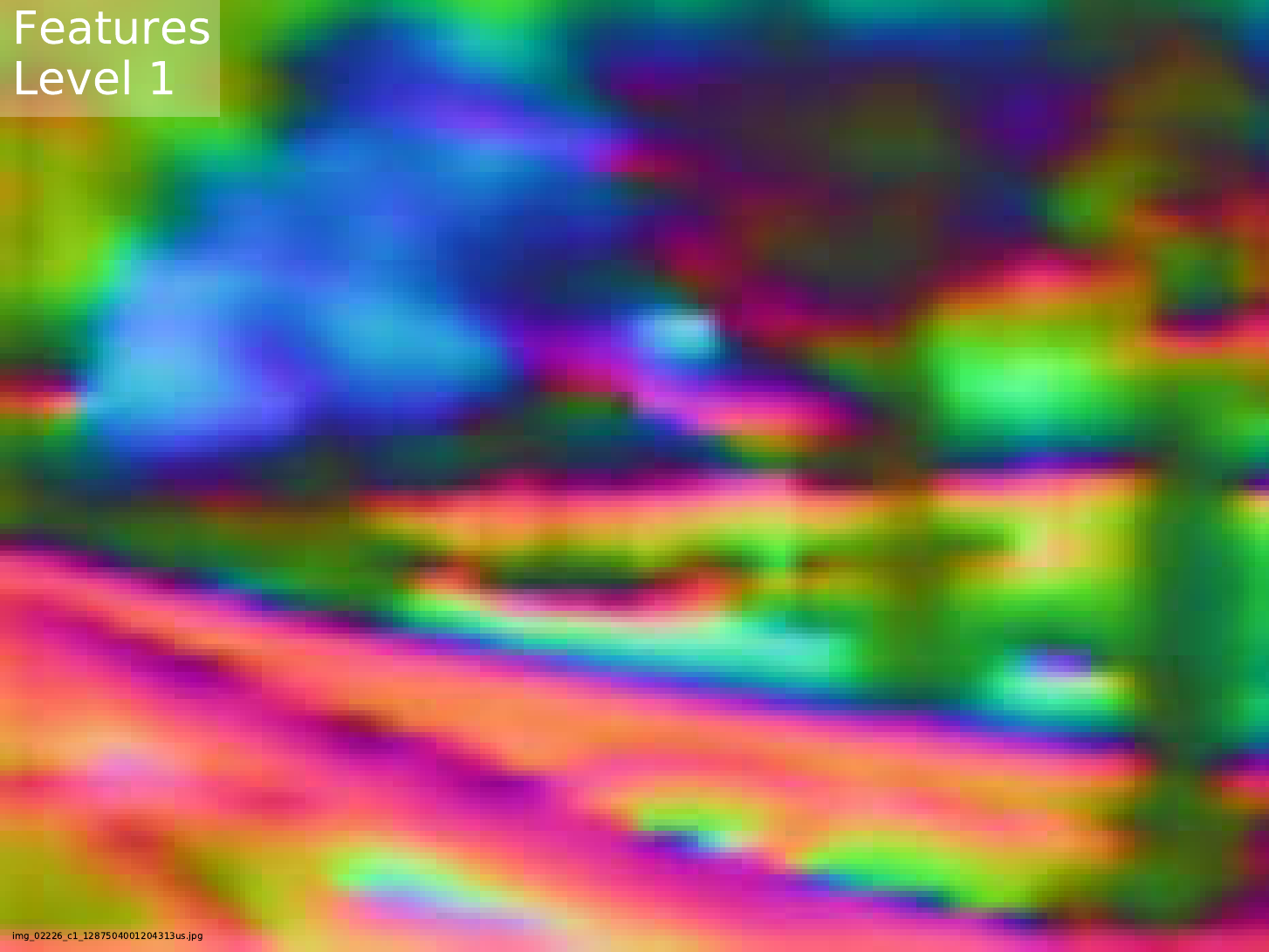}
\end{minipage}%
\begin{minipage}{\iwidth\textwidth}
    \centering
    \includegraphics[width=\pwidth\linewidth]{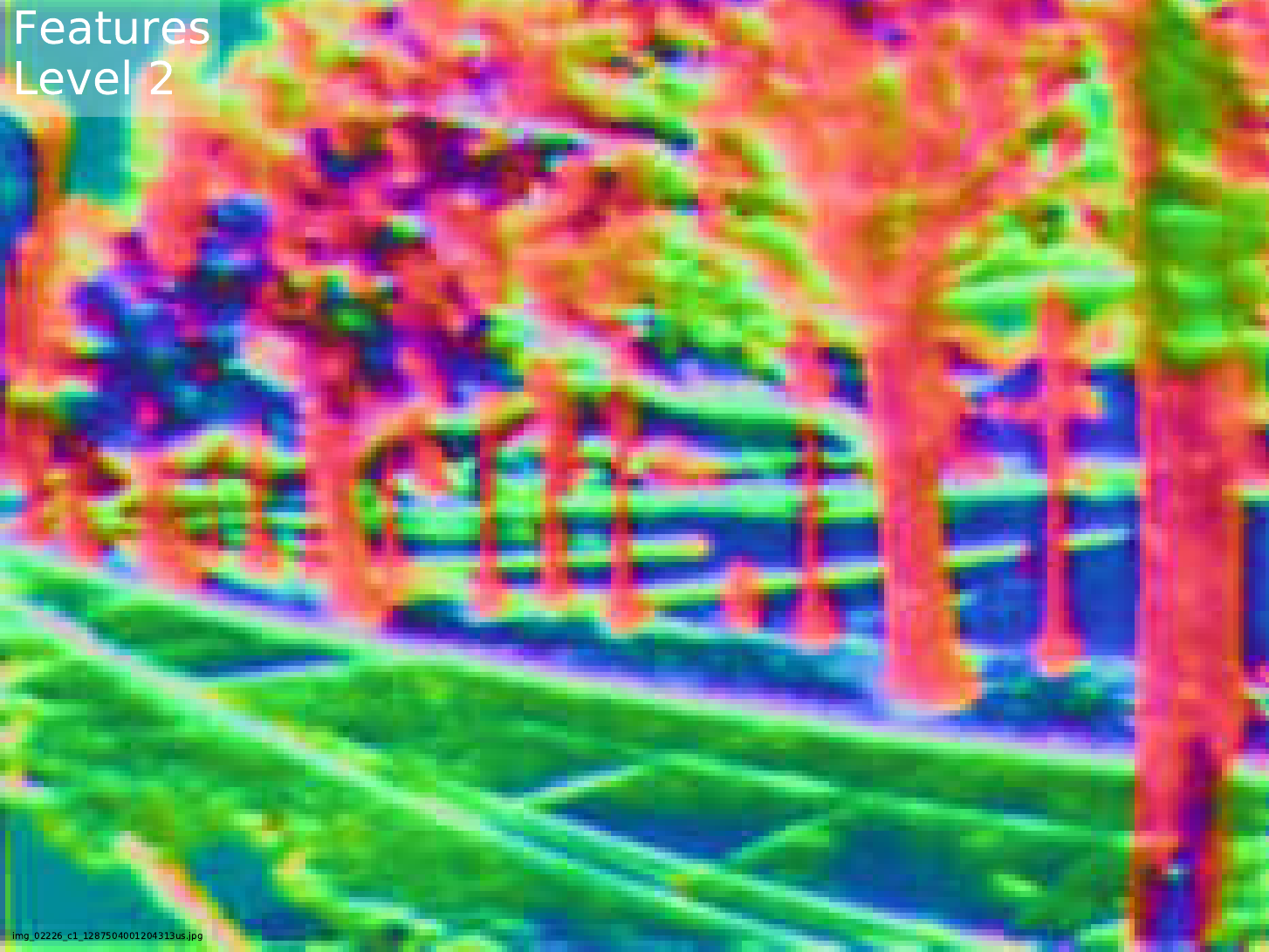}
\end{minipage}%
\begin{minipage}{\iwidth\textwidth}
    \centering
    \includegraphics[width=\pwidth\linewidth]{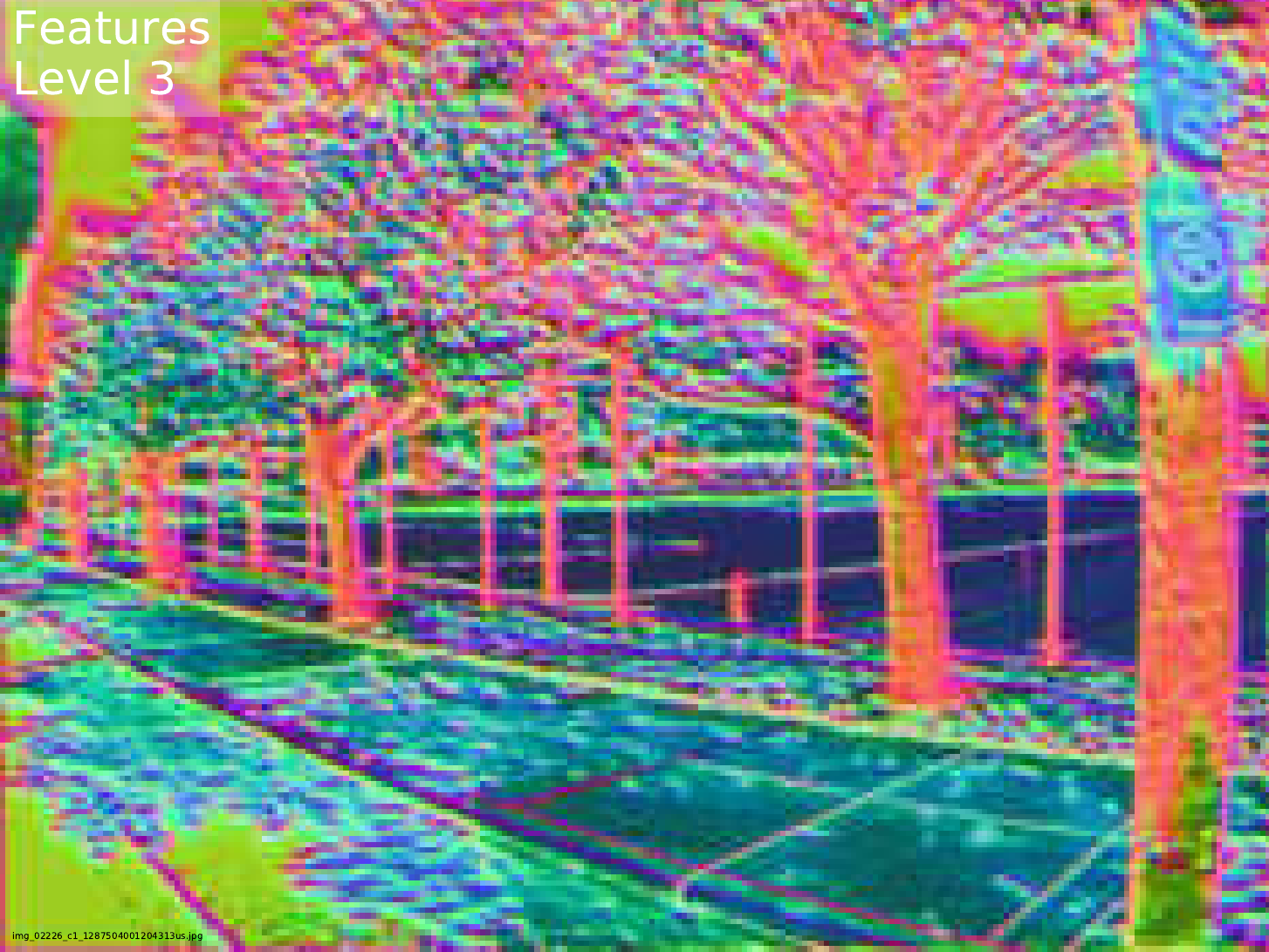}
\end{minipage}%
\begin{minipage}{\iwidth\textwidth}
    \centering
    \includegraphics[width=\pwidth\linewidth]{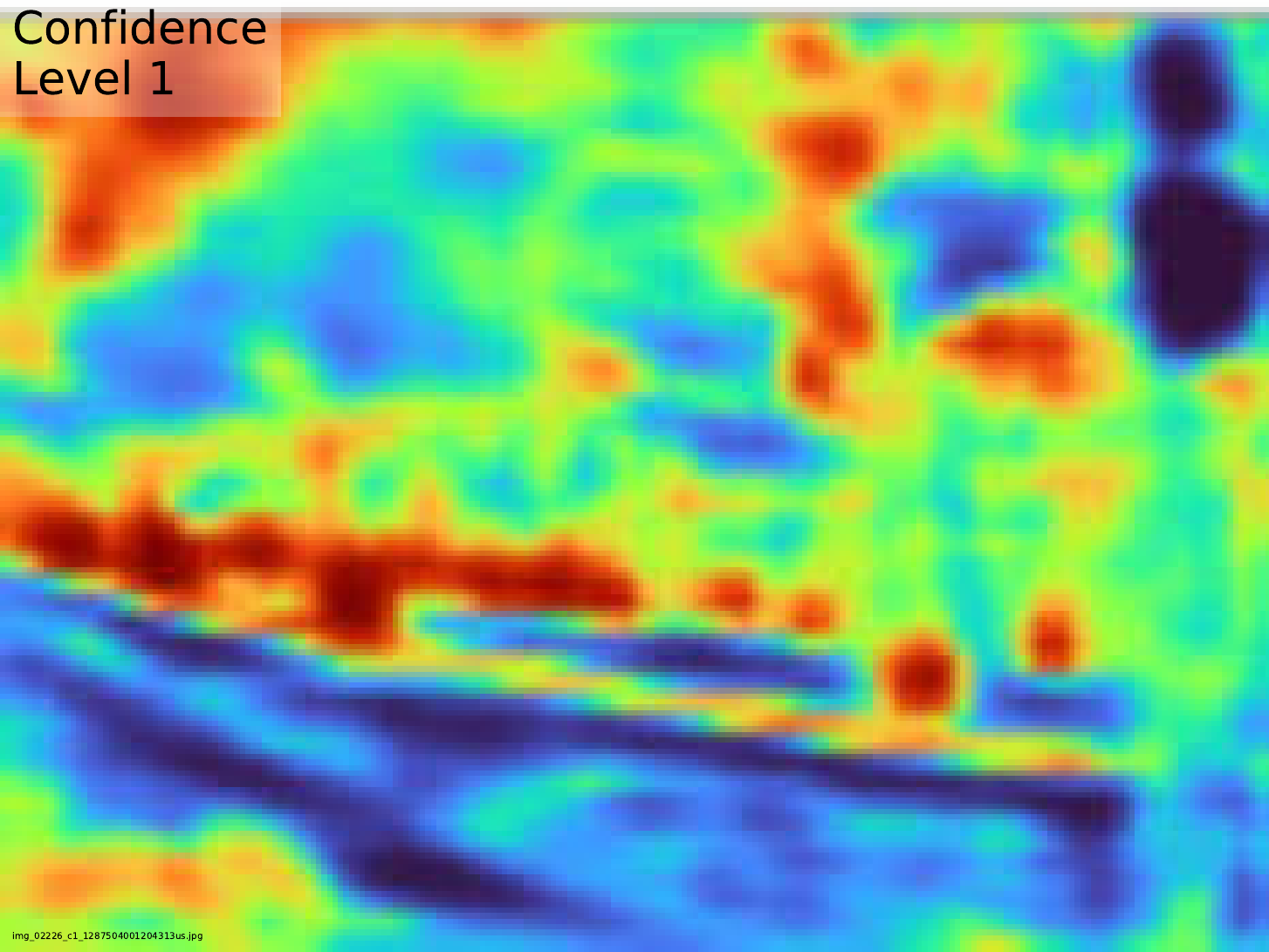}
\end{minipage}%
\begin{minipage}{\iwidth\textwidth}
    \centering
    \includegraphics[width=\pwidth\linewidth]{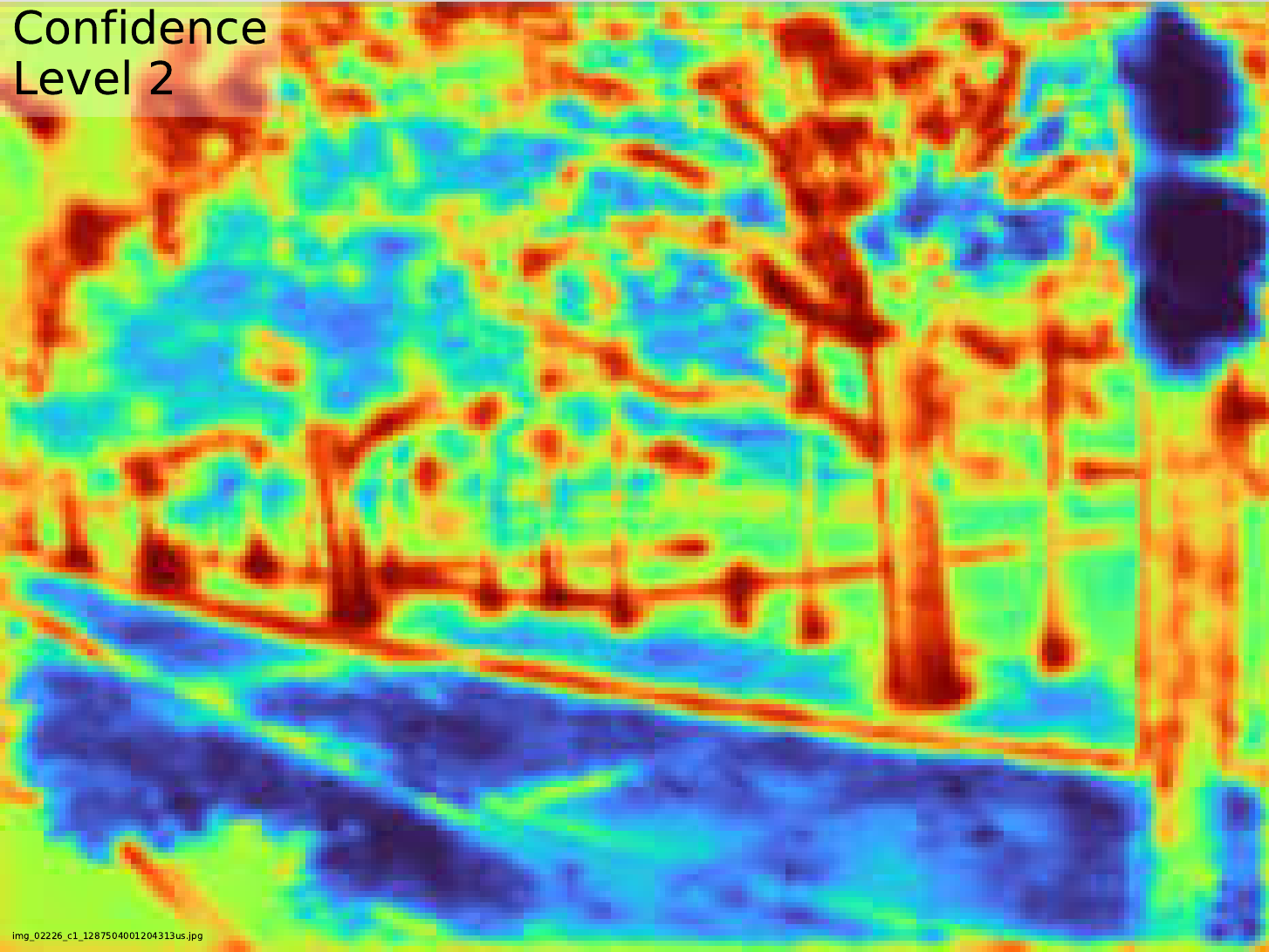}
\end{minipage}%
\begin{minipage}{\iwidth\textwidth}
    \centering
    \includegraphics[width=\pwidth\linewidth]{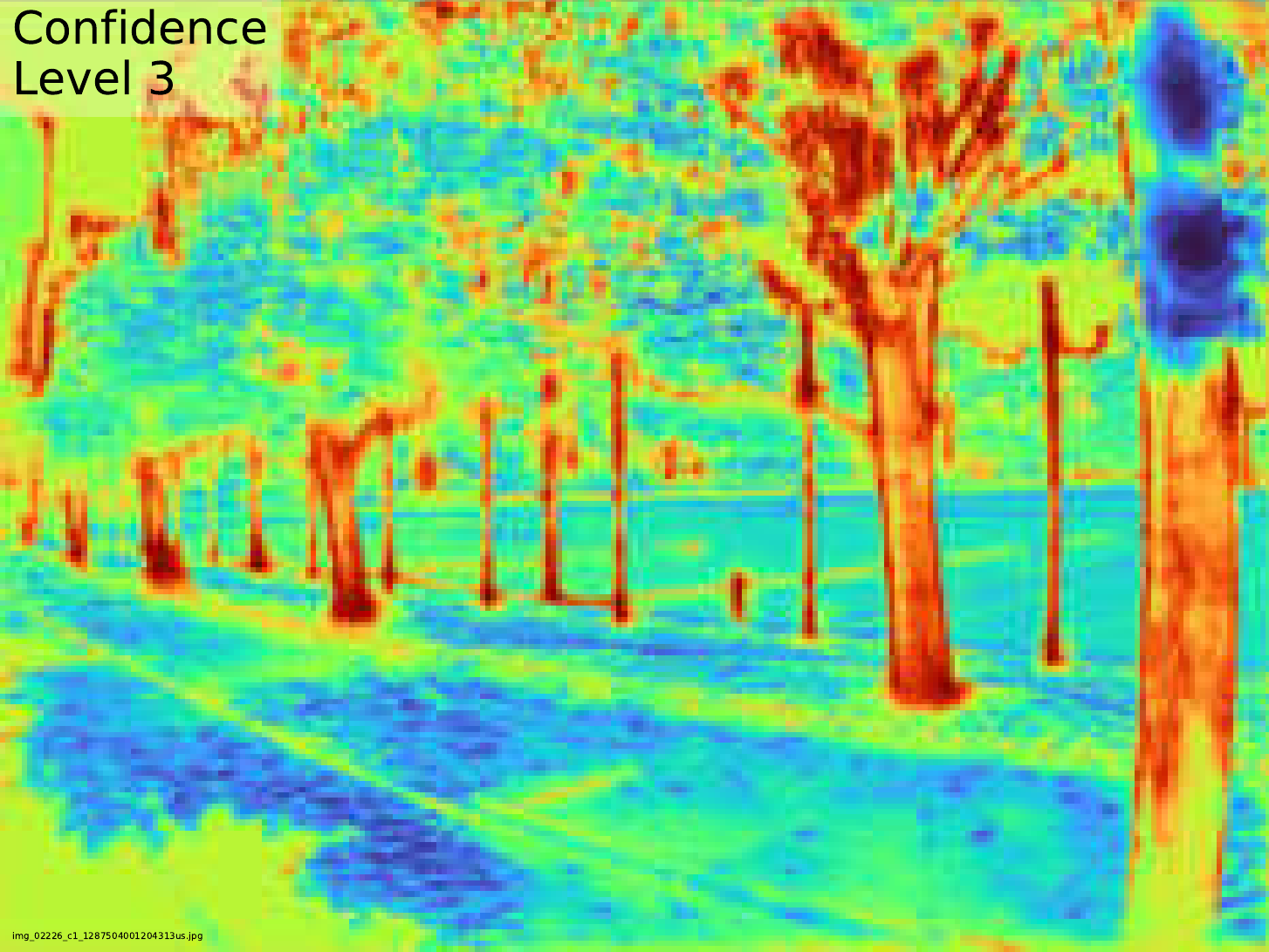}
\end{minipage}
\begin{minipage}{\lwidth\textwidth}
\rotatebox[origin=c]{90}{Reference}
\end{minipage}%
\begin{minipage}{\iwidth\textwidth}
    \centering
    \includegraphics[width=\pwidth\linewidth]{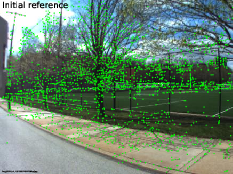}
\end{minipage}%
\begin{minipage}{\iwidth\textwidth}
    \centering
    \includegraphics[width=\pwidth\linewidth]{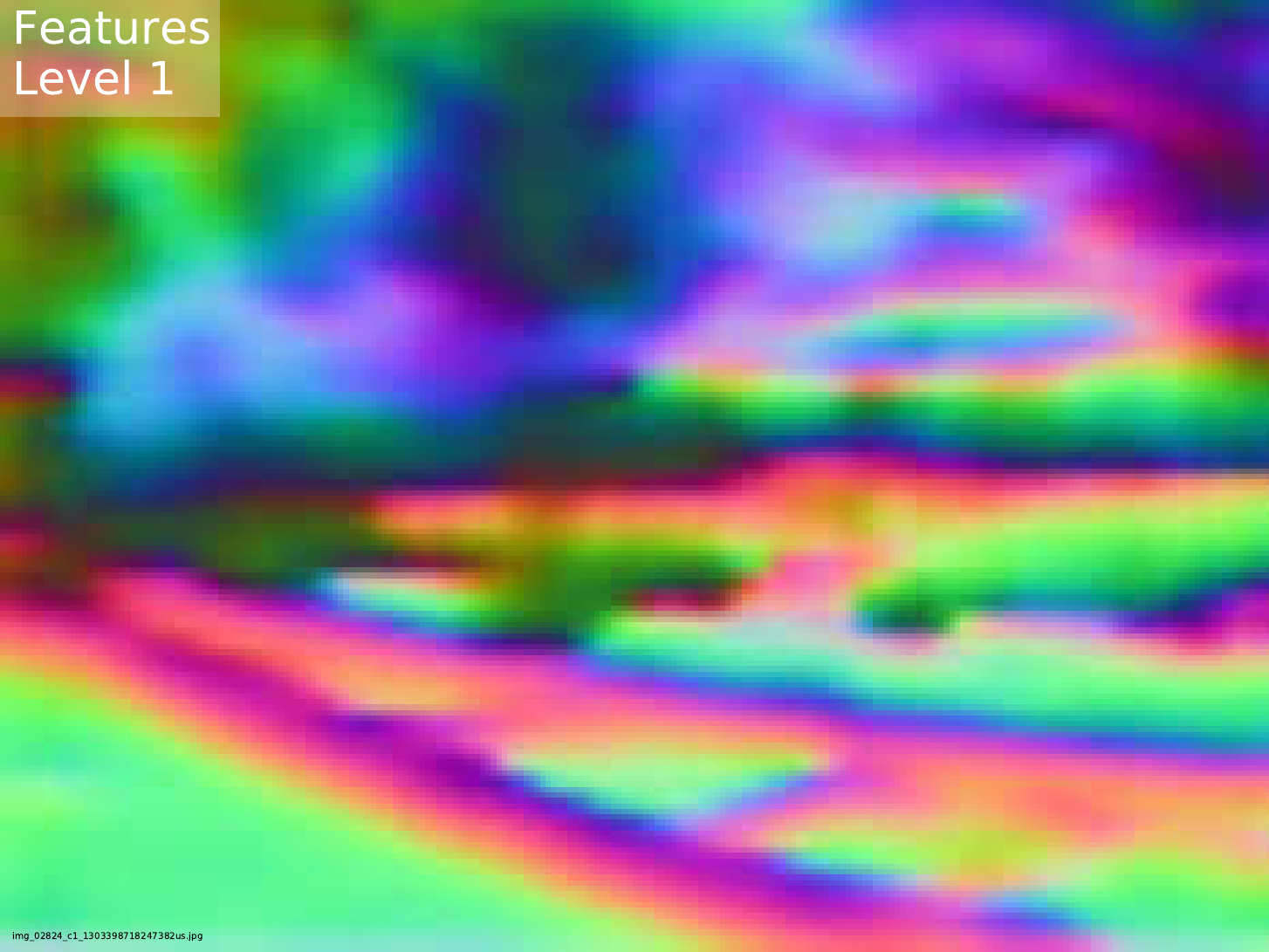}
\end{minipage}%
\begin{minipage}{\iwidth\textwidth}
    \centering
    \includegraphics[width=\pwidth\linewidth]{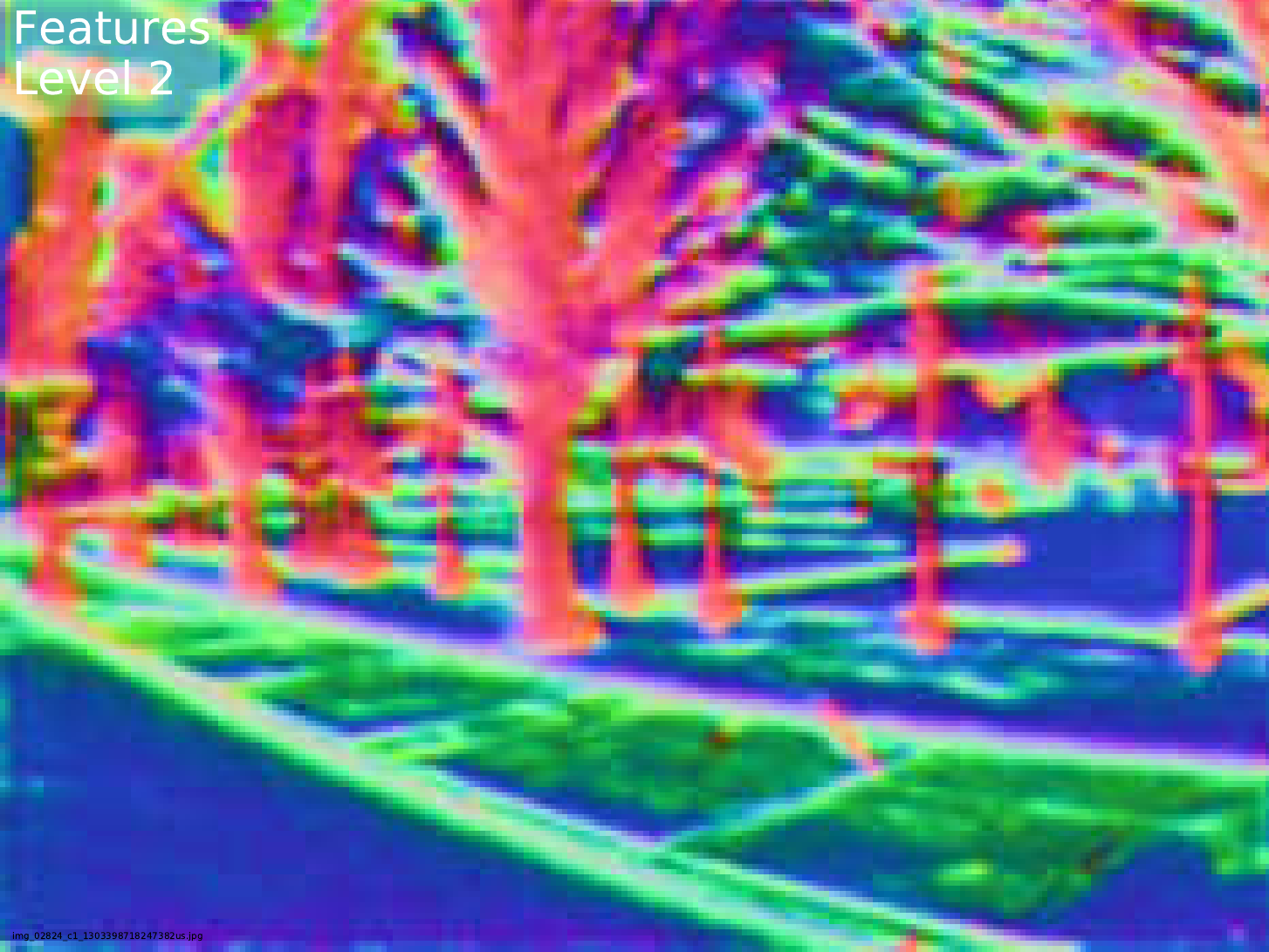}
\end{minipage}%
\begin{minipage}{\iwidth\textwidth}
    \centering
    \includegraphics[width=\pwidth\linewidth]{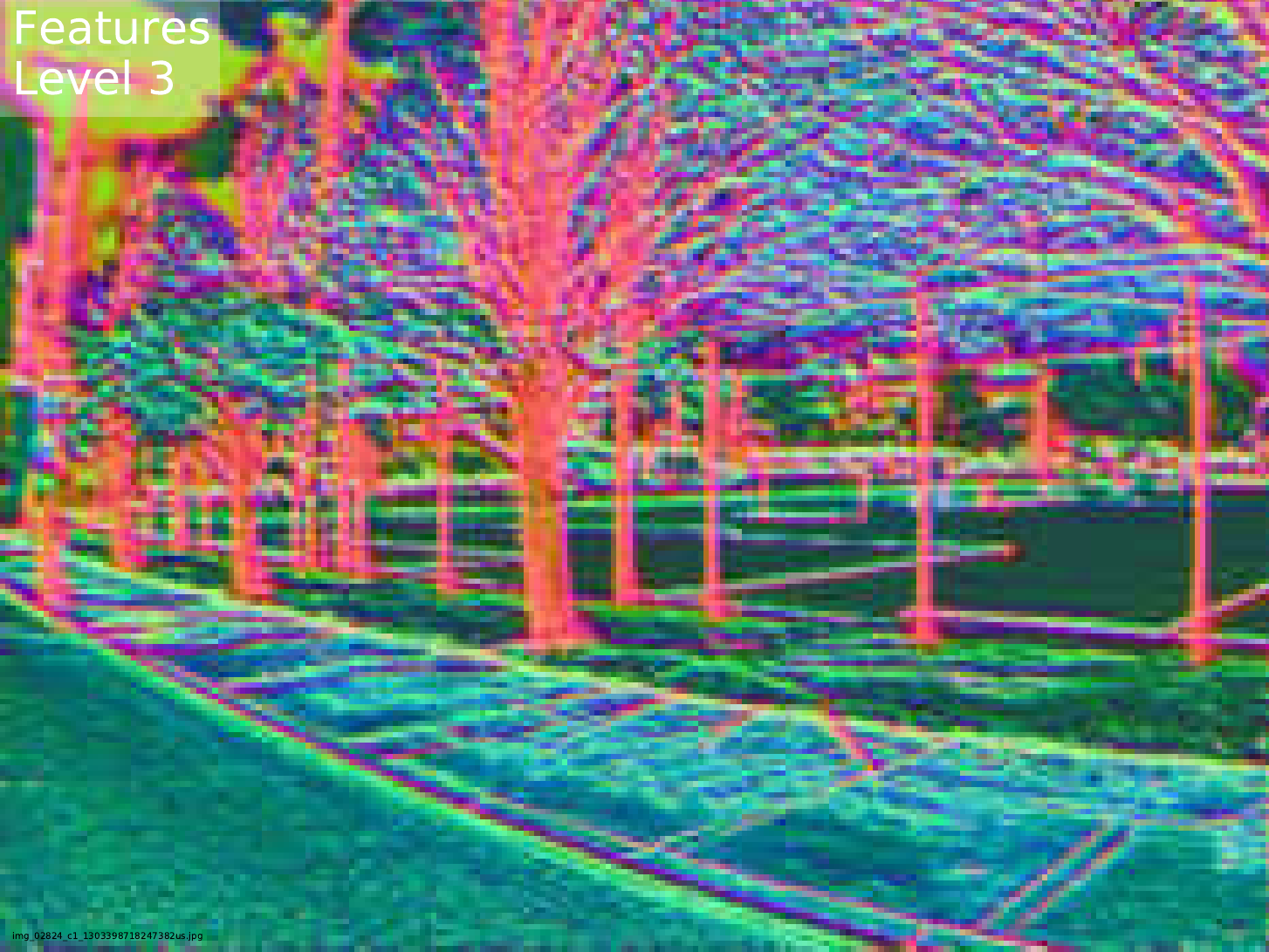}
\end{minipage}%
\begin{minipage}{\iwidth\textwidth}
    \centering
    \includegraphics[width=\pwidth\linewidth]{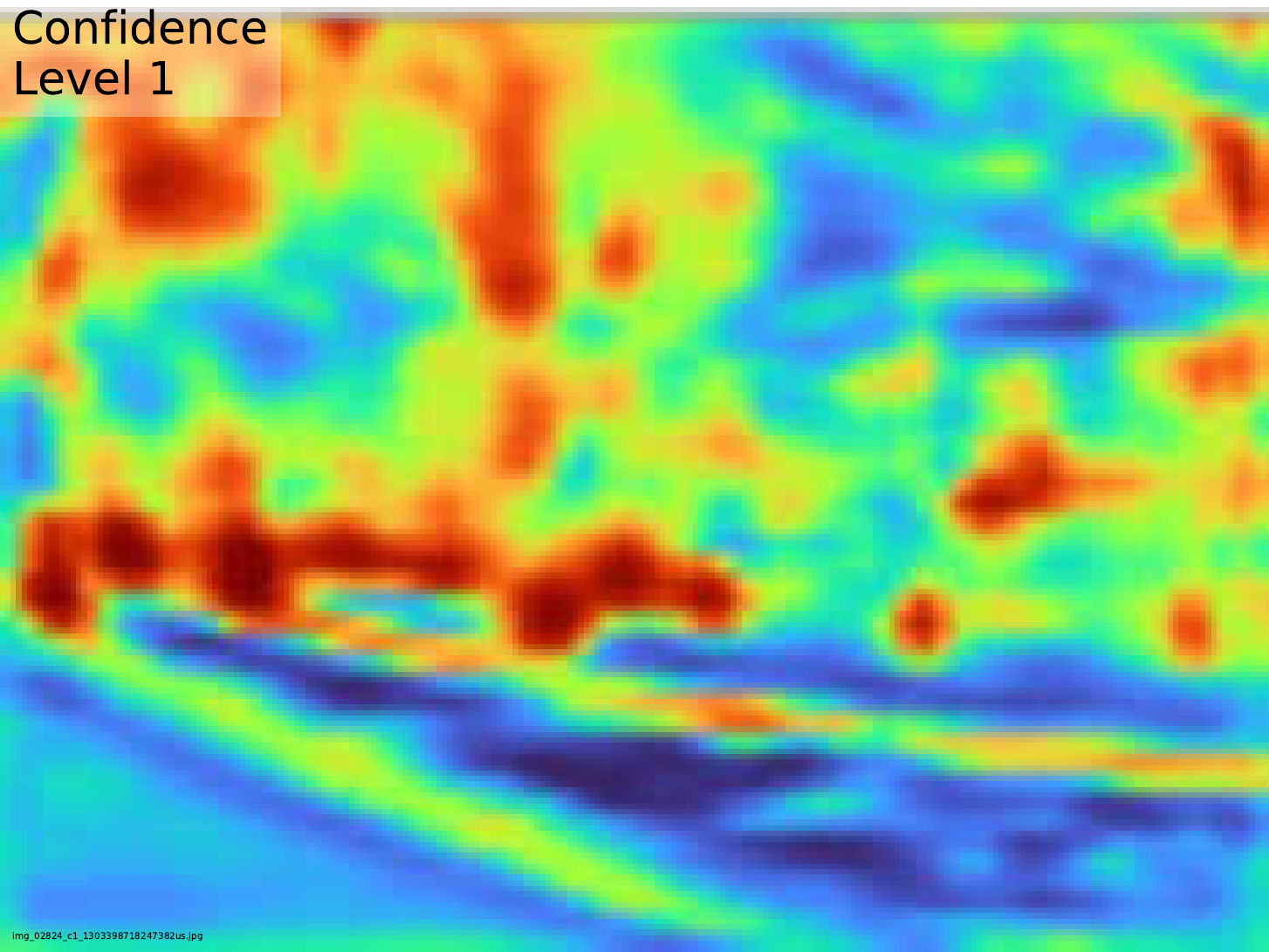}
\end{minipage}%
\begin{minipage}{\iwidth\textwidth}
    \centering
    \includegraphics[width=\pwidth\linewidth]{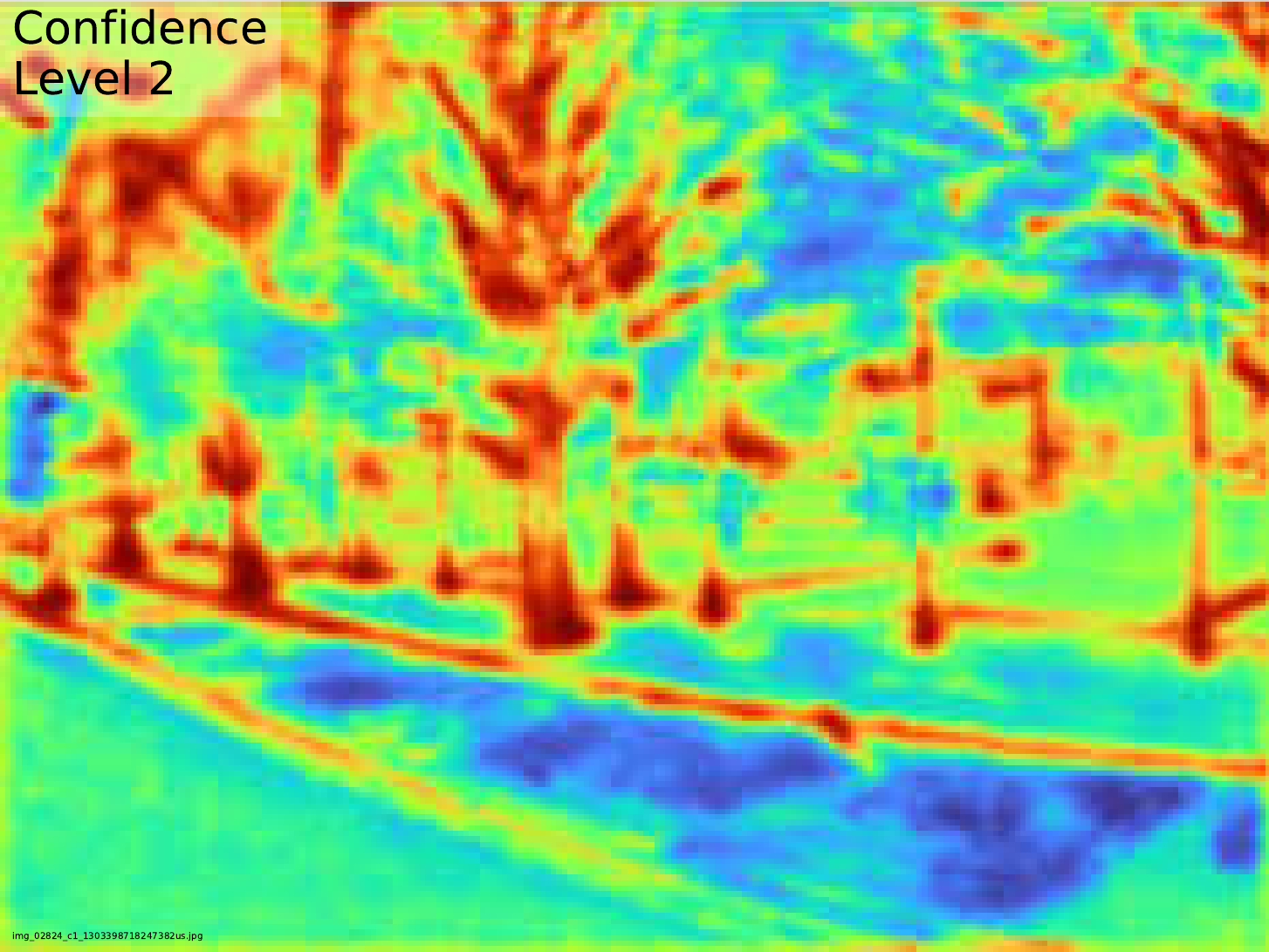}
\end{minipage}%
\begin{minipage}{\iwidth\textwidth}
    \centering
    \includegraphics[width=\pwidth\linewidth]{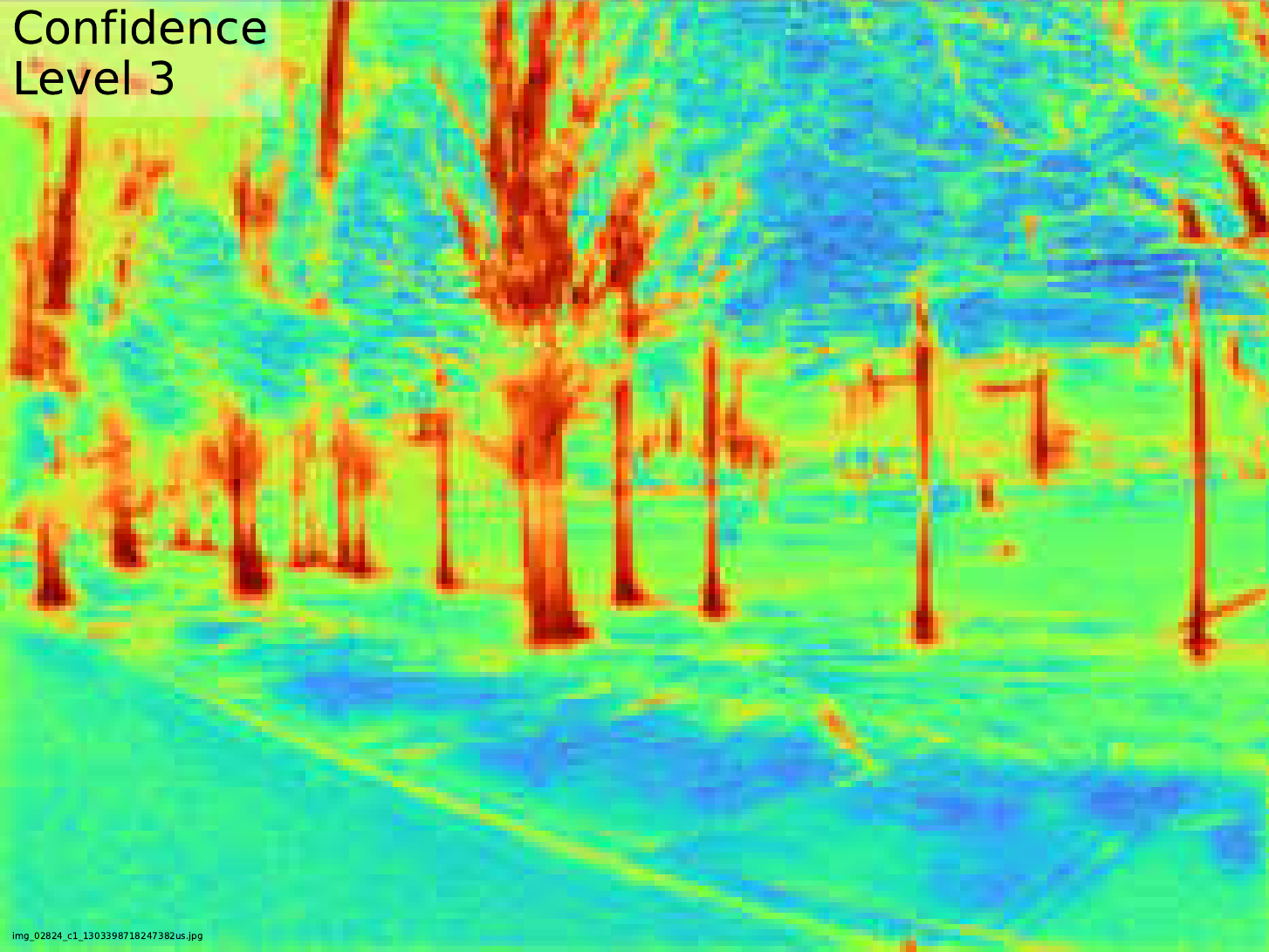}
\end{minipage}
\vspace{2mm}

\begin{minipage}{\lwidth\textwidth}
\rotatebox[origin=c]{90}{Query}
\end{minipage}%
\begin{minipage}{\iwidth\textwidth}
    \centering
    \includegraphics[width=\pwidth\linewidth]{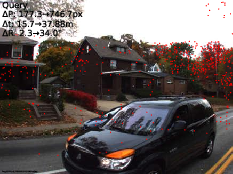}
\end{minipage}%
\begin{minipage}{\iwidth\textwidth}
    \centering
    \includegraphics[width=\pwidth\linewidth]{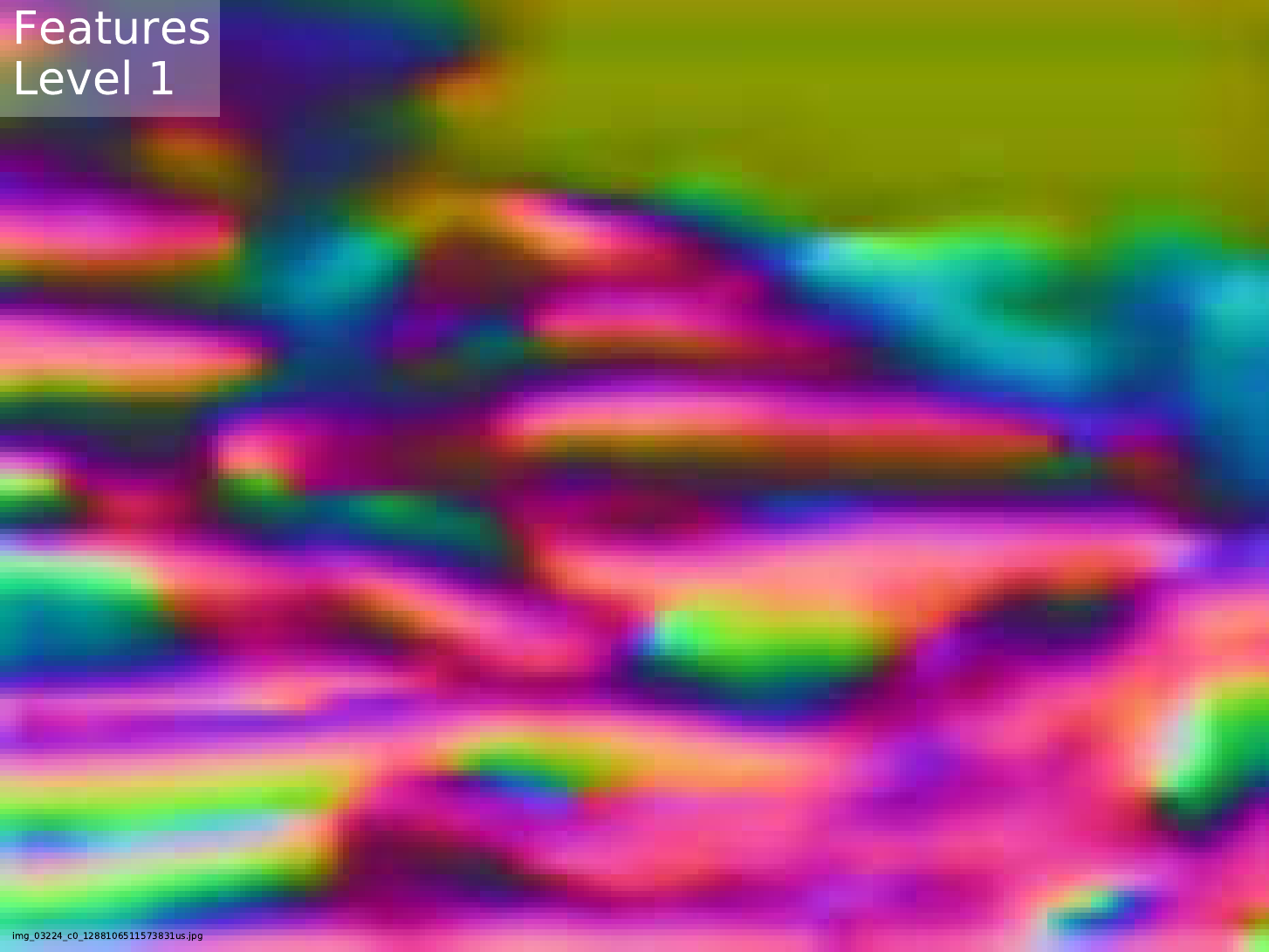}
\end{minipage}%
\begin{minipage}{\iwidth\textwidth}
    \centering
    \includegraphics[width=\pwidth\linewidth]{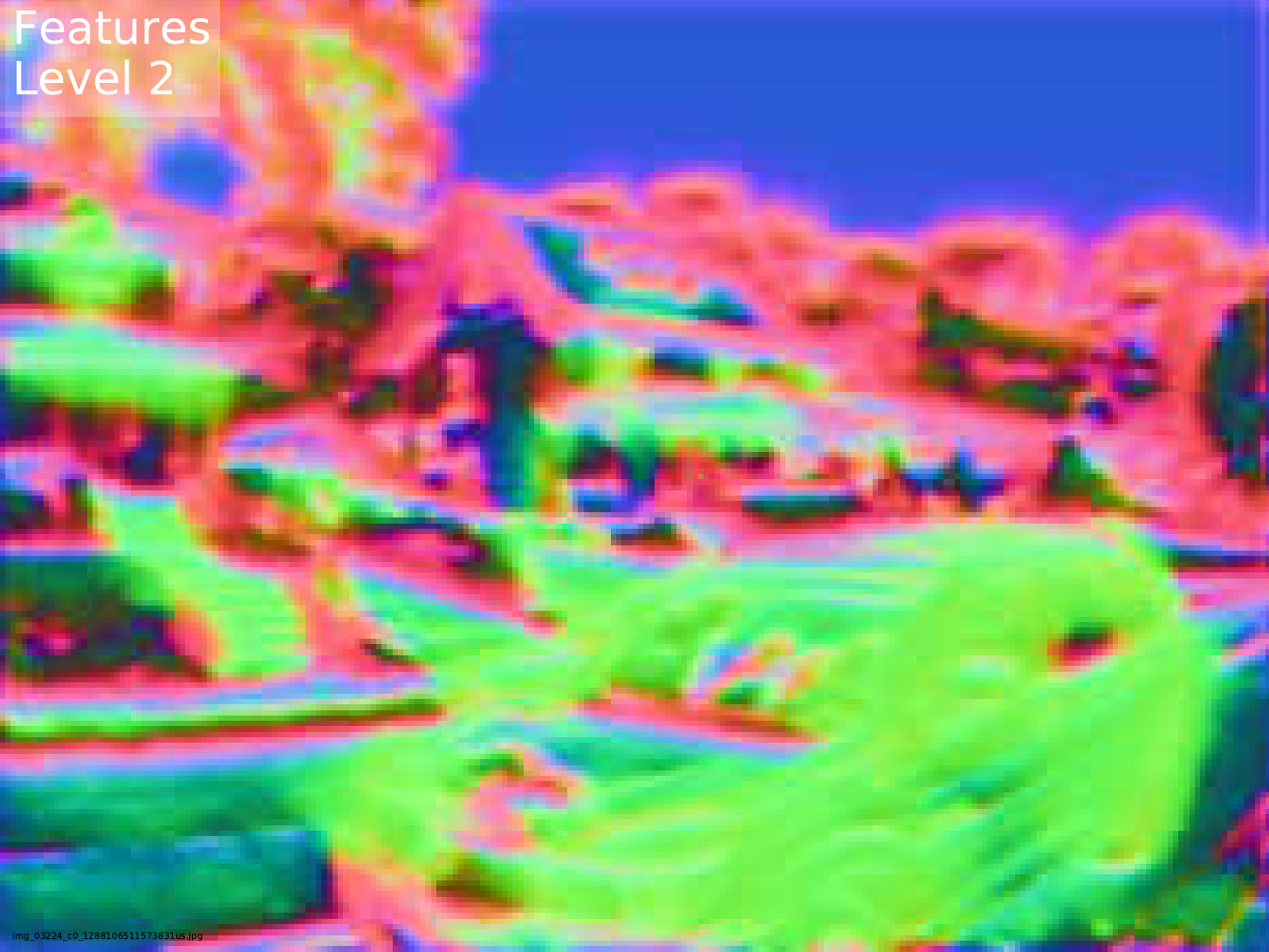}
\end{minipage}%
\begin{minipage}{\iwidth\textwidth}
    \centering
    \includegraphics[width=\pwidth\linewidth]{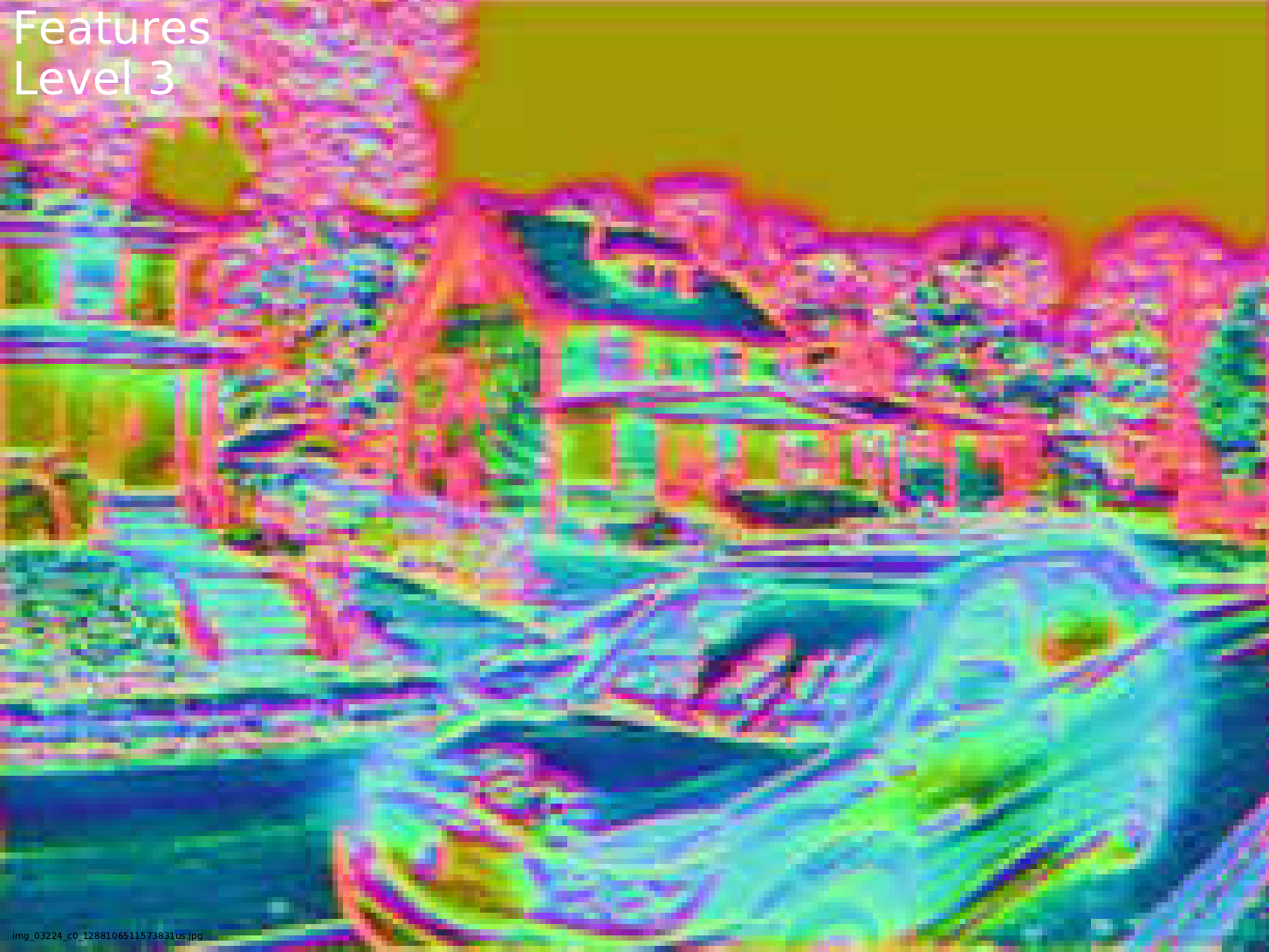}
\end{minipage}%
\begin{minipage}{\iwidth\textwidth}
    \centering
    \includegraphics[width=\pwidth\linewidth]{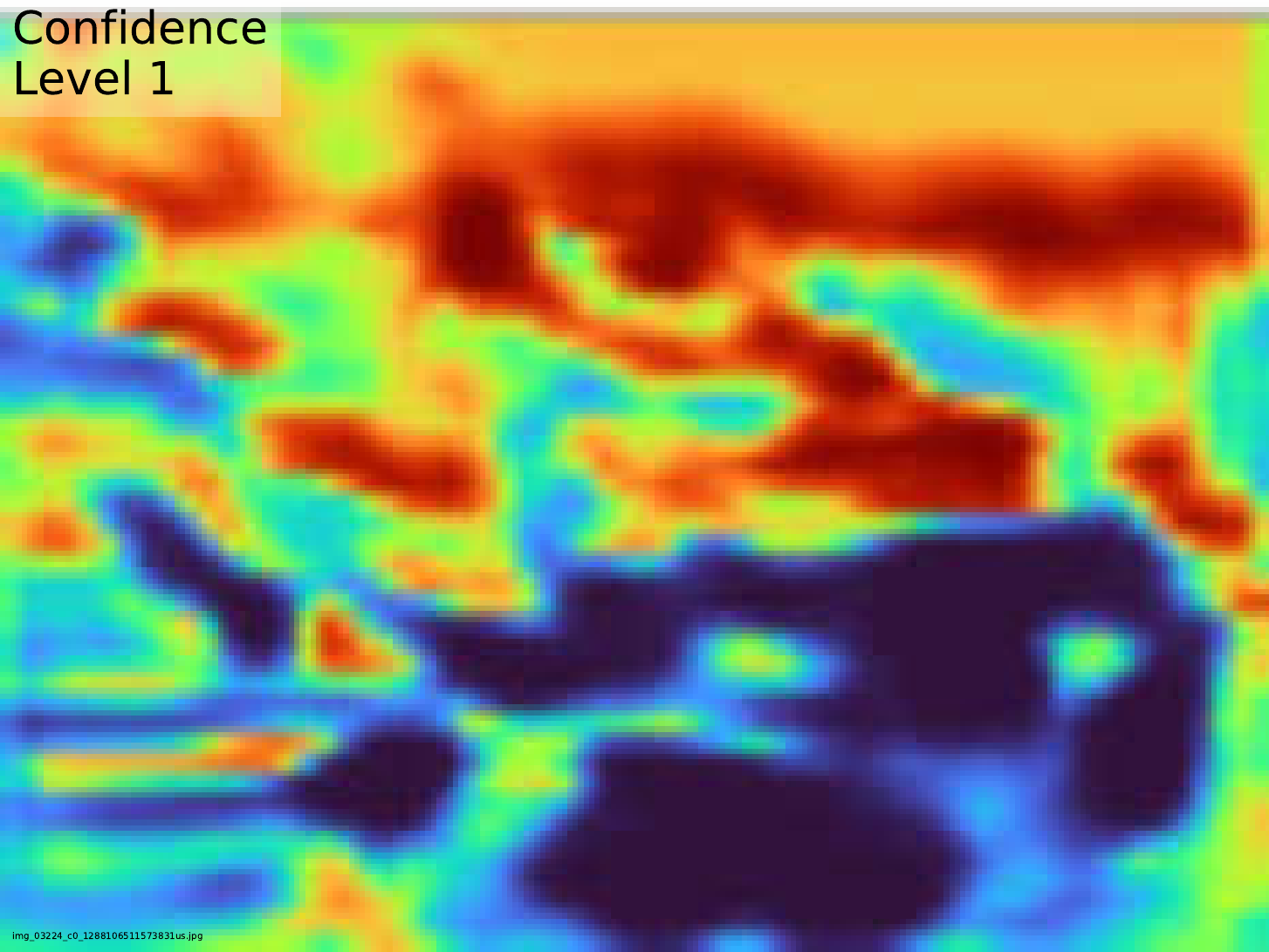}
\end{minipage}%
\begin{minipage}{\iwidth\textwidth}
    \centering
    \includegraphics[width=\pwidth\linewidth]{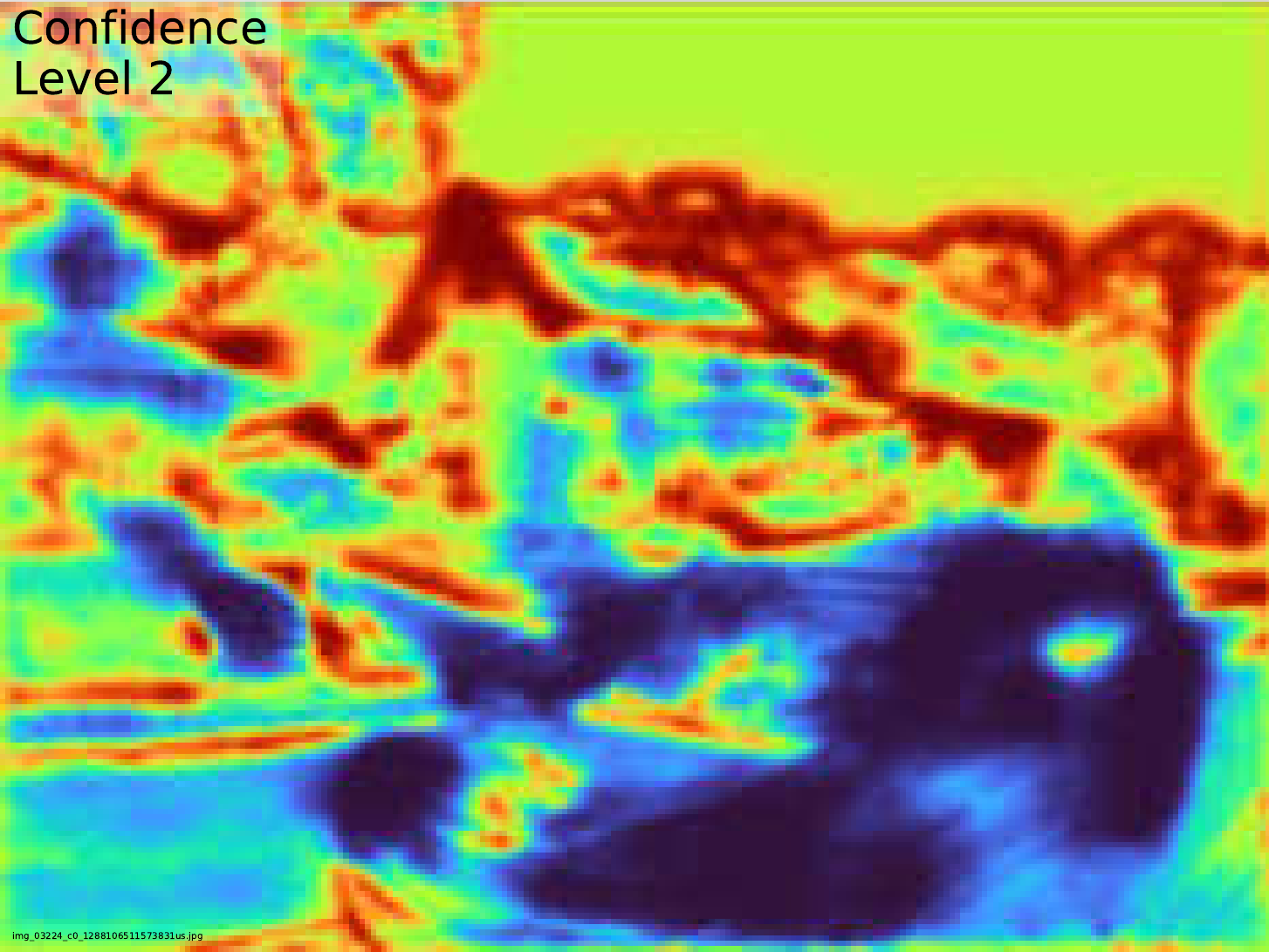}
\end{minipage}%
\begin{minipage}{\iwidth\textwidth}
    \centering
    \includegraphics[width=\pwidth\linewidth]{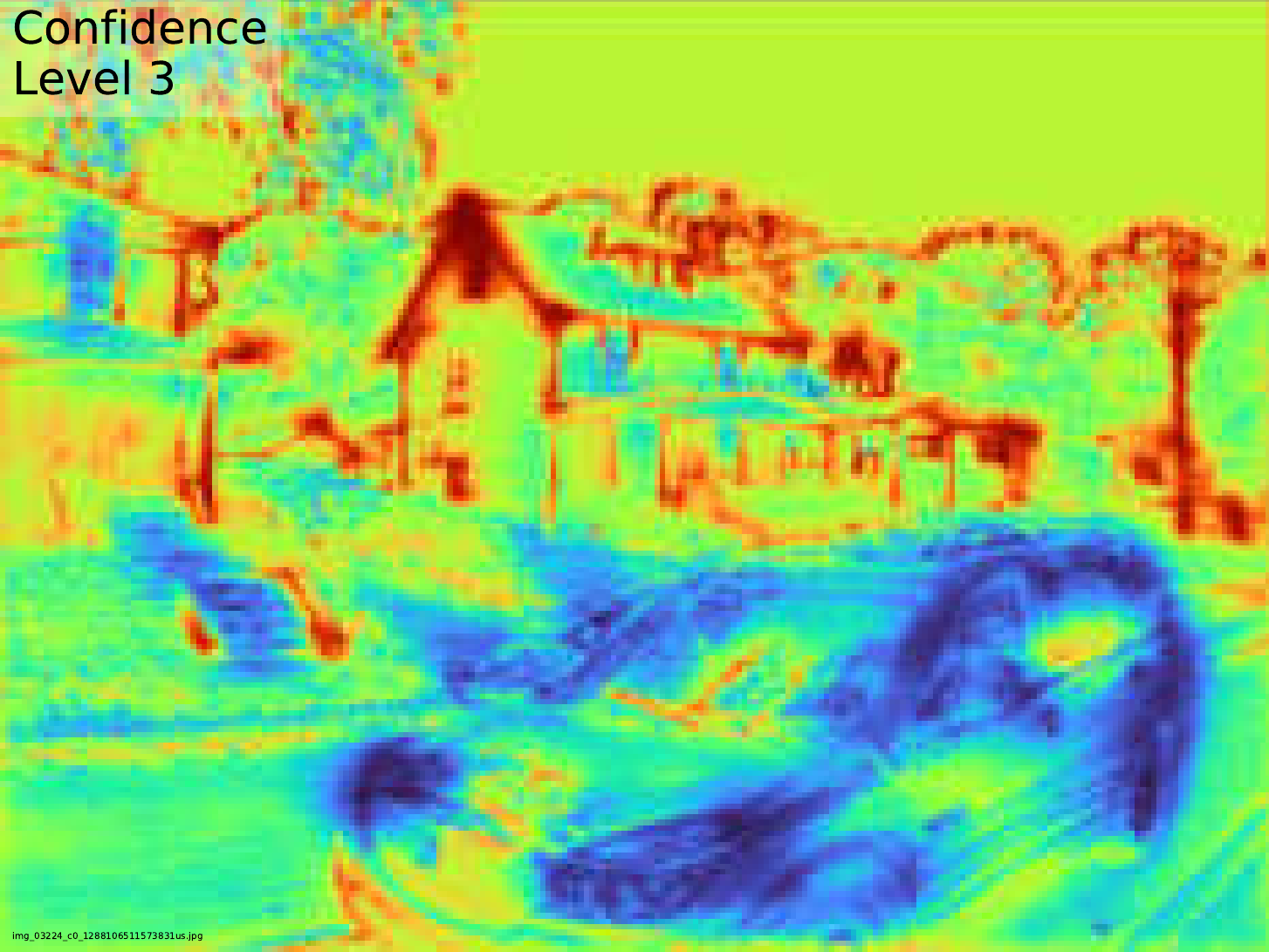}
\end{minipage}
\begin{minipage}{\lwidth\textwidth}
\rotatebox[origin=c]{90}{Reference}
\end{minipage}%
\begin{minipage}{\iwidth\textwidth}
    \centering
    \includegraphics[width=\pwidth\linewidth]{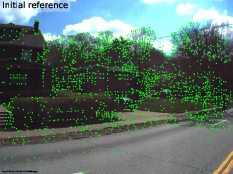}
\end{minipage}%
\begin{minipage}{\iwidth\textwidth}
    \centering
    \includegraphics[width=\pwidth\linewidth]{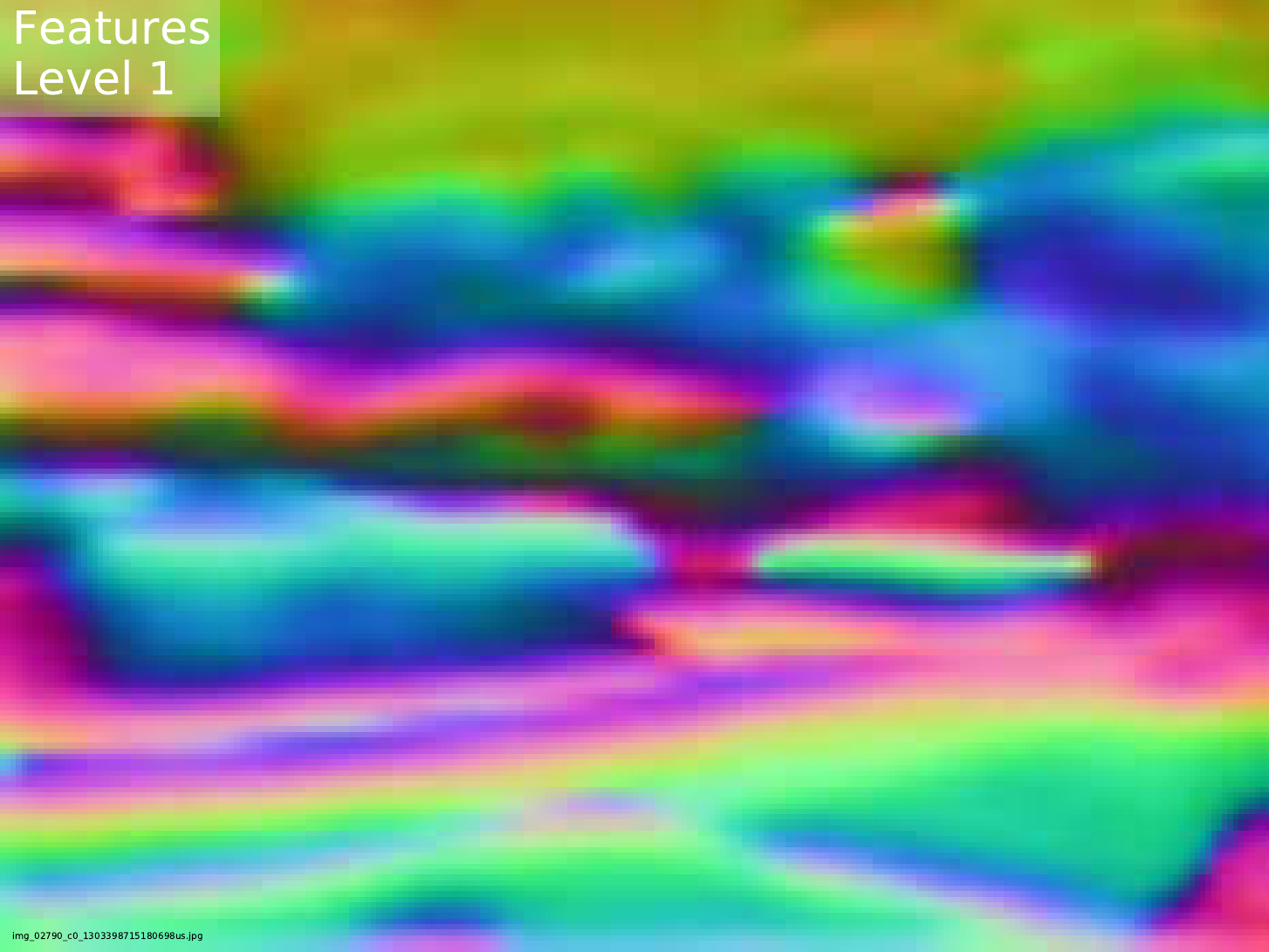}
\end{minipage}%
\begin{minipage}{\iwidth\textwidth}
    \centering
    \includegraphics[width=\pwidth\linewidth]{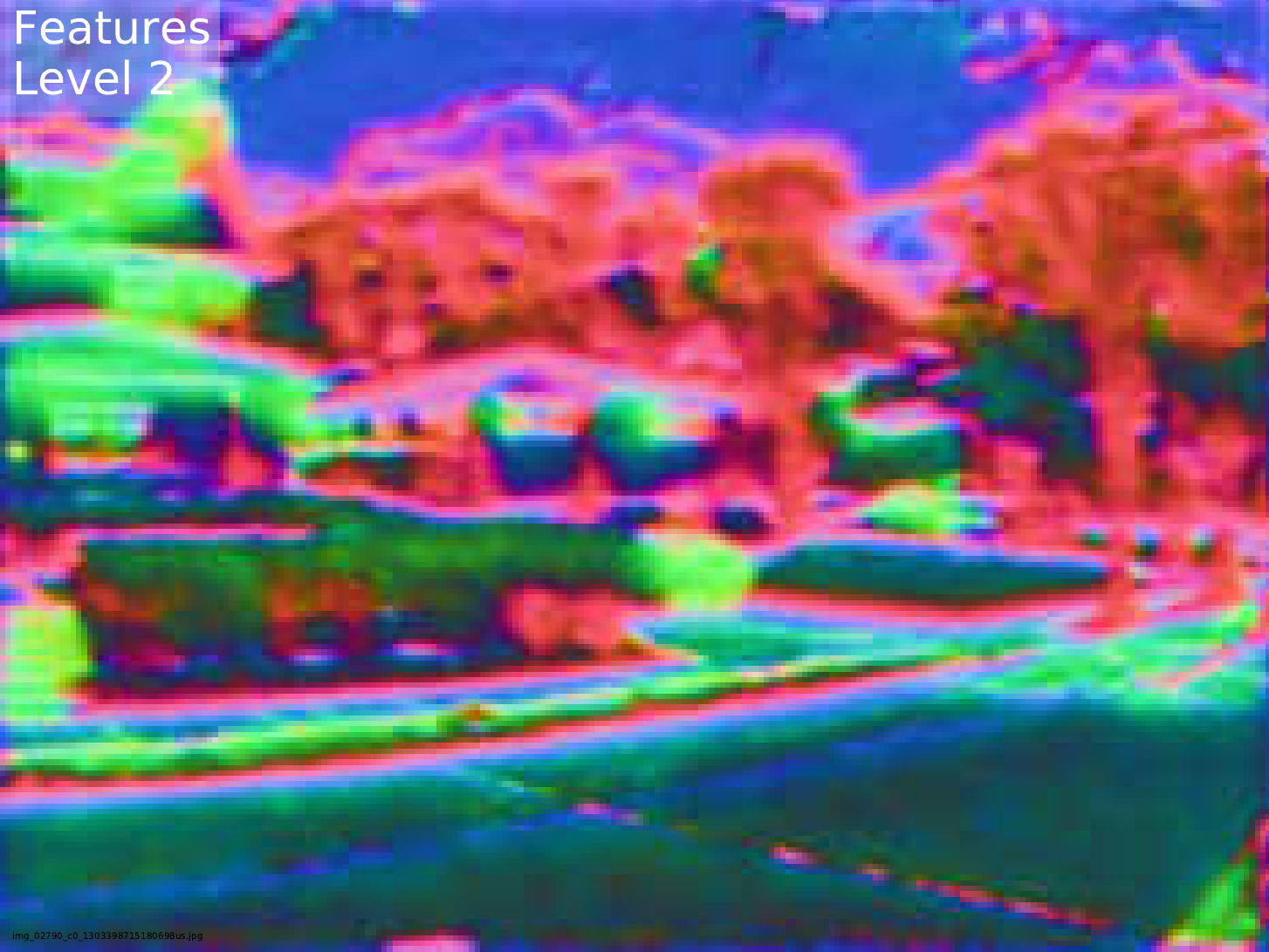}
\end{minipage}%
\begin{minipage}{\iwidth\textwidth}
    \centering
    \includegraphics[width=\pwidth\linewidth]{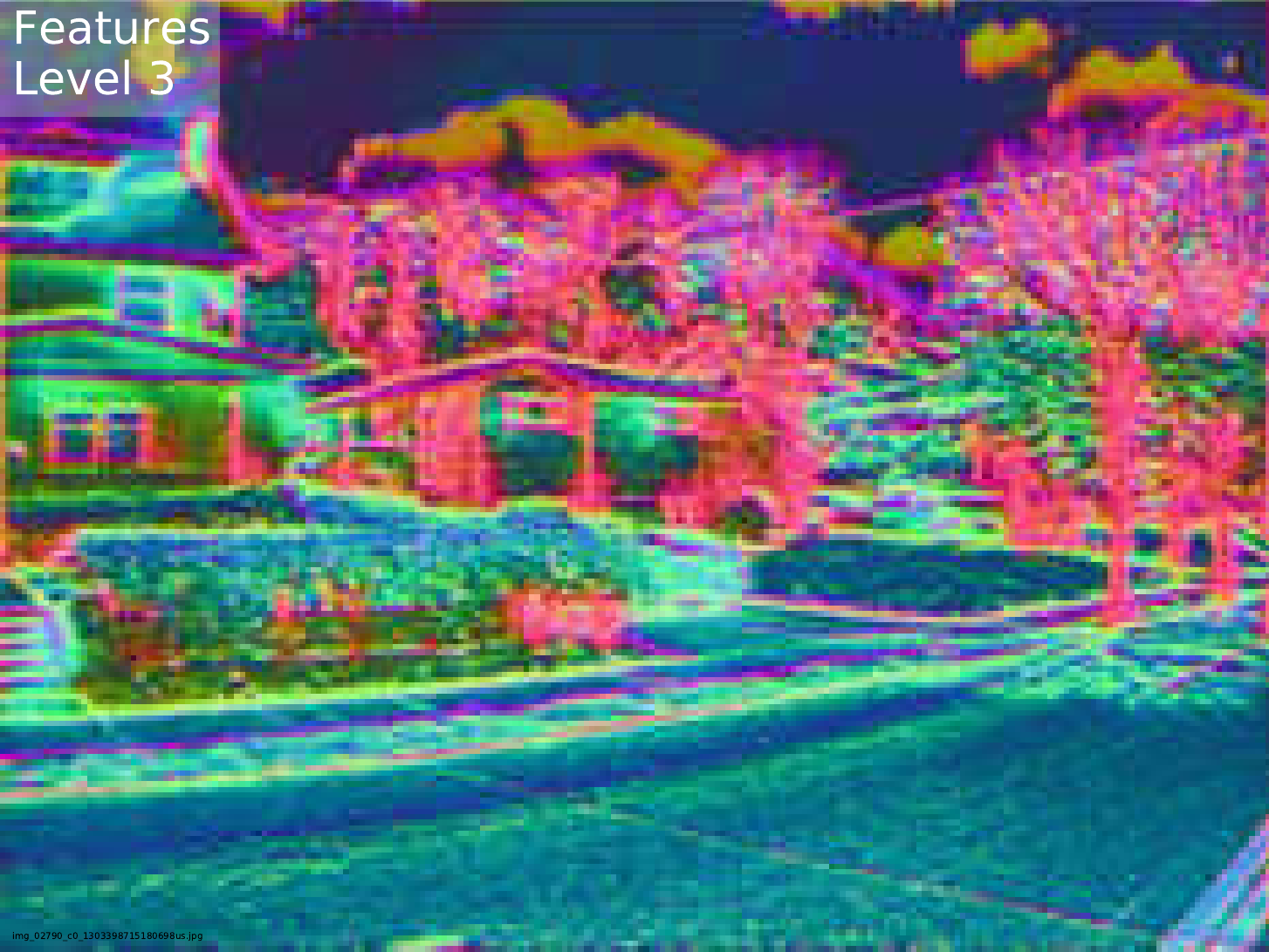}
\end{minipage}%
\begin{minipage}{\iwidth\textwidth}
    \centering
    \includegraphics[width=\pwidth\linewidth]{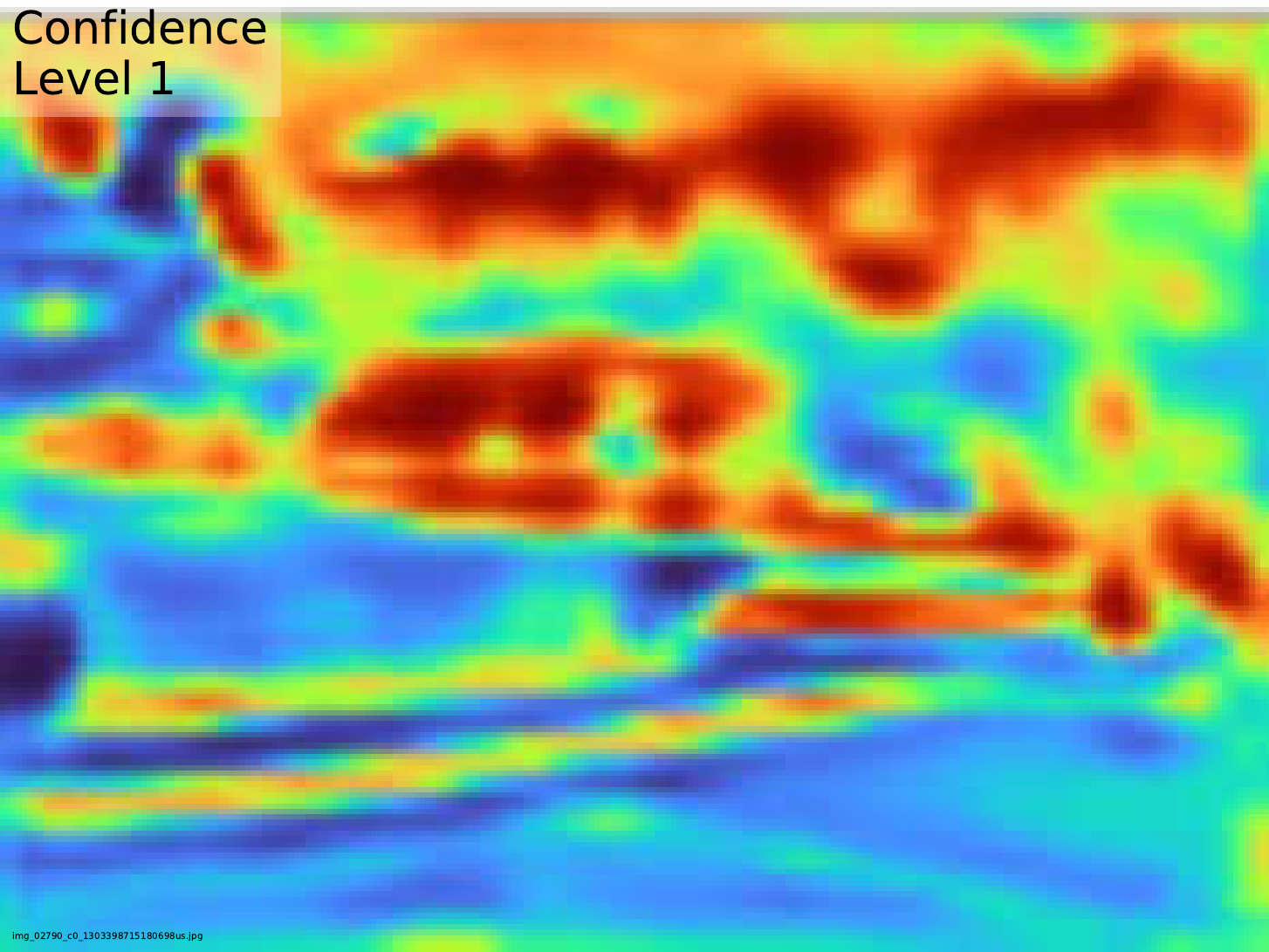}
\end{minipage}%
\begin{minipage}{\iwidth\textwidth}
    \centering
    \includegraphics[width=\pwidth\linewidth]{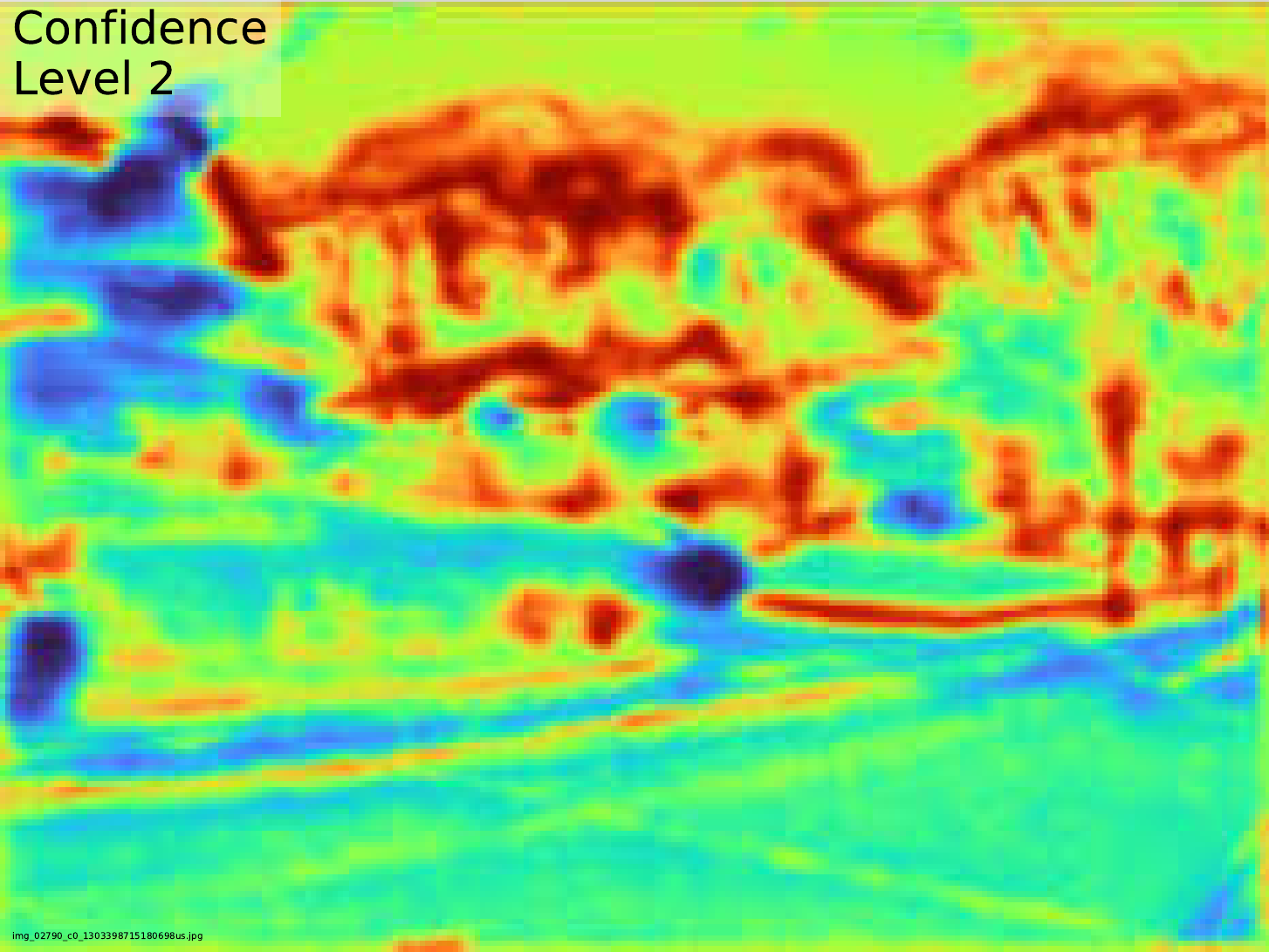}
\end{minipage}%
\begin{minipage}{\iwidth\textwidth}
    \centering
    \includegraphics[width=\pwidth\linewidth]{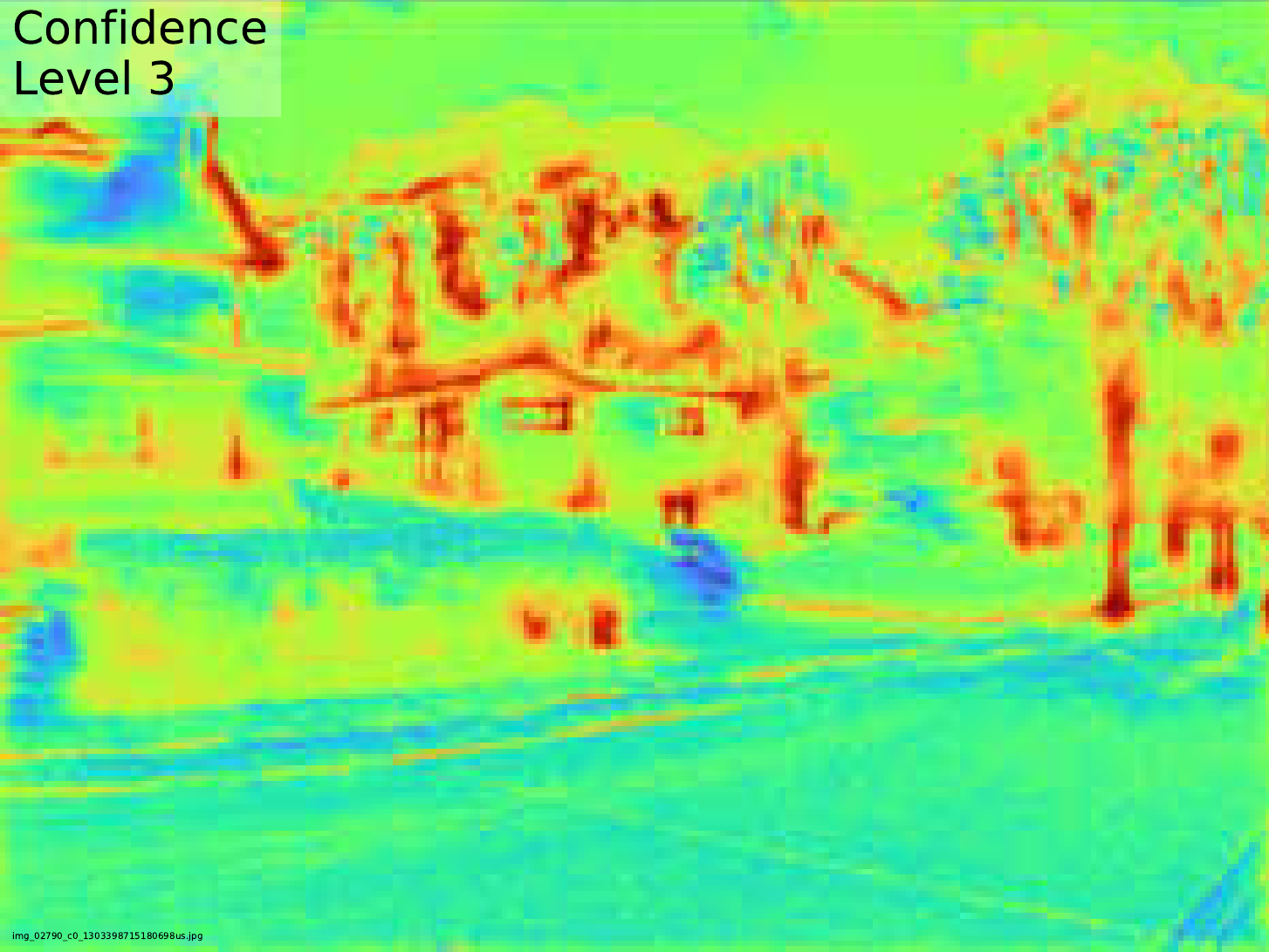}
\end{minipage}
\vspace{2mm}

\begin{minipage}{\lwidth\textwidth}
\rotatebox[origin=c]{90}{Query}
\end{minipage}%
\begin{minipage}{\iwidth\textwidth}
    \centering
    \includegraphics[width=\pwidth\linewidth]{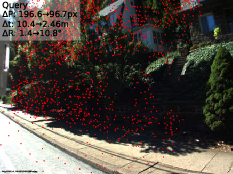}
\end{minipage}%
\begin{minipage}{\iwidth\textwidth}
    \centering
    \includegraphics[width=\pwidth\linewidth]{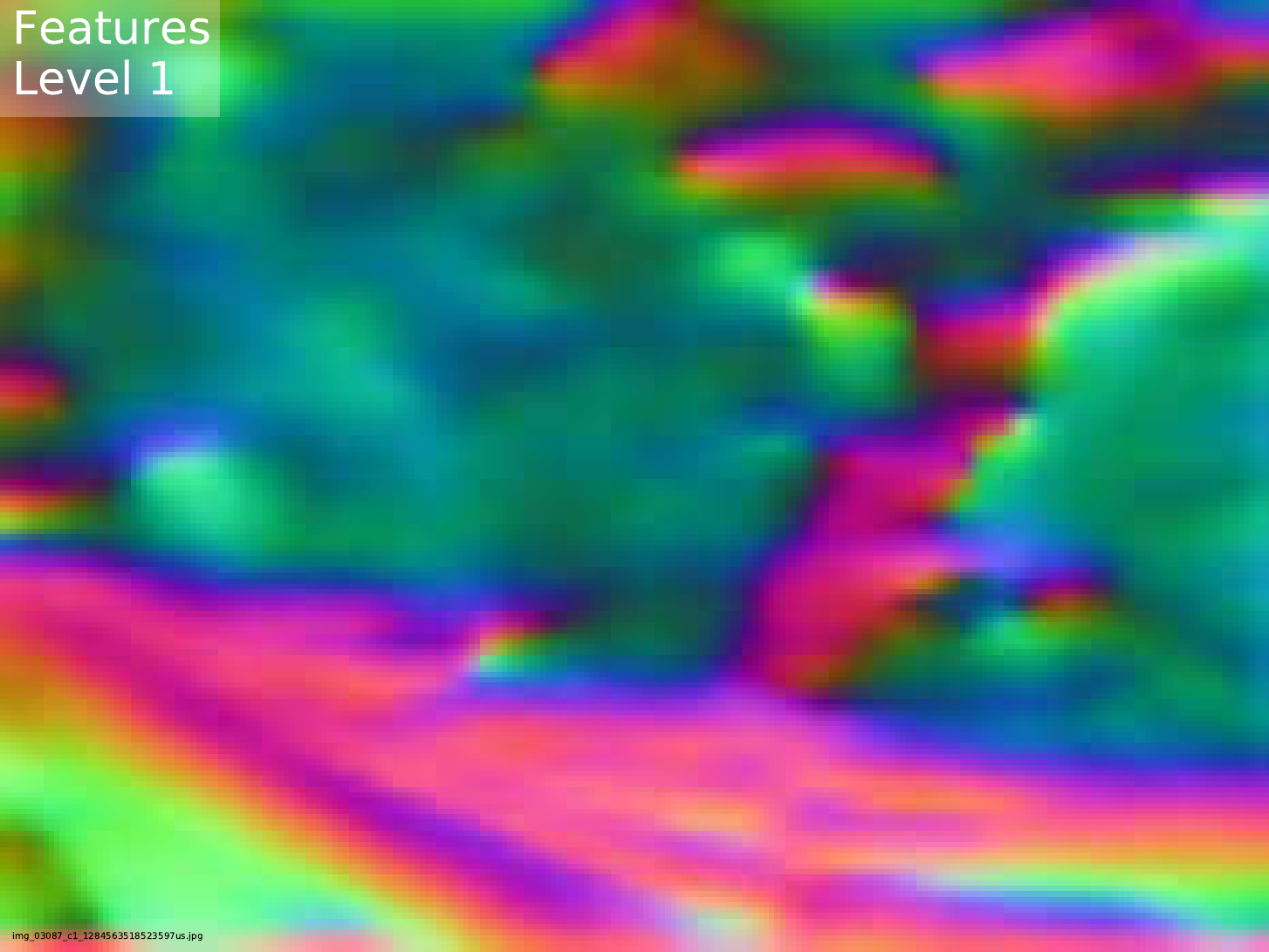}
\end{minipage}%
\begin{minipage}{\iwidth\textwidth}
    \centering
    \includegraphics[width=\pwidth\linewidth]{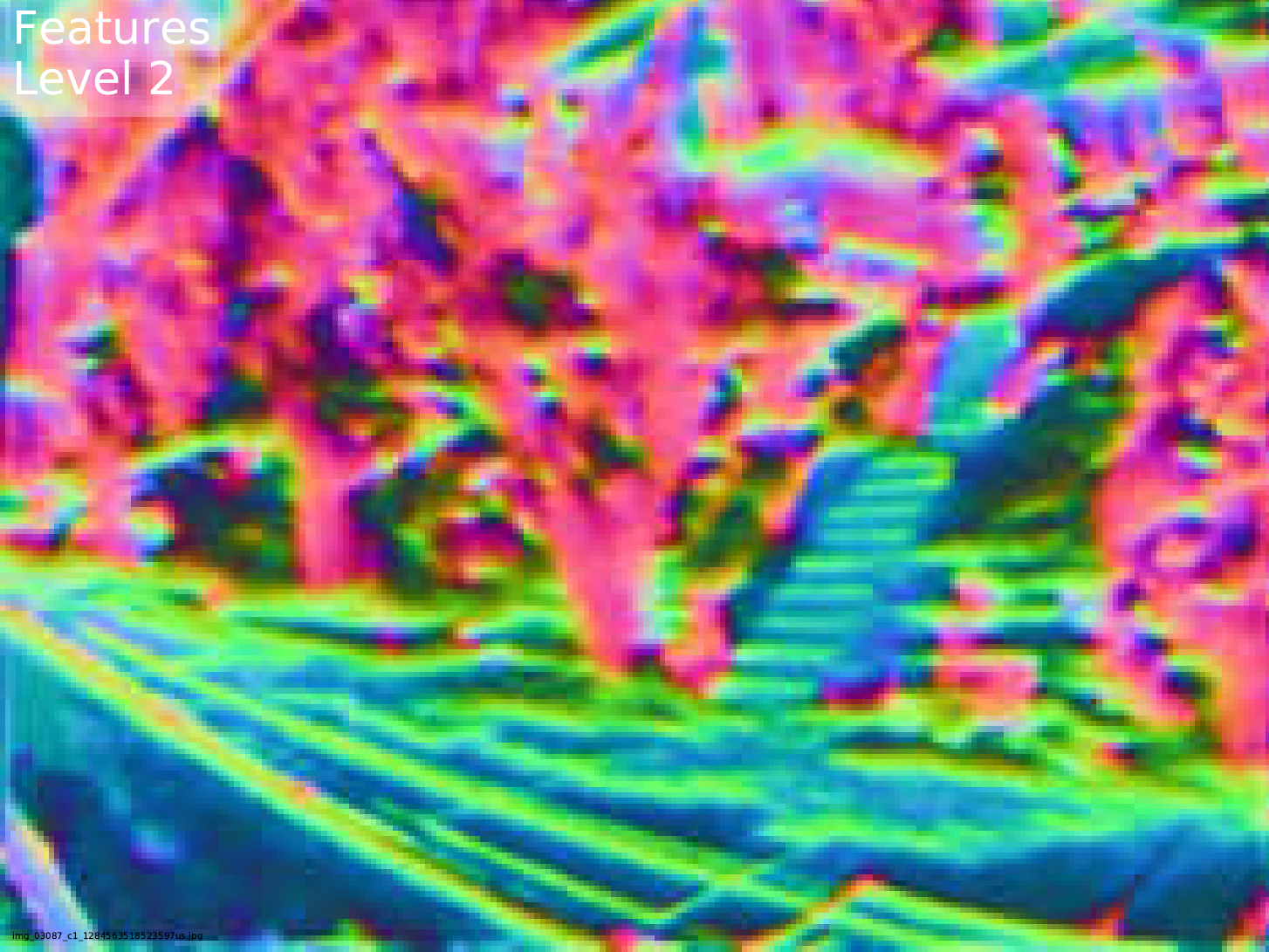}
\end{minipage}%
\begin{minipage}{\iwidth\textwidth}
    \centering
    \includegraphics[width=\pwidth\linewidth]{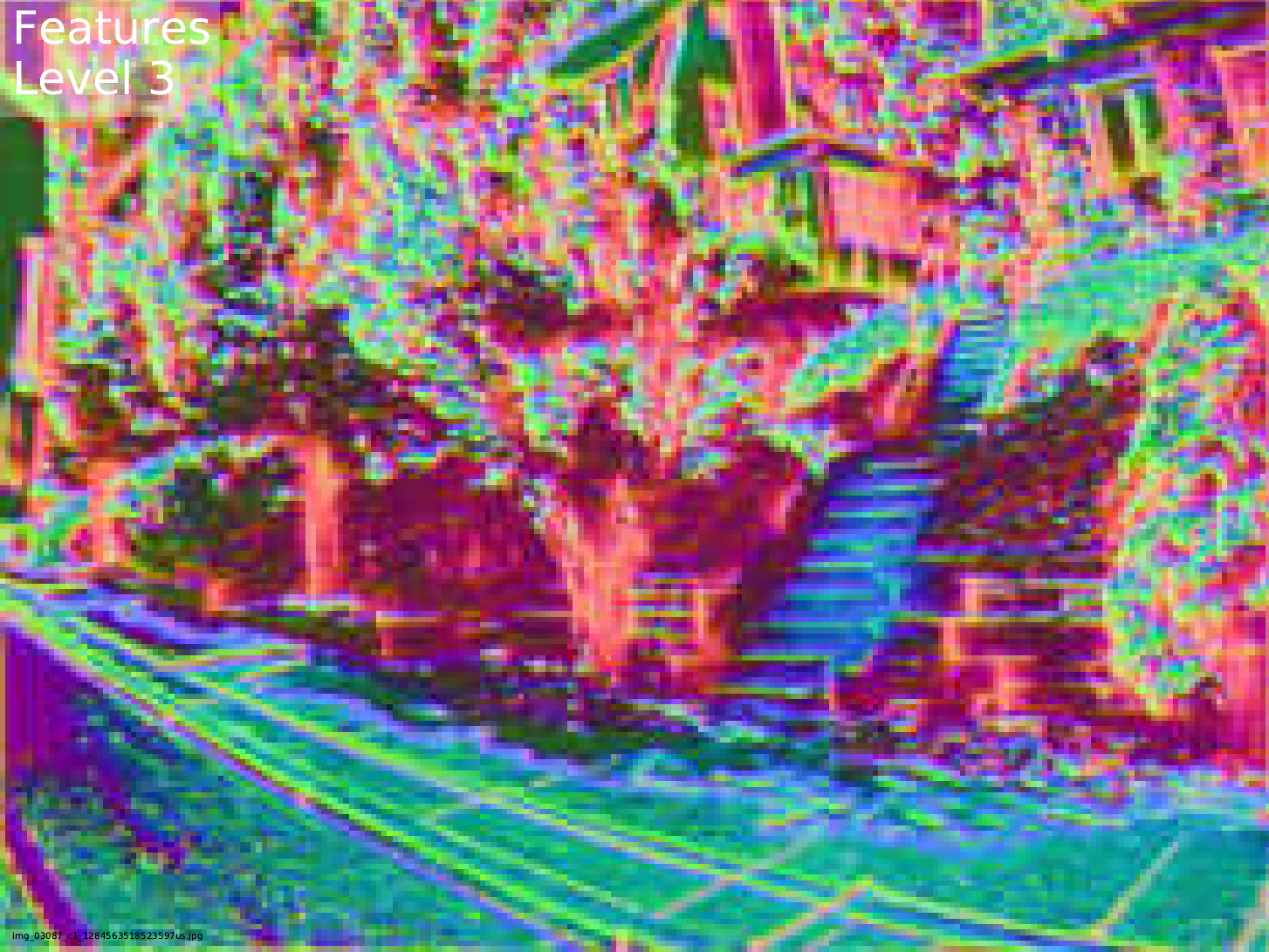}
\end{minipage}%
\begin{minipage}{\iwidth\textwidth}
    \centering
    \includegraphics[width=\pwidth\linewidth]{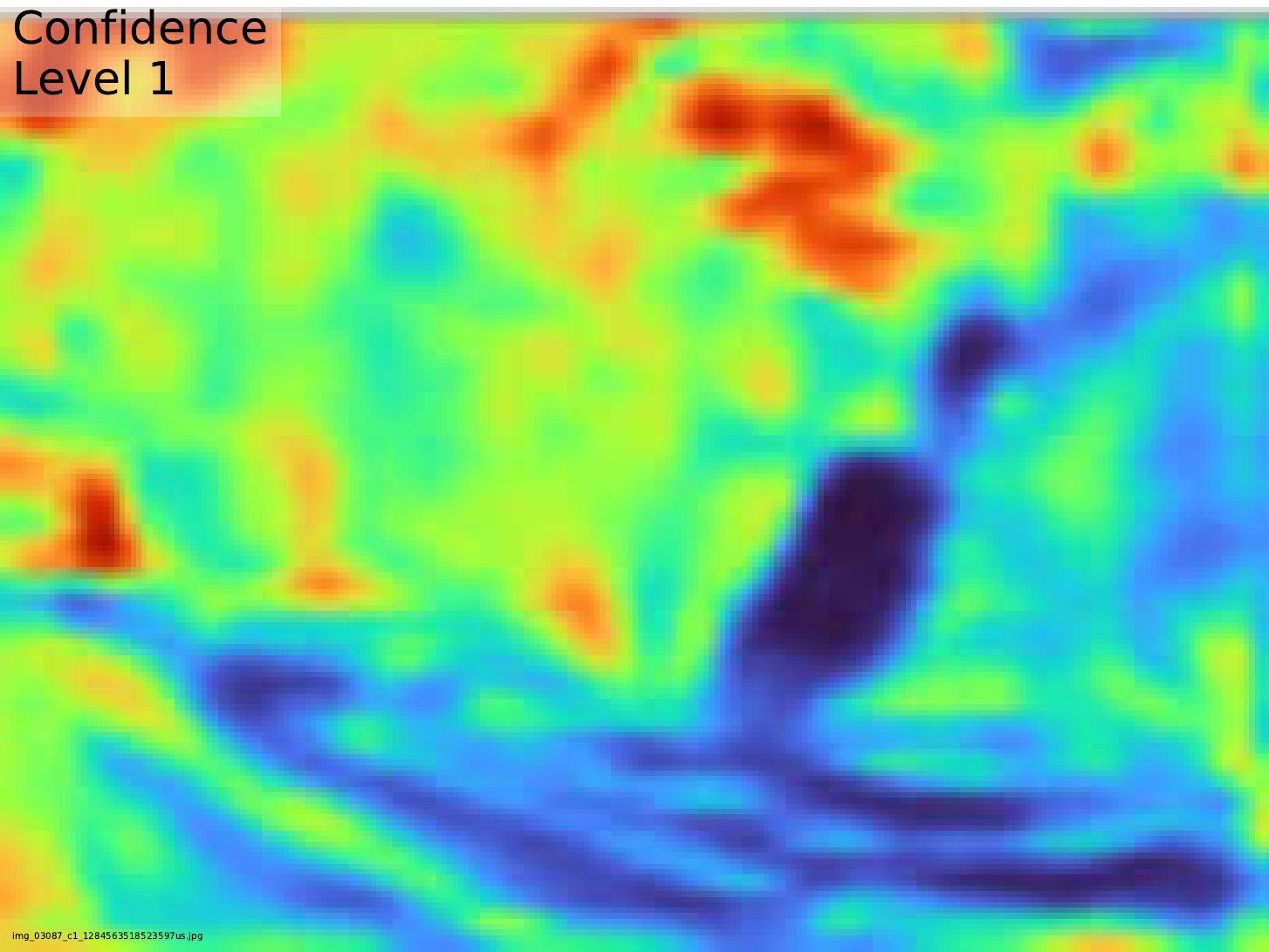}
\end{minipage}%
\begin{minipage}{\iwidth\textwidth}
    \centering
    \includegraphics[width=\pwidth\linewidth]{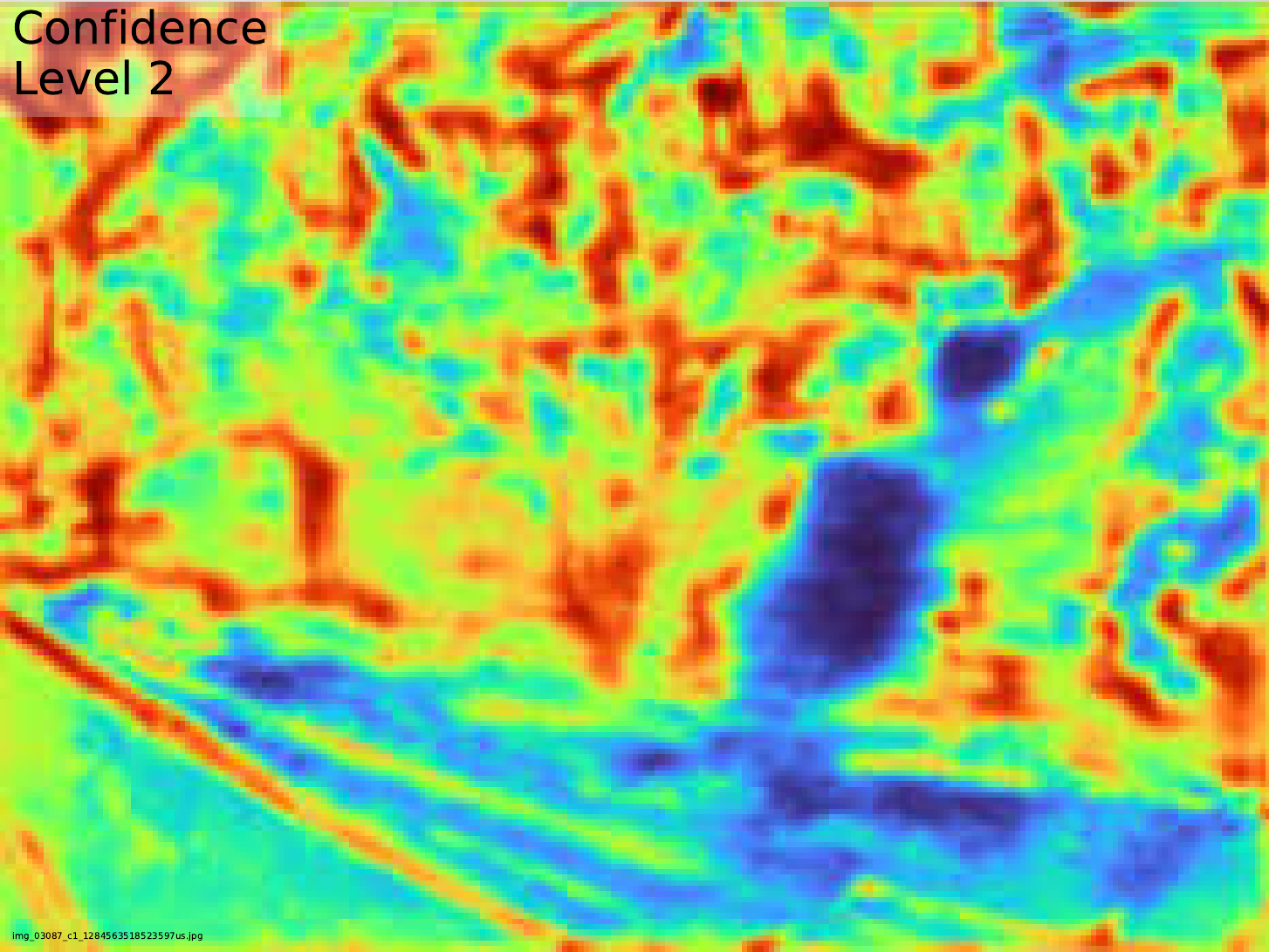}
\end{minipage}%
\begin{minipage}{\iwidth\textwidth}
    \centering
    \includegraphics[width=\pwidth\linewidth]{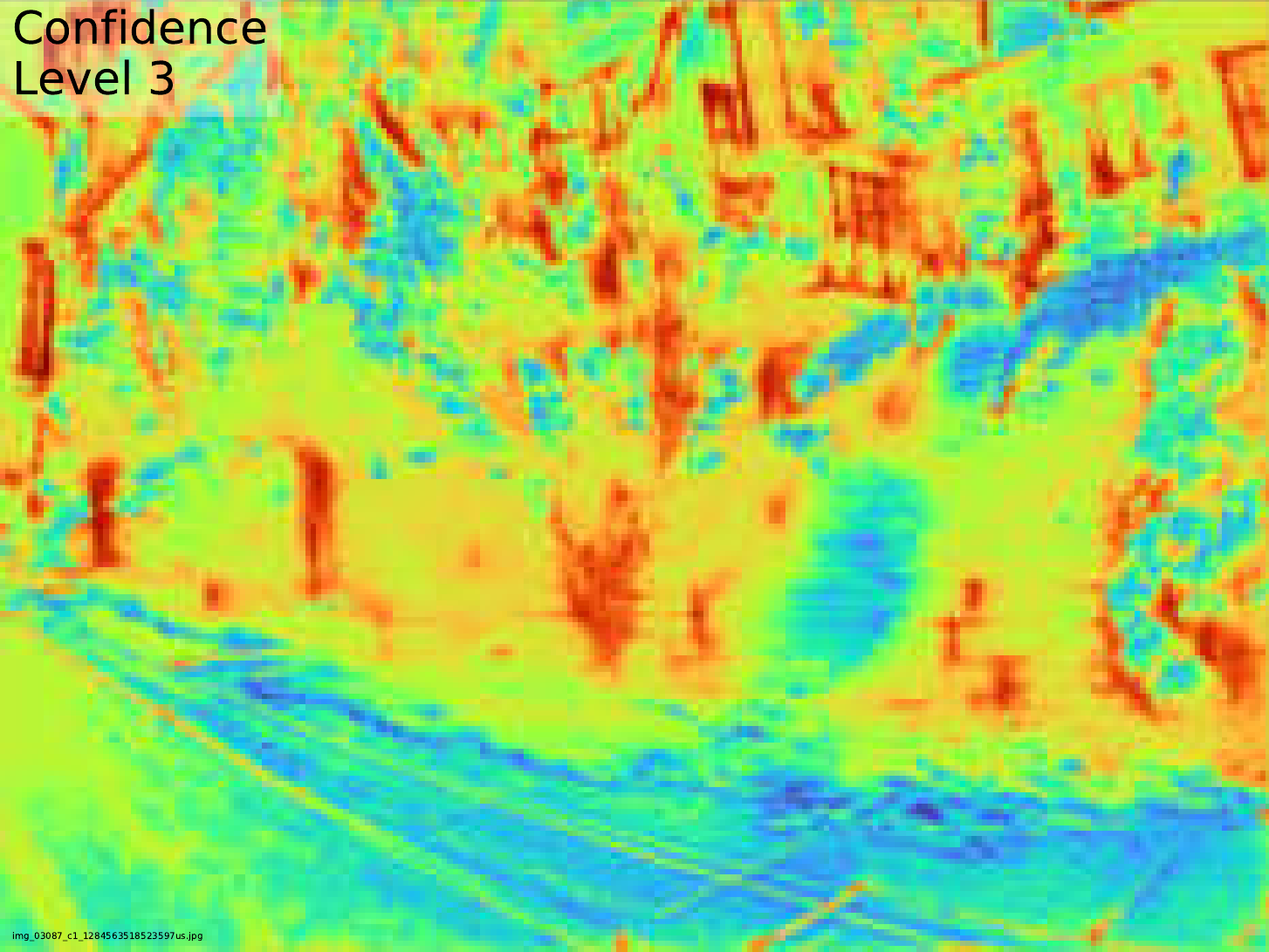}
\end{minipage}
\begin{minipage}{\lwidth\textwidth}
\rotatebox[origin=c]{90}{Reference}
\end{minipage}%
\begin{minipage}{\iwidth\textwidth}
    \centering
    \includegraphics[width=\pwidth\linewidth]{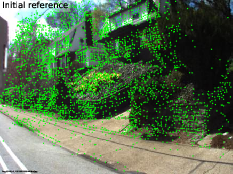}
\end{minipage}%
\begin{minipage}{\iwidth\textwidth}
    \centering
    \includegraphics[width=\pwidth\linewidth]{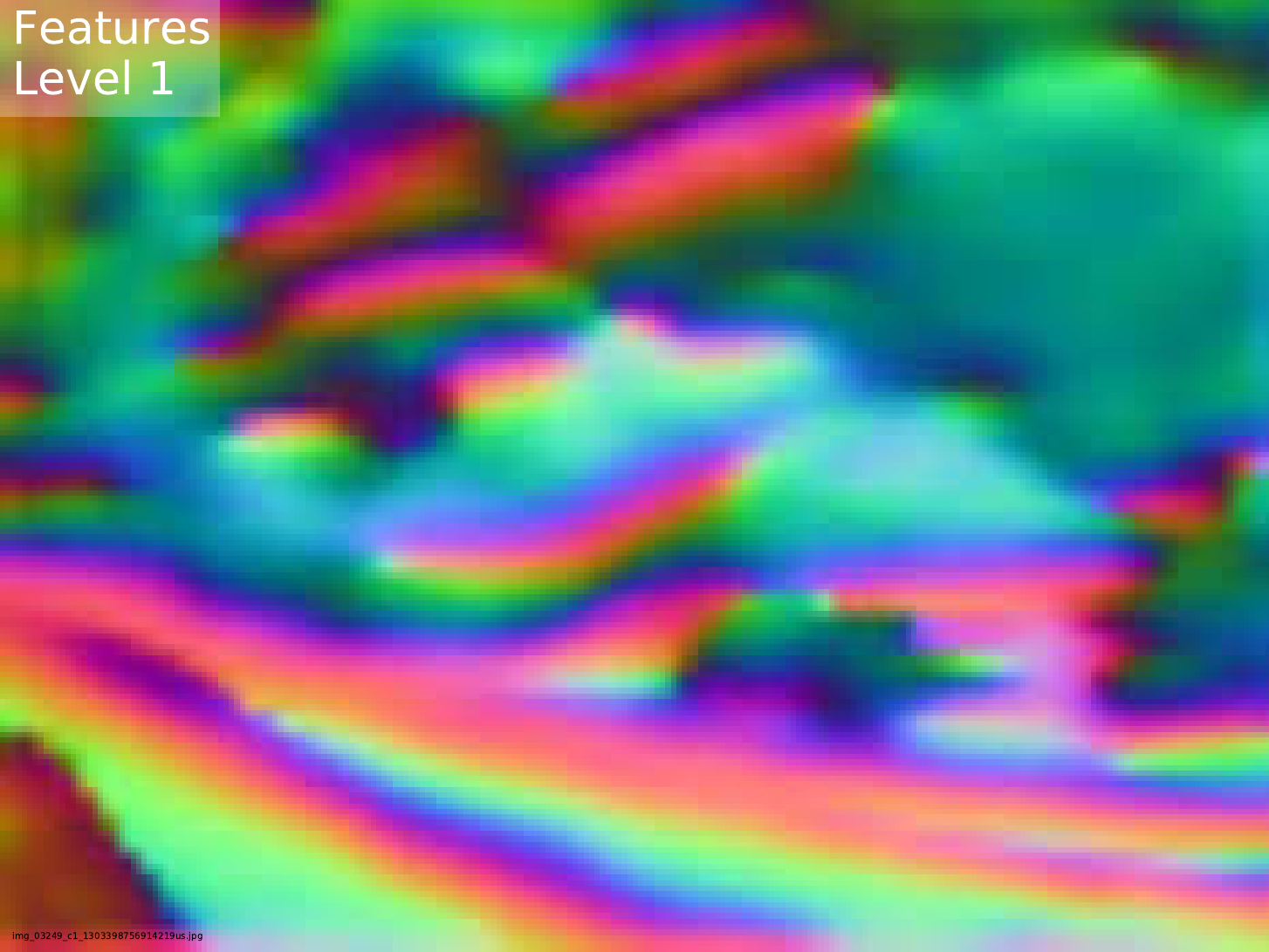}
\end{minipage}%
\begin{minipage}{\iwidth\textwidth}
    \centering
    \includegraphics[width=\pwidth\linewidth]{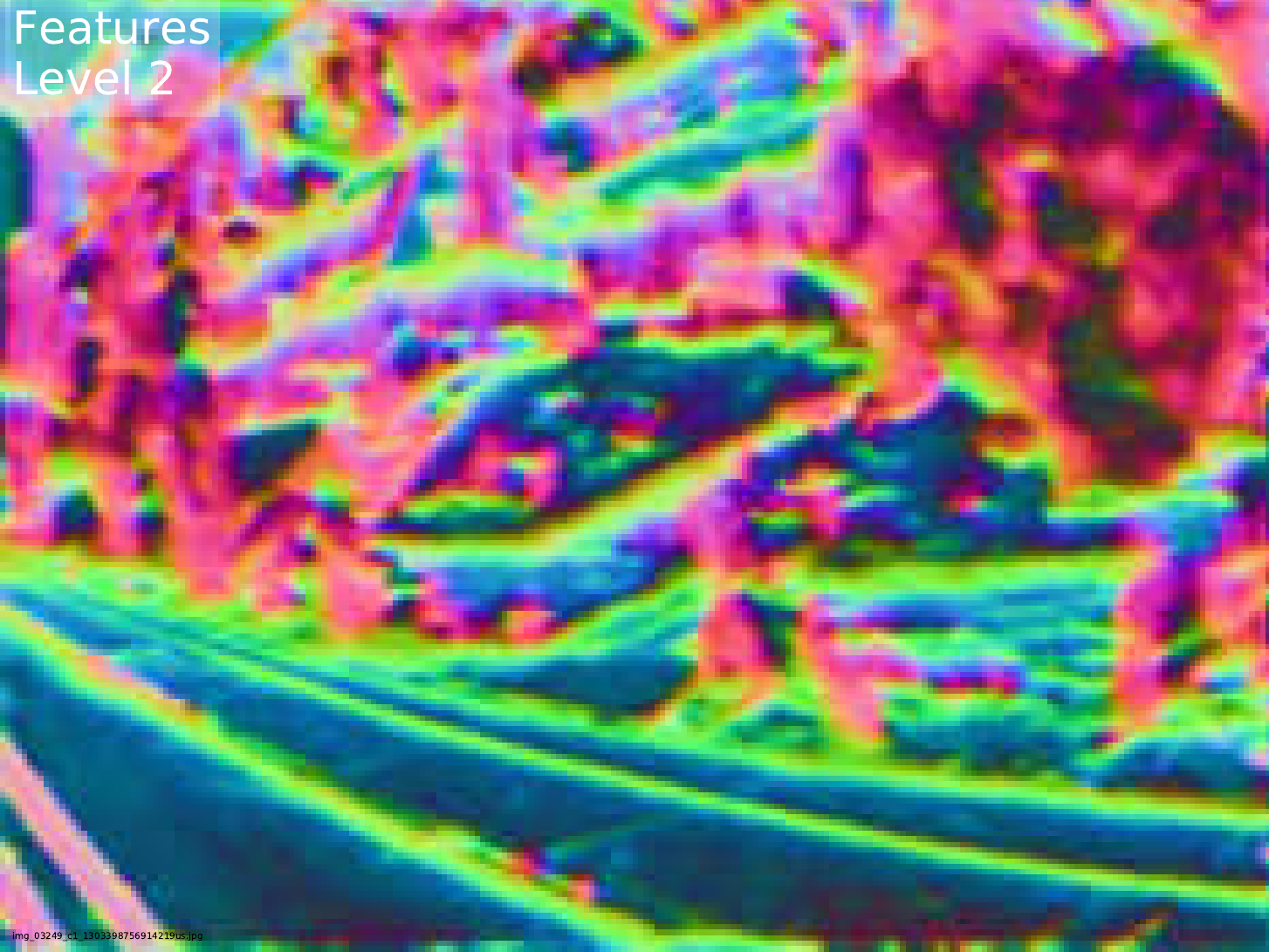}
\end{minipage}%
\begin{minipage}{\iwidth\textwidth}
    \centering
    \includegraphics[width=\pwidth\linewidth]{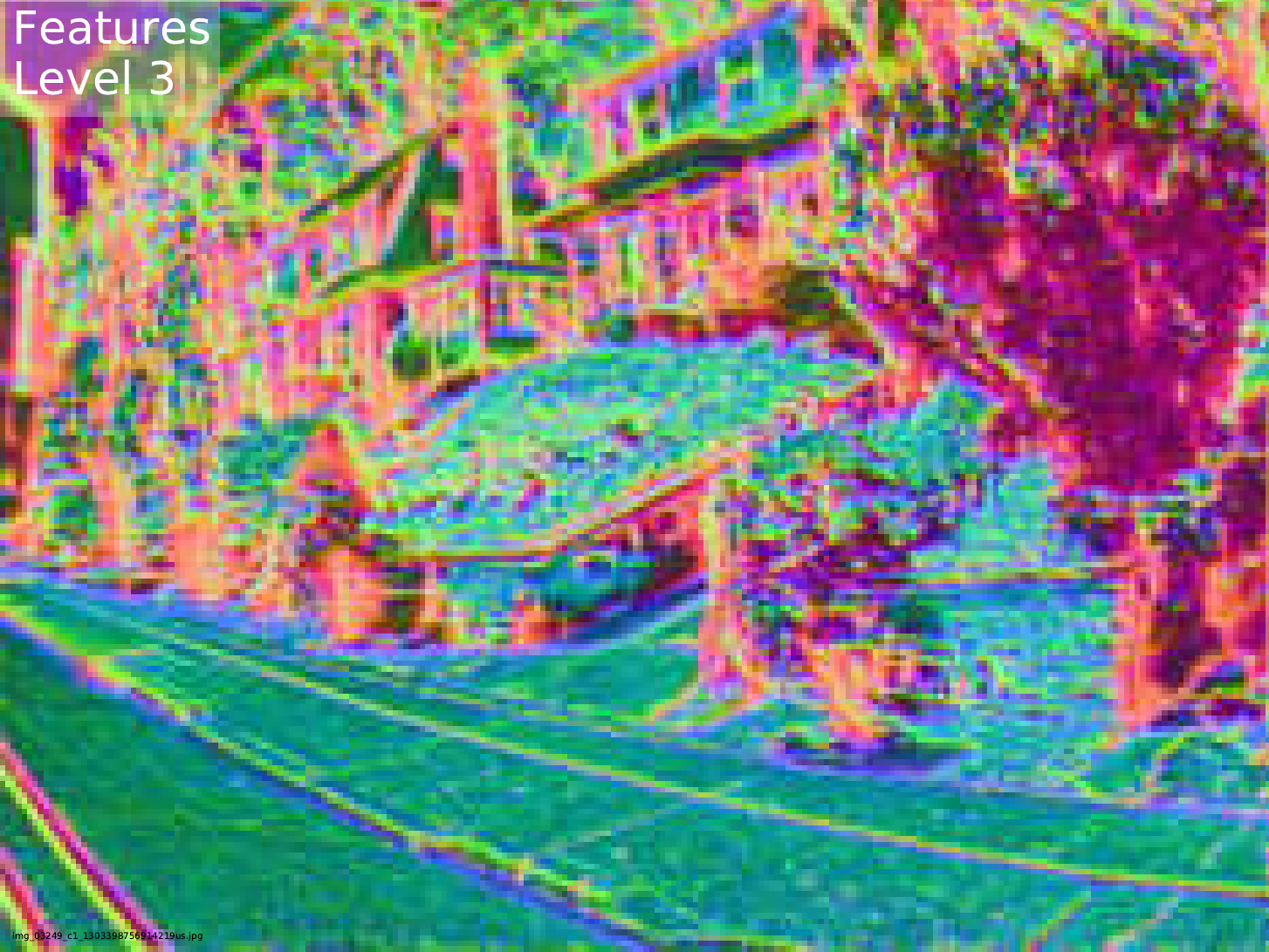}
\end{minipage}%
\begin{minipage}{\iwidth\textwidth}
    \centering
    \includegraphics[width=\pwidth\linewidth]{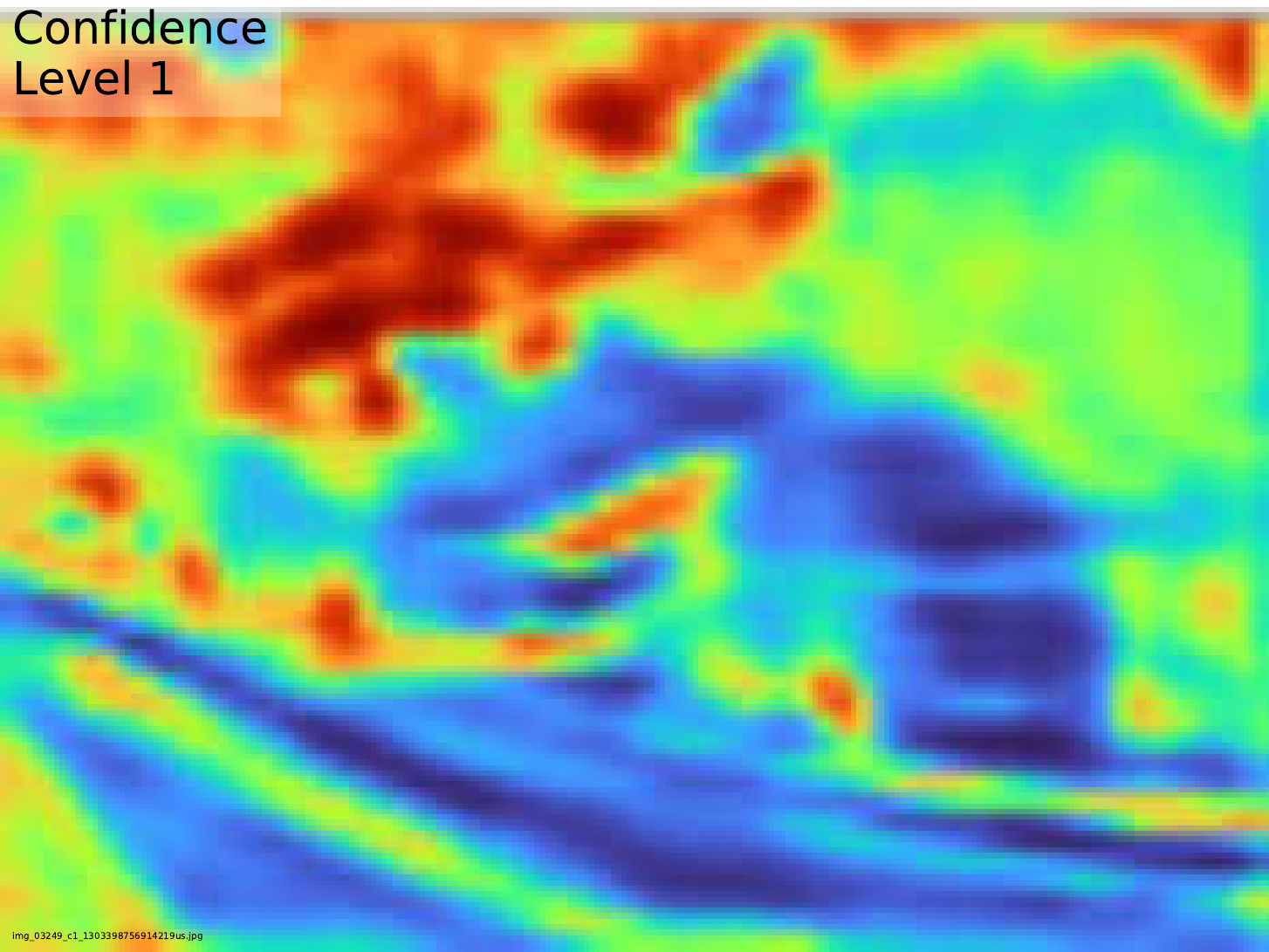}
\end{minipage}%
\begin{minipage}{\iwidth\textwidth}
    \centering
    \includegraphics[width=\pwidth\linewidth]{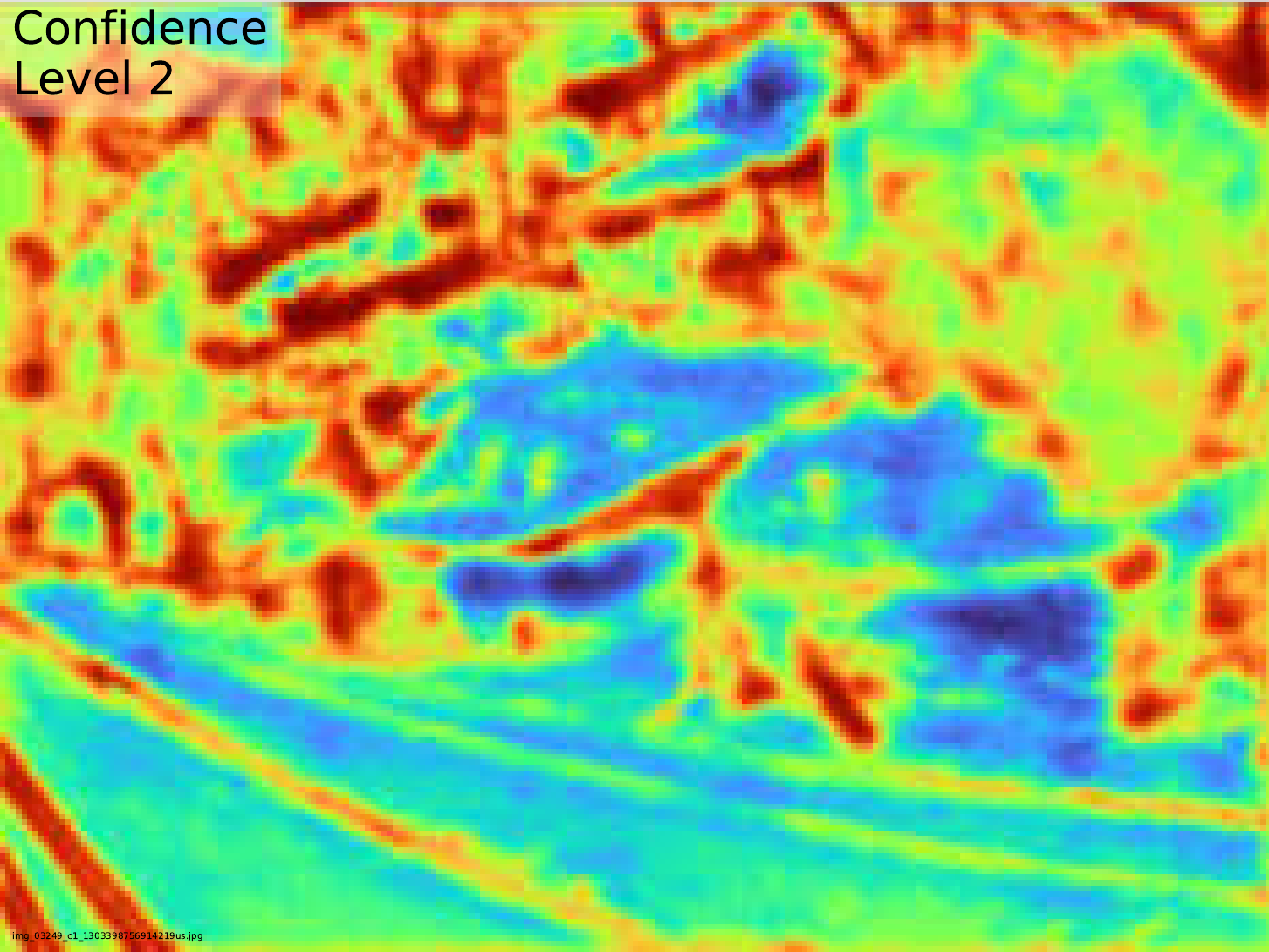}
\end{minipage}%
\begin{minipage}{\iwidth\textwidth}
    \centering
    \includegraphics[width=\pwidth\linewidth]{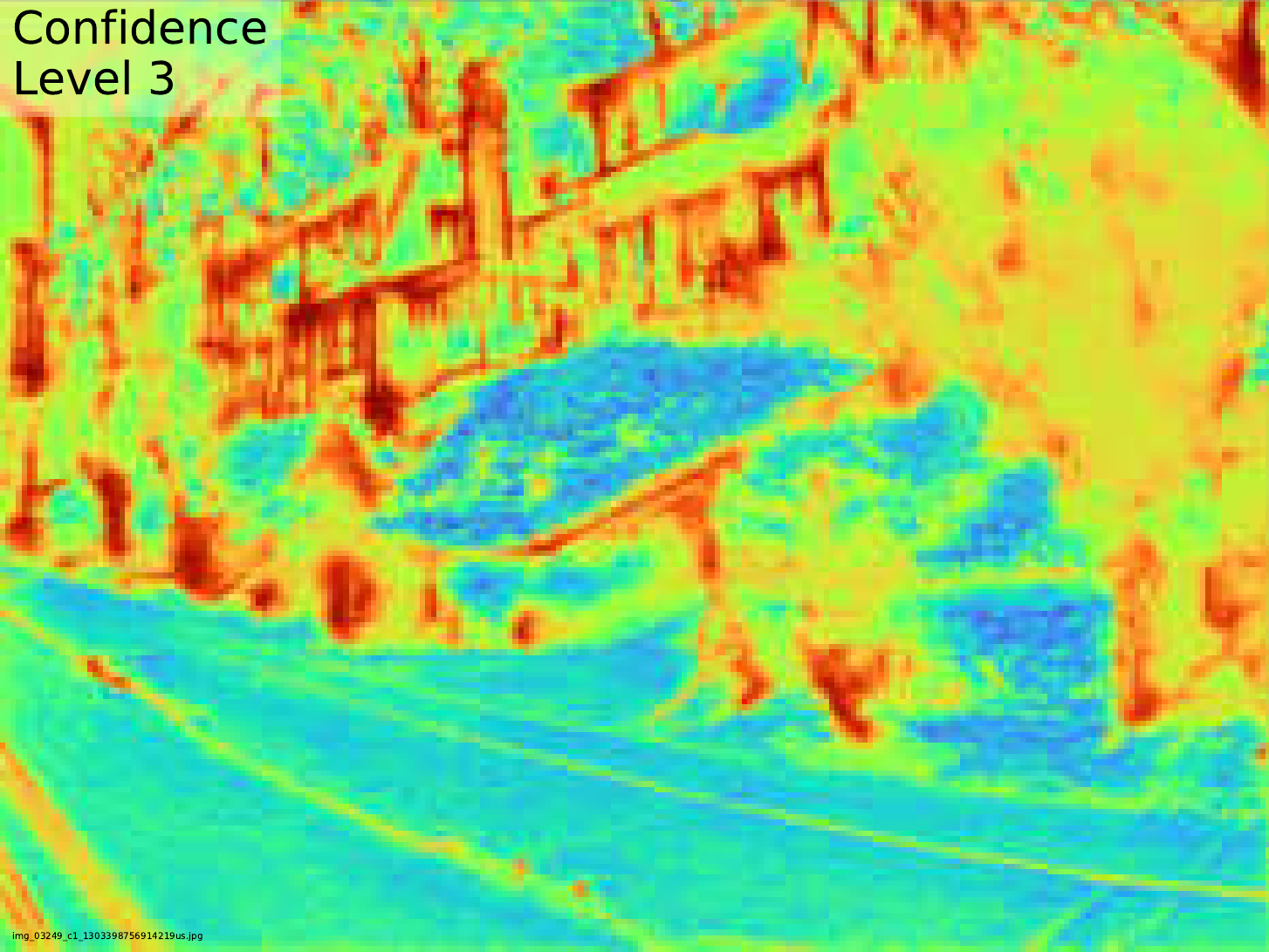}
\end{minipage}
\vspace{2mm}

\begin{minipage}{\lwidth\textwidth}
\rotatebox[origin=c]{90}{Query}
\end{minipage}%
\begin{minipage}{\iwidth\textwidth}
    \centering
    \includegraphics[width=\pwidth\linewidth]{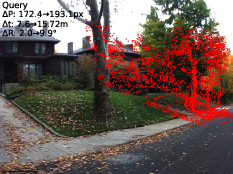}
\end{minipage}%
\begin{minipage}{\iwidth\textwidth}
    \centering
    \includegraphics[width=\pwidth\linewidth]{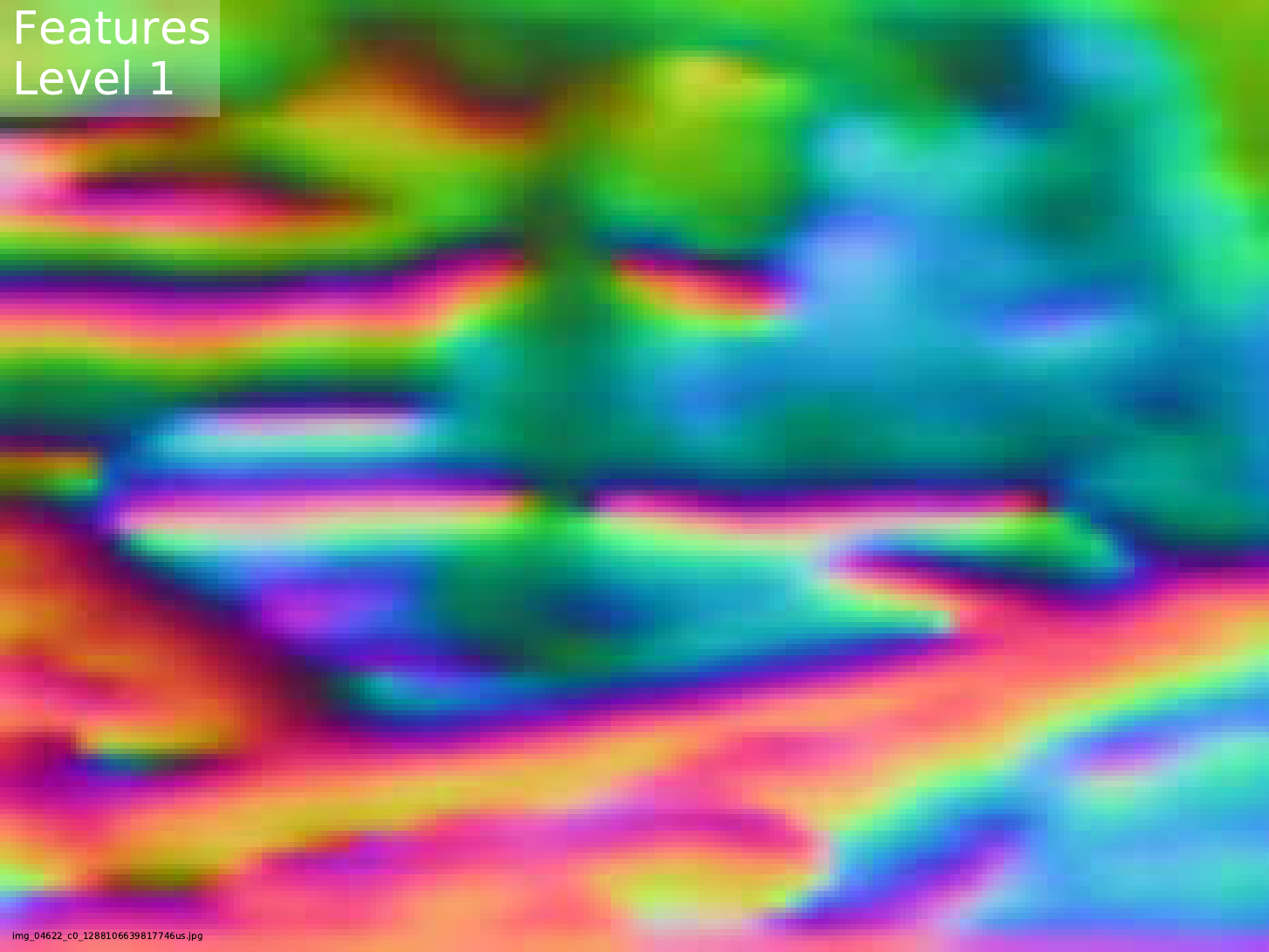}
\end{minipage}%
\begin{minipage}{\iwidth\textwidth}
    \centering
    \includegraphics[width=\pwidth\linewidth]{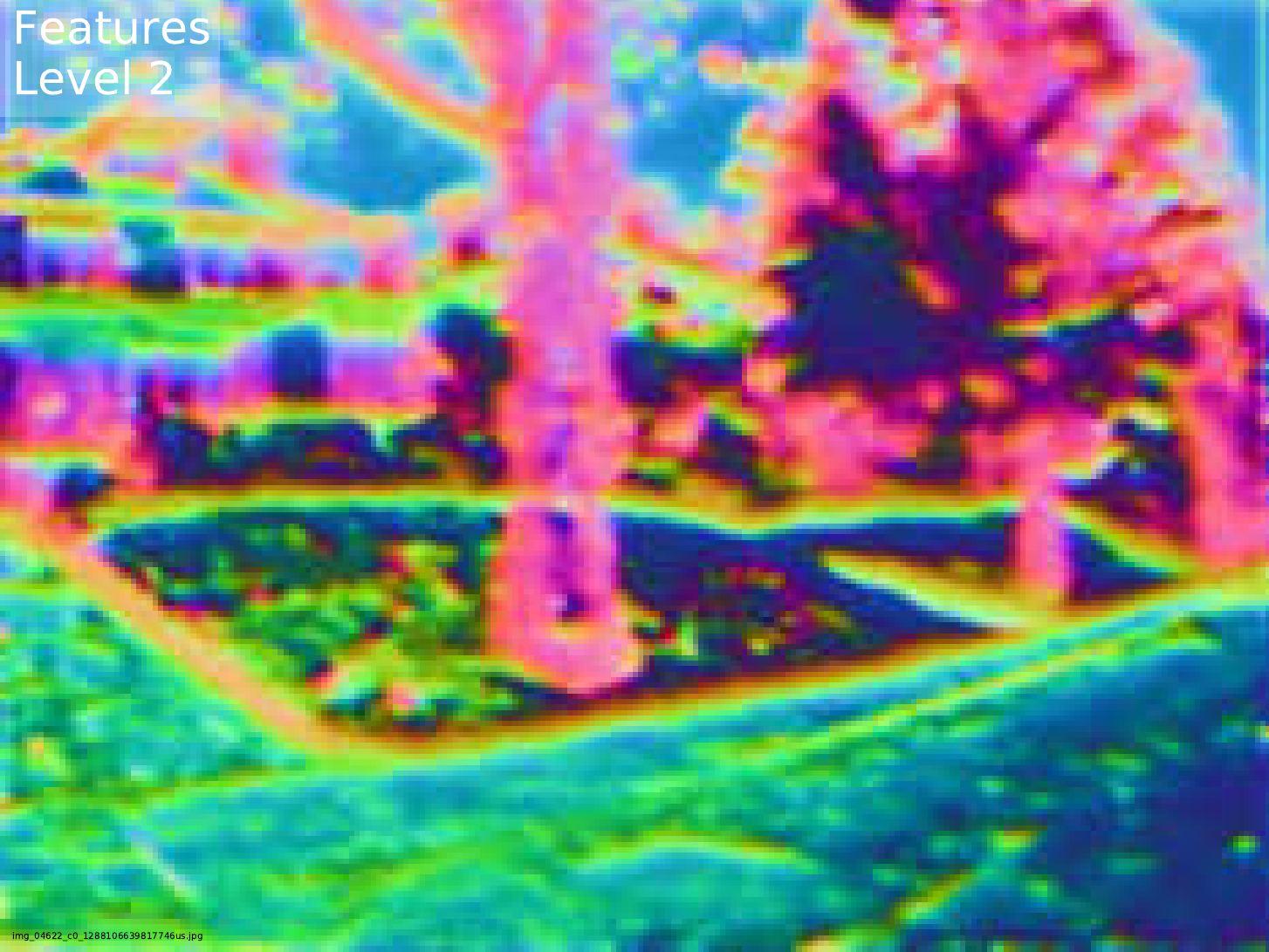}
\end{minipage}%
\begin{minipage}{\iwidth\textwidth}
    \centering
    \includegraphics[width=\pwidth\linewidth]{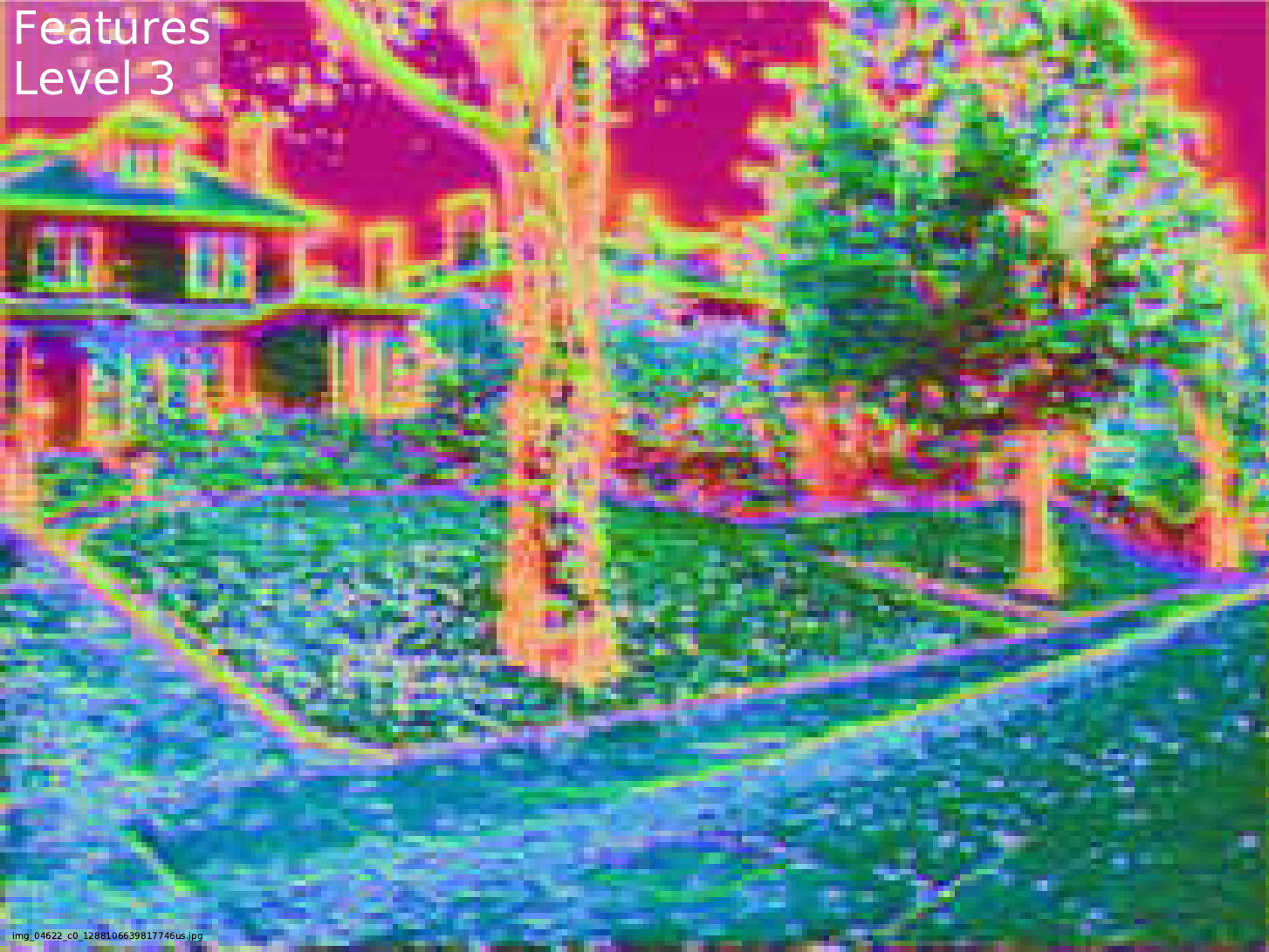}
\end{minipage}%
\begin{minipage}{\iwidth\textwidth}
    \centering
    \includegraphics[width=\pwidth\linewidth]{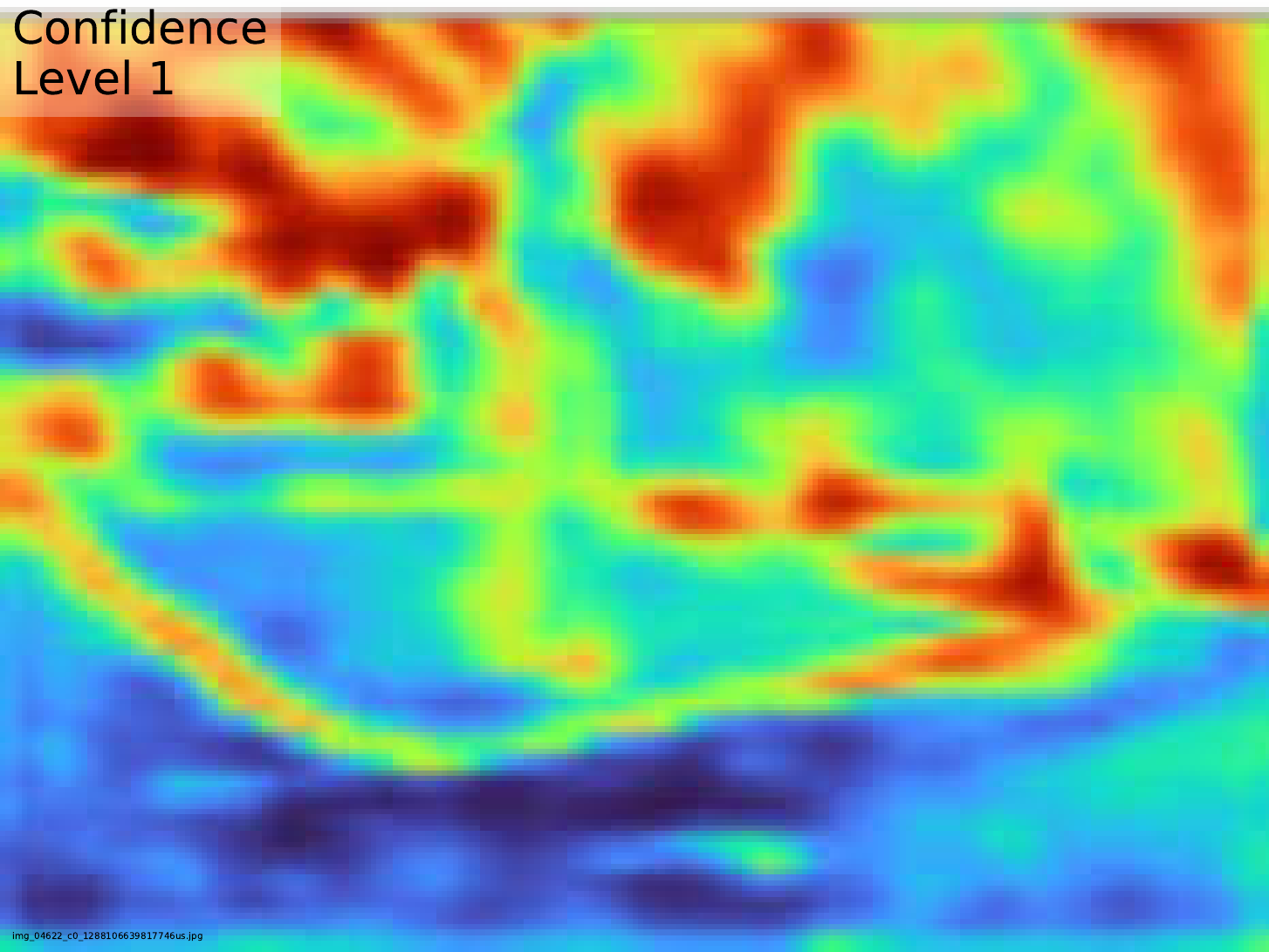}
\end{minipage}%
\begin{minipage}{\iwidth\textwidth}
    \centering
    \includegraphics[width=\pwidth\linewidth]{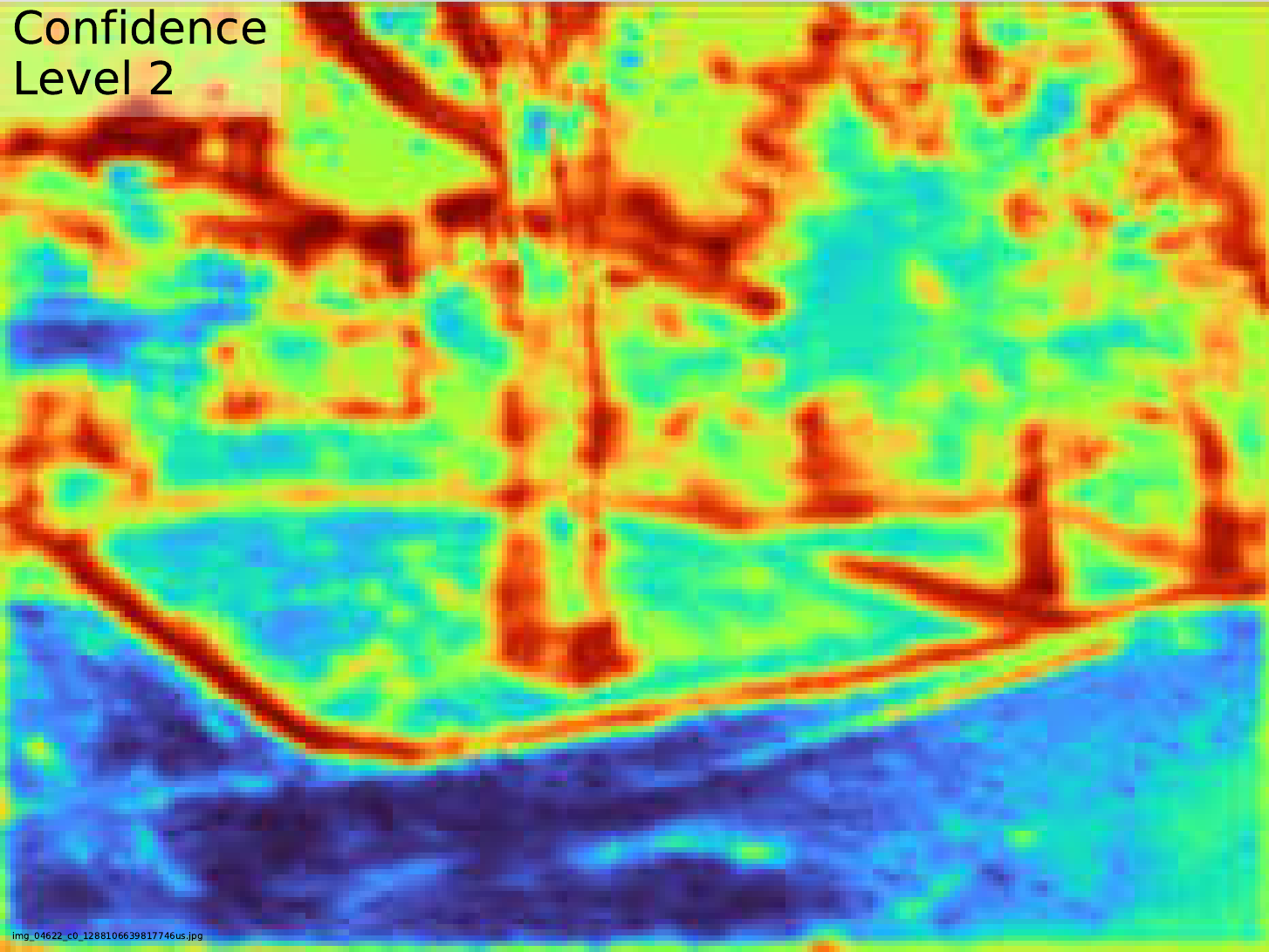}
\end{minipage}%
\begin{minipage}{\iwidth\textwidth}
    \centering
    \includegraphics[width=\pwidth\linewidth]{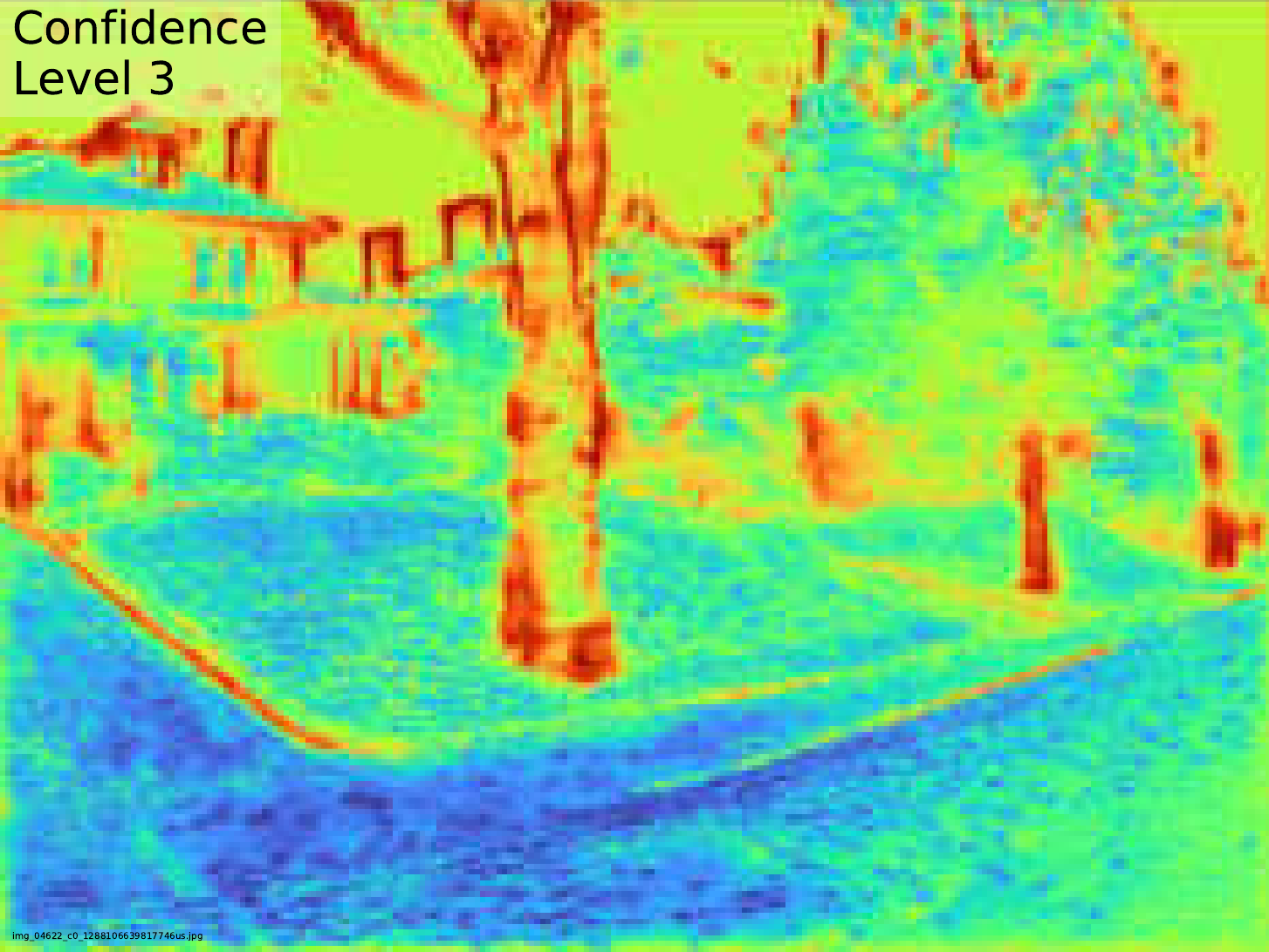}
\end{minipage}
\begin{minipage}{\lwidth\textwidth}
\rotatebox[origin=c]{90}{Reference}
\end{minipage}%
\begin{minipage}{\iwidth\textwidth}
    \centering
    \includegraphics[width=\pwidth\linewidth]{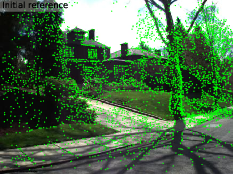}
\end{minipage}%
\begin{minipage}{\iwidth\textwidth}
    \centering
    \includegraphics[width=\pwidth\linewidth]{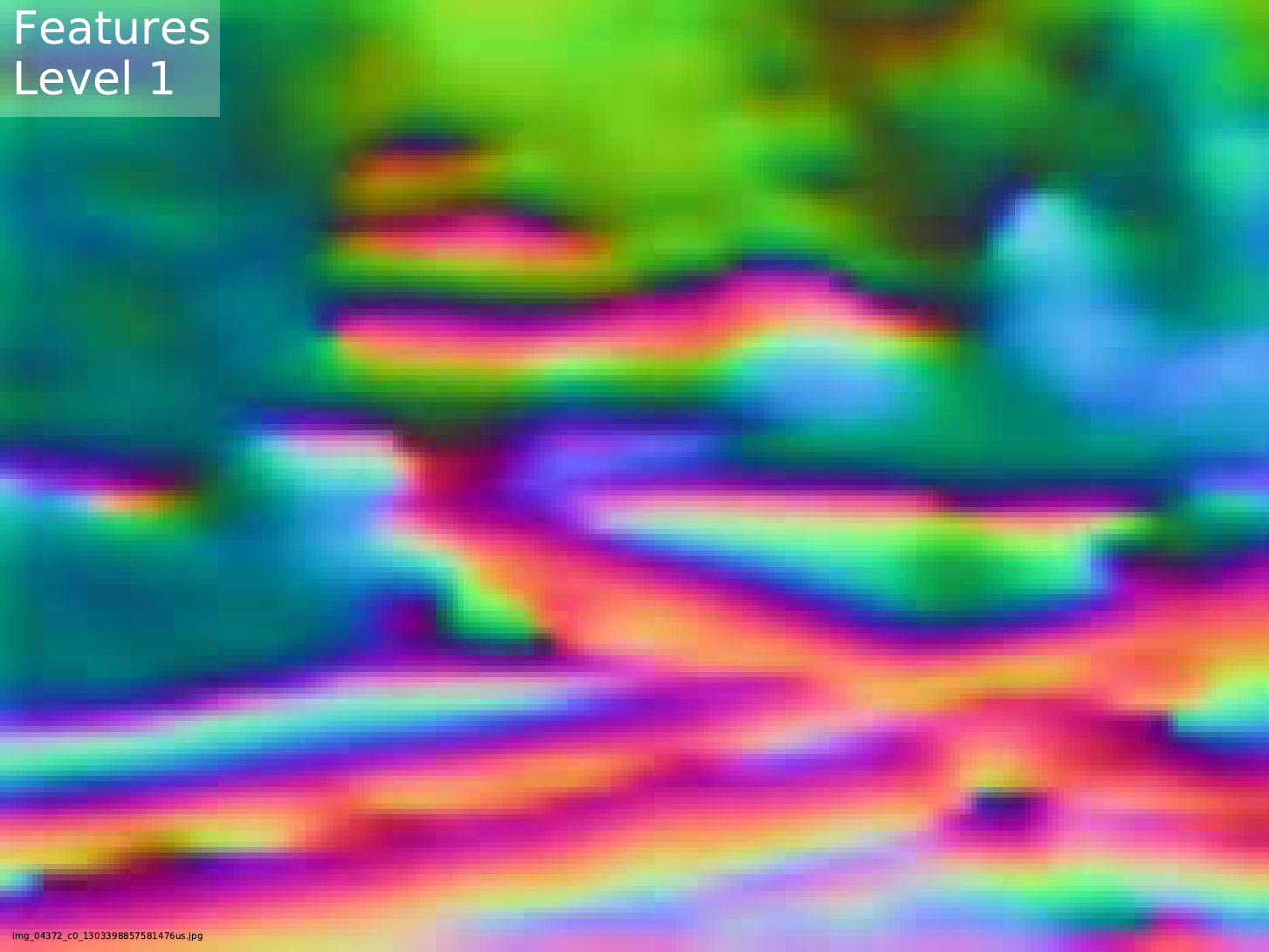}
\end{minipage}%
\begin{minipage}{\iwidth\textwidth}
    \centering
    \includegraphics[width=\pwidth\linewidth]{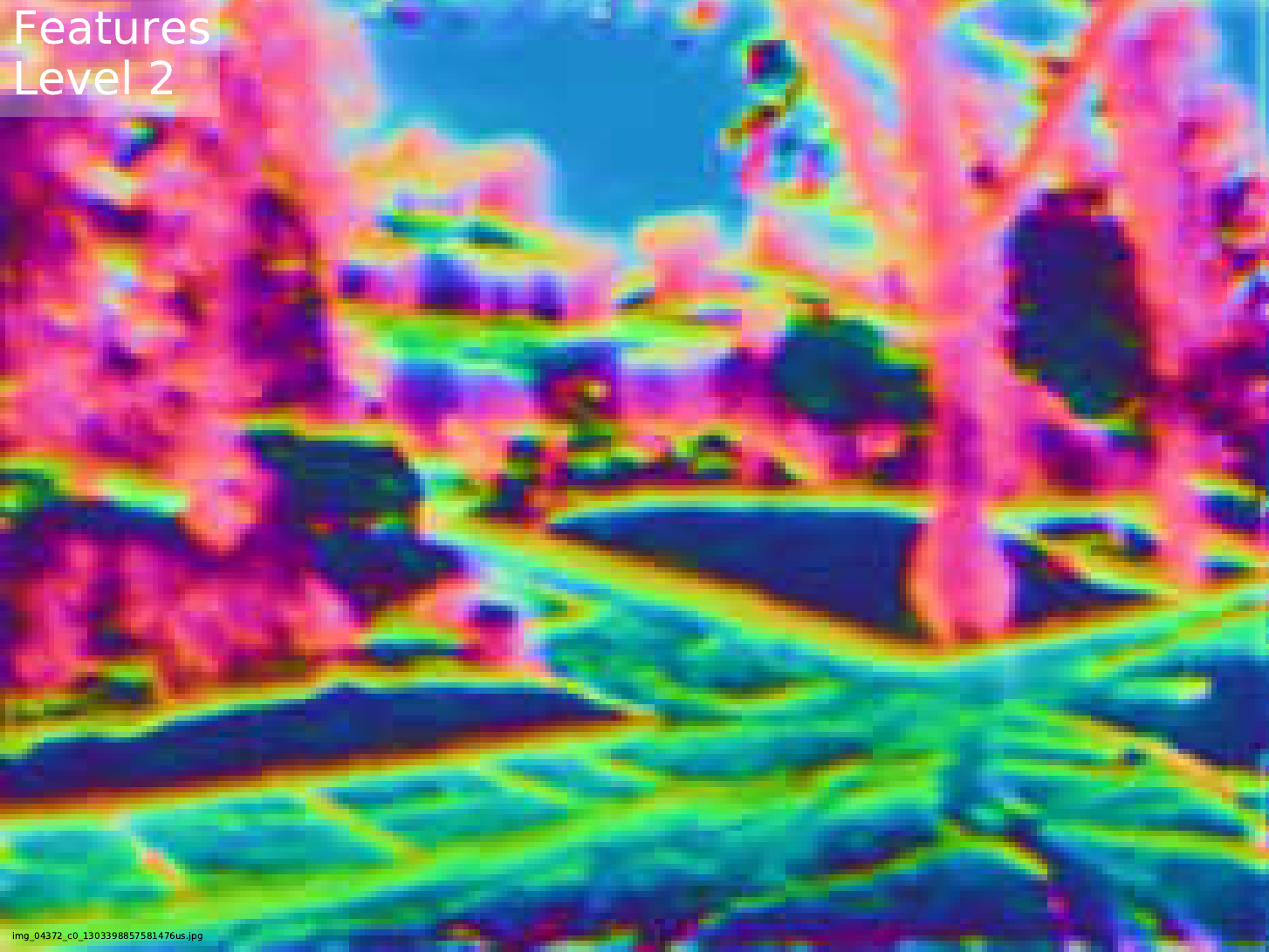}
\end{minipage}%
\begin{minipage}{\iwidth\textwidth}
    \centering
    \includegraphics[width=\pwidth\linewidth]{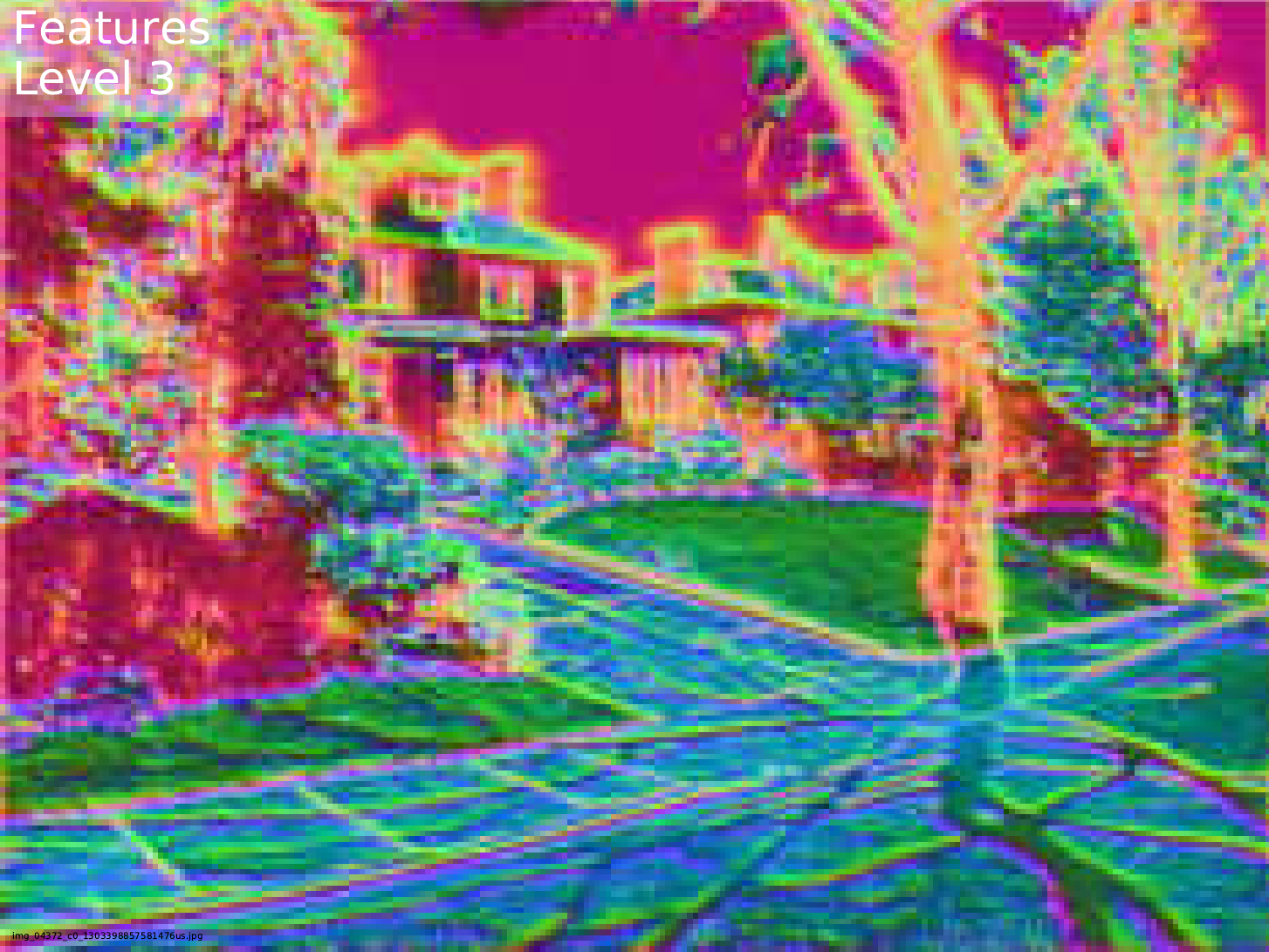}
\end{minipage}%
\begin{minipage}{\iwidth\textwidth}
    \centering
    \includegraphics[width=\pwidth\linewidth]{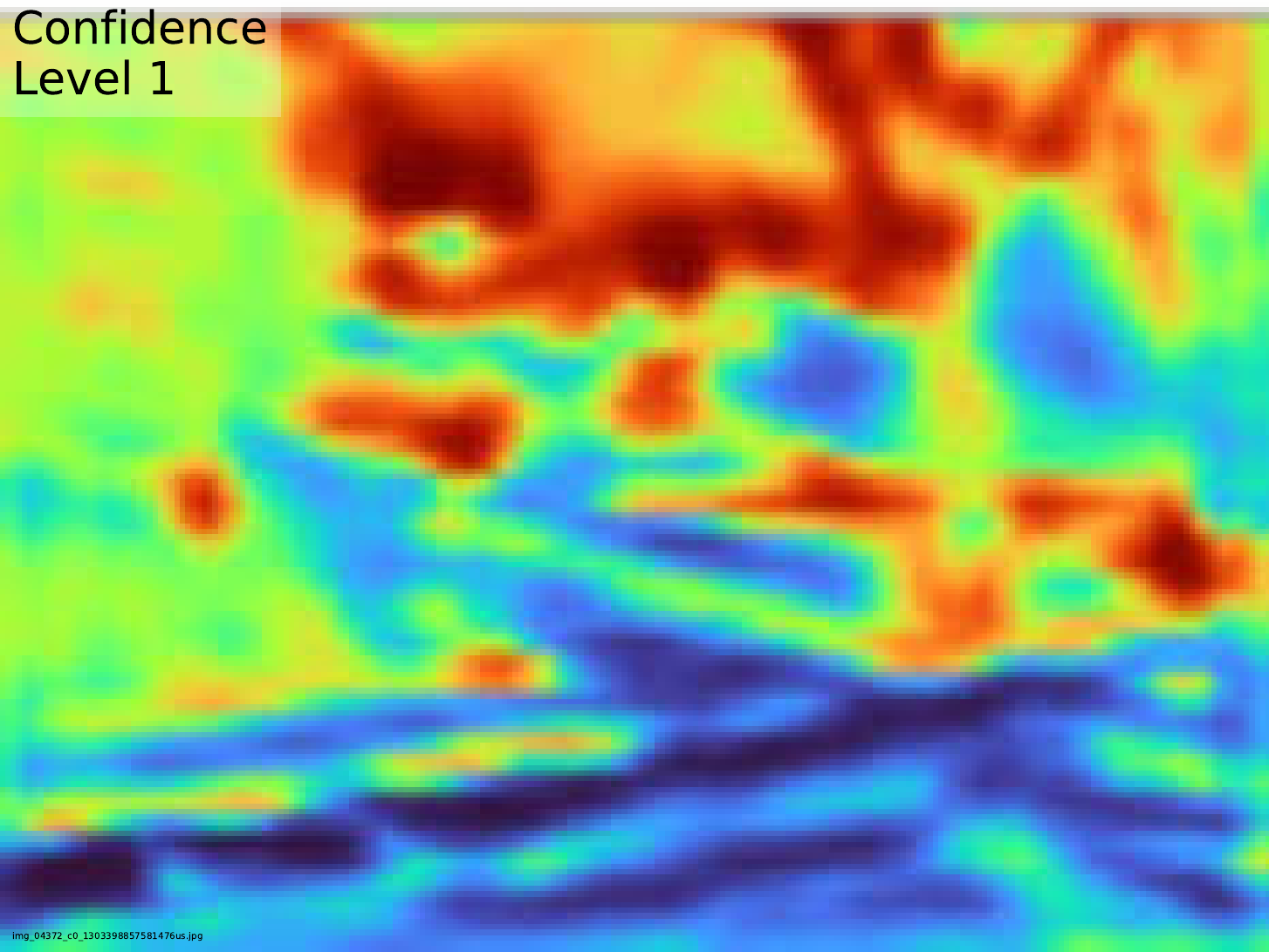}
\end{minipage}%
\begin{minipage}{\iwidth\textwidth}
    \centering
    \includegraphics[width=\pwidth\linewidth]{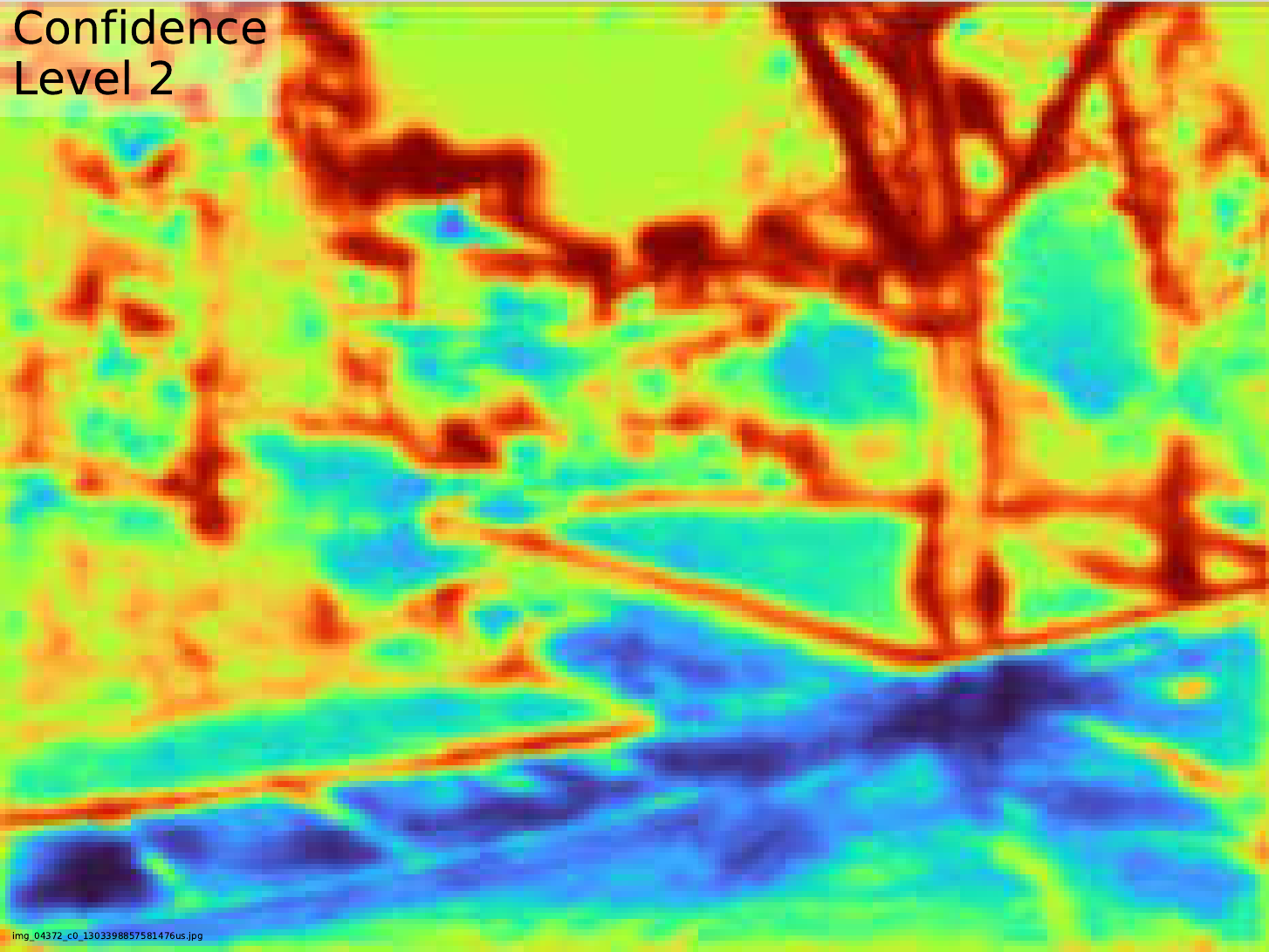}
\end{minipage}%
\begin{minipage}{\iwidth\textwidth}
    \centering
    \includegraphics[width=\pwidth\linewidth]{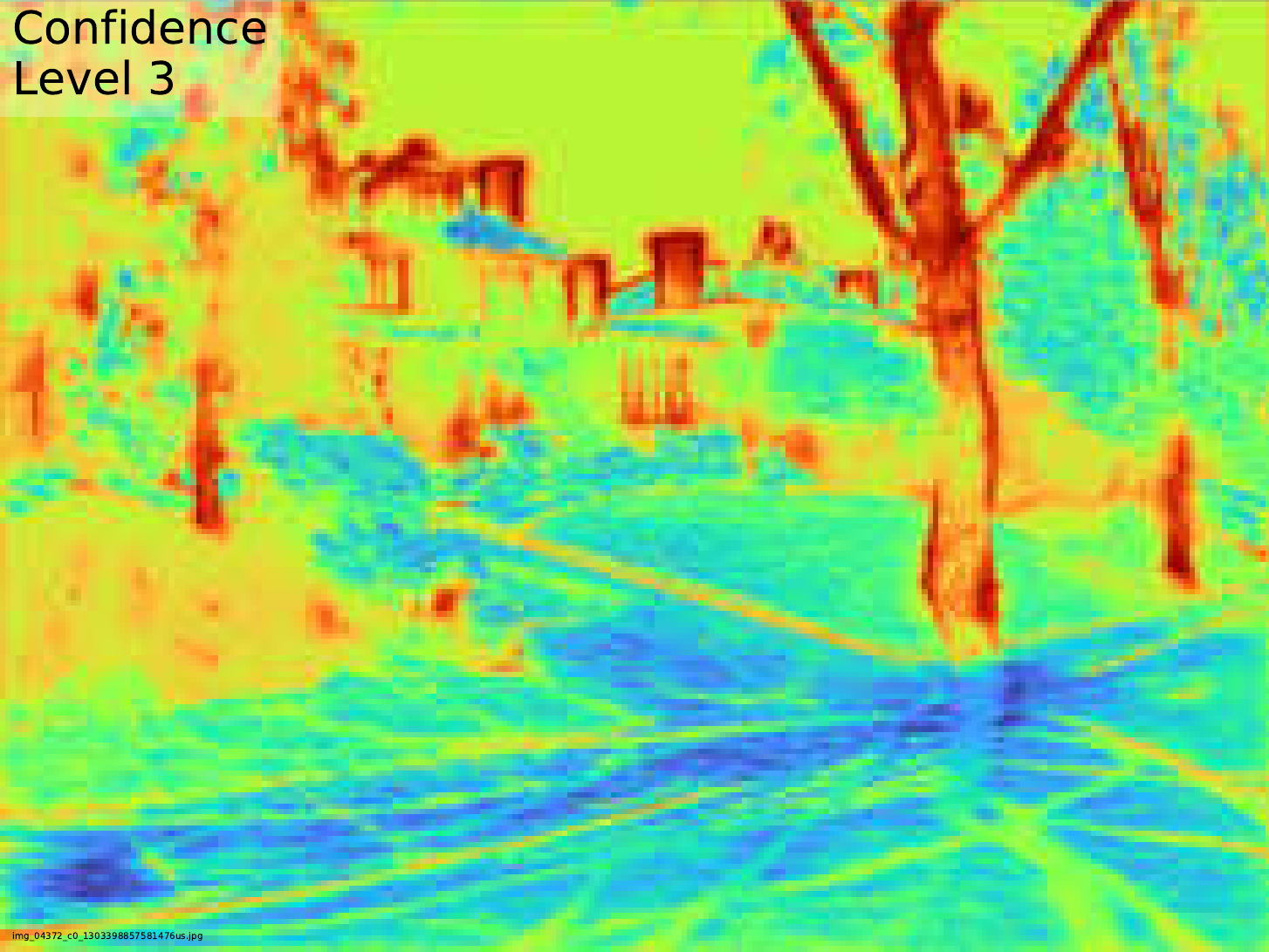}
\end{minipage}
\vspace{2mm}

\begin{minipage}{\lwidth\textwidth}
\rotatebox[origin=c]{90}{Query}
\end{minipage}%
\begin{minipage}{\iwidth\textwidth}
    \centering
    \includegraphics[width=\pwidth\linewidth]{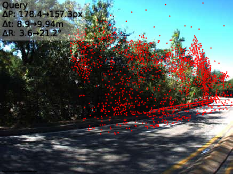}
\end{minipage}%
\begin{minipage}{\iwidth\textwidth}
    \centering
    \includegraphics[width=\pwidth\linewidth]{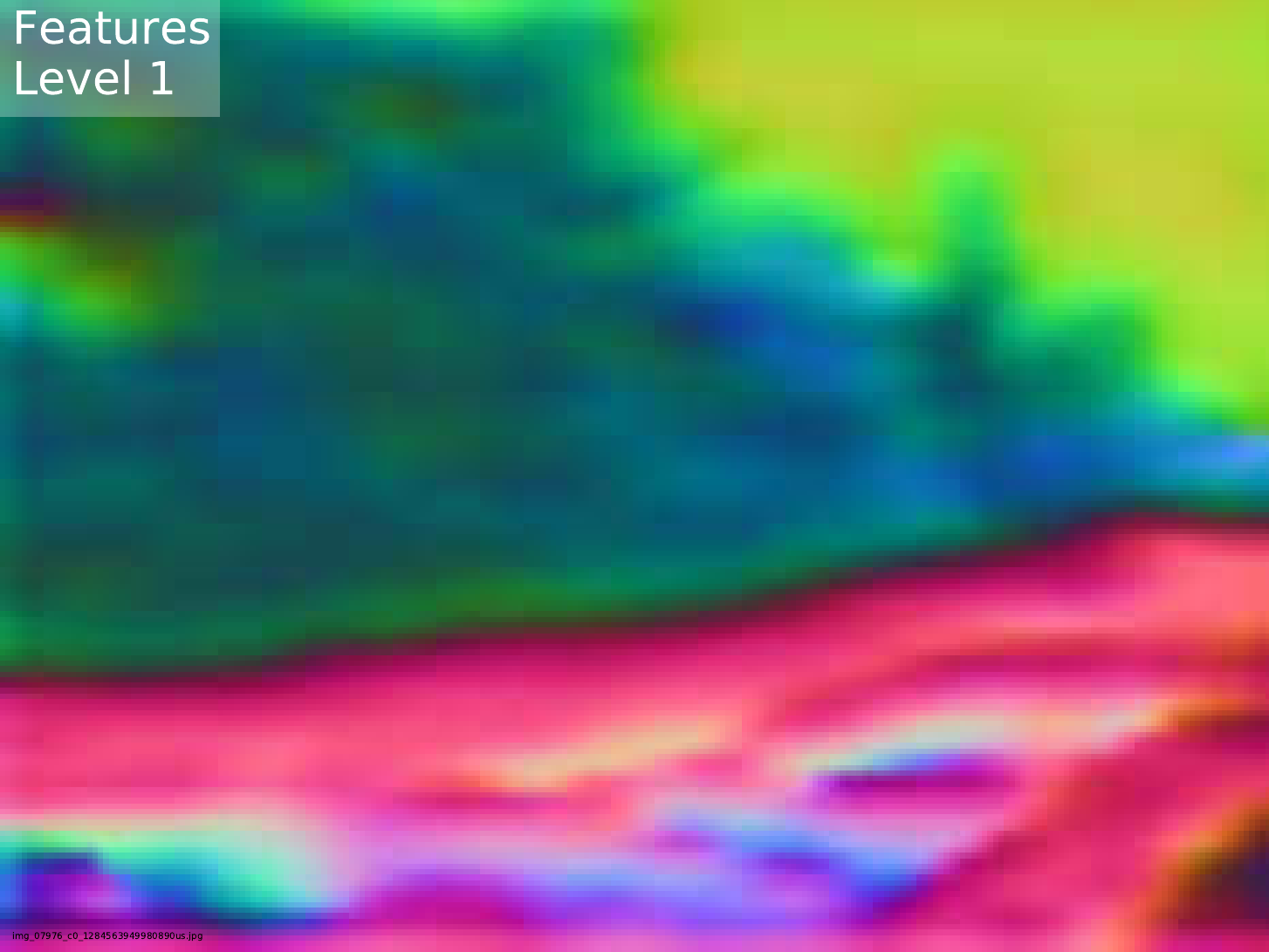}
\end{minipage}%
\begin{minipage}{\iwidth\textwidth}
    \centering
    \includegraphics[width=\pwidth\linewidth]{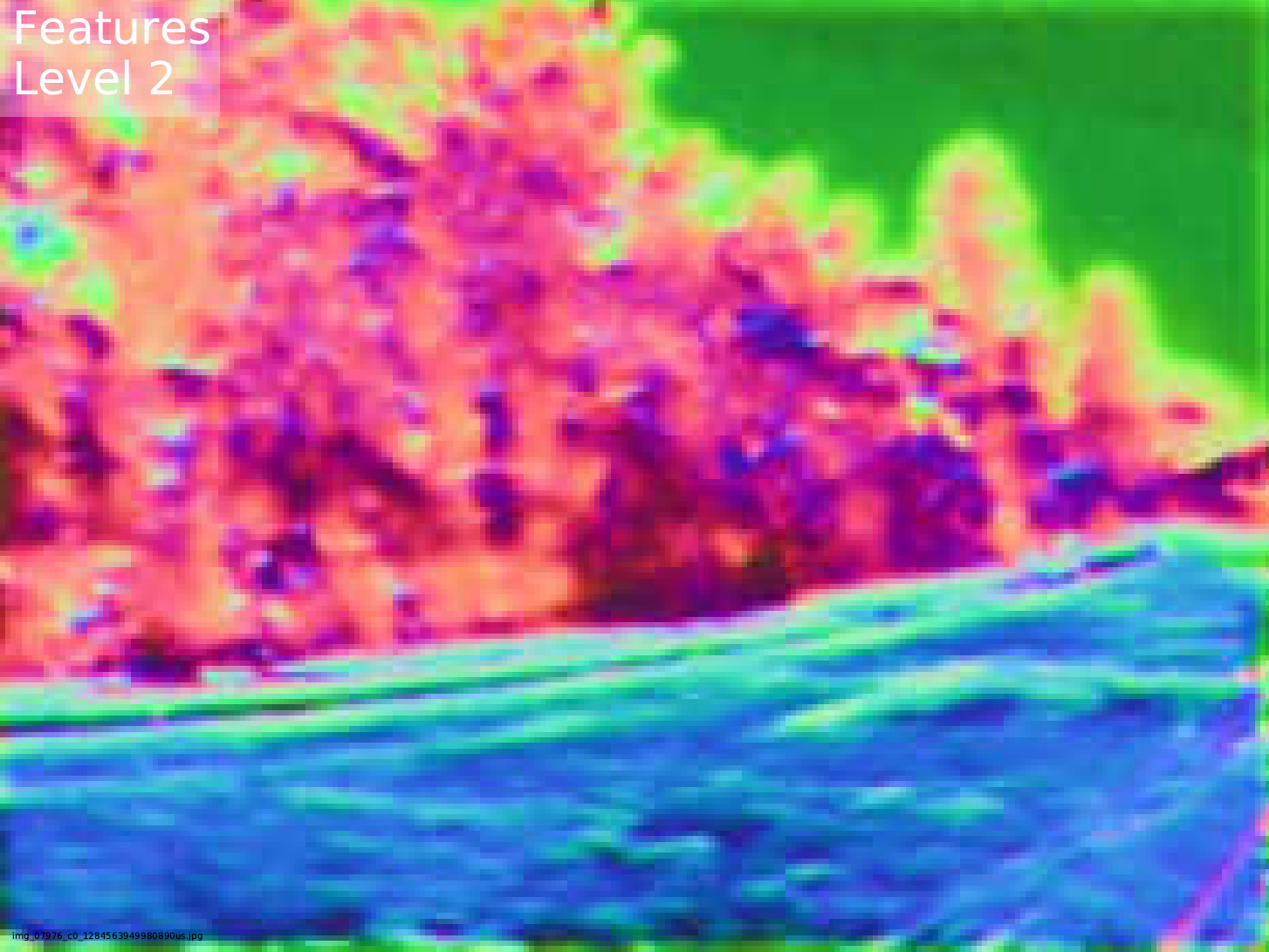}
\end{minipage}%
\begin{minipage}{\iwidth\textwidth}
    \centering
    \includegraphics[width=\pwidth\linewidth]{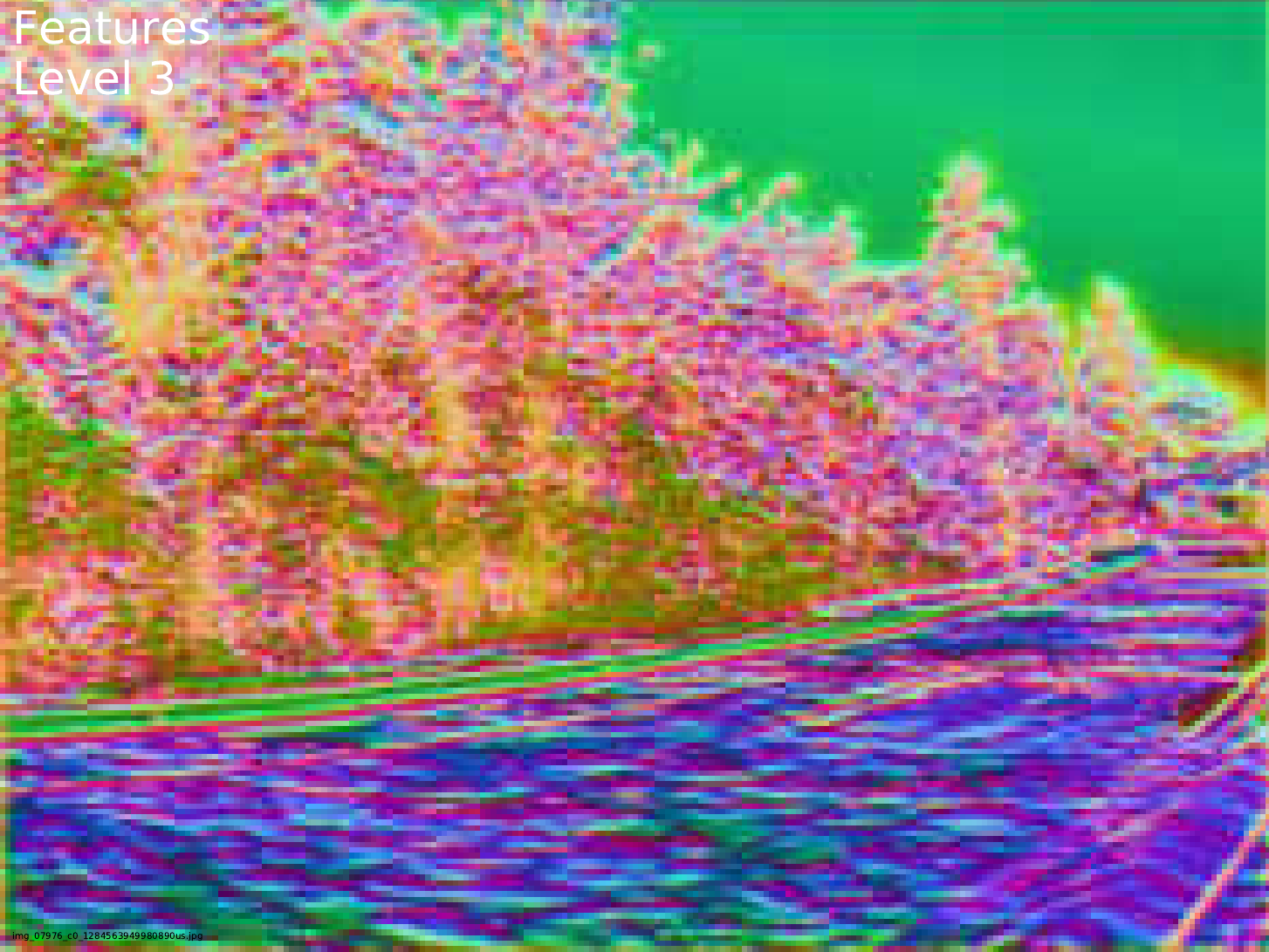}
\end{minipage}%
\begin{minipage}{\iwidth\textwidth}
    \centering
    \includegraphics[width=\pwidth\linewidth]{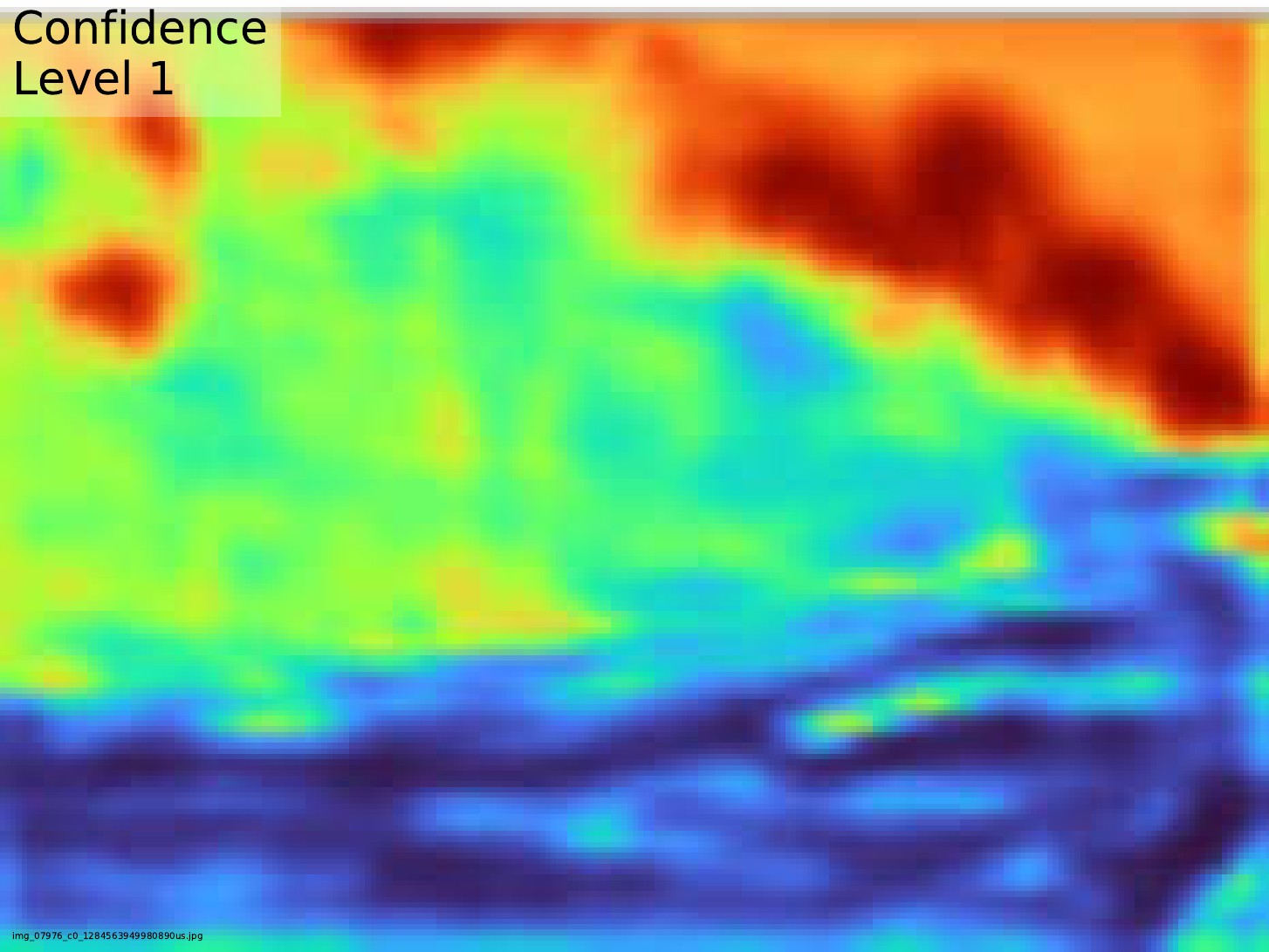}
\end{minipage}%
\begin{minipage}{\iwidth\textwidth}
    \centering
    \includegraphics[width=\pwidth\linewidth]{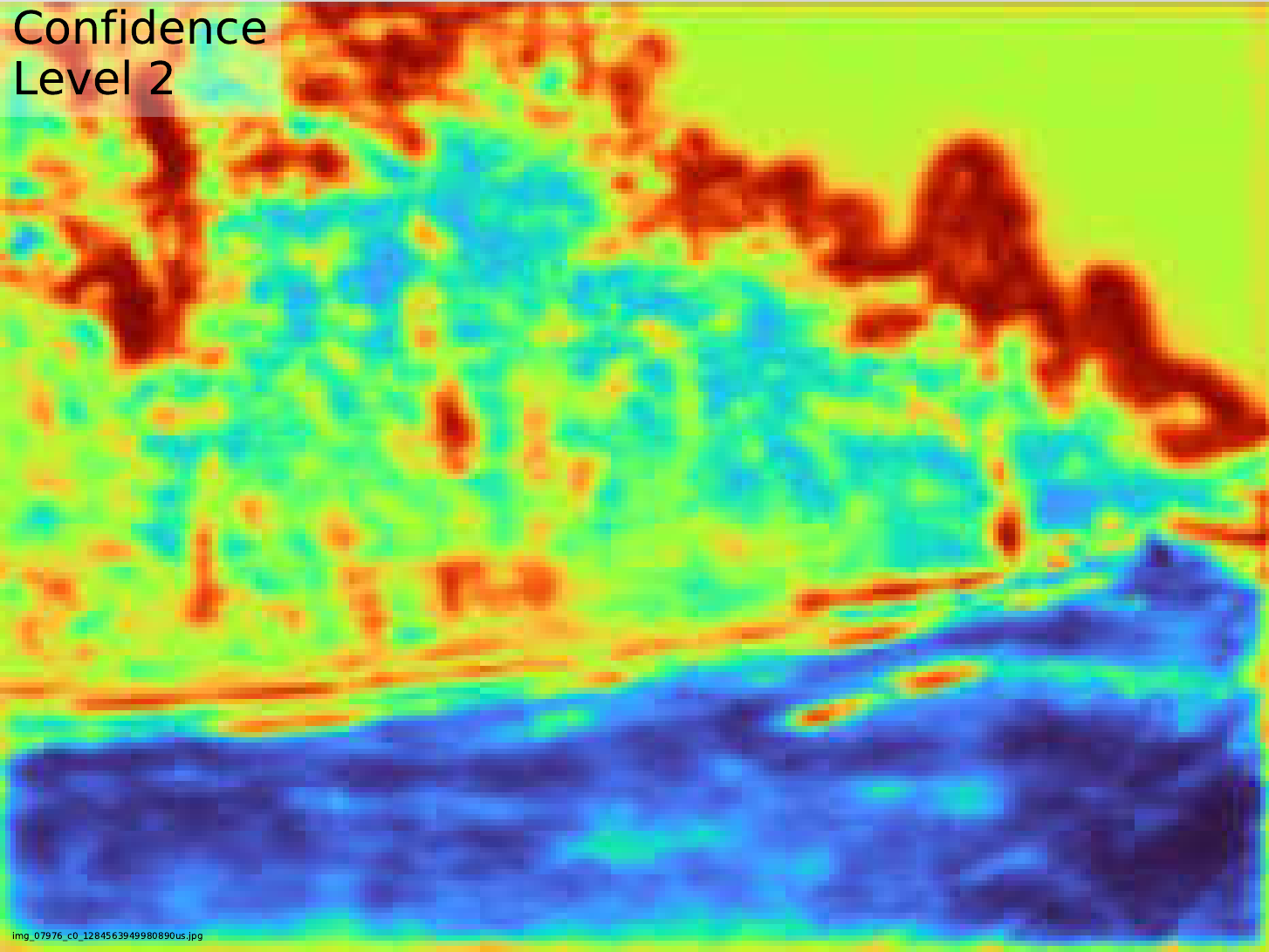}
\end{minipage}%
\begin{minipage}{\iwidth\textwidth}
    \centering
    \includegraphics[width=\pwidth\linewidth]{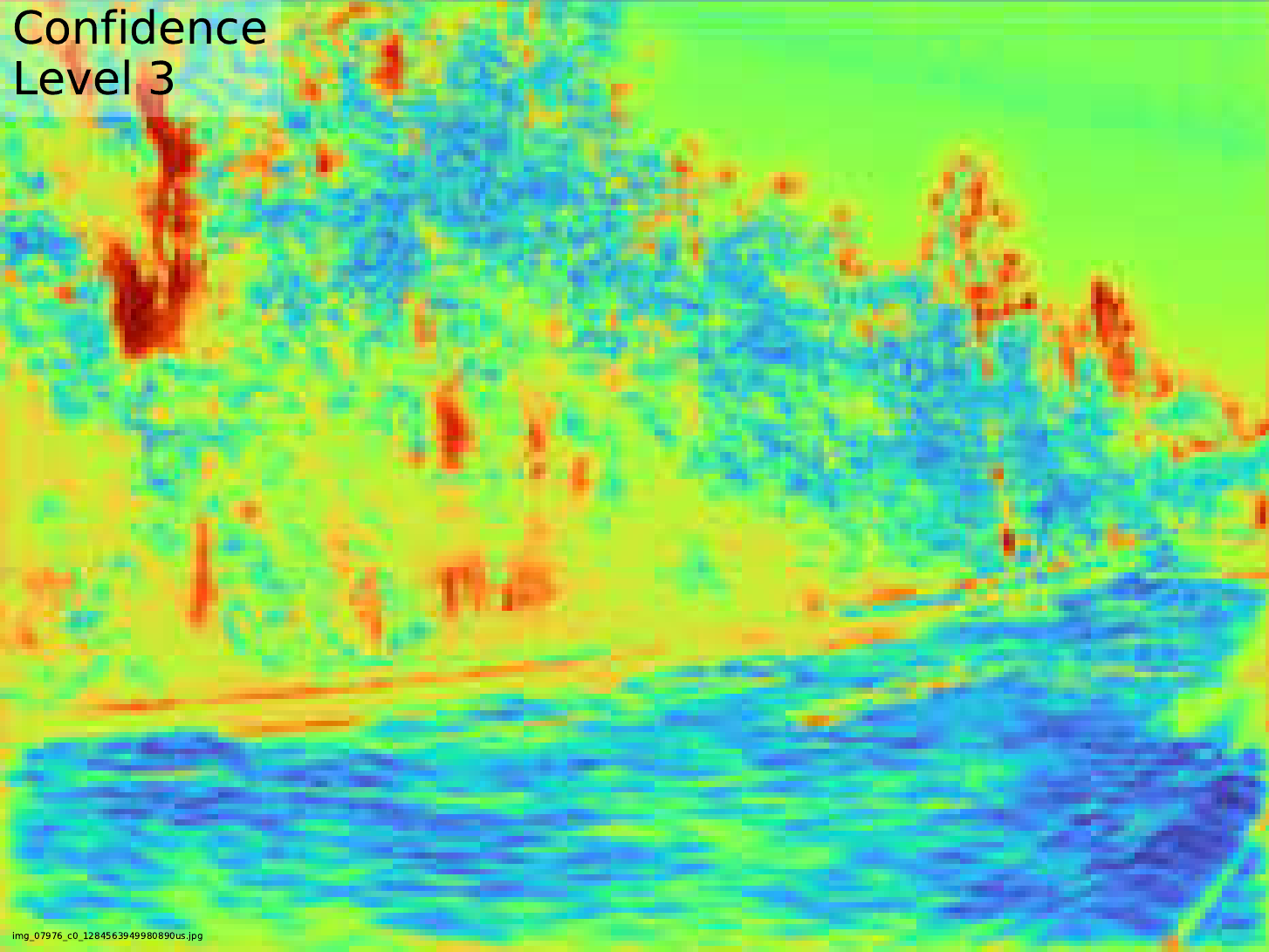}
\end{minipage}
\begin{minipage}{\lwidth\textwidth}
\rotatebox[origin=c]{90}{Reference}
\end{minipage}%
\begin{minipage}{\iwidth\textwidth}
    \centering
    \includegraphics[width=\pwidth\linewidth]{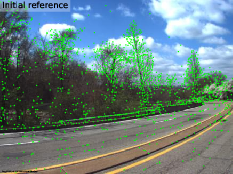}
\end{minipage}%
\begin{minipage}{\iwidth\textwidth}
    \centering
    \includegraphics[width=\pwidth\linewidth]{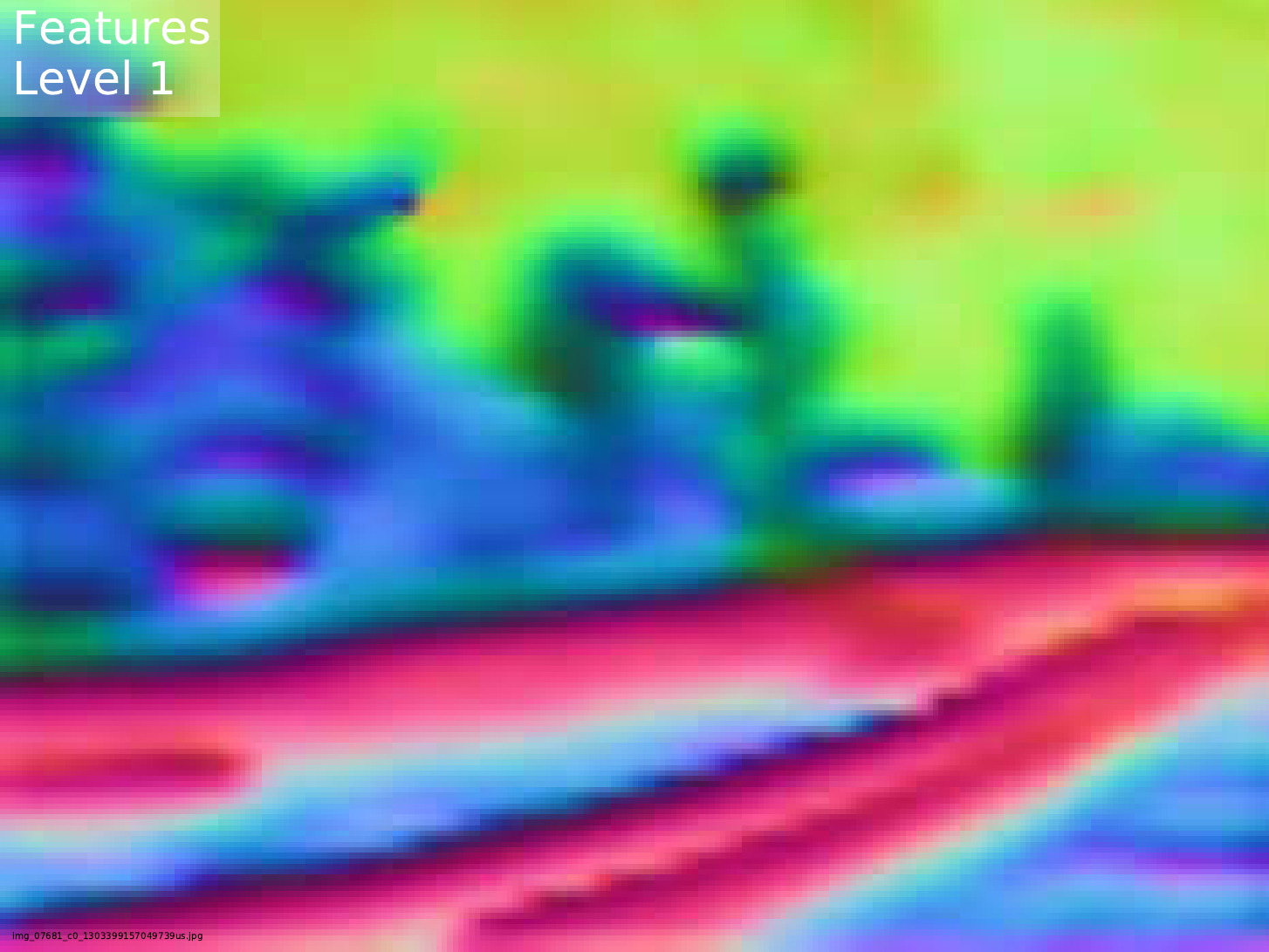}
\end{minipage}%
\begin{minipage}{\iwidth\textwidth}
    \centering
    \includegraphics[width=\pwidth\linewidth]{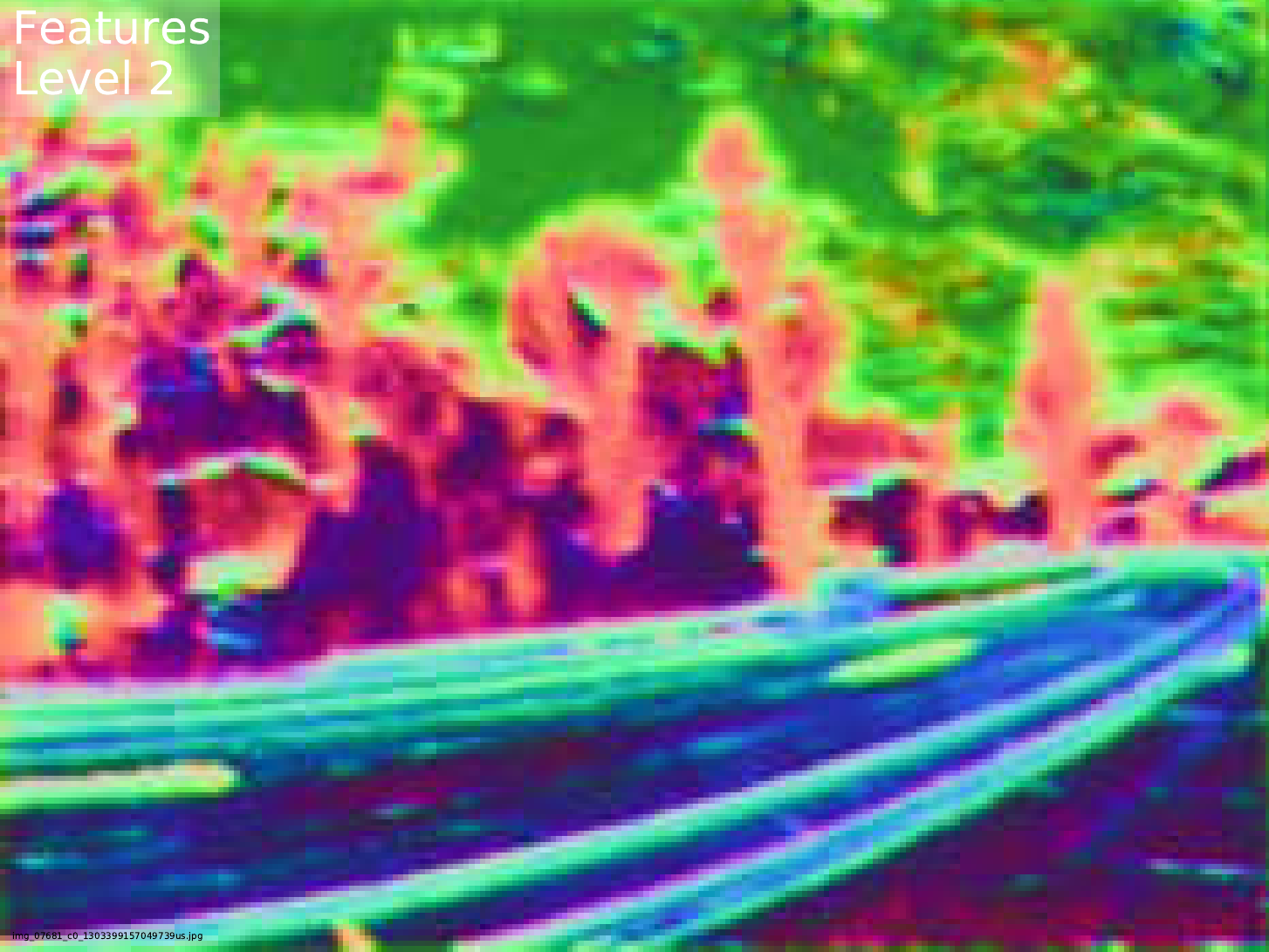}
\end{minipage}%
\begin{minipage}{\iwidth\textwidth}
    \centering
    \includegraphics[width=\pwidth\linewidth]{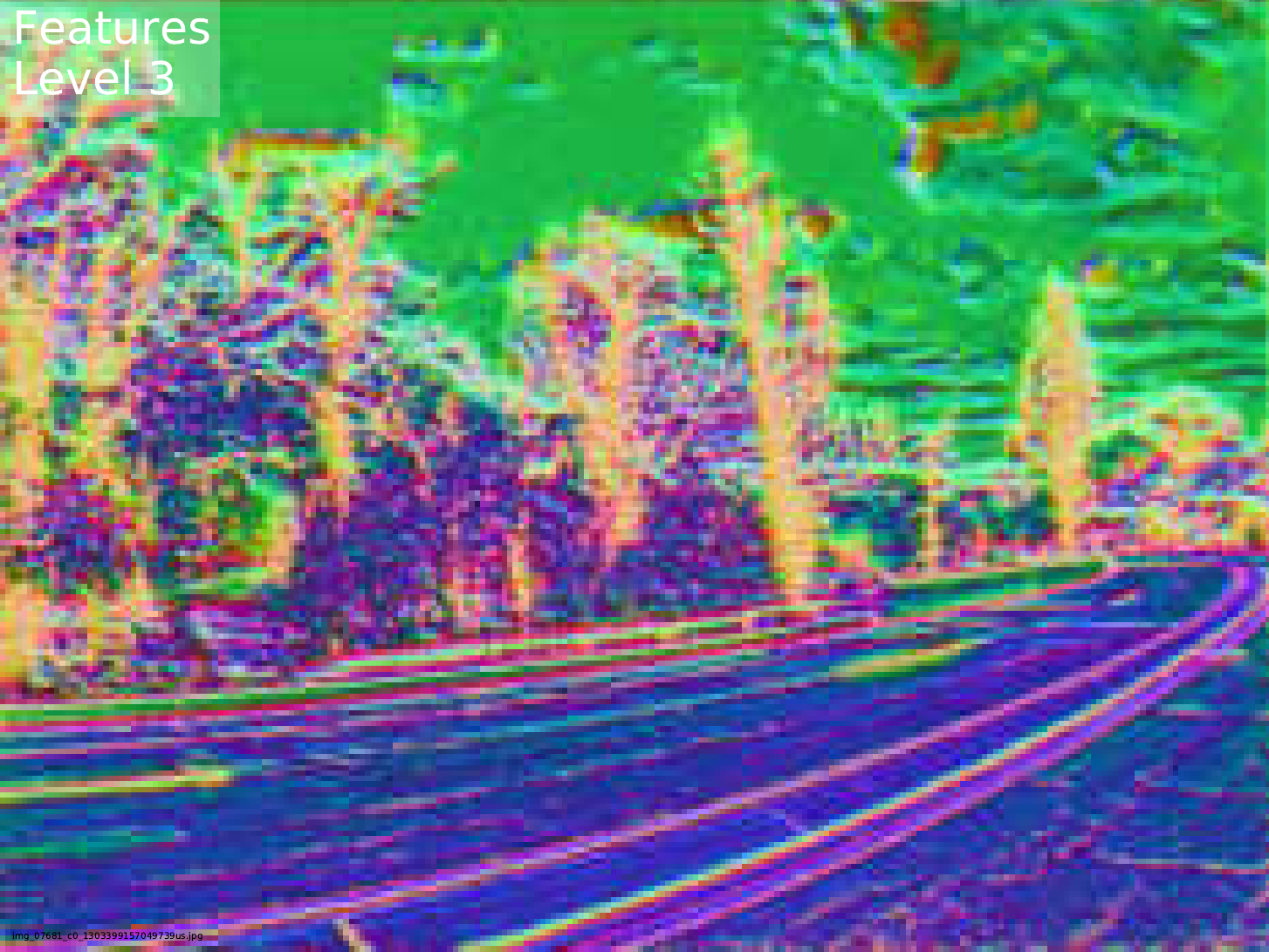}
\end{minipage}%
\begin{minipage}{\iwidth\textwidth}
    \centering
    \includegraphics[width=\pwidth\linewidth]{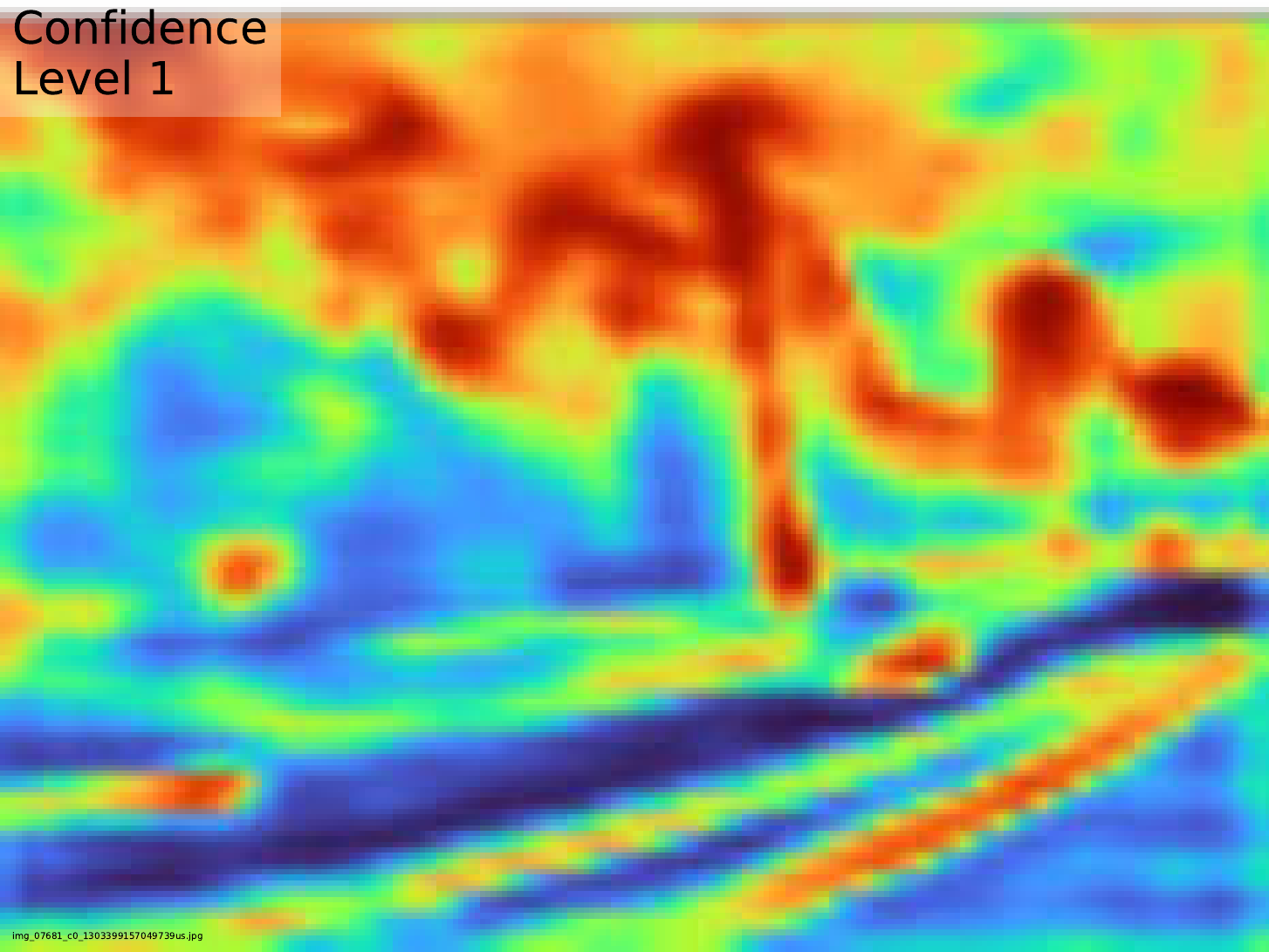}
\end{minipage}%
\begin{minipage}{\iwidth\textwidth}
    \centering
    \includegraphics[width=\pwidth\linewidth]{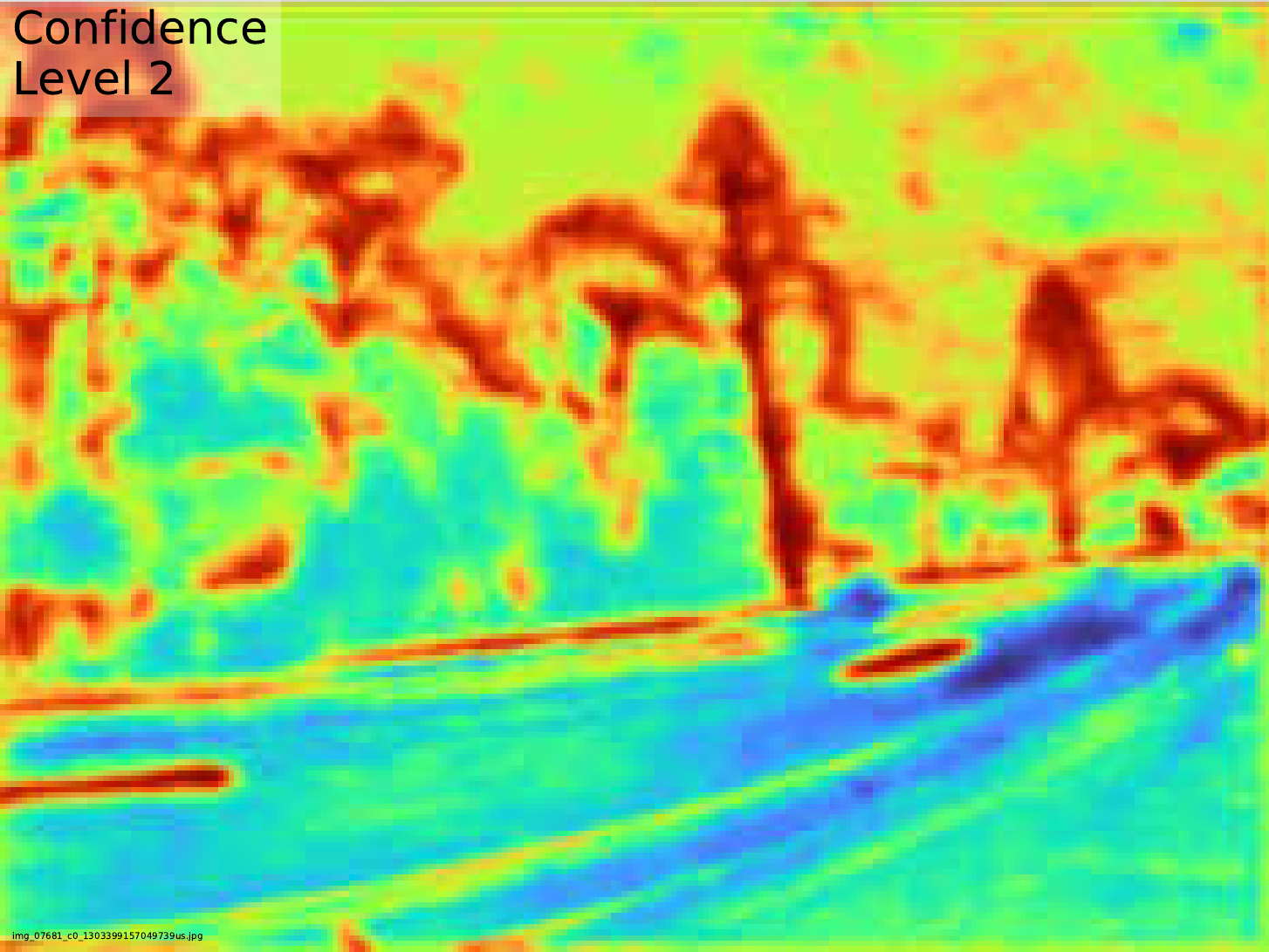}
\end{minipage}%
\begin{minipage}{\iwidth\textwidth}
    \centering
    \includegraphics[width=\pwidth\linewidth]{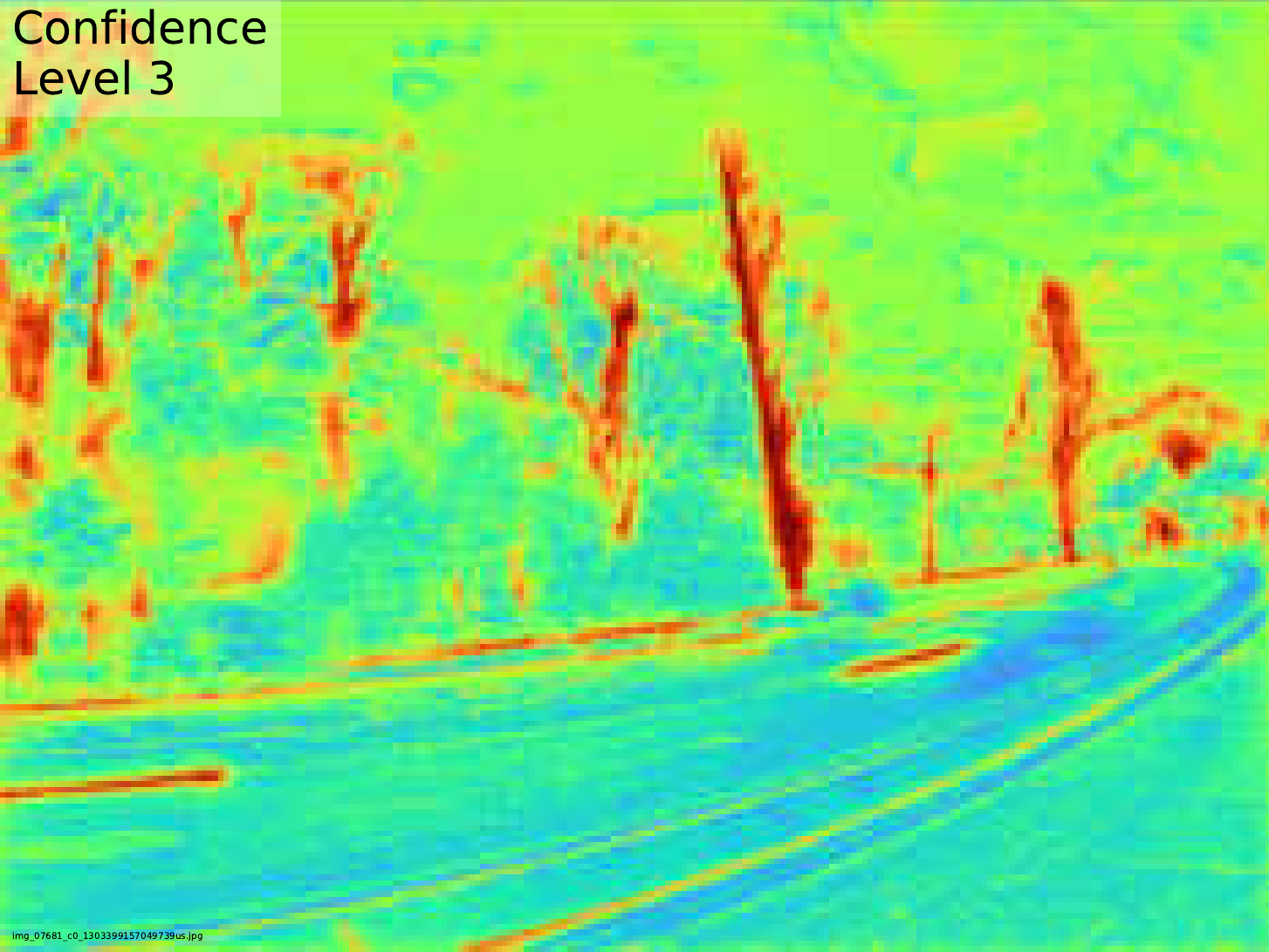}
\end{minipage}
\vspace{2mm}

    \caption{\textbf{Failure cases on the CMU dataset.}
    We show examples for which the optimization results in a large final error.
    This is often due to repeated elements or to a lack of spatial context of the coarse features or a lack of distinctive elements.
    Natural scenes are be particularly challenging when tree trunks and vegetation cannot be easily distinguished.
    }
    \label{fig:qualitative_cmu_failure}%
\end{figure*}

\begin{figure*}[t]
    \centering
\def\iwidth{0.14}
\def\pwidth{0.99}
\def\lwidth{0.020}
\def\rcwidth{0.14}

\begin{minipage}{\lwidth\textwidth}
\hfill
\end{minipage}%
\begin{minipage}{\iwidth\textwidth}
    \centering
    \small{Images}
\end{minipage}%
\begin{minipage}{\iwidth\textwidth*\real{3.0}}
    \centering
    \small{Features}
    \vspace{0.5mm}
    \hrule width 0.99\linewidth
    \vspace{0.2mm}
\end{minipage}%
\begin{minipage}{\iwidth\textwidth*\real{3.0}}
    \centering
    \small{Confidence}
    \vspace{0.5mm}
    \hrule width 0.99\linewidth
    \vspace{0.2mm}
\end{minipage}%

\begin{minipage}{\lwidth\textwidth}
\rotatebox[origin=c]{90}{Query}
\end{minipage}%
\begin{minipage}{\iwidth\textwidth}
    \centering
    \includegraphics[width=\pwidth\linewidth]{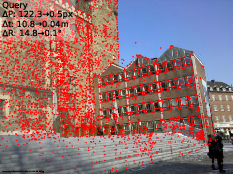}
\end{minipage}%
\begin{minipage}{\iwidth\textwidth}
    \centering
    \includegraphics[width=\pwidth\linewidth]{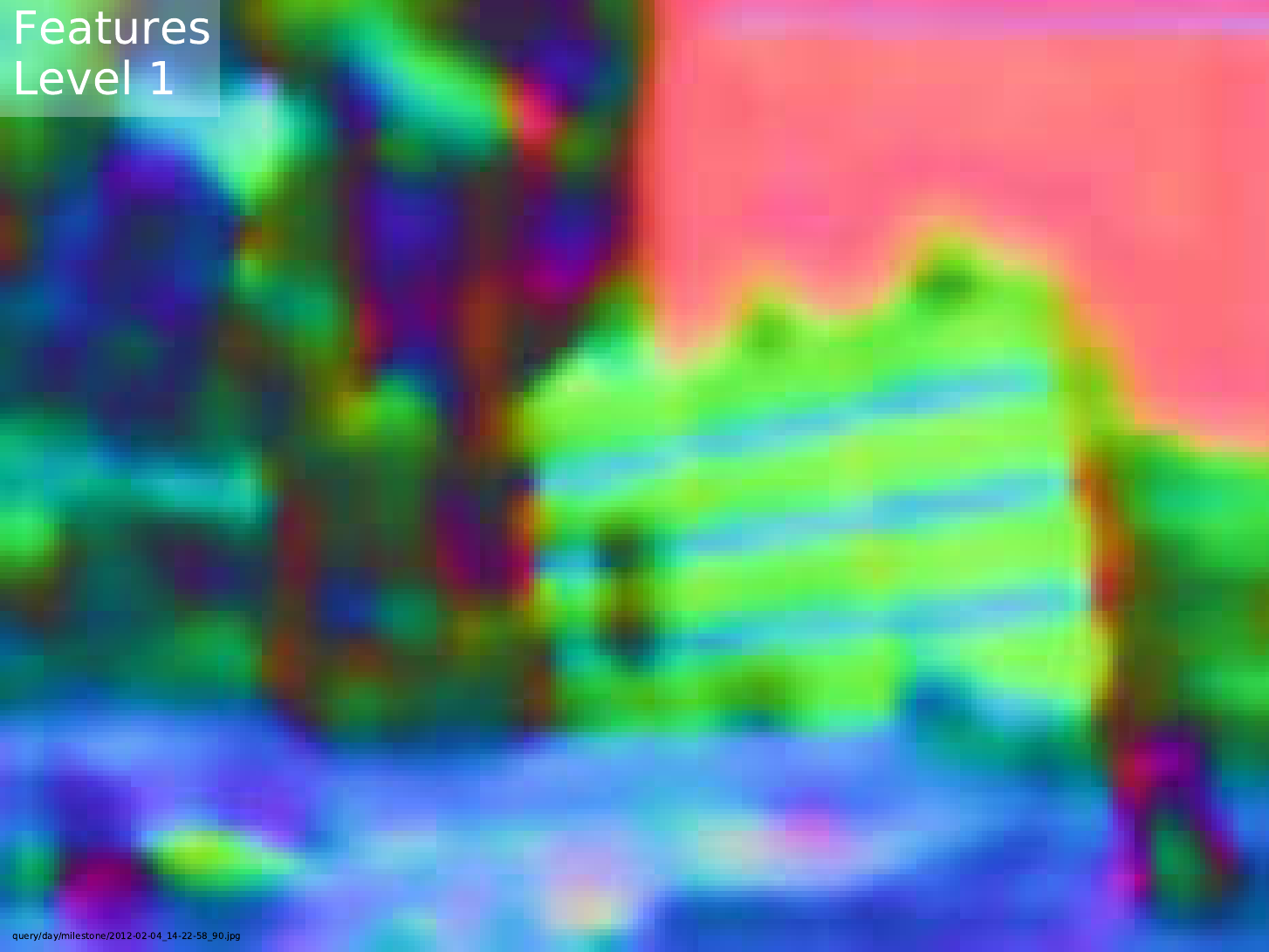}
\end{minipage}%
\begin{minipage}{\iwidth\textwidth}
    \centering
    \includegraphics[width=\pwidth\linewidth]{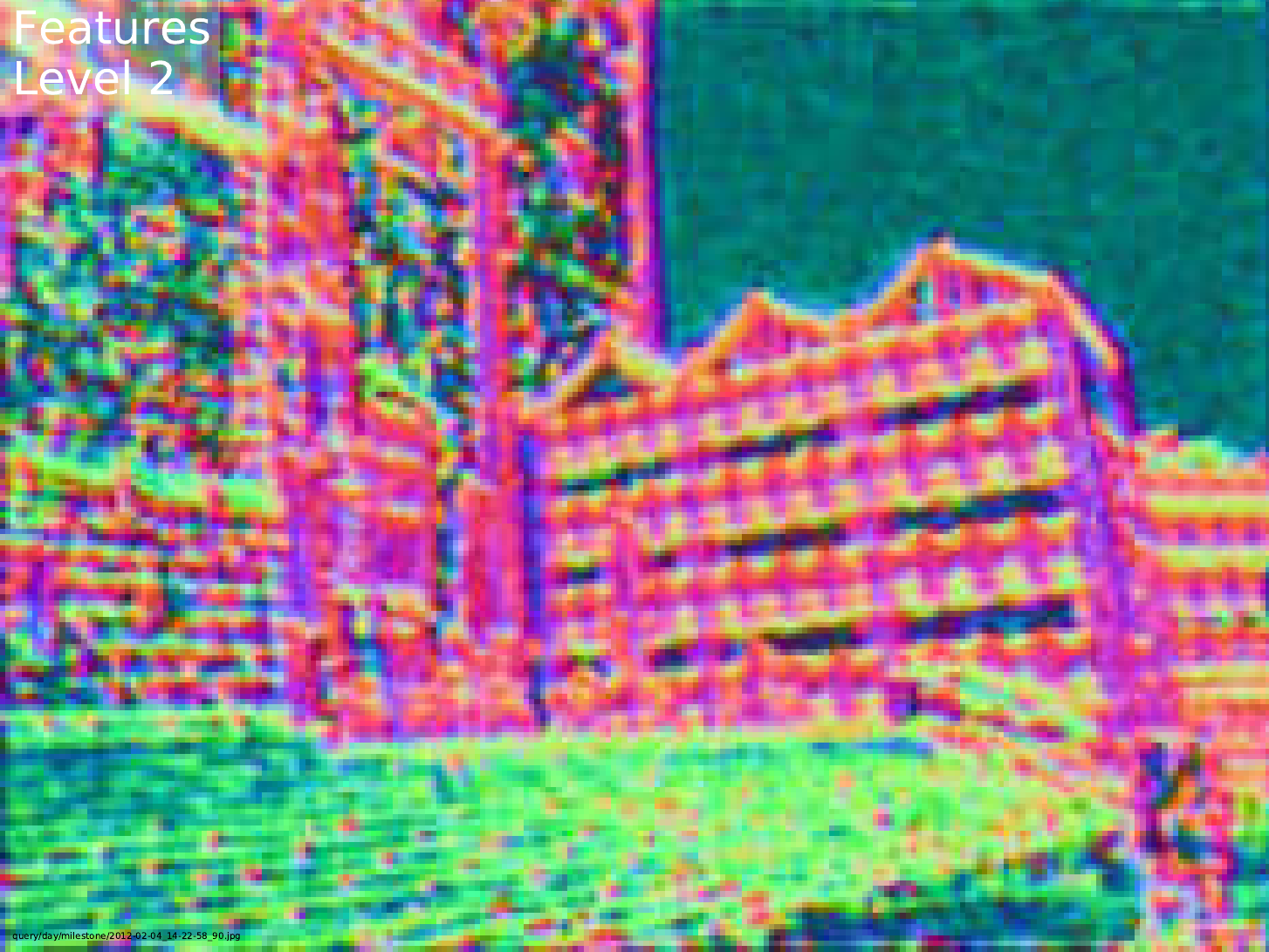}
\end{minipage}%
\begin{minipage}{\iwidth\textwidth}
    \centering
    \includegraphics[width=\pwidth\linewidth]{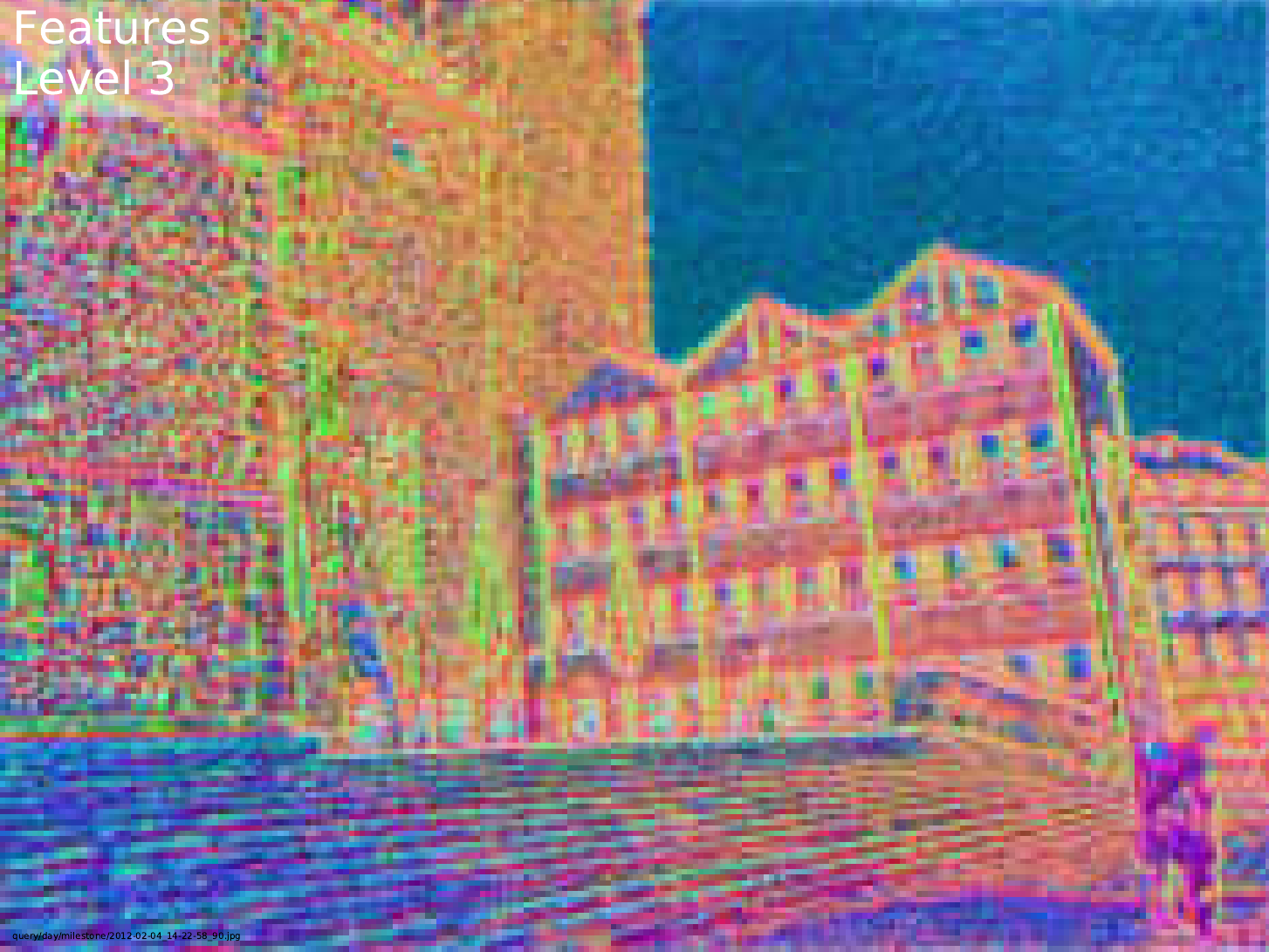}
\end{minipage}%
\begin{minipage}{\iwidth\textwidth}
    \centering
    \includegraphics[width=\pwidth\linewidth]{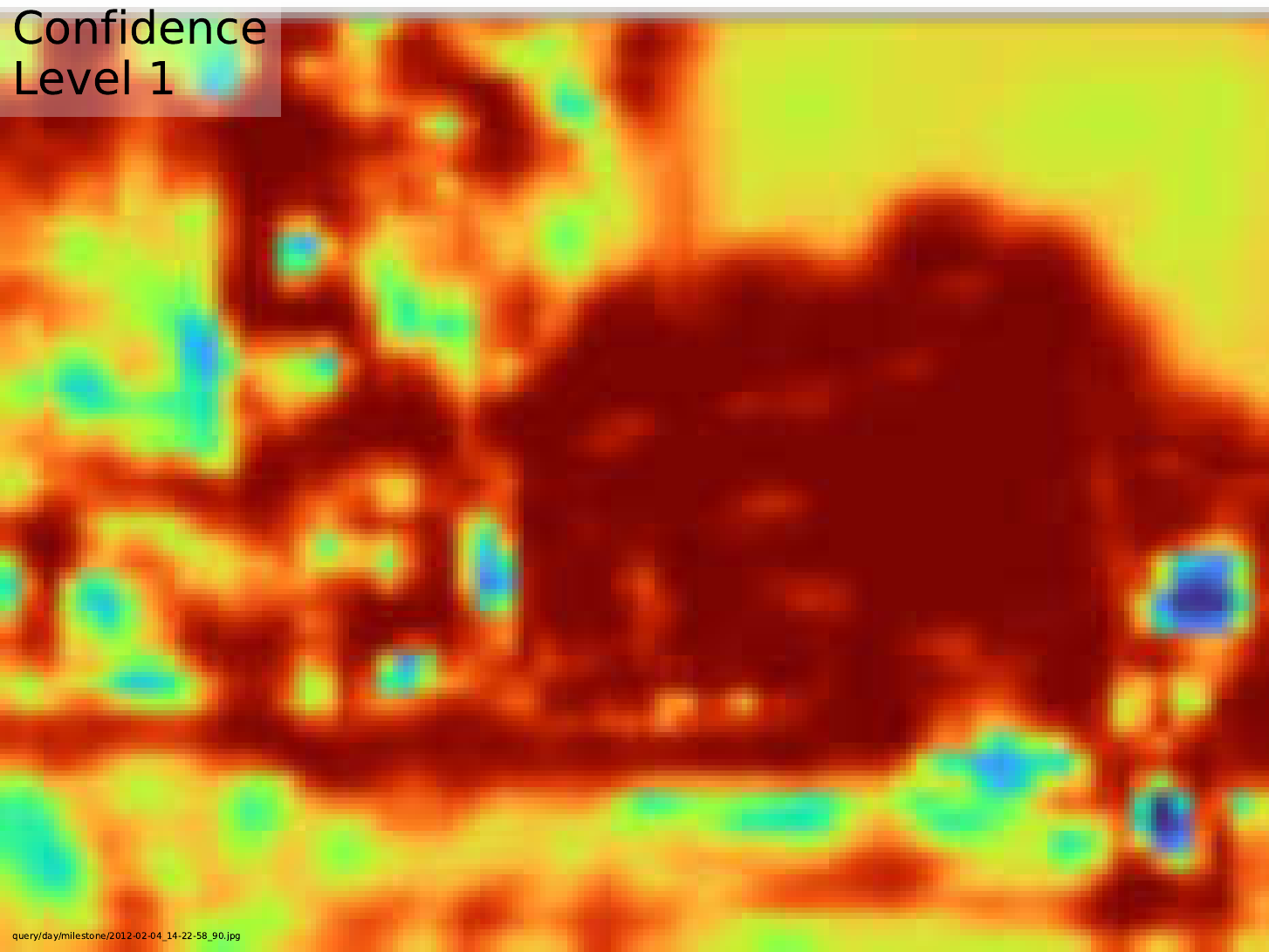}
\end{minipage}%
\begin{minipage}{\iwidth\textwidth}
    \centering
    \includegraphics[width=\pwidth\linewidth]{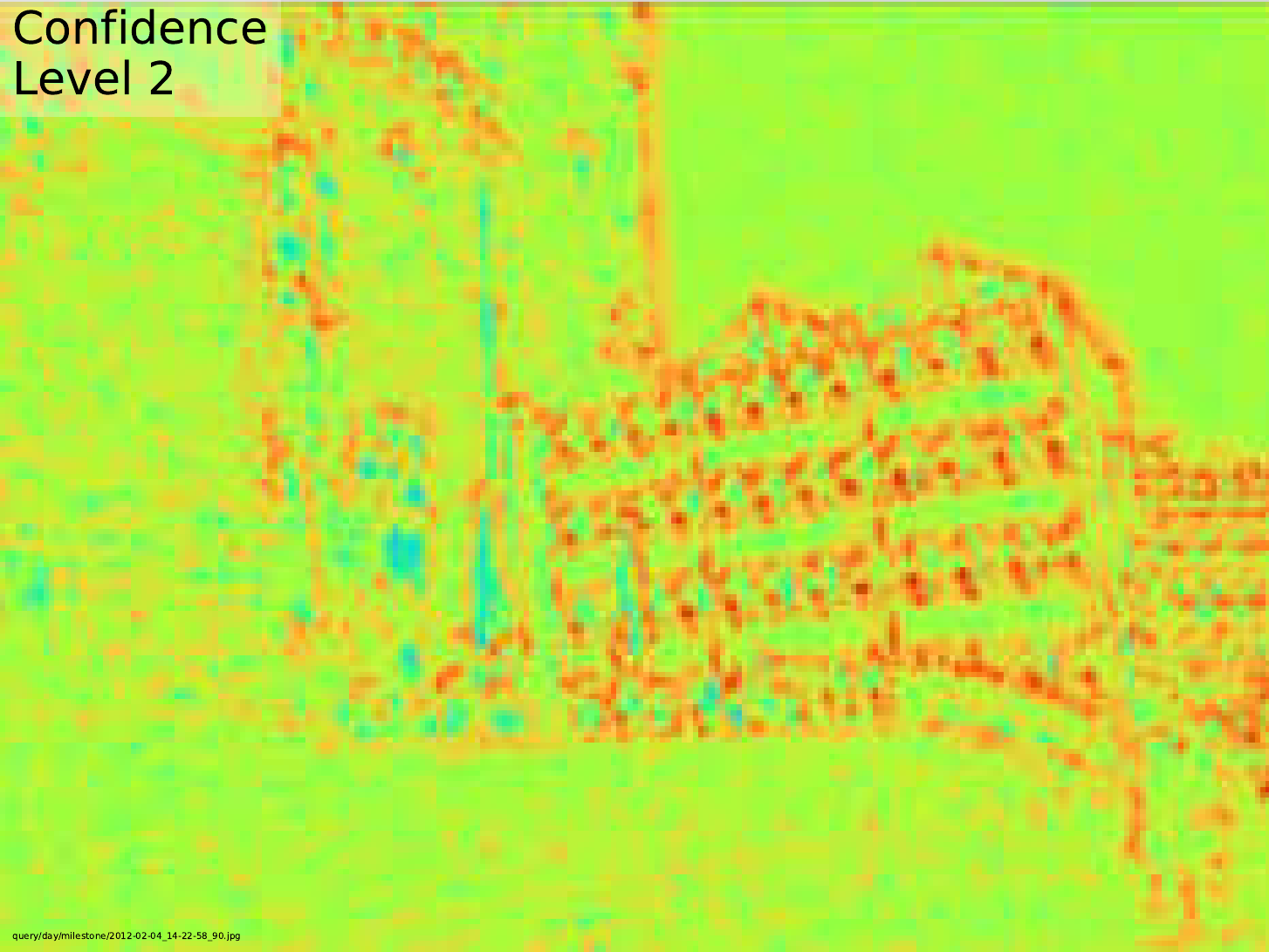}
\end{minipage}%
\begin{minipage}{\iwidth\textwidth}
    \centering
    \includegraphics[width=\pwidth\linewidth]{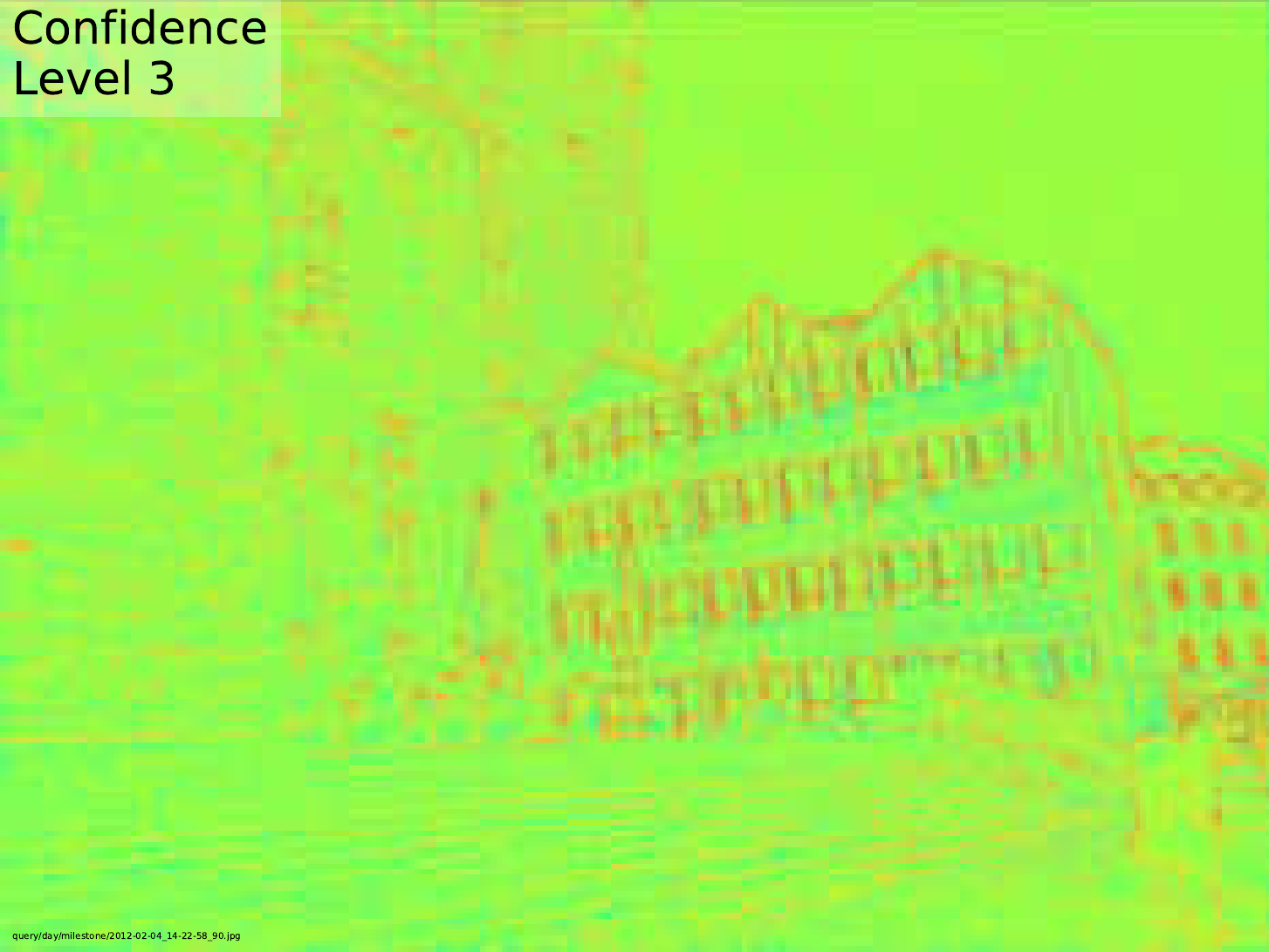}
\end{minipage}
\begin{minipage}{\lwidth\textwidth}
\rotatebox[origin=c]{90}{Reference}
\end{minipage}%
\begin{minipage}{\iwidth\textwidth}
    \centering
    \includegraphics[width=\pwidth\linewidth]{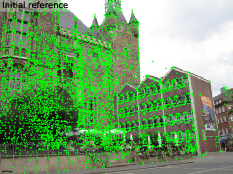}
\end{minipage}%
\begin{minipage}{\iwidth\textwidth}
    \centering
    \includegraphics[width=\pwidth\linewidth]{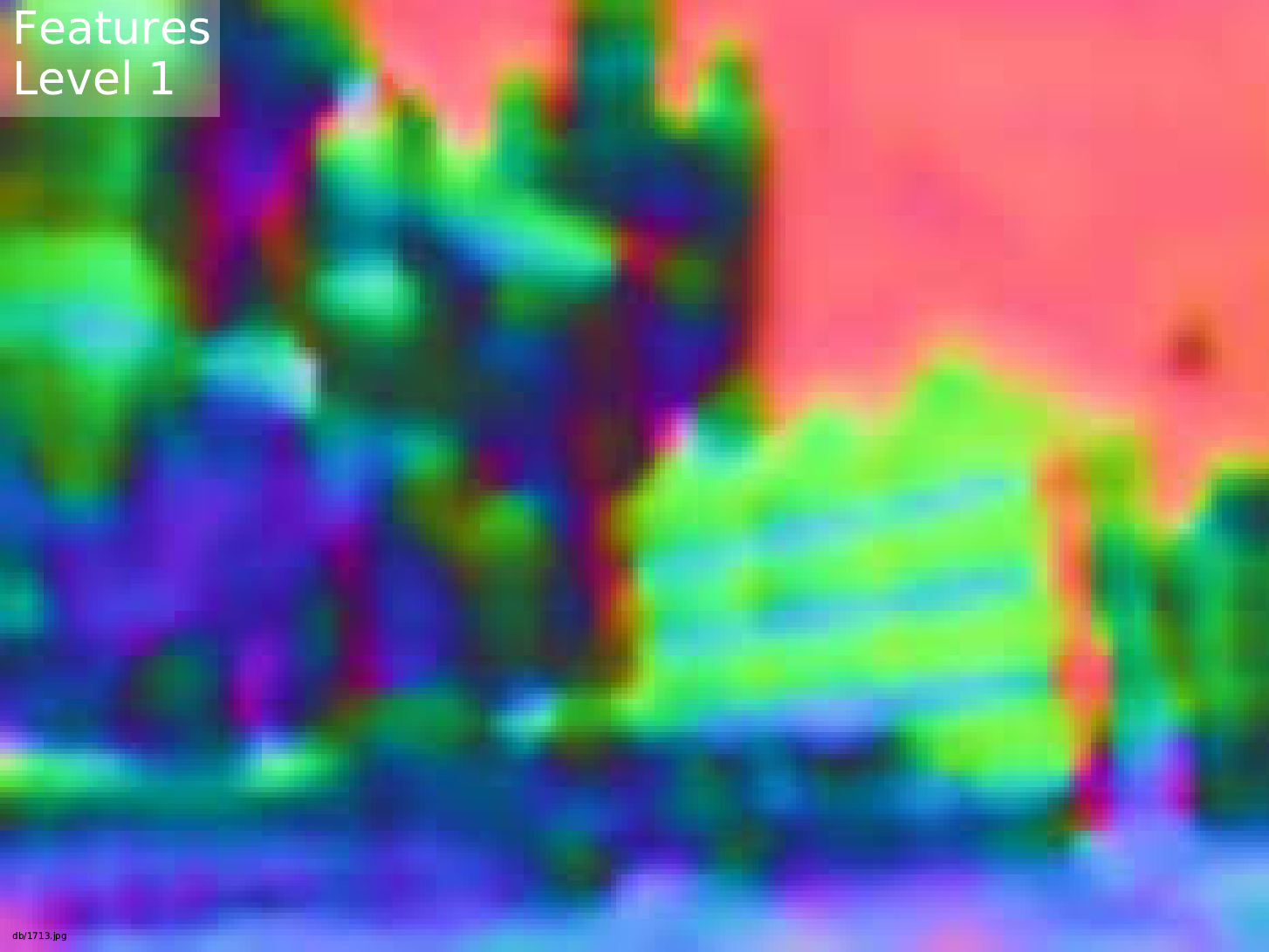}
\end{minipage}%
\begin{minipage}{\iwidth\textwidth}
    \centering
    \includegraphics[width=\pwidth\linewidth]{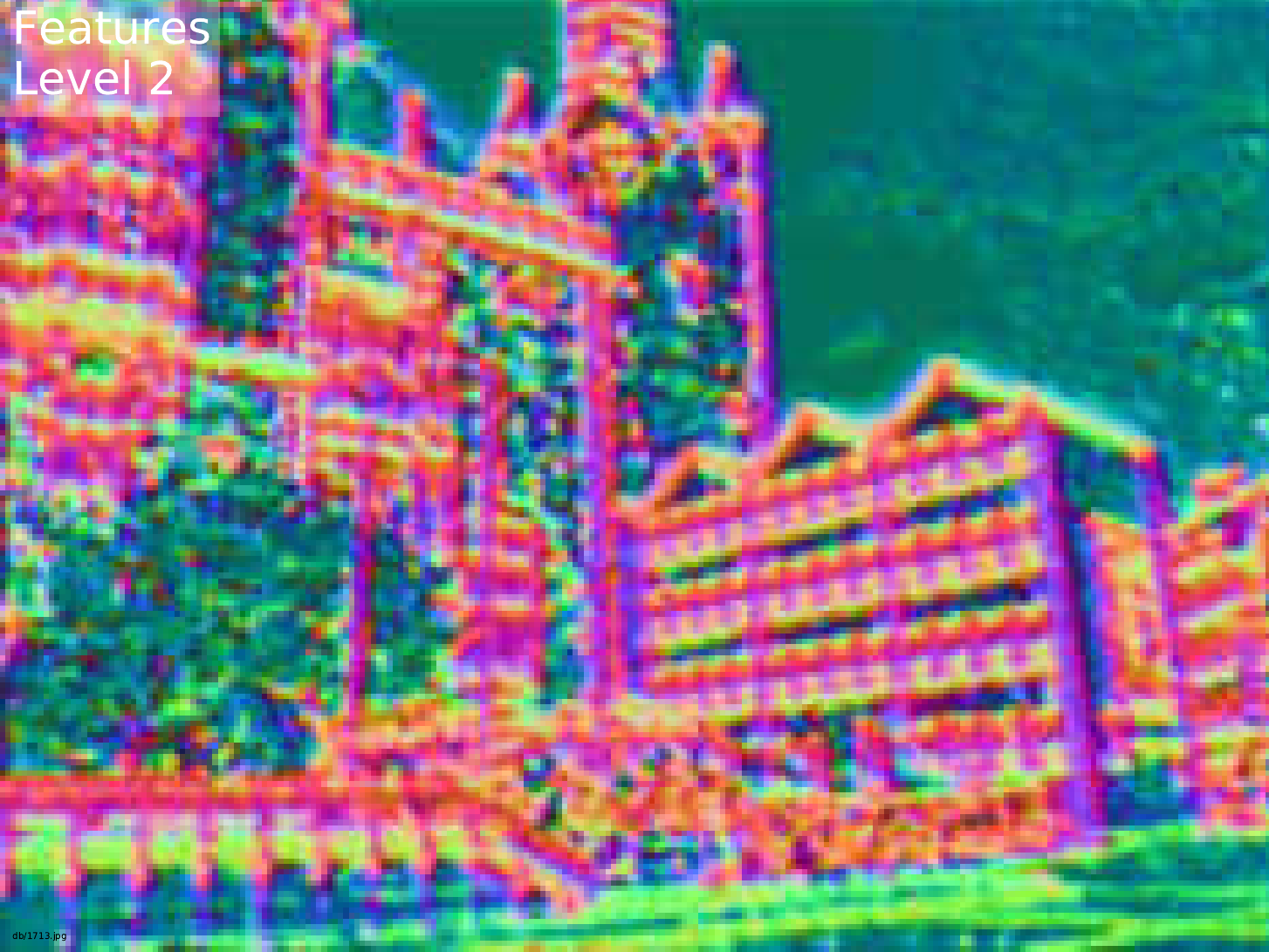}
\end{minipage}%
\begin{minipage}{\iwidth\textwidth}
    \centering
    \includegraphics[width=\pwidth\linewidth]{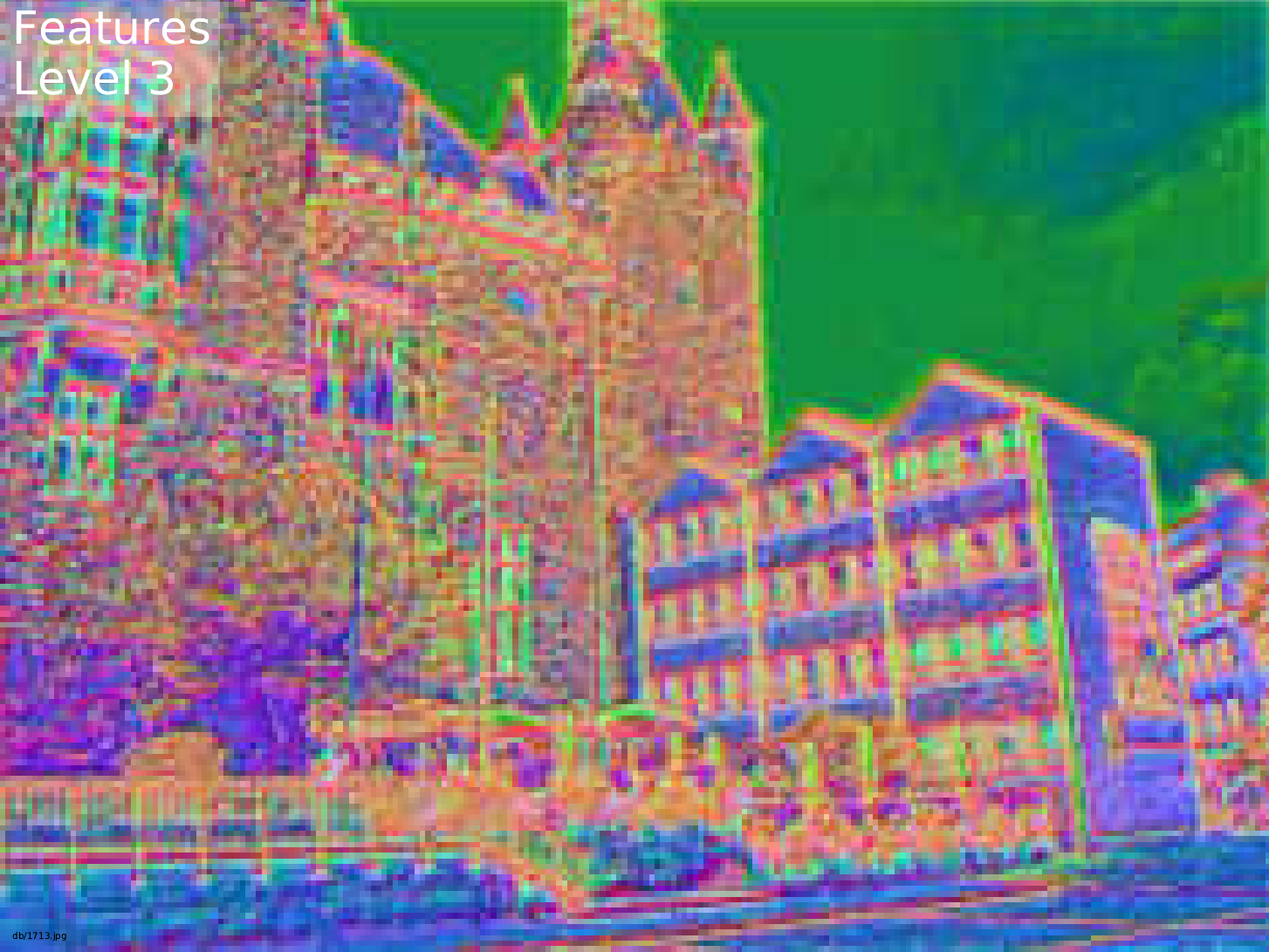}
\end{minipage}%
\begin{minipage}{\iwidth\textwidth}
    \centering
    \includegraphics[width=\pwidth\linewidth]{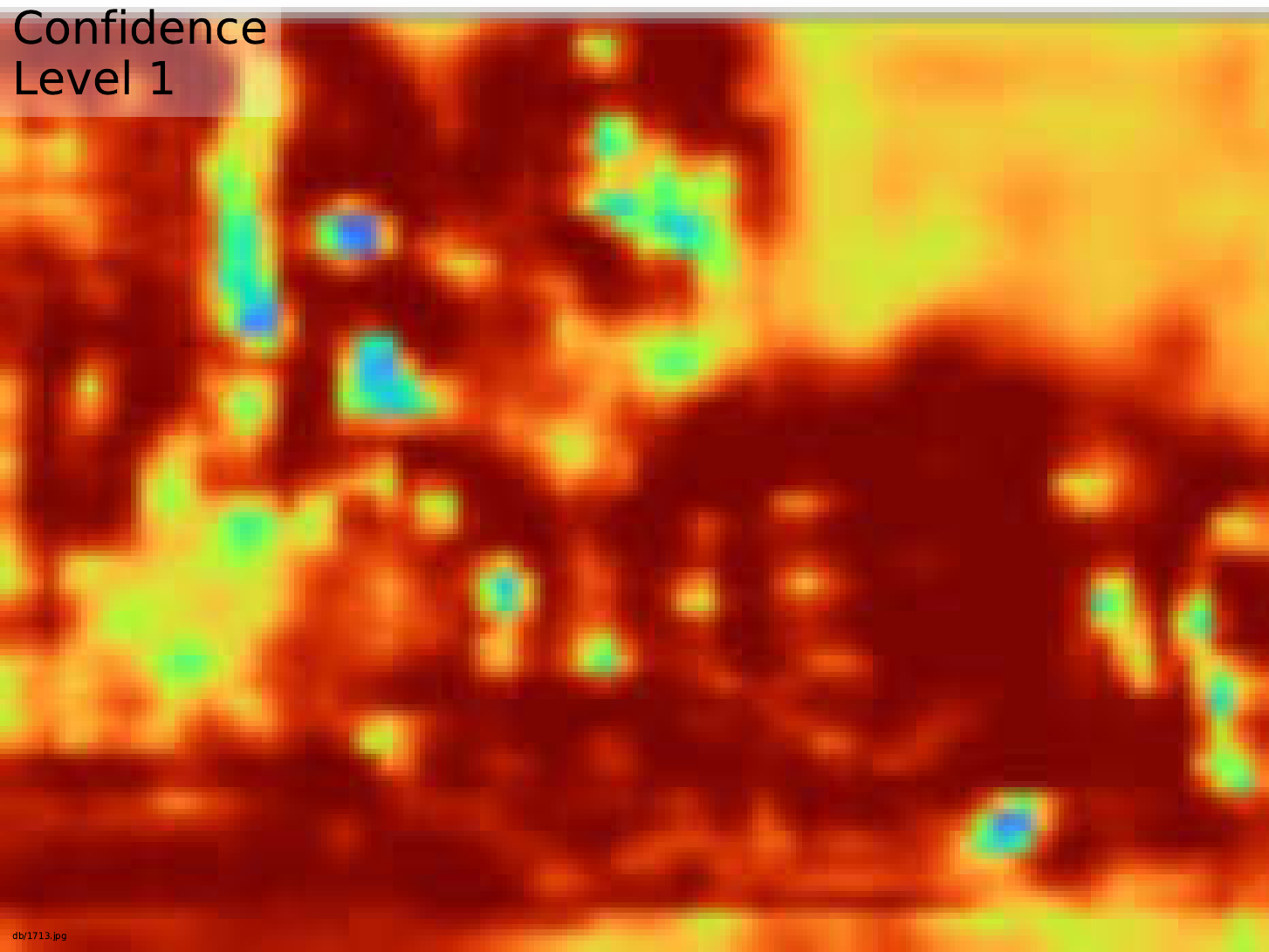}
\end{minipage}%
\begin{minipage}{\iwidth\textwidth}
    \centering
    \includegraphics[width=\pwidth\linewidth]{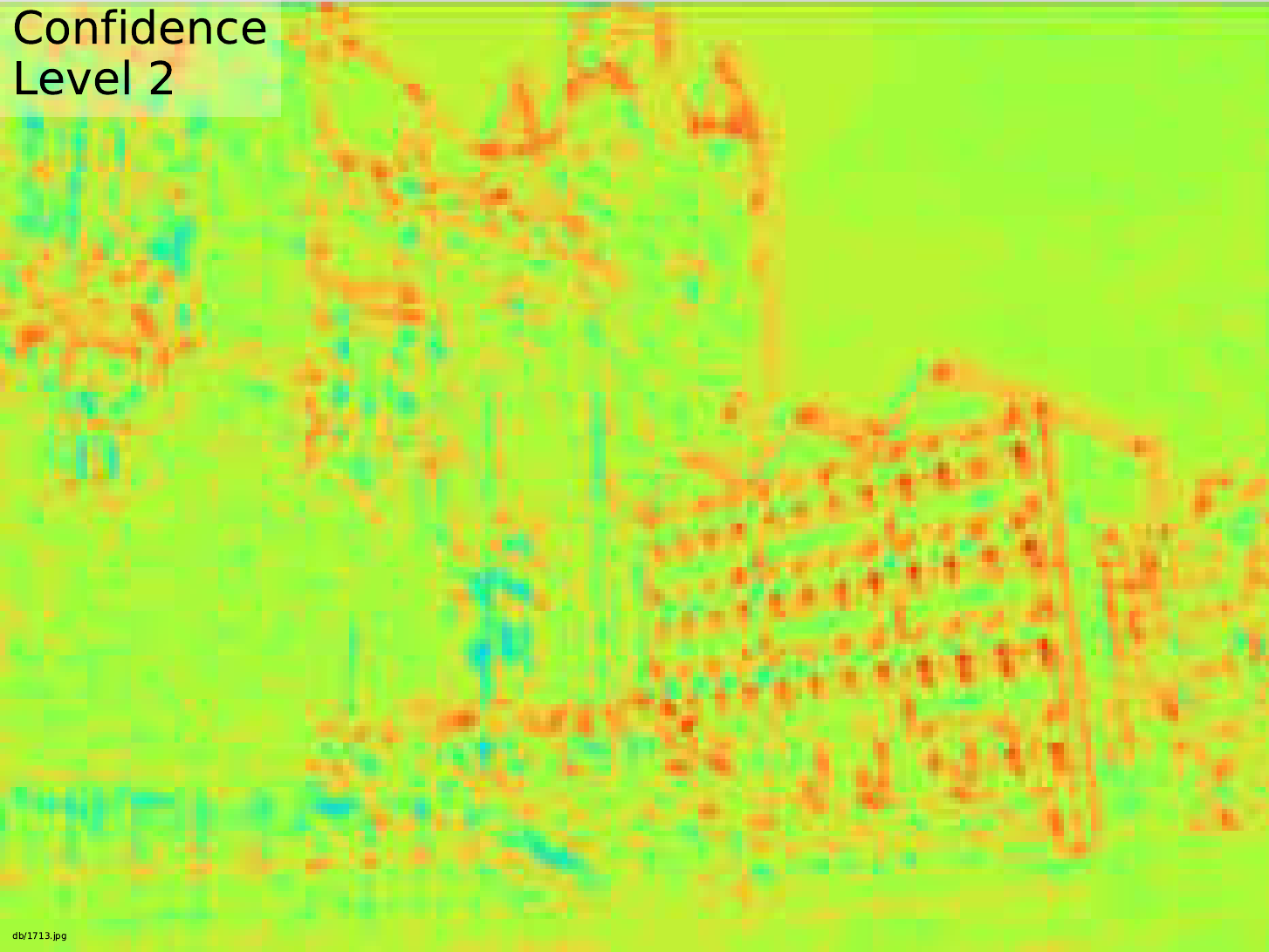}
\end{minipage}%
\begin{minipage}{\iwidth\textwidth}
    \centering
    \includegraphics[width=\pwidth\linewidth]{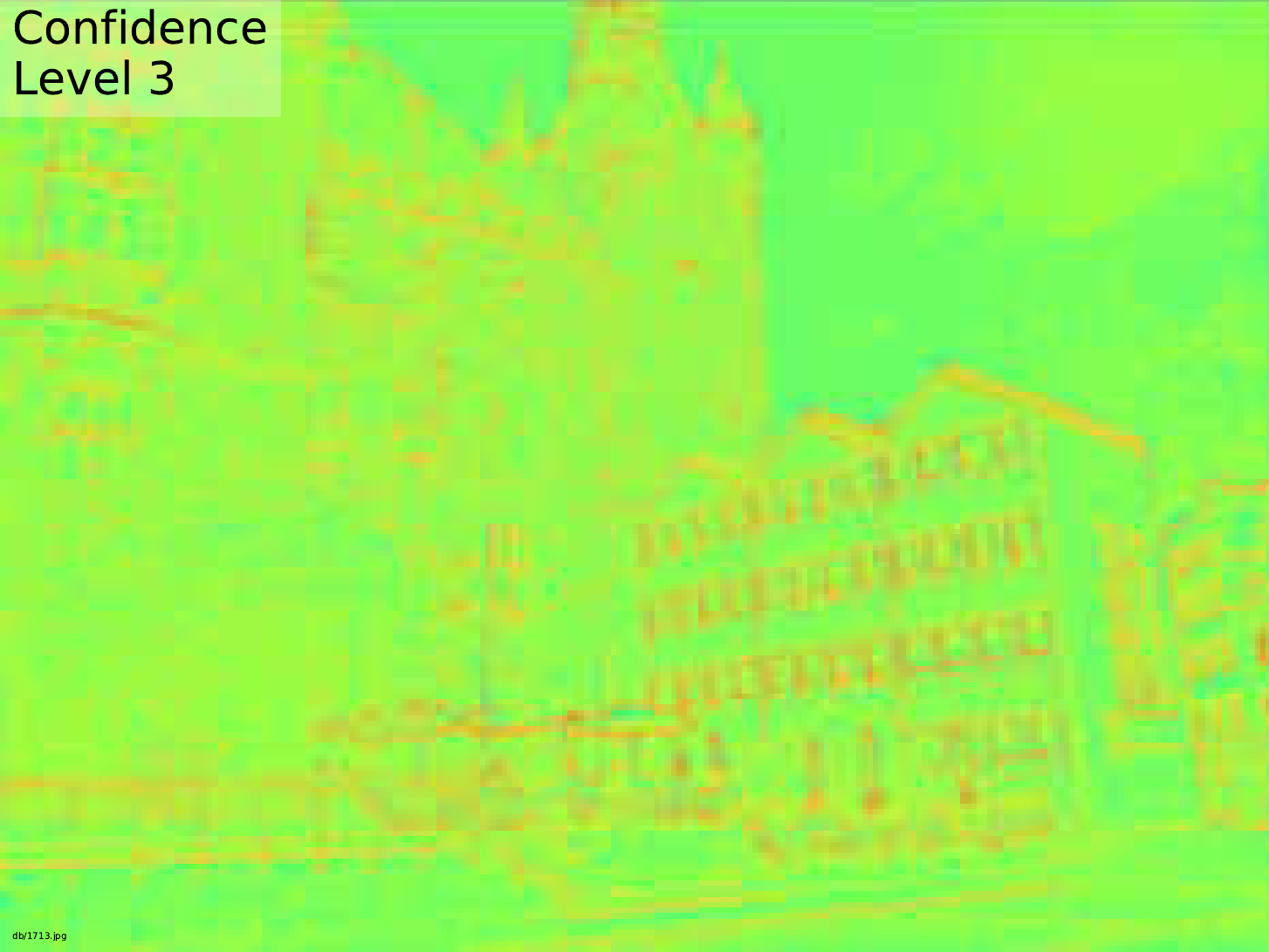}
\end{minipage}
\vspace{2mm}

\begin{minipage}{\lwidth\textwidth}
\rotatebox[origin=c]{90}{Query}
\end{minipage}%
\begin{minipage}{\iwidth\textwidth}
    \centering
    \includegraphics[width=\pwidth\linewidth]{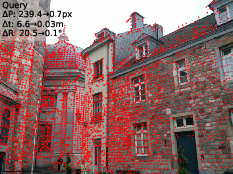}
\end{minipage}%
\begin{minipage}{\iwidth\textwidth}
    \centering
    \includegraphics[width=\pwidth\linewidth]{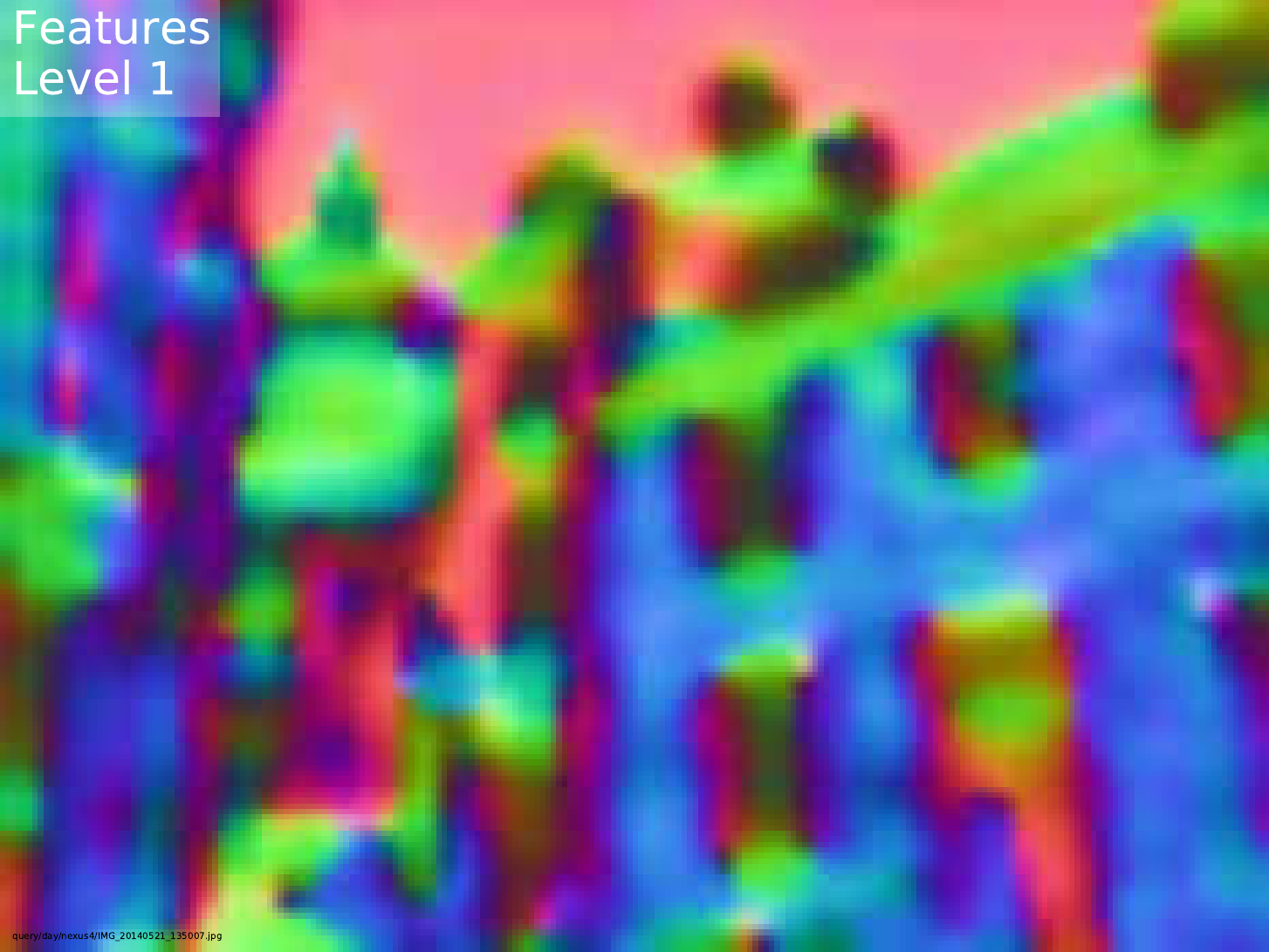}
\end{minipage}%
\begin{minipage}{\iwidth\textwidth}
    \centering
    \includegraphics[width=\pwidth\linewidth]{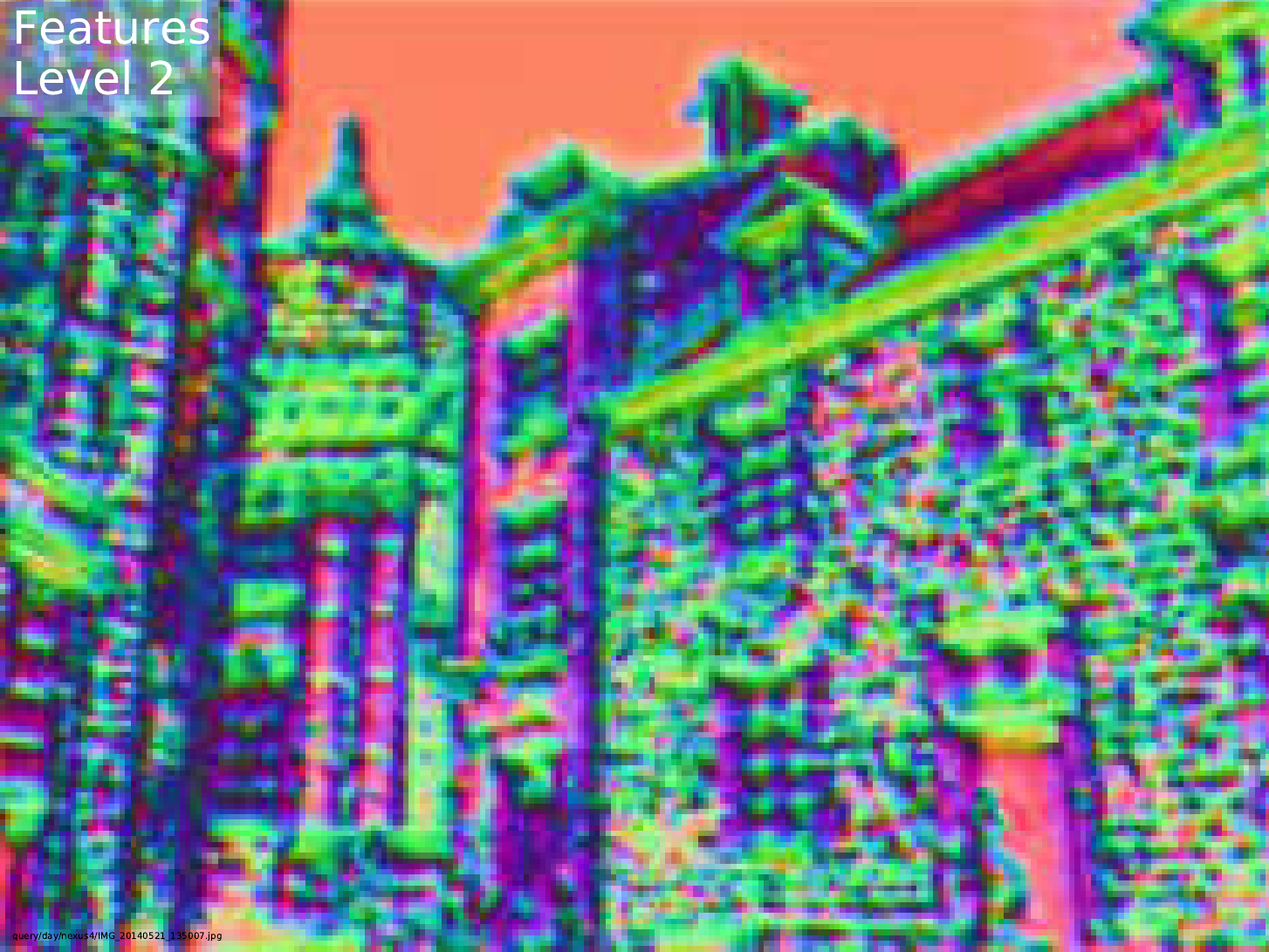}
\end{minipage}%
\begin{minipage}{\iwidth\textwidth}
    \centering
    \includegraphics[width=\pwidth\linewidth]{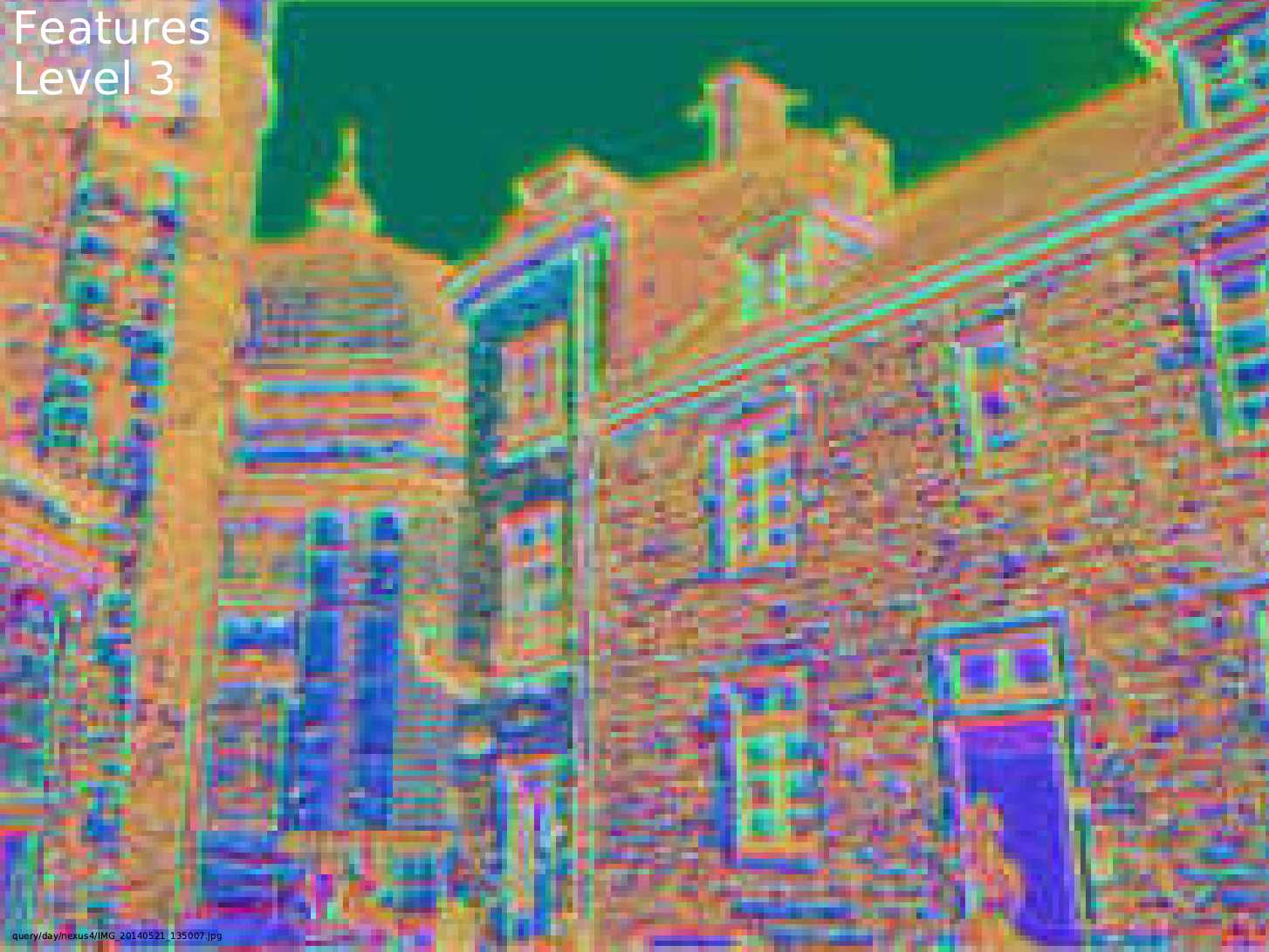}
\end{minipage}%
\begin{minipage}{\iwidth\textwidth}
    \centering
    \includegraphics[width=\pwidth\linewidth]{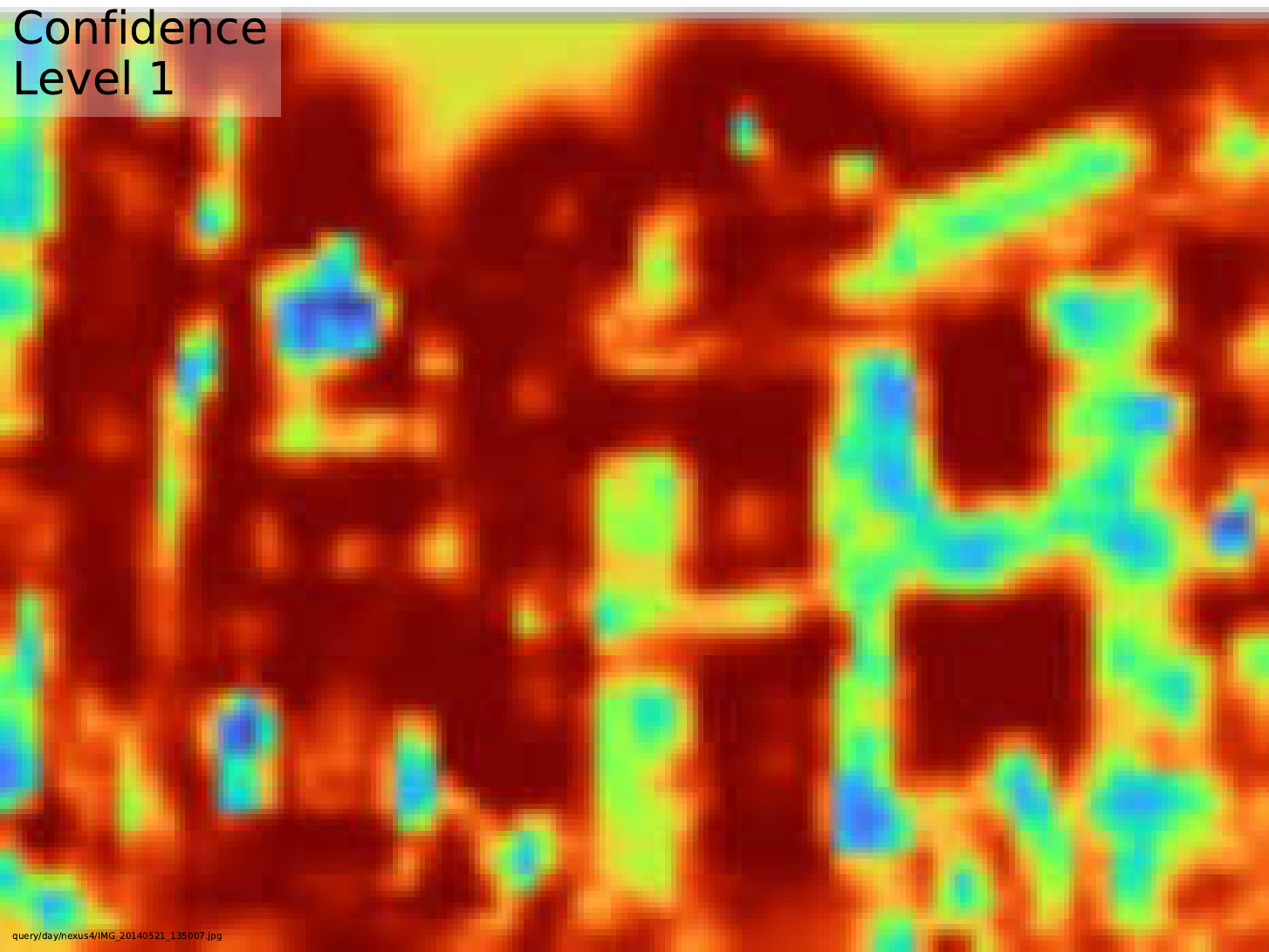}
\end{minipage}%
\begin{minipage}{\iwidth\textwidth}
    \centering
    \includegraphics[width=\pwidth\linewidth]{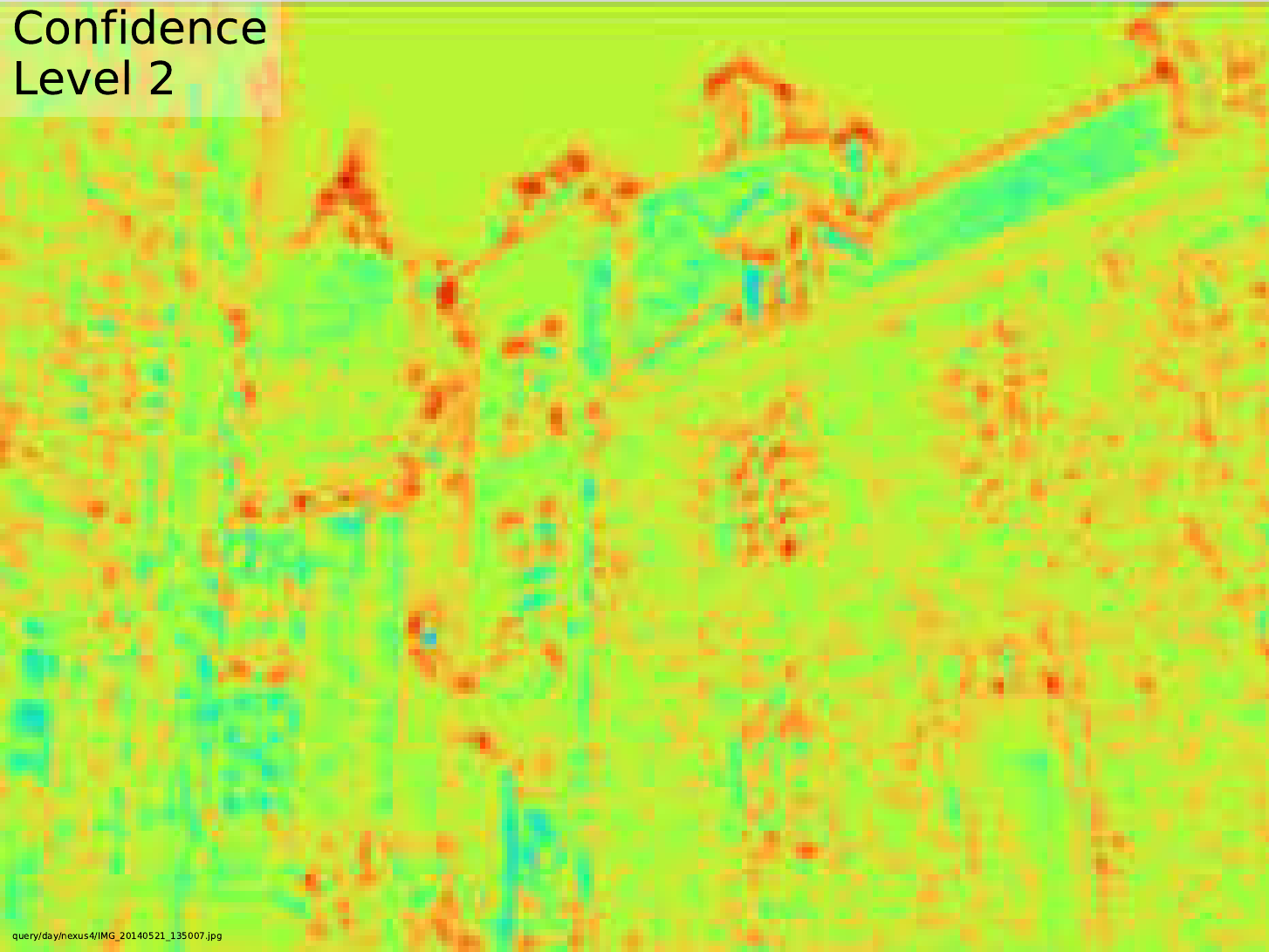}
\end{minipage}%
\begin{minipage}{\iwidth\textwidth}
    \centering
    \includegraphics[width=\pwidth\linewidth]{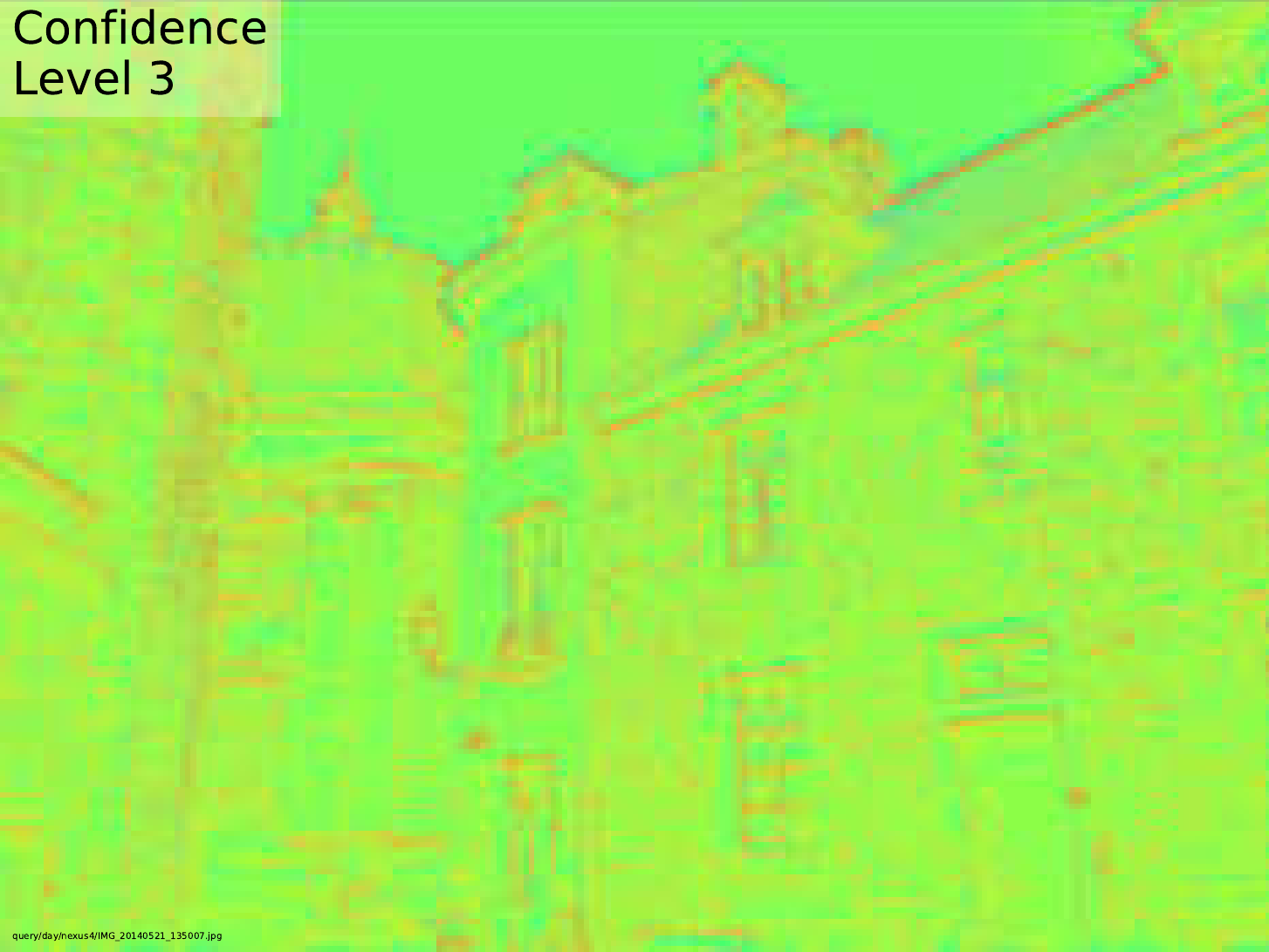}
\end{minipage}
\begin{minipage}{\lwidth\textwidth}
\rotatebox[origin=c]{90}{Reference}
\end{minipage}%
\begin{minipage}{\iwidth\textwidth}
    \centering
    \includegraphics[width=\pwidth\linewidth]{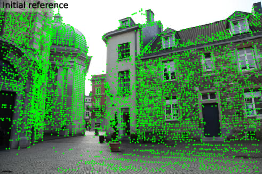}
\end{minipage}%
\begin{minipage}{\iwidth\textwidth}
    \centering
    \includegraphics[width=\pwidth\linewidth]{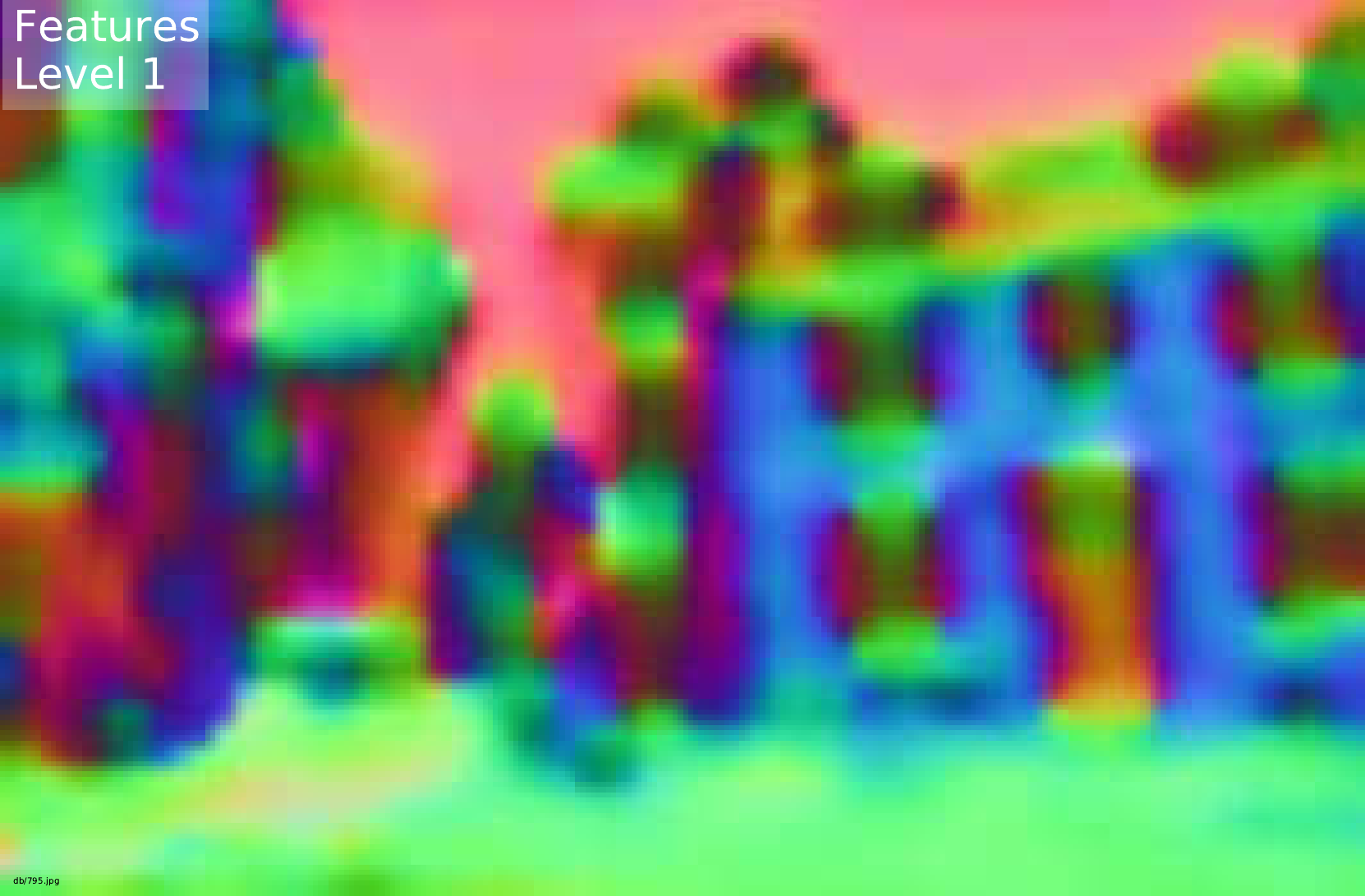}
\end{minipage}%
\begin{minipage}{\iwidth\textwidth}
    \centering
    \includegraphics[width=\pwidth\linewidth]{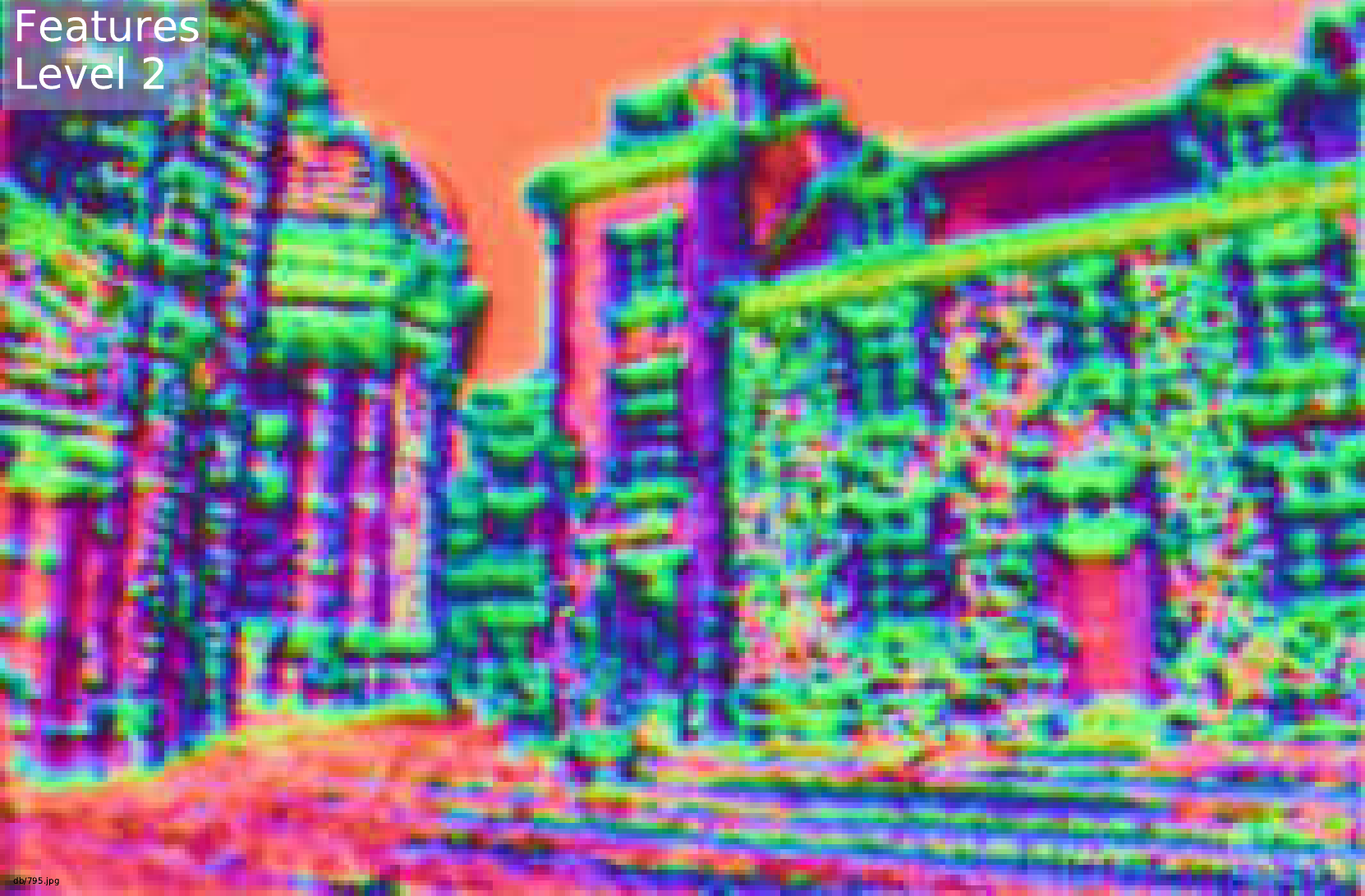}
\end{minipage}%
\begin{minipage}{\iwidth\textwidth}
    \centering
    \includegraphics[width=\pwidth\linewidth]{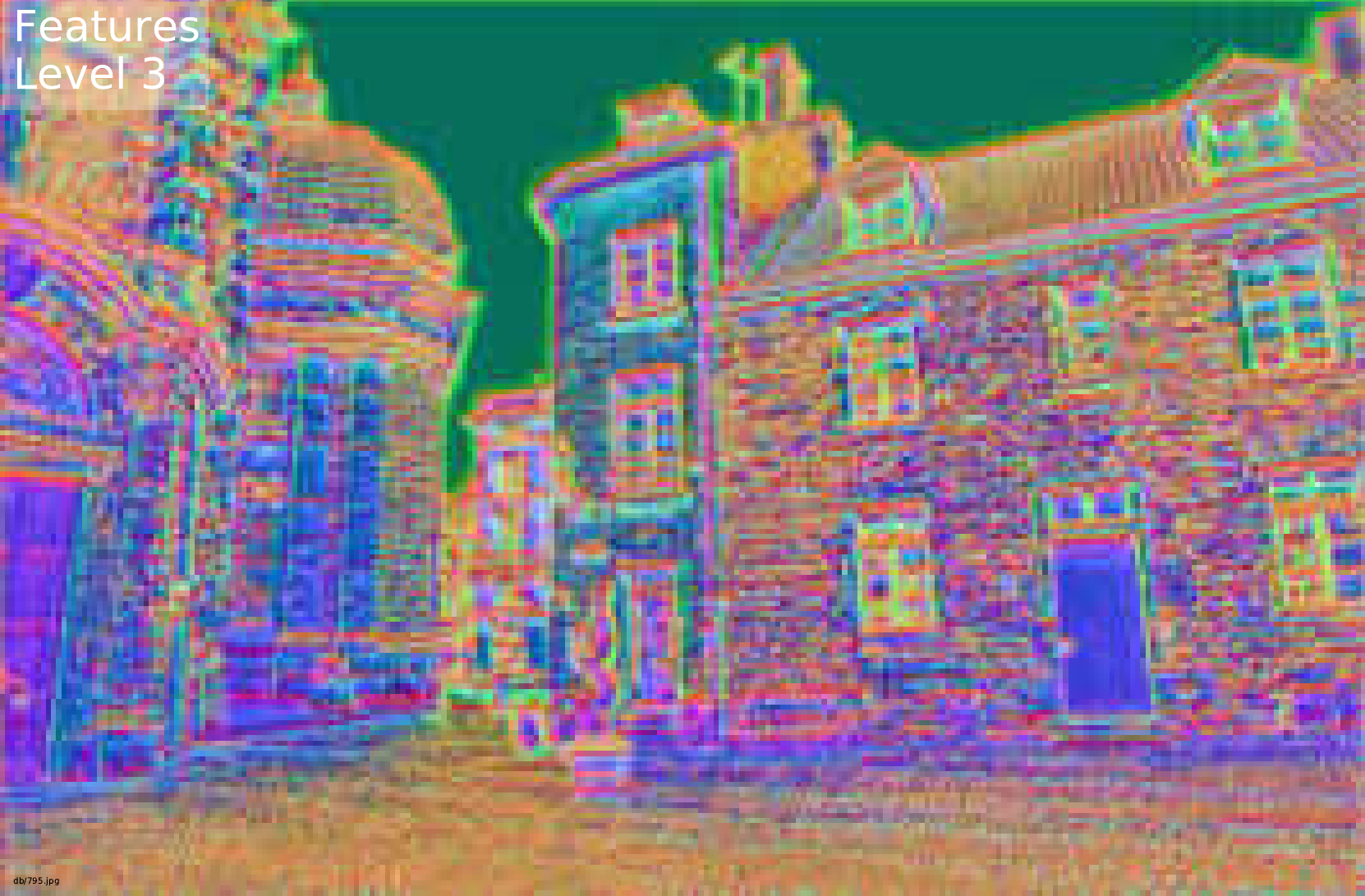}
\end{minipage}%
\begin{minipage}{\iwidth\textwidth}
    \centering
    \includegraphics[width=\pwidth\linewidth]{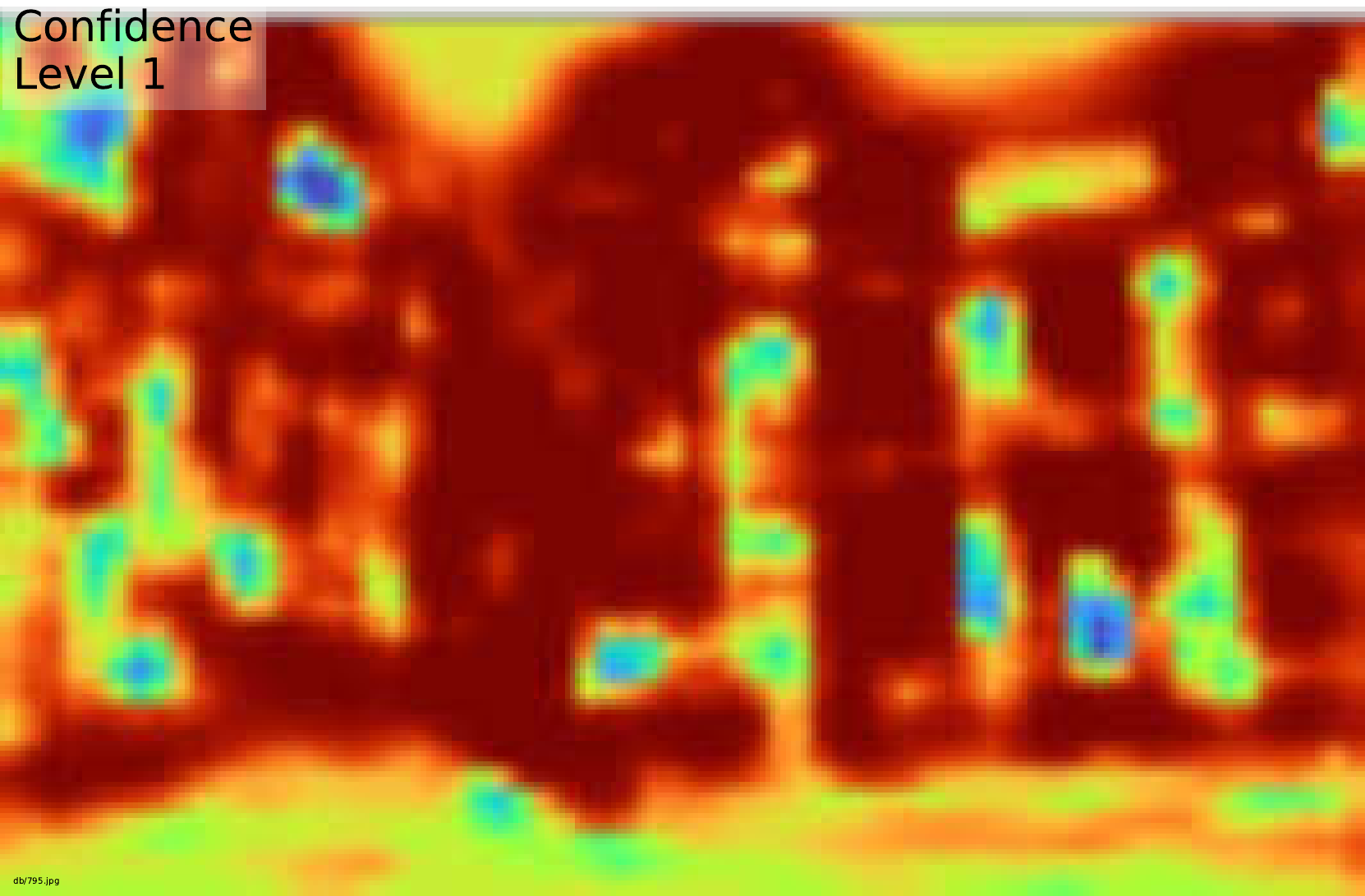}
\end{minipage}%
\begin{minipage}{\iwidth\textwidth}
    \centering
    \includegraphics[width=\pwidth\linewidth]{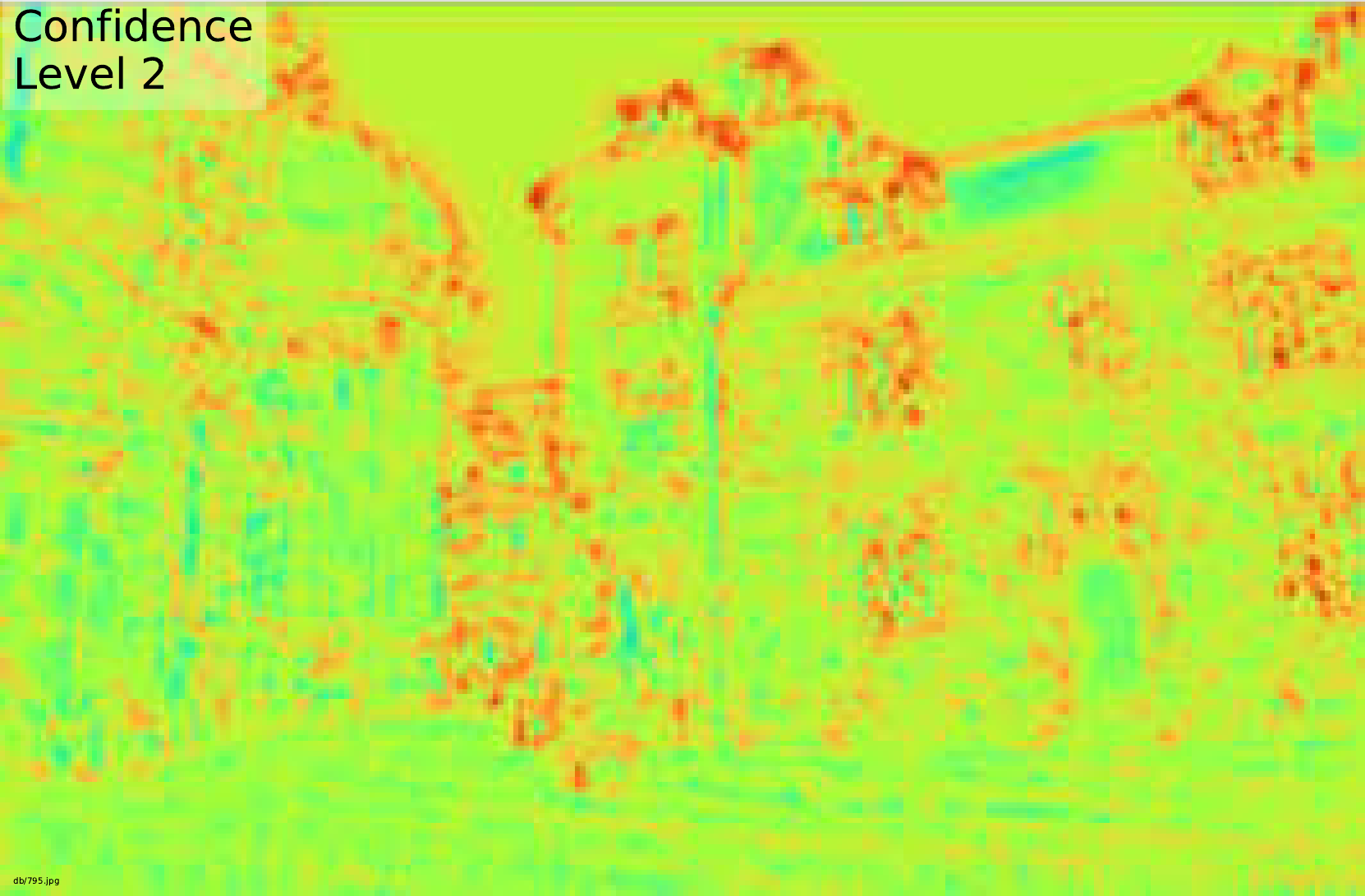}
\end{minipage}%
\begin{minipage}{\iwidth\textwidth}
    \centering
    \includegraphics[width=\pwidth\linewidth]{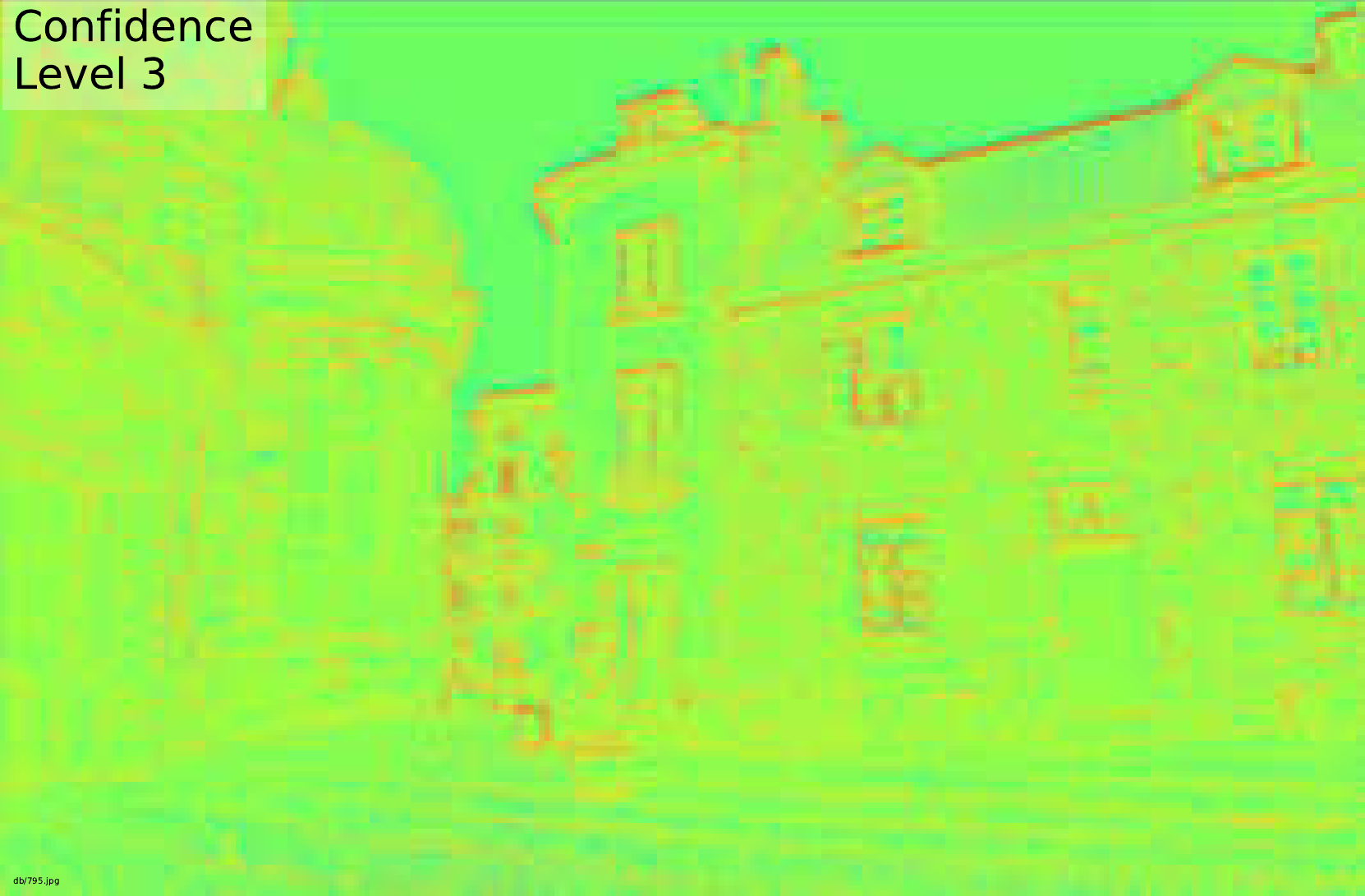}
\end{minipage}
\vspace{2mm}

\begin{minipage}{\lwidth\textwidth}
\rotatebox[origin=c]{90}{Query}
\end{minipage}%
\begin{minipage}{\iwidth\textwidth}
    \centering
    \includegraphics[width=\pwidth\linewidth]{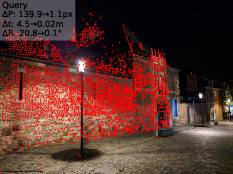}
\end{minipage}%
\begin{minipage}{\iwidth\textwidth}
    \centering
    \includegraphics[width=\pwidth\linewidth]{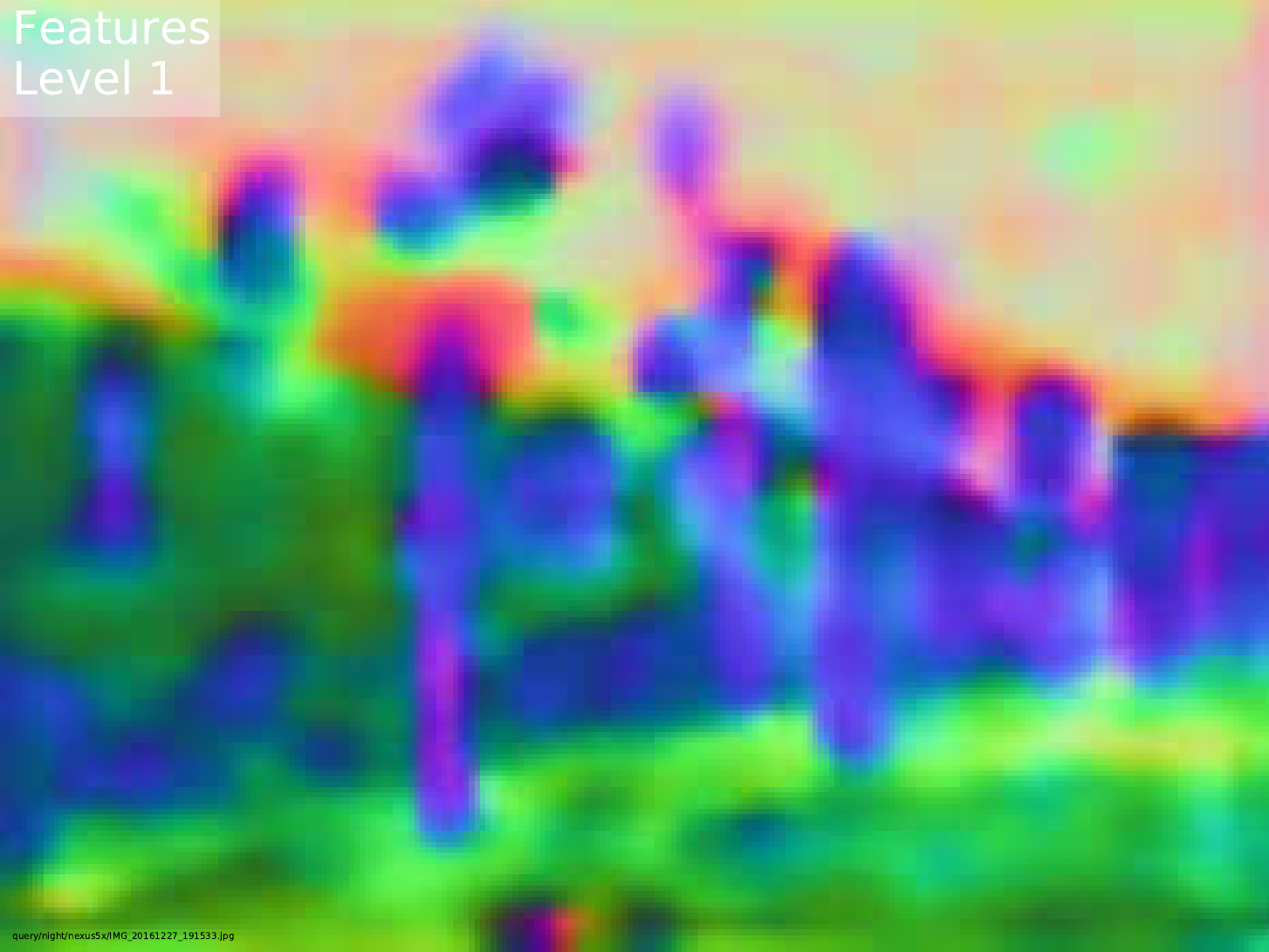}
\end{minipage}%
\begin{minipage}{\iwidth\textwidth}
    \centering
    \includegraphics[width=\pwidth\linewidth]{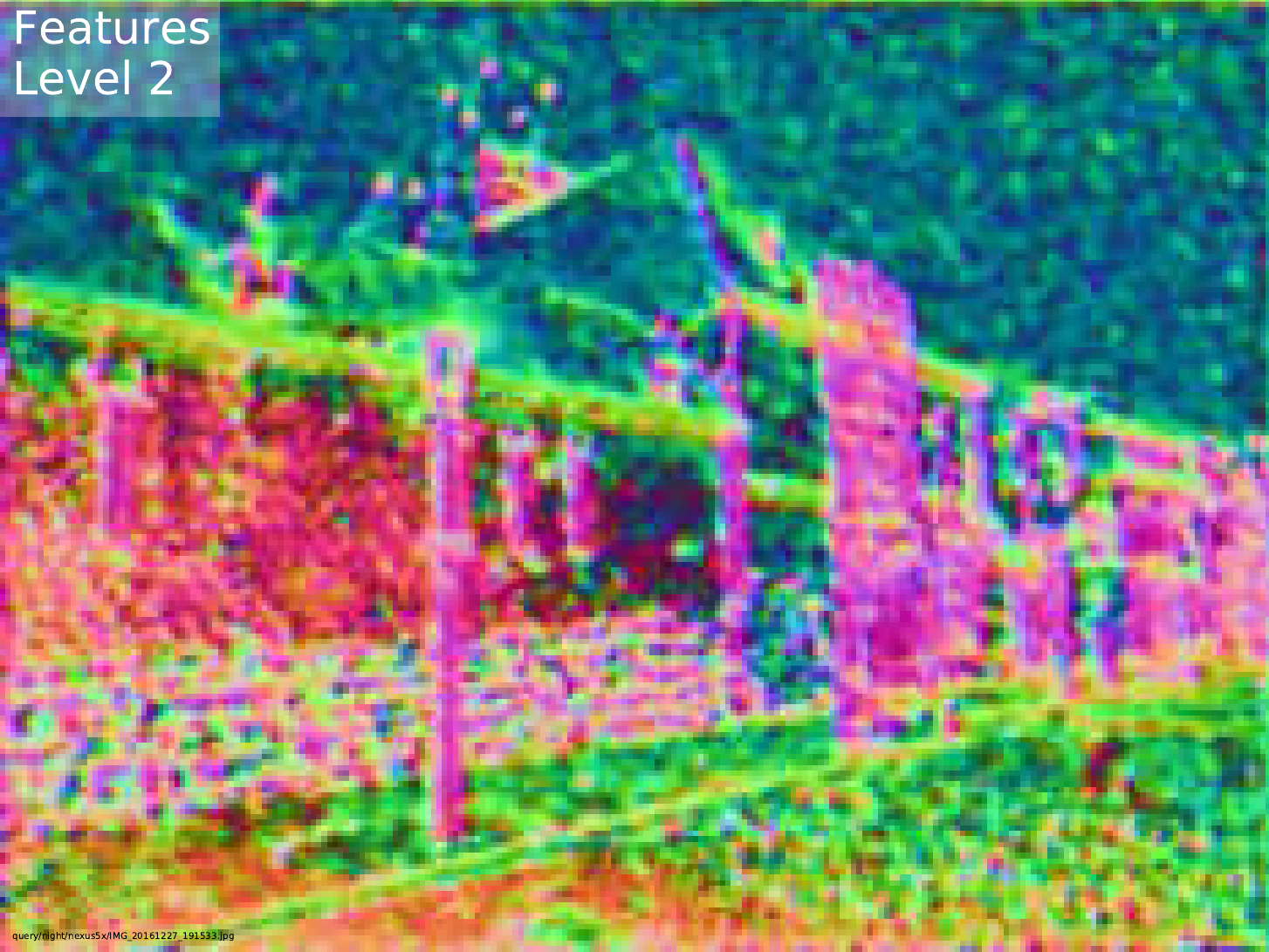}
\end{minipage}%
\begin{minipage}{\iwidth\textwidth}
    \centering
    \includegraphics[width=\pwidth\linewidth]{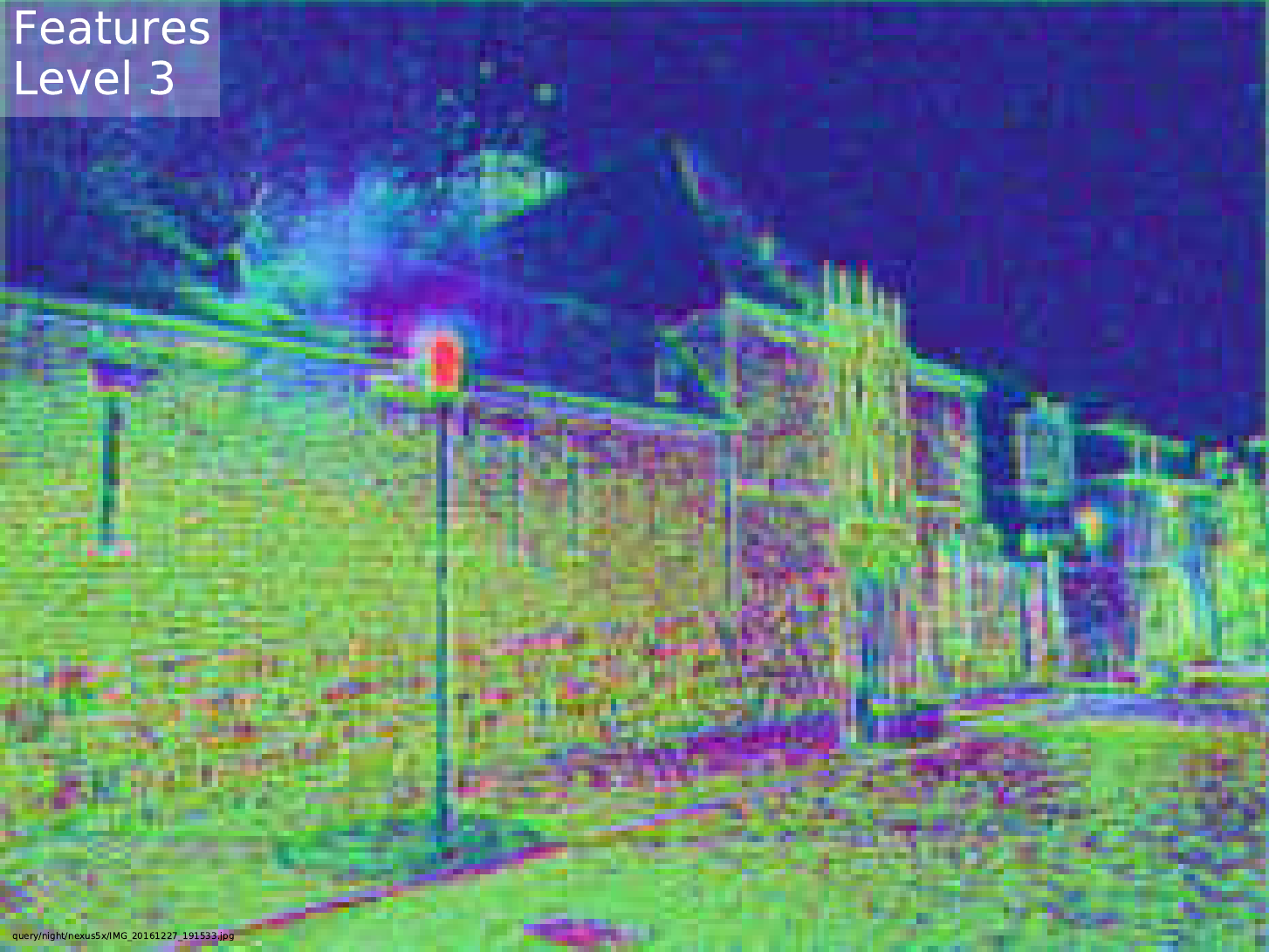}
\end{minipage}%
\begin{minipage}{\iwidth\textwidth}
    \centering
    \includegraphics[width=\pwidth\linewidth]{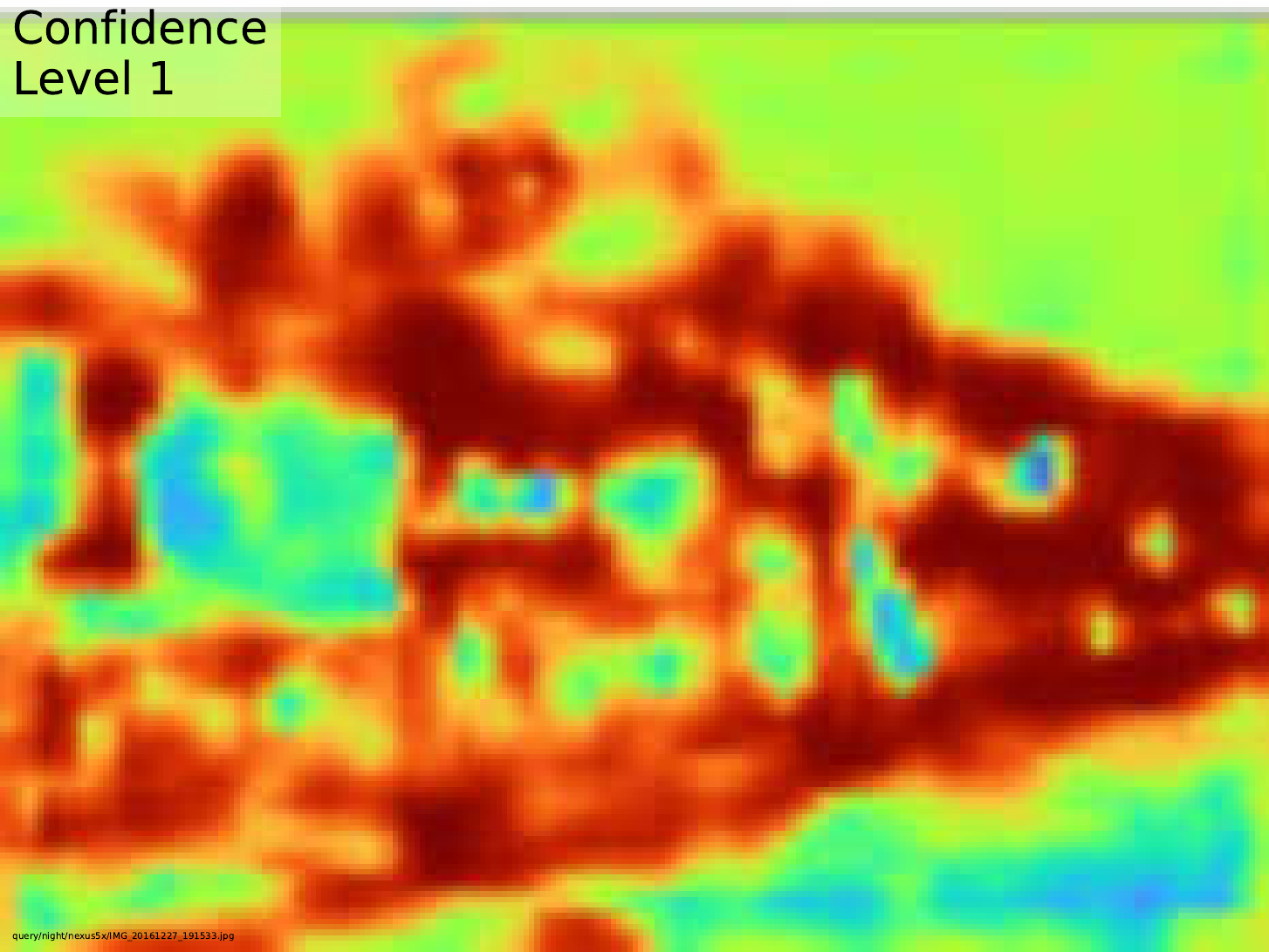}
\end{minipage}%
\begin{minipage}{\iwidth\textwidth}
    \centering
    \includegraphics[width=\pwidth\linewidth]{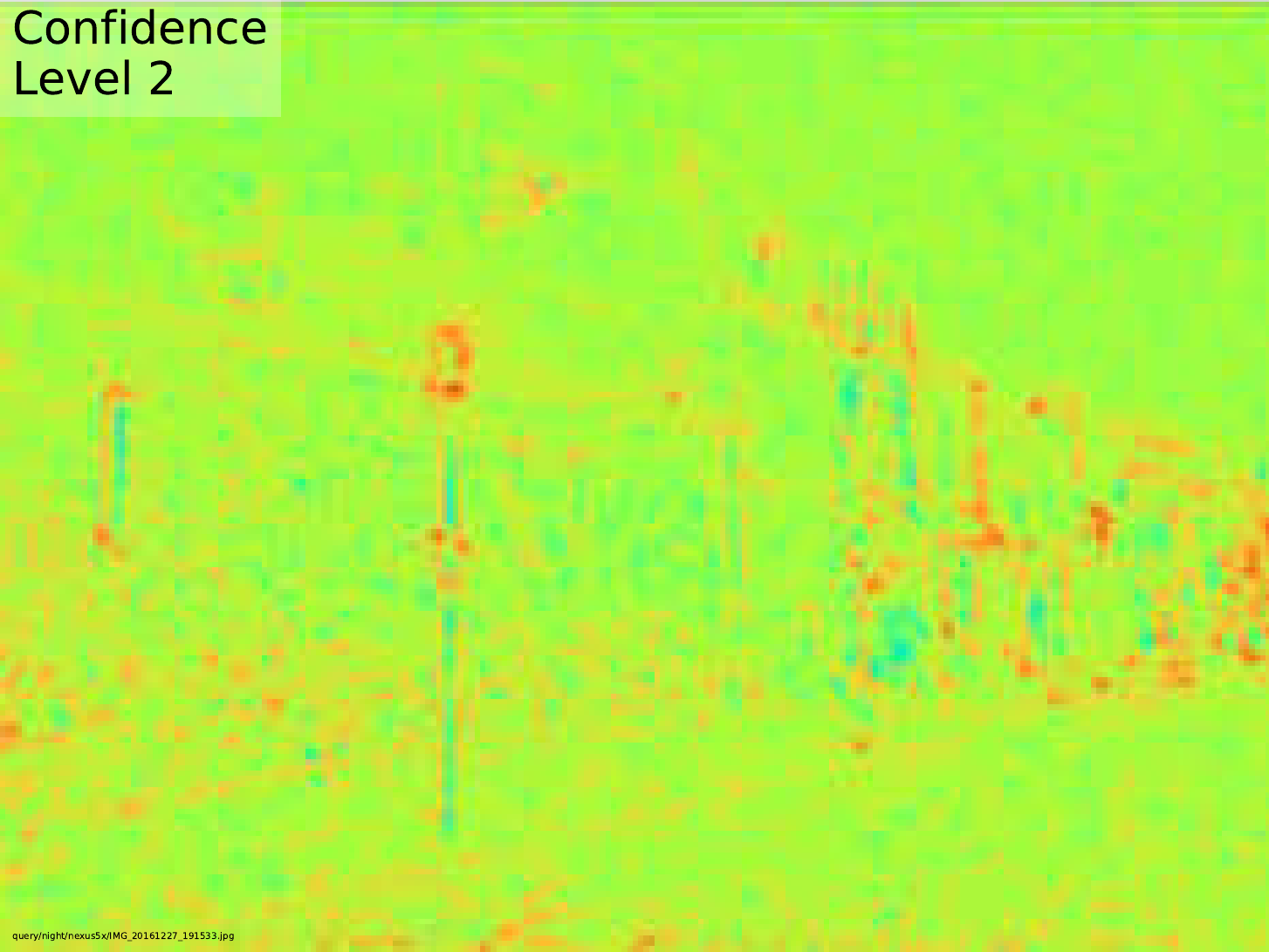}
\end{minipage}%
\begin{minipage}{\iwidth\textwidth}
    \centering
    \includegraphics[width=\pwidth\linewidth]{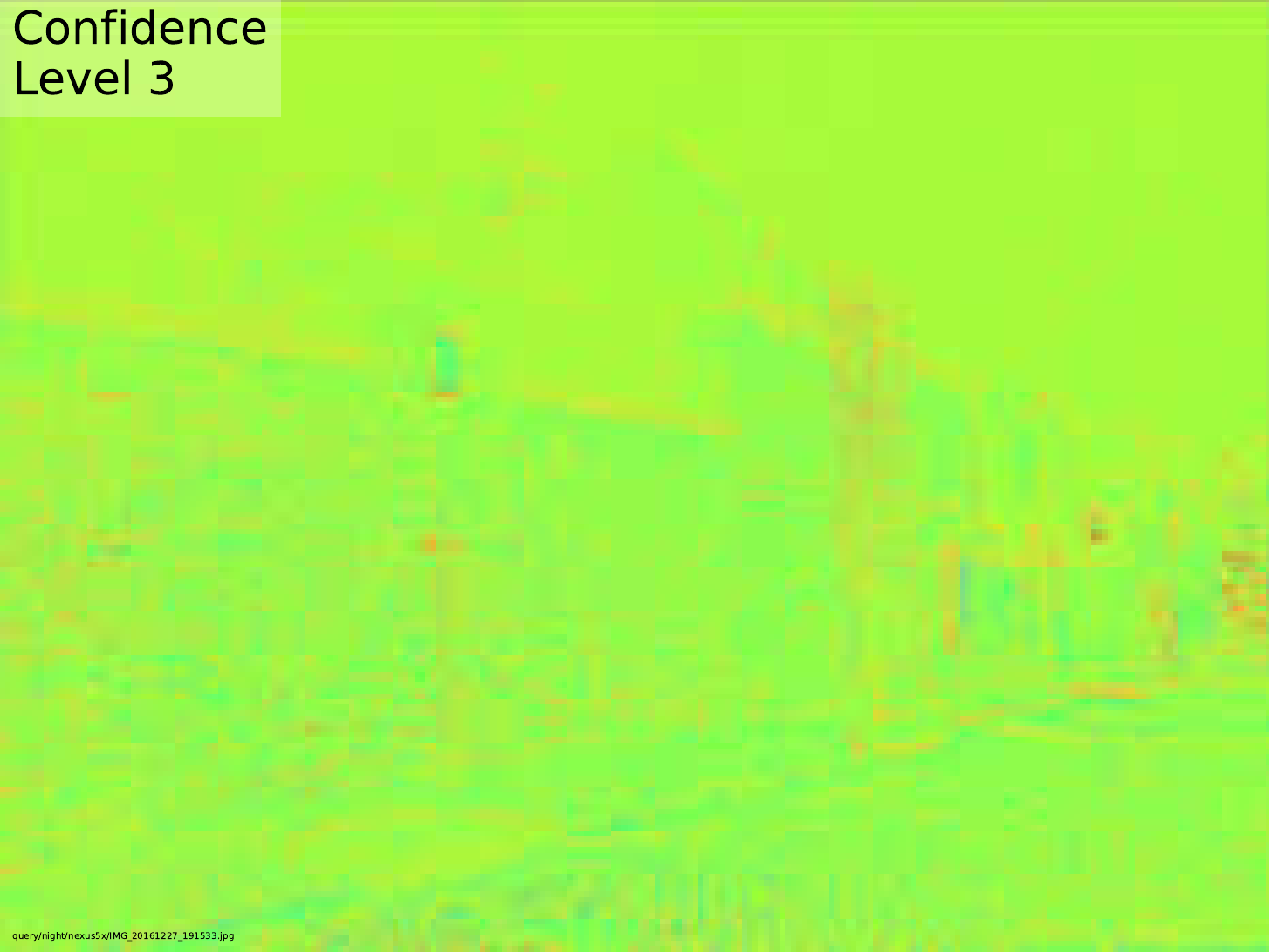}
\end{minipage}
\begin{minipage}{\lwidth\textwidth}
\rotatebox[origin=c]{90}{Reference}
\end{minipage}%
\begin{minipage}{\iwidth\textwidth}
    \centering
    \includegraphics[width=\pwidth\linewidth]{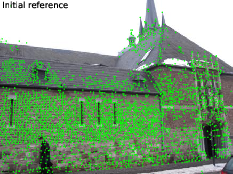}
\end{minipage}%
\begin{minipage}{\iwidth\textwidth}
    \centering
    \includegraphics[width=\pwidth\linewidth]{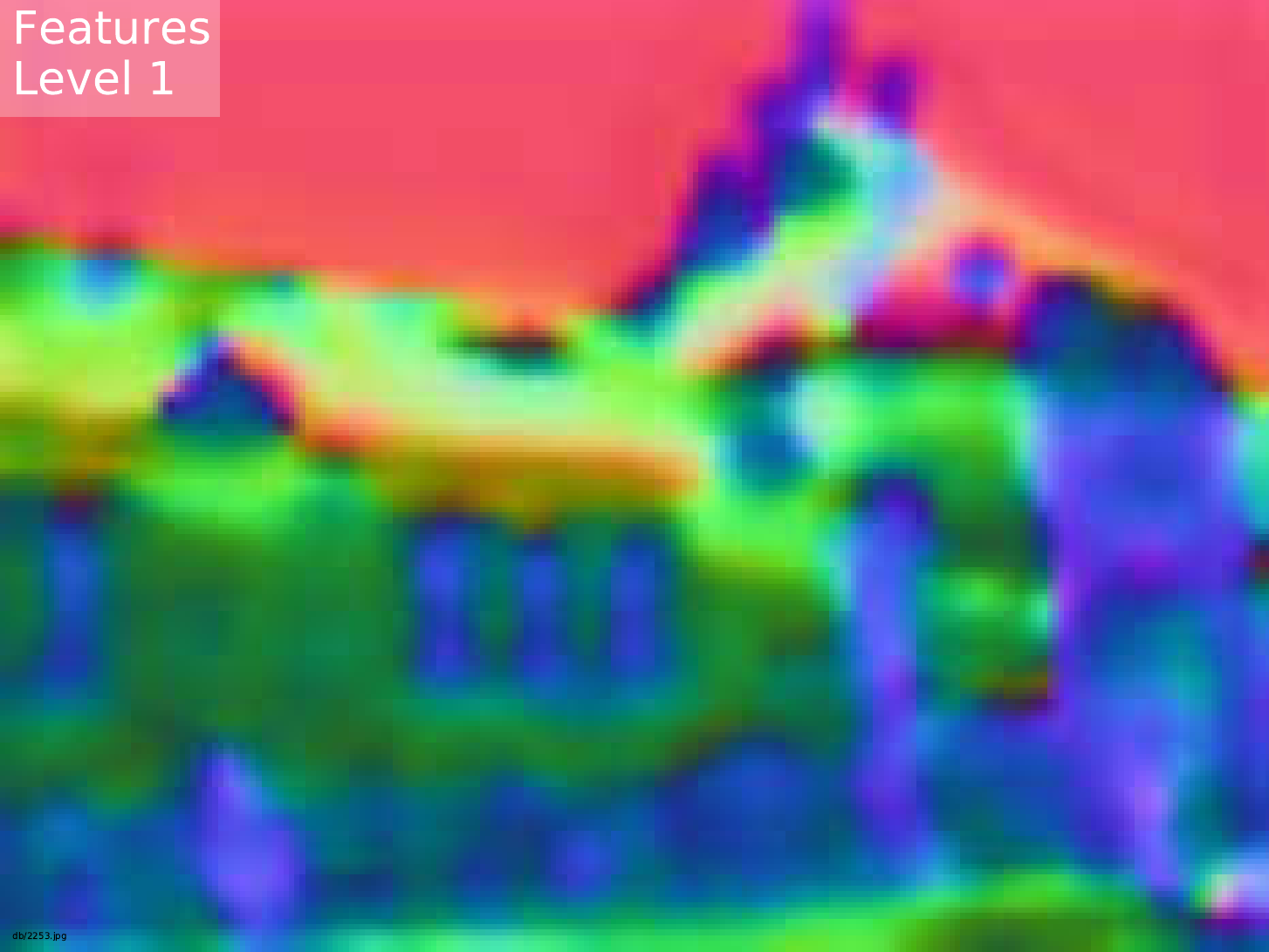}
\end{minipage}%
\begin{minipage}{\iwidth\textwidth}
    \centering
    \includegraphics[width=\pwidth\linewidth]{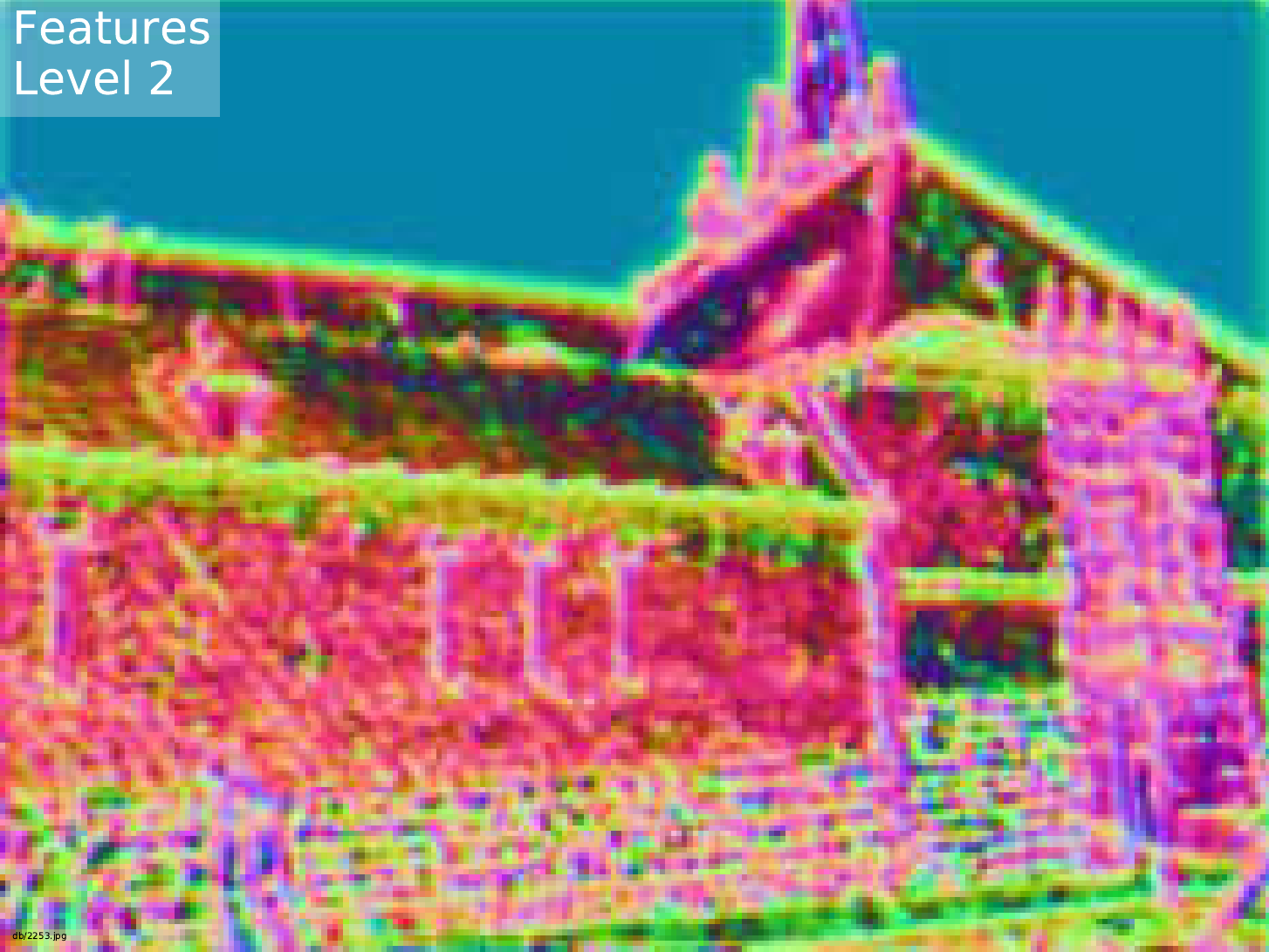}
\end{minipage}%
\begin{minipage}{\iwidth\textwidth}
    \centering
    \includegraphics[width=\pwidth\linewidth]{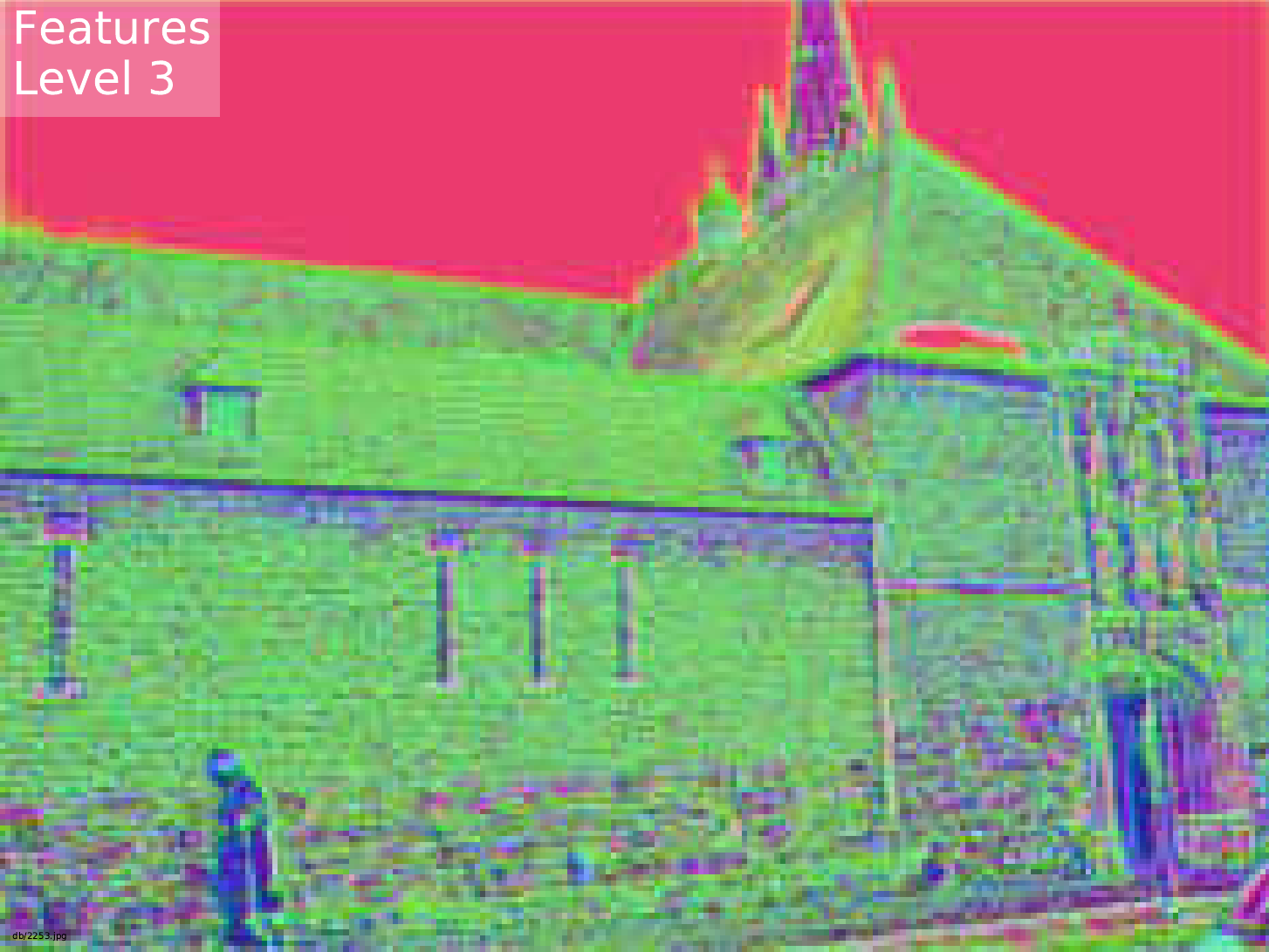}
\end{minipage}%
\begin{minipage}{\iwidth\textwidth}
    \centering
    \includegraphics[width=\pwidth\linewidth]{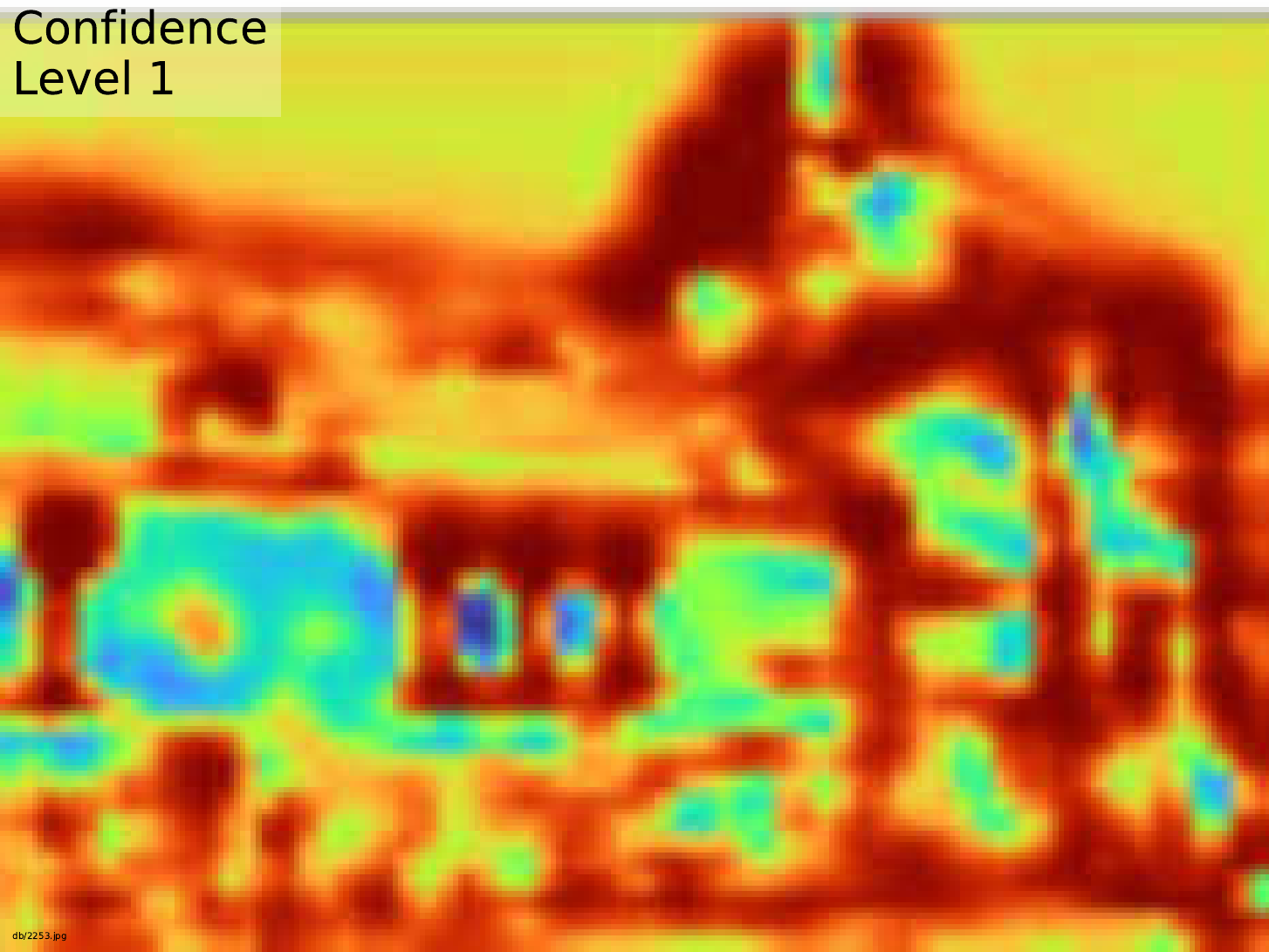}
\end{minipage}%
\begin{minipage}{\iwidth\textwidth}
    \centering
    \includegraphics[width=\pwidth\linewidth]{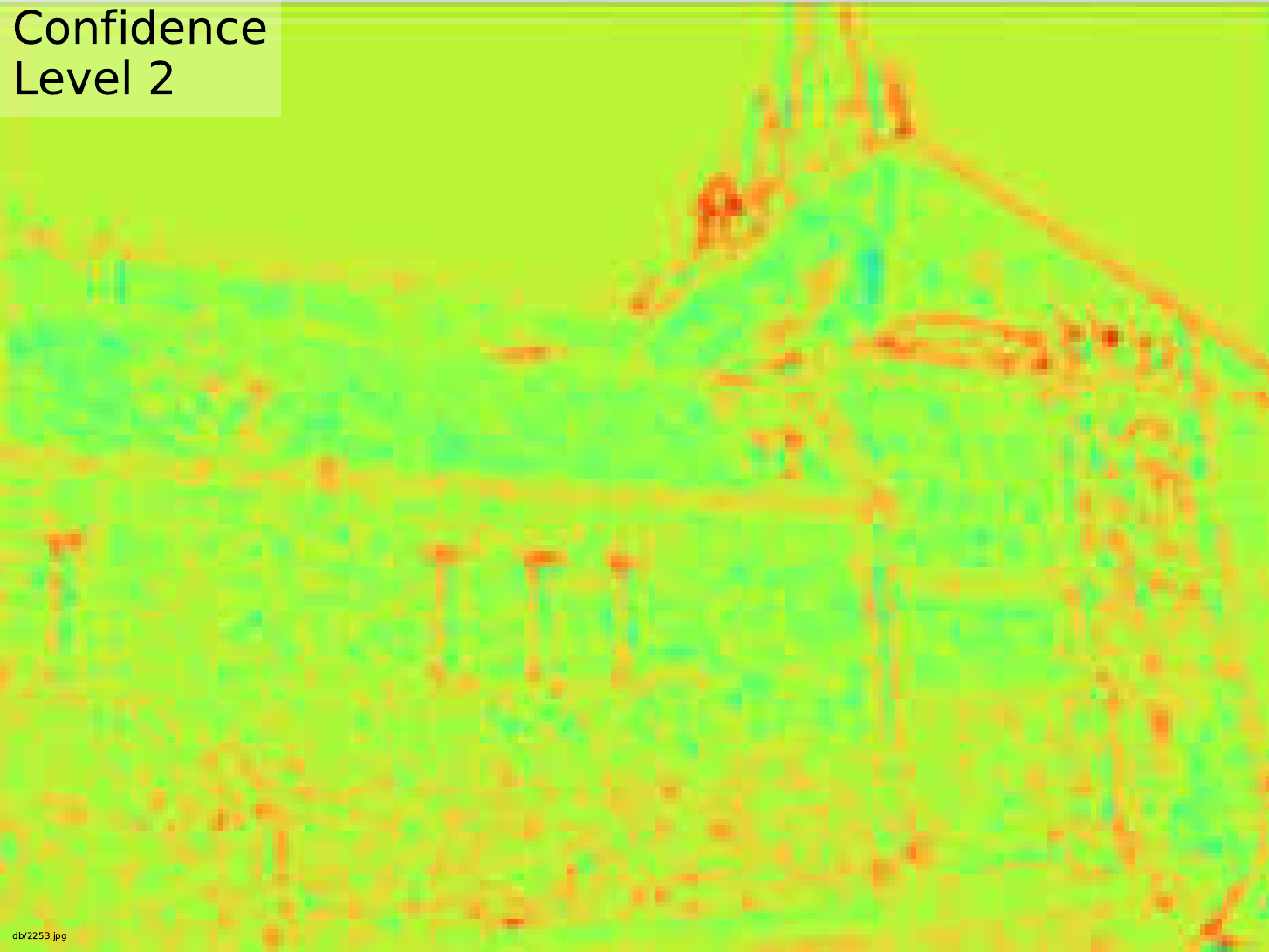}
\end{minipage}%
\begin{minipage}{\iwidth\textwidth}
    \centering
    \includegraphics[width=\pwidth\linewidth]{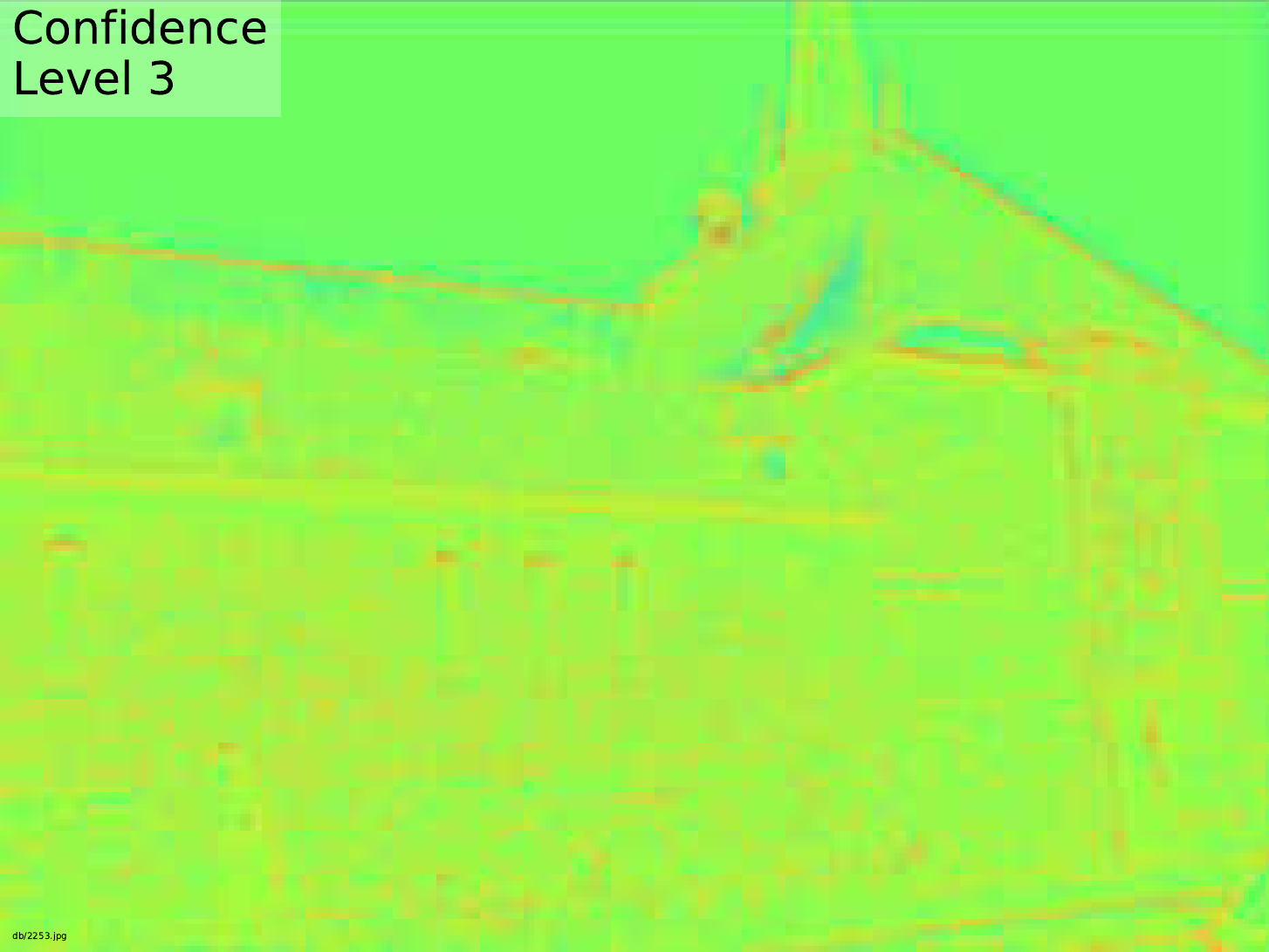}
\end{minipage}
\vspace{2mm}

\begin{minipage}{\lwidth\textwidth}
\rotatebox[origin=c]{90}{Query}
\end{minipage}%
\begin{minipage}{\iwidth\textwidth}
    \centering
    \includegraphics[width=\pwidth\linewidth]{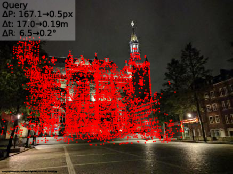}
\end{minipage}%
\begin{minipage}{\iwidth\textwidth}
    \centering
    \includegraphics[width=\pwidth\linewidth]{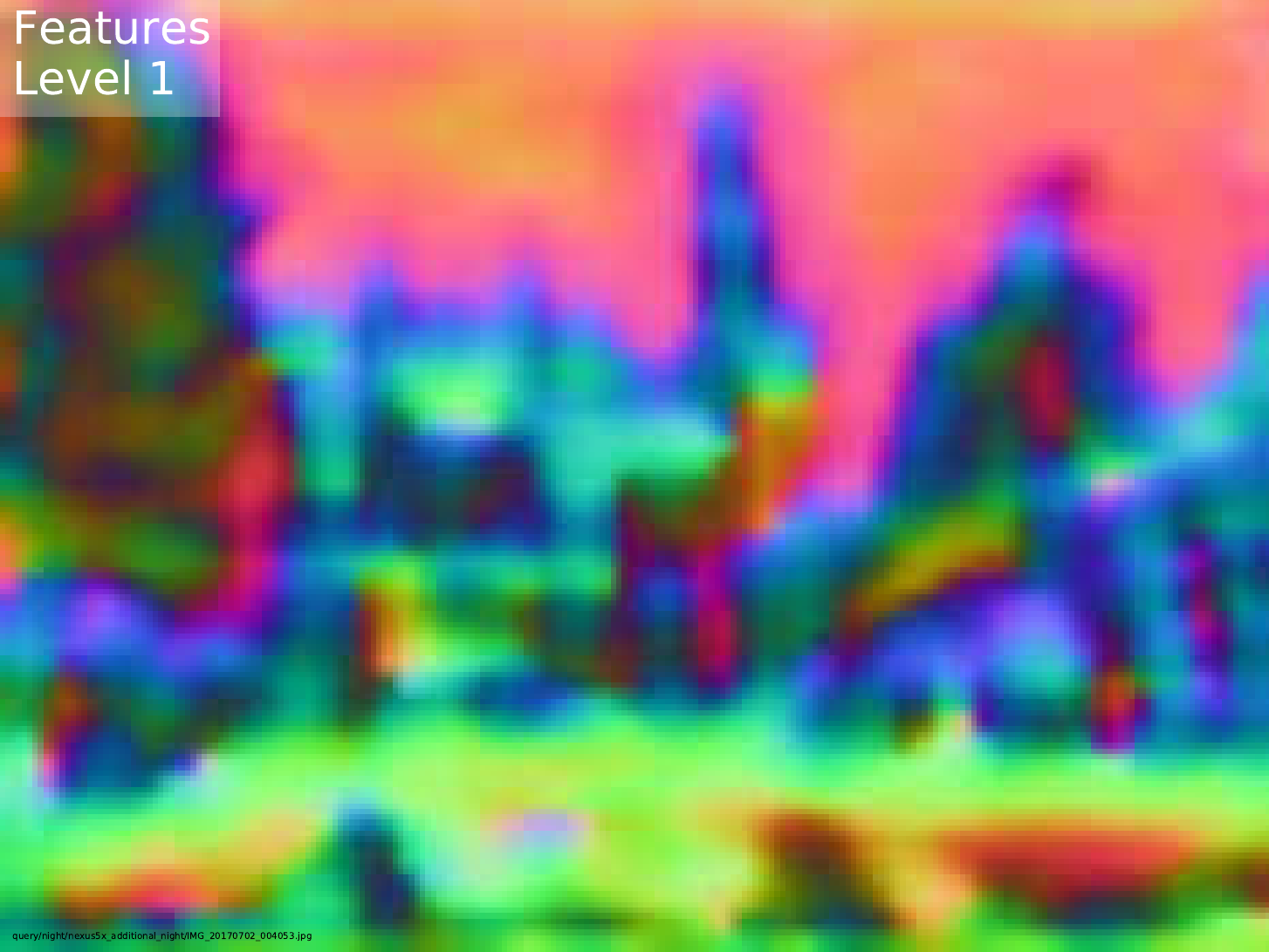}
\end{minipage}%
\begin{minipage}{\iwidth\textwidth}
    \centering
    \includegraphics[width=\pwidth\linewidth]{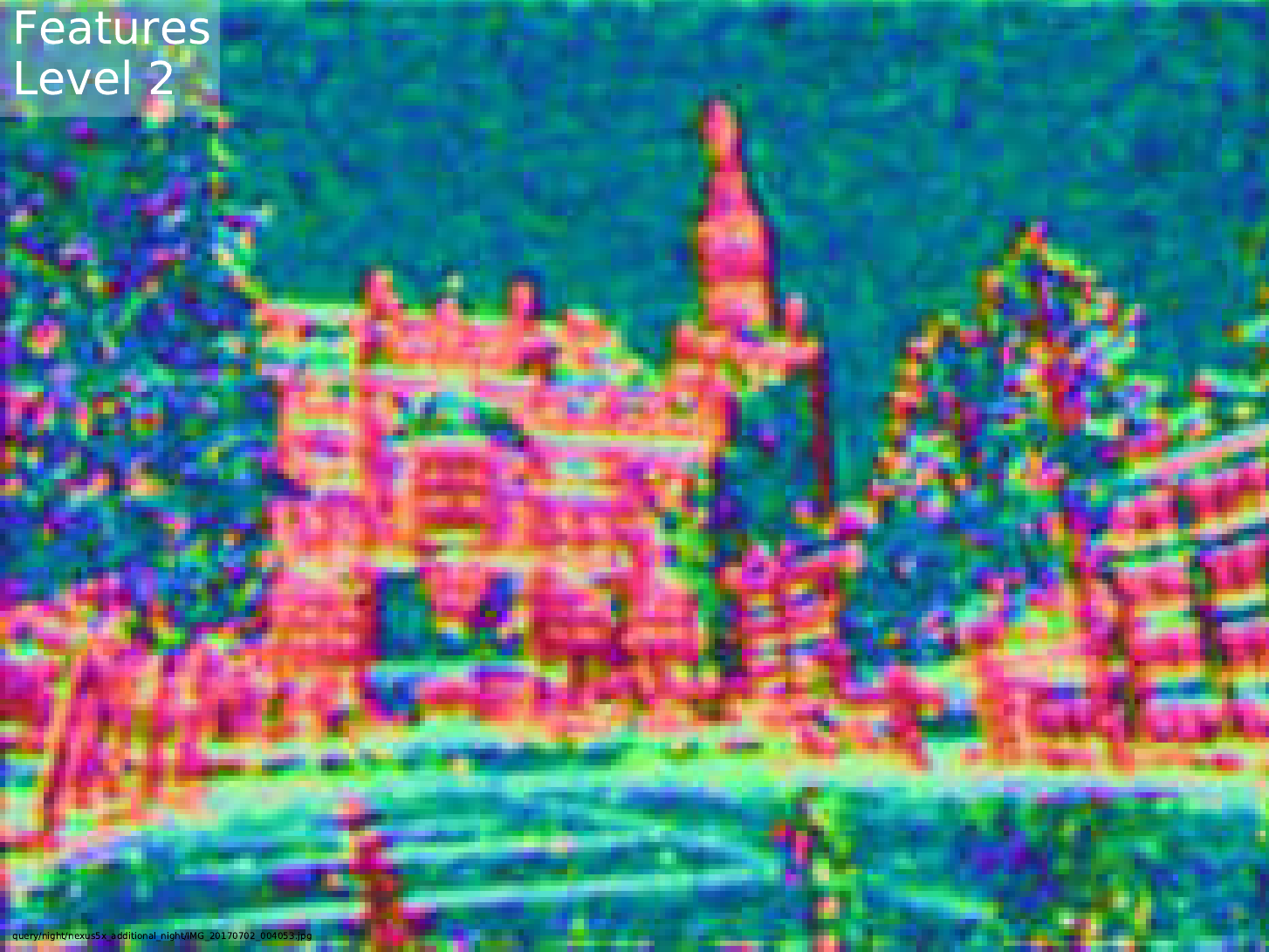}
\end{minipage}%
\begin{minipage}{\iwidth\textwidth}
    \centering
    \includegraphics[width=\pwidth\linewidth]{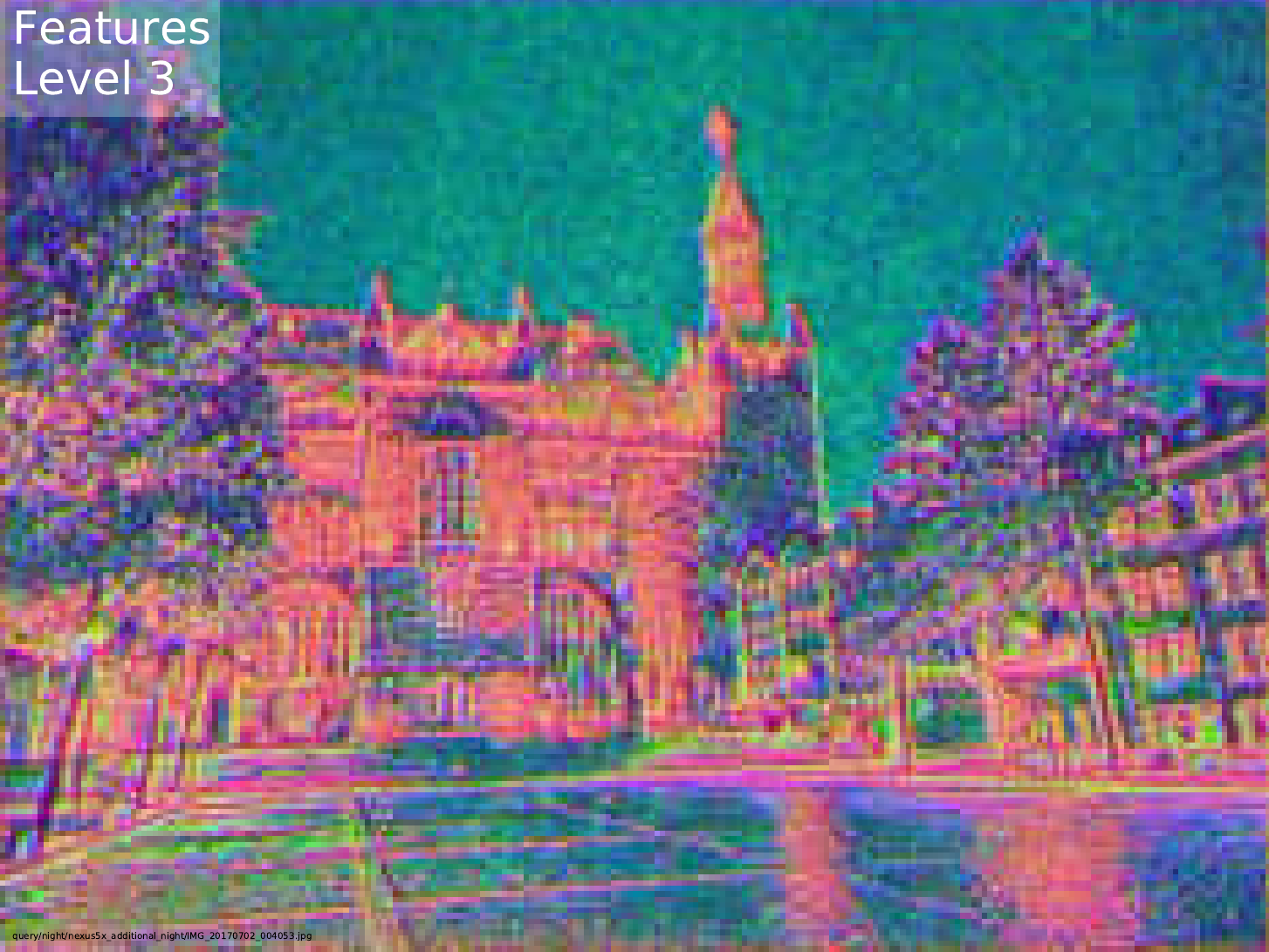}
\end{minipage}%
\begin{minipage}{\iwidth\textwidth}
    \centering
    \includegraphics[width=\pwidth\linewidth]{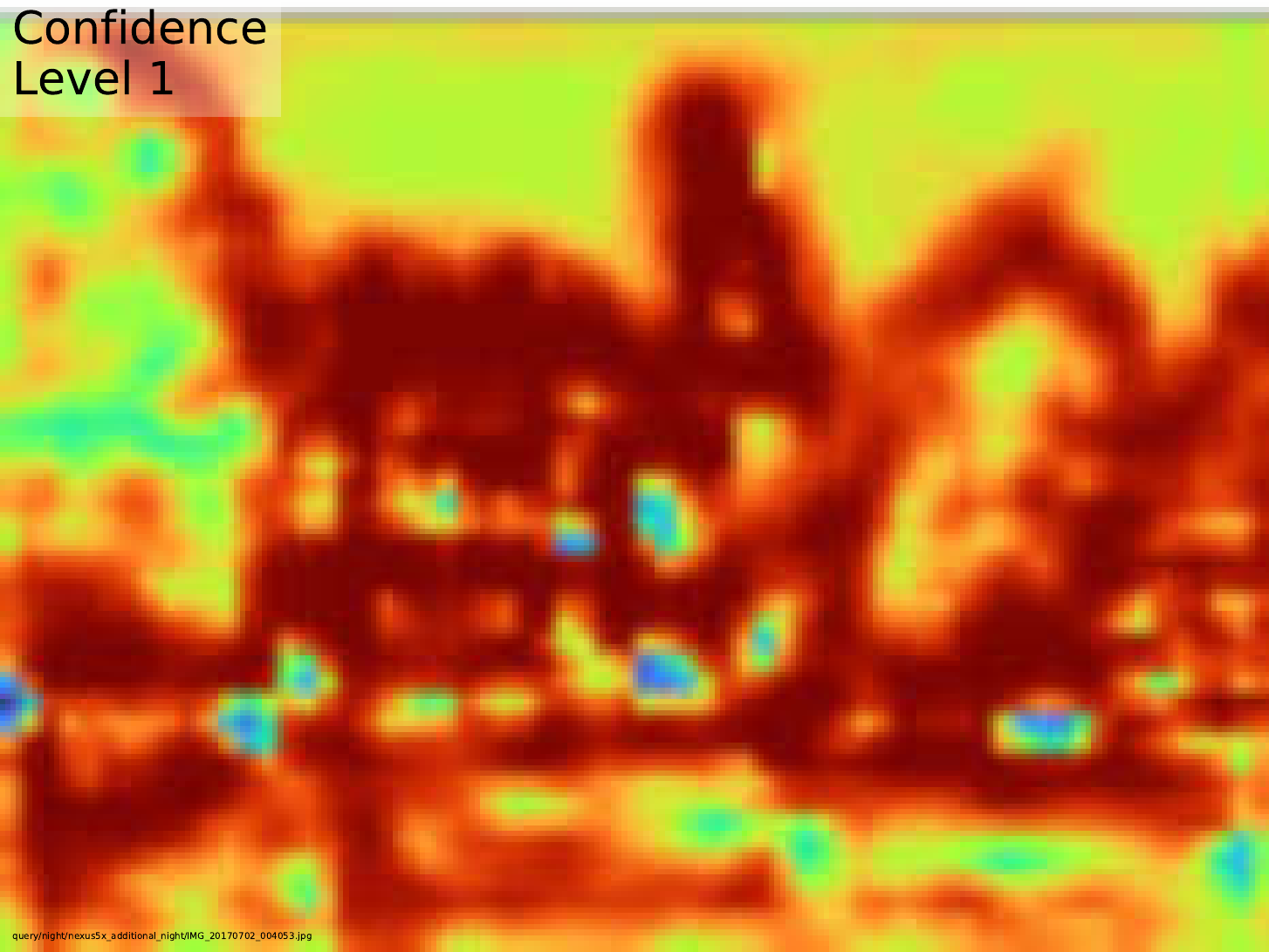}
\end{minipage}%
\begin{minipage}{\iwidth\textwidth}
    \centering
    \includegraphics[width=\pwidth\linewidth]{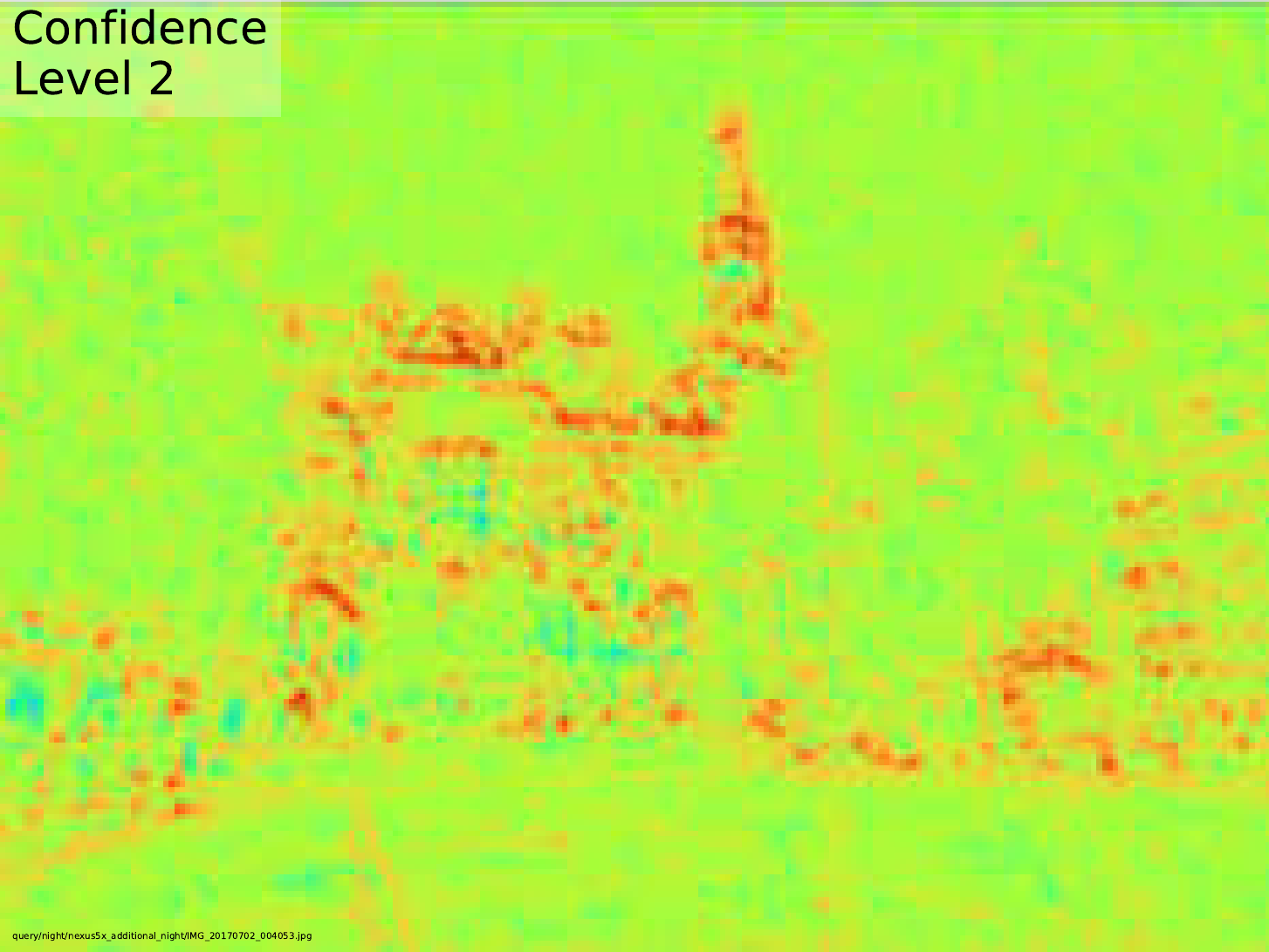}
\end{minipage}%
\begin{minipage}{\iwidth\textwidth}
    \centering
    \includegraphics[width=\pwidth\linewidth]{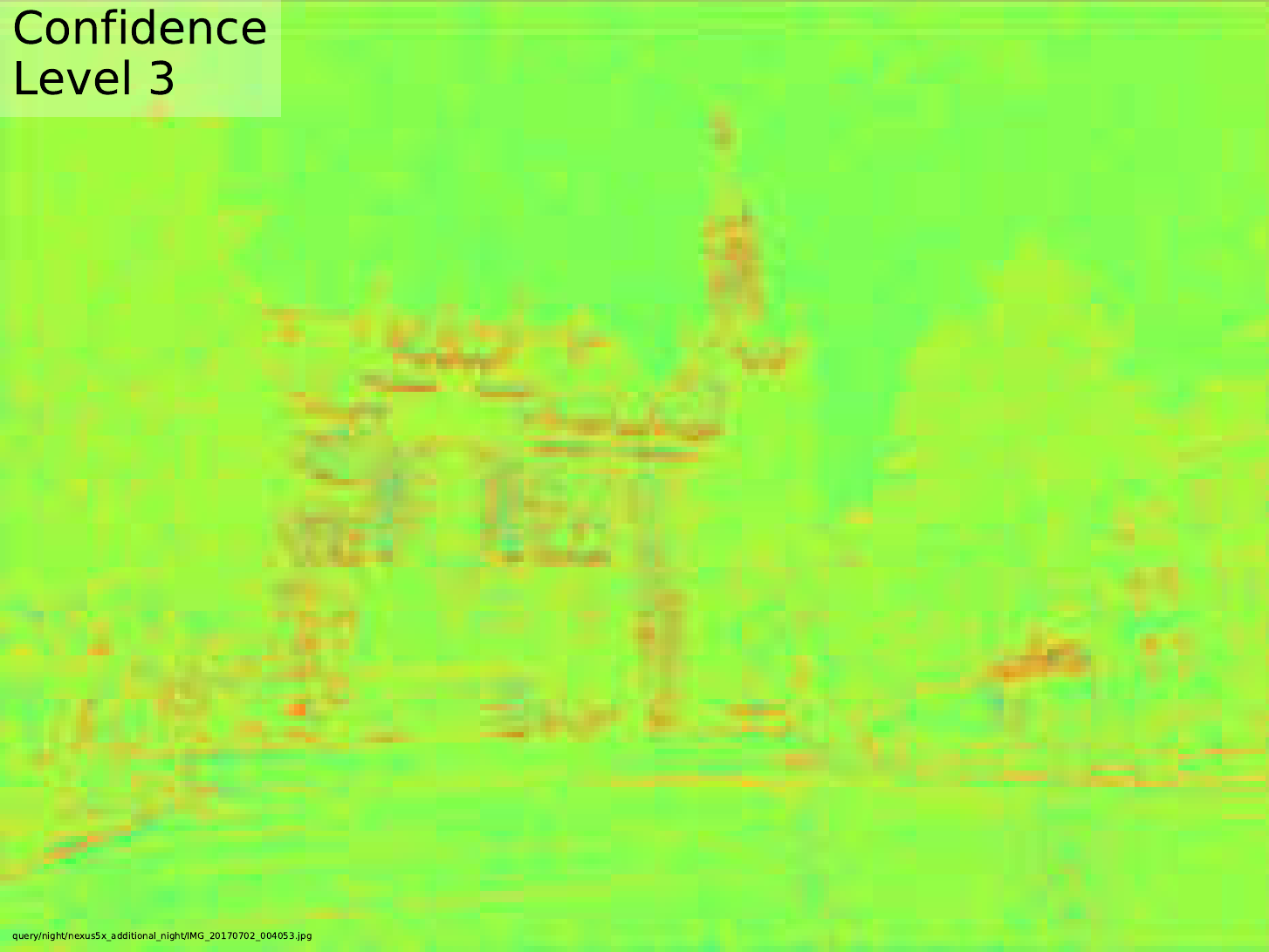}
\end{minipage}
\begin{minipage}{\lwidth\textwidth}
\rotatebox[origin=c]{90}{Reference}
\end{minipage}%
\begin{minipage}{\iwidth\textwidth}
    \centering
    \includegraphics[width=\pwidth\linewidth]{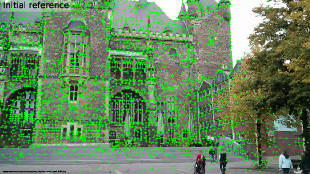}
\end{minipage}%
\begin{minipage}{\iwidth\textwidth}
    \centering
    \includegraphics[width=\pwidth\linewidth]{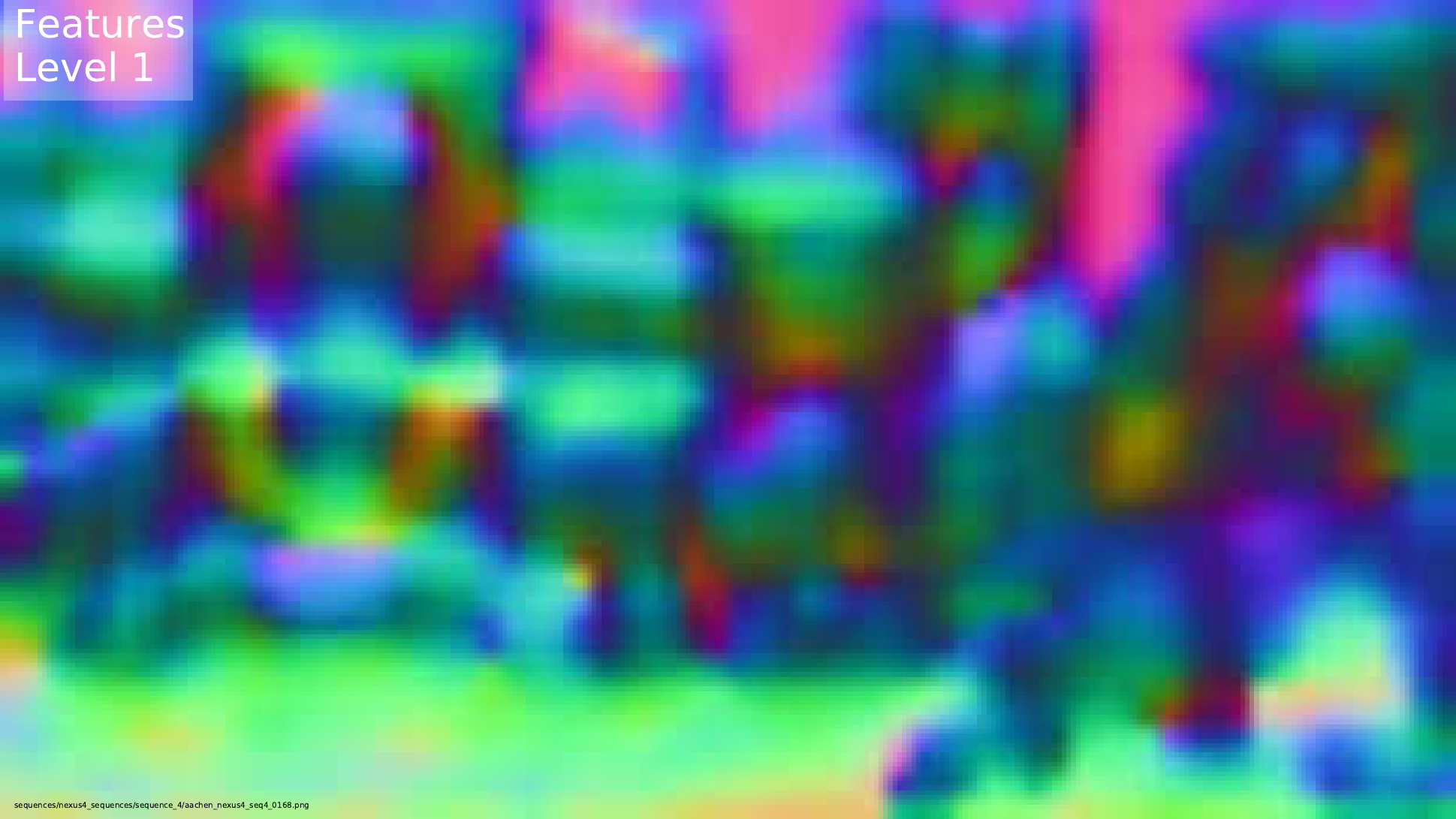}
\end{minipage}%
\begin{minipage}{\iwidth\textwidth}
    \centering
    \includegraphics[width=\pwidth\linewidth]{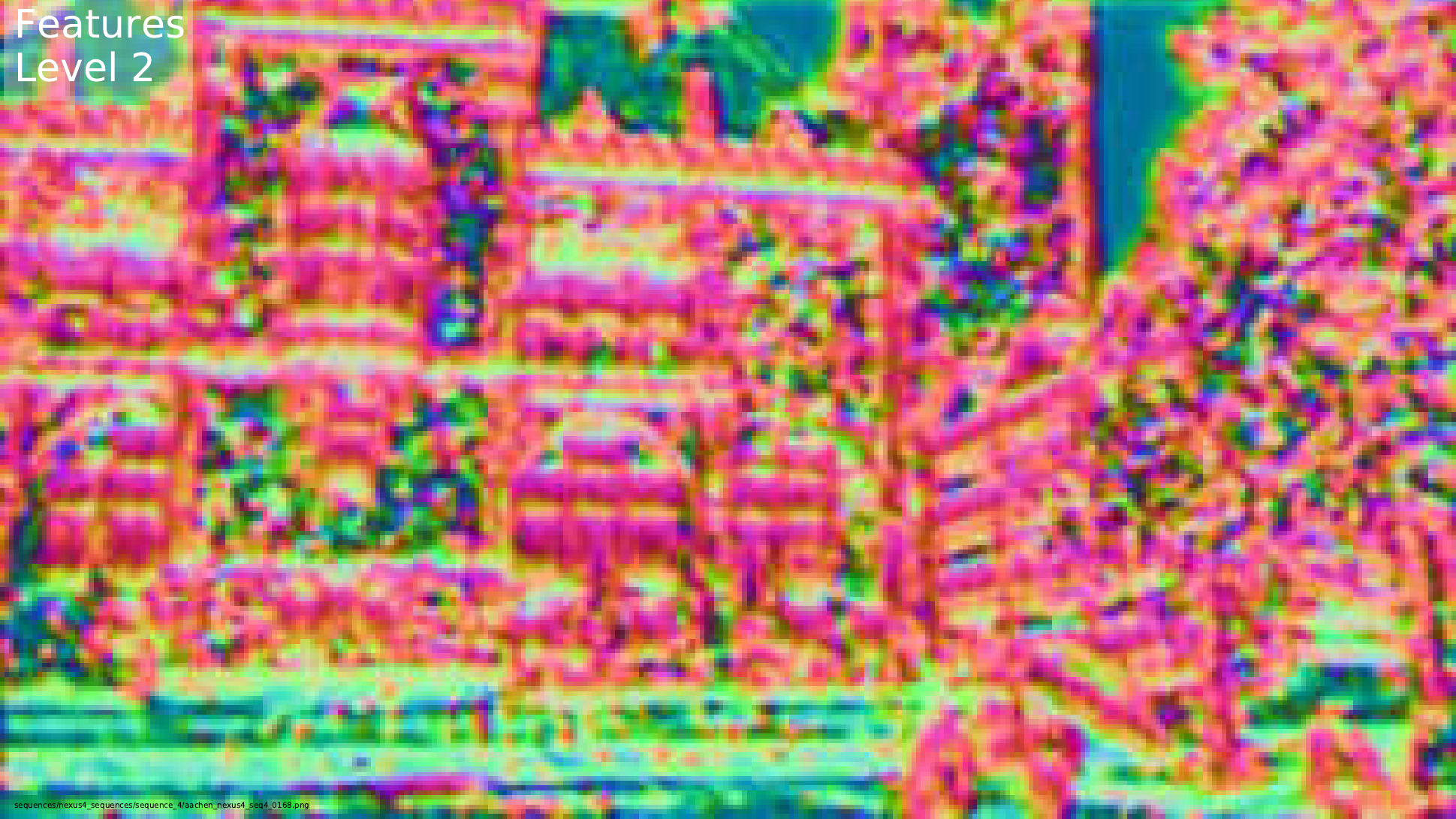}
\end{minipage}%
\begin{minipage}{\iwidth\textwidth}
    \centering
    \includegraphics[width=\pwidth\linewidth]{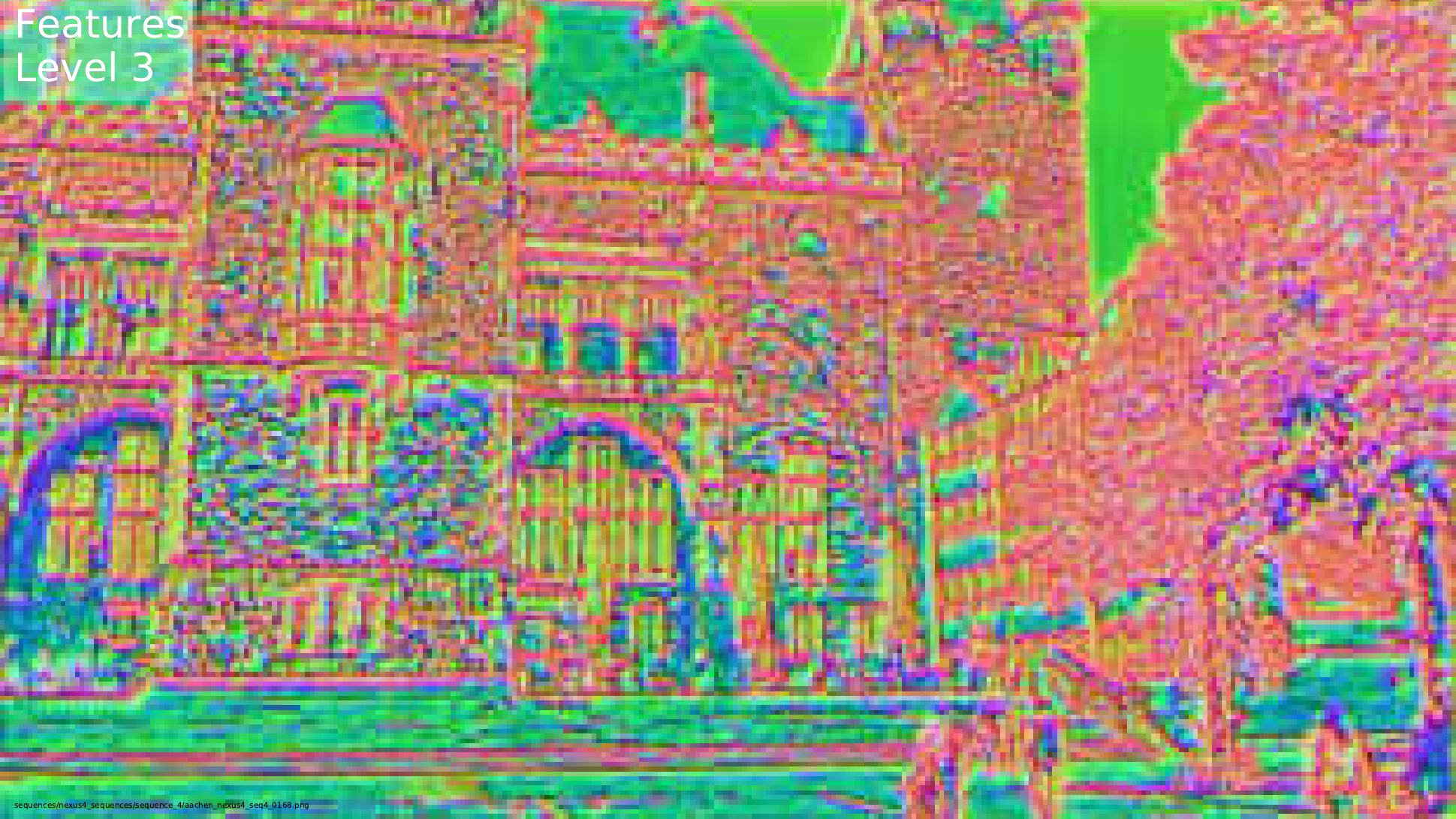}
\end{minipage}%
\begin{minipage}{\iwidth\textwidth}
    \centering
    \includegraphics[width=\pwidth\linewidth]{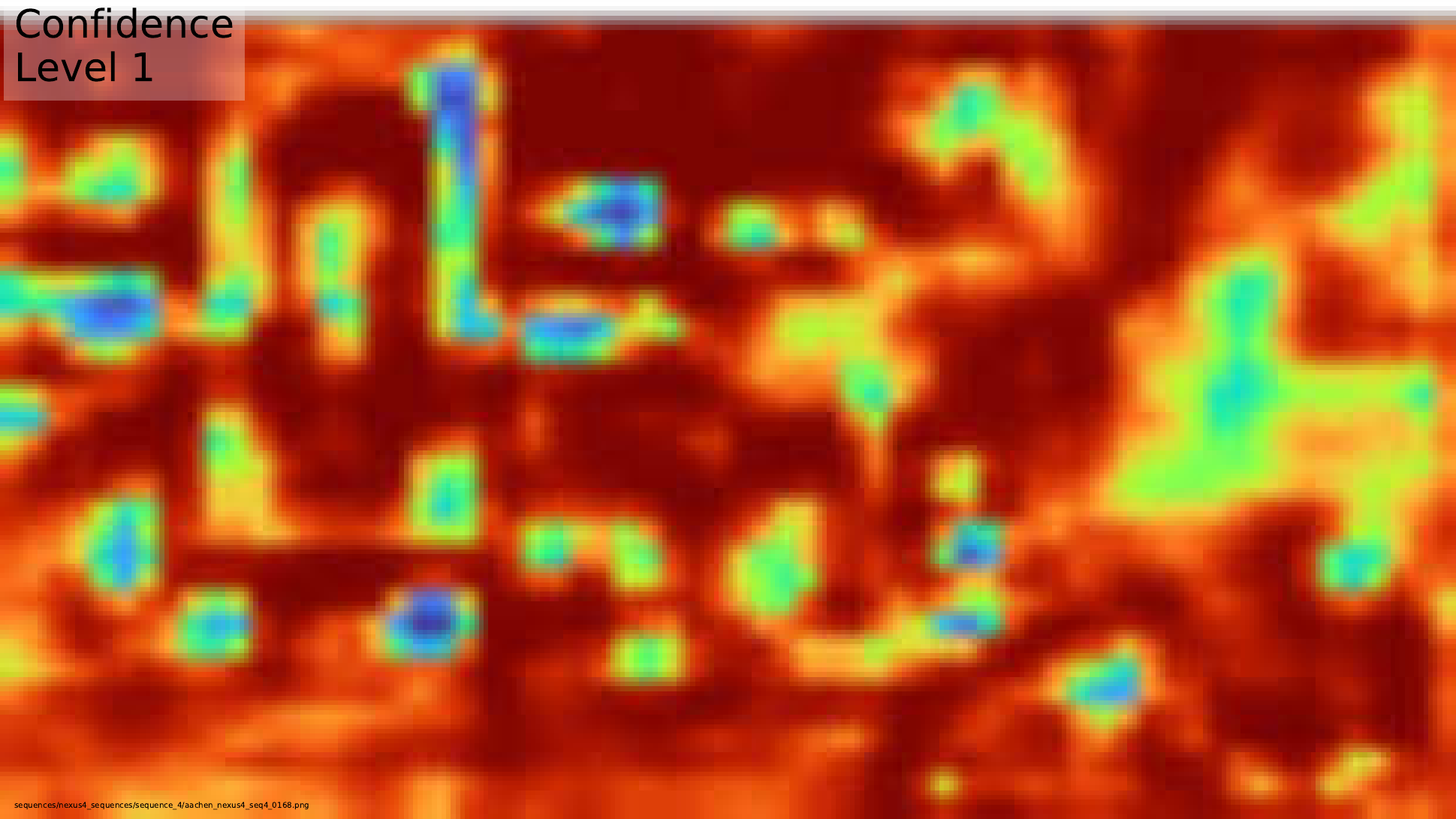}
\end{minipage}%
\begin{minipage}{\iwidth\textwidth}
    \centering
    \includegraphics[width=\pwidth\linewidth]{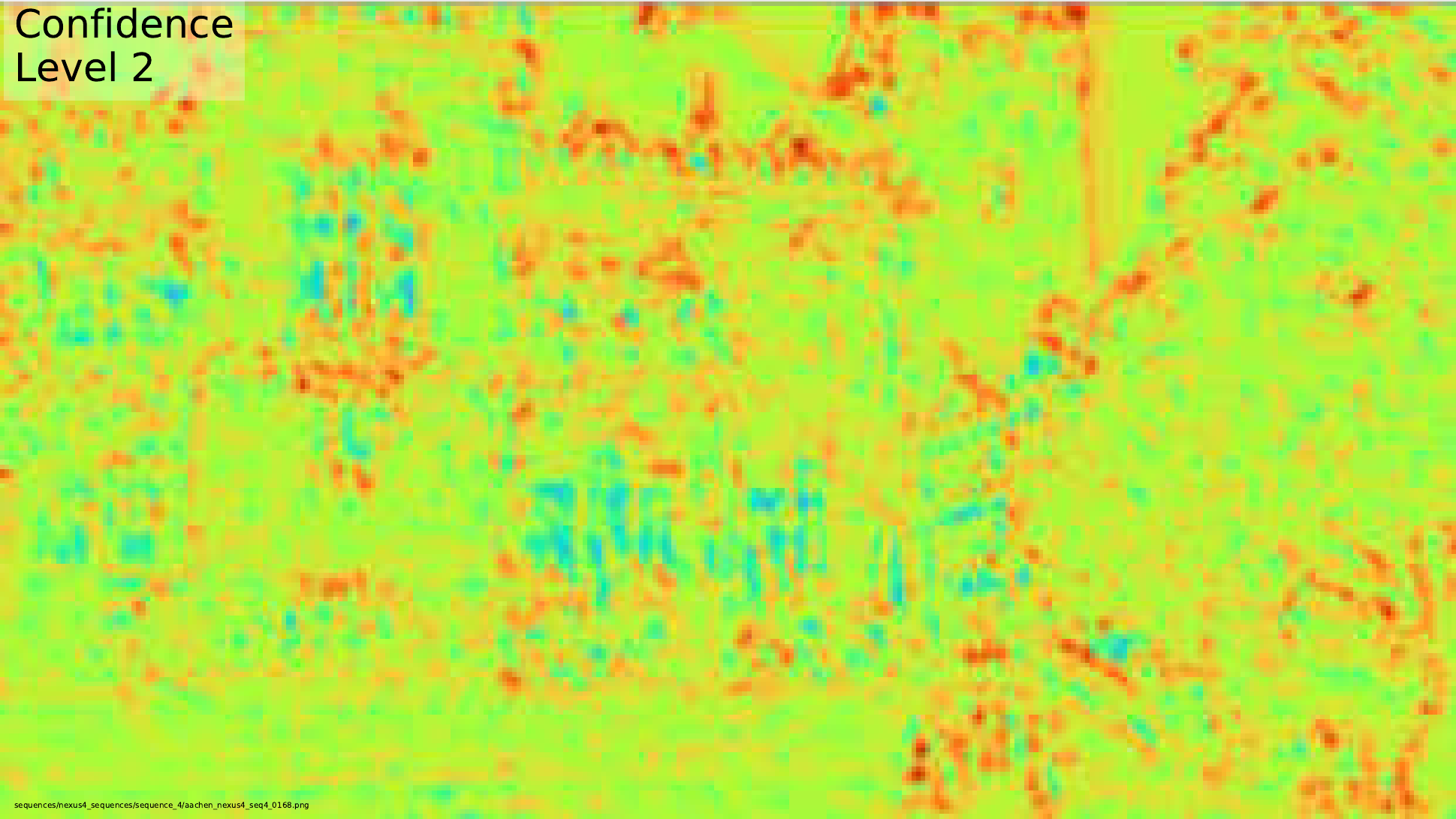}
\end{minipage}%
\begin{minipage}{\iwidth\textwidth}
    \centering
    \includegraphics[width=\pwidth\linewidth]{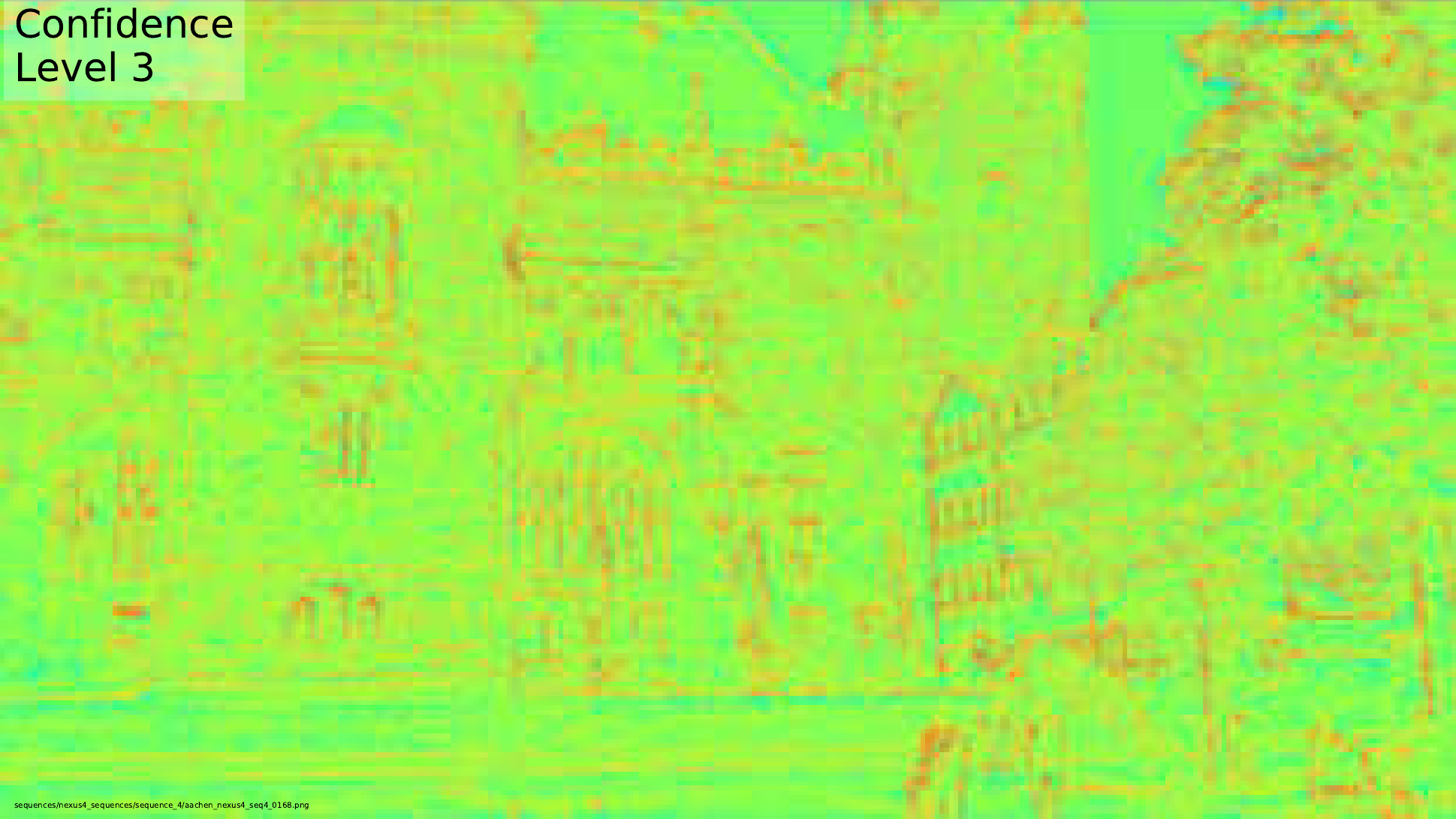}
\end{minipage}
\vspace{2mm}

\begin{minipage}{\lwidth\textwidth}
\rotatebox[origin=c]{90}{Query}
\end{minipage}%
\begin{minipage}{\iwidth\textwidth}
    \centering
    \includegraphics[width=\pwidth\linewidth]{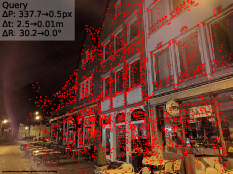}
\end{minipage}%
\begin{minipage}{\iwidth\textwidth}
    \centering
    \includegraphics[width=\pwidth\linewidth]{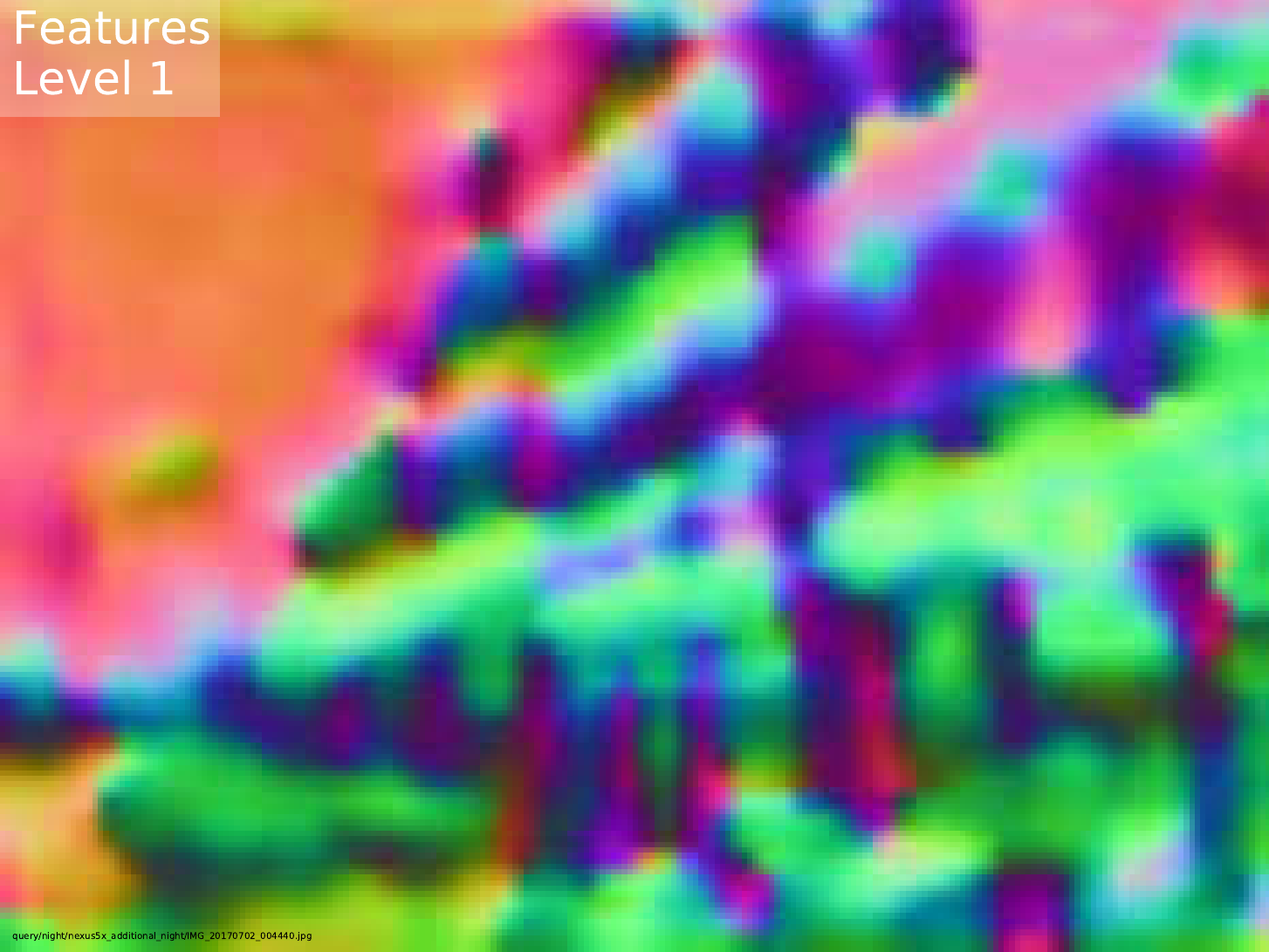}
\end{minipage}%
\begin{minipage}{\iwidth\textwidth}
    \centering
    \includegraphics[width=\pwidth\linewidth]{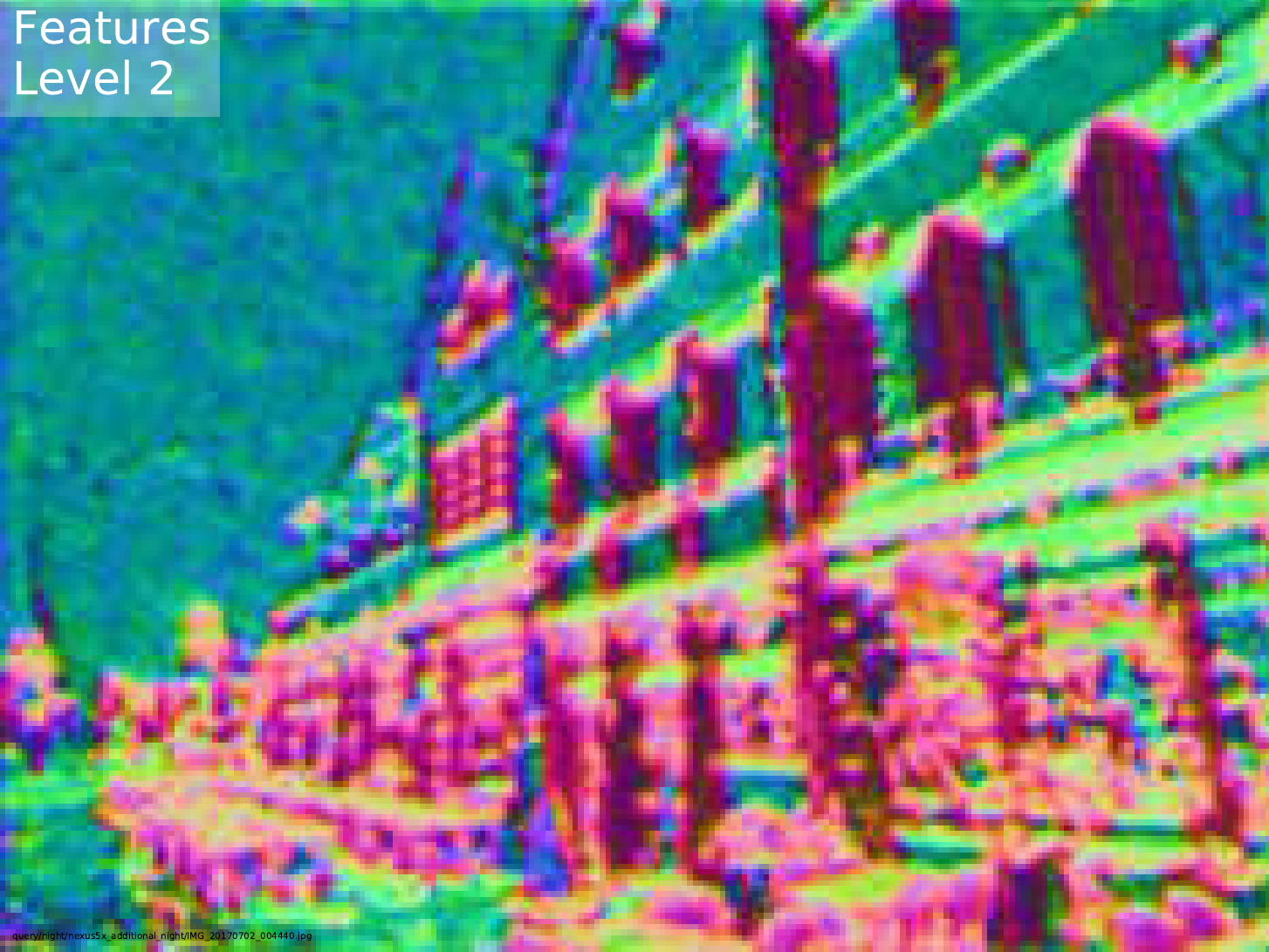}
\end{minipage}%
\begin{minipage}{\iwidth\textwidth}
    \centering
    \includegraphics[width=\pwidth\linewidth]{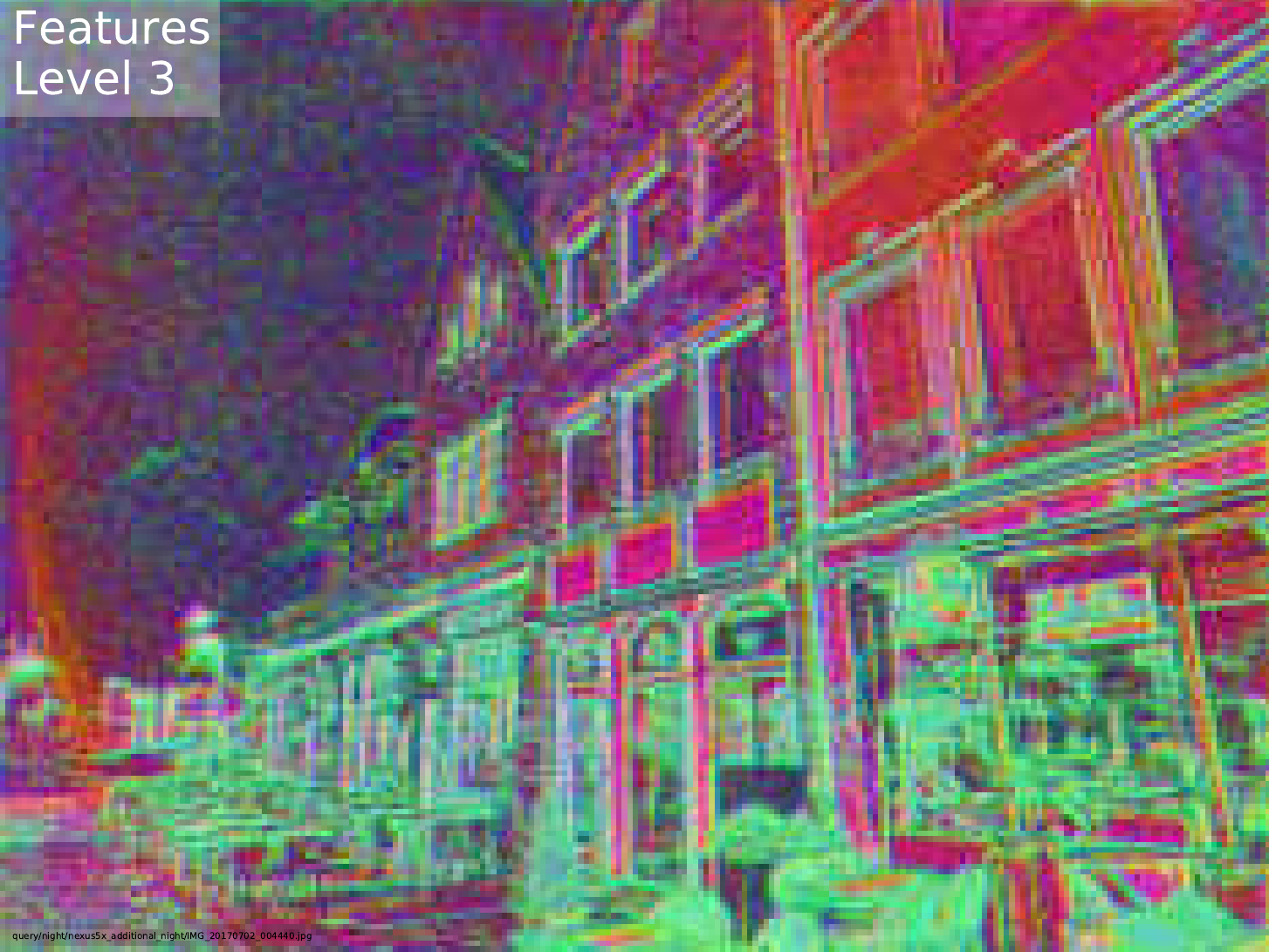}
\end{minipage}%
\begin{minipage}{\iwidth\textwidth}
    \centering
    \includegraphics[width=\pwidth\linewidth]{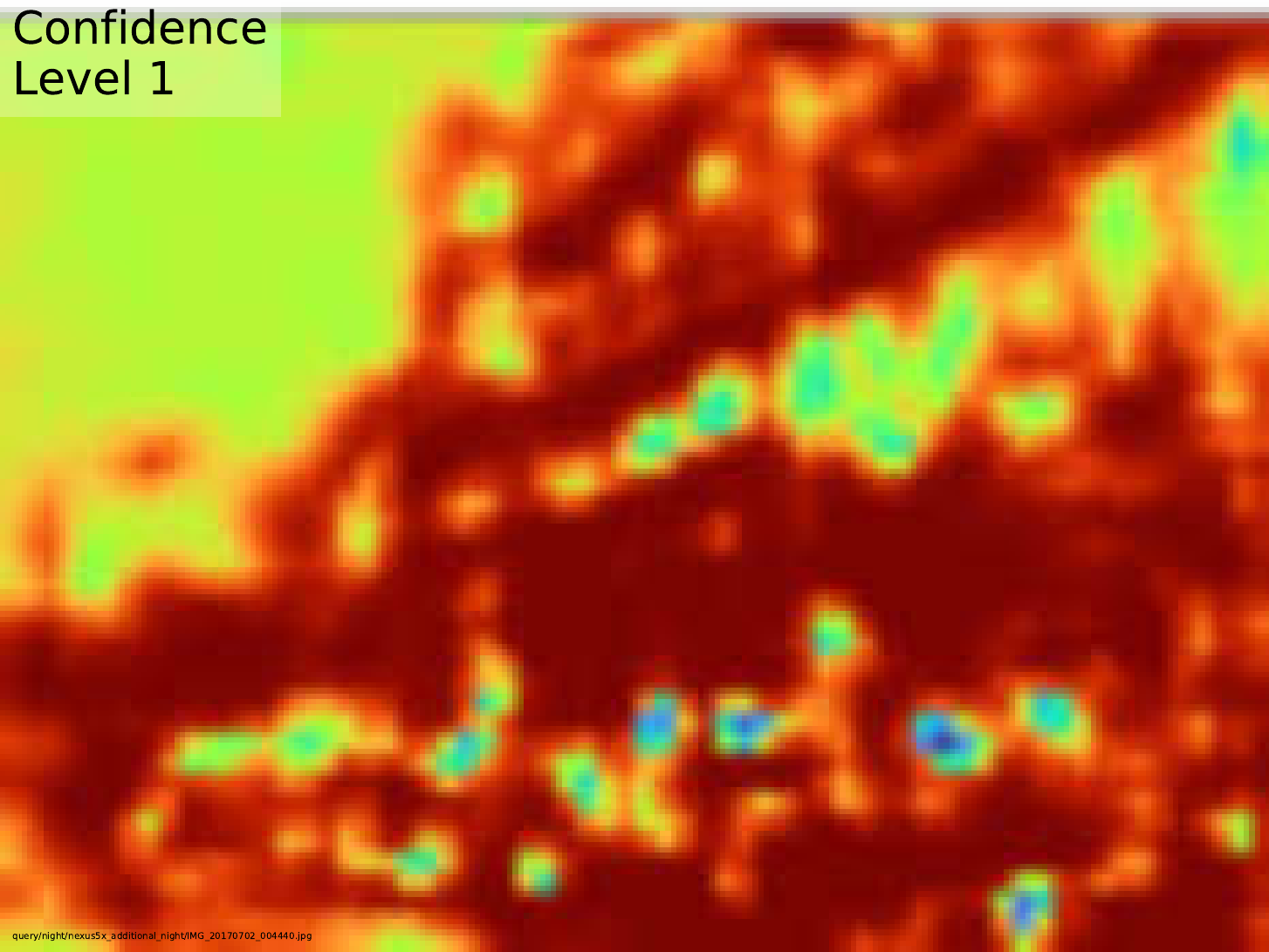}
\end{minipage}%
\begin{minipage}{\iwidth\textwidth}
    \centering
    \includegraphics[width=\pwidth\linewidth]{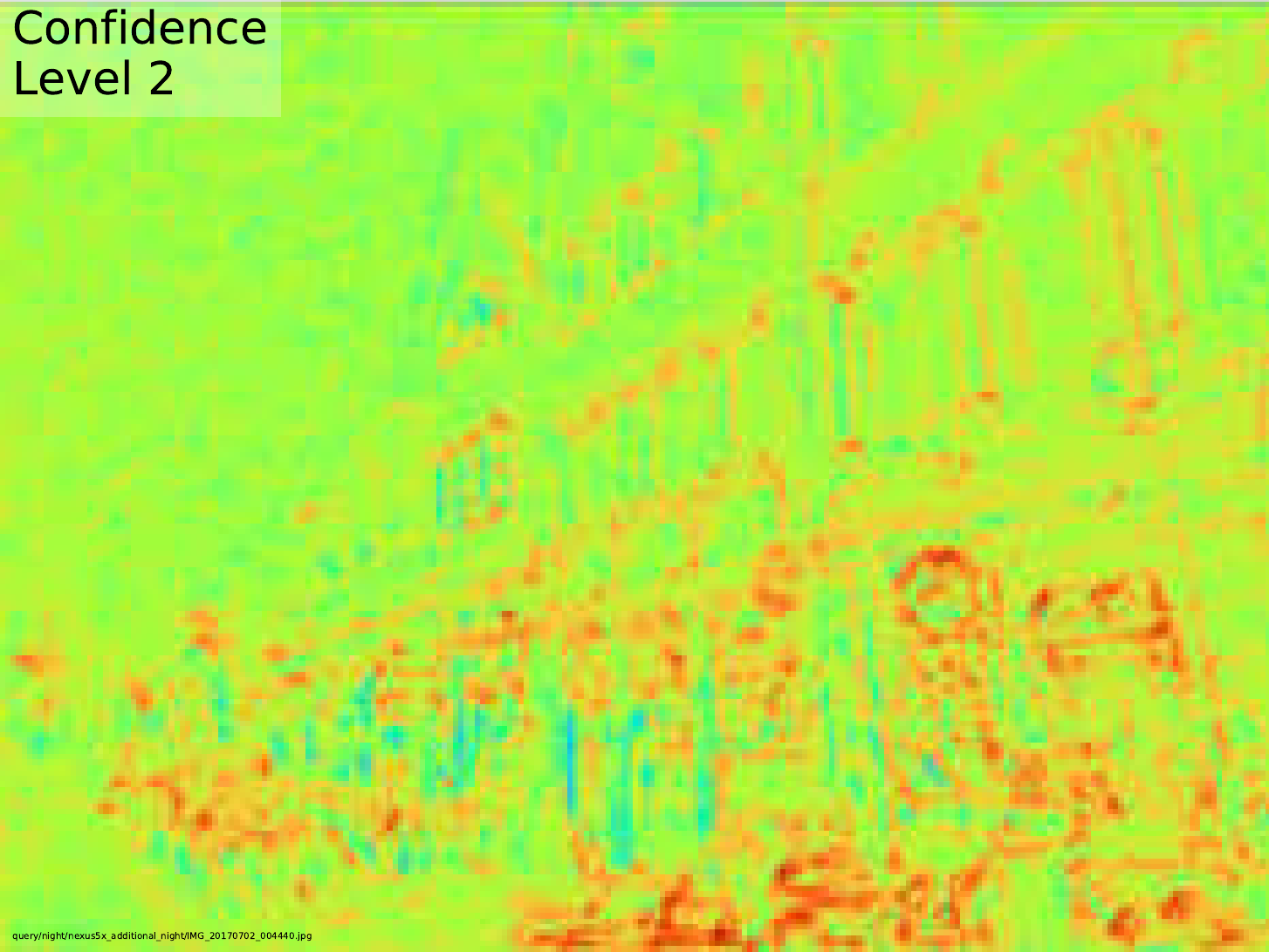}
\end{minipage}%
\begin{minipage}{\iwidth\textwidth}
    \centering
    \includegraphics[width=\pwidth\linewidth]{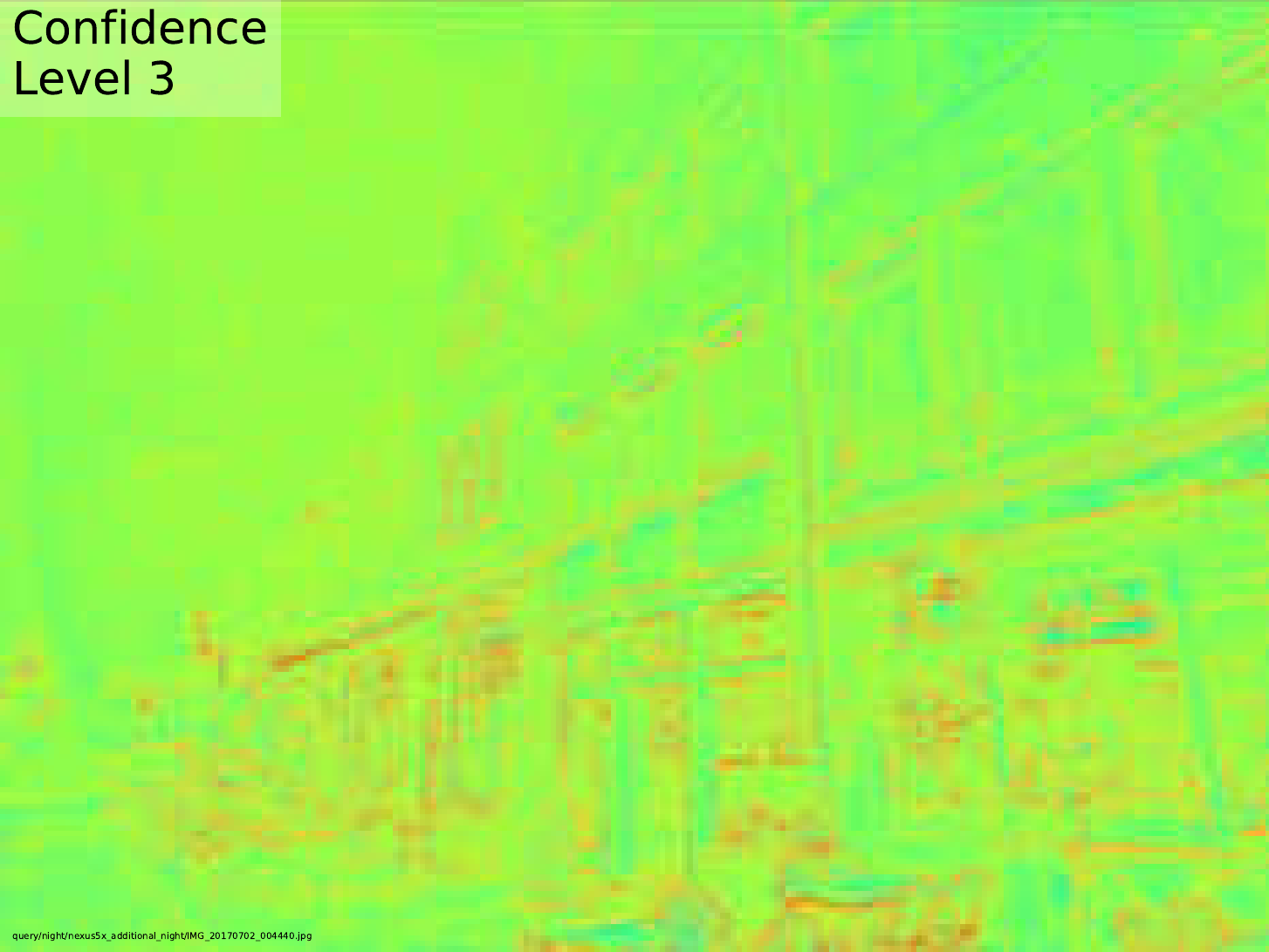}
\end{minipage}
\begin{minipage}{\lwidth\textwidth}
\rotatebox[origin=c]{90}{Reference}
\end{minipage}%
\begin{minipage}{\iwidth\textwidth}
    \centering
    \includegraphics[width=\pwidth\linewidth]{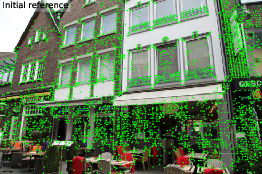}
\end{minipage}%
\begin{minipage}{\iwidth\textwidth}
    \centering
    \includegraphics[width=\pwidth\linewidth]{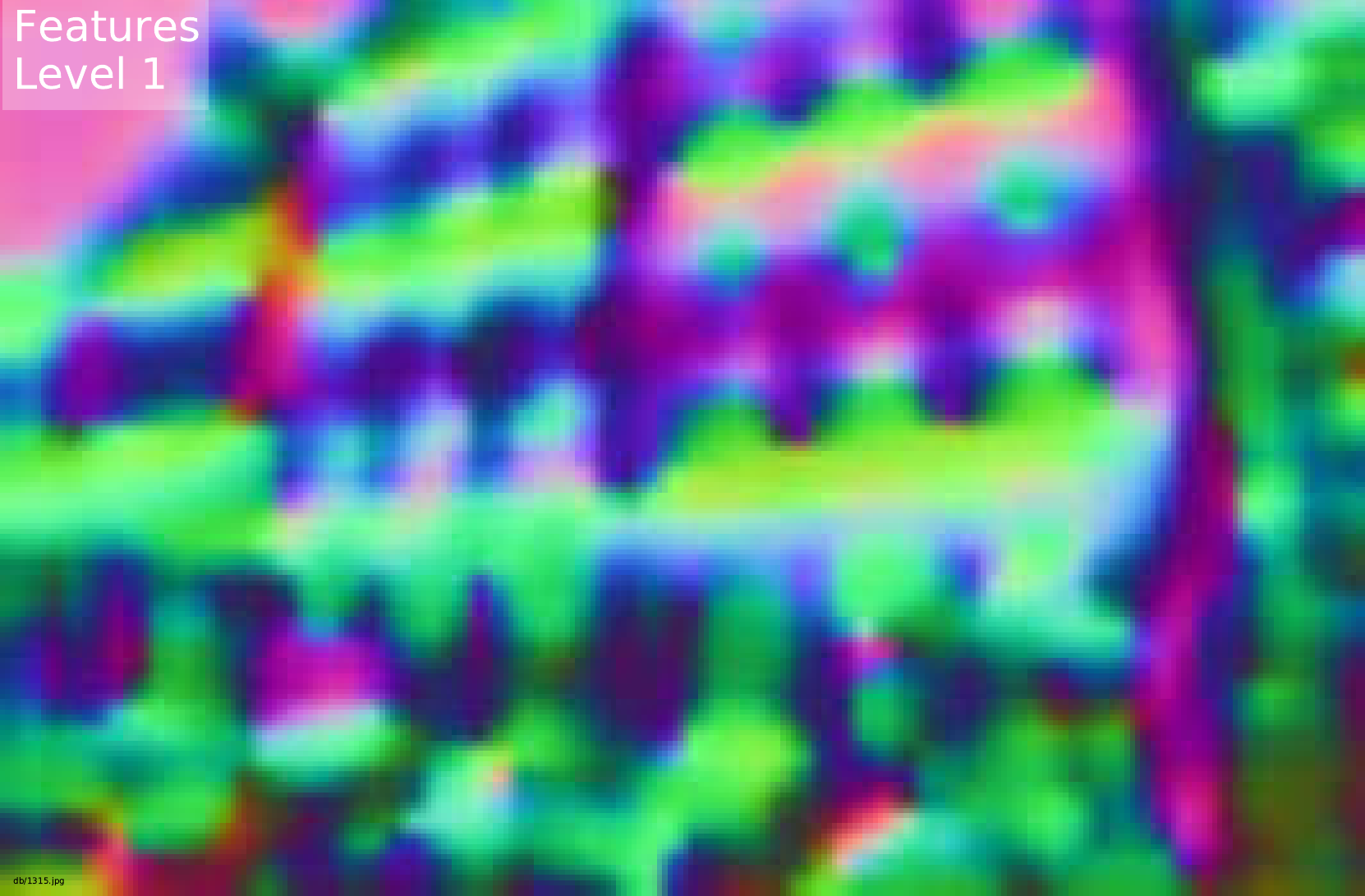}
\end{minipage}%
\begin{minipage}{\iwidth\textwidth}
    \centering
    \includegraphics[width=\pwidth\linewidth]{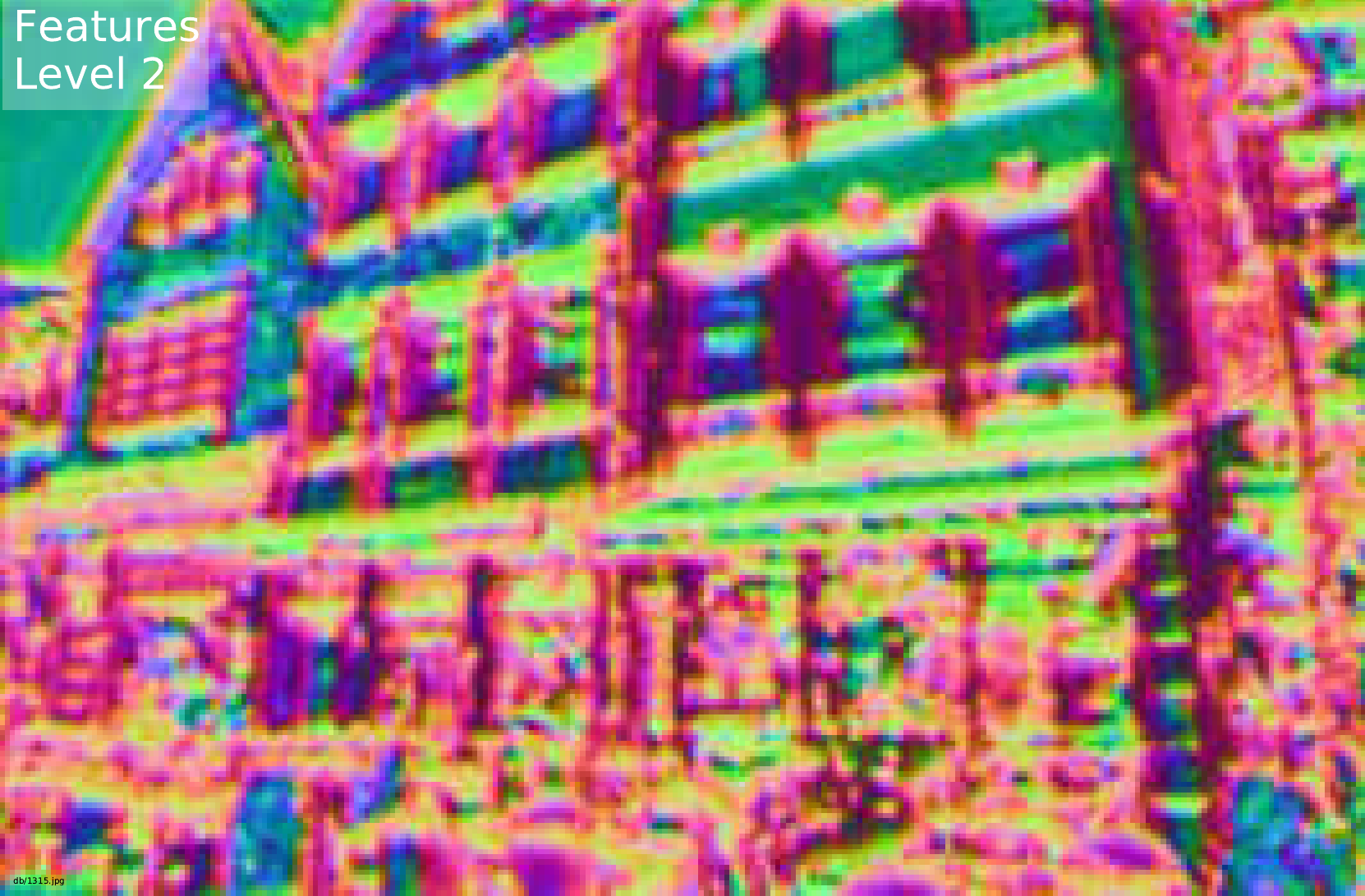}
\end{minipage}%
\begin{minipage}{\iwidth\textwidth}
    \centering
    \includegraphics[width=\pwidth\linewidth]{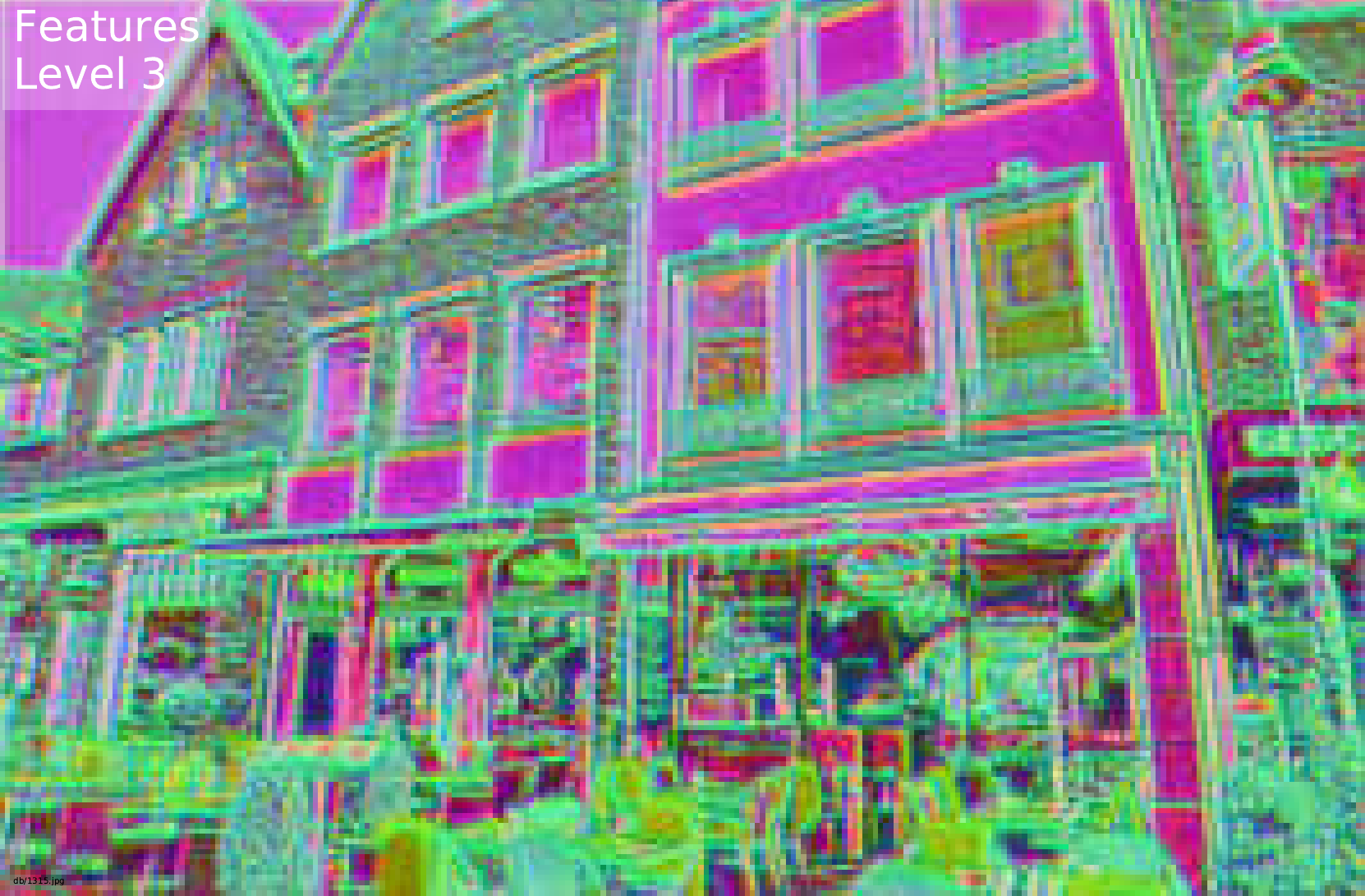}
\end{minipage}%
\begin{minipage}{\iwidth\textwidth}
    \centering
    \includegraphics[width=\pwidth\linewidth]{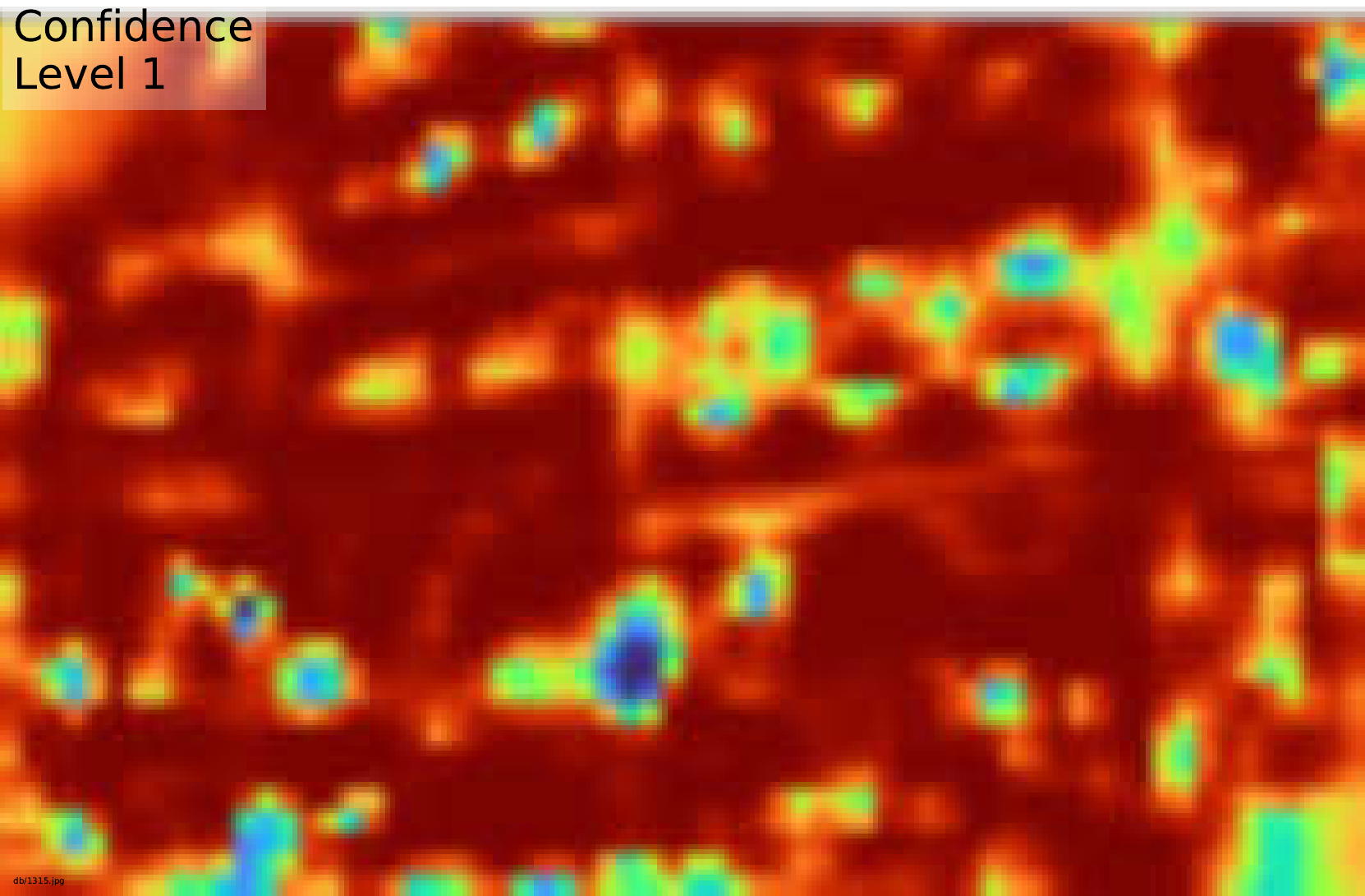}
\end{minipage}%
\begin{minipage}{\iwidth\textwidth}
    \centering
    \includegraphics[width=\pwidth\linewidth]{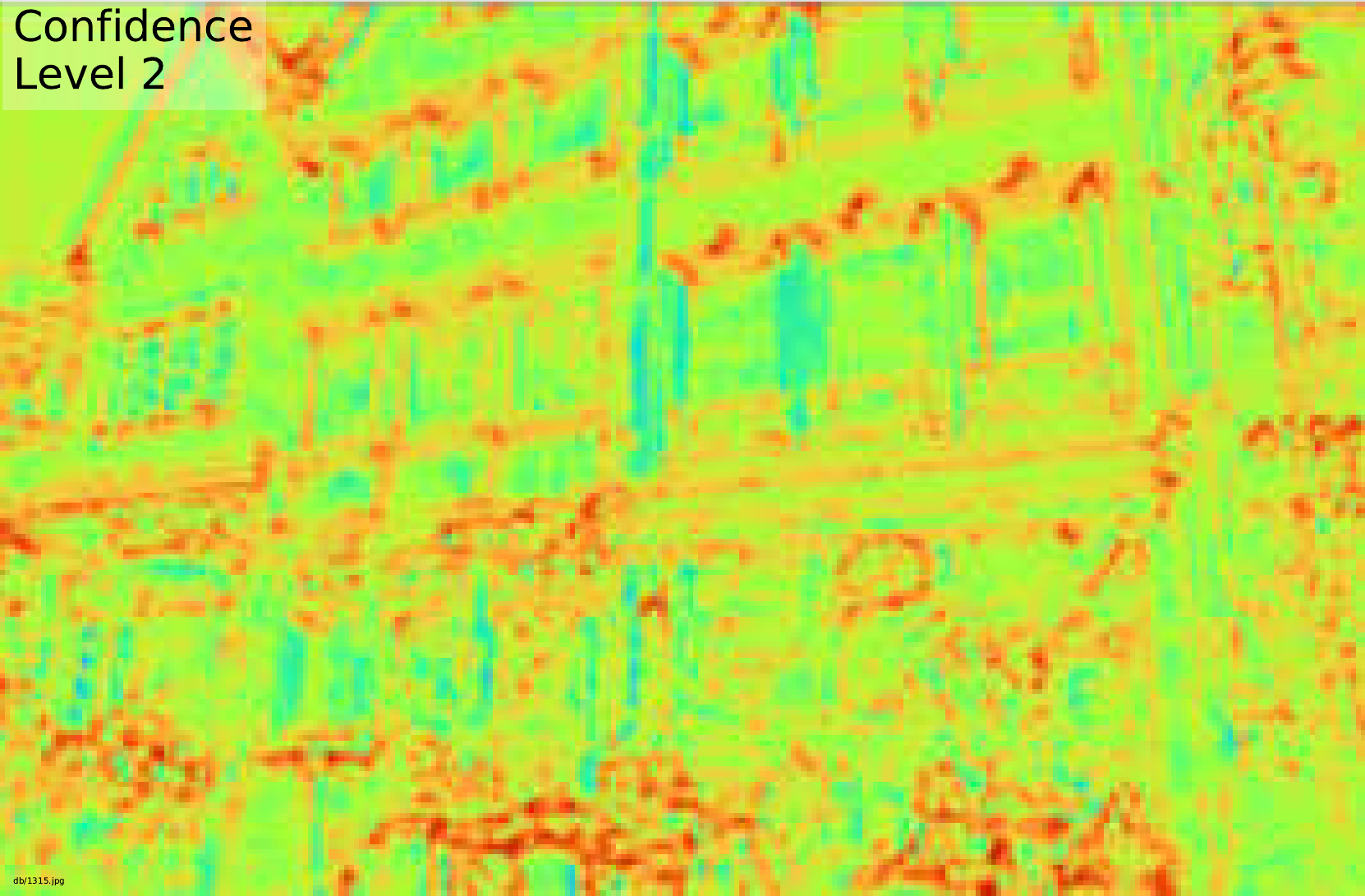}
\end{minipage}%
\begin{minipage}{\iwidth\textwidth}
    \centering
    \includegraphics[width=\pwidth\linewidth]{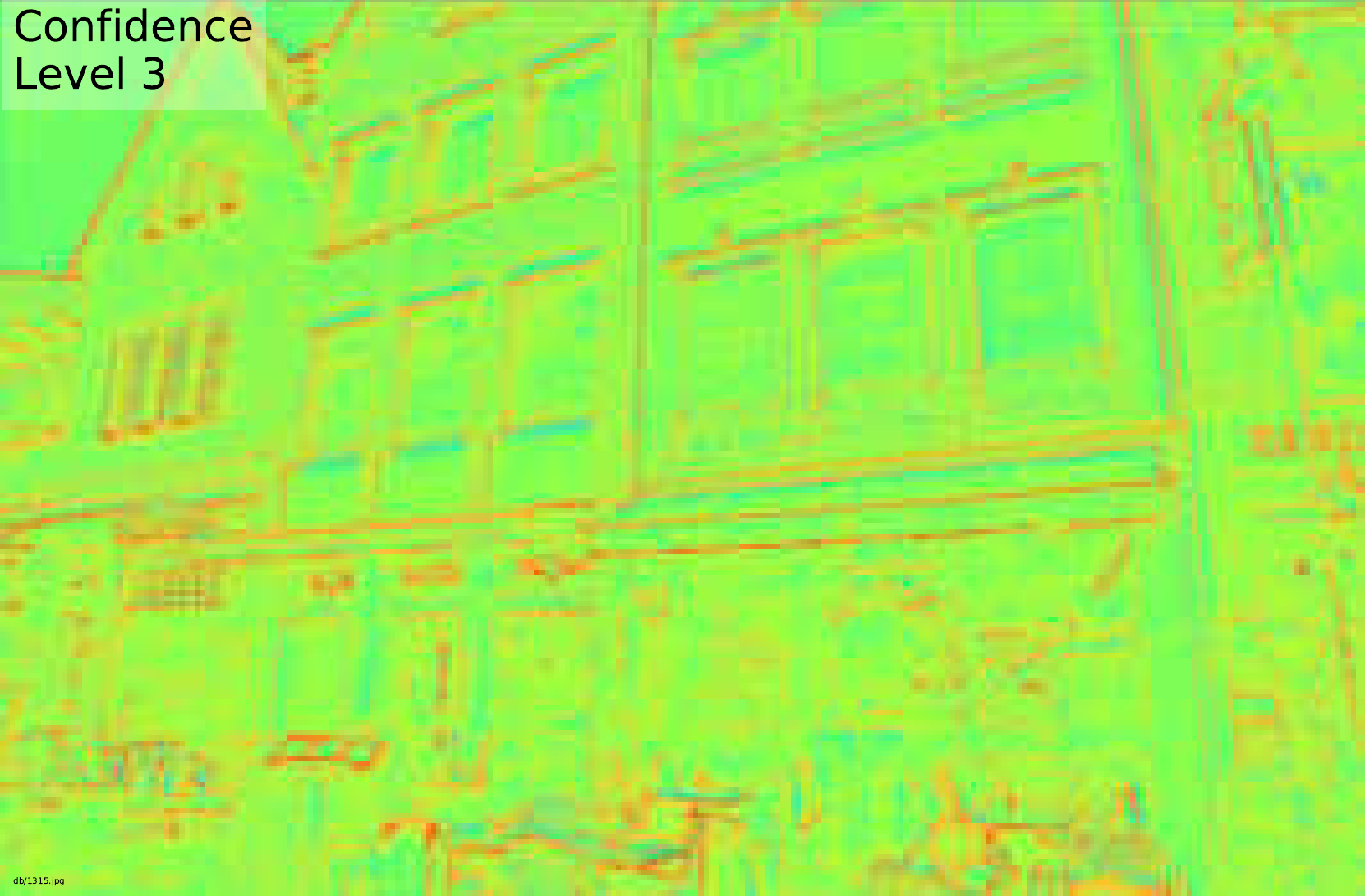}
\end{minipage}
\vspace{2mm}

    \caption{\textbf{Successful localization on the Aachen dataset.}
    We show 5 challenging queries with large initial errors and large day-night appearance changes that are successfully localized by PixLoc.
    The reprojection and pose errors are computed with respect to the pose estimated by hloc.
    }
    \label{fig:qualitative_aachen_success}%
\end{figure*}

\begin{figure*}[t]
    \centering
\def\iwidth{0.14}
\def\pwidth{0.99}
\def\lwidth{0.020}
\def\rcwidth{0.14}

\begin{minipage}{\lwidth\textwidth}
\hfill
\end{minipage}%
\begin{minipage}{\iwidth\textwidth}
    \centering
    \small{Images}
\end{minipage}%
\begin{minipage}{\iwidth\textwidth*\real{3.0}}
    \centering
    \small{Features}
    \vspace{0.5mm}
    \hrule width 0.99\linewidth
    \vspace{0.2mm}
\end{minipage}%
\begin{minipage}{\iwidth\textwidth*\real{3.0}}
    \centering
    \small{Confidence}
    \vspace{0.5mm}
    \hrule width 0.99\linewidth
    \vspace{0.2mm}
\end{minipage}%

\begin{minipage}{\lwidth\textwidth}
\rotatebox[origin=c]{90}{Query}
\end{minipage}%
\begin{minipage}{\iwidth\textwidth}
    \centering
    \includegraphics[width=\pwidth\linewidth]{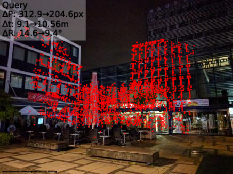}
\end{minipage}%
\begin{minipage}{\iwidth\textwidth}
    \centering
    \includegraphics[width=\pwidth\linewidth]{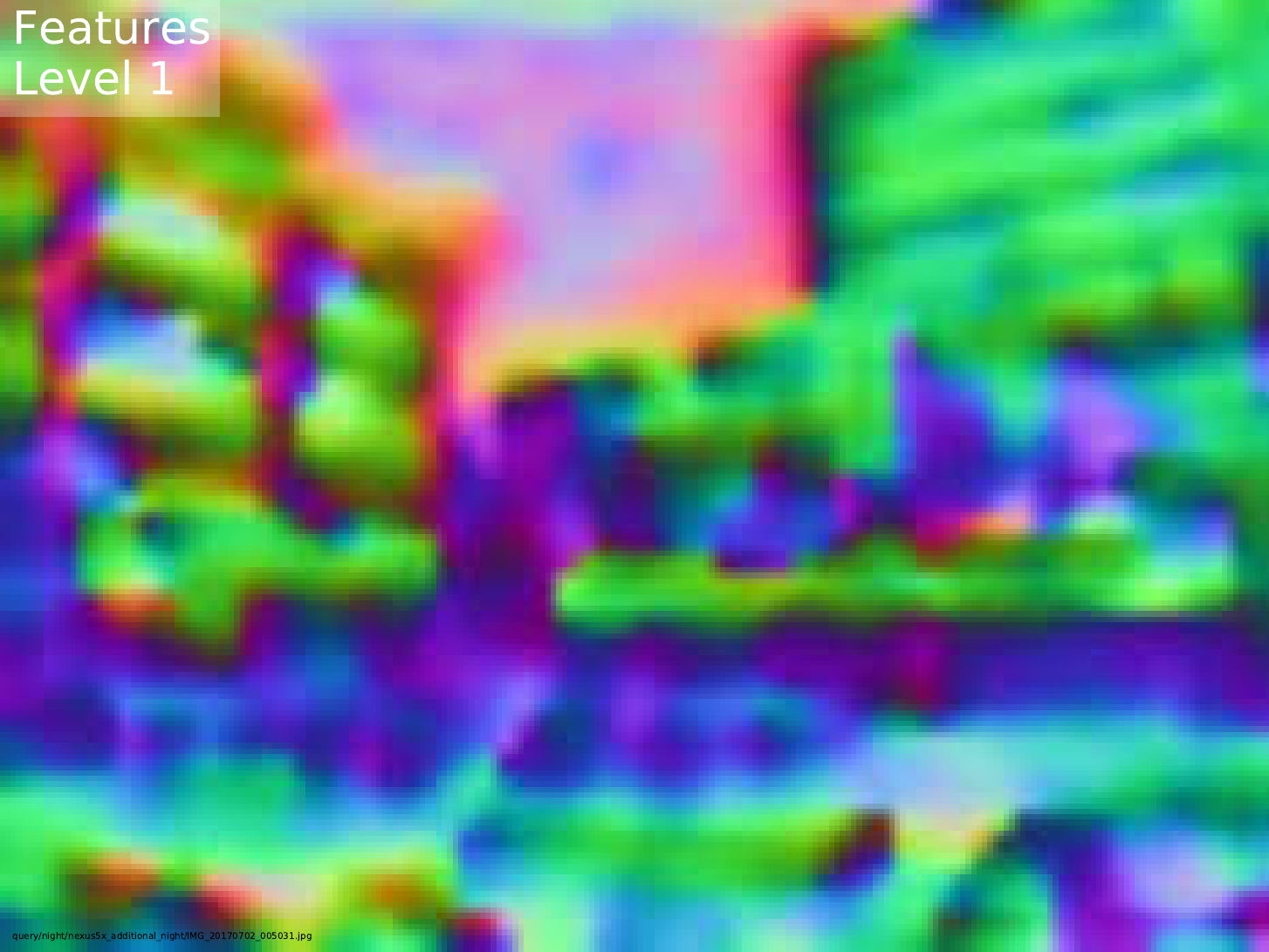}
\end{minipage}%
\begin{minipage}{\iwidth\textwidth}
    \centering
    \includegraphics[width=\pwidth\linewidth]{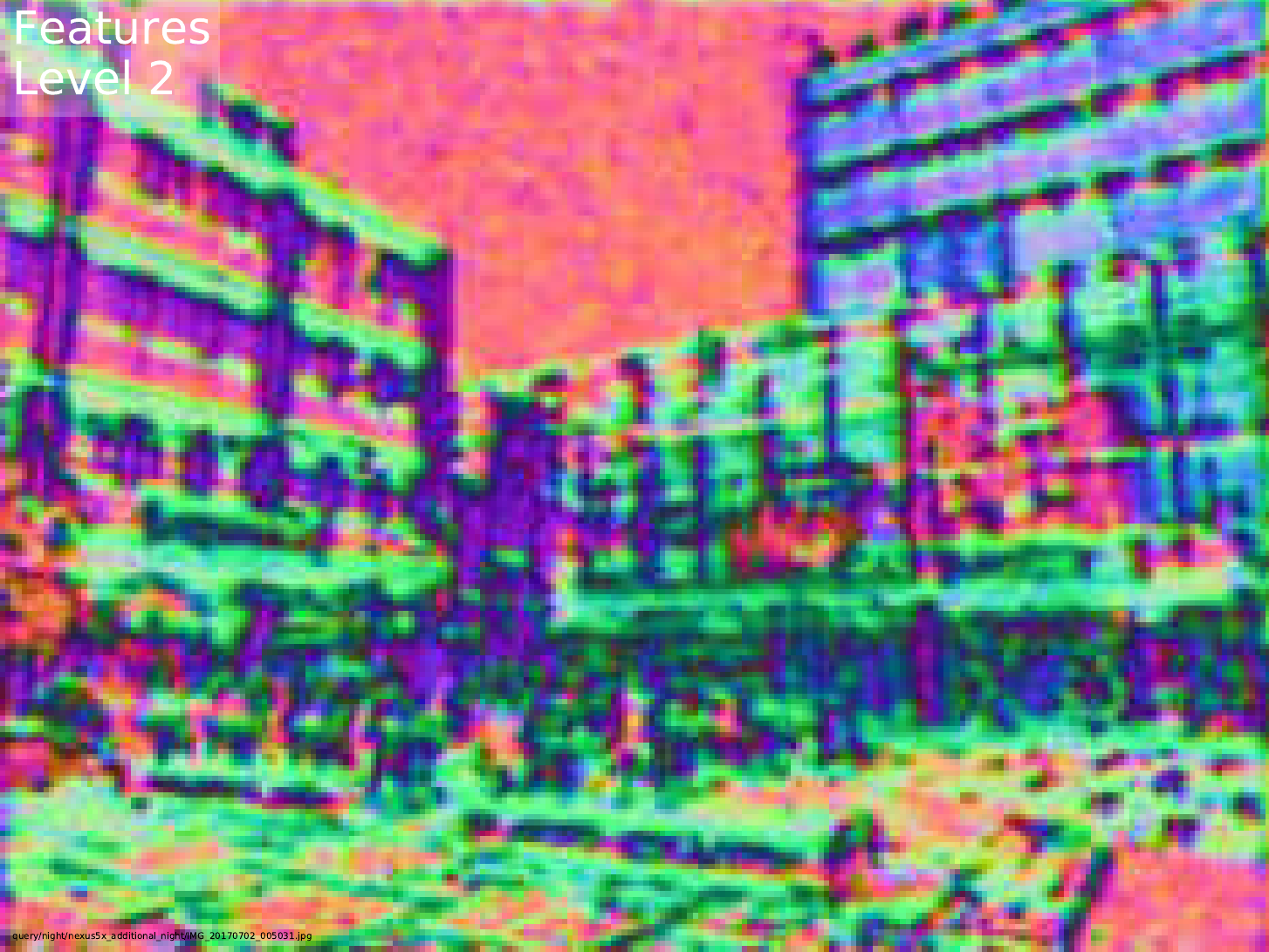}
\end{minipage}%
\begin{minipage}{\iwidth\textwidth}
    \centering
    \includegraphics[width=\pwidth\linewidth]{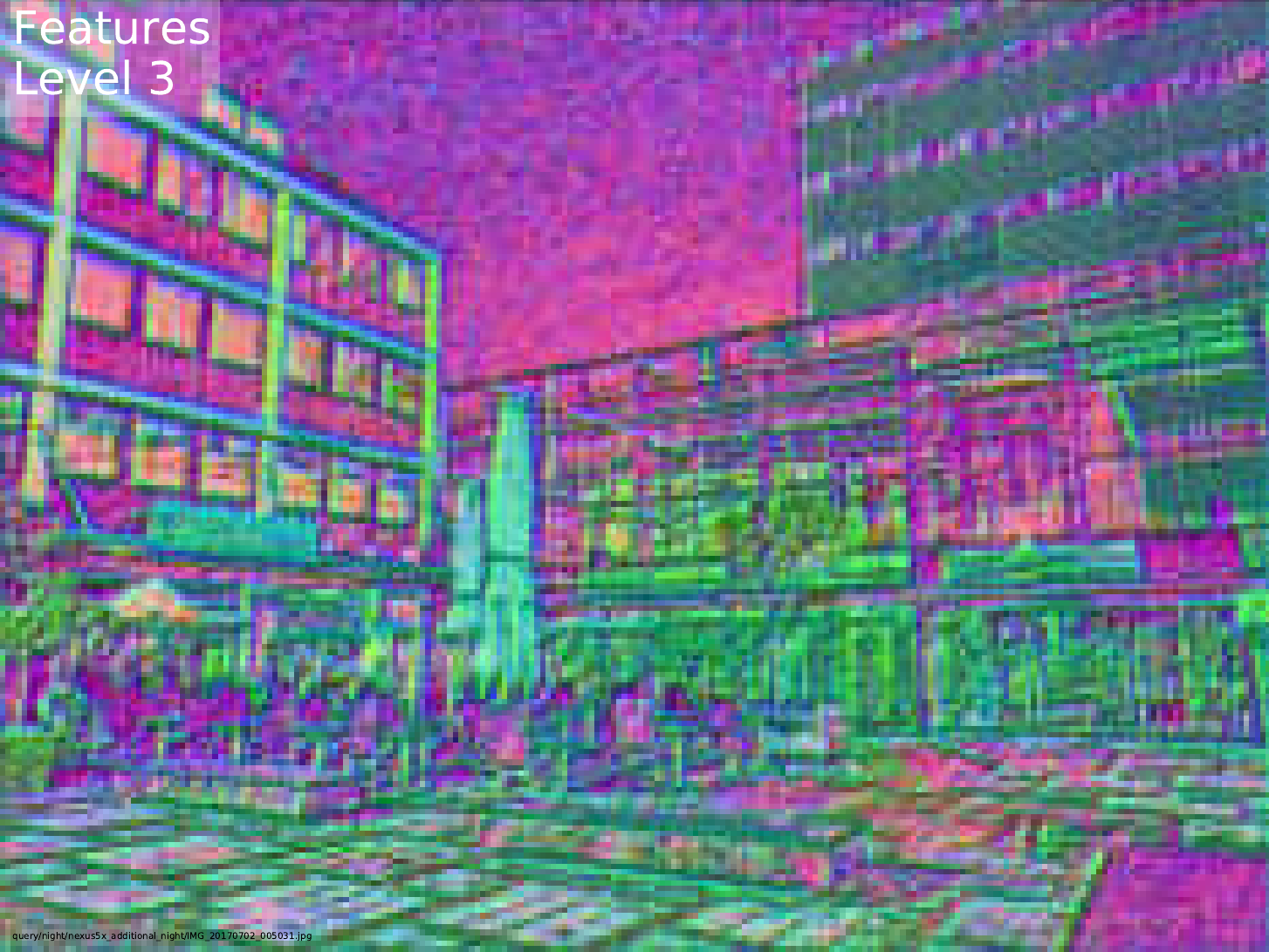}
\end{minipage}%
\begin{minipage}{\iwidth\textwidth}
    \centering
    \includegraphics[width=\pwidth\linewidth]{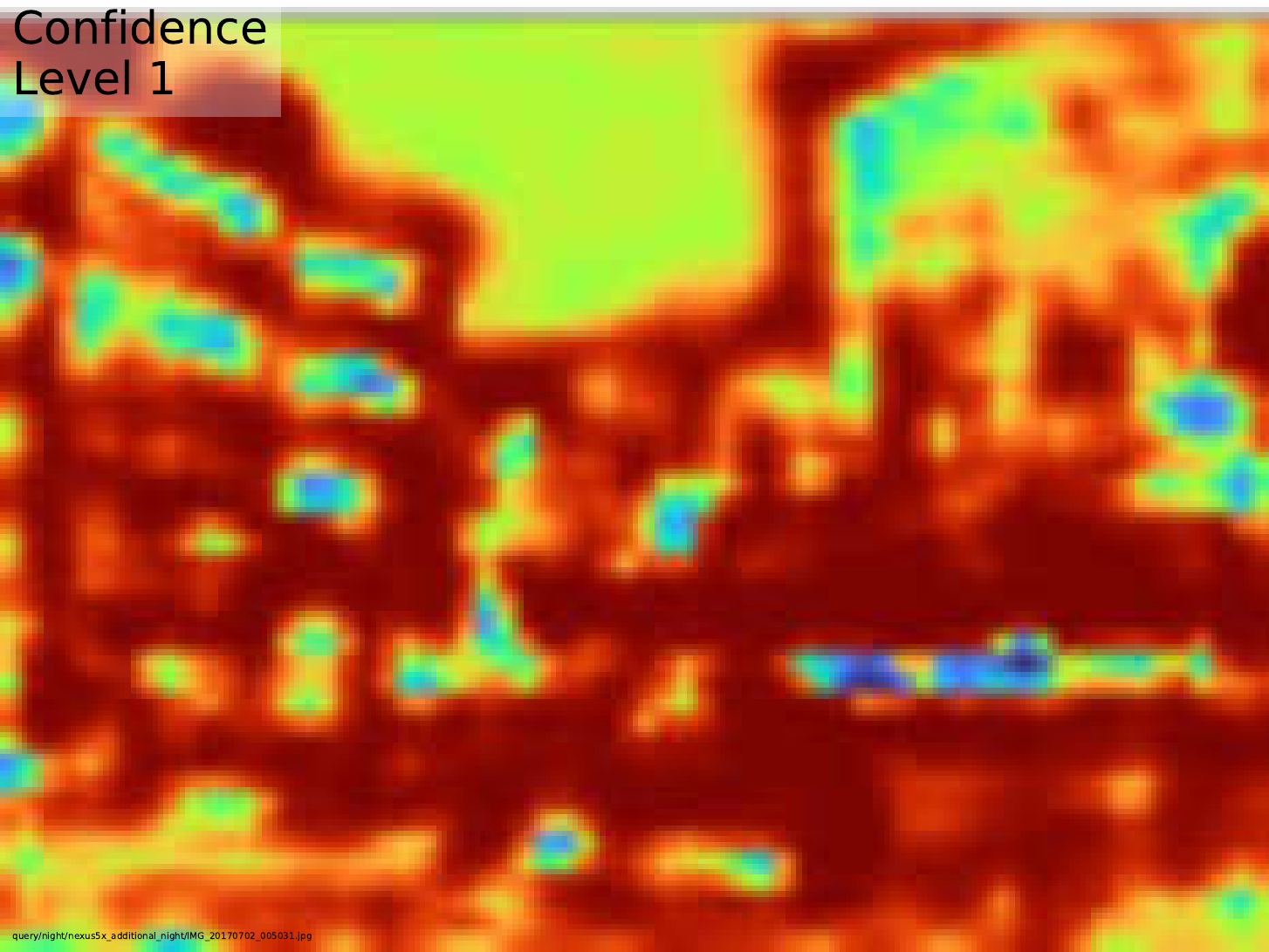}
\end{minipage}%
\begin{minipage}{\iwidth\textwidth}
    \centering
    \includegraphics[width=\pwidth\linewidth]{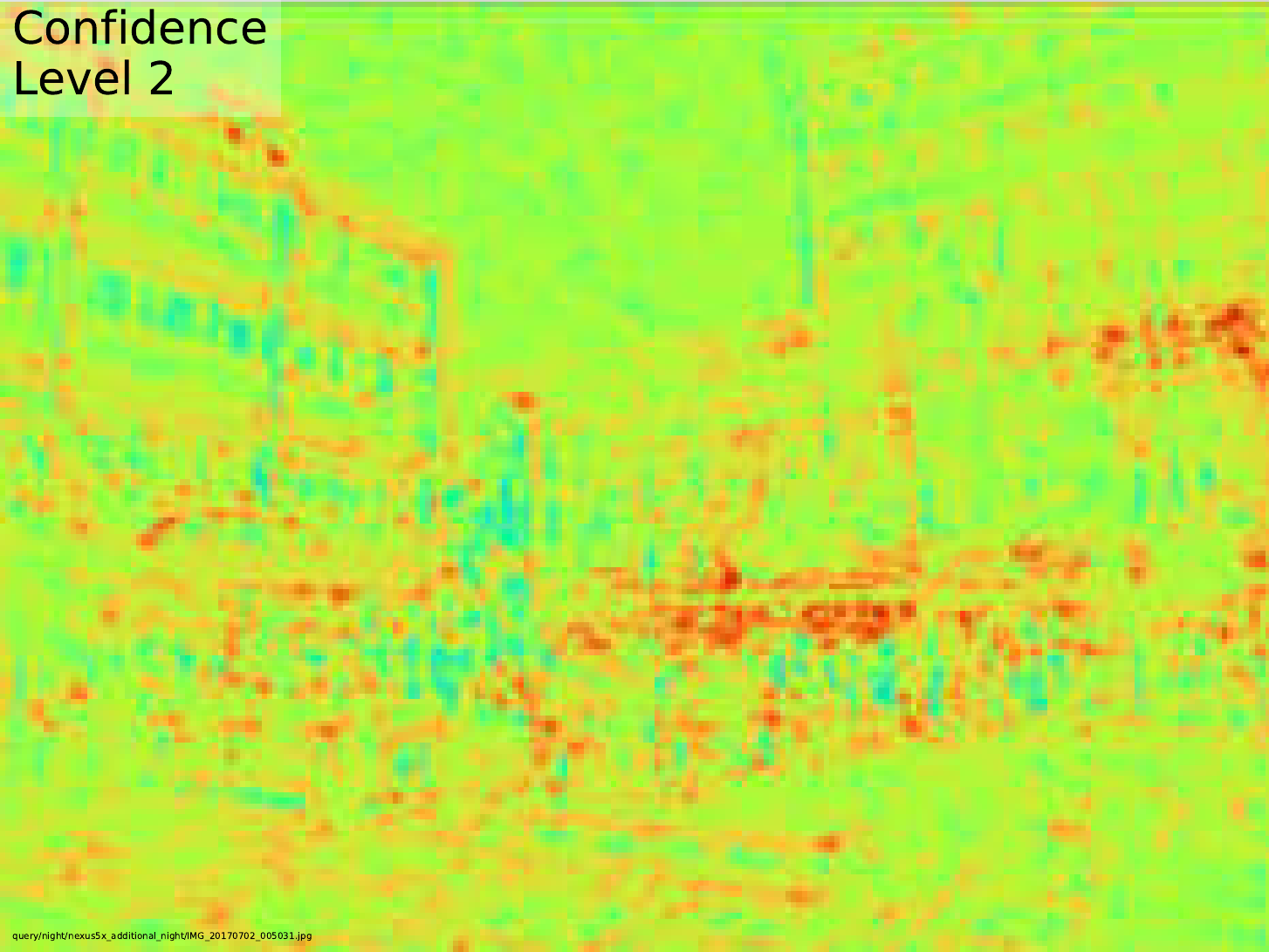}
\end{minipage}%
\begin{minipage}{\iwidth\textwidth}
    \centering
    \includegraphics[width=\pwidth\linewidth]{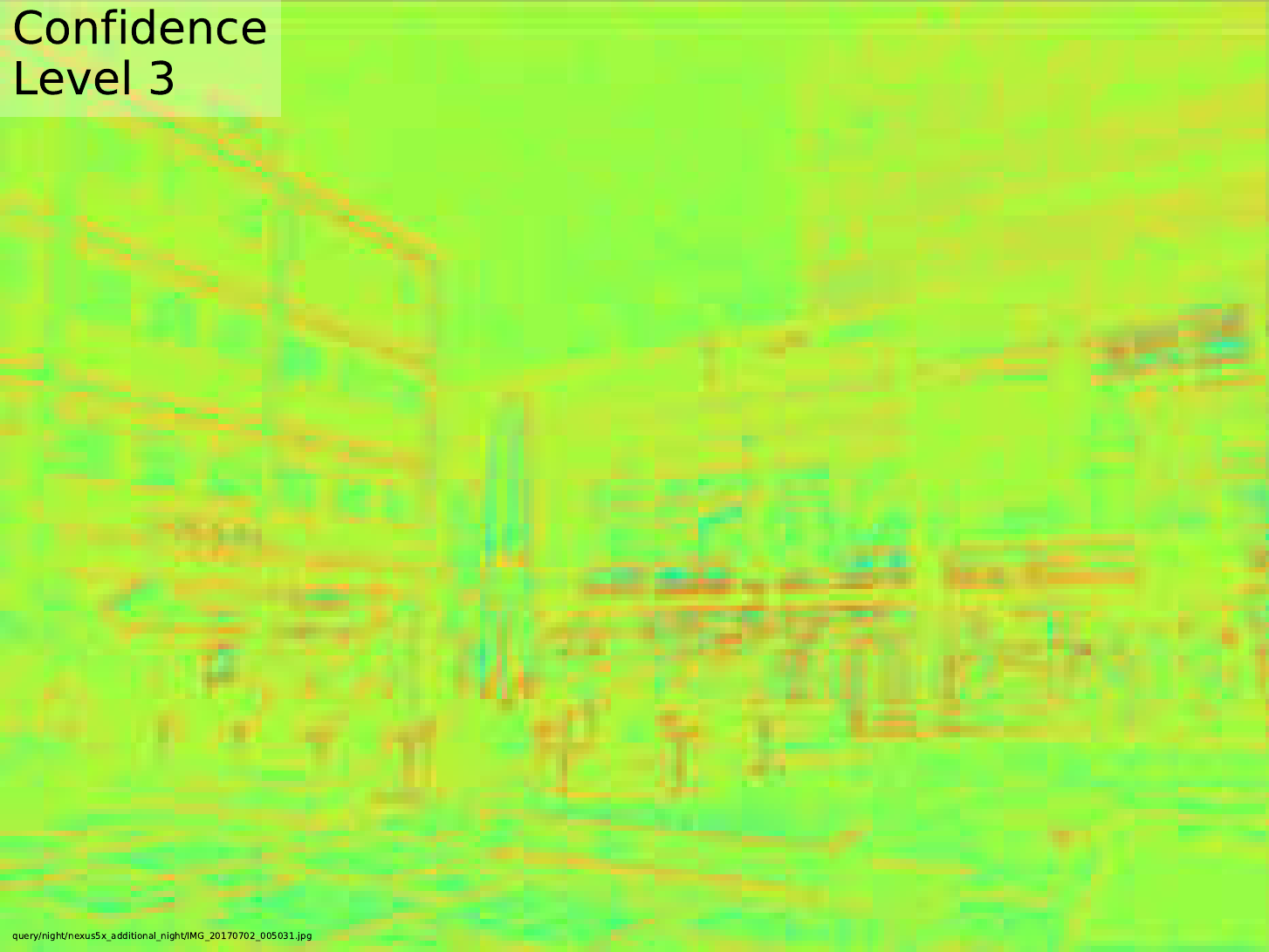}
\end{minipage}
\begin{minipage}{\lwidth\textwidth}
\rotatebox[origin=c]{90}{Reference}
\end{minipage}%
\begin{minipage}{\iwidth\textwidth}
    \centering
    \includegraphics[width=\pwidth\linewidth]{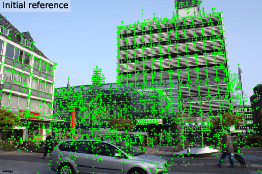}
\end{minipage}%
\begin{minipage}{\iwidth\textwidth}
    \centering
    \includegraphics[width=\pwidth\linewidth]{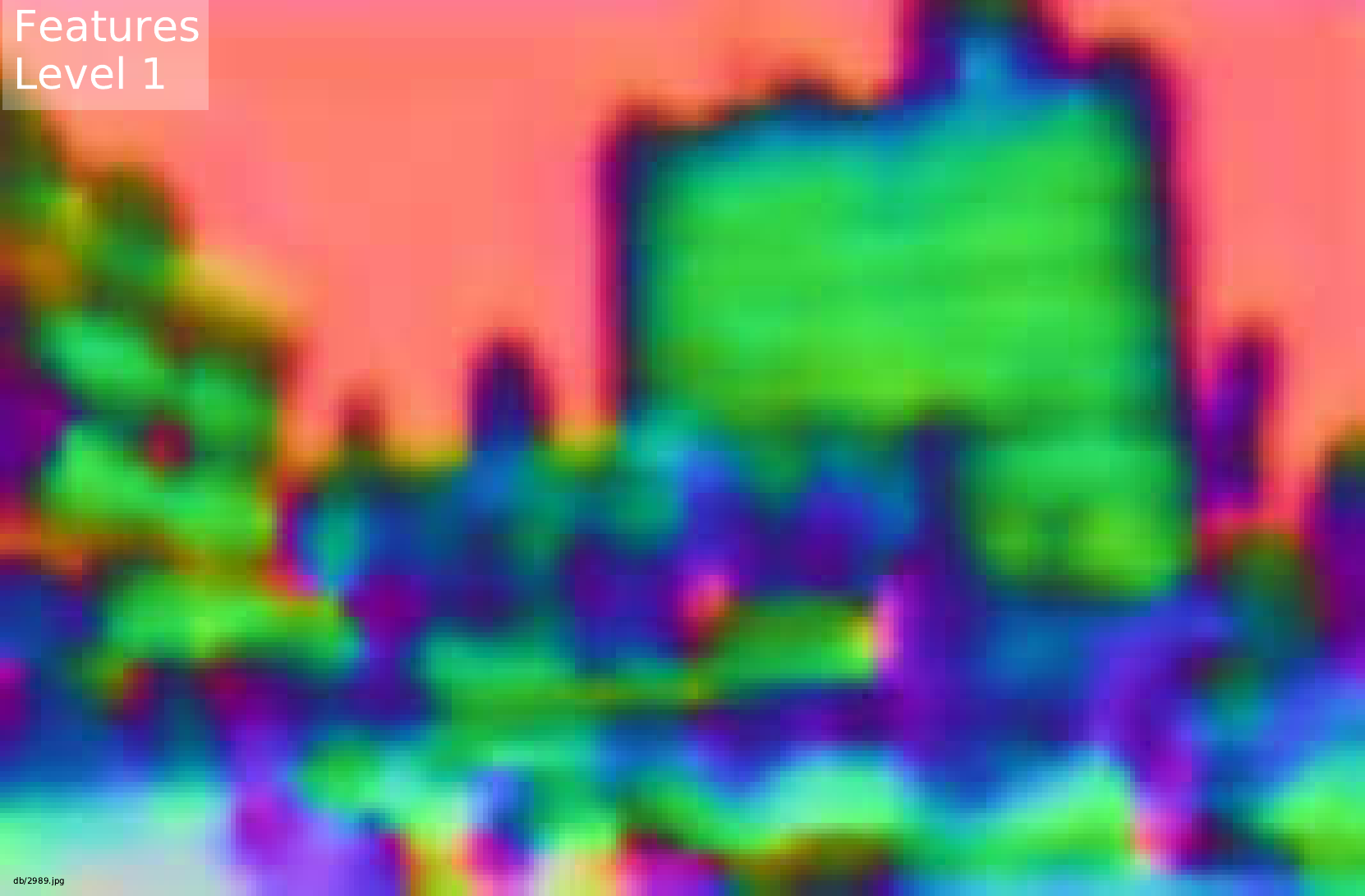}
\end{minipage}%
\begin{minipage}{\iwidth\textwidth}
    \centering
    \includegraphics[width=\pwidth\linewidth]{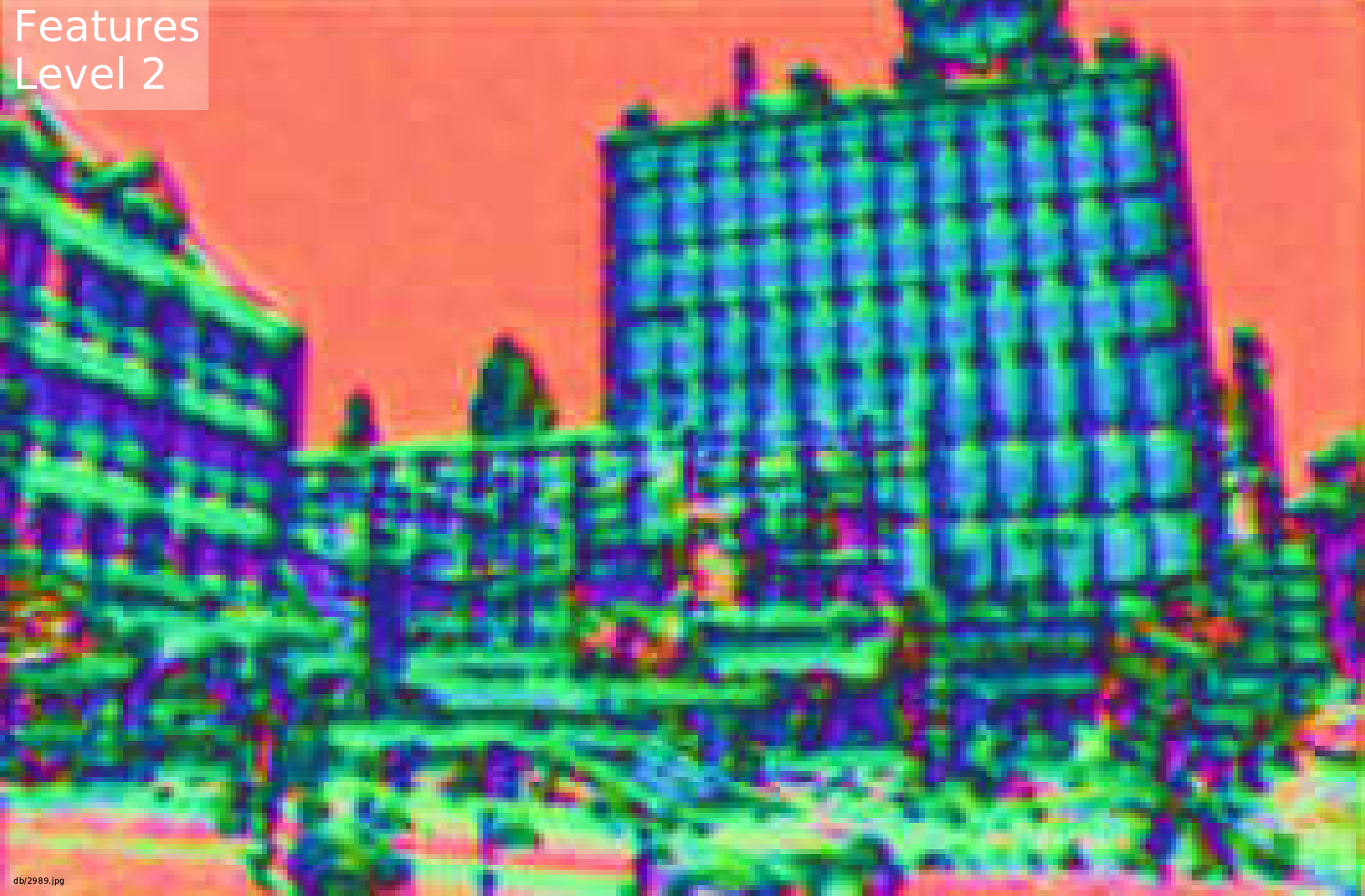}
\end{minipage}%
\begin{minipage}{\iwidth\textwidth}
    \centering
    \includegraphics[width=\pwidth\linewidth]{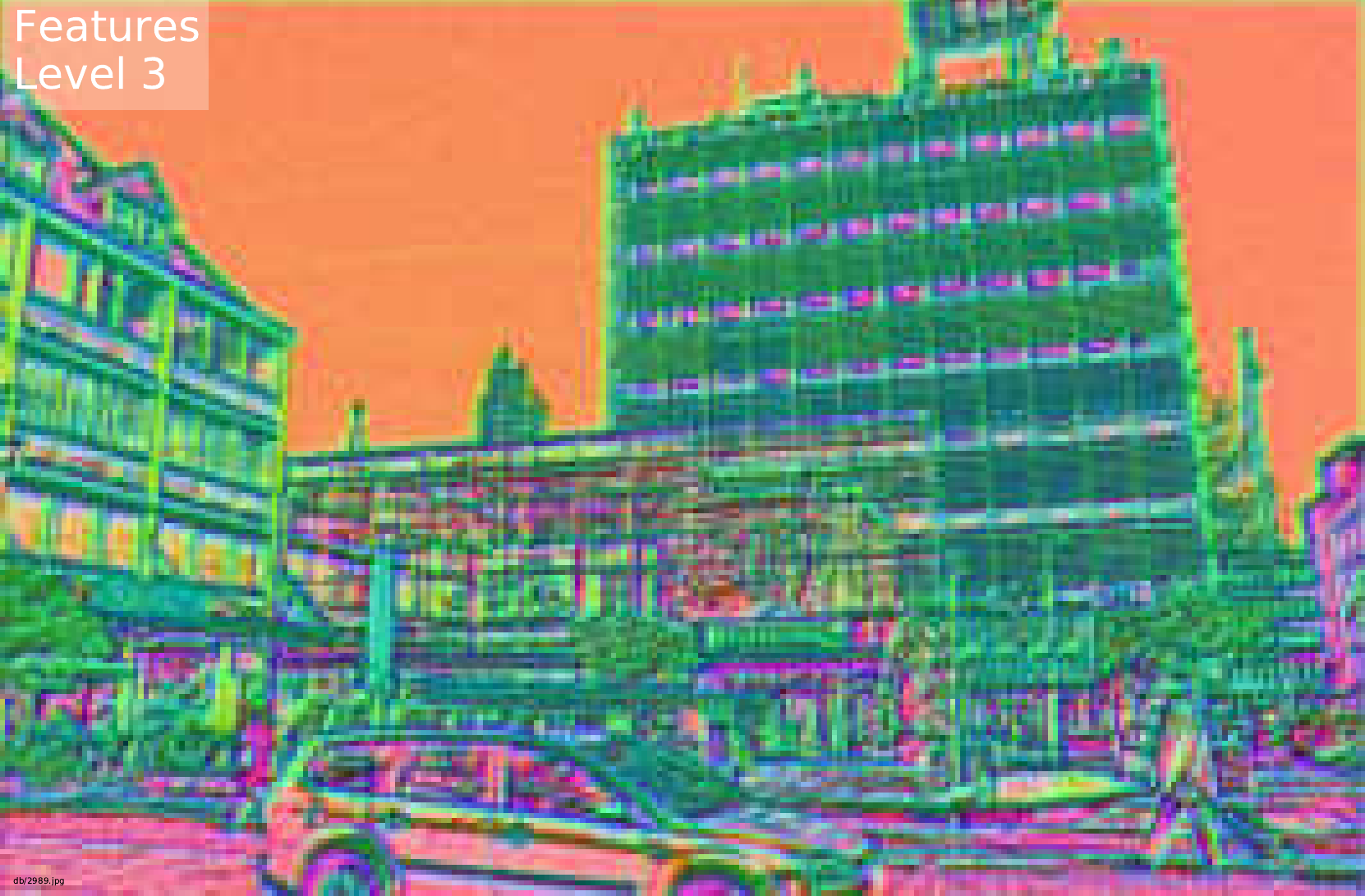}
\end{minipage}%
\begin{minipage}{\iwidth\textwidth}
    \centering
    \includegraphics[width=\pwidth\linewidth]{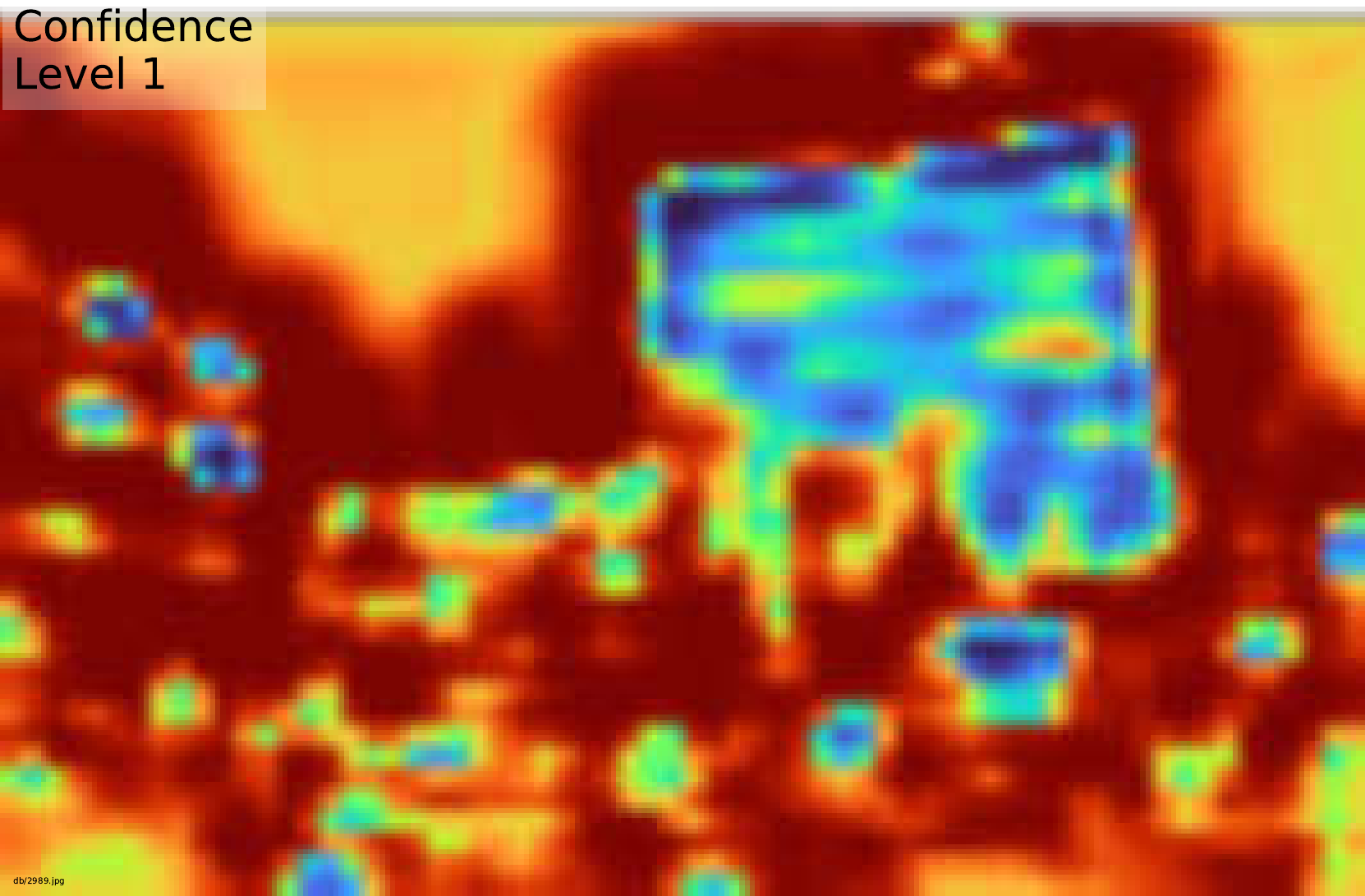}
\end{minipage}%
\begin{minipage}{\iwidth\textwidth}
    \centering
    \includegraphics[width=\pwidth\linewidth]{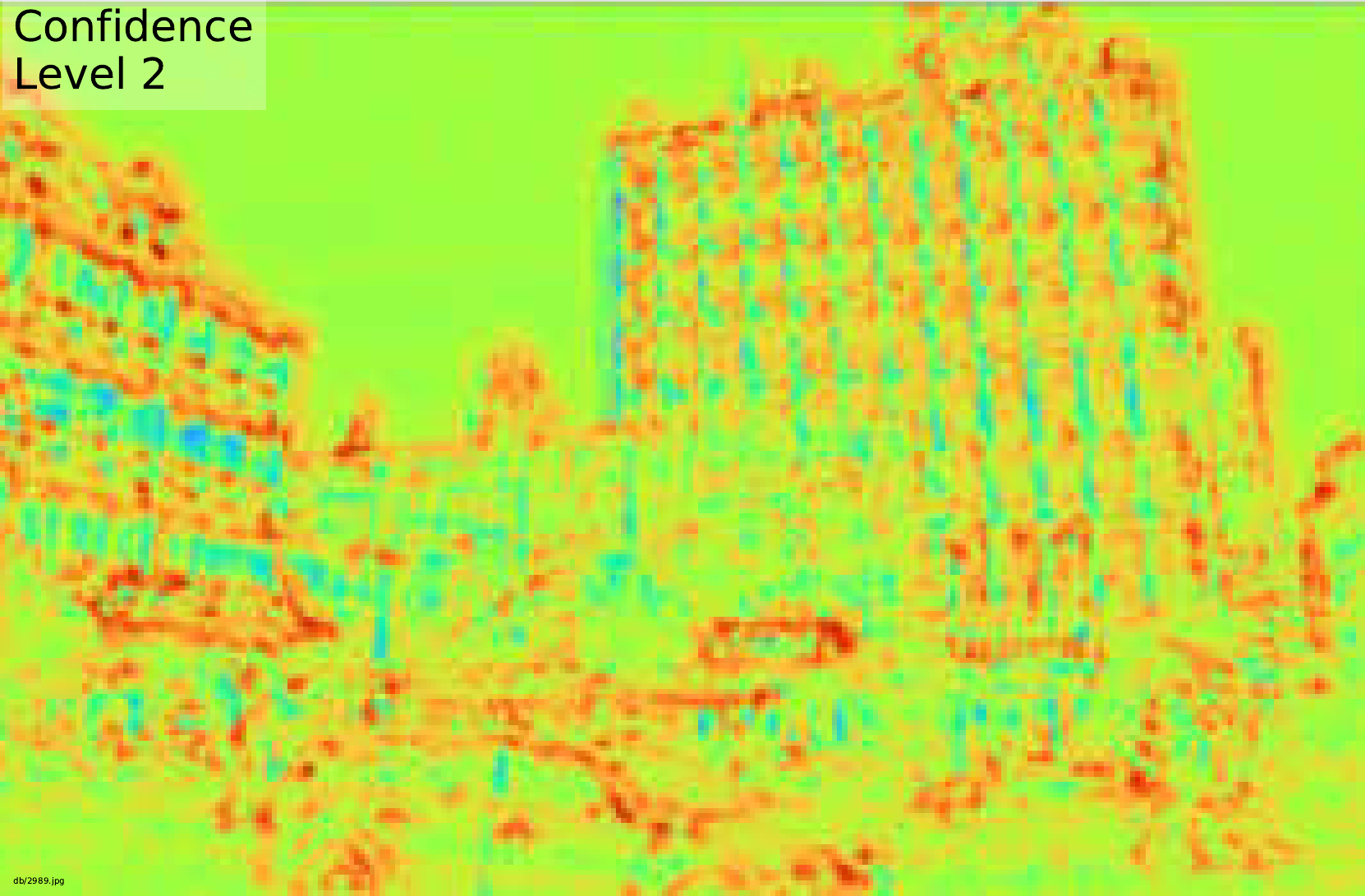}
\end{minipage}%
\begin{minipage}{\iwidth\textwidth}
    \centering
    \includegraphics[width=\pwidth\linewidth]{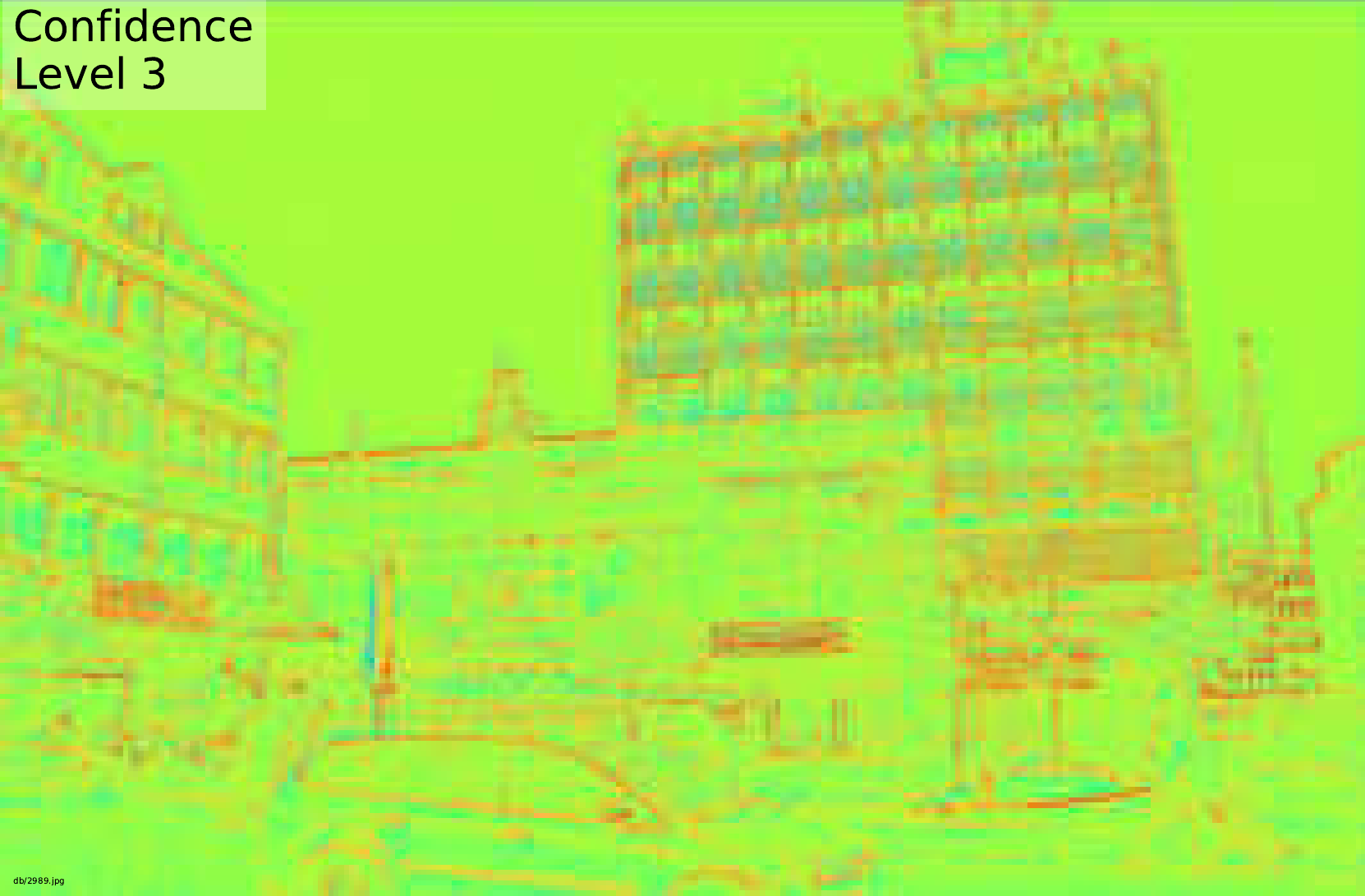}
\end{minipage}
\vspace{2mm}

\begin{minipage}{\lwidth\textwidth}
\rotatebox[origin=c]{90}{Query}
\end{minipage}%
\begin{minipage}{\iwidth\textwidth}
    \centering
    \includegraphics[width=\pwidth\linewidth]{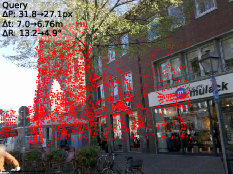}
\end{minipage}%
\begin{minipage}{\iwidth\textwidth}
    \centering
    \includegraphics[width=\pwidth\linewidth]{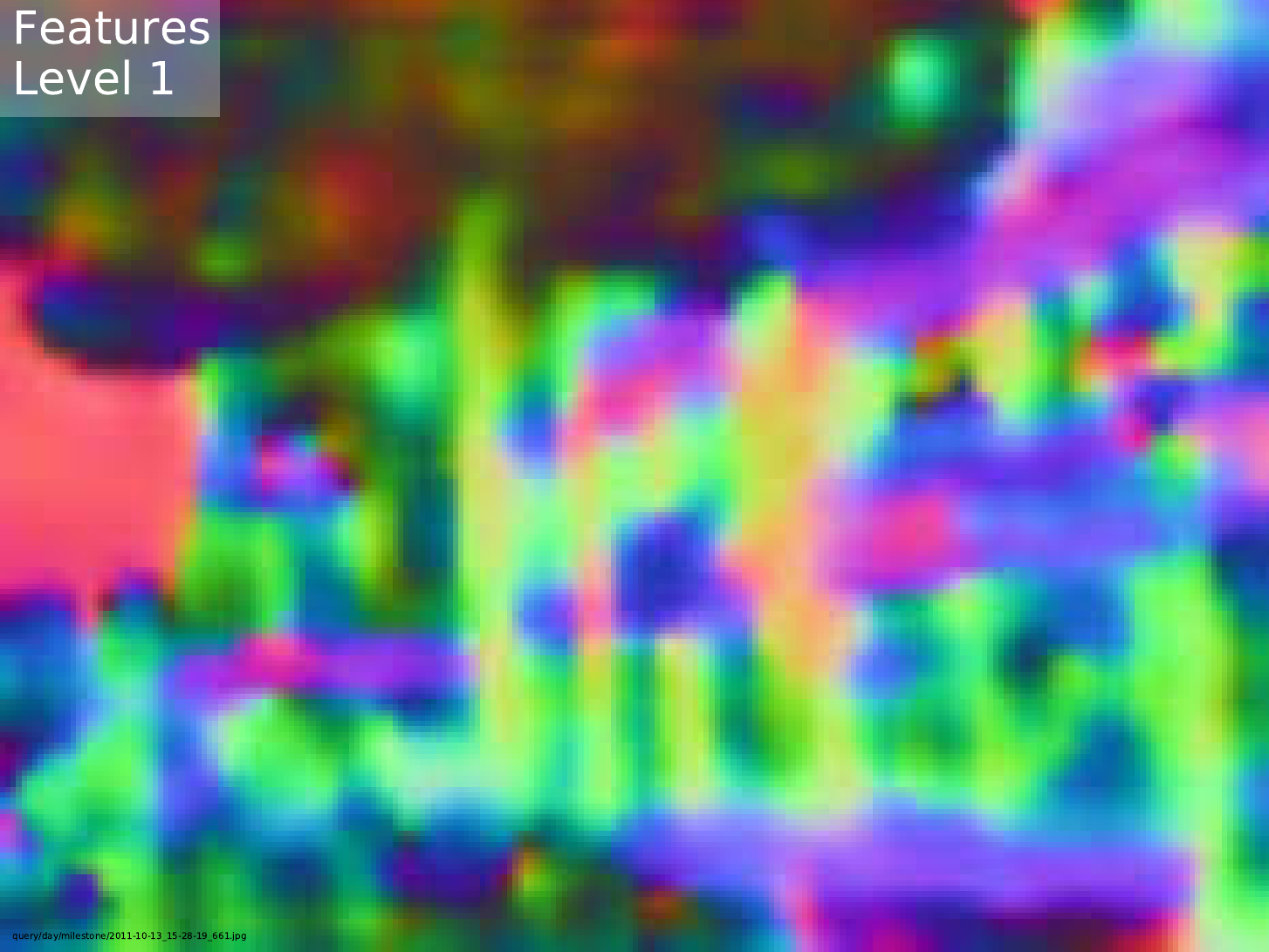}
\end{minipage}%
\begin{minipage}{\iwidth\textwidth}
    \centering
    \includegraphics[width=\pwidth\linewidth]{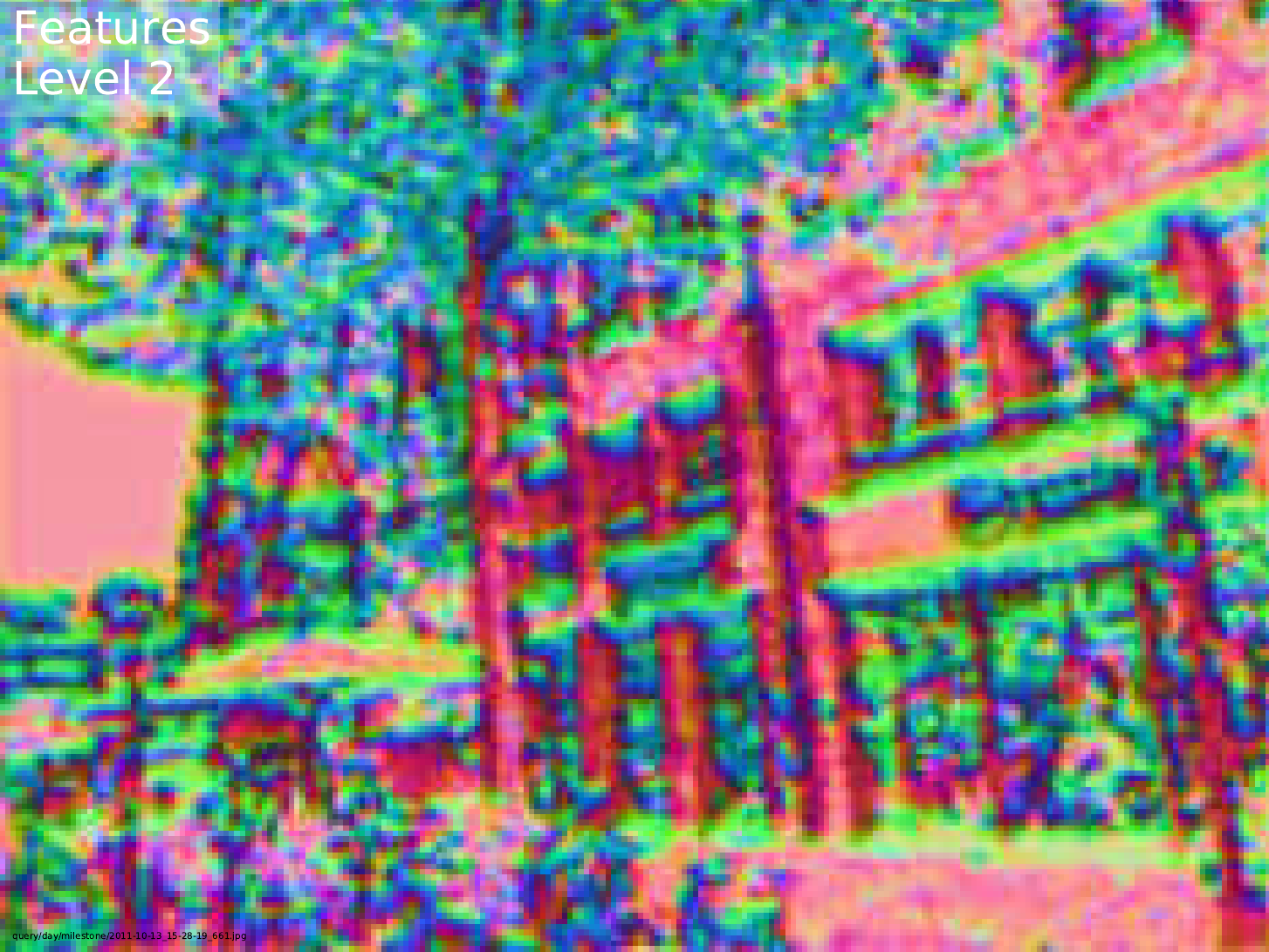}
\end{minipage}%
\begin{minipage}{\iwidth\textwidth}
    \centering
    \includegraphics[width=\pwidth\linewidth]{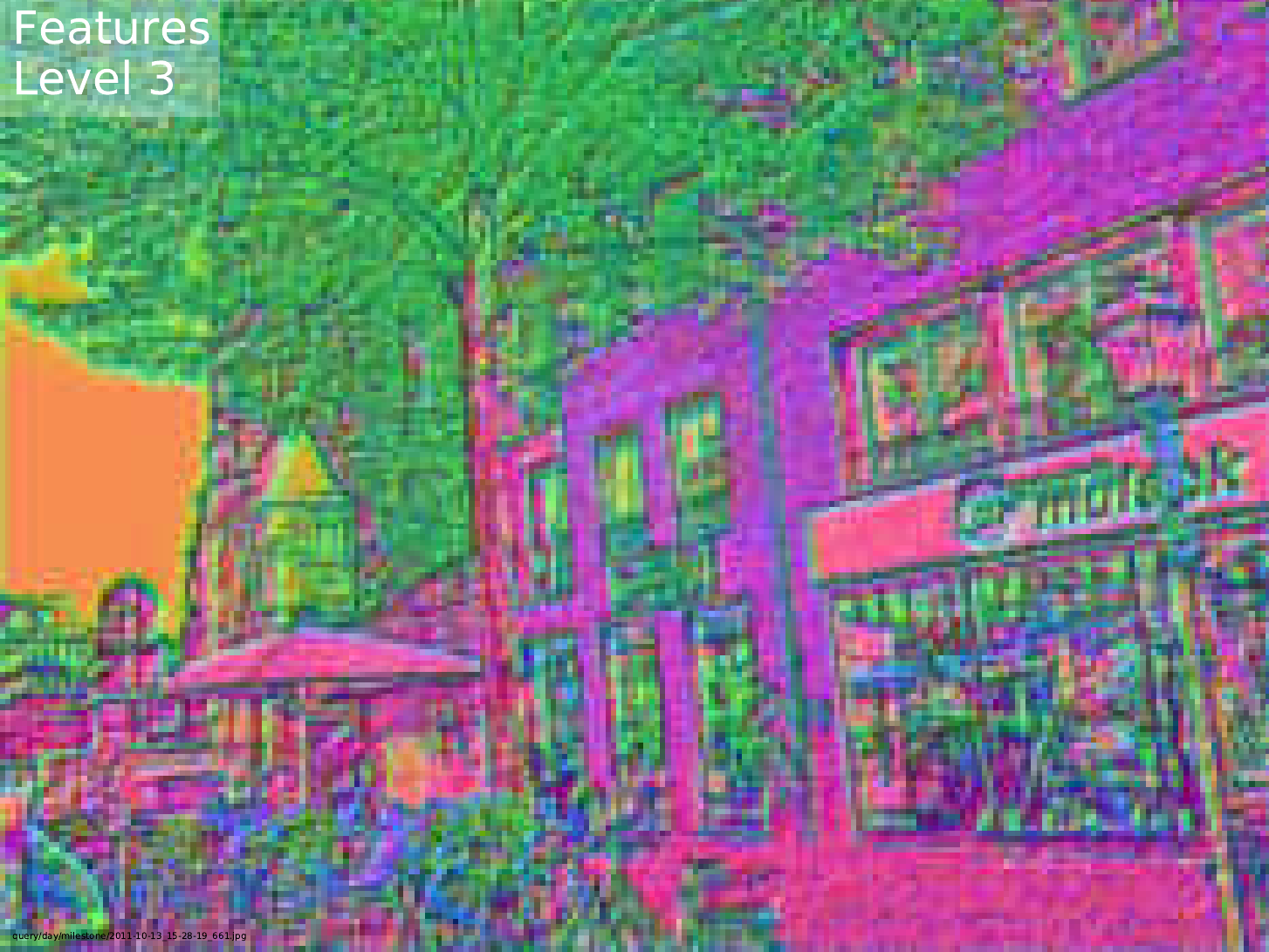}
\end{minipage}%
\begin{minipage}{\iwidth\textwidth}
    \centering
    \includegraphics[width=\pwidth\linewidth]{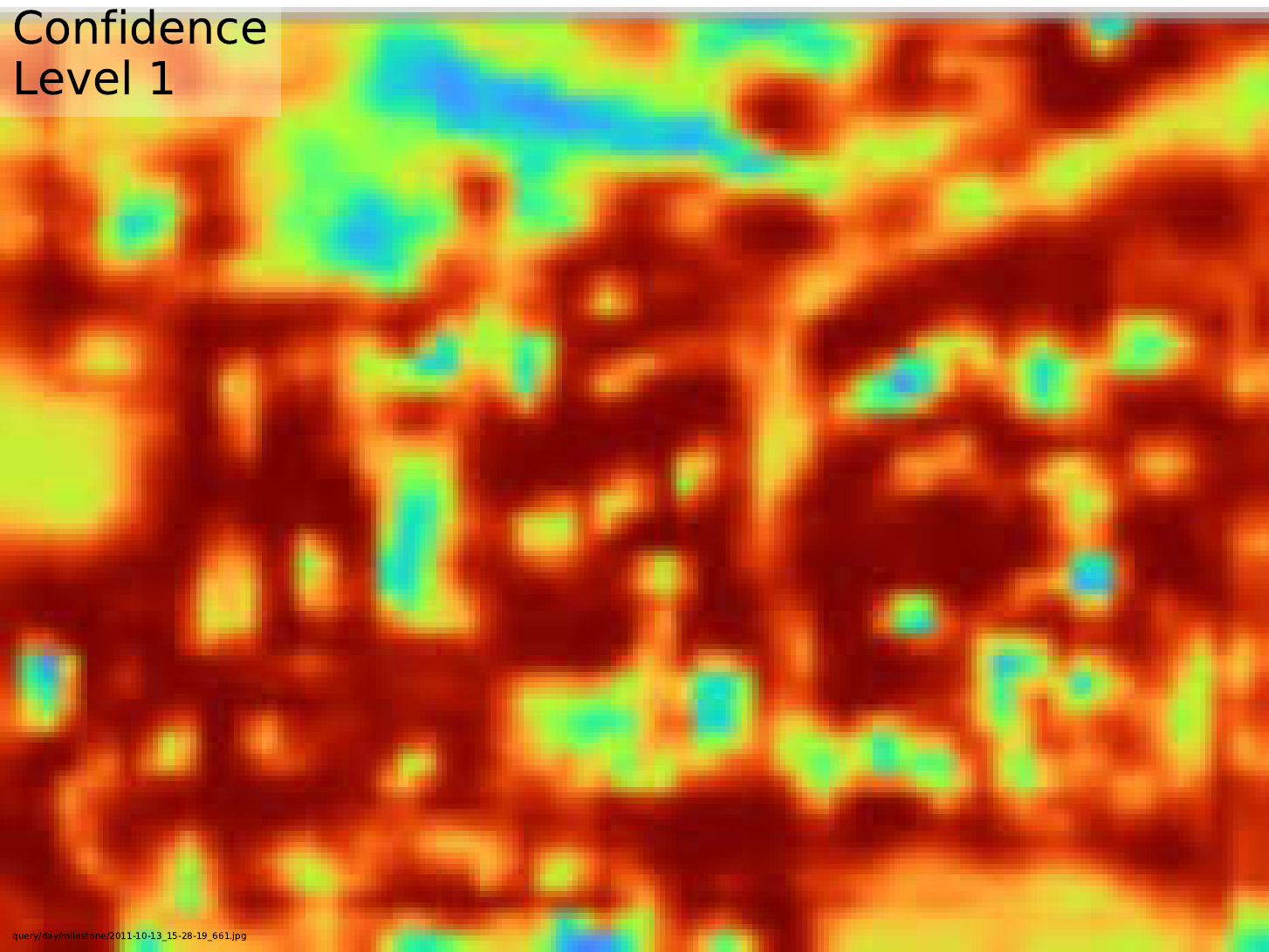}
\end{minipage}%
\begin{minipage}{\iwidth\textwidth}
    \centering
    \includegraphics[width=\pwidth\linewidth]{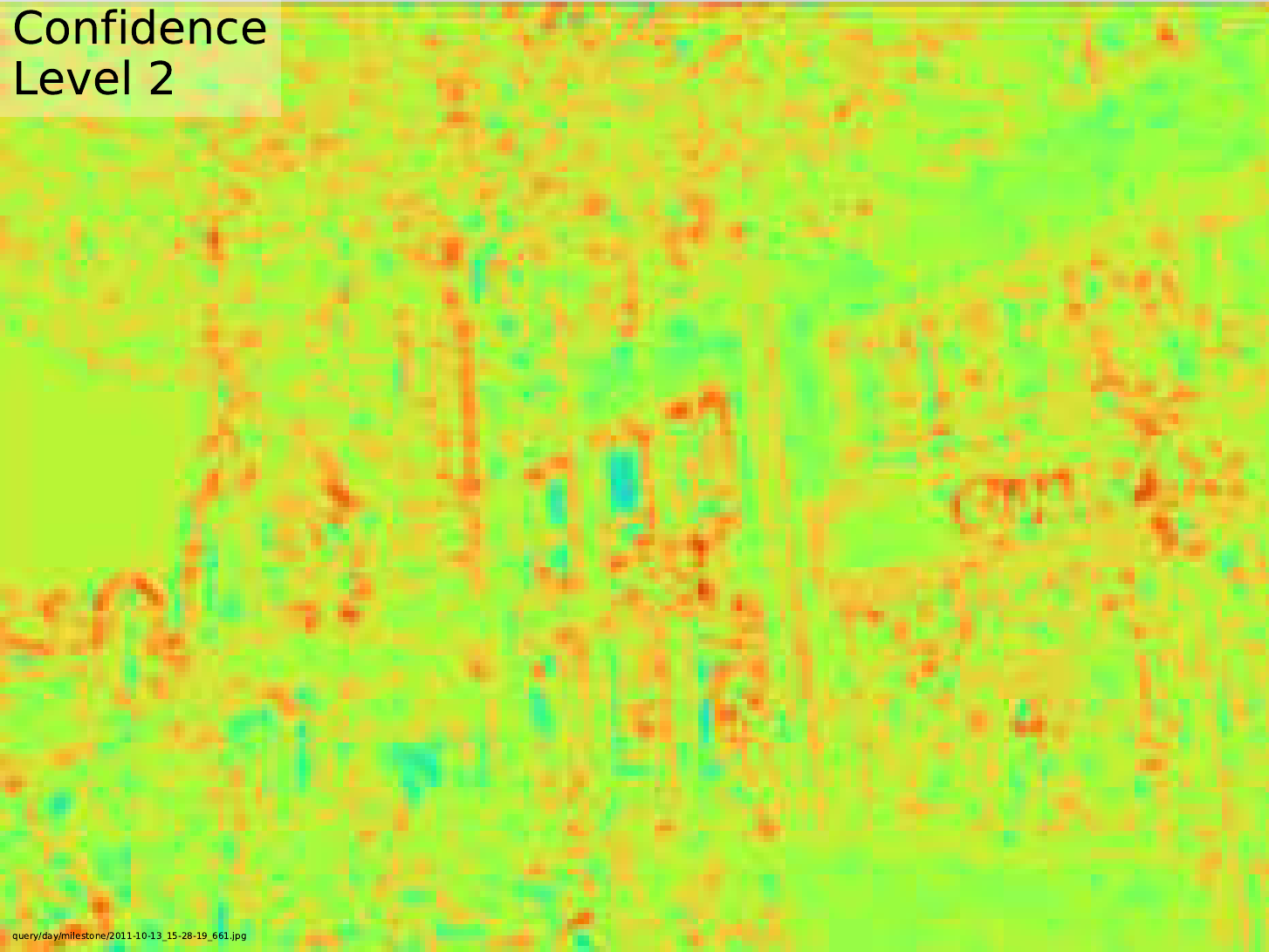}
\end{minipage}%
\begin{minipage}{\iwidth\textwidth}
    \centering
    \includegraphics[width=\pwidth\linewidth]{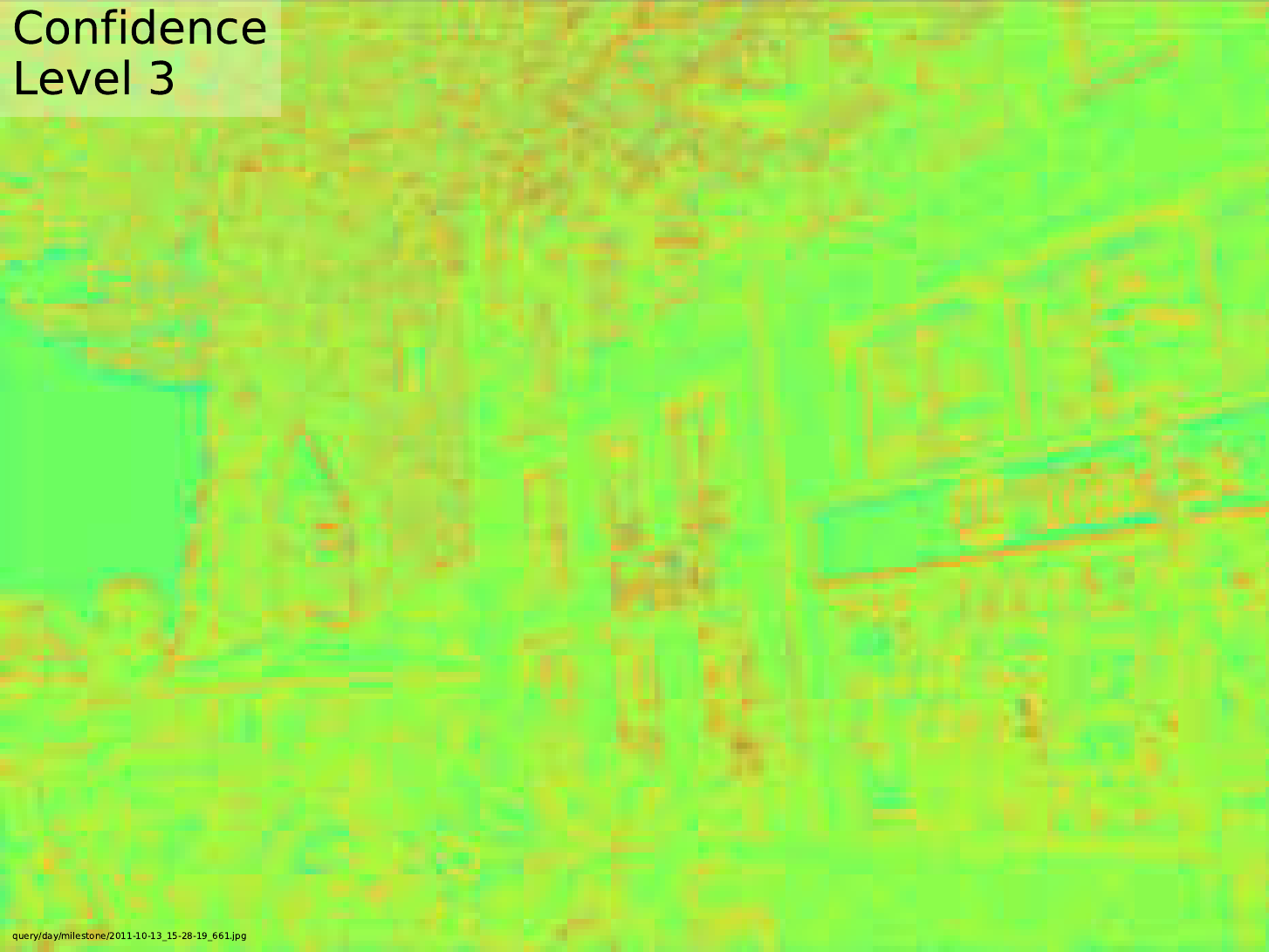}
\end{minipage}
\begin{minipage}{\lwidth\textwidth}
\rotatebox[origin=c]{90}{Reference}
\end{minipage}%
\begin{minipage}{\iwidth\textwidth}
    \centering
    \includegraphics[width=\pwidth\linewidth]{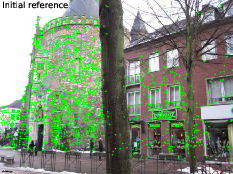}
\end{minipage}%
\begin{minipage}{\iwidth\textwidth}
    \centering
    \includegraphics[width=\pwidth\linewidth]{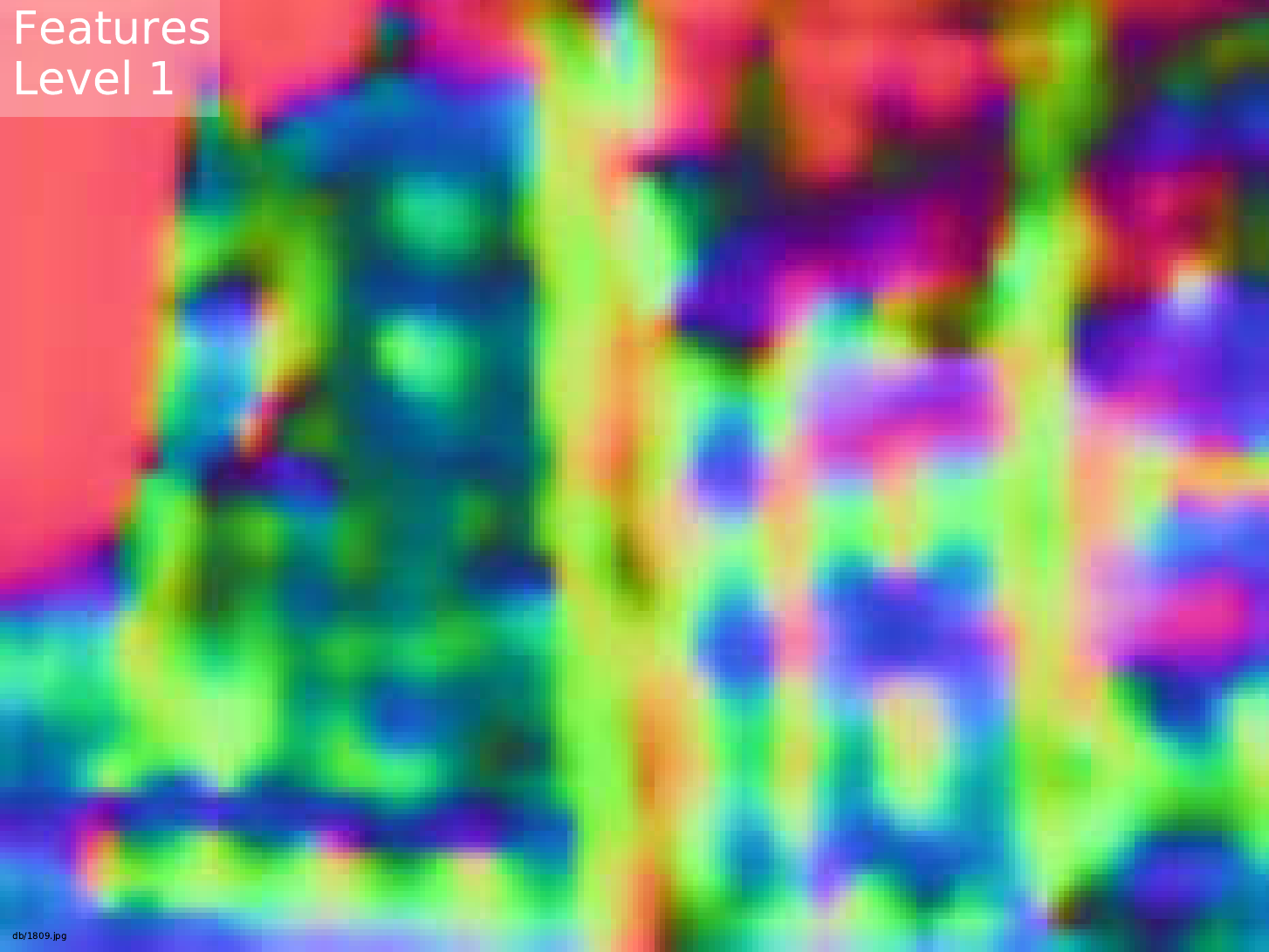}
\end{minipage}%
\begin{minipage}{\iwidth\textwidth}
    \centering
    \includegraphics[width=\pwidth\linewidth]{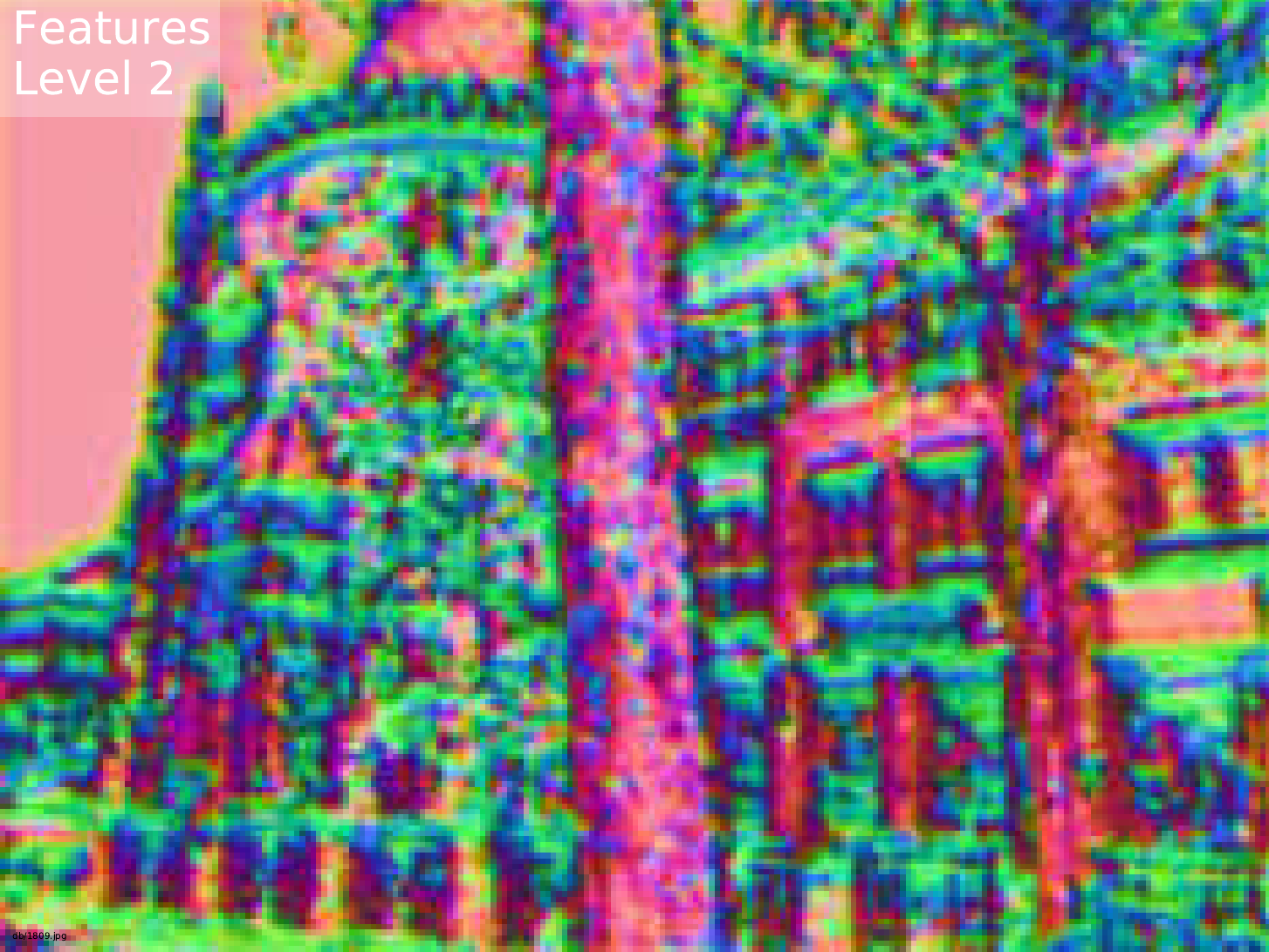}
\end{minipage}%
\begin{minipage}{\iwidth\textwidth}
    \centering
    \includegraphics[width=\pwidth\linewidth]{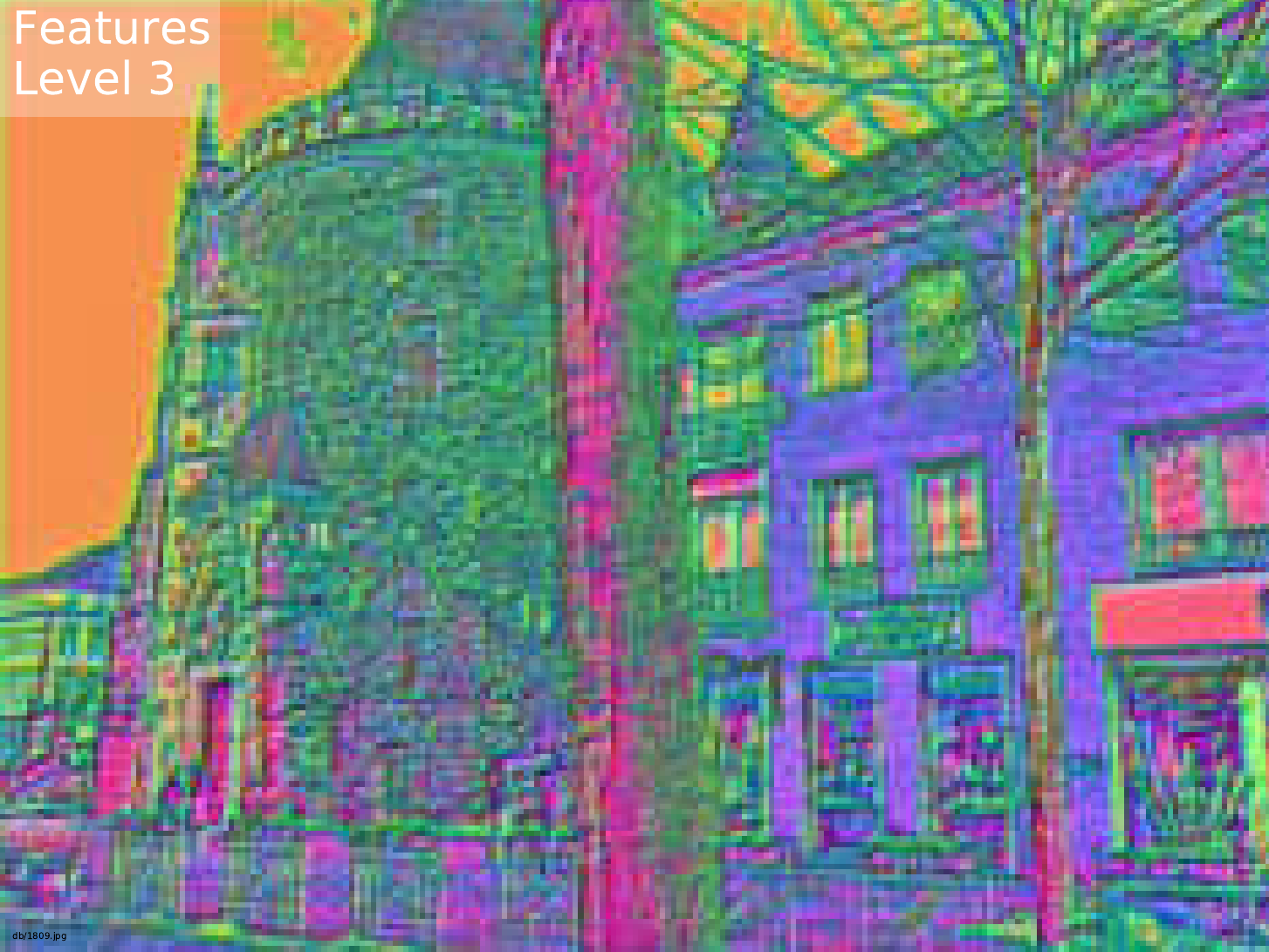}
\end{minipage}%
\begin{minipage}{\iwidth\textwidth}
    \centering
    \includegraphics[width=\pwidth\linewidth]{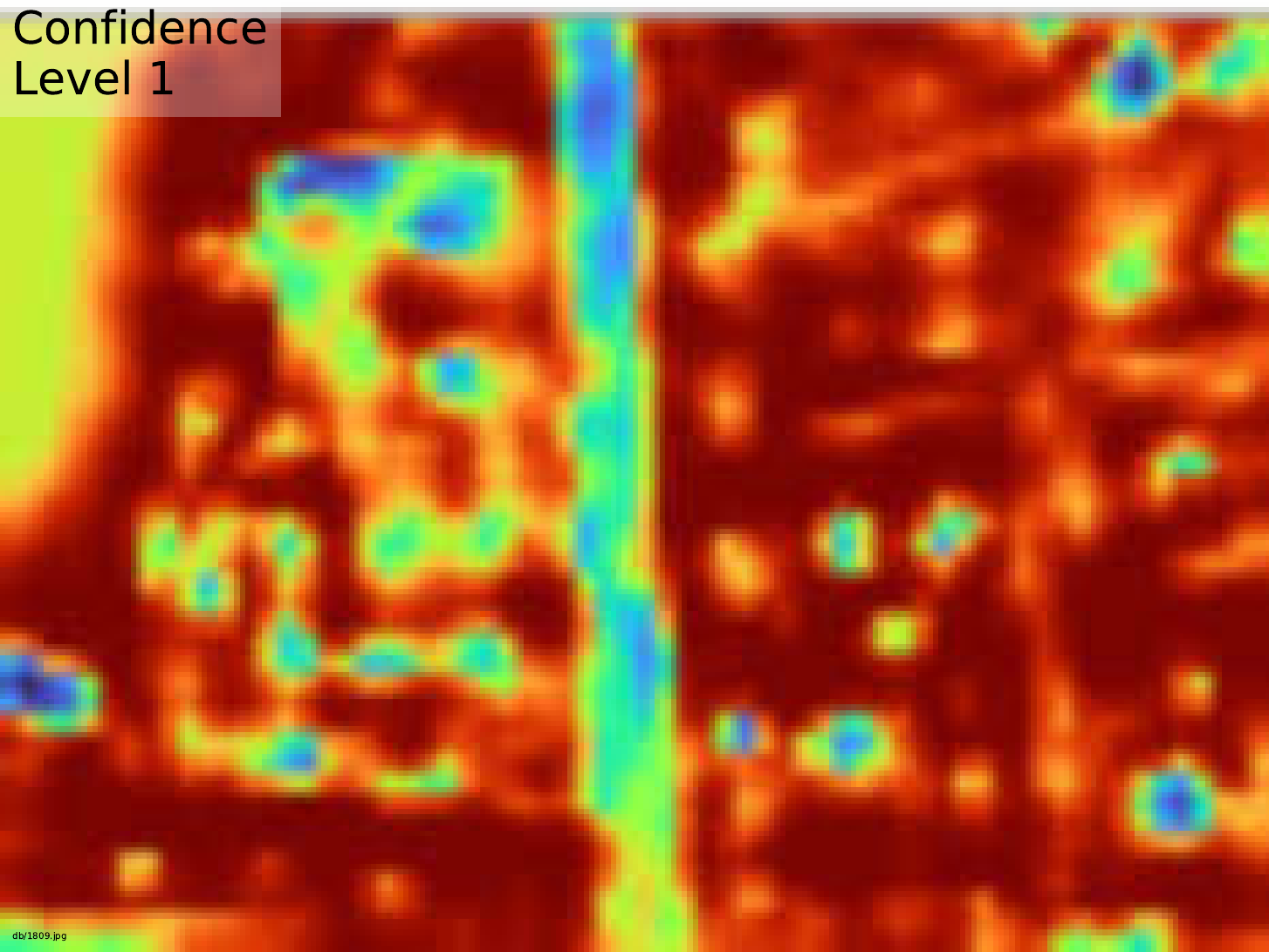}
\end{minipage}%
\begin{minipage}{\iwidth\textwidth}
    \centering
    \includegraphics[width=\pwidth\linewidth]{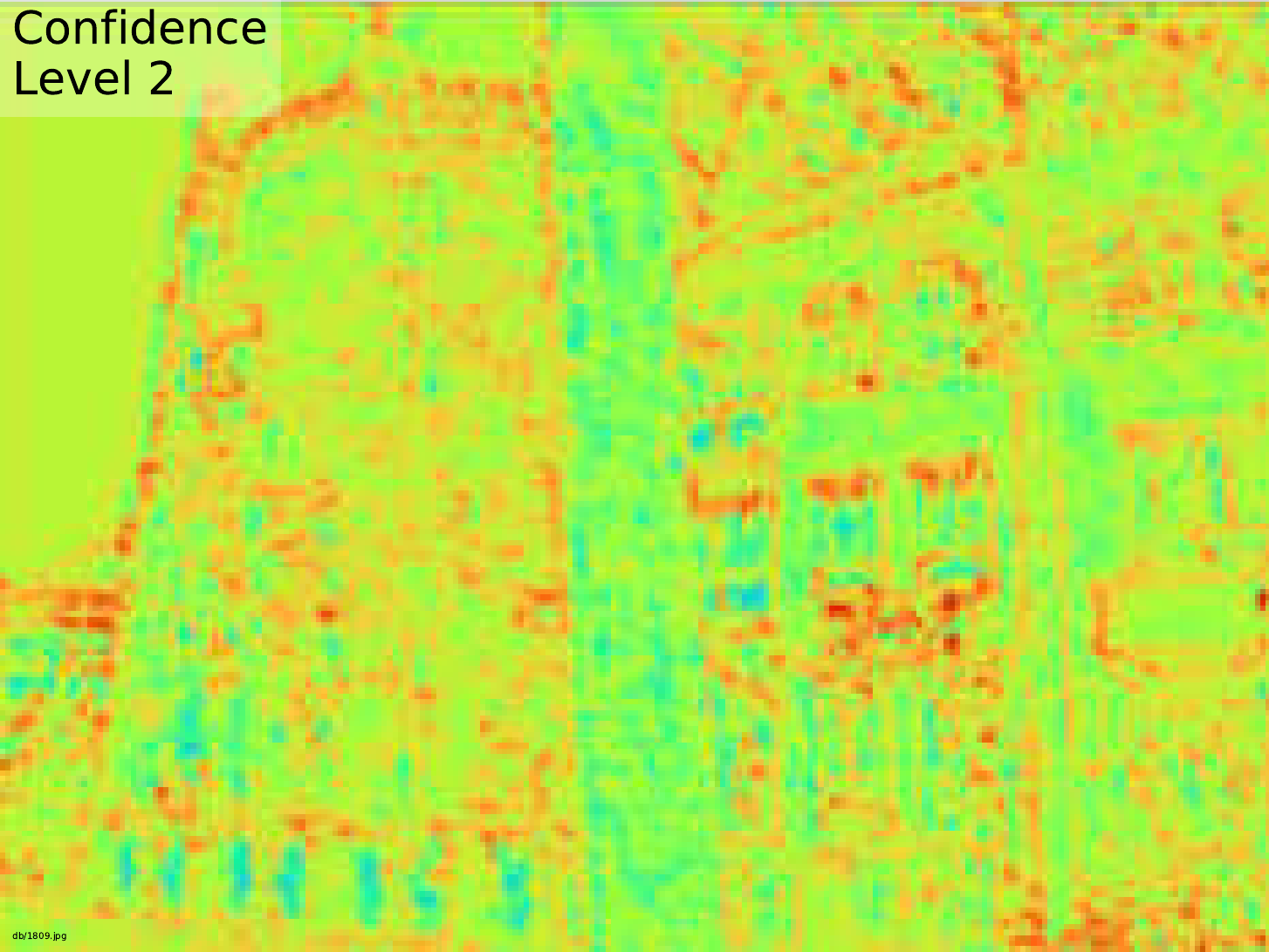}
\end{minipage}%
\begin{minipage}{\iwidth\textwidth}
    \centering
    \includegraphics[width=\pwidth\linewidth]{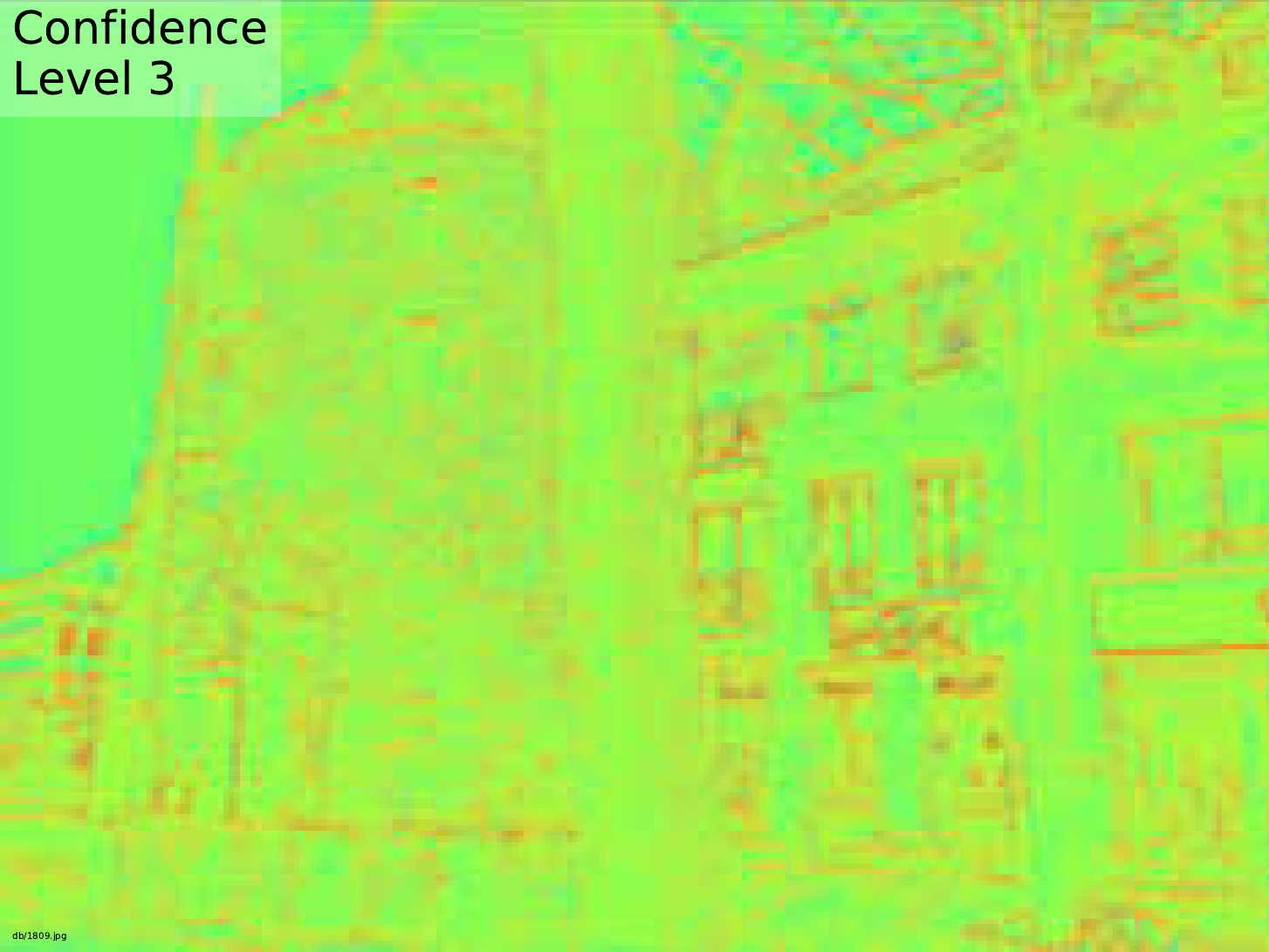}
\end{minipage}
\vspace{2mm}

\begin{minipage}{\lwidth\textwidth}
\rotatebox[origin=c]{90}{Query}
\end{minipage}%
\begin{minipage}{\iwidth\textwidth}
    \centering
    \includegraphics[width=\pwidth\linewidth]{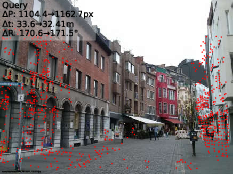}
\end{minipage}%
\begin{minipage}{\iwidth\textwidth}
    \centering
    \includegraphics[width=\pwidth\linewidth]{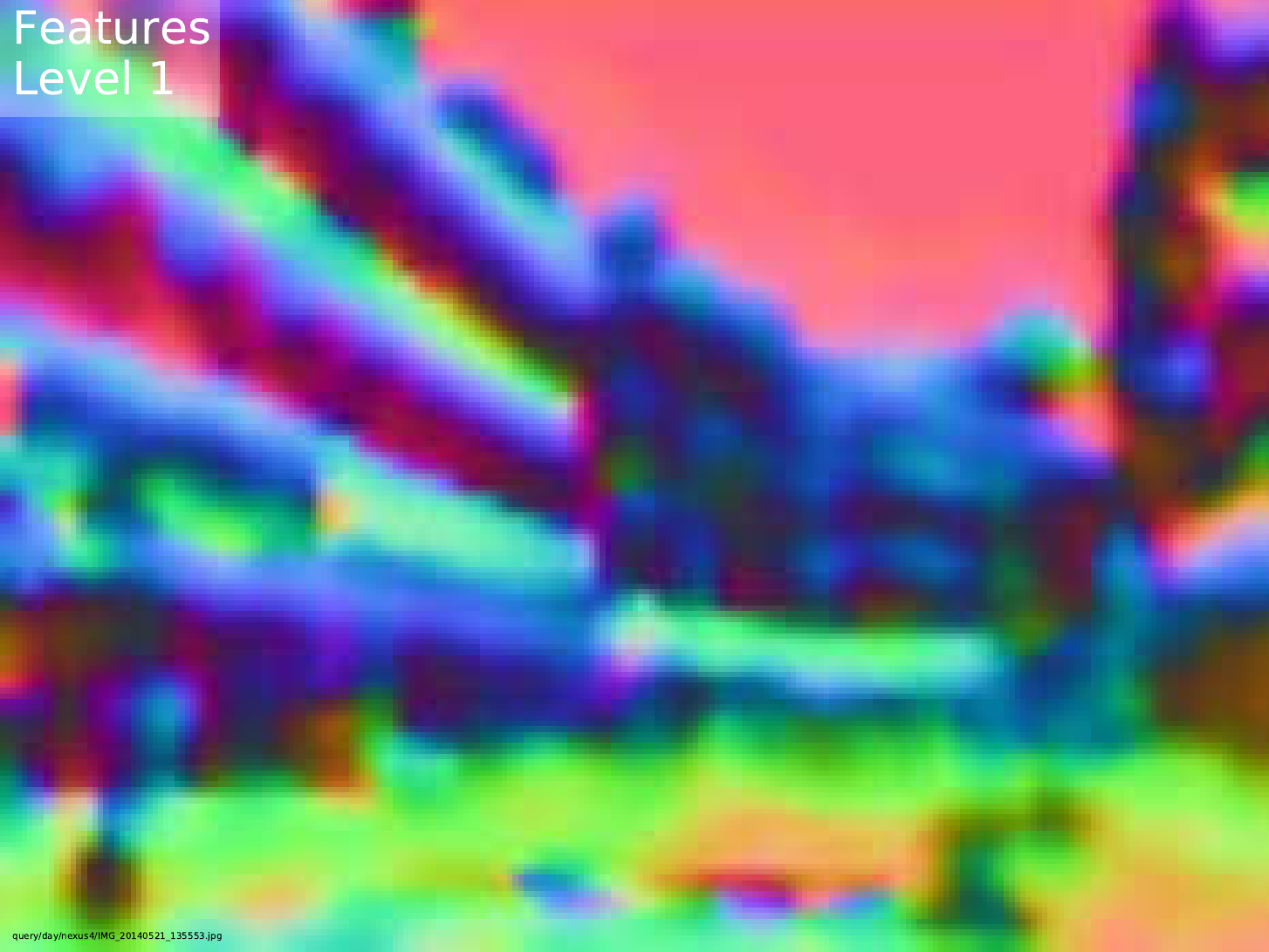}
\end{minipage}%
\begin{minipage}{\iwidth\textwidth}
    \centering
    \includegraphics[width=\pwidth\linewidth]{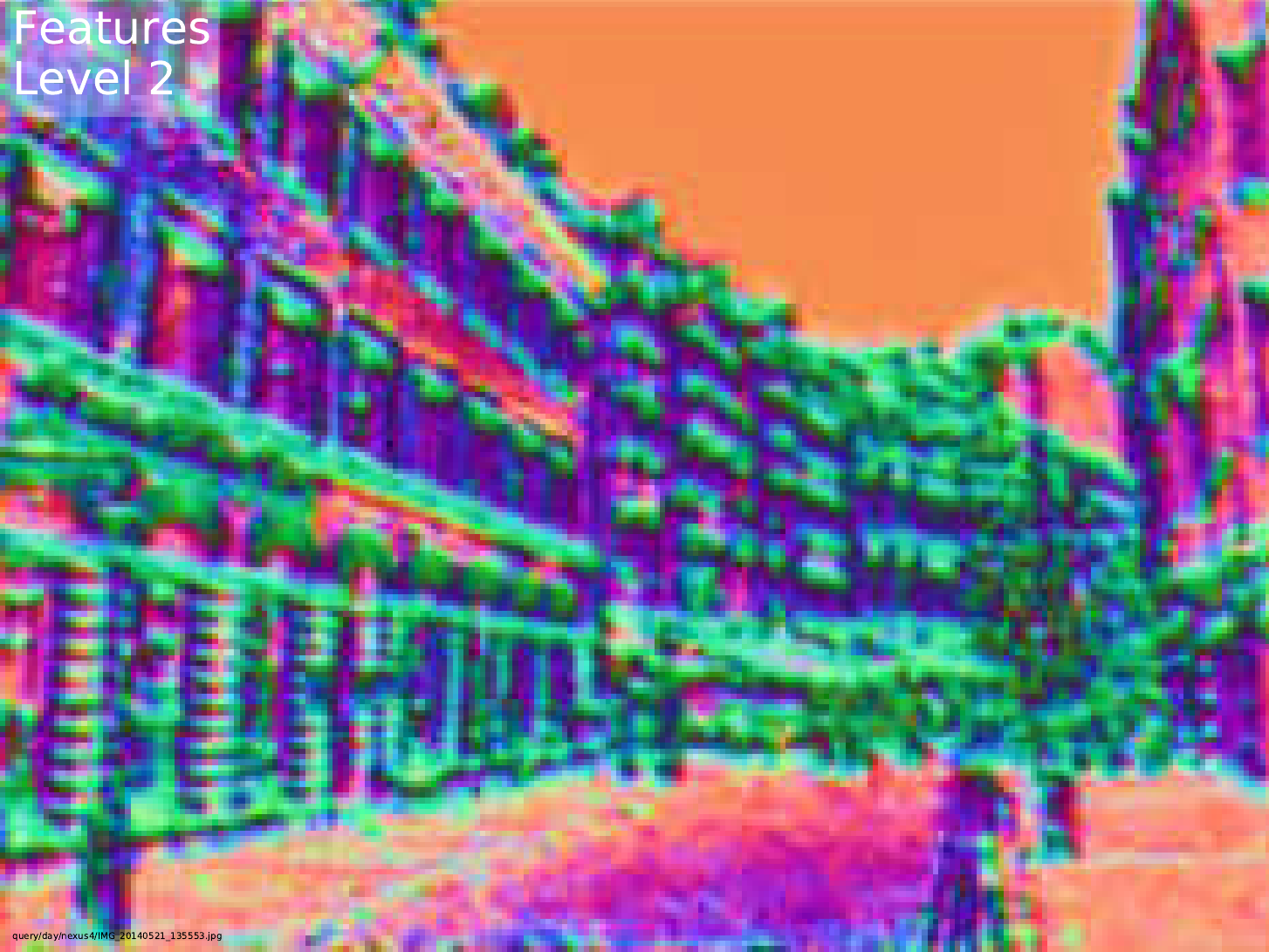}
\end{minipage}%
\begin{minipage}{\iwidth\textwidth}
    \centering
    \includegraphics[width=\pwidth\linewidth]{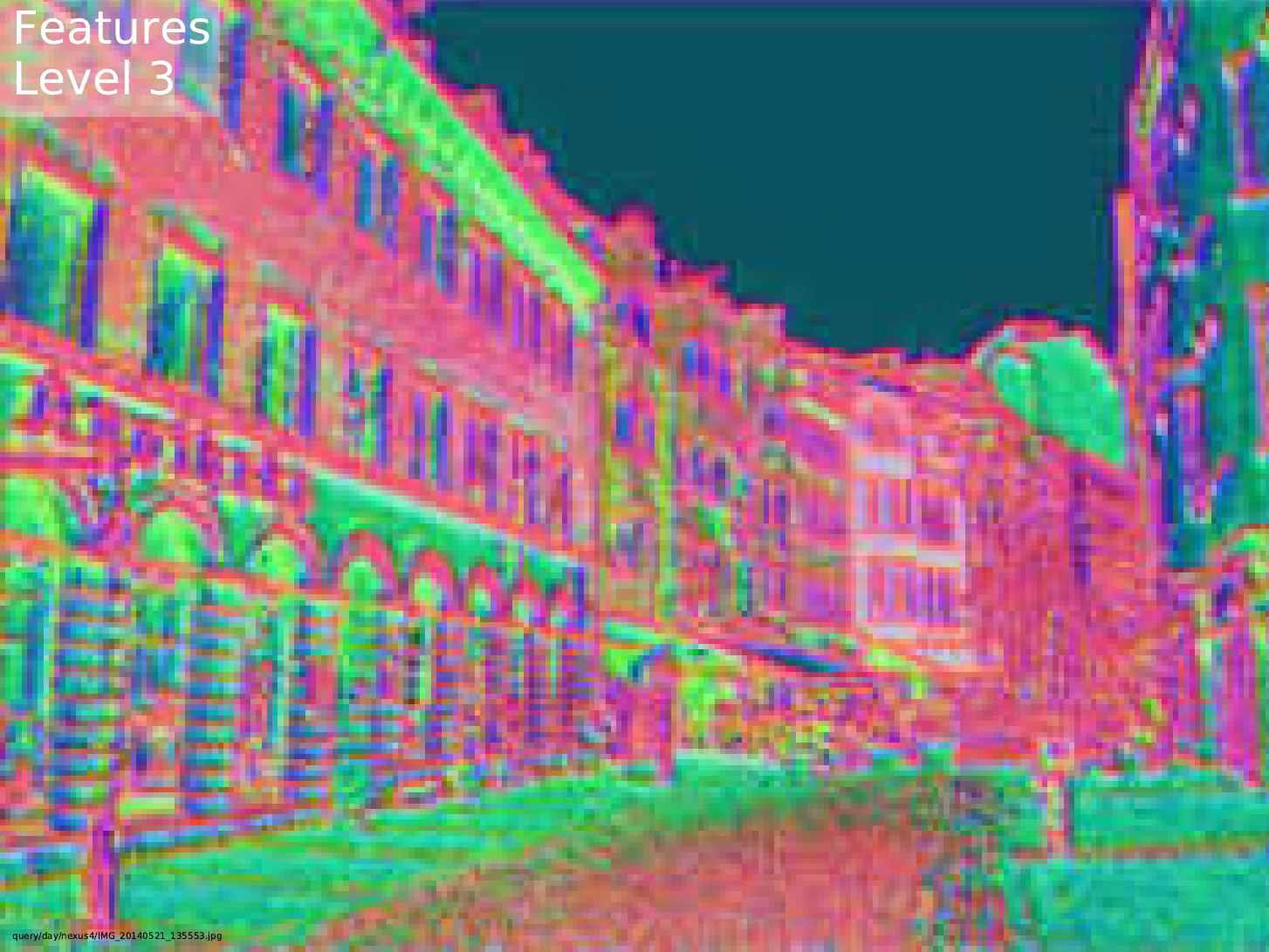}
\end{minipage}%
\begin{minipage}{\iwidth\textwidth}
    \centering
    \includegraphics[width=\pwidth\linewidth]{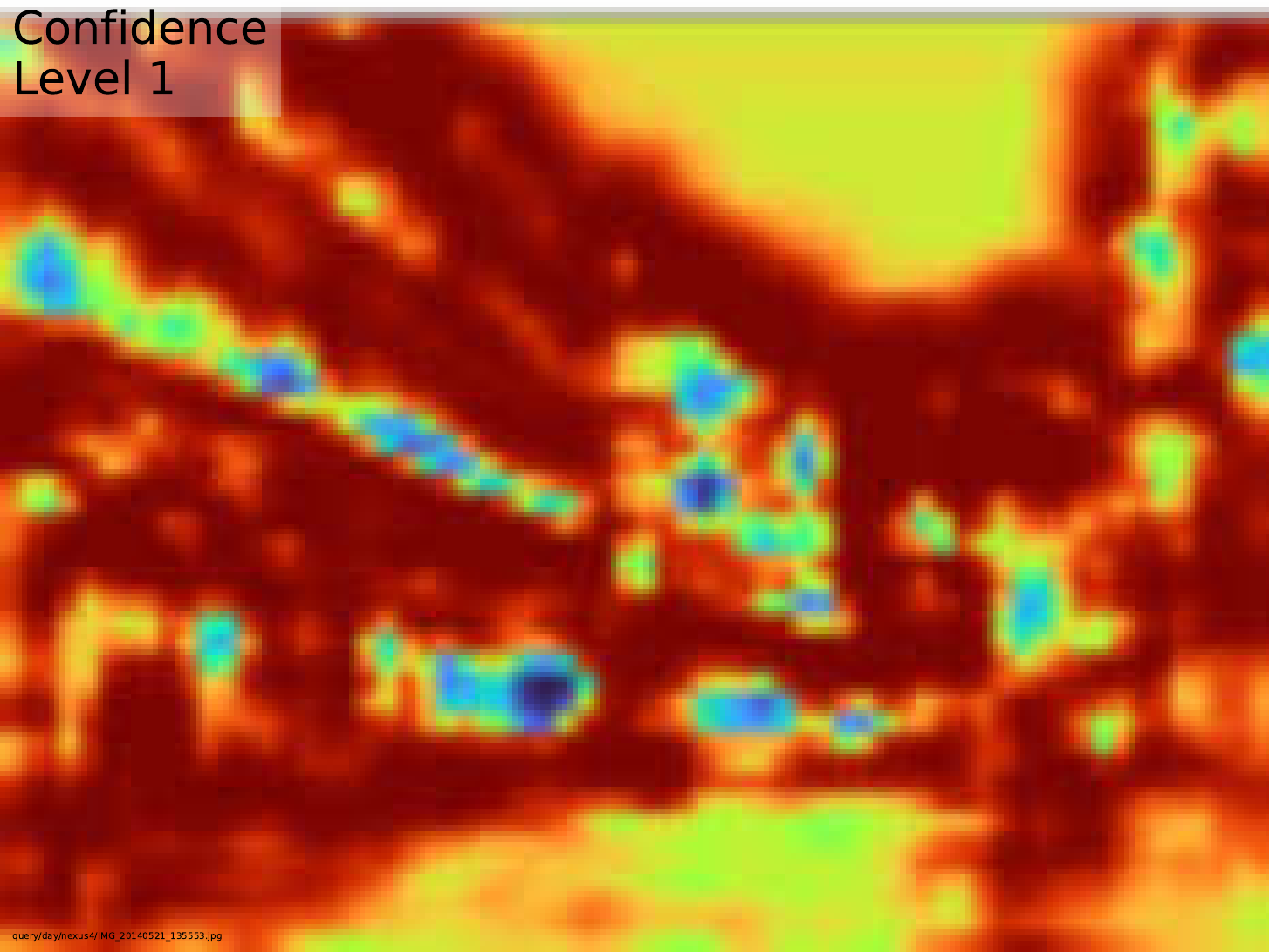}
\end{minipage}%
\begin{minipage}{\iwidth\textwidth}
    \centering
    \includegraphics[width=\pwidth\linewidth]{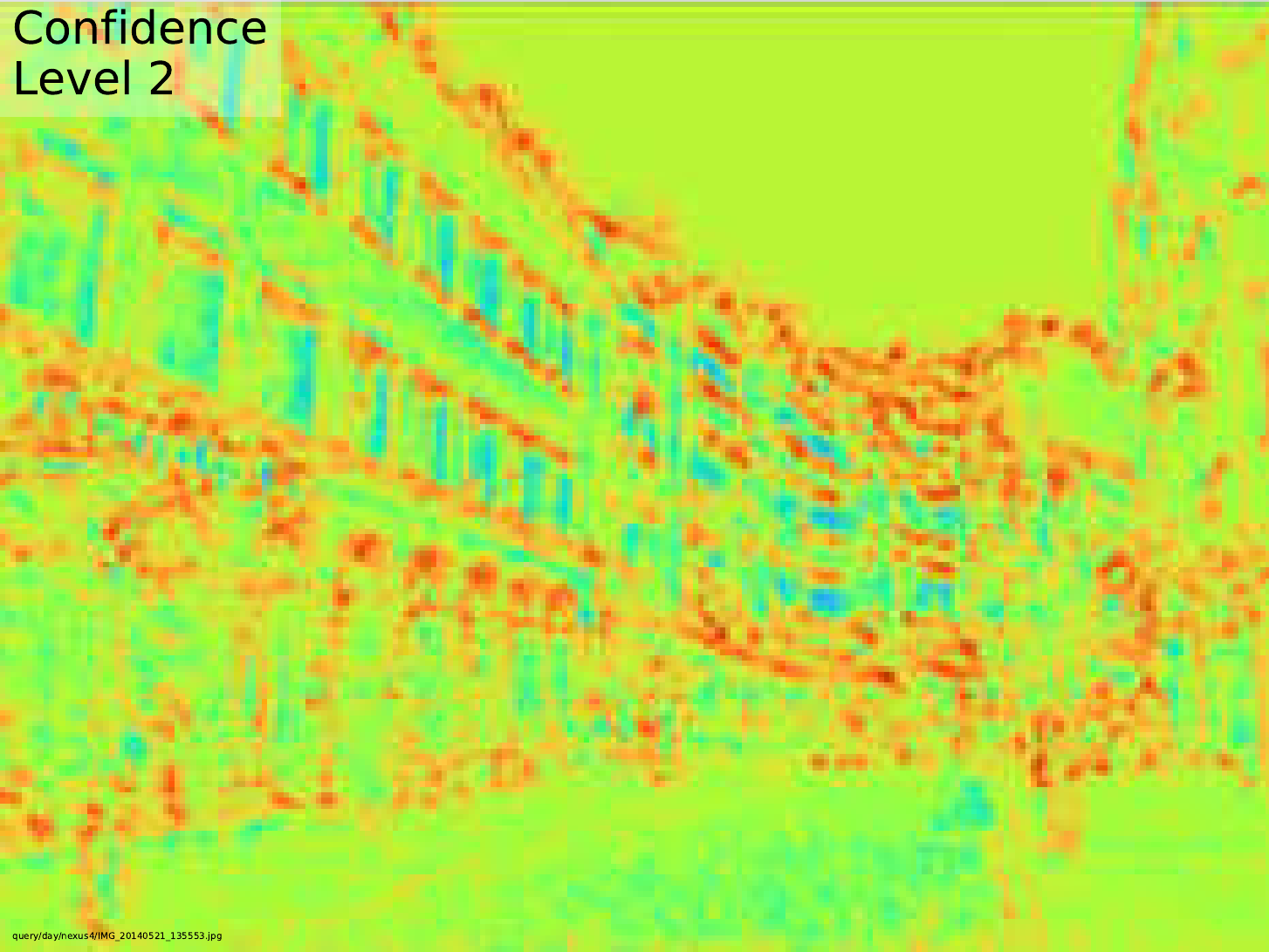}
\end{minipage}%
\begin{minipage}{\iwidth\textwidth}
    \centering
    \includegraphics[width=\pwidth\linewidth]{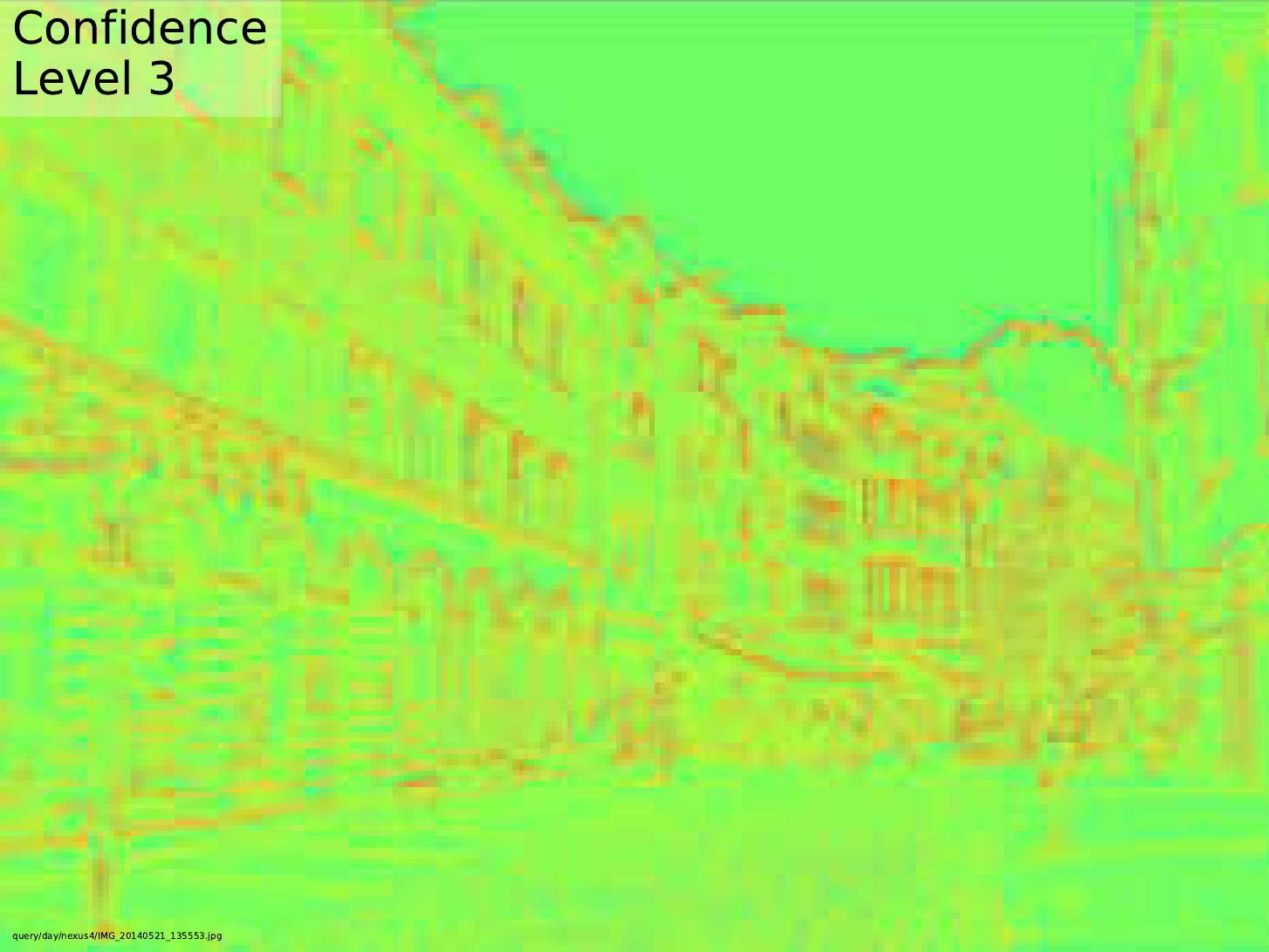}
\end{minipage}
\begin{minipage}{\lwidth\textwidth}
\rotatebox[origin=c]{90}{Reference}
\end{minipage}%
\begin{minipage}{\iwidth\textwidth}
    \centering
    \includegraphics[width=\pwidth\linewidth]{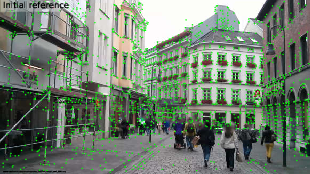}
\end{minipage}%
\begin{minipage}{\iwidth\textwidth}
    \centering
    \includegraphics[width=\pwidth\linewidth]{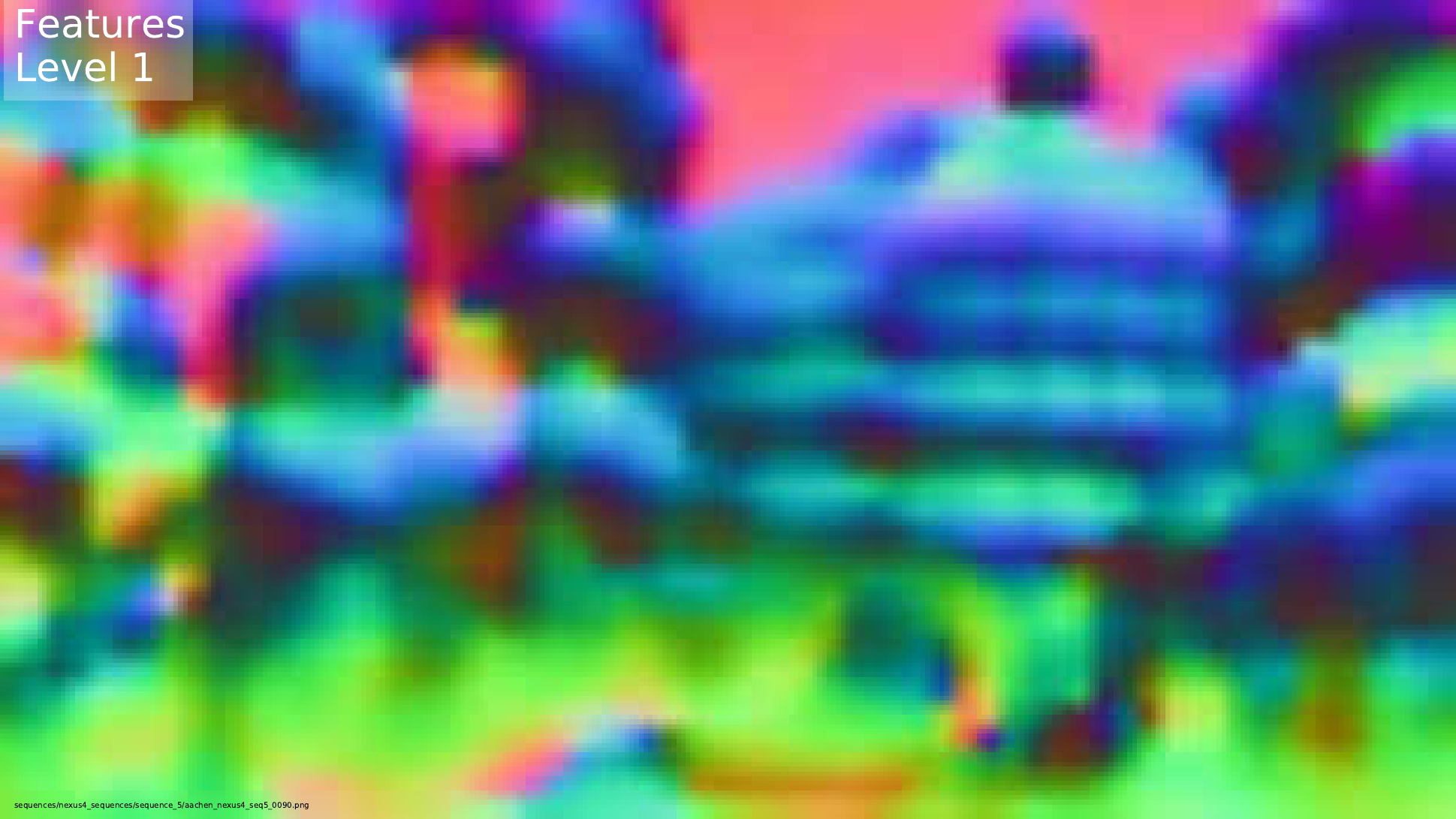}
\end{minipage}%
\begin{minipage}{\iwidth\textwidth}
    \centering
    \includegraphics[width=\pwidth\linewidth]{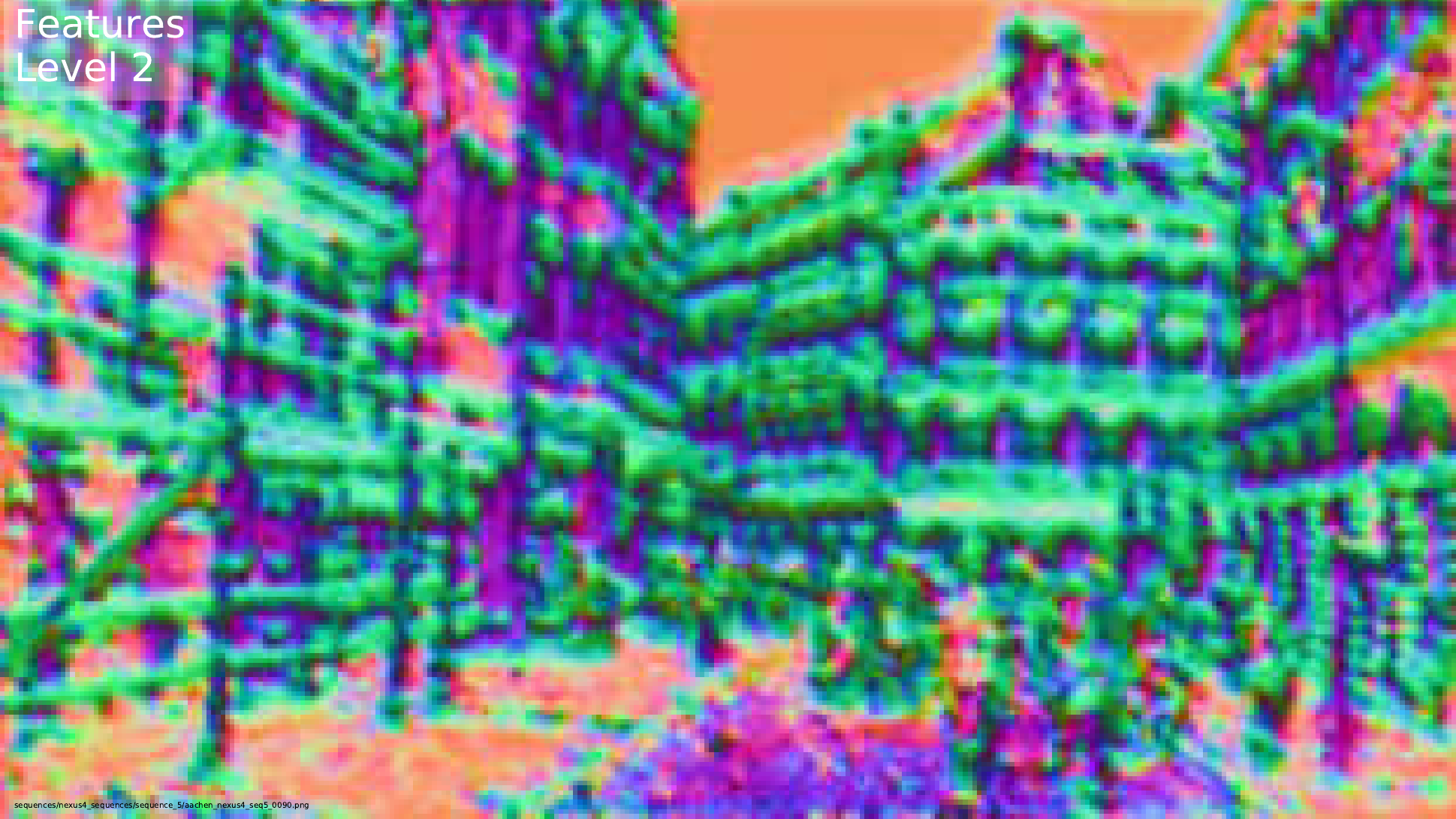}
\end{minipage}%
\begin{minipage}{\iwidth\textwidth}
    \centering
    \includegraphics[width=\pwidth\linewidth]{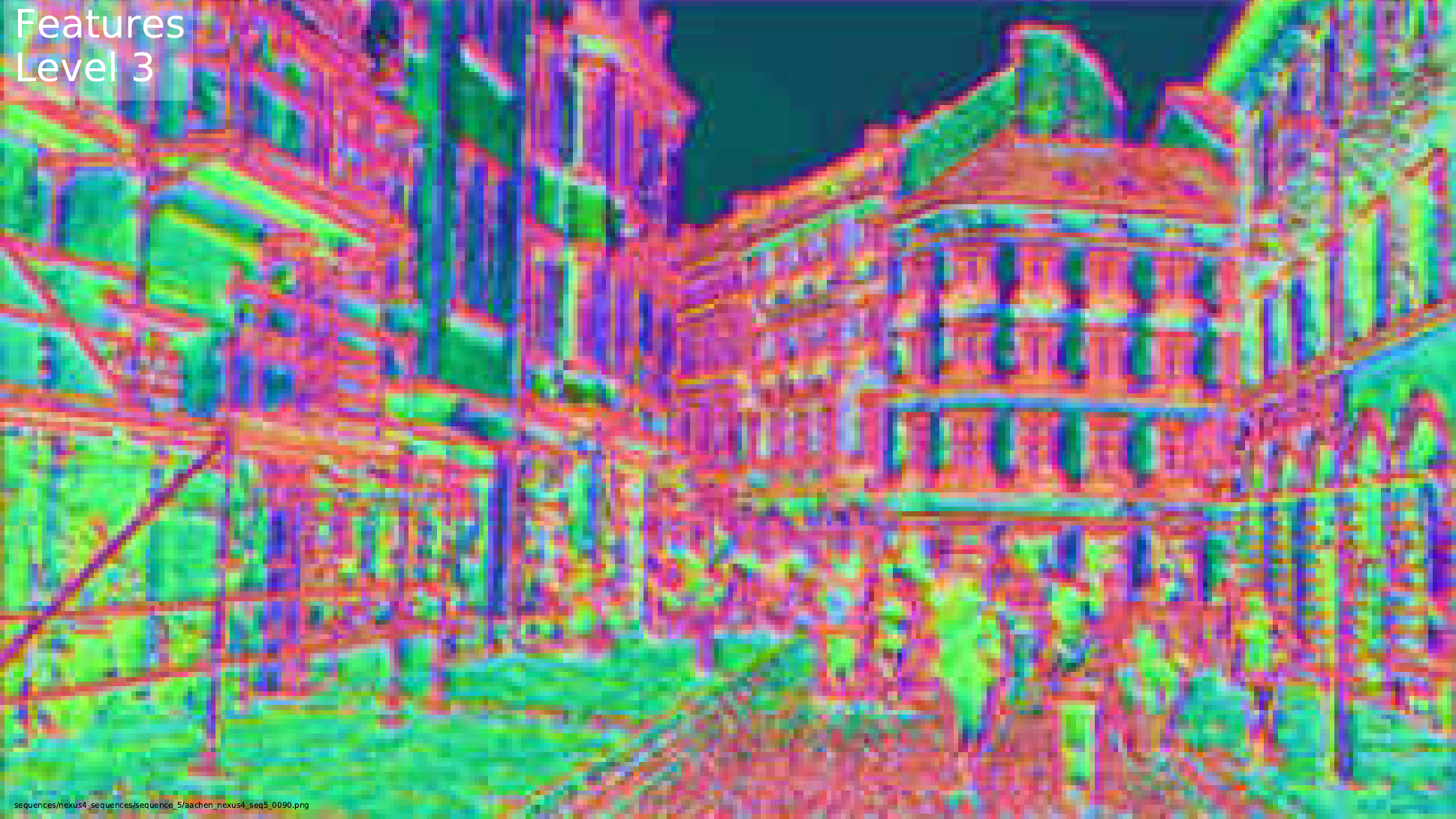}
\end{minipage}%
\begin{minipage}{\iwidth\textwidth}
    \centering
    \includegraphics[width=\pwidth\linewidth]{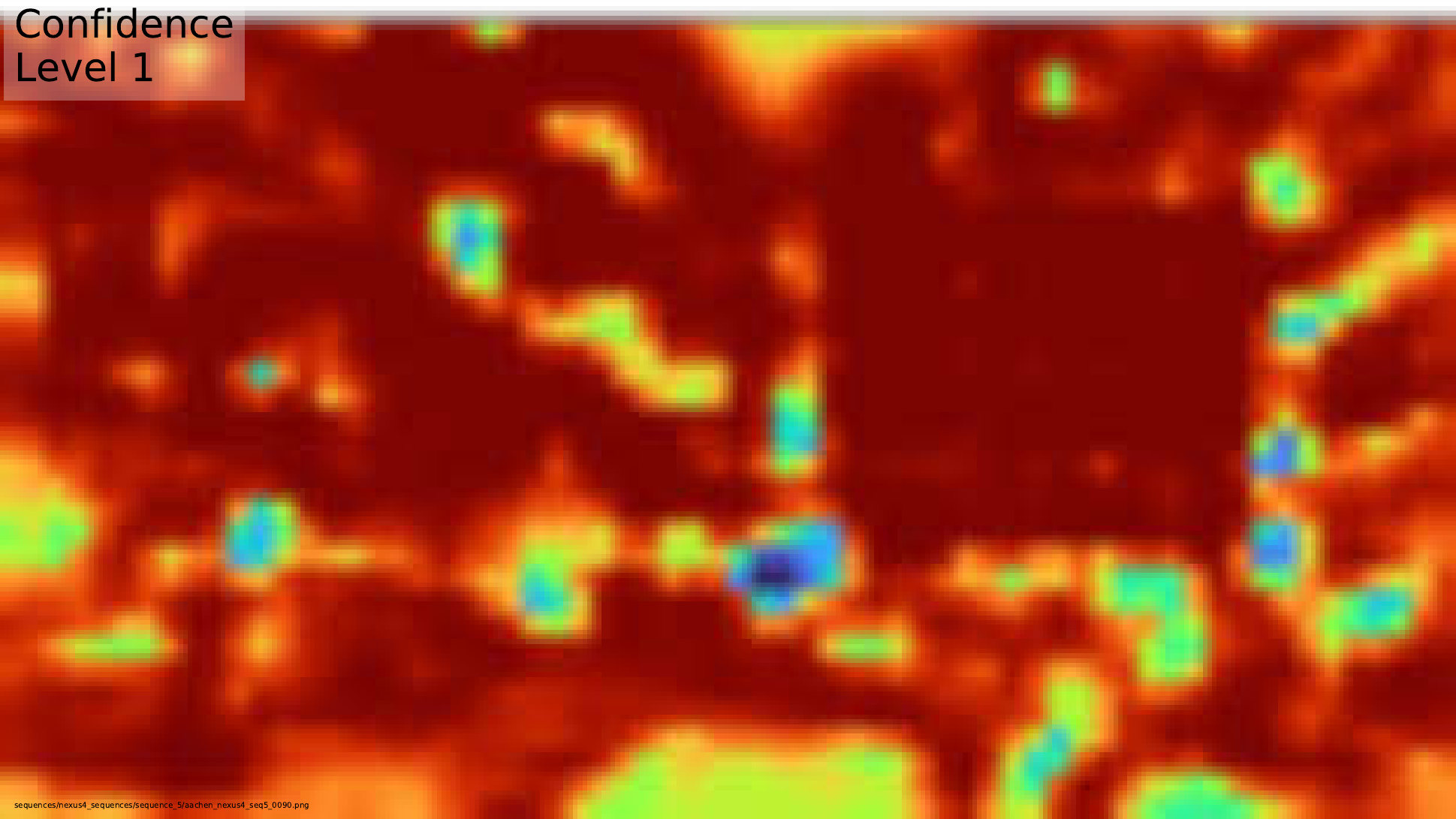}
\end{minipage}%
\begin{minipage}{\iwidth\textwidth}
    \centering
    \includegraphics[width=\pwidth\linewidth]{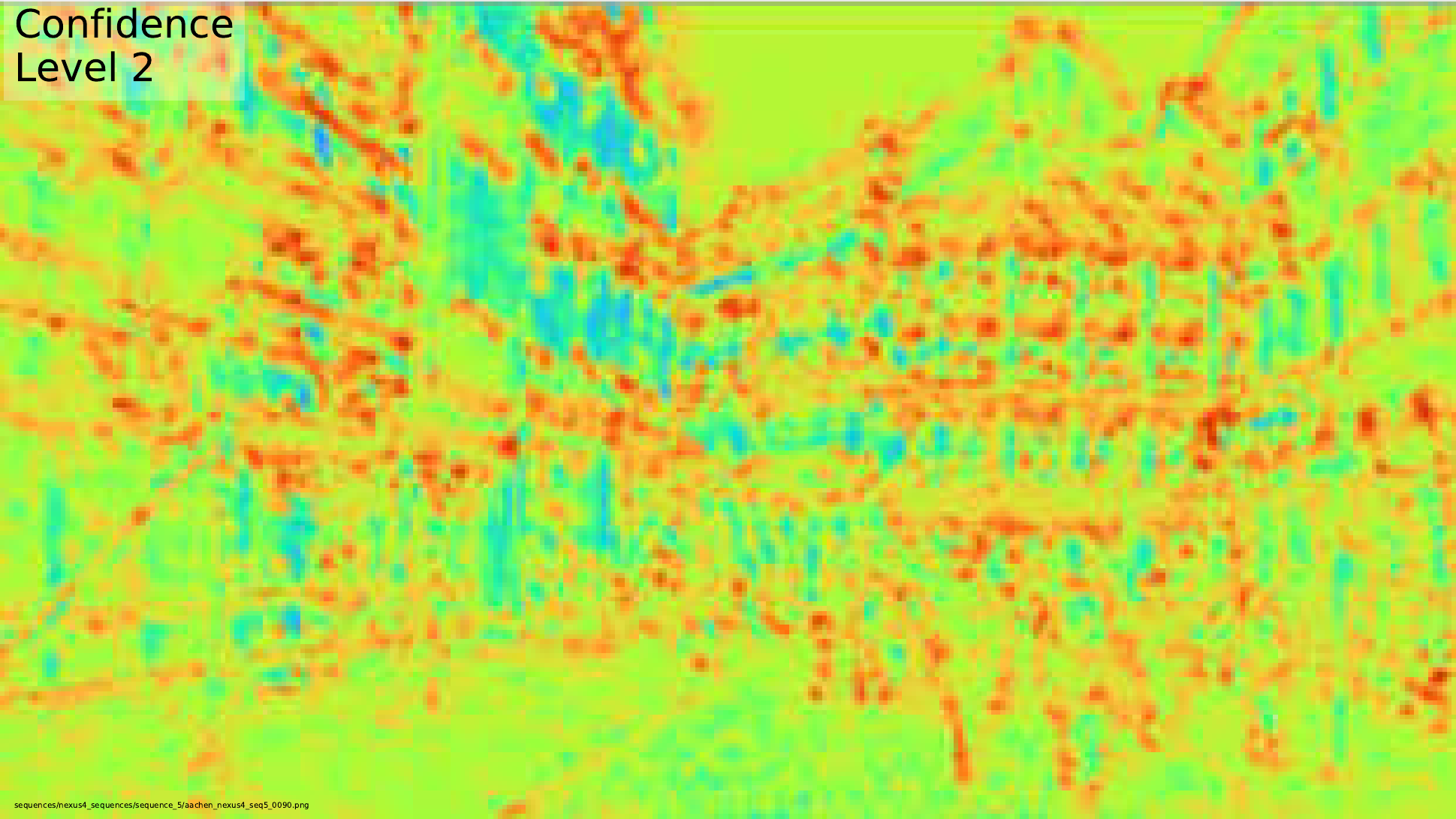}
\end{minipage}%
\begin{minipage}{\iwidth\textwidth}
    \centering
    \includegraphics[width=\pwidth\linewidth]{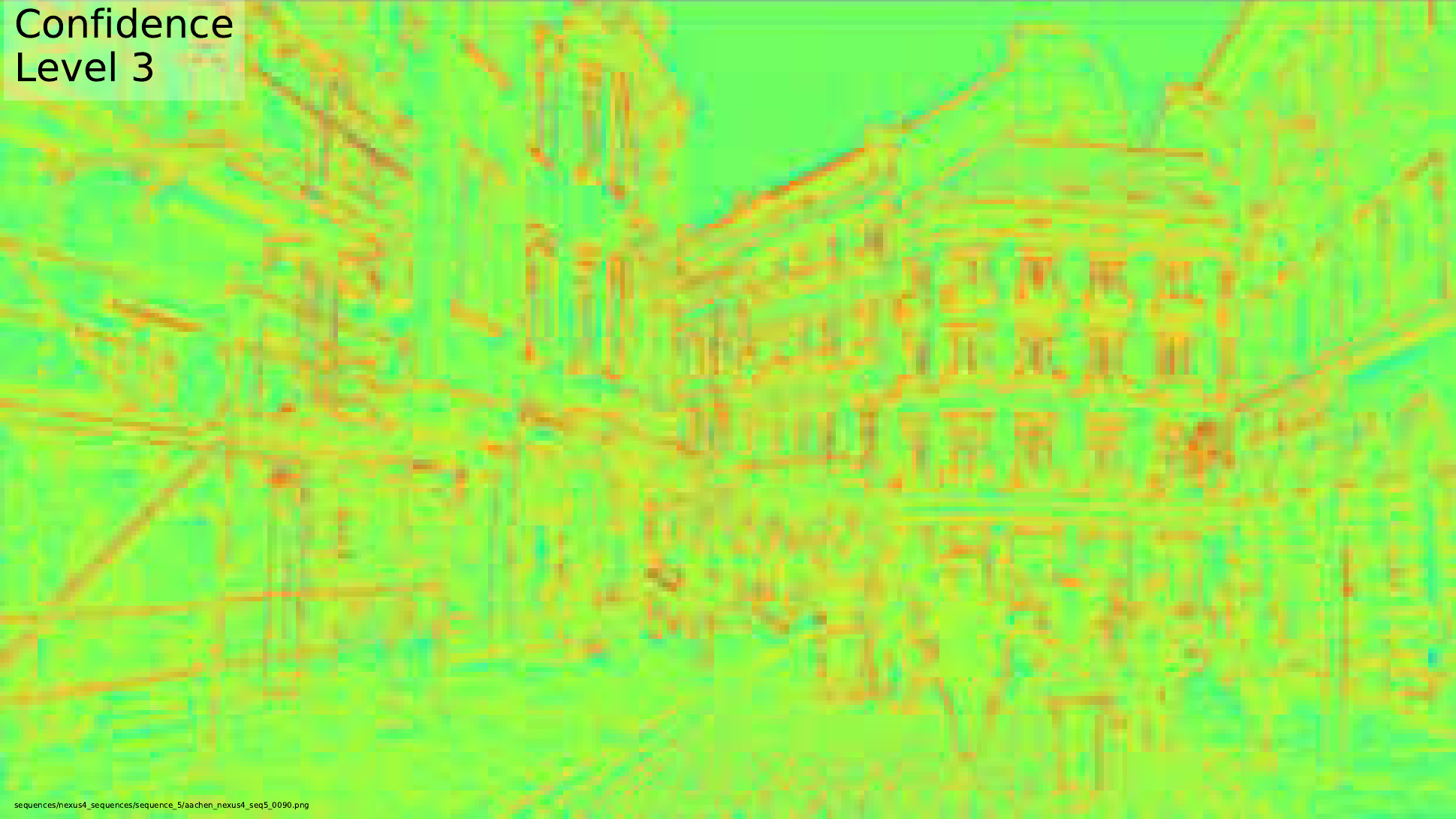}
\end{minipage}
\vspace{2mm}

\begin{minipage}{\lwidth\textwidth}
\rotatebox[origin=c]{90}{Query}
\end{minipage}%
\begin{minipage}{\iwidth\textwidth}
    \centering
    \includegraphics[width=\pwidth\linewidth]{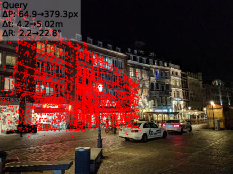}
\end{minipage}%
\begin{minipage}{\iwidth\textwidth}
    \centering
    \includegraphics[width=\pwidth\linewidth]{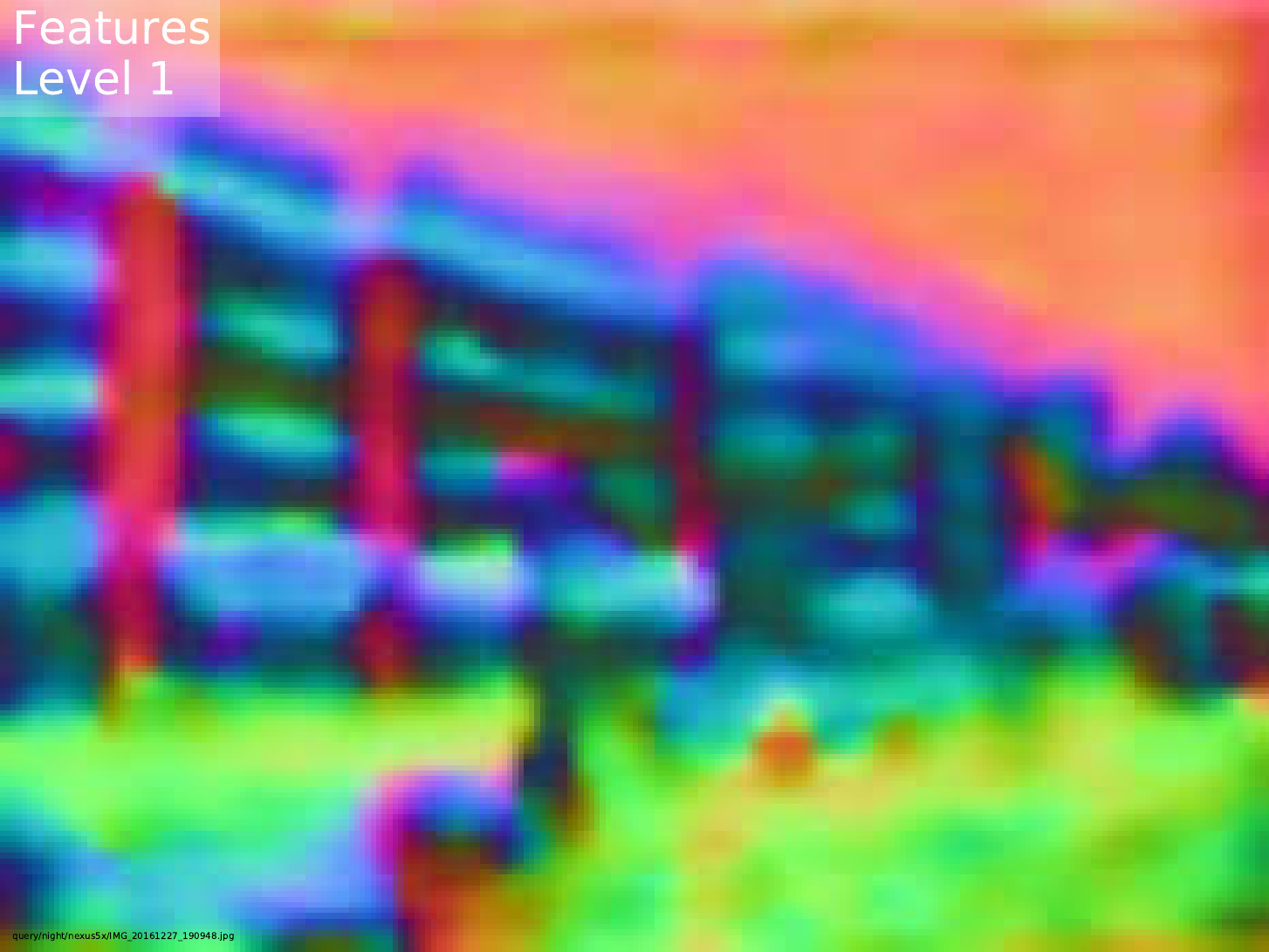}
\end{minipage}%
\begin{minipage}{\iwidth\textwidth}
    \centering
    \includegraphics[width=\pwidth\linewidth]{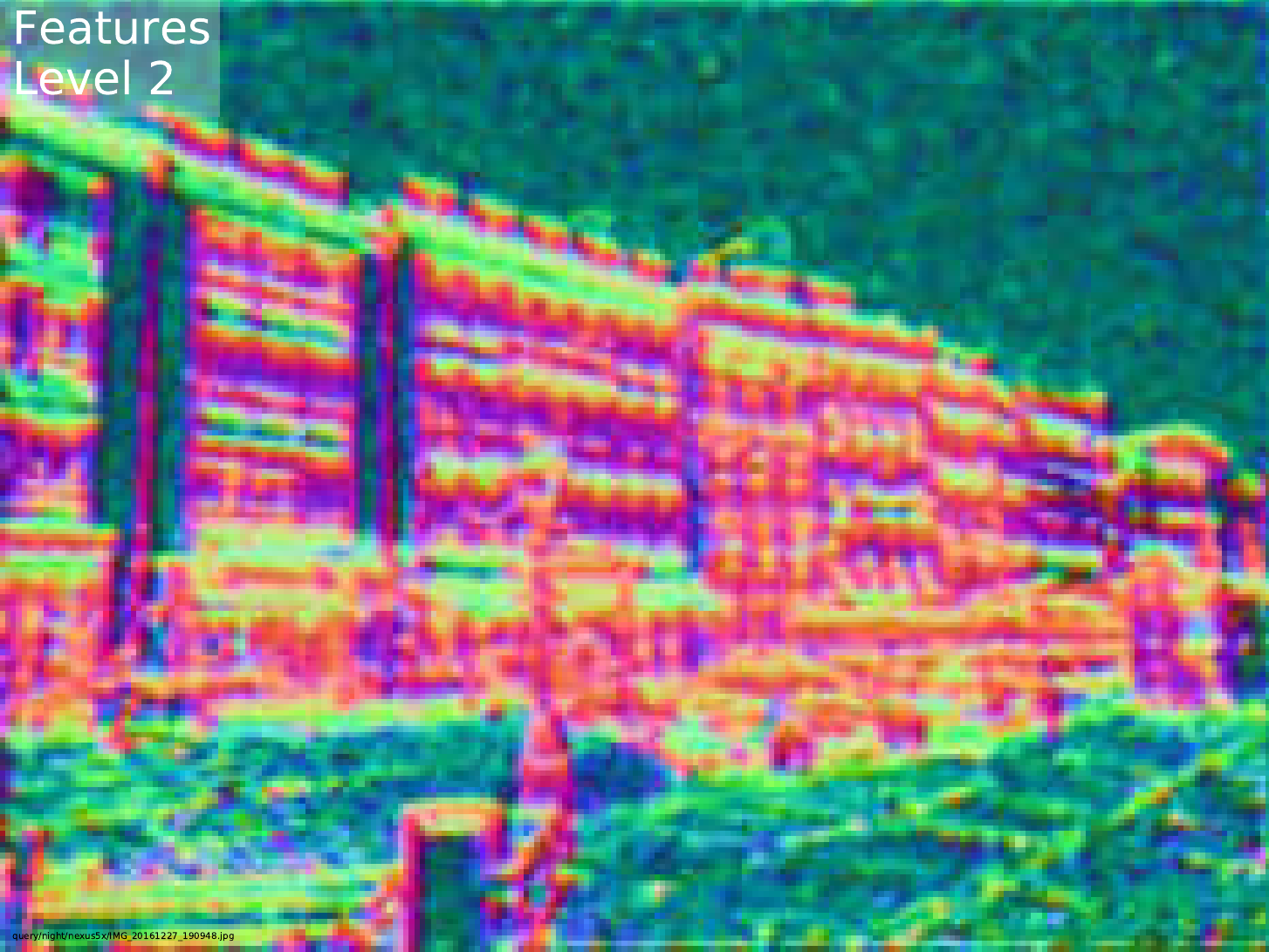}
\end{minipage}%
\begin{minipage}{\iwidth\textwidth}
    \centering
    \includegraphics[width=\pwidth\linewidth]{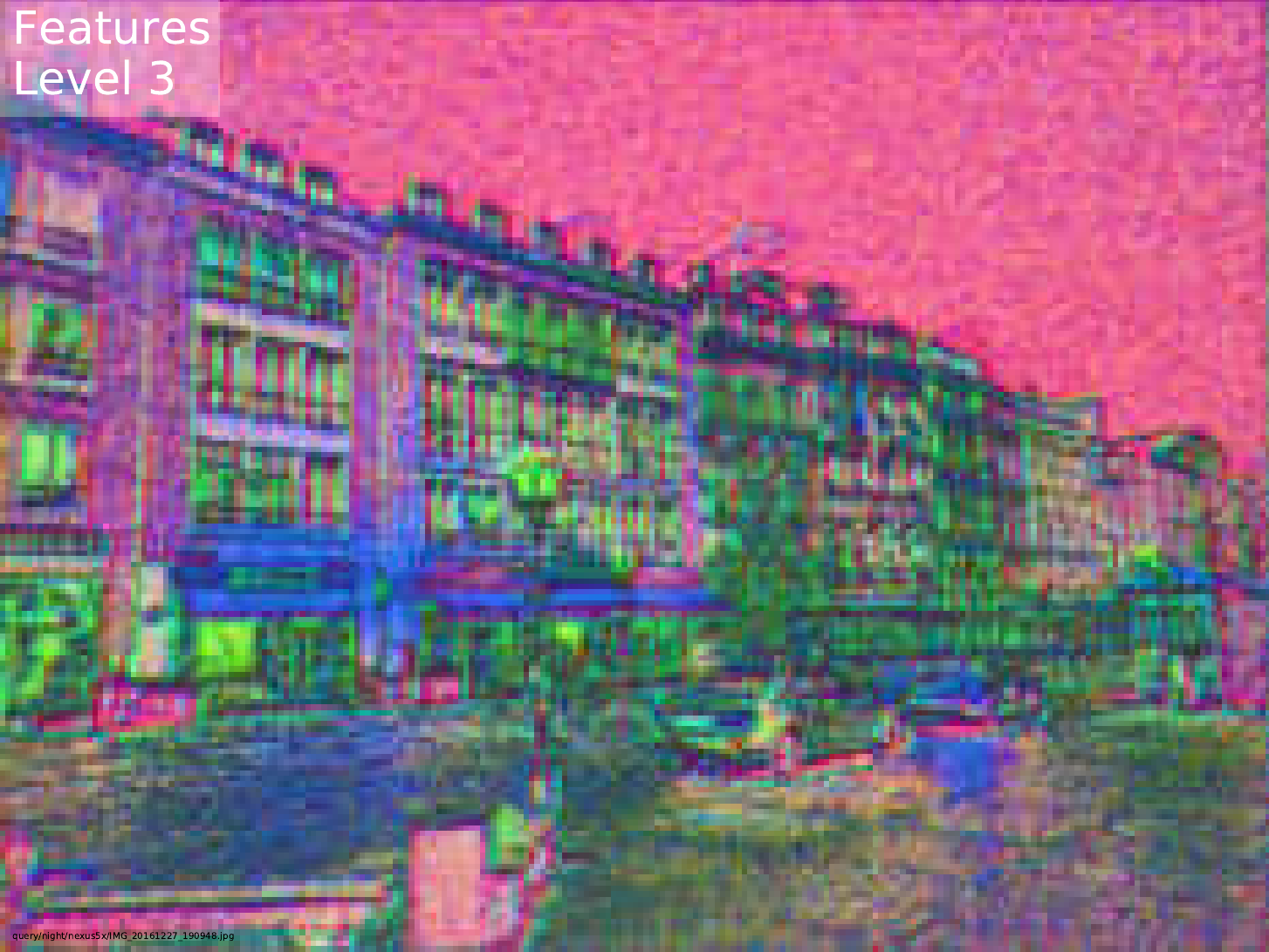}
\end{minipage}%
\begin{minipage}{\iwidth\textwidth}
    \centering
    \includegraphics[width=\pwidth\linewidth]{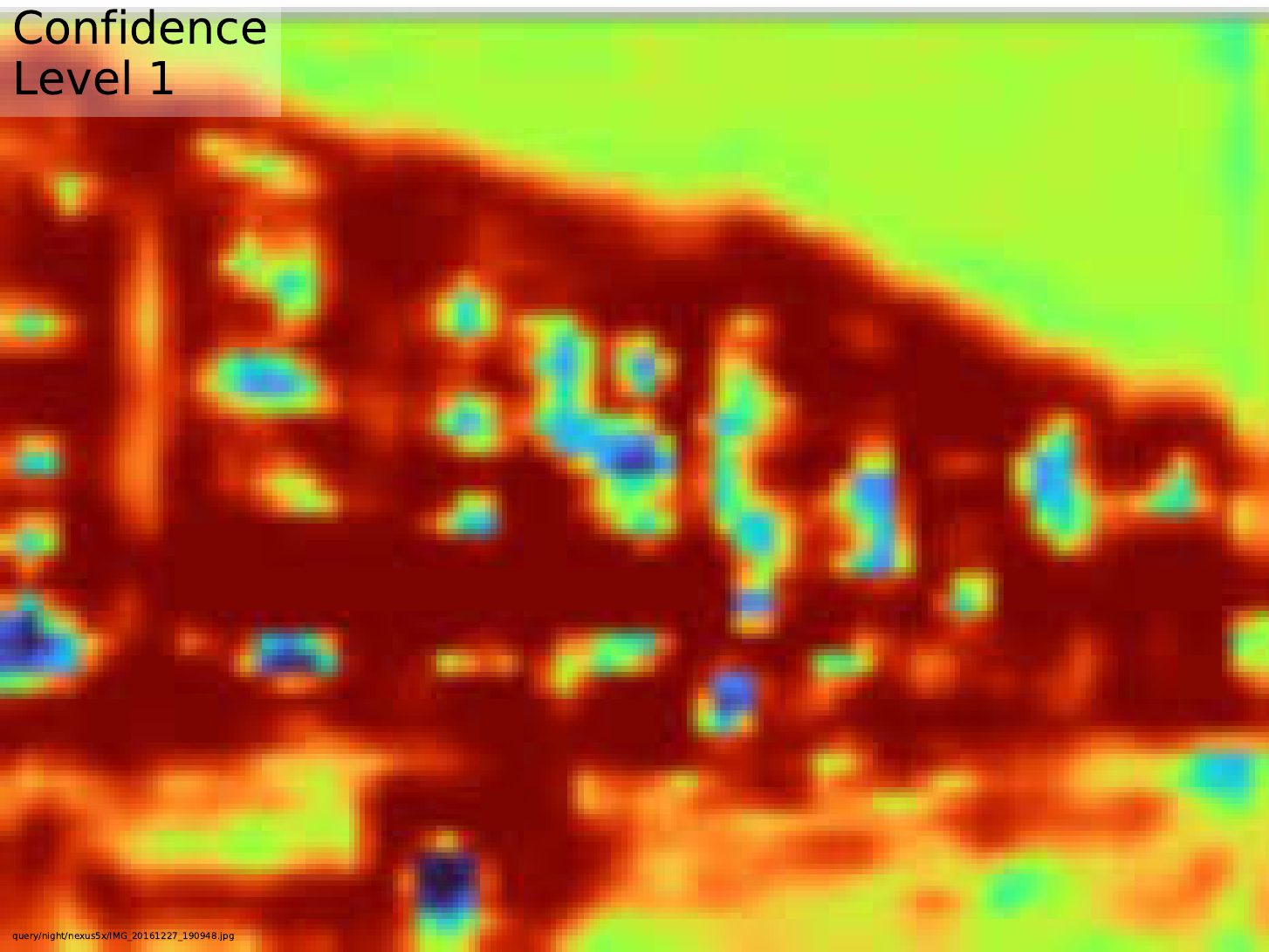}
\end{minipage}%
\begin{minipage}{\iwidth\textwidth}
    \centering
    \includegraphics[width=\pwidth\linewidth]{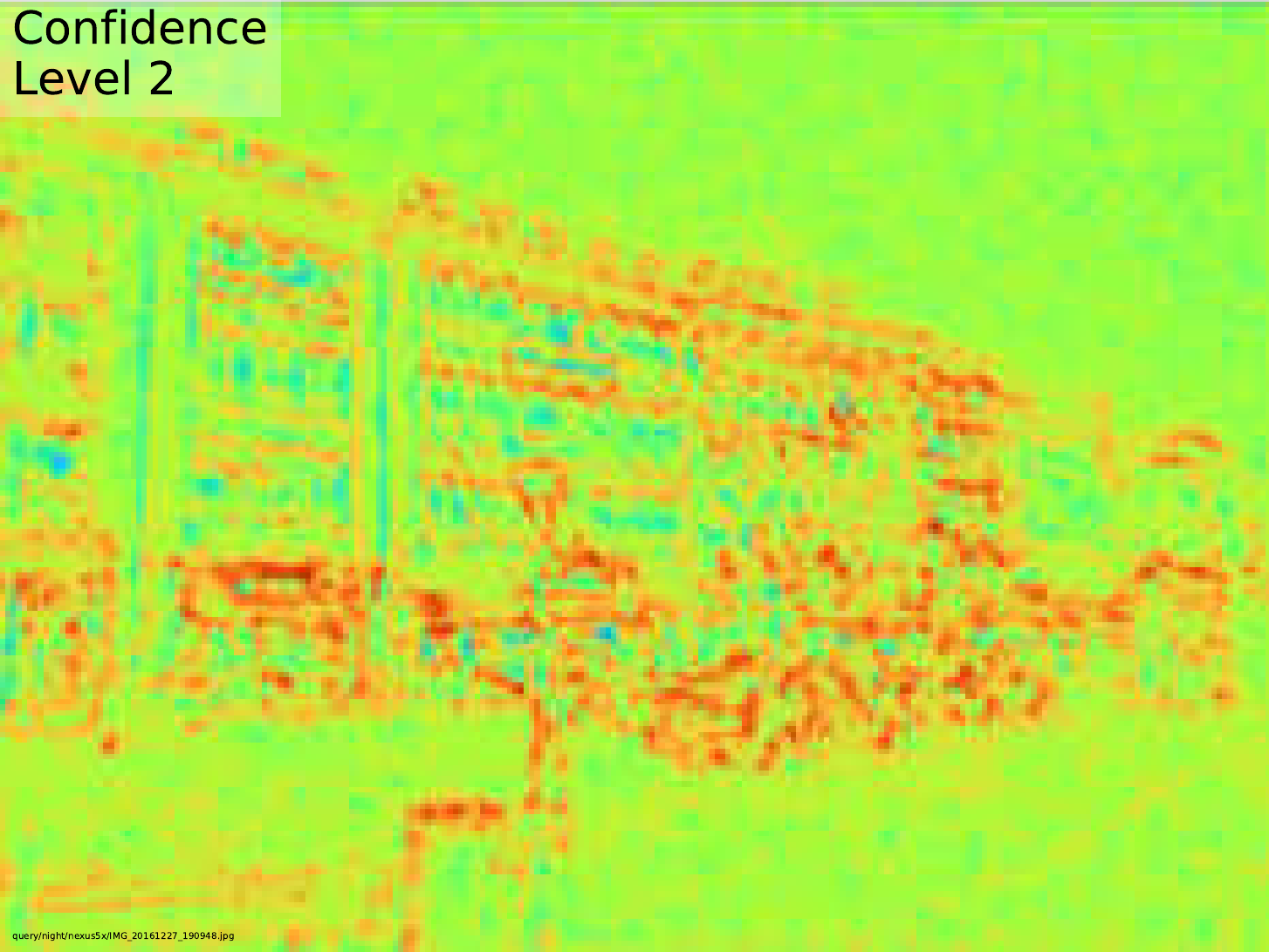}
\end{minipage}%
\begin{minipage}{\iwidth\textwidth}
    \centering
    \includegraphics[width=\pwidth\linewidth]{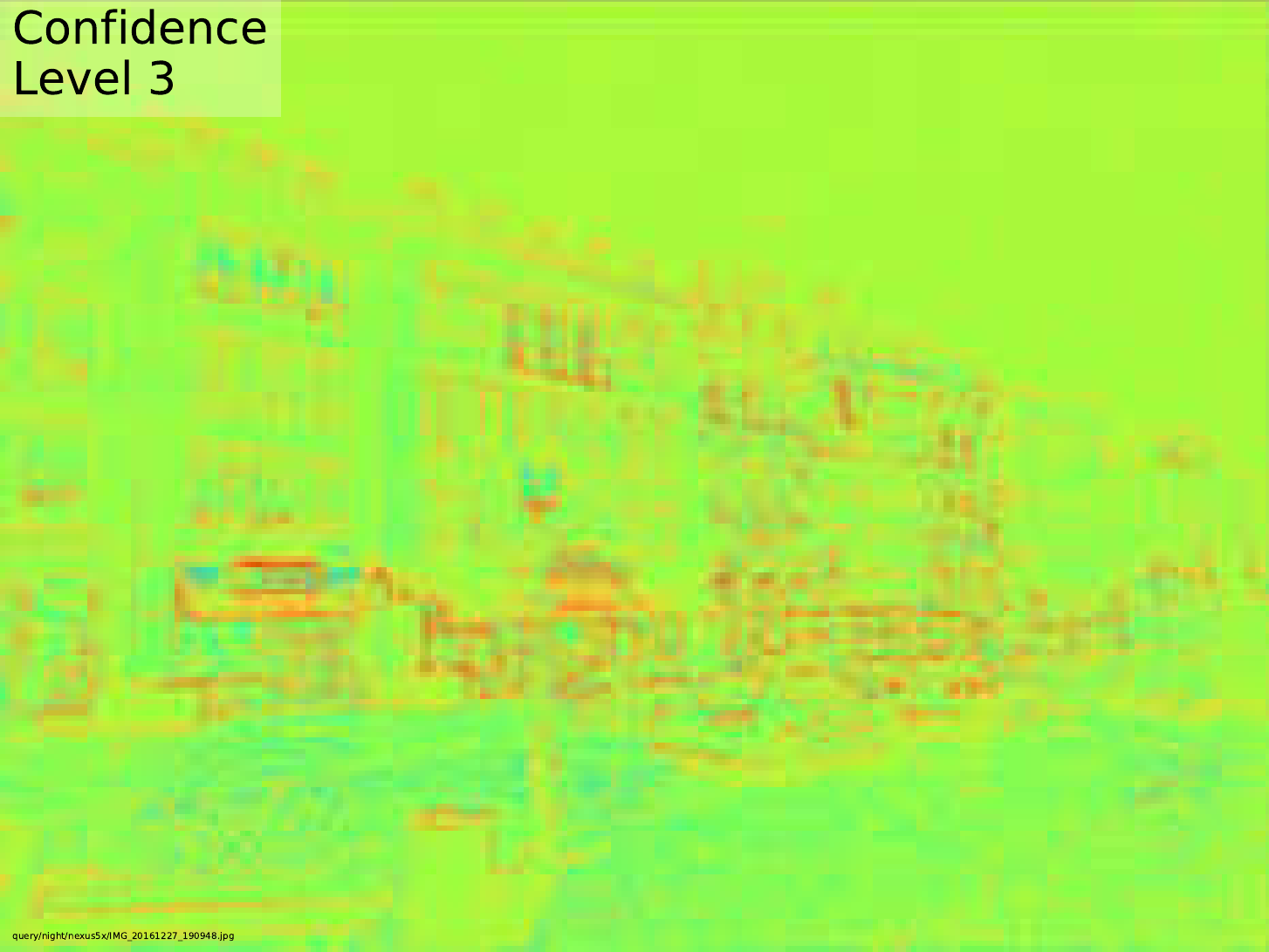}
\end{minipage}
\begin{minipage}{\lwidth\textwidth}
\rotatebox[origin=c]{90}{Reference}
\end{minipage}%
\begin{minipage}{\iwidth\textwidth}
    \centering
    \includegraphics[width=\pwidth\linewidth]{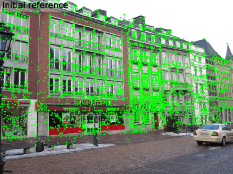}
\end{minipage}%
\begin{minipage}{\iwidth\textwidth}
    \centering
    \includegraphics[width=\pwidth\linewidth]{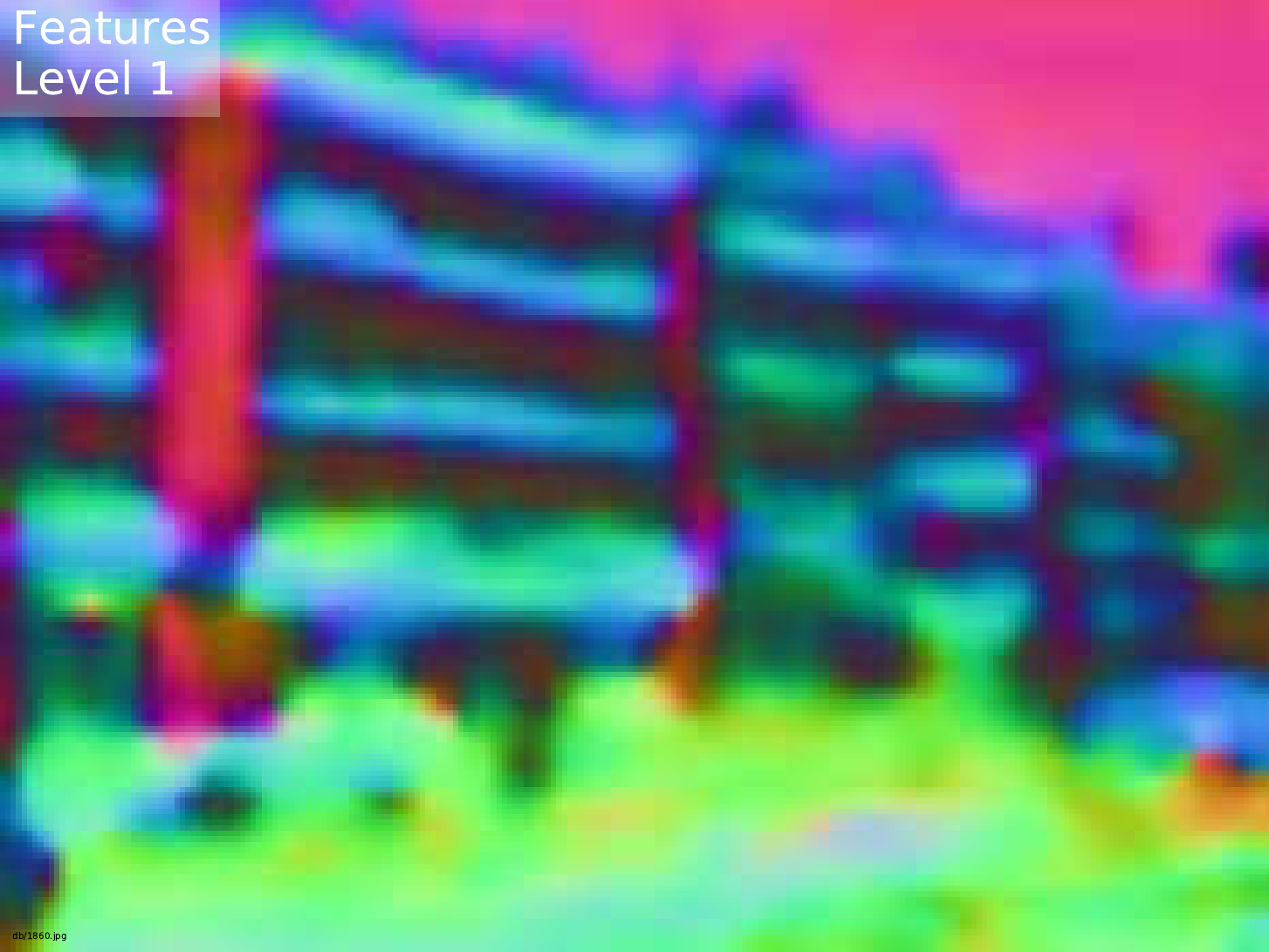}
\end{minipage}%
\begin{minipage}{\iwidth\textwidth}
    \centering
    \includegraphics[width=\pwidth\linewidth]{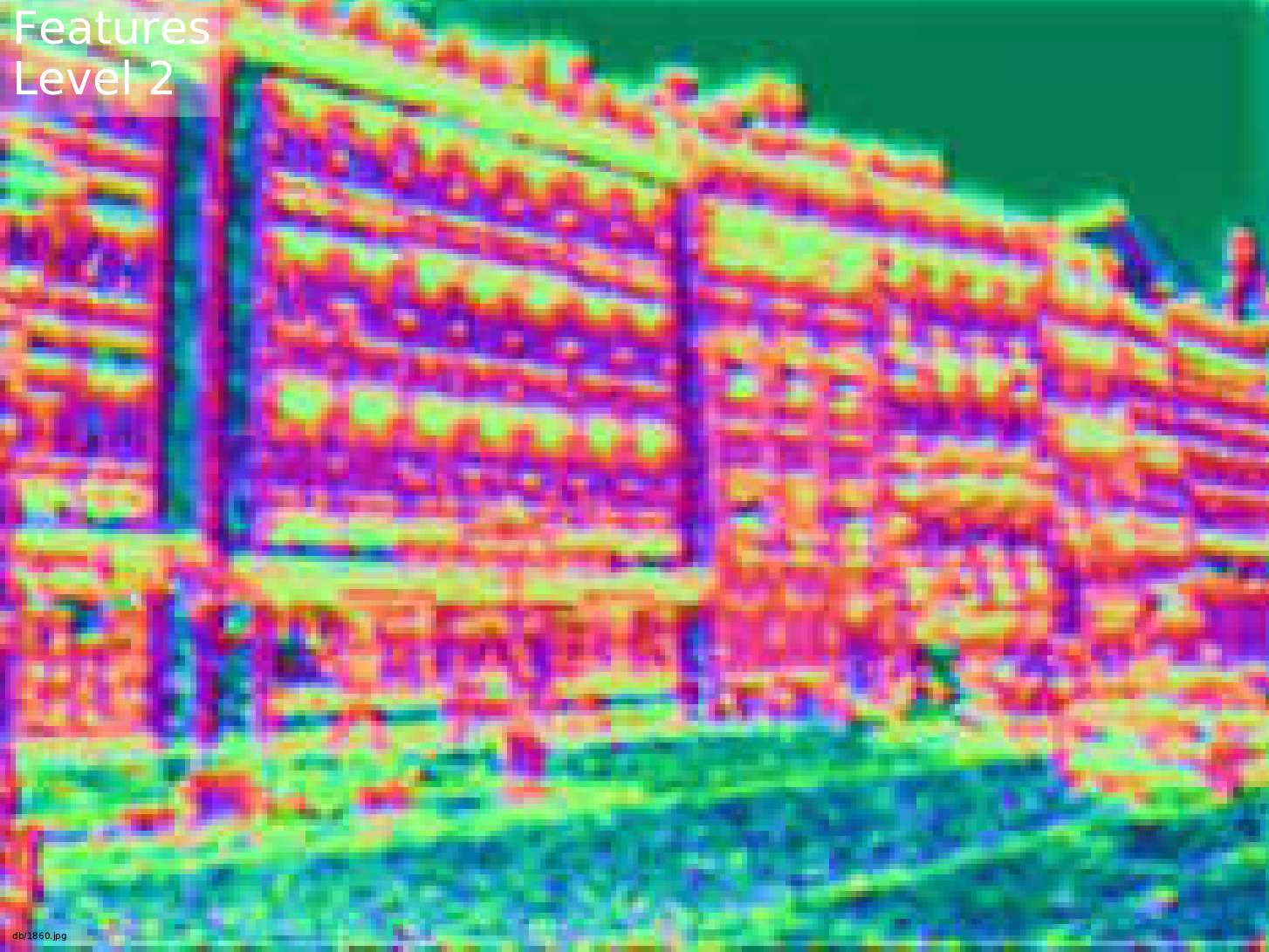}
\end{minipage}%
\begin{minipage}{\iwidth\textwidth}
    \centering
    \includegraphics[width=\pwidth\linewidth]{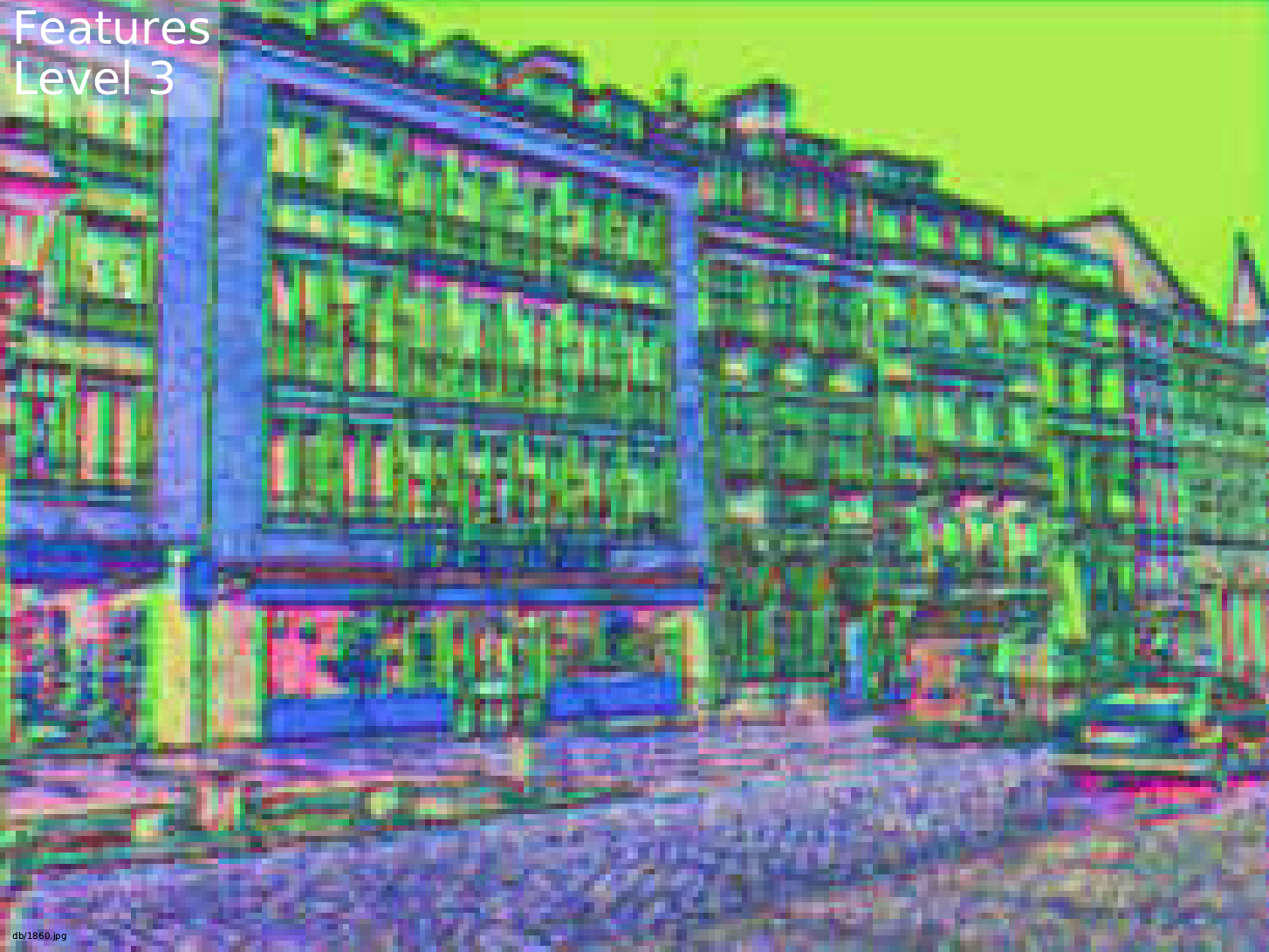}
\end{minipage}%
\begin{minipage}{\iwidth\textwidth}
    \centering
    \includegraphics[width=\pwidth\linewidth]{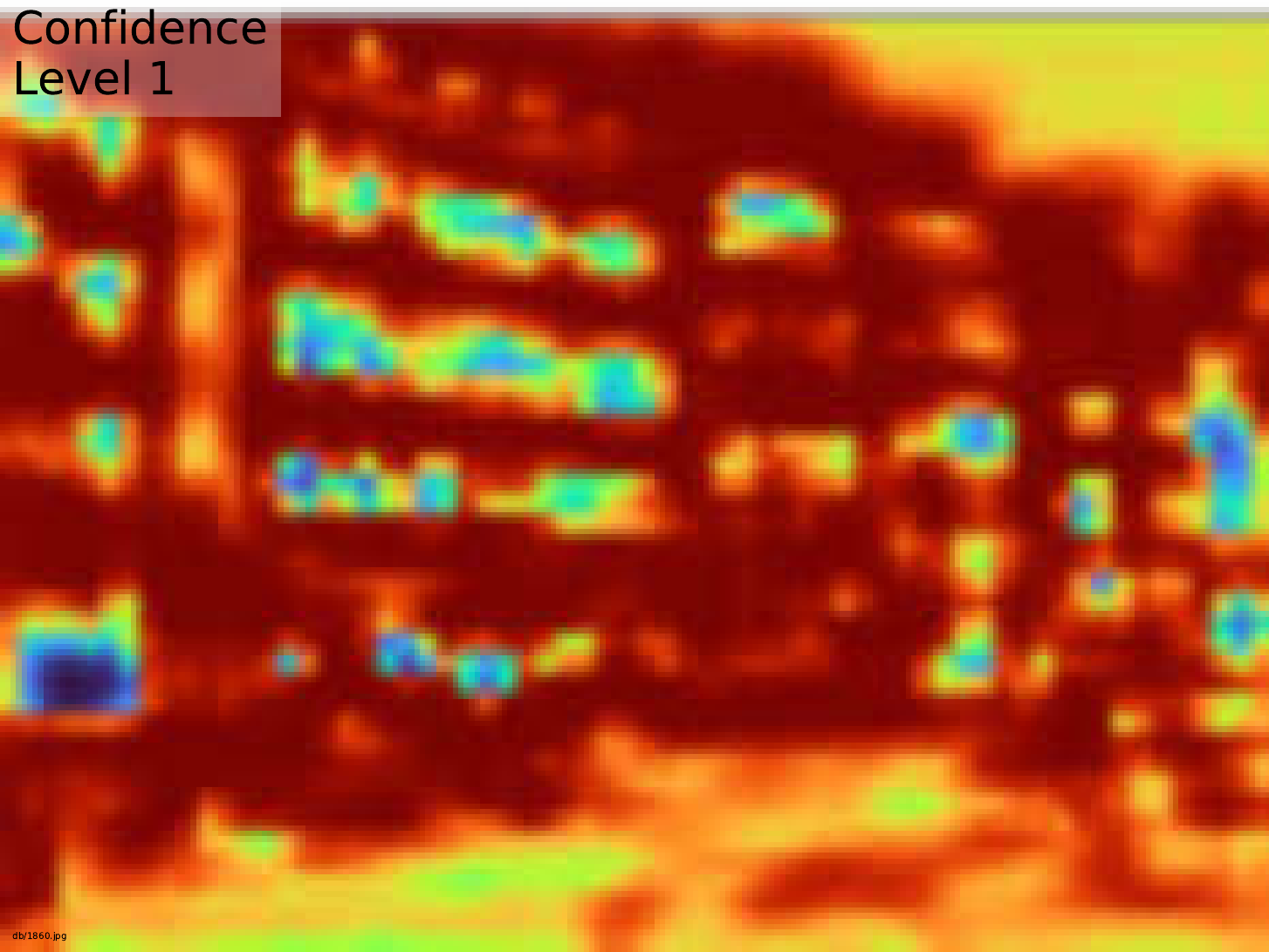}
\end{minipage}%
\begin{minipage}{\iwidth\textwidth}
    \centering
    \includegraphics[width=\pwidth\linewidth]{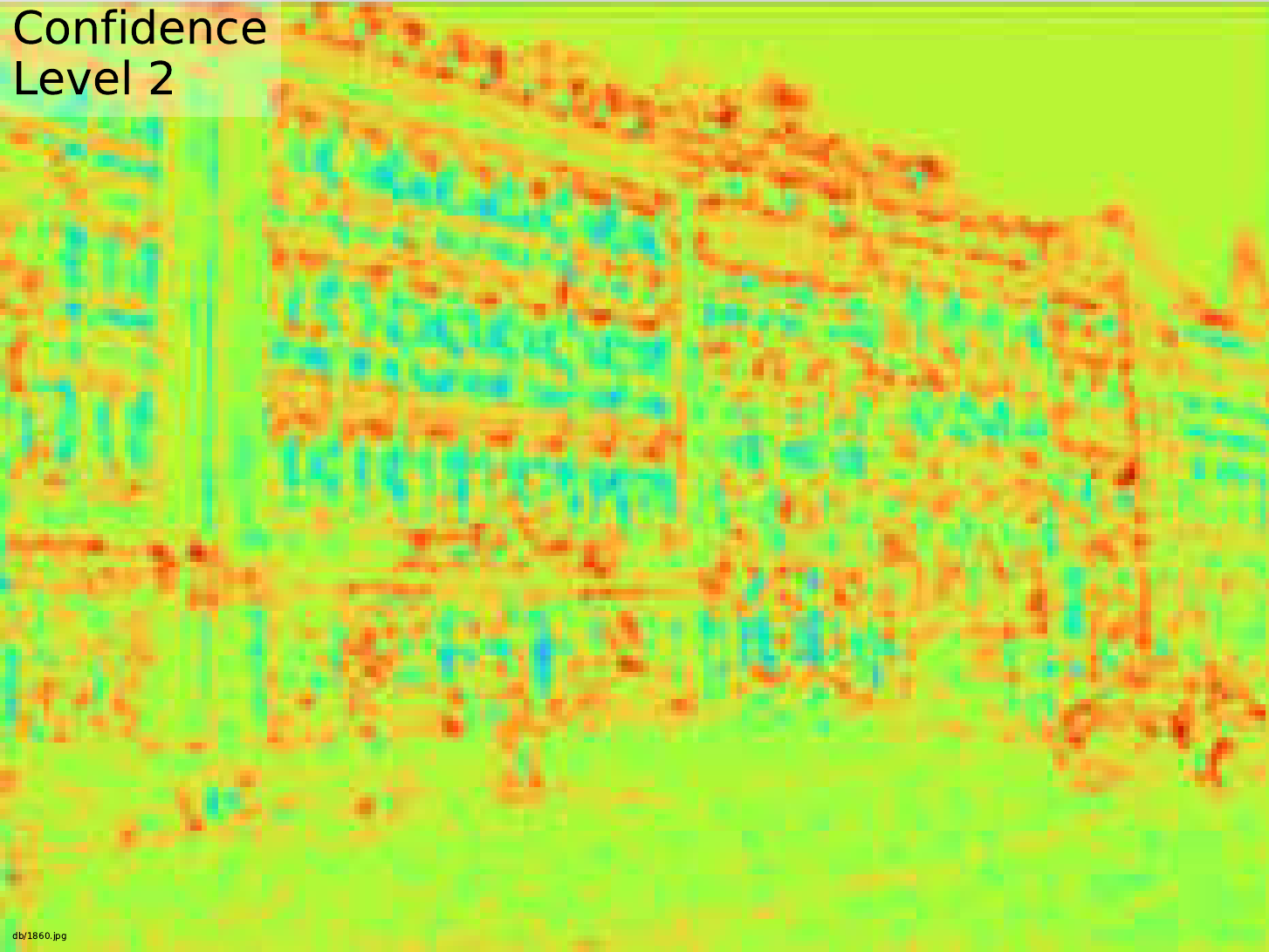}
\end{minipage}%
\begin{minipage}{\iwidth\textwidth}
    \centering
    \includegraphics[width=\pwidth\linewidth]{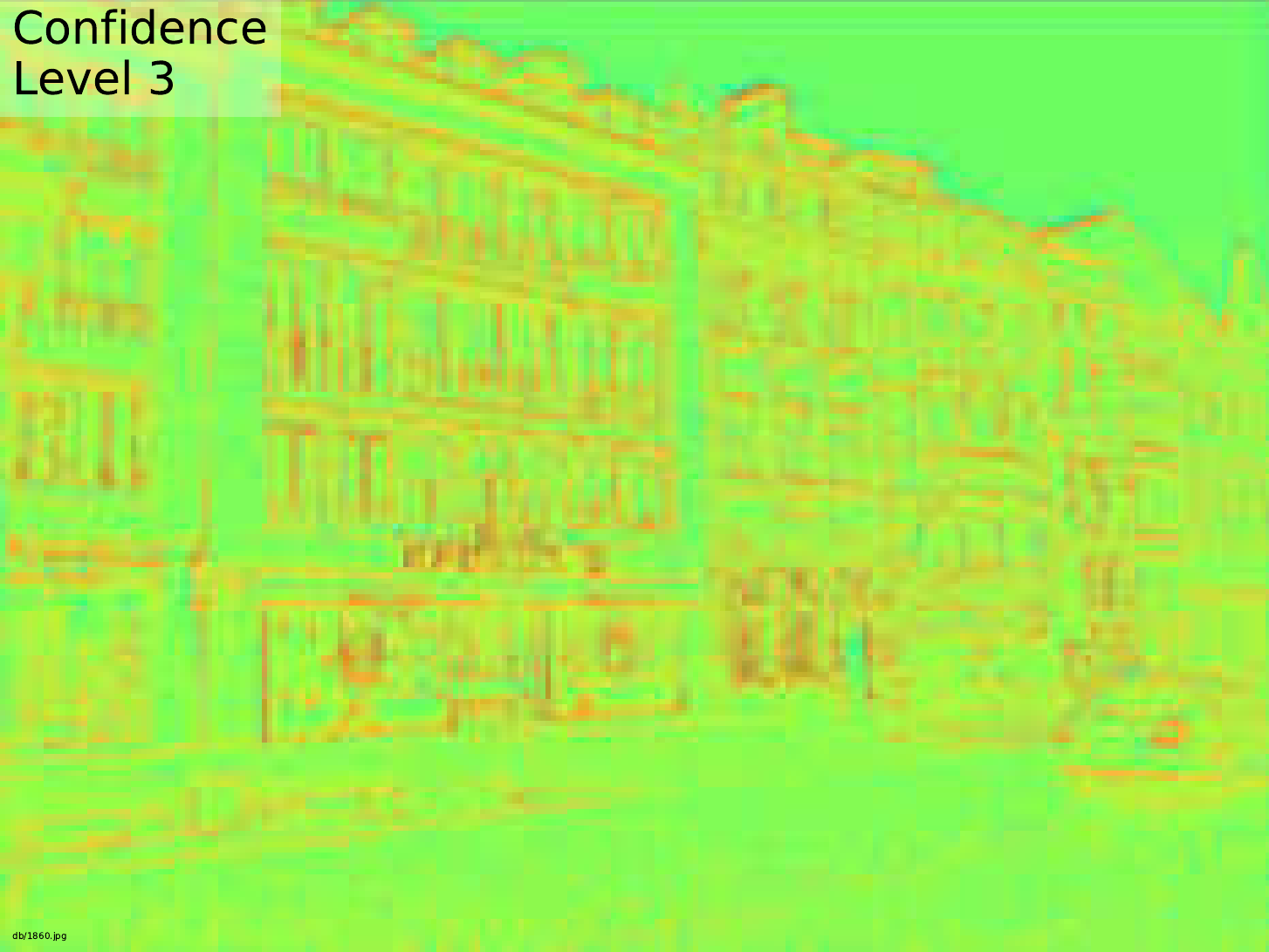}
\end{minipage}
\vspace{2mm}

\begin{minipage}{\lwidth\textwidth}
\rotatebox[origin=c]{90}{Query}
\end{minipage}%
\begin{minipage}{\iwidth\textwidth}
    \centering
    \includegraphics[width=\pwidth\linewidth]{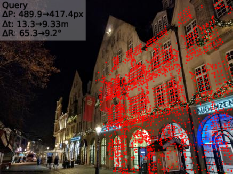}
\end{minipage}%
\begin{minipage}{\iwidth\textwidth}
    \centering
    \includegraphics[width=\pwidth\linewidth]{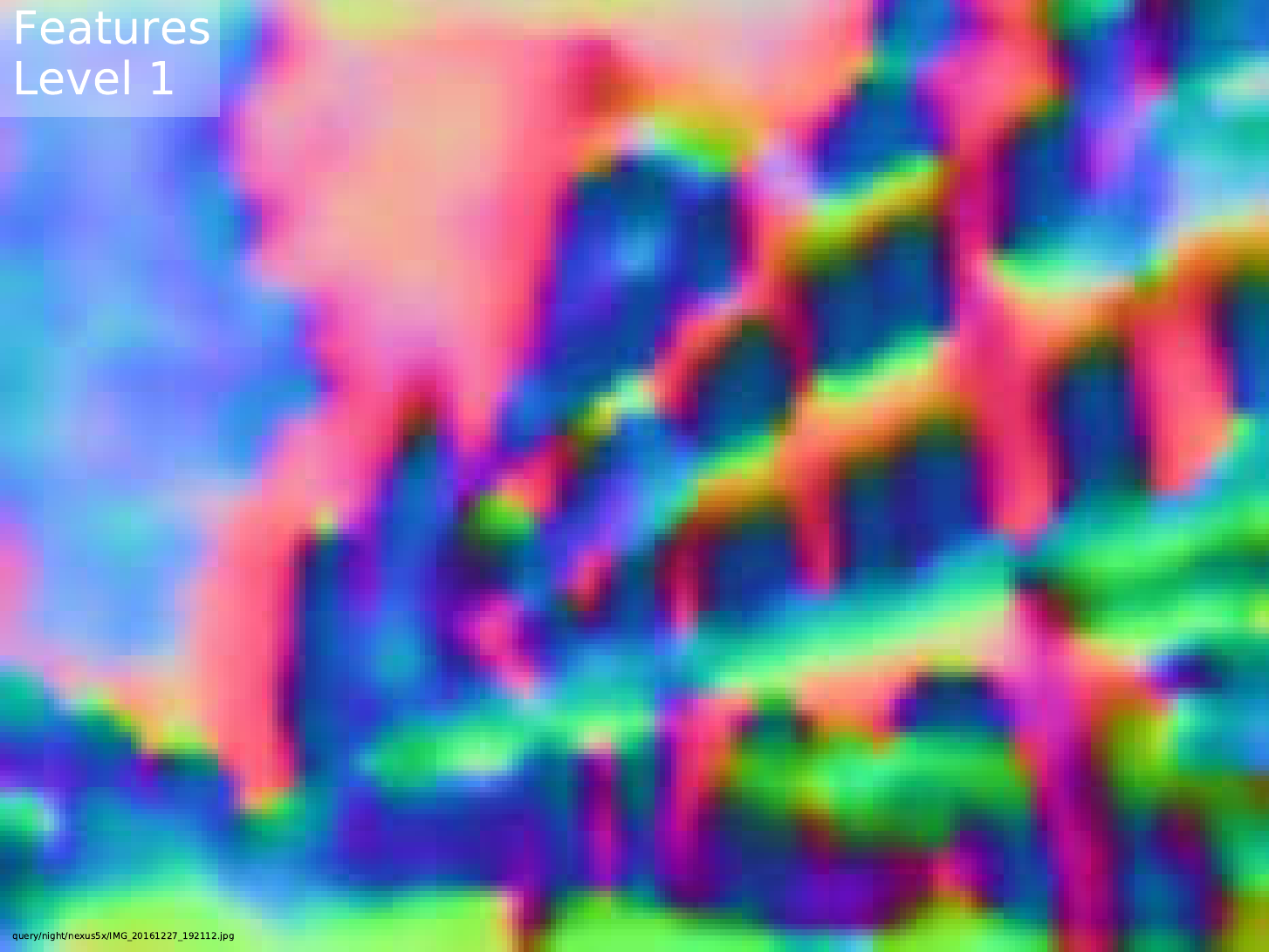}
\end{minipage}%
\begin{minipage}{\iwidth\textwidth}
    \centering
    \includegraphics[width=\pwidth\linewidth]{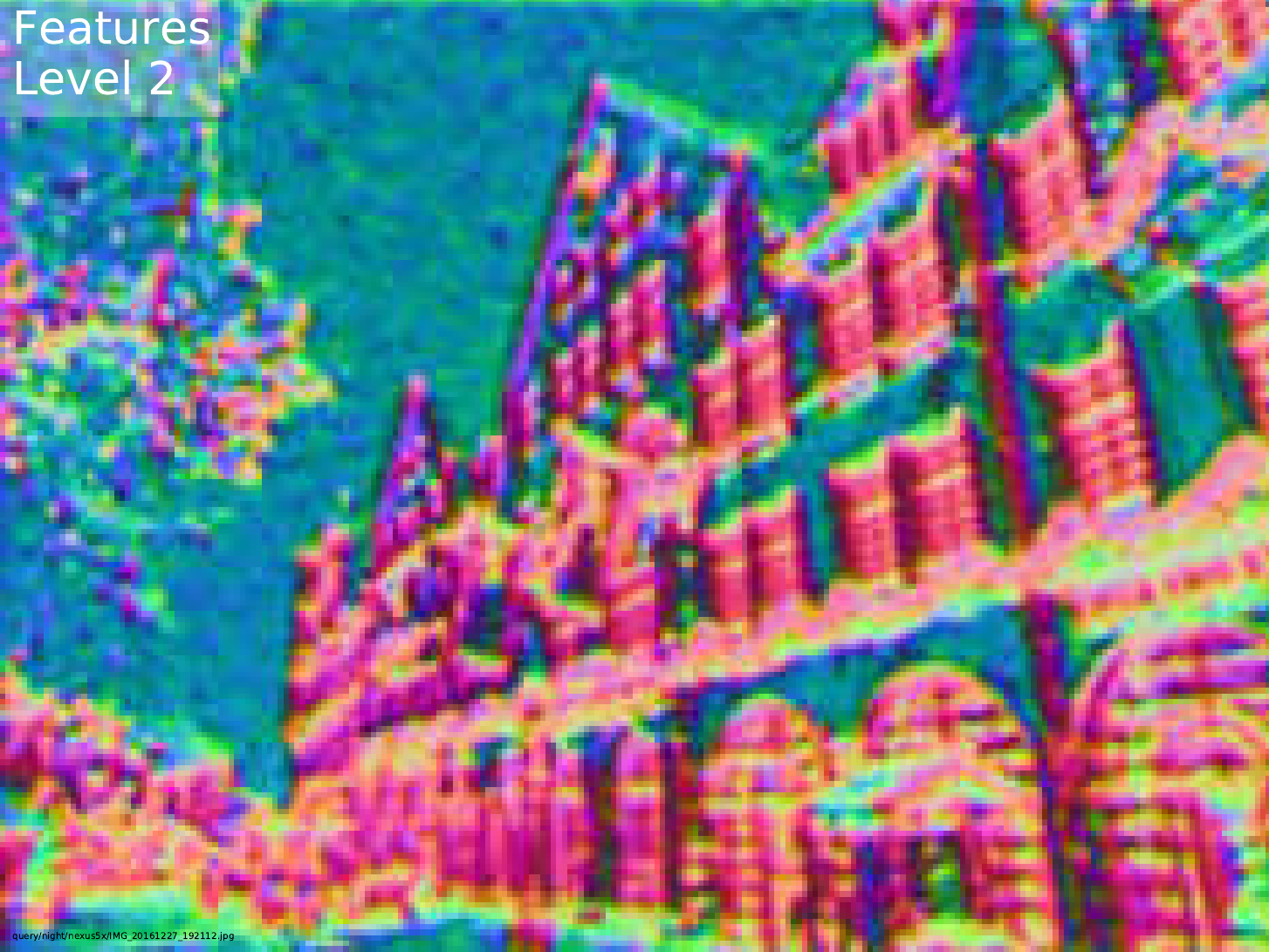}
\end{minipage}%
\begin{minipage}{\iwidth\textwidth}
    \centering
    \includegraphics[width=\pwidth\linewidth]{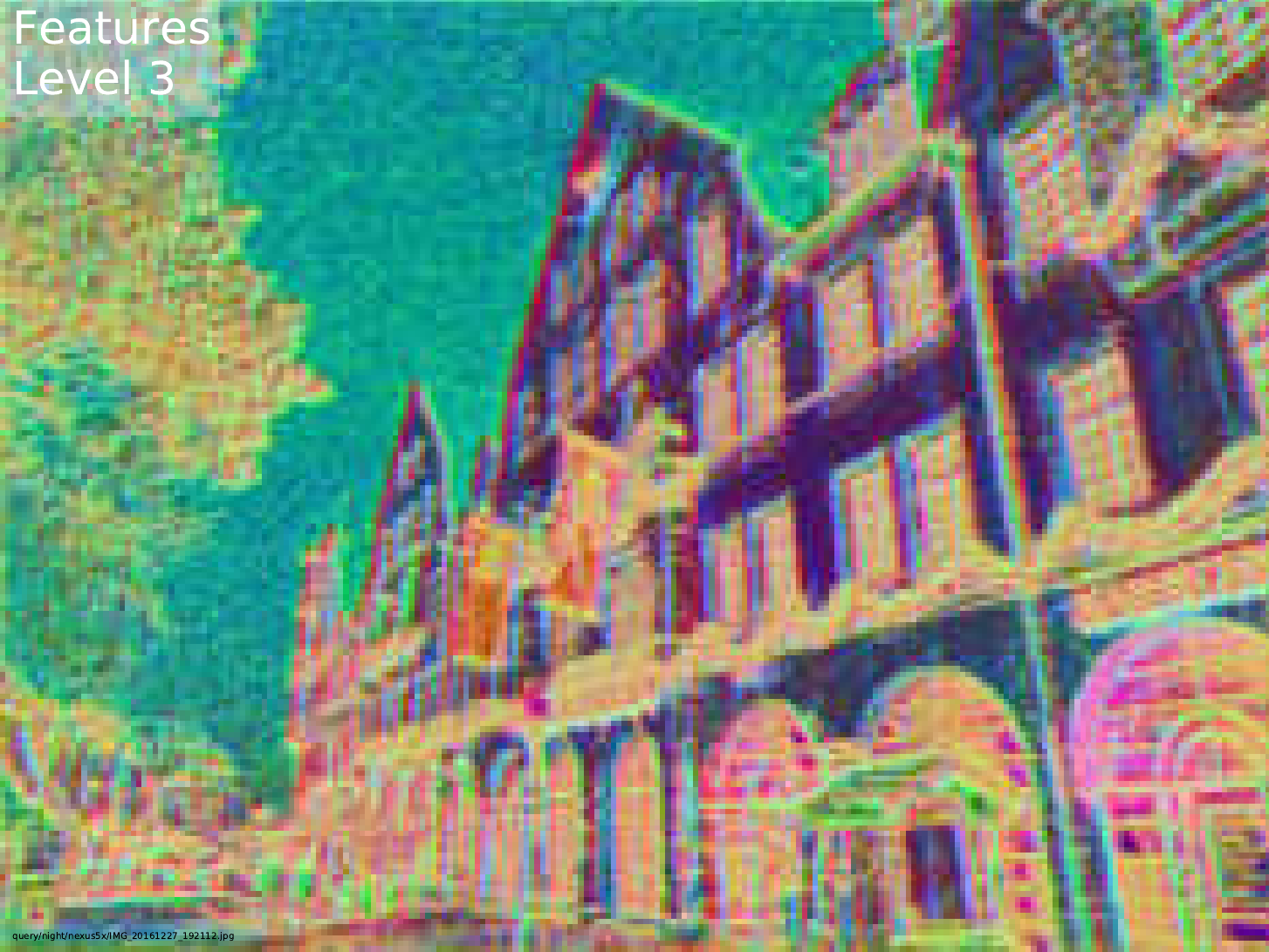}
\end{minipage}%
\begin{minipage}{\iwidth\textwidth}
    \centering
    \includegraphics[width=\pwidth\linewidth]{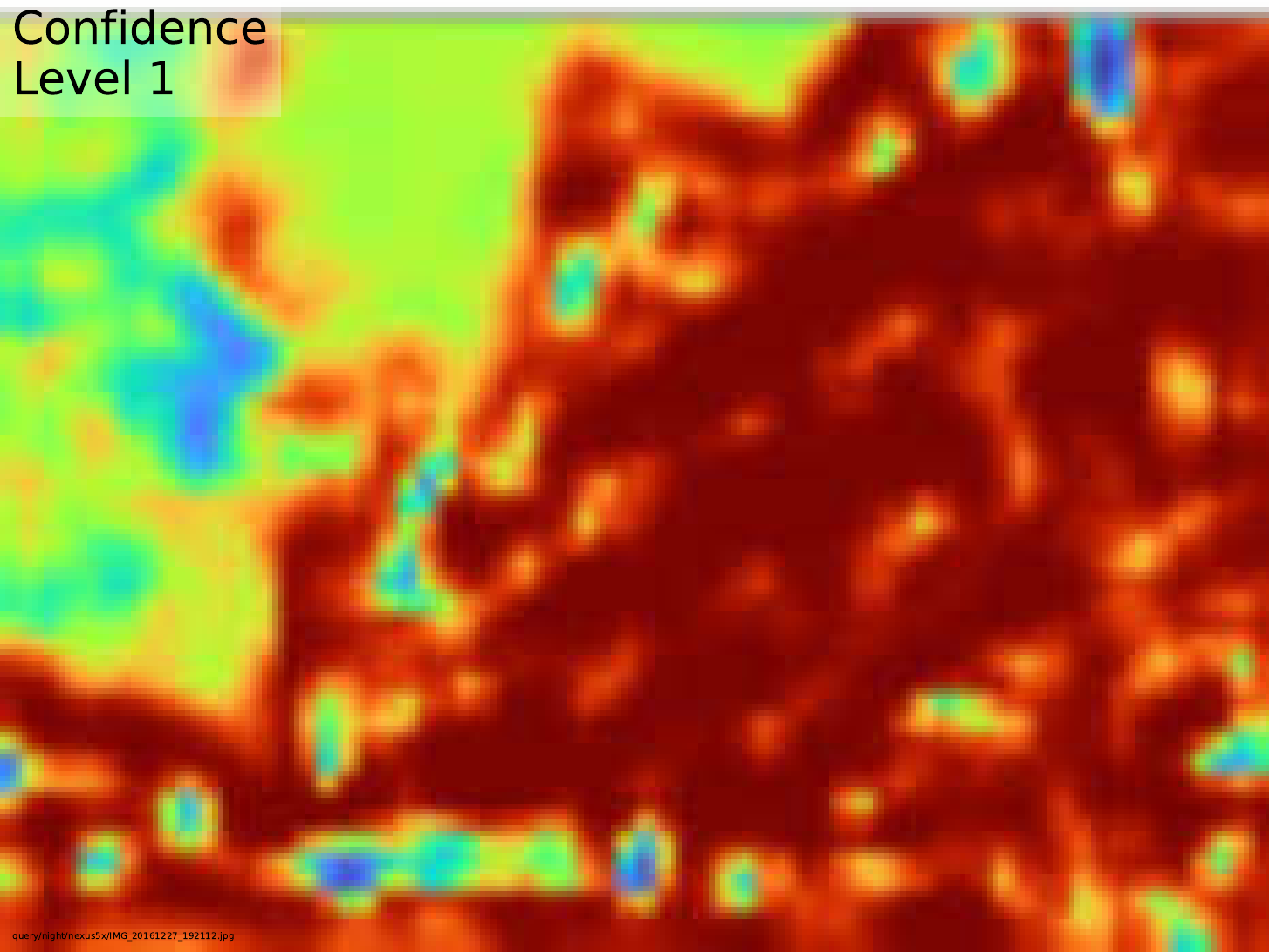}
\end{minipage}%
\begin{minipage}{\iwidth\textwidth}
    \centering
    \includegraphics[width=\pwidth\linewidth]{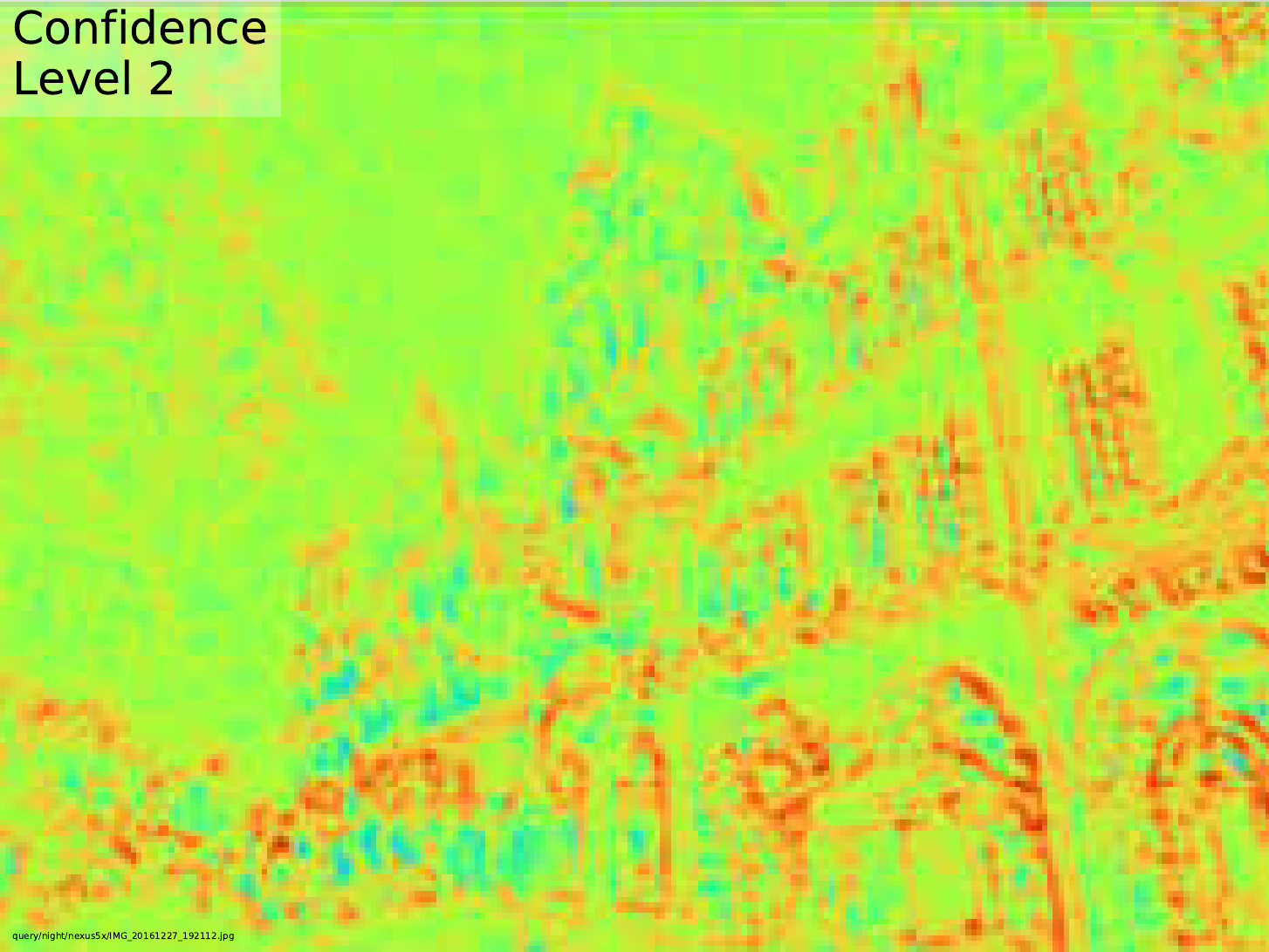}
\end{minipage}%
\begin{minipage}{\iwidth\textwidth}
    \centering
    \includegraphics[width=\pwidth\linewidth]{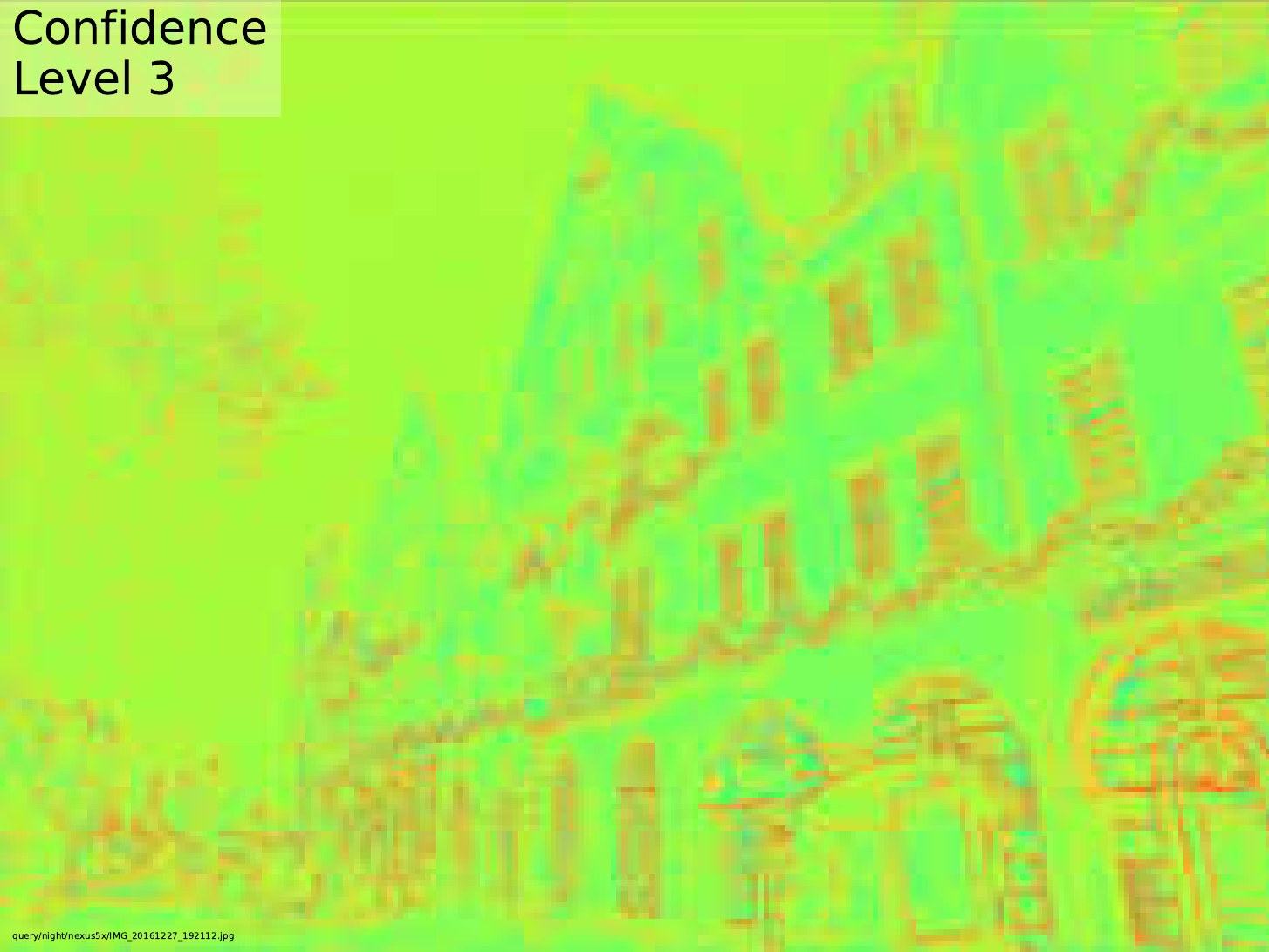}
\end{minipage}
\begin{minipage}{\lwidth\textwidth}
\rotatebox[origin=c]{90}{Reference}
\end{minipage}%
\begin{minipage}{\iwidth\textwidth}
    \centering
    \includegraphics[width=\pwidth\linewidth]{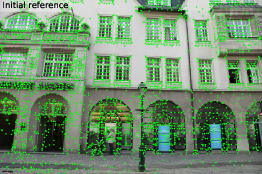}
\end{minipage}%
\begin{minipage}{\iwidth\textwidth}
    \centering
    \includegraphics[width=\pwidth\linewidth]{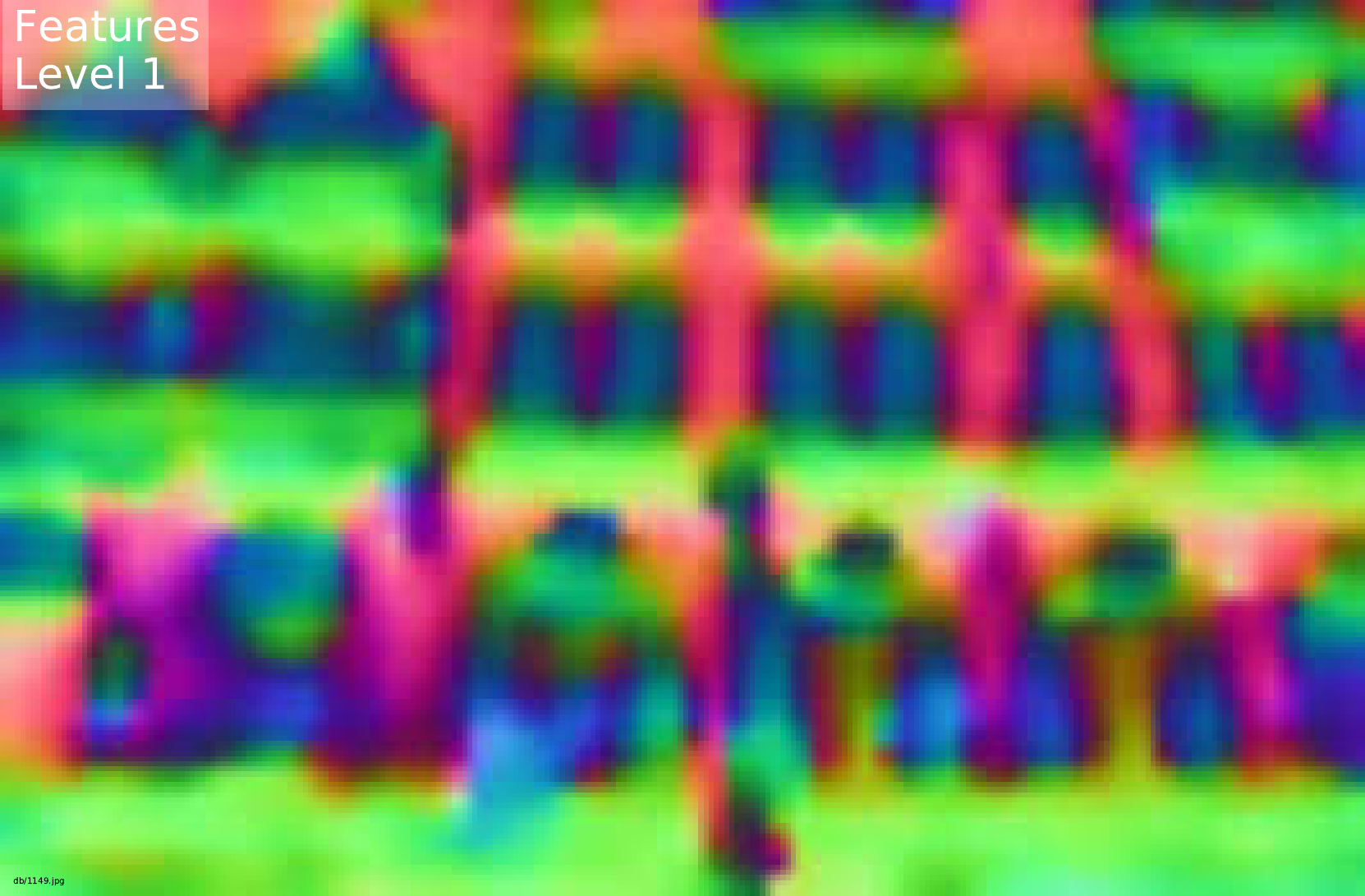}
\end{minipage}%
\begin{minipage}{\iwidth\textwidth}
    \centering
    \includegraphics[width=\pwidth\linewidth]{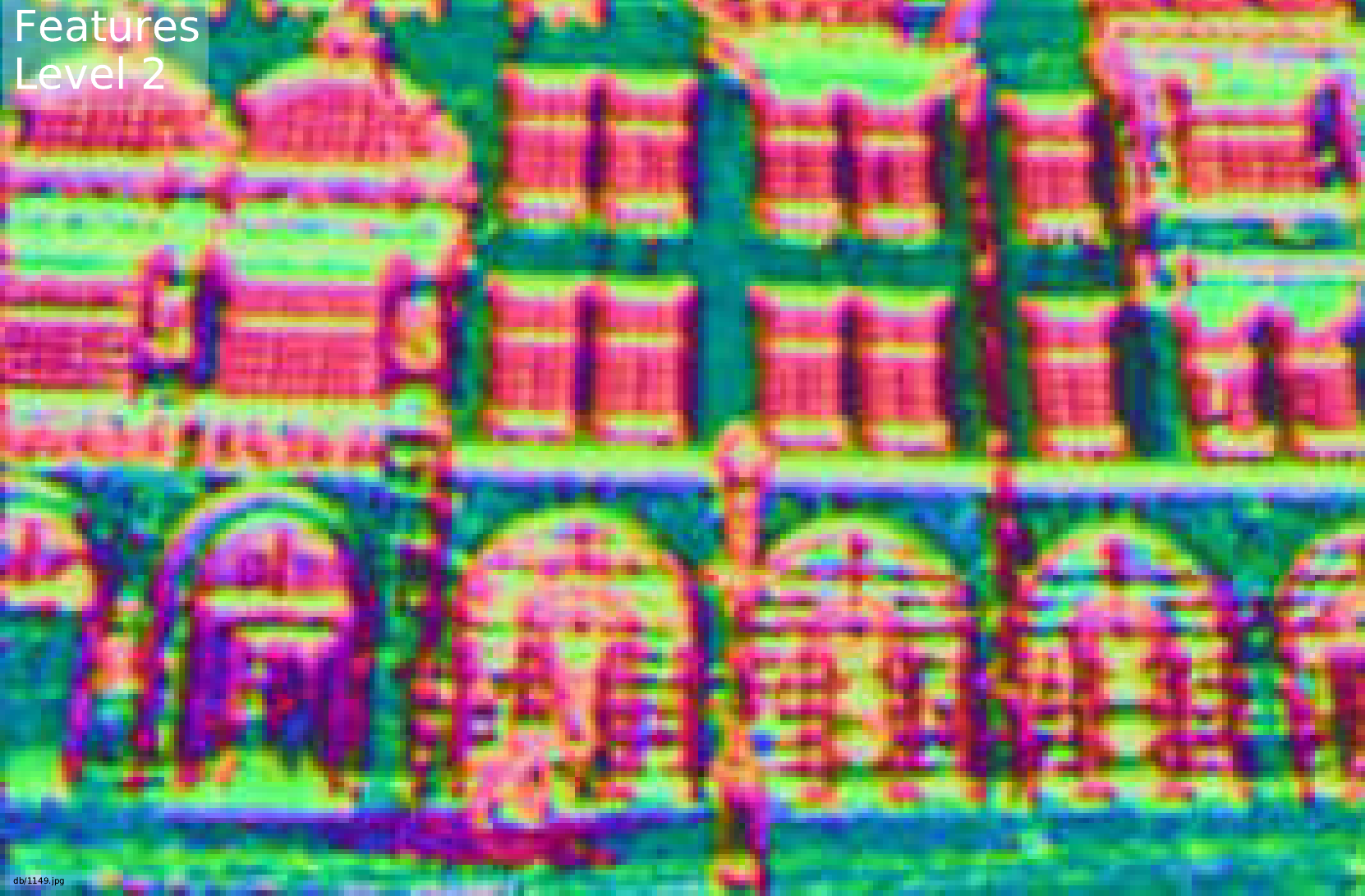}
\end{minipage}%
\begin{minipage}{\iwidth\textwidth}
    \centering
    \includegraphics[width=\pwidth\linewidth]{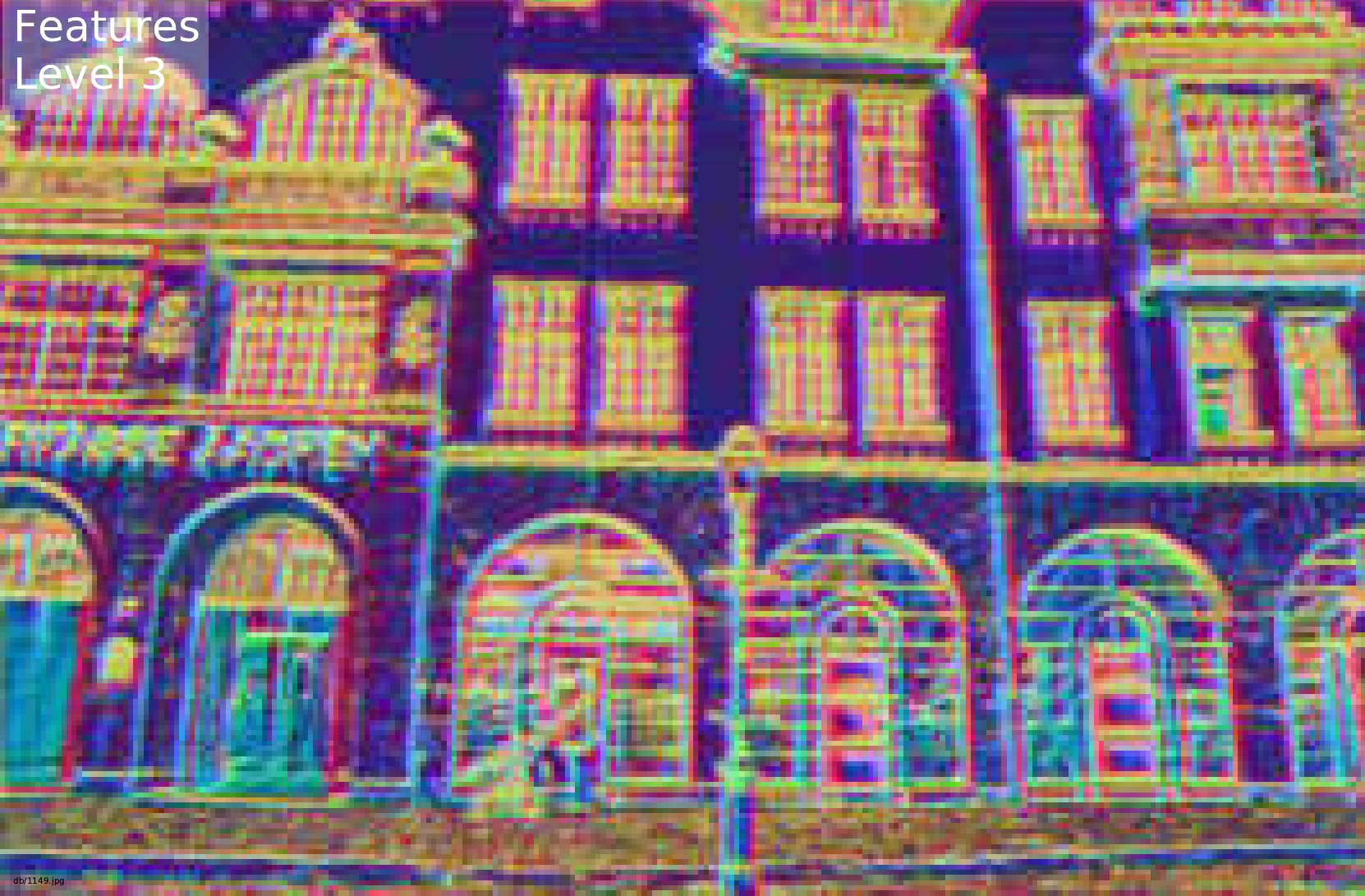}
\end{minipage}%
\begin{minipage}{\iwidth\textwidth}
    \centering
    \includegraphics[width=\pwidth\linewidth]{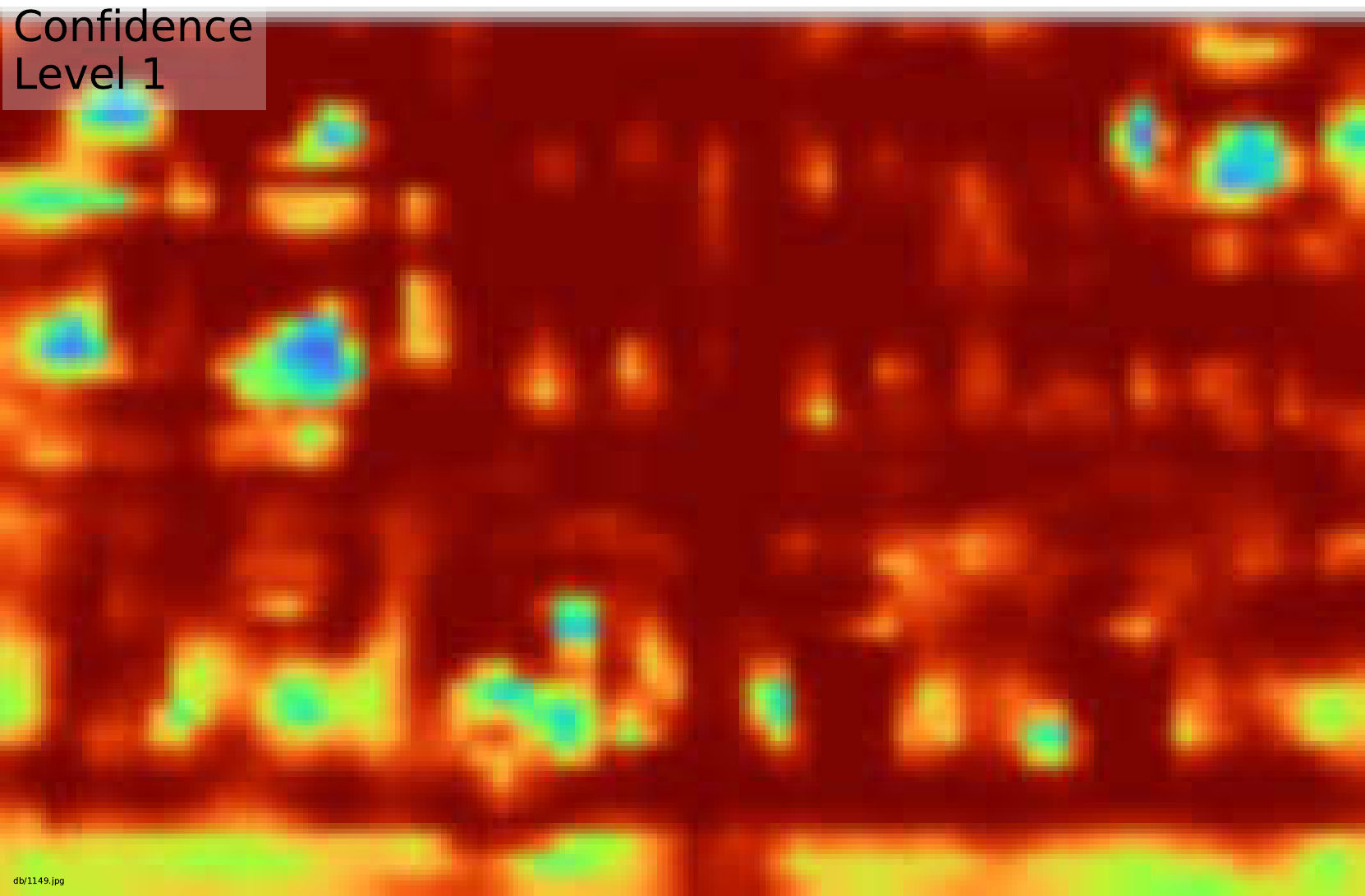}
\end{minipage}%
\begin{minipage}{\iwidth\textwidth}
    \centering
    \includegraphics[width=\pwidth\linewidth]{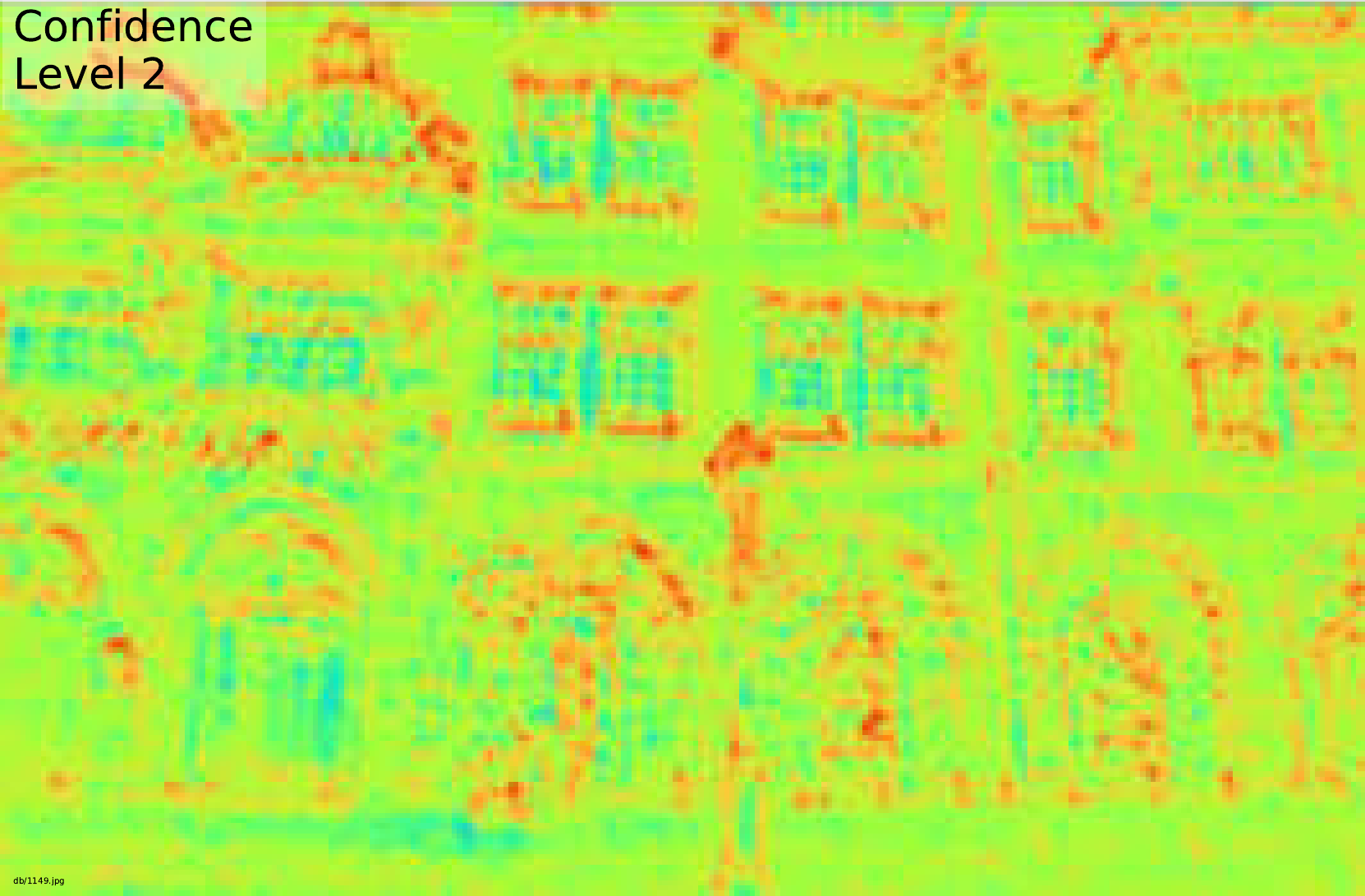}
\end{minipage}%
\begin{minipage}{\iwidth\textwidth}
    \centering
    \includegraphics[width=\pwidth\linewidth]{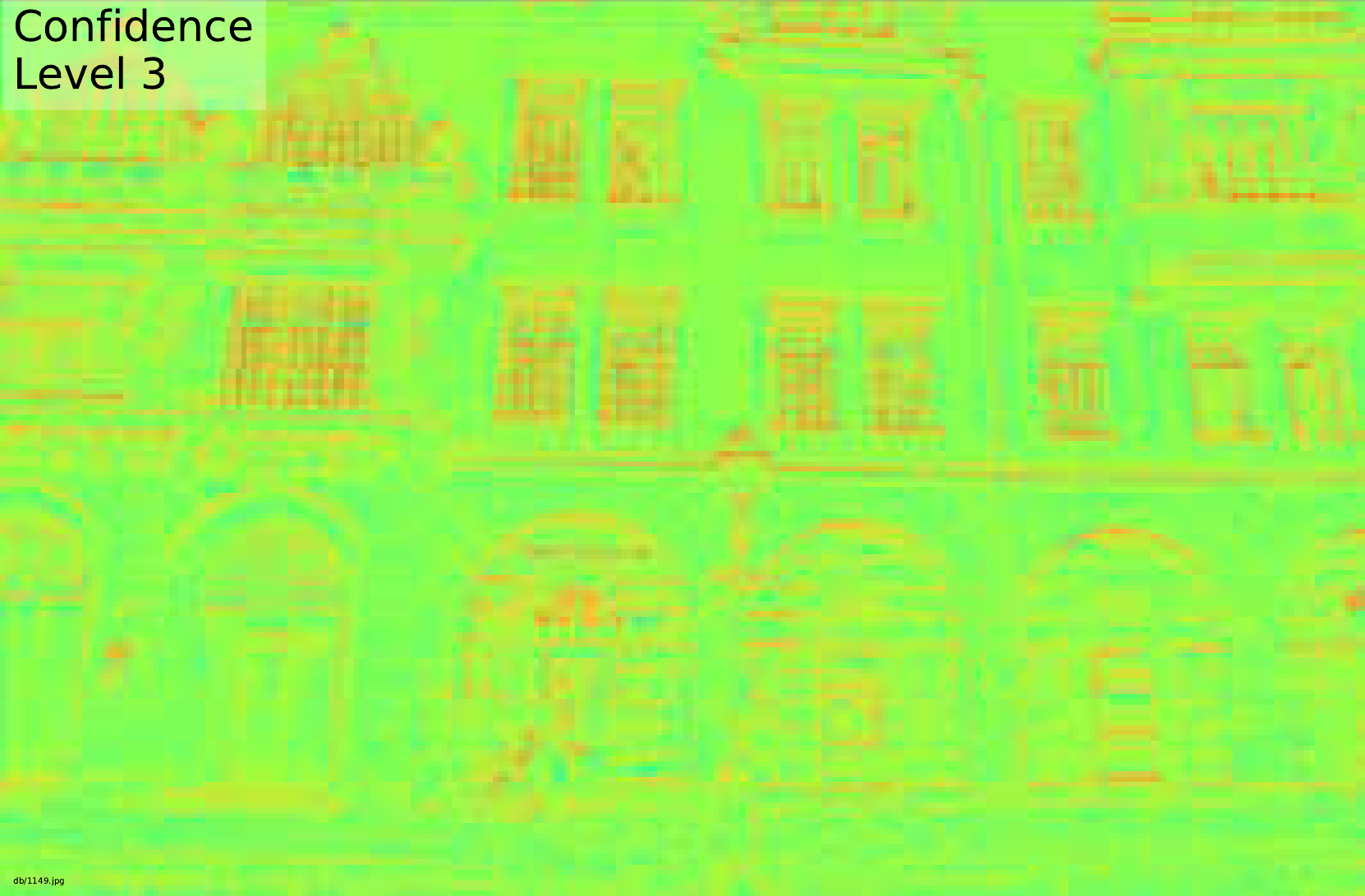}
\end{minipage}
\vspace{2mm}

    \caption{\textbf{Failure cases on the Aachen dataset.}
    Convergence to a local and incorrect minima can be due to large appearance changes (row 1), occlusion (row 2), large viewpoint change (row 3) or repeated structures on facades (rows 4 and 5).
    }
    \label{fig:qualitative_aachen_failure}%
\end{figure*}

\begin{figure*}[t]
    \centering
    \def\iwidth{0.23}
    \begin{minipage}{\iwidth\linewidth}
        \includegraphics[width=\linewidth]{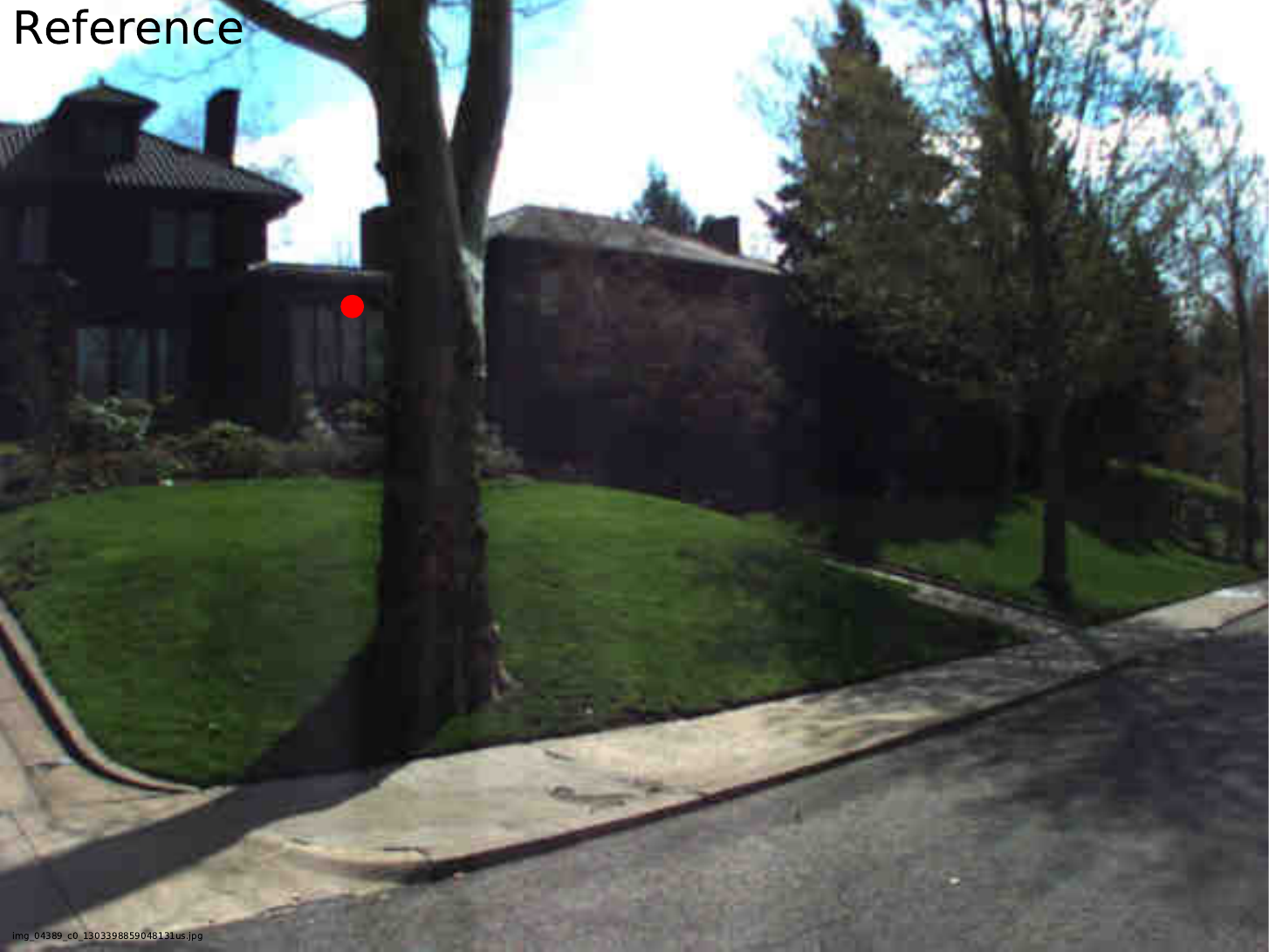}
        
        \vspace{2mm}
        \includegraphics[width=\linewidth]{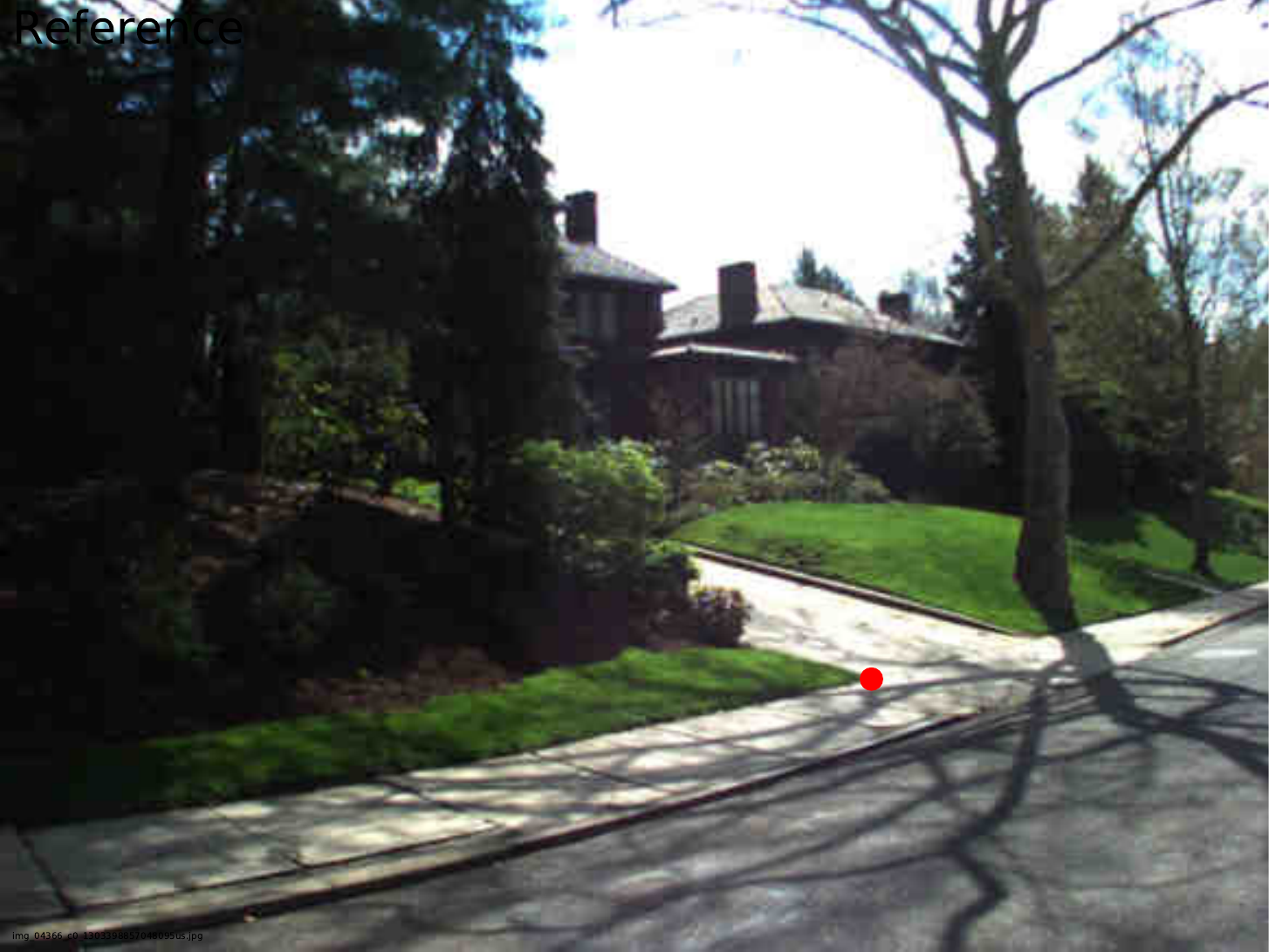}
        \vspace{2mm}
        \includegraphics[width=\linewidth]{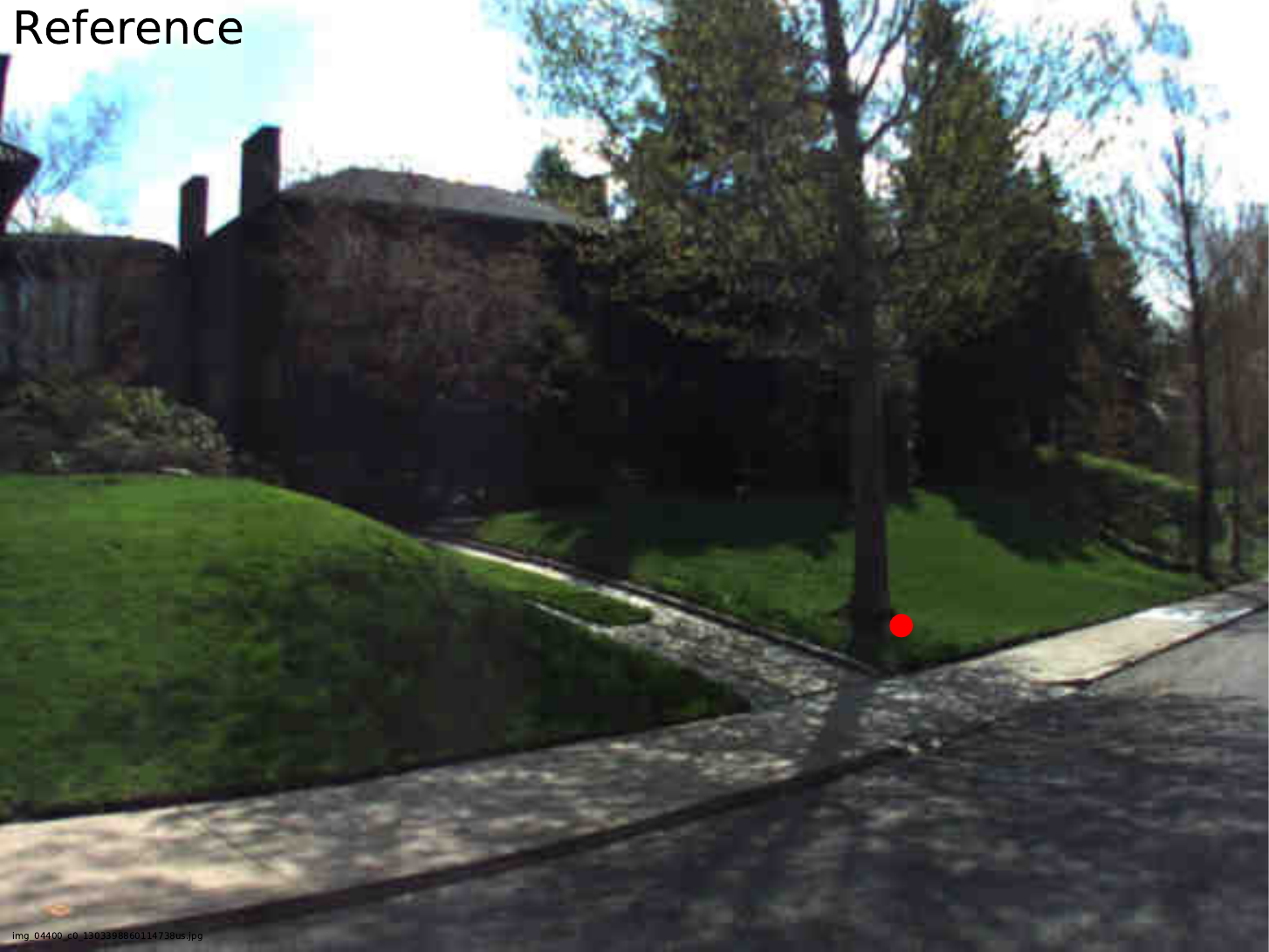}
        \vspace{2mm}
        \includegraphics[width=\linewidth]{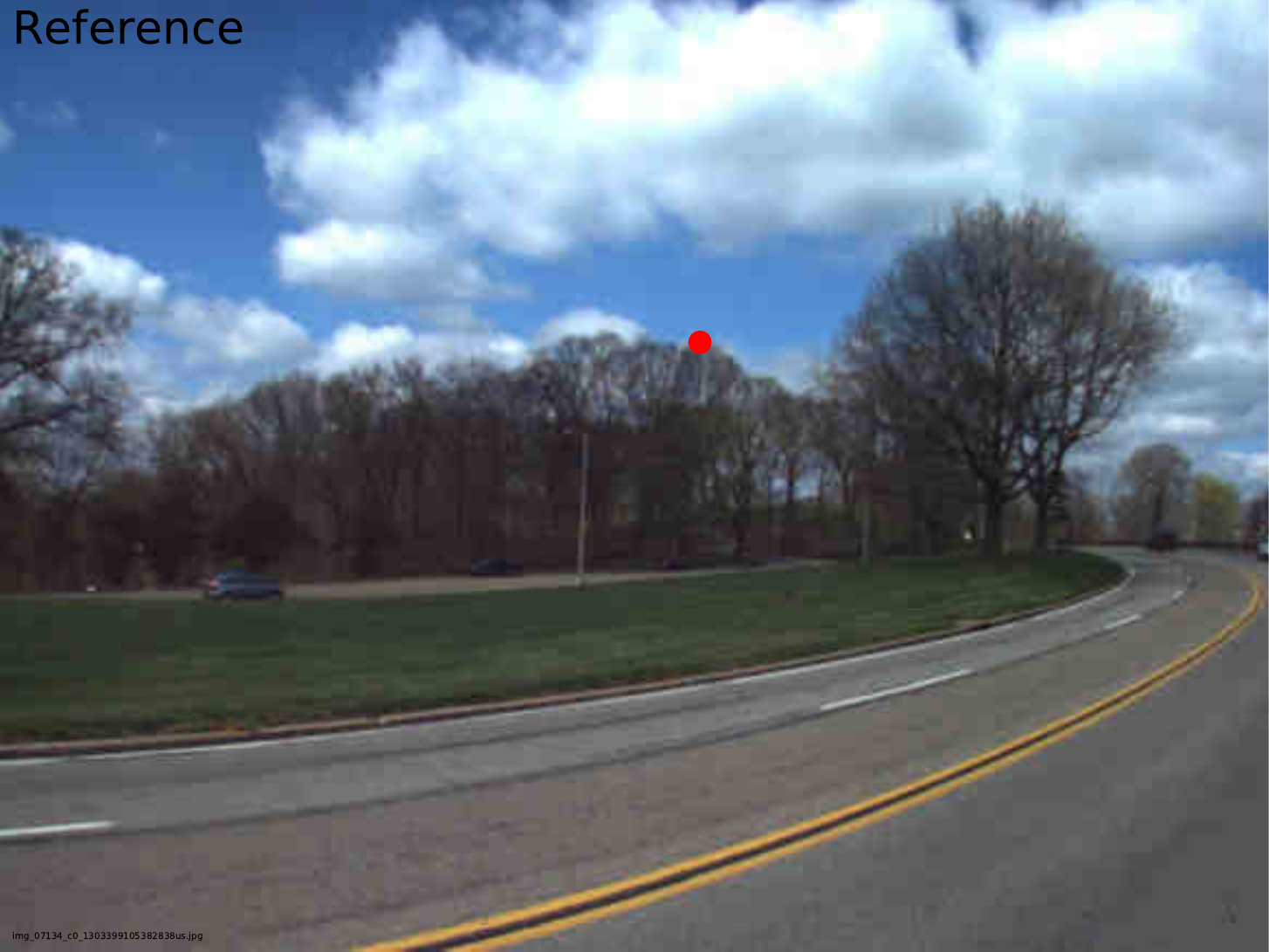}
    \end{minipage}%
    \begin{minipage}{\iwidth\linewidth}
        \includegraphics[width=\linewidth]{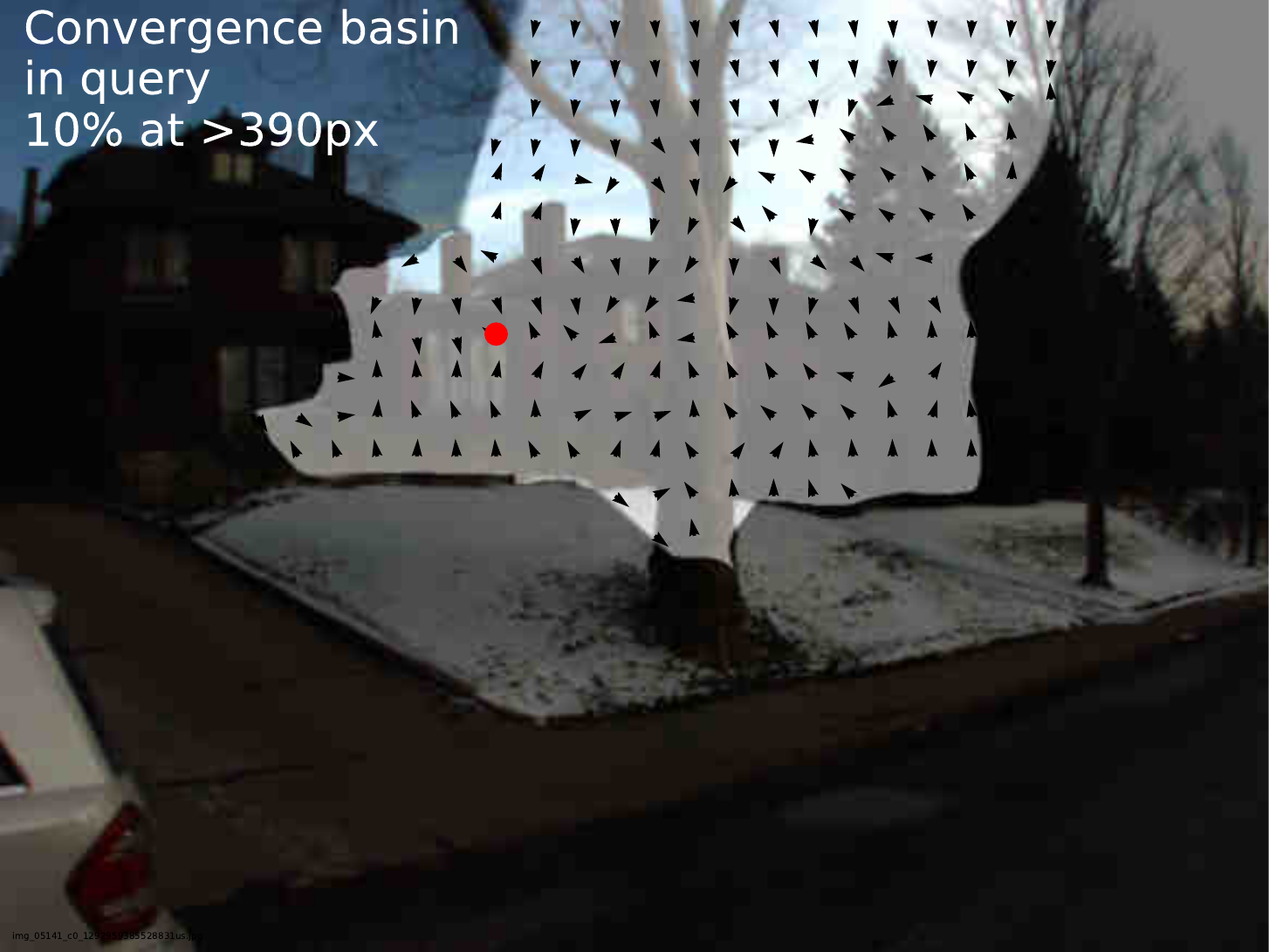}
        
        \vspace{2mm}
        \includegraphics[width=\linewidth]{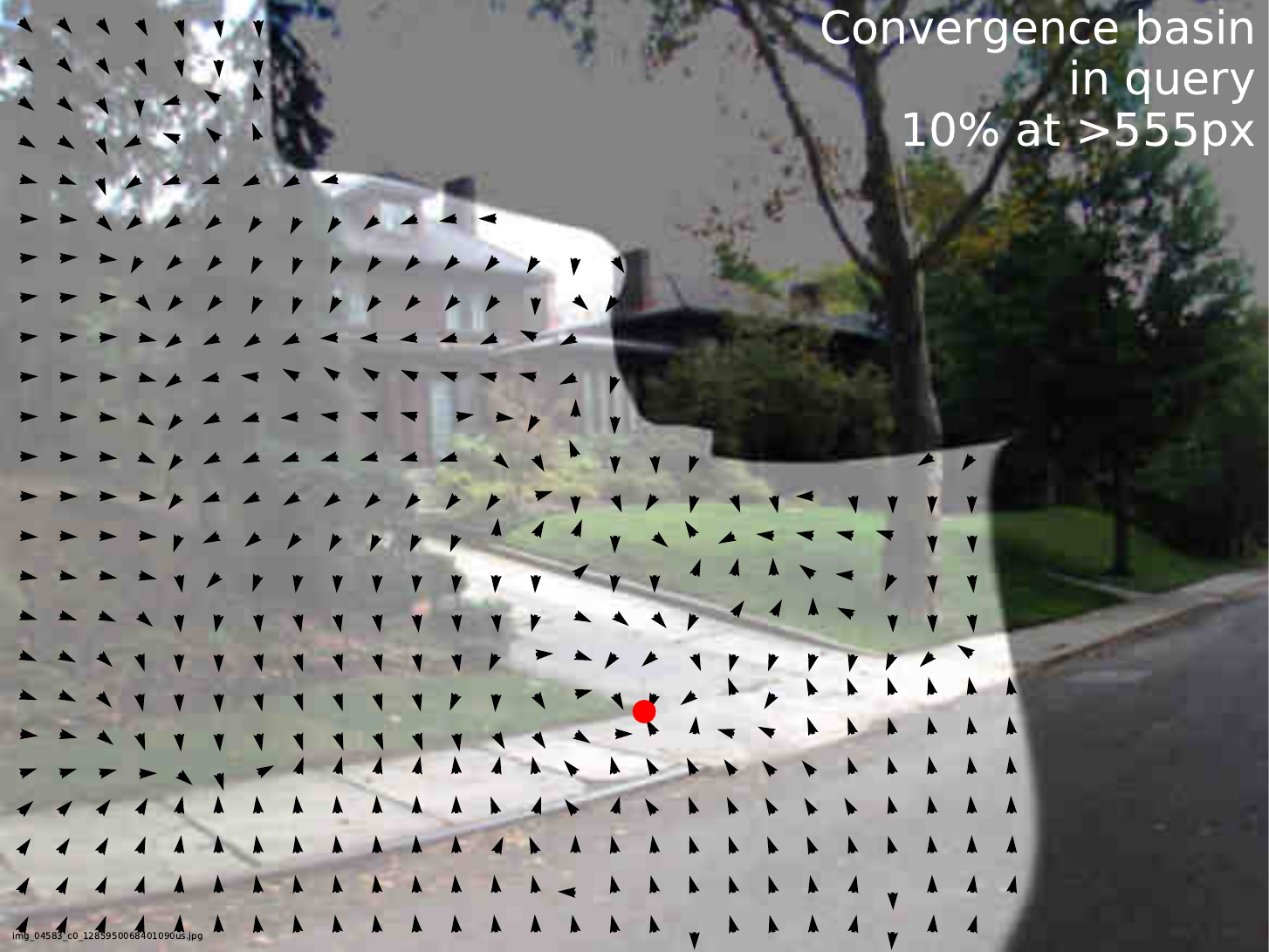}
        \vspace{2mm}
        \includegraphics[width=\linewidth]{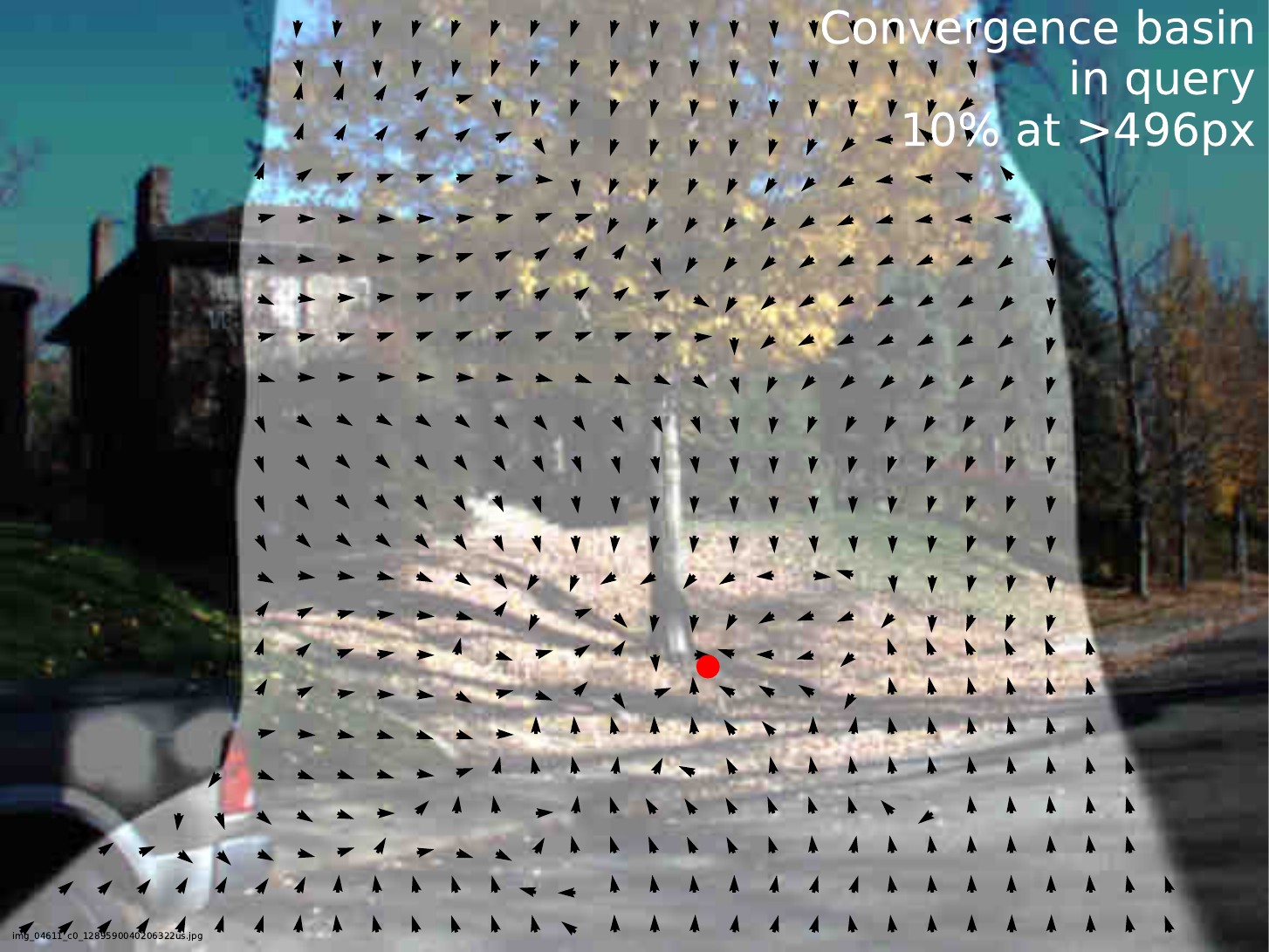}
        \vspace{2mm}
        \includegraphics[width=\linewidth]{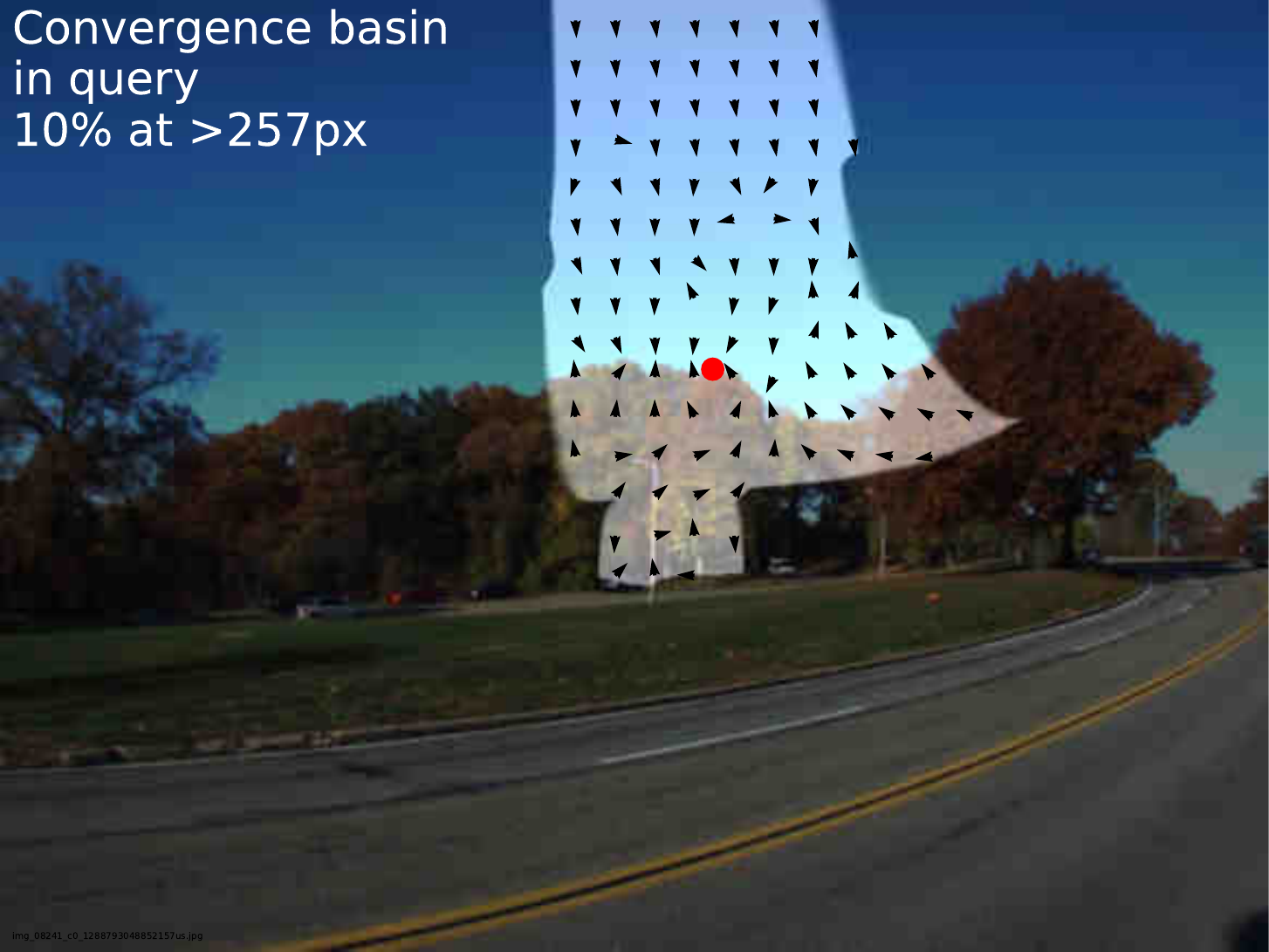}
    \end{minipage}%
    \hspace{2mm}
    \begin{minipage}{\iwidth\linewidth}
        \includegraphics[width=\linewidth]{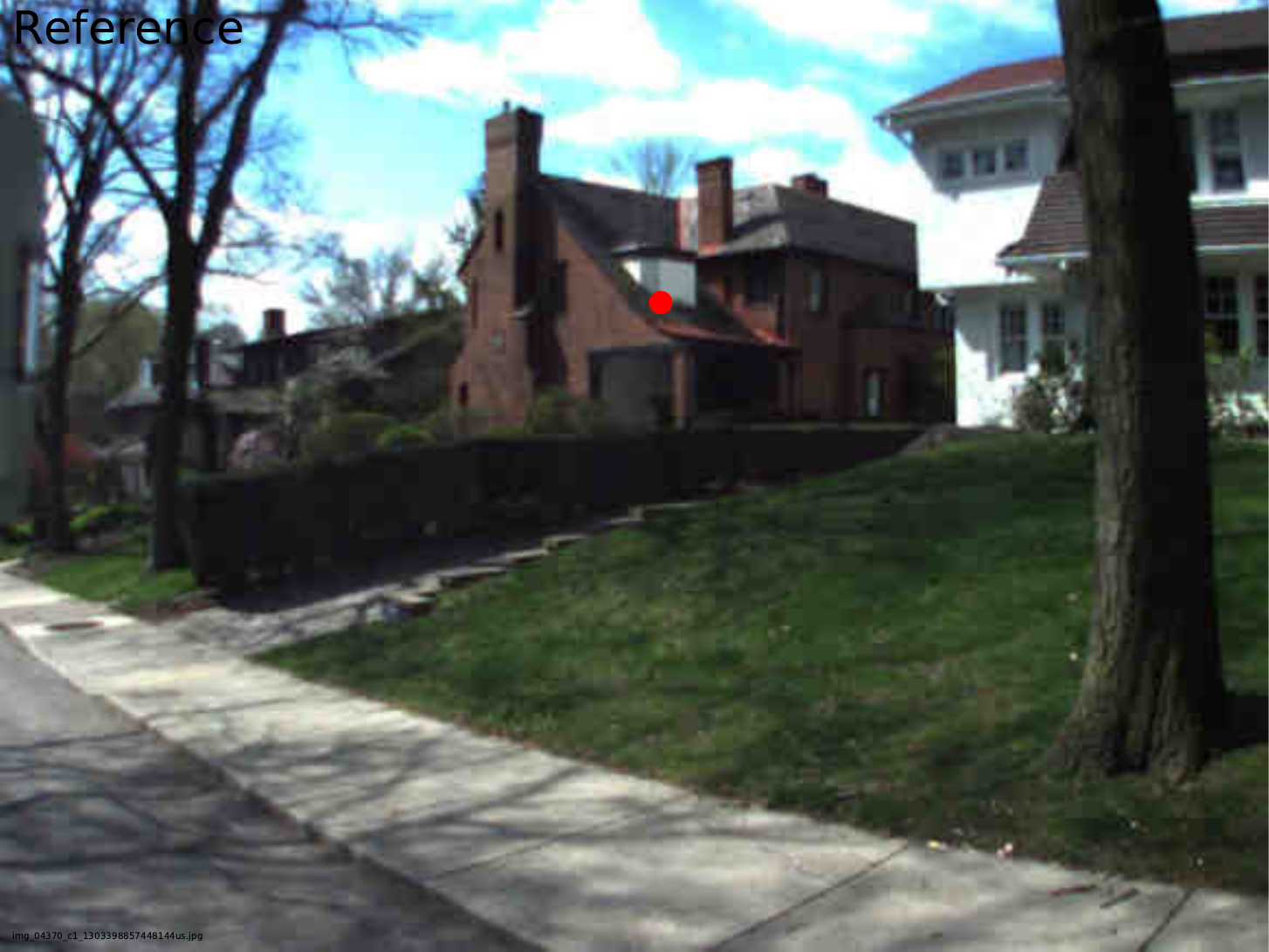}
        
        \vspace{2mm}
        \includegraphics[width=\linewidth]{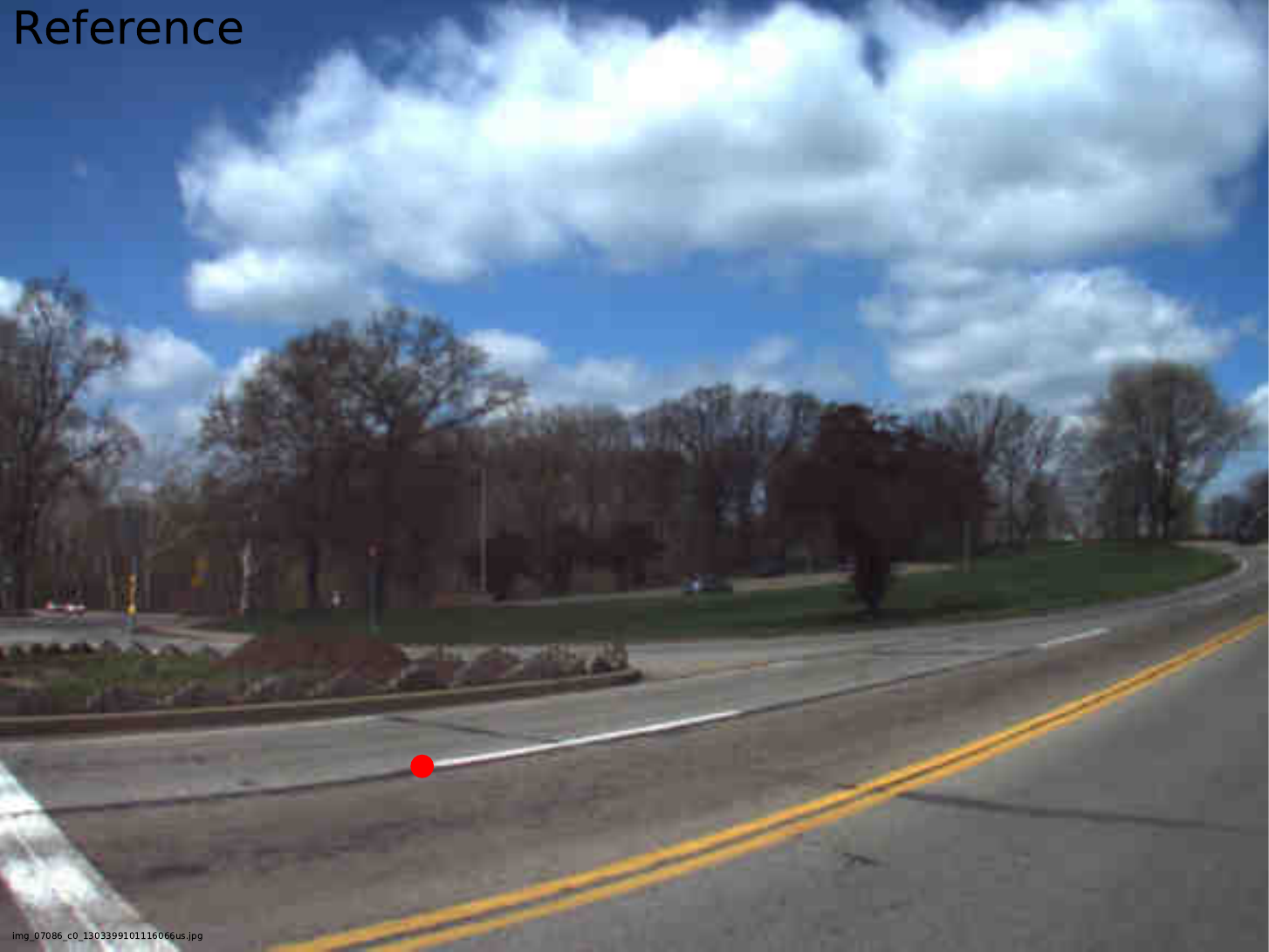}
        \vspace{2mm}
        \includegraphics[width=\linewidth]{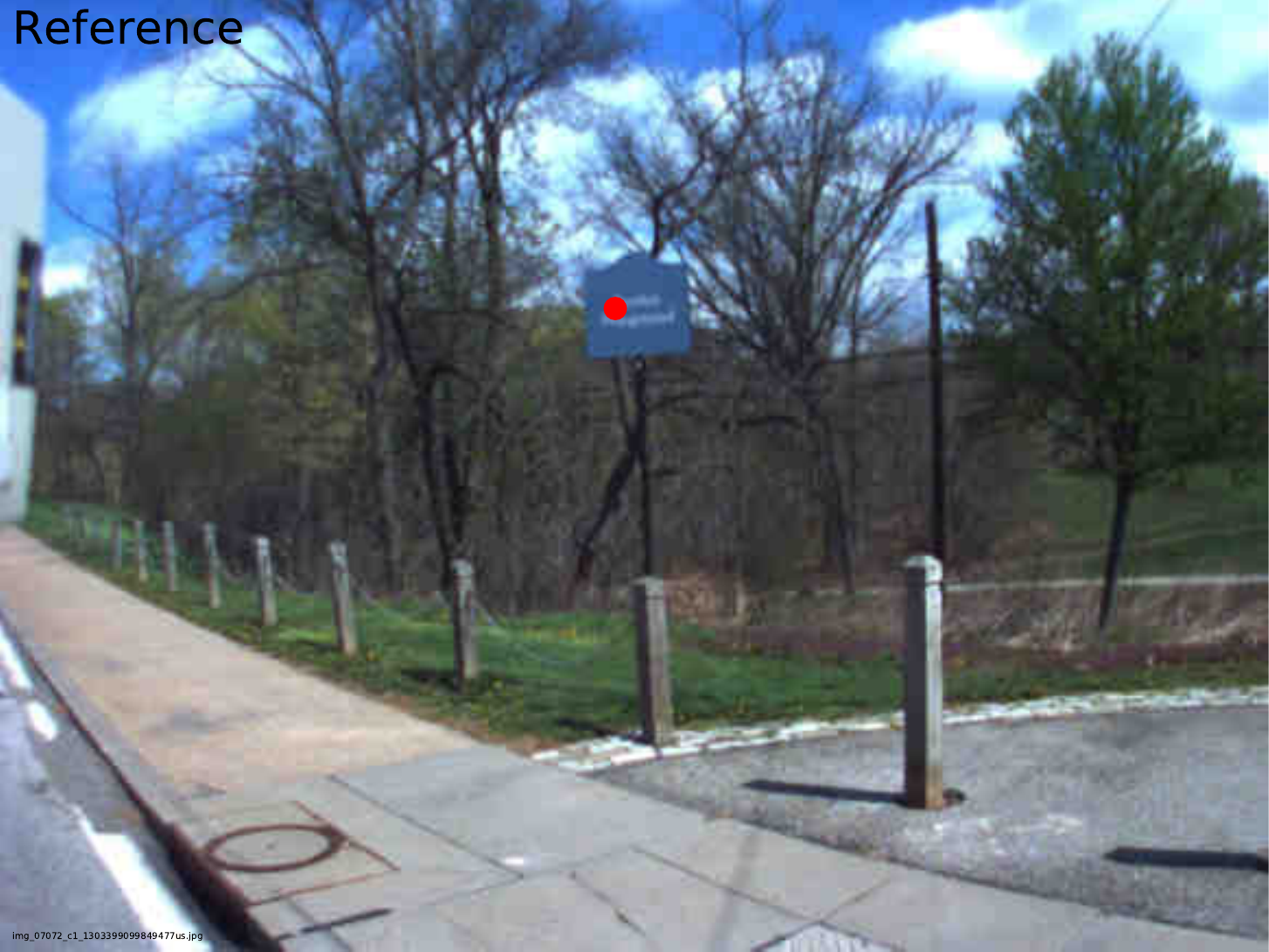}
        \vspace{2mm}
        \includegraphics[width=\linewidth]{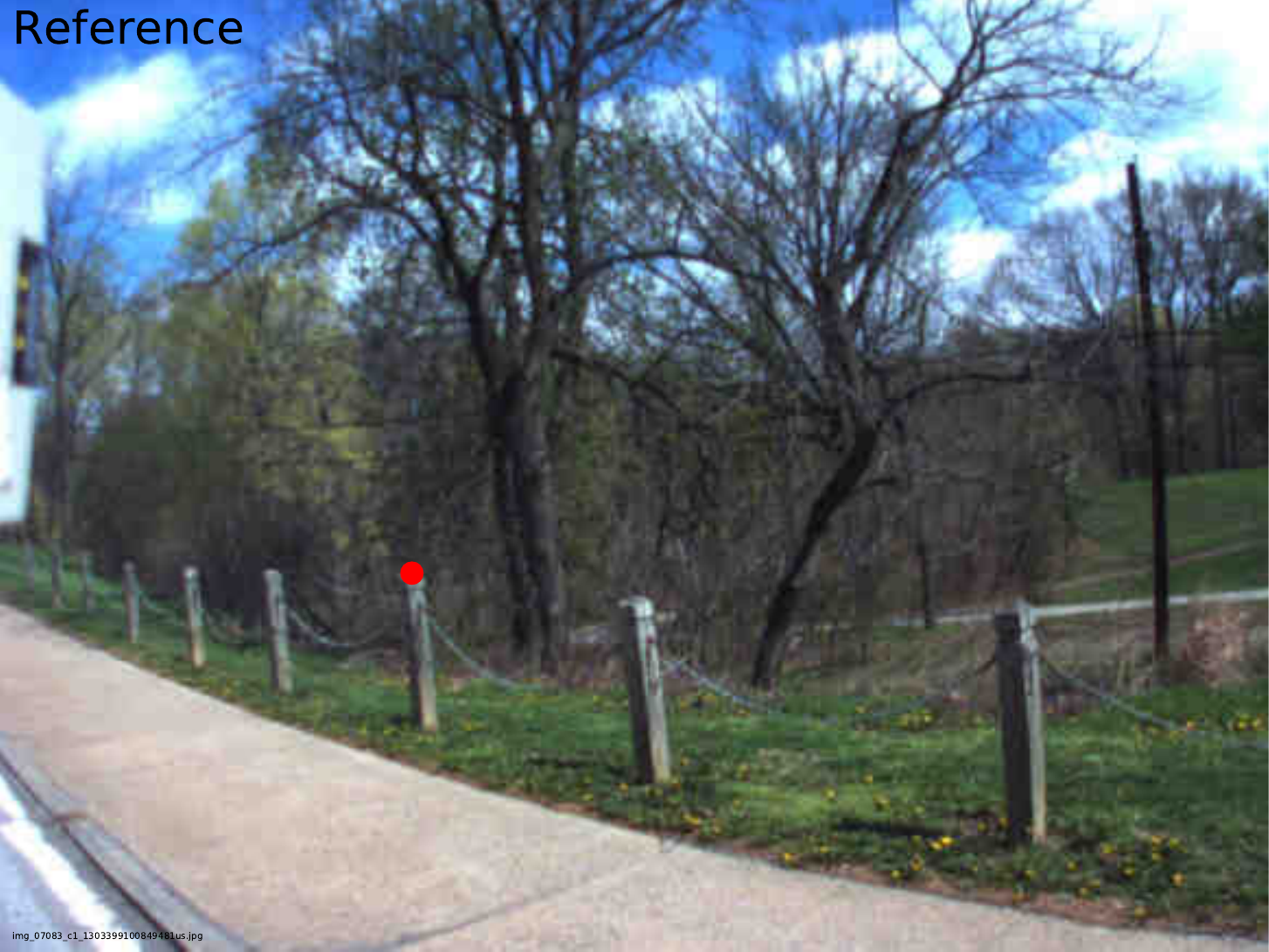}
    \end{minipage}%
    \begin{minipage}{\iwidth\linewidth}
        \includegraphics[width=\linewidth]{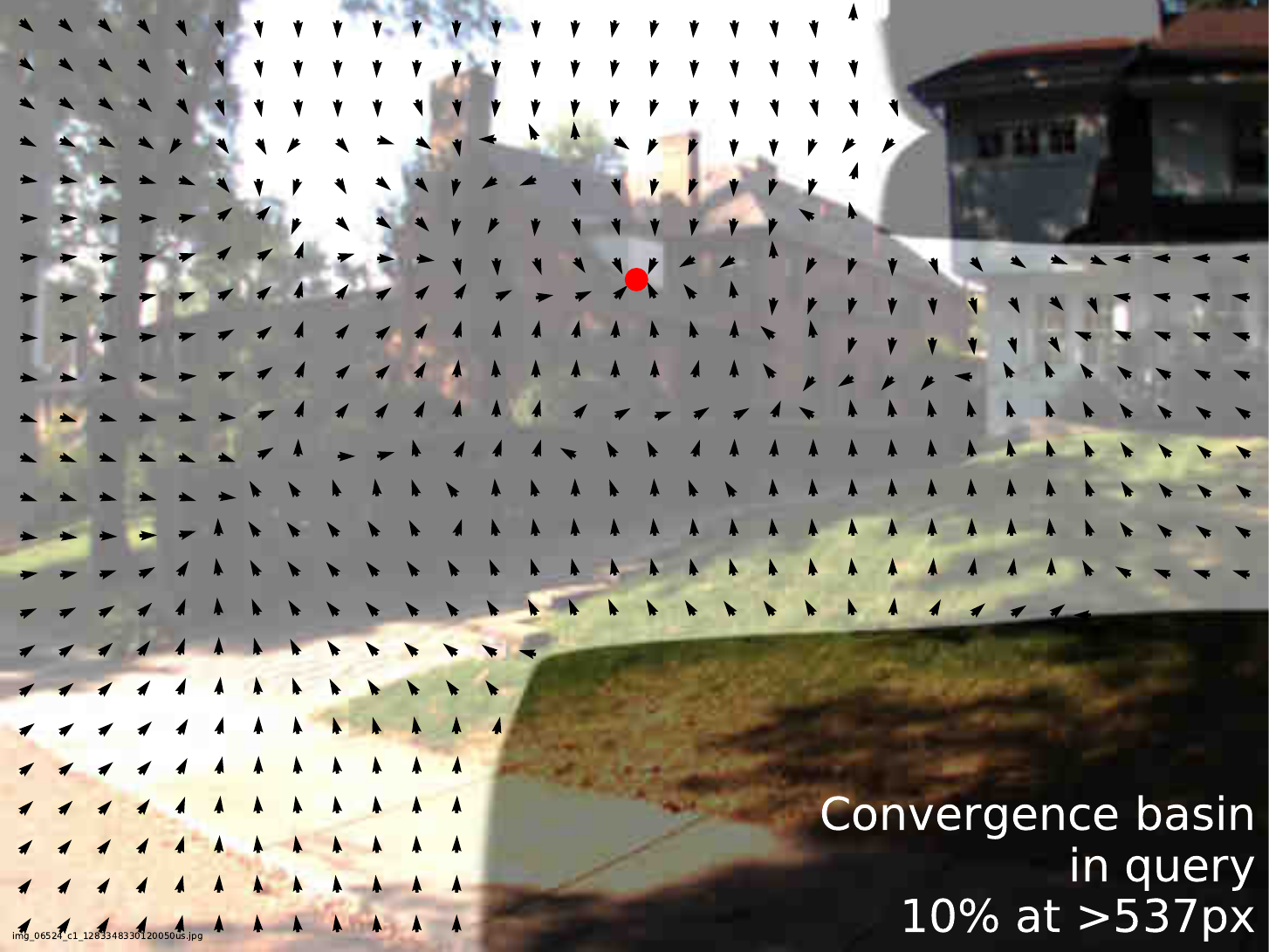}
        
        \vspace{2mm}
        \includegraphics[width=\linewidth]{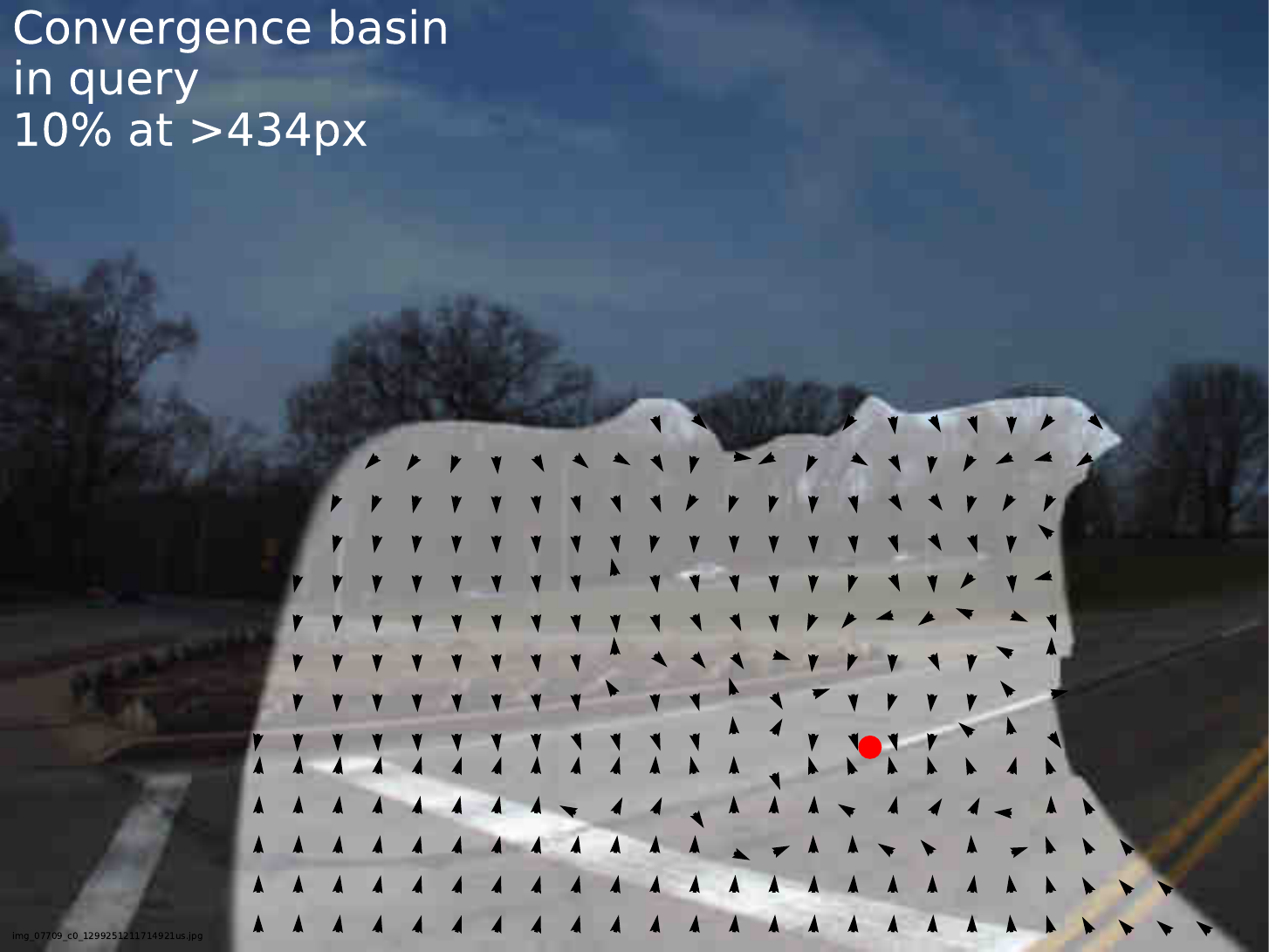}
        \vspace{2mm}
        \includegraphics[width=\linewidth]{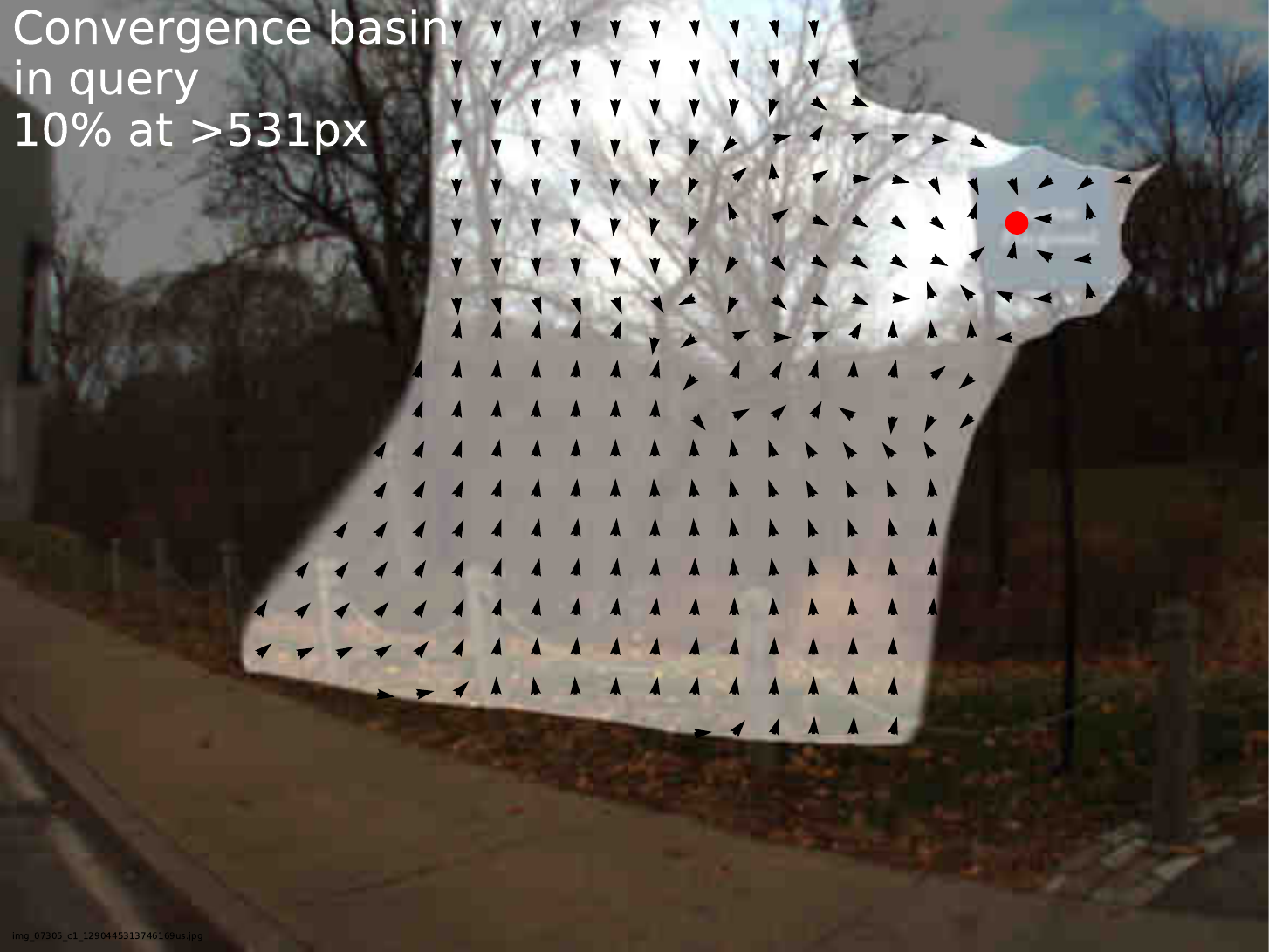}
        \vspace{2mm}
        \includegraphics[width=\linewidth]{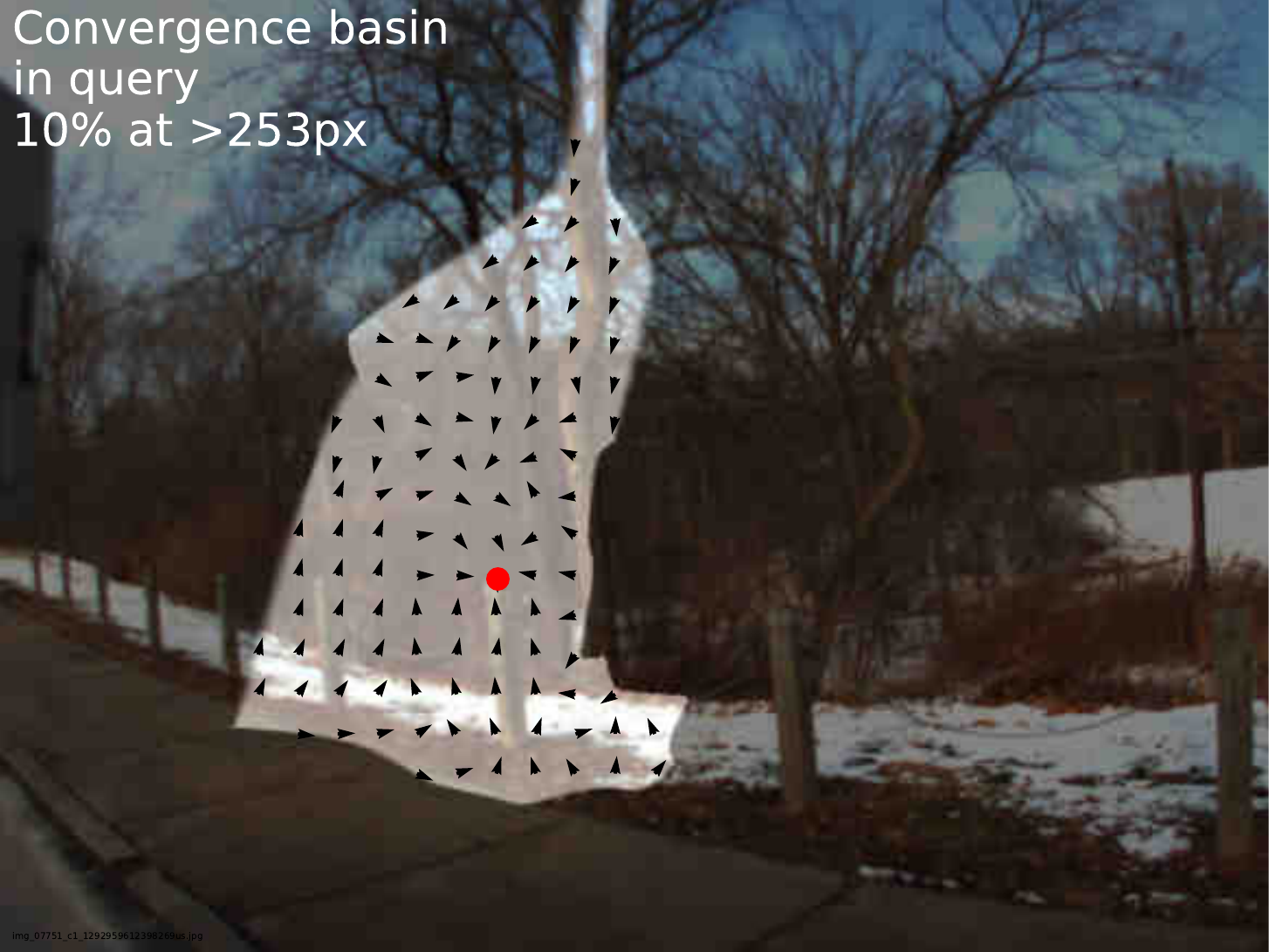}
    \end{minipage}
    \vspace{1mm}
    \caption{\textbf{Convergence basin.}
    We show the convergence basins of individual selected points given cross-season query and reference images from the CMU dataset.
    The last row shows smaller basins due to repeated patterns like poles or tree silhouettes.
    }
    \label{fig:convergence_supp}%
\end{figure*}
    \fi
\else
    
\fi

\clearpage
{\small
\bibliographystyle{ieee_fullname}
\bibliography{mybib}
}

\end{document}